\documentclass[10pt,a4paper,twoside,openright,onecolumn]{report}
\pdfoutput=1
\usepackage[left=3.75cm,right=2.5cm,top=2.5cm,bottom=2.5cm]{geometry}
\usepackage[misc]{cmpcover}
\usepackage{pdfpages} %
\usepackage[toc,page]{appendix} %

\usepackage[caption=false,font=normalsize,labelfont=sf,textfont=sf]{subfig} %
\usepackage[dvipsnames,table]{xcolor}

\usepackage{algpseudocode}
\usepackage{algorithm}
\usepackage{array}
\usepackage{amssymb}
\usepackage{graphicx}
\usepackage{array}
\usepackage{eqparbox}
\usepackage{url}
\usepackage{gensymb}
\usepackage{rotating}
\usepackage{dcolumn}
\usepackage{xspace}
\usepackage{algpseudocode}

\usepackage{tabu}
\usepackage{setspace}
\usepackage{appendix}
\usepackage{changepage}
\usepackage[numbers]{natbib}
\usepackage{afterpage}
\usepackage{mathtools}
\usepackage{booktabs}    %
\usepackage{amsfonts}    %
\usepackage{nicefrac}    %
\usepackage{microtype}   %
\usepackage{amsmath}
\usepackage{hyphenat}

\usepackage{tabularx}
\usepackage{bbm}
\usepackage{multirow}
\usepackage[pagebackref=true,breaklinks=true,colorlinks,bookmarks=false]{hyperref}
\usepackage{bbding}
\usepackage{pifont}
\usepackage{wasysym}
\usepackage[inline]{enumitem}
\usepackage{adjustbox}
\usepackage{stmaryrd}
\usepackage{trimclip}

\usepackage{titlesec}
\usepackage{lipsum}
\usepackage[boldmath]{numprint}

\usepackage{setspace}
\usepackage{eqnarray}
\usepackage{breqn}
\usepackage{capt-of}

\usepackage{wrapfig}

\usepackage[T1,T2A]{fontenc}
\usepackage[utf8]{inputenc} %
\usepackage[ukrainian,english]{babel}

\titleformat{\chapter}[display]
{\normalfont\Large\filcenter\sffamily}
{%
 \titlerule[1pt]%
 \vspace{1pt}%
 \titlerule
 \vspace{1pc}%
 \LARGE\MakeUppercase{\chaptertitlename}~\thechapter}
{1pc}
{\titlerule\Huge}
\usepackage{fancyhdr}
\DeclareMathOperator*{\argmin}{arg\,min}

\makeatletter
\DeclareRobustCommand\onedot{\futurelet\@let@token\@onedot}
\def\@onedot{\ifx\@let@token.\else.\null\fi\xspace}
\newcommand{\ra}[1]{\renewcommand{\arraystretch}{#1}}
\def\eg{\emph{e.g}\onedot} 
\def\ie{\emph{i.e}\onedot}

\def\etal{\emph{et al}\onedot}

\def\jm{Ji\v{r}\'{i} Matas}
\def\fr{Filip Radenovi\'{c}}

\date{\today}
\newcommand{\algrule}[1][.2pt]{\par\vskip.5\baselineskip\hrule height #1\par\vskip.5\baselineskip}

\newcommand{\hiddensubsection}[1]{
    \stepcounter{subsection}
    \subsection*{\arabic{chapter}.\arabic{section}.\arabic{subsection}\hspace{1em}{#1}}
}

\makeatother

\newcommand{\comment}[1]{}

\newcommand{\bI}{\mathbf{I}}

\newcommand{\bF}{\mathbf{F}}

\newcommand{\bK}{\mathbf{K}}

\newcommand{\fig}[1]{Fig.~\ref{fig:#1}}

\newcommand{\tbl}[1]{Table~\ref{tbl:#1}}
\newcommand{\secref}[1]{Section~\ref{imc:#1}}

\definecolor{color1}{RGB}{0,199,1}
\definecolor{color2}{RGB}{224,43,28}

\newcommand{\ky}[1]{{\color{orange}{#1}}}
\newcommand{\KY}[1]{{\color{orange}{\bf [Kwang: #1]}}}
\newcommand{\et}[1]{{\color{magenta}{#1}}}
\newcommand{\ET}[1]{{\color{magenta}{\bf [Eduard: #1]}}}

\newcommand{\PF}[1]{{\color{blue}{\bf [Pascal: #1]}}}

\newcommand{\AM}[1]{{\color{OliveGreen}{\bf [Anastasiia: #1]}}}

\newcommand{\DM}[1]{{\color{Purple}{\bf [Dmytro: #1]}}}

\newcommand{\YJ}[1]{{\color{Bittersweet}{\bf [Yuhe: #1]}}}
\newcommand{\edit}[1]{{\color{Cyan}{#1}}}
\newcommand{\revision}[1]{{\color{Cyan}{#1}}}

\renewcommand{\ky}[1]{#1}
\renewcommand{\KY}[1]{}
\renewcommand{\et}[1]{#1}
\renewcommand{\ET}[1]{}

\renewcommand{\PF}[1]{}

\renewcommand{\AM}[1]{}

\renewcommand{\DM}[1]{}

\renewcommand{\YJ}[1]{}
\renewcommand{\edit}[1]{#1}
\renewcommand{\revision}[1]{#1}

\newcommand{\bE}{\mathbf{E}}

\newcommand{\imsize}{1cm}
\newcommand{\hspacer}{0.3mm}
\newcommand{\vspacer}{0.1mm}

\newcommand{\numim}{N}
\newcommand{\numfeat}{K}

\newcommand{\covisibility}{v}

\newcommand{\imscl}{0.01}

\newcommand{\ransacConf}{\tau}

\newcommand{\ransacReprojTh}{\eta}
\newcommand{\ransacMaxIters}{\Gamma}

\newcommand{\ratiotest}{r}

\makeatletter
\DeclareRobustCommand{\shortto}{%
  \mathrel{\mathpalette\short@to\relax}%
}

\newcommand{\short@to}[2]{%
  \mkern2mu
  \clipbox{{.5\width} 0 0 0}{$\m@th#1\vphantom{+}{\shortrightarrow}$}%
  }
\makeatother

\renewcommand{\paragraph}[1]{\vspace{0.5em}\noindent{\bf#1}}

\makeatletter
\newcommand\footnoteref[1]{\protected@xdef\@thefnmark{\ref{#1}}\@footnotemark}
\makeatother

\makeatletter
\renewcommand{\paragraph}{%
  \@startsection{paragraph}{4}%
  {\z@}{0.5ex \@plus 1ex \@minus .2ex}{-1em}%
  {\normalfont\normalsize\bfseries}%
}

\def\ccaEF#1{\cellcolor{black!#1!black!#1}\ifnum #1>50\color{white}\fi{#1}}
\def\ccaEVD#1{\cellcolor{black!#10}\ifnum #1>7\color{white}\fi{#1}}
\def\ccaMMS#1{\cellcolor{black!#1}\ifnum #1>50\color{white}\fi{#1}}
\def\ccaWGABS#1{\cellcolor{black!#10}\ifnum #1>3\color{white}\fi{#1}}
\def\ccaWGALBS#1{\cellcolor{black!#10}\ifnum #1>5\color{white}\fi{#1}}
\def\ccaWGLBS#1{\cellcolor{black!#10}\ifnum #1>5\color{white}\fi{#1}}
\def\ccaWGSBS#1{\cellcolor{black!#10}\ifnum #1>3\color{white}\fi{#1}}
\def\ccaWLABS#1{\cellcolor{black!#10}\ifnum #1>3\color{white}\fi{#1}}
\def\ccalostinpast#1{\cellcolor{black!#1}\ifnum #1>80\color{white}\fi{#1}}
\def\ccaoxford#1{\cellcolor{black!#1!black!#1}\ifnum #1>50\color{white}\fi{#1}}
\def\ccasnavely#1{\cellcolor{black!#1!black!#1}\ifnum #1>50\color{white}\fi{#1}}
\def\ccastewart#1{\cellcolor{black!#1}\ifnum #1>55\color{white}\fi{#1}}

\usepackage{scrextend}
\title{Learning and Crafting \\for the\\ Wide Multiple Baseline Stereo}
\CMPAdvisor{prof. \jm}
\author{Dmytro Mishkin}

\CMPSideBarText{PhD Thesis} %
\CMPSubtitle{A Doctoral Thesis presented to the Faculty of the Electrical Engineering
of the Czech Technical University in Prague in fulfillment of the
requirements for the Ph.D. Degree in Study Programme No. P2612 -
Electrical Engineering and Information Technology, Branch No.
3902V035 - Articial Intelligence and Biocybernetics, by}

\CMPEmail{mishkdmy@cmp.felk.cvut.cz} 

\CMPDocumentURL{http://cmp.felk.cvut.cz/~mishkdmy/dissertation_DM.pdf}

\CMPAcknowledgement{Supported by Czech Science Foundation Project GACR P103/12/G084, Austrian Ministry for Transport, Innovation and Technology, the Federal Ministry of Science, Research and Economy, and the Province of Upper Austria in the frame of the COMET center SCCH, the CTU student grant SGS17/185/OHK3/3T/13, and OP VVV funded project CZ.02.1.01/0.0/0.0/16 019/0000765 “Research Center for Informatics”}

\begin{document}

\chapter*{\centering Abstract}

This thesis introduces the wide multiple baseline stereo (WxBS) problem.
WxBS, a generalization of the standard wide baseline stereo problem,
considers the matching of images that simultaneously differ in more than one image acquisition factor such as viewpoint, illumination, sensor type, or where object appearance changes significantly, e.g., over time. A new dataset with the ground truth, evaluation metric and baselines has been introduced. 

The thesis presents the following improvements of the WxBS pipeline. (i) A loss function, called HardNeg, for learning a local image descriptor that relies on hard negative mining within a mini-batch and on the maximization of the distance between the closest positive and the closest negative patches. (ii) The descriptor trained with the HardNeg loss, called HardNet, is compact and shows state-of-the-art performance in standard matching, patch verification and retrieval benchmarks. (iii) A method for learning the affine shape, orientation, and potentially other parameters related to geometric and appearance properties of local features. (iv) A tentative correspondences generation strategy which generalizes the standard first to second closest distance ratio is presented. The selection strategy, which shows performance superior to the standard method, is applicable to either hard-engineered descriptors like SIFT, LIOP, and MROGH or deeply learned like HardNet. (v) A feedback loop is introduced for the two-view matching problem, resulting in MODS -- matching with on-demand view synthesis -- algorithm. MODS is an algorithm that handles a viewing angle difference even larger than the previous state-of-the-art ASIFT algorithm, without a significant increase of computational cost over ``standard'' wide and narrow baseline approaches. 

Last, but not least, a comprehensive benchmark for local features and robust estimation algorithms is introduced. The modular structure of its pipeline allows easy integration, configuration, and combination of methods and heuristics.

\thispagestyle{empty} 

\chapter*{\centering Abstrakt}

Tato práce představuje problém wide multiple baseline stereo (WxBS), který je zobecněním standardního problému wide baseline stereo. WxBS se zabývá párováním obrázků, které se současně liší více než v jednom faktoru získání obrázku, jako je stanoviště a zorný úhel, osvětlení, typ senzoru, nebo kde se vzhled objektu zásadně mění, např. v průběhu času. Byl představen nový dataset s anotacemi, evaluační metrikou a výchozími metodami.

Disertační práce představuje následující vylepšení algoritmu WxBS: Ztrátovou funkci (i) pro učení lokálního deskriptoru obrázku založeného na výběru obtížných negativních příkladů v mini-batchi a na maximalizaci rozdílu vzdálenosti od nejbližšího pozitivního a nejbližšího negativního příkladu. Výsledný deskriptor (ii), nazvaný HardNet, který je kompaktní a dosahuje vynikajících výsledků v standardním párování, ověřování malých částí obrázků a vyhledávání obrázků. Metodu pro učení afinního tvaru, orientace a potenciálně dalších parametrů souvisejících s geometrií a vzhledem lokálních příznaků (iii). Strategii pro generování tentativních korespondencí (iv), která zobecňuje standardní poměr prvních dvou nejbližších vzdáleností. Strategie výběru, která dosahuje lepších výsledků než standardní metoda, je aplikovatelná jak pro manuálně navržené deskriptory jako SIFT, LIOP či MROGH, tak pro hluboce učené deskriptory jako HardNet. Zavedením zpětnovazební smyčky pro problém párování ze dvou pohledů vznikl algoritmus MODS -- matching with on-demand view synthesis (v). Algoritmus MODS zvládá i větší rozdíly v úhlu pohledu než ASIFT algoritmus, a to bez výrazného navýšení výpočetní ceny proti standardním wide-baseline a narrow-baseline přístupům.

V neposlední řadě je představen komplexní benchmark pro porovnávání lokálních příznaků a algoritmů pro robustní odhady (vi). Jeho modulární struktura umožňuje jednoduchou integraci, konfiguraci a kombinaci metod a heuristik.
\thispagestyle{empty} 
\newpage

\begin{otherlanguage*}{ukrainian}
\chapter*{\centering Анотація}
У даній дисертацiї досліджується задача стереозору з широкою мультибазою (англ. Wide Multiple Baseline Stereo, WxBS), яка є узагальненням стандартної задачi стереозору iз широкою базою (англ. Wide Baseline Stereo, WBS). Задача WxBS розглядає спiвставлення зображень однiєї сцени, якi одночасно вiдрiзняються бiльш нiж в одному аспектi. Такими аспектами можуть бути позицiя камери, освiтленiсть, тип датчика або iстотна змiна зовнiшнього вигляду об’єкта, наприклад, з часом. Для тестування алгоритмiв розв'язування задачі WxBS представлено новий датасет iз еталонними даними та метриками.

У дисертацiї запропоновано такі вдосконалення алгоритму для розв’язання задачi стереозору з широкою мультибазою. Розроблено функцiю втрат (i) для тренування локального дескриптора, яка покладається на пошук найскладнiшого негативного прикладу у мiнi-батчi та максимiзацiї вiдстанi мiж найближчим позитивним та найближчим негативним фрагментами зображення. Отриманий дескриптор (ii), який одержав назву HardNet, є компактним та демонструє найкращi результати у тестах на порiвняння локальних дескрипторiв. Запропоновано метод навчання детекторів елiптичної форми, орiєнтацiї та потенцiйно iнших параметрiв локальних ознак зображення (iii). Удосконалено стратегiю генерацiї початкових вiдповiдностей (iv) шляхом узагальнення стандартного спiввiдношення першої та другої найближчих вiдстаней дескрипторiв. Нова стратегiя генерацiї покращує якiсть отриманих вiдповiдностей порiвняно зi стандартним методом та може застосовуватися для будь-якого типу дескрипторів, вiд спроєктованих вручну, таких як SIFT, LIOP i MROGH, до навчених за допомогою глибокого навчання, таких як HardNet. Запропоновано включення зворотного зв’язку до алгоритму спiвставлення двох зображень однiєї сцени, що дозволило побудувати алгоритм спiвставлення шляхом генерування синтетичних зображень на вимогу (v), який одержав назву MODS (від англ. Matching with On-Demand View Synthesis). MODS дозволяє знаходити вiдповiдностi на зображеннях, одержаних у ширшому діапазоні рiзниць в позицiях камер, ніж попередній найкращий алгоритм розв'язання даної задачі, ASIFT. Окрім того, на вiдмiну вiд ASIFT, MODS спроможний швидко опрацьовувати пари зображень невисокої складностi, наприклад, отриманих з камер, які розташовані близько одна до одної.

Також у дисертації представлено бенчмарк для детекторiв та дескрипторiв локальних ознак, алгоритмiв стiйкої оцiнки геометрiї сцени. Його модульна структура дозволяє легко додавати, налаштовувати, поєднувати та порiвнювати різноманітні методи та еврістики.
\end{otherlanguage*}

\thispagestyle{empty} 
\newpage

\chapter*{\centering Acknowledgements}

Pursuing a PhD has been the longest and hardest endeavour in my life so far. It started with an internship at Center for Machine Perception in 2012 -- 2013. That internship literally changed my life and not least because of the wonderful people I met and admired since then. Jan, Michal, Kostya, Sasha, Nataliya, Hongping, Matej, Jana, Javier, Jimmy, Michal, Milan, Filip -- our lunches together will always be in my heart. My first close supervisor --  Michal Perdoch spent a lot of time answering my stupid questions about callbacks, coding, and geometry. 

Professor \jm{} -- was and is much more than just a supervisor to me. He always has time for help and advice -- be it the deadline paper writing session at midnight, fight for the open-science on social media, discussion about scientific reforms, or not kicking me out for leaving the note "I am going for Euromaidan tomorrow" on his table and disappearing to Kyiv. He -- together with all other people of the CMP -- have shown me a bright side of academia -- free, supportive, and inspiring. It is a blessing to work with you.

Eva Matyskova -- thank you for making Prague feel like a home and treating us --  PhD students -- as your kids in the best sense of this word. My life in Czech Republic would be much more difficult without your help. Sergey Shelpuk -- your talk in Odesa, 2014 about deep learning was pivotal for me. Thank you.

My co-authors -- Michal, Karel, Anastasiia, Filip, Edgar, Eduard, Milan, Milan, Kwang, Yuhe, Javier, Alexey, Nikolaj, Vladlen, Orest, Mykola, Volodymyr, Gary, Daniel, Daniel, Ivan, Ethan, Ilia, Amy -- it was a pleasure to work together and I hope to do it once again. 

Amy Tabb, James Pritts, and Karel Lenc -- thank you for reading the early drafts of this thesis and helping to make it better. 

Finally, nothing above would happen without the endless support of my family -- my parents Sergiy and Tetyana, my wife Olya, and my daughter Sofia. Whatever I do, you believe in me. Hugs :)

\thispagestyle{empty}

{
  \hypersetup{linkcolor=black}
  \tableofcontents
}
\newpage

\chapter[Introduction]{Introduction  \\
\includegraphics[width=0.95\linewidth]{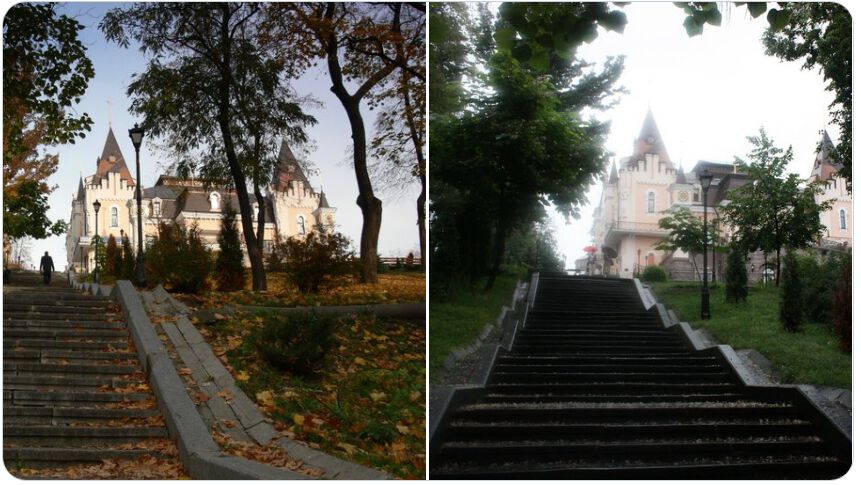}}
\label{ch:intro}
\section{What is wide multiple baseline stereo?}
\label{intro:first}

Imagine you have a nice photo you took in autumn and would like to take one in summer, from the same spot. How would you achieve that? You go to the place and you start to compare what you see on the camera screen and on the printed photo. Specifically, you would probably try to locate the same objects, e.g., ``that high lamppost'' or ``this wall clock''. Then one would estimate how differently they are arranged on the old photo and camera screen. For example, by checking whether the lamppost is occluding the clock on the tower or not. That would give an idea of how you should move your camera.%

Now, what if you are not allowed to take that photo with you, because it is a museum photo and taking pictures is prohibited there. Instead you can create a description of it. In that case, it is likely that you would try to make a list of features and objects in the photo together with the descriptions, which are sufficient to distinguish the objects. For example, ``a long staircase on the left side'', ``The nice building with a dark roof and two towers'' or ``the top of the lampost''. It would be useful to also describe where these objects and features are pictured in the photo, ``The lamp posts are on the left, the closest to the viewer is in front of the left tower with a clock. The clock tower is not the part of the building and stands on its own''. Then when arriving, you would try to find those objects, match them to the description you have, and try to estimate where you should go. You repeat the procedure until the camera screen shows a picture which is fitting the description you have and the image you have in your memory.%

Congratulations! You just have successfully registered the two images, which have a significant difference in viewpoint, appearance, and illumination. In the process of doing so, you were solving multiple times the wide multiple-baseline stereo problem (WxBS) -- estimating the relative camera pose from a pair of images, different in many aspects, yet depicting the same scene.

A bit more formally\footnote{The formal definition of the WxBS will be given in Section~\ref{wxbs:definition}.}, the \emph{wide multiple baseline stereo (WxBS)} is a process of establishing a sufficient number of pixel or region correspondences from two or more images depicting the same scene to estimate the geometric relationship between cameras, which produced these images. Typically, WxBS relies on the scene rigidity -- the assumption that there is no motion in the scene except the motion of the camera itself. The stereo problem is called wide multiple baseline if the images are significantly different in more than one aspect: viewpoint, illumination, time of acquisition, and so on. Historically, people were focused on the simpler problem with a single baseline, which was geometrical, i.e., viewpoint difference between cameras, and the area was known as wide baseline stereo. Nowadays, the field is mature and research is focused on solving more challenging multi-baseline problems. 
WxBS is a building block of many popular computer vision applications, where spatial localization or 3D world understanding is required -- panorama stitching, 3D reconstruction, image retrieval, SLAM, etc. 

\section{Areas related to the wide multiple baseline stereo}
\label{intro:not-wbs}
Let us position the WxBS problem in computer vision research by showing its similarities and differences to related areas and concepts. 

\emph{Image registration} is a general term for the task of establishing dense correspondences between images of the same or related scenes, surfaces of volumes, often in 2D. Unlike WxBS, image registration typically assumes a small geometric baseline and no occlusions, but does not assume the rigidity of the scene. Image registration often assumes some underlying model, e.g., elastic deformation, whose parameters need to be estimated.  Image registration also allows images of a different modality, e.g., registering the X-ray image to the ultrasound image, or aligning different but consecutive cuts of some tissue, each of those is coloured with a different chemical pigment. 

A related concept --  \emph{optical flow}  -- is a process of establishing dense correspondences between images, which are taken in consecutive moments of time. Optical flow does not have the rigidity assumption unlike the WxBS but instead relies on the narrow temporal baseline and often on the (soft) brightness constancy assumption. Unlike image registration, optical flow typically deals with occlusions, but does not estimate camera movement or scene geometry.

\emph{Narrow baseline stereo} or -- simply -- \emph{stereo} estimates the disparity between the left and right frames. The disparity is a difference in horizontal coordinates between the corresponding points in the left and right images. The disparity definition suggests that the camera setup is a calibrated camera pair, or, at least, that the geometric relationship between cameras is known beforehand. Moreover, as the name suggests, the stereo algorithm assumes a small geometrical baseline. Gradient-based methods often can be successfully applied to solve the stereo, image registration and optical flow problems, unlike the wide baseline stereo. 

The \emph{3D reconstruction} or \emph{Structure-from-Motion (SfM)} aims to reconstruct a 3D scene structure, e.g., in the form of a point cloud, from a set of 2D images. WxBS, unlike SfM, does not estimate the underlying point cloud. Moreover, the SfM, be it sparse or dense, benefits from having as many correspondences as possible to obtain a good 3D model. WxBS, in its turn, tends to have just enough correspondences needed for estimating the geometry model. 

\emph{Visual localization} is a task of estimating the camera pose from which a query image was acquired, given the database of training images with a known pose. WxBS can be seen as one component of the general system, which is used for solving the visual localization. For example, one can first perform an image search to produce the list of candidate images most similar to the query image. Then the WxBS is applied to estimate the query image camera pose. However, that is only one of the possible ways of solving the visual localization problem. For example, one could train a model, which directly estimates the camera pose given the image. 

\section{How to solve the wide multiple baseline stereo problem?}
\label{intro:pipeline}

\begin{figure}[tb]
\centering
\includegraphics[width=1\linewidth]{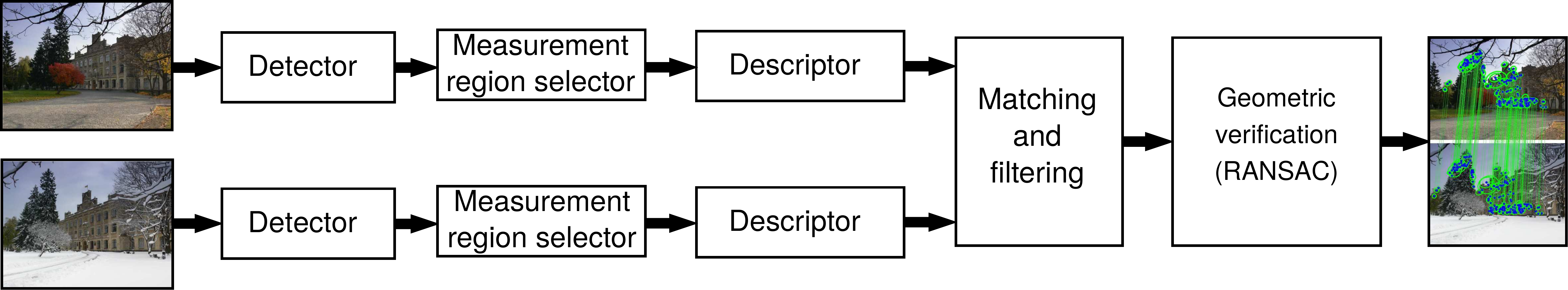}
\caption{WxBS: Wide multiple baseline stereo pipeline.}
\label{fig:wbs-pipeline}
\end{figure}

The wide baseline stereo problem is commonly addressed by a family of algorithms, the general structure of which is shown in Figure~\ref{fig:wbs-pipeline}. We will be referring to it as the WxBS pipeline or the WxBS algorithm. Let us describe it in more detail and discuss the reasoning behind each block. 

\begin{enumerate}

\item A set of the local features\footnote{Also known as keypoints, local regions, distinguished regions, salient regions, salient points, etc.} is detected in each image independently. In automated systems the local features are usually low level structures like corners, blobs and so on. However, they can also be more high level semantic structures, as we used in the example in a Section~\ref{intro:first}: ``a long staircase on the left side'', ``the top of the lampost'' and so on. An important detail is that detection is typically done in each image separately. Why is it the case? If the task is to match only a single image pair, that would be an unnecessary restriction. It is even benefitial to process the images jointly, as a human would do, by placing both images side-by-side and looking at them back and forth. However, the wide baseline stereo task rarely arises by itself, more often it is only a part of a bigger system, e.g. visual localization or 3D recontruction from the collection of images. Therefore, one needs to match an image to not the one, but multiple other images. That is why it is benefitial to perform feature extraction only once per image and then load the stored results. Moreover, independent feature extraction is a task which is easy to parallelize and that is typically done in most of libraries and frameworks for the WxBS. One could be wondering if the local feature detection process is necessary at all? Indeed, it is possible to avoid feature detection and consider all the pixels as ``detections''. The problem with such approach is the high computational and memory complexity -- even a small $800 \times 600$ image contains half a million pixels, which need to be matched to half a million pixels in another image. 

\item A region around the local feature to be described is selected. If one considers a keypoint to be literally a point, it is impractical to distinguish between them based only on coordinates and, maybe, the single RGB value of the pixel. On the other extreme, part of the image, which really far from the current keypoint helps little to nothing in terms of finding a correspondence. Thus, a reasonable trade-off needs to be made. Keypoint therefore can be think of as the ``reference point of the distinguished region'', e.g. a center of the blob. It worth mention that some detectors return a region by default, so this step is omitted, or, to be precise, included into step 1 ``local features detection''. However, it is useful to have it discussed separately .

\item A patch around each local feature is described with a local feature descriptor, i.e. converted to a compact format. Such procedure also should be robust to changes in acquisition conditions so that descriptors related to the same 3D points are similar and dissimilar otherwise. The local feature descriptors are then used for the efficient generation of tentative correspondences. Could one skip this stage? Yes, but as with the local feature detection, the skipping is not desirable from a computational point of view -- the benefits are discussed in the next stage -- matching. 
Local feature detection, measurement region selection and description together convert an image into a sparse representation, which is suitable for the correspondence search. Such representation is more robust to the acquisition conditions and can be further indexed if used for image retrieval.

\item Tentative correspondences between detected features are established and then filtered. The simplest and common way to generate tentative correspondences is to perform a nearest neighbor search in the descriptor space. The commonly used descriptors are the binary or float point vectors, which allows to employ various algorithms for approximate nearest neighbor search and trade a small amount of accuracy for orders of magnitude speed-up. 
Such correspondences need to be filtered, that is why the are called ``tentative'' or ``putative'' -- a significant percantage of them is incorrect. There are many reasons for that -- imprerfection of the local feature detector, descriptor, matching process and simply the fact that some parts of the scene are visible only on one image, but not another. 

\item The geometric relationship between the two images is recovered, which is the final goal of the whole process. In addition, the tentative correspondences, which are not consistent with the found geometry, are called outliers and are discarded. The most common way of robust model estimation in the presense of outliers is called RANSAC -- random sample consensus. There are other methods as well, e.g. re-weighted least squares, but RANSAC predominately used in practice. 

\end{enumerate}

The class of algorithms that we have just described significantly changed the computer vision landscape after its introduction in middle 1990-s\footnote{The brief history of the wide (single) baseline stereo will be given in Section~\ref{intro:history}} and is still at the core of many computer vision tasks in 2021, such as 3D reconstruction~\cite{PhotoTourism2006, RomeInDay2009, COLMAP2016}, SLAM(Simultaneous localization and mapping)~\cite{Se02, PTAM2007, Mur15}, panorama stiching~\cite{Brown07}, feature-based image registration~\cite{DualBootstrap2003}, visual localization~\cite{pion2020benchmarking} and image retrieval~\cite{DELF2017}.

This thesis is devoted to the wide multiple baseline stereo problem and to improving the WxBS algorithm  in the form shown in Figure~\ref{fig:wbs-pipeline} in particular. We present ways of improving individual components of the WxBS algorithm, as well as adding new parts to it\footnote{See Section~\ref{intro:contributions} for the detailed list of contributions}.

\section{History of the wide multiple baseline stereo and its role in the deep learning world}
\label{intro:history} 

As often happens, a new problem arises from the old -- narrow or short baseline stereo. In the narrow baseline stereo, images are taken from nearby positions, often exactly at the same time. %
One could find correspondence for the point $(x,y)$ from the image $I_1$ in the image $I_2$ by simply searching in some small window around $(x,y)$~\cite{Hannah1974ComputerMO, Moravec1980} or, assuming that a camera pair is calibrated and the images are rectified -- by searching along the known epipolar line~\cite{Hartley2004}.

One of the first, if not the first, approaches to the wide baseline stereo problem was proposed by Schmid and Mohr~\cite{Schmid1995} in 1995. Given the difficulty of the wide multiple baseline stereo task at the moment, only a single --- geometrical -- baseline was considered, thus the name -- wide baseline stereo. The idea of Schmid and Mohr was to equip each keypoint with an invariant descriptor. This allowed establishing tentative correspondences between keypoints under viewpoint and illumination changes, as well as occlusions. One of the stepping stones was the corner detector by Harris and Stevens~\cite{Harris88}, initially used for the application of tracking. It is worth a mention, that there were other good choices for the local feature detector at the time, starting with the Forstner~\cite{forstner1987fast}, Moravec~\cite{Moravec1980} and Beaudet feature detectors~\cite{Hessian78}.

The Schmid and Mohr approach was later extended by Beardsley, Torr and Zisserman~\cite{Beardsley96} by adding RANSAC~\cite{RANSAC1981} robust geometry estimation and later refined by Pritchett and Zisserman~\cite{Pritchett1998, Pritchett1998b} in 1998. The general pipeline remains mostly the same until now~\cite{WBSTorr99, CsurkaReview2018, IMW2020}, which is shown in Figure~\ref{fig:wbs-pipeline}. %

Soon after the introduction of the WBS algorithm, it became clear that its quality significantly depends on the quality of each component, i.e., local feature detector, descriptor, and geometry estimation. Local feature detectors were designed to be as invariant as possible, backed up by the scale-space theory, most notable developed by Lindenberg~\cite{Lindeberg1993, Lindeberg1998, lindeberg2013scale}. A plethora of new detectors and descriptors were proposed in that time. We refer the interested reader to these two surveys: by Tuytelaars and Mikolajczyk~\cite{Tuytelaars2008} (2008) and by Csurka~\etal~\cite{CsurkaReview2018} (2018). Among the proposed local features is one of the most cited computer vision papers ever -- SIFT local feature~\cite{Lowe99, SIFT2004}. Besides the SIFT descriptor itself, Lowe`s paper incorporated several important steps, proposed earlier with his co-authors, to the matching pipeline. Specifically, they are quadratic fitting of the feature responses for precise keypoint localization~\cite{QuadInterp2002}, using the Best-Bin-First kd-tree~\cite{aknn1997} as an approximate nearest neightbor search engine to speed-up the tentative correspondences generation, and using second-nearest neighbor (SNN) ratio to filter the tentative matches. 
It is worth noting that SIFT feature became popular only after Mikolajczyk benchmark paper~\cite{MikoDescEval2003, Mikolajczyk05} that showed its superiority to the rest of alternatives. 
 
Robust geometry estimation was also a hot topic: a lot of improvements over vanilla RANSAC were proposed. For example, LO-RANSAC~\cite{LOransac2003} proposed an additional local optimization step into RANSAC to significantly decrease the number of required steps. PROSAC~\cite{PROSAC2005} takes into account the tentative correspondences matching score during sampling to speed up the procedure.  DEGENSAC~\cite{Degensac2005} improved the quality of the geometry estimation in the presence of a dominant plane in the images, which is the typical case for urban images. We refer the interested reader to the survey by Choi~\etal~\cite{RANSACSurvey2009}. 

Wide baseline stereo framework became the main part of such applications as 3D reconstruction, SLAM, and image localization. The success of the wide baseline stereo pipeline with SIFT features led to aplication of its components to other computer vision tasks, which were reformulated through the wide baseline stereo lens. We took the illustration figures from the iconic papers of that time in Figure~\ref{fig:cv-as-wbs-problems}. To name a few:

\begin{figure}
\centering
\begin{subfloat}[]{
\includegraphics[width=0.31\linewidth]{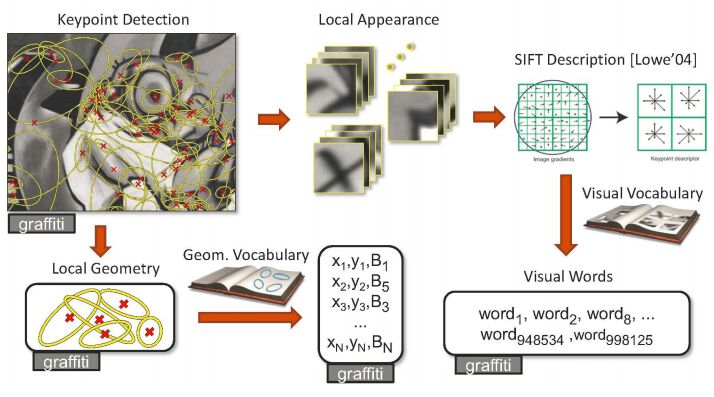}}
\end{subfloat}%
\begin{subfloat}[]{
\includegraphics[width=0.31\linewidth]{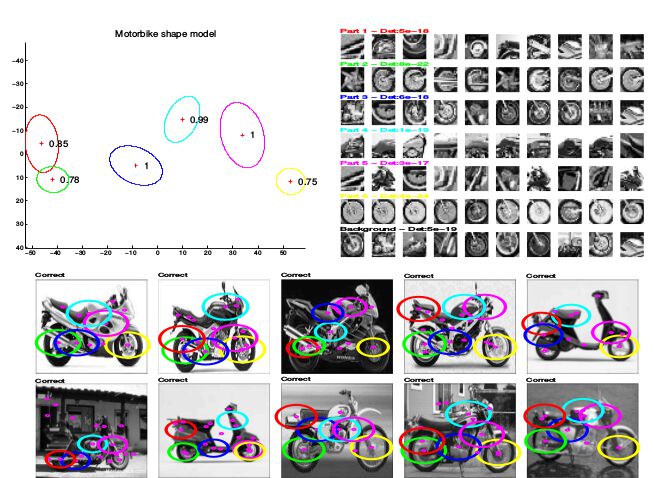}}
\end{subfloat}
\begin{subfloat}[]{
\includegraphics[width=0.31\linewidth]{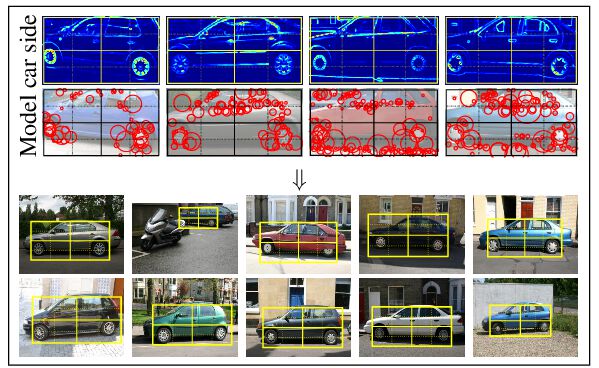}}
\end{subfloat}\\
\begin{subfloat}[]{
\includegraphics[width=0.95\linewidth]{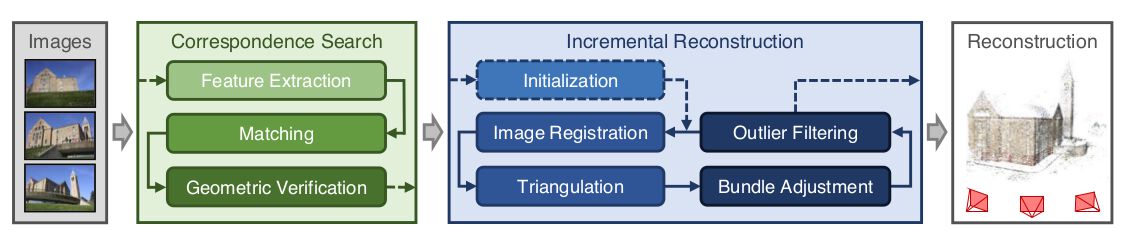}}
\end{subfloat}%

\caption{Application of the wide baseline stereo in the different computer vision tasks. All the images are taken from the cited papers respectively. (a) Bag of words image search. Image credit: Ondra Chum and \fr. (b) Bag of local features representation for classification~\cite{Fergus03} (c) Exemplar-representation of the classes using local features~\cite{Chum2007Exemplar}, (d) SfM pipeline from COLMAP~\cite{COLMAP2016}}
\label{fig:cv-as-wbs-problems}
\end{figure}

\begin{itemize}
\item
  \textbf{Scalable image search}. Sivic and Zisserman in the famous "Video  Google" paper~\cite{VideoGoogle2003} proposed to treat local features as "visual words" and use ideas from text processing for searching in
  image collections. Later, even more WBS elements were reintroduced to image search, most notably -- \textbf{spatial
  verification}~\cite{Philbin07}: a simplified RANSAC procedure to verify if visual word matches were spatially consistent.
\item
  \textbf{Image classification} was performed by placing a classifier (SVM, random forest, etc) on top of some encoding of the SIFT-like descriptors, extracted sparsely -- by Fergus~\etal~\cite{Fergus03}, Dance~\etal\cite{CsurkaBoK2004} or densely by Lazebnik~\etal~\cite{Lazebnik06}.

\item
  \textbf{Object detection} was formulated as a relaxed wide baseline stereo problem -- by Chum and Zisserman~\cite{Chum2007Exemplar} or as a classification of SIFT-like features inside a sliding window -- by Dalal and Triggs~\cite{HoG2005}.

\item
  \textbf{Semantic segmentation} was performed by a classification of the local region descriptors, typically SIFT and color features, and postprocessing afterwards -- by Tighe and Lazebnik~\cite{Superparsing2010}.
\end{itemize}

\begin{figure}
\centering
\includegraphics[width=0.95\linewidth]{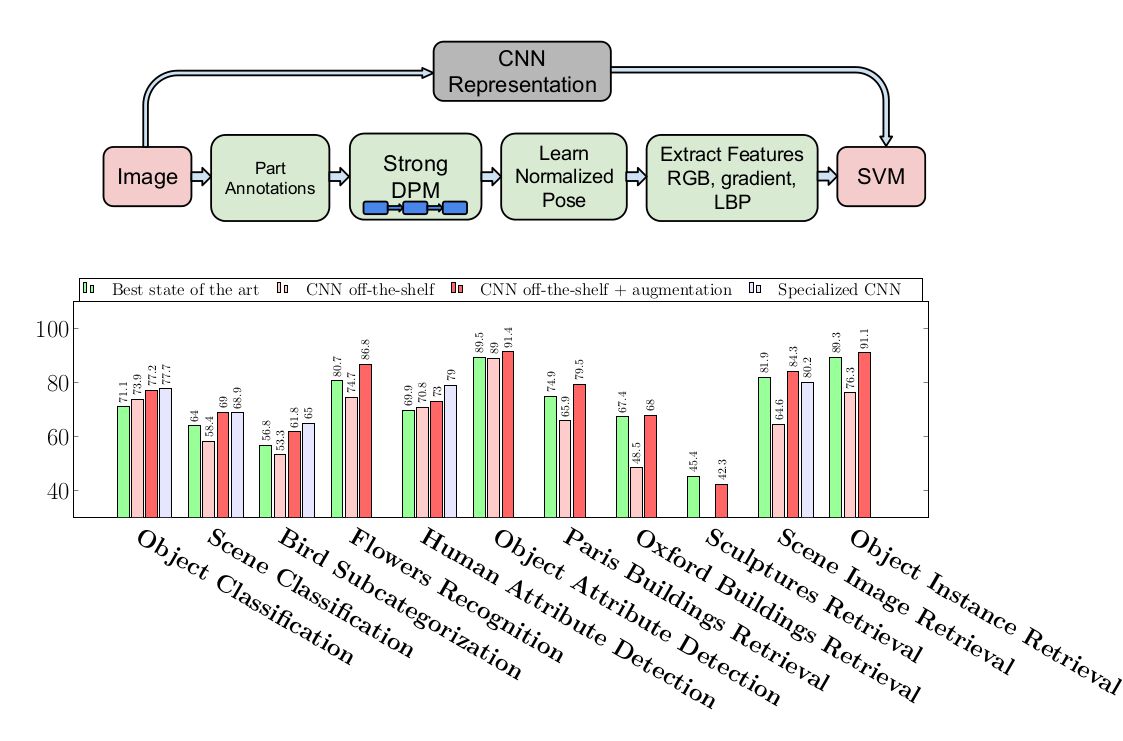}
\caption{CNN representation beats complex traditional pipelines. Reds are CNN-based and greens are the handcrafted. From \cite{Astounding2014}.}
\label{fig:cnn-beats-handcrafted}
\end{figure}

\textbf{Deep learning invasion: Retreat to the geometrical fortress}.  In 2012 the deep learning-based AlexNet~\cite{AlexNet2012} approach beat all methods in image classification at the ImageNet Large Scale Visual Recognition Challenge (ILSVRC). Soon after, Razavian~\etal~\cite{Astounding2014} have shown that convolutional neural networks (CNNs) pre-trained on the Imagenet outperform more complex traditional solutions in image and scene classification, object detection and image search, see Figure~\ref{fig:cnn-beats-handcrafted}. The performance gap between deep leaning and ``classical'' solutions was large and quickly increasing. In addition, deep learning pipelines, be it off-the-shelf pretrained, fine-tuned or the end-to-end learned networks, are simple from the engineering perspective. That is why the deep learning algorithms quickly become the default option for lots of computer vision problems. 
It worth noting, that the first paper, which uses CNNs and deep learning for the local feature description came out 4 years before the AlexNet -- in 2008, proposen by Jahrer~\etal~\cite{Jahrer2008}~\footnote{Thanks to Krystian Mikolajczyk for the pointing that out!}.%

However, there was still a domain, where deep learned solutions failed, sometimes spectacularly: geometry-related tasks. Wide baseline stereo~\cite{Melekhov2017relativePoseCnn}, visual localization~\cite{PoseNet2015} and SLAM are still areas, where the classical wide baseline stereo dominates~\cite{sattler2019understanding, zhou2019learn, pion2020benchmarking}. The full reasons why convolution neural network pipelines are struggling to perform tasks that are related to geometry, and how to fix that, are yet to be understood. The observations from the recent papers are following:

\begin{itemize}
\item
  CNN-based pose predictions predictions are roughly equivalent to the retrieval of the most similar image from the training set and outputing its pose~\cite{sattler2019understanding}. This kind of behaviour is also observed in a related area: single-view 3D reconstruction performed by deep networks is essentially a retrieval of the most similar 3D model from the training set~\cite{Tatarchenko2019}.
\item
  Geometric and arithmetic operations are hard to represent via vanilla neural networks (i.e., matrix multiplication followed by non-linearity) and they may require specialized building blocks, approximating operations of algorithmic or geometric methods, e.g. spatial transformers~\cite{STN2015} and arithmetic
  units~\cite{NALU2018,NAU2020}. Even with such special-purpose components, the deep learning solutions require "careful initialization, restricting parameter space, and regularizing for sparsity"~\cite{NAU2020}.
\item
  Vanilla CNNs suffer from sensitivity to geometric transformations like scaling and rotation~\cite{GroupEqCNN2016} or even translation~\cite{MakeCNNShiftInvariant2019}. The sensitivity to translations might sound counter-intuitive, because the concolution operation by definition is translation-covariant. However, a typical CNN contains also zero-padding and downscaling operations, which break the covariance~\cite{MakeCNNShiftInvariant2019, AbsPositionCNN2020}. Unlike them, classical local feature detectors are grounded on scale-space~\cite{lindeberg2013scale} and image processing theories. Some of the classical methods deal with the issue by explicit geometric normalization of the patches before description.
\item
  CNNs predictions can be altered by a change in a small localized area~\cite{AdvPatch2017} or even a single pixel~\cite{OnePixelAttack2019}, while the wide baseline stereo methods require the consensus of different independent regions.
\end{itemize}

\begin{figure*}
\centering
\begin{subfloat}[]{
\includegraphics[width=0.23\linewidth]{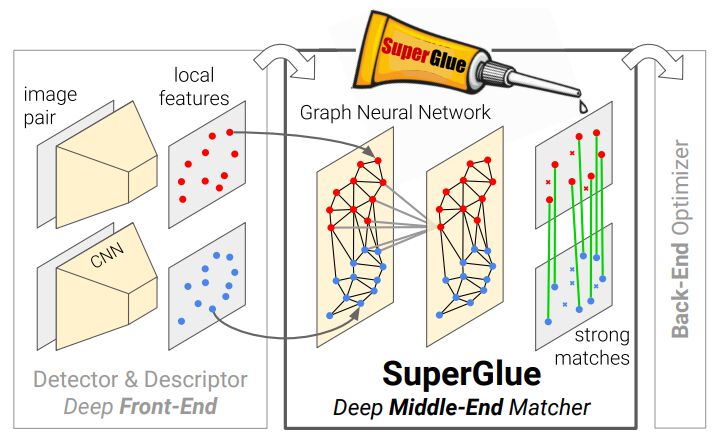}}
\end{subfloat}
\begin{subfloat}[]{
\includegraphics[width=0.73\linewidth]{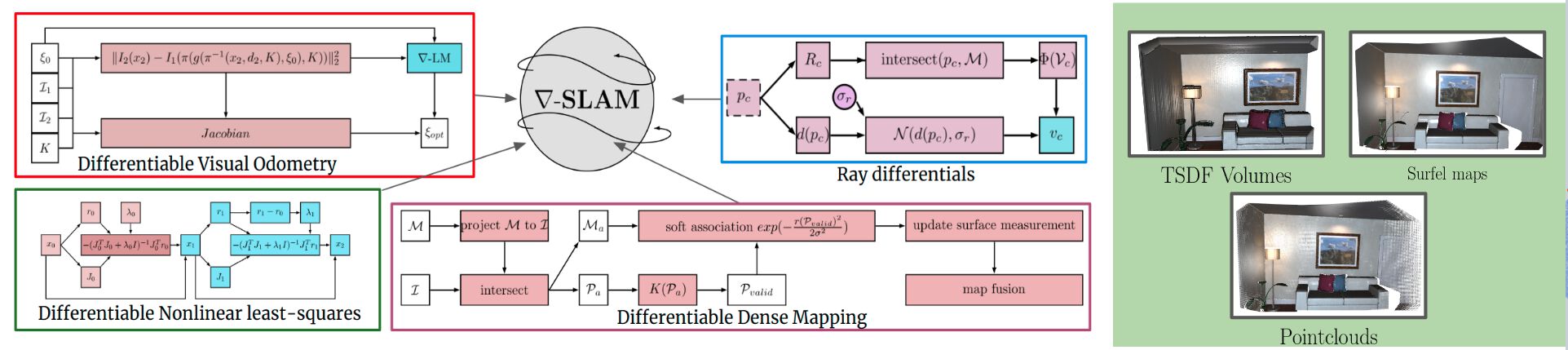}}
\end{subfloat}
\caption{Modern approach to wide multiple baseline stereo: use learned components in a classical structure pipeline or make the whole pipeline differentiable. Images taken from the cited papers. (a) SuperGlue~\cite{sarlin2019superglue}: separate matching module for handcrafted and learned features. (b) gradSLAM: differentiable formulation of SLAM pipeline~\cite{gradslam2020}}
\label{fig:wbs-as-task}
\end{figure*}

This leads us to the following question -- \textbf{is deep learning helping WxBS today?} The answer is yes. After the quick interest in the black-box-style models, the current trend is to design deep learning solutions for the wide baseline stereo in a modular fashion~\cite{cv4action2019}, resembling the one in Figure~\ref{fig:wbs-pipeline}. Such modules are learned separately. For example, the HardNet~\cite{HardNet2017} descriptor replaces SIFT local descriptor. The Hessian detector can be replaced by deep learned detectors like KeyNet~\cite{KeyNet2019} or the joint
detector-descriptor~\cite{SuperPoint2017, R2D22019, D2Net2019}. The matching and filtering are performed by the SuperGlue~\cite{sarlin2019superglue} matching network, etc. There have been attempts to formulate the full pipeline solving problem like SLAM~\cite{gradslam2020} in a differentiable way, combining the advantages of structured and learning-based approaches. Examples of such structured and modular architectures are shown in Figure~\ref{fig:wbs-as-task}.

\begin{figure*}
\centering
\begin{subfloat}[]{
\includegraphics[height=3cm]{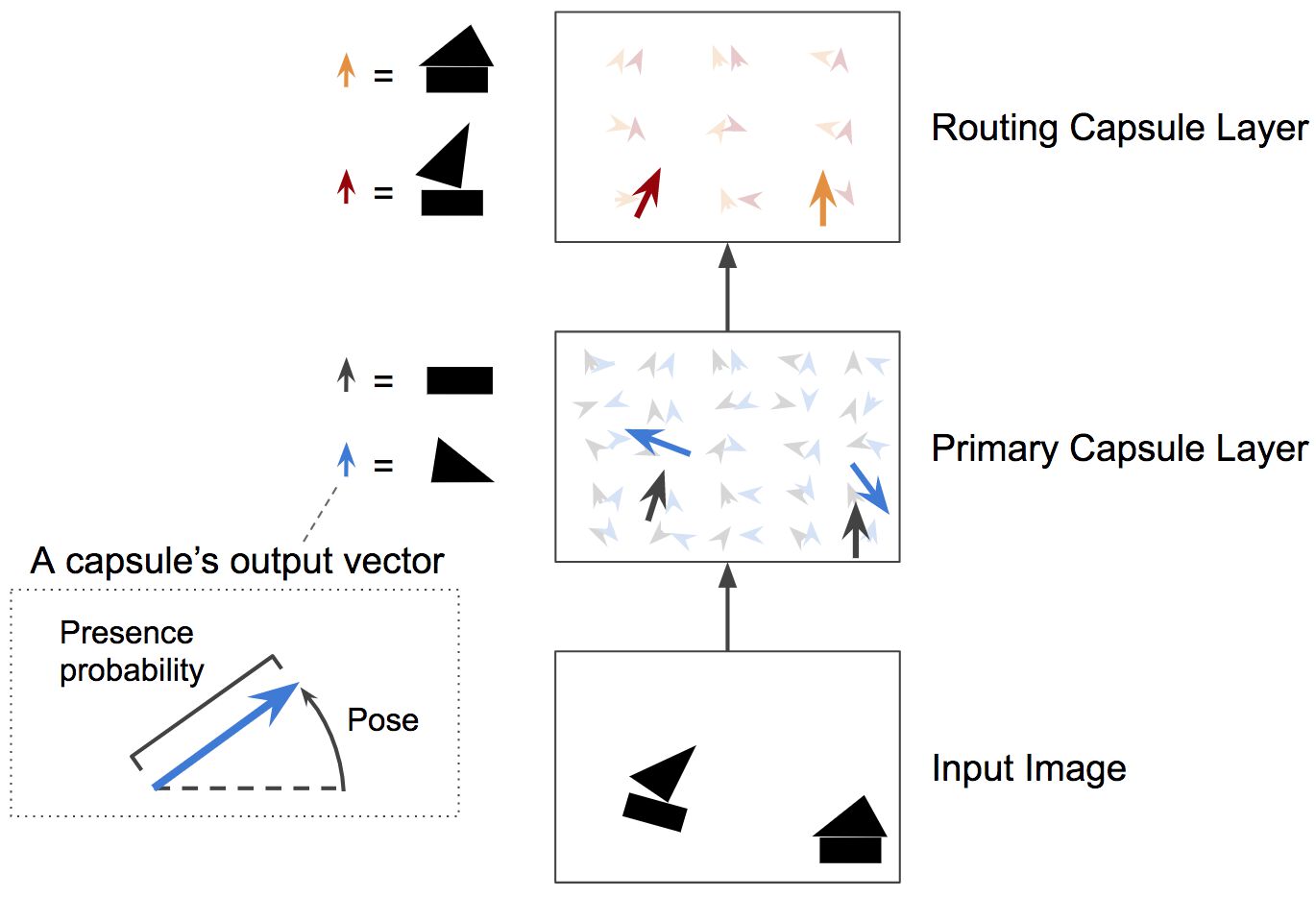}}
\end{subfloat}%
\begin{subfloat}[]{
\includegraphics[height=3cm]{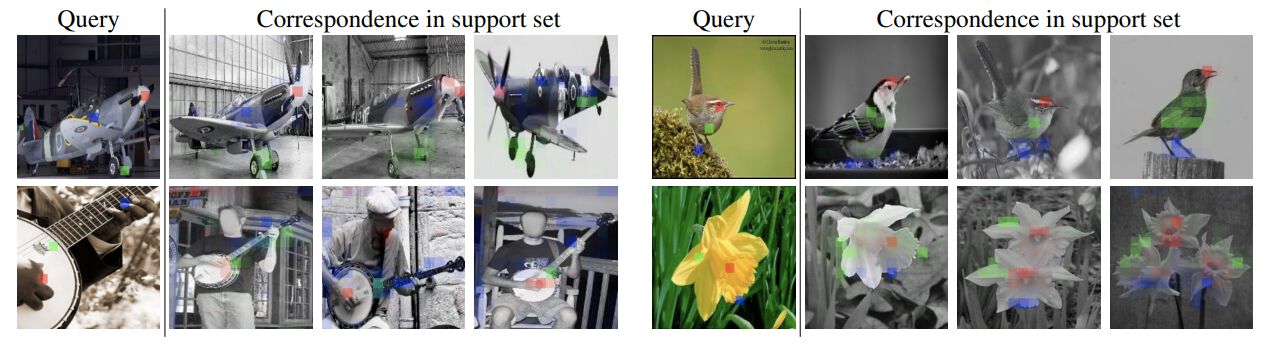}}
\end{subfloat}\\
\caption{Wide baseline stereo ideas in the modern neural network architectures. (a) Capsule networks~\cite{CapsNet2017}: revisiting the WBS idea. Each feature response is accompanied with its pose. Feature poses should be in agreement, otherwise object would not be recognized. Image by Aurélien Géron %
(b) Cross-transformers: spatial attention helps to select relevant feature for few-shot learning~\cite{doersch2020}.
}
\label{fig:wbs-ideas}
\end{figure*}

\textbf{Is WxBS helping deep learning?} The WxBS is helping the progress of deep learning-based computer vision research in three main ways. First is providing supervision and labeling of the data as a result of running structure-from-motion pipelines like COLMAP. Produced 3D models of the scenes are used for training image retrieval~\cite{Radenovic-ECCV2016CNNfromBOW, Radenovic-arXiv17}, monocular depth estimation~\cite{MegaDepth2018}, local features~\cite{LFNet2018,LogPolar2019}, correspondence filtering~\cite{CNe2018, Sun19} and other models. The second way is to incorporate geometry-based constraints into the learning~\cite{LeftRightMonoDepth2017, Monet2019, CameraPoseSupervisionDesc2020}. 
The third is the influence of the ideas from WxBS on the design of new model types.  Specifically, capsule networks~\cite{CapsNet2011,CapsNet2017} are based on two ideas, rooted in wide baseline stereo: an image should be represented as a set of local parts, robust to occlusion. The second idea is that the decision should be based on the spatial consensus of local feature correspondences. Unlike vanilla CNNs, capsule networks encode not only the intensity of feature response, but also its location. Geometric agreement between ``object parts'' is a requirement for outputing a confident prediction. Similar ideas are explored for improving the adversarial robustness of CNNs~\cite{li2020extreme} or for semi-supervised learning~\cite{doersch2020}, where the spatial consistency is employed to provide supervision in the few-shot learning setup. Examples of such consensus are shown in Figure~\ref{fig:wbs-ideas}.

While wide multiple baseline stereo is a mature field now and does not attract even nearly as much attention as before, it continues to play an important role in computer vision.

\section{Contributions}
\label{intro:contributions}

The contributions of this thesis are:

\begin{enumerate}

\item 
The introduction of the concept of the wide multiple baseline stereo (\textsc{WxBS})~\cite{WXBS2015} problem. WxBS is a generalization of the standard wide baseline stereo problem that considers matching of images that simultaneously differ in more than one image acquisition factor such as viewpoint, illumination, sensor type or where object appearance changes significantly, e.g. over time. A new dataset with the ground truth, evaluation metric and baselines was introduced.

\item
A loss function~\cite{HardNet2017} for learning a local image descriptor that relies on hard negative mining within a mini-batch and the maximization of the distance between the closest positive and closest negative patches. The resulting descriptor called HardNet is compact -- it has the same dimensionality as SIFT (128), it shows state-of-art performance in standard matching, patch verification and retrieval benchmarks. HardNet is fast to compute on a GPU. This work was done in collaboration with Anastasiia Mishchuk, who is first author of the paper~\cite{HardNet2017}. 

\item A loss function for descriptor-based registration and learning, named the hard negative-constant loss~\cite{AffNet2018}. It combines the advantages of the triplet and contrastive positive losses.

\item We experimentally show that geometric repeatability of the local feature is not a sufficient condition for successful feature matching~\cite{AffNet2018}. The learning of affine shape increases the number of corrected matches if it steers the estimators towards discriminative regions and therefore must involve optimization of descriptor-related loss.

\item A method for learning the affine shape, orientation and potentially other parameters related to  geometric and appearance properties of local features~\cite{AffNet2018}. The learning method  does not require a precise ground truth which reduces the need for manual annotation. 
    
\item  A tentative correspondences generation strategy is presented which generalizes the standard first to second closest distance ratio~\cite{MODS2015}. The selection strategy which shows performance~\cite{PatchExtraction2014} superior to the standard method is applicable to either hard-engineered descriptor like SIFT, LIOP and MROGH or deeply learned like HardNet.

\item  MODS -- matching with on-demand view synthesis. MODS is a two-view matching algorithm that handles viewing angle difference even larger than the previous state-of-the-art ASIFT algorithm, without a significant increase of computational costs over ``standard'' wide and narrow baseline approaches~\cite{Mishkin2013, MODS2015}. 

\item  EVD -- extreme view dataset for benchmarking the two-view matching algorithms under extreme viewpoint changes~\cite{Mishkin2013, MODS2015}. 

\item A comprehensive benchmark~\cite{IMW2020} for local features and robust estimation algorithms. The modular structure of its pipeline allows to easily integrate, configure, and combine methods and heuristics. This work is done in a wide collaboration across four labs. I focused on the benchmark methodology, specifically importance of RANSAC and matching tuning in order to properly evaluate local feature methods. 

\end{enumerate}

\section{Publications}
\label{intro:publications}

The presented work is published in the following papers:

\begin{labeling}{[200]}

\item[\cite{MODS2015}] \textbf{Dmytro Mishkin}, \jm, and Michal Perdoch. MODS: Fast and robust method for two-view matching. Computer Vision and Image Understanding, 141:81 – 93, 2015.

\item[\cite{WXBS2015}]\textbf{Dmytro Mishkin}, \jm, Michal Perdoch, and Karel Lenc. WxBS: Wide baseline stereo generalizations. In BMVC, 2015.

\item[\cite{HardNet2017}]Anastasiia Mishchuk, \textbf{Dmytro Mishkin}, \fr{} and \jm. Working Hard to Know Your Neighbor’s Margins: Local Descriptor Learning Loss. In NeurIPS, 2017.

\item[\cite{AffNet2018}]\textbf{Dmytro Mishkin}, \fr{}, and \jm. Repeatability is Not Enough: Learning Affine Regions via Discriminability. In ECCV, 2018.

\item[\cite{IMW2020}]Yuhe Jin, \textbf{Dmytro Mishkin}, Anastasiia Mishchuk, \jm, Pascal Fua, Kwang Moo Yi, and Eduard Trulls. Image matching across wide baselines: From paper to practice. IJCV, 2020.

\end{labeling}

The differentiable implementation of some of the abovementioned contributions is included into an open source library named \textbf{kornia}~\cite{kornia2020}. The paper describing the library is however, of little scientific interest and we do not include it into the thesis. 

\begin{labeling}{[200]}

\item[\cite{kornia2020}] Edgar Riba, \textbf{Dmytro Mishkin}, Daniel Ponsa, Ethan Rublee, Gary Bradski. Kornia: an open source differentiable computer vision library for PyTorch. In WACV, 2020.

\end{labeling}

Some of the results presented in this thesis and related to HardNet descriptor, were obtained with Milan Pultar, who completed the master thesis under my supervision. They are presented in full detail in his master thesis~\cite{Pultar2020improving}. 

Finally, the following publications have not been included into the thesis, either because they are less relevant to the thesis main topic (\cite{Mishkin2015VPRICE},\cite{Mishkin2016LSUV},  \cite{Systematic2017}, \cite{deblurgan2018}, \cite{barath2020efficient}), or because the author contribution is small(\cite{Saddle2016}, \cite{deblurgan2018}, \cite{HardNetAMOS2019}, \cite{saddle2019}):

\begin{labeling}{[200]}

\item[\cite{Mishkin2013}] \textbf{Dmytro Mishkin}, Michal Perdoch, and \jm. Two-view matching with view synthesis revisited.
In Proc of Image and Vision Computing New Zealand (IVCNZ), 2013.

\item[\cite{PatchExtraction2014}] Karel Lenc, \jm, and \textbf{Dmytro Mishkin}. A few things one should know about feature extraction. Computer Vision Winter Workshop, pages 67 -- 74, 2014

\item[\cite{Mishkin2015VPRICE}] \textbf{Dmytro Mishkin}, Michal Perdoch and \jm. Place Recognition with WxBS Retrieval. CVPR 2015 Workshop on Visual Place Recognition in Changing Environments.

\item[\cite{Mishkin2016LSUV}]\textbf{Dmytro Mishkin} and \jm. All you need is a good init. In ICLR, 2016.

\item[\cite{Systematic2017}] \textbf{Dmytro Mishkin}, Nikolay Sergievskiy and \jm. Systematic evaluation of convolution neural network advances on the imagenet. Computer Vision and Image Understanding, 161:11 – 19, 2017.

\item[\cite{Saddle2016}]Javier Aldana-Iuit, \textbf{Dmytro Mishkin}, Ondra Chum and \jm. In the Saddle: Chasing Fast and Repeatable Features. In ICPR, 2016. 

\item[\cite{deblurgan2018}]Orest Kupyn,  Volodymyr Budzan, Mykola Mykhailych, \textbf{Dmytro Mishkin} and \jm. DeblurGAN: Blind Motion Deblurring Using Conditional Adversarial Networks. In CVPR 2018

\item[\cite{HardNetAMOS2019}]Milan Pultar, \textbf{Dmytro Mishkin} and \jm. Leveraging Outdoor Webcams for Local Descriptor Learning. In Computer Vision Winter Workshop, 2019.

\item[\cite{saddle2019}]Javier Aldana-Iuit, \textbf{Dmytro Mishkin}, Ondej Chum and \jm. Saddle: Fast and repeatable features with good coverage. Image and Vision Computing, 2019.

\item[\cite{barath2020efficient}] Daniel Barath, \textbf{Dmytro Mishkin}, Ivan Eichhardt and \jm. Efficient Initial Pose-graph Generation for Global SfM. In CVPR, 2021 (to appear).

\end{labeling}

\chapter[WxBS: Wide multiple baseline stereo problem]{Wide multiple baseline stereo \\
\includegraphics[height=24mm,width=4cm]{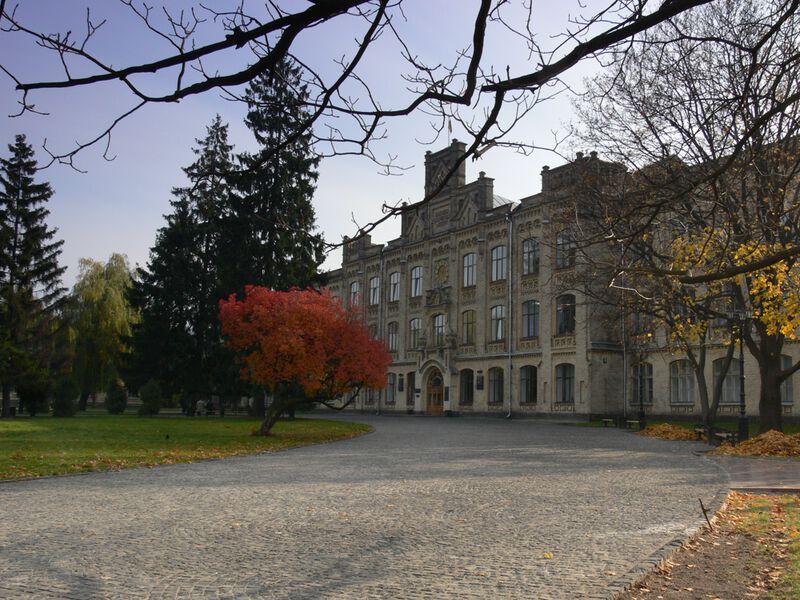}
\includegraphics[height=24mm,width=4cm]{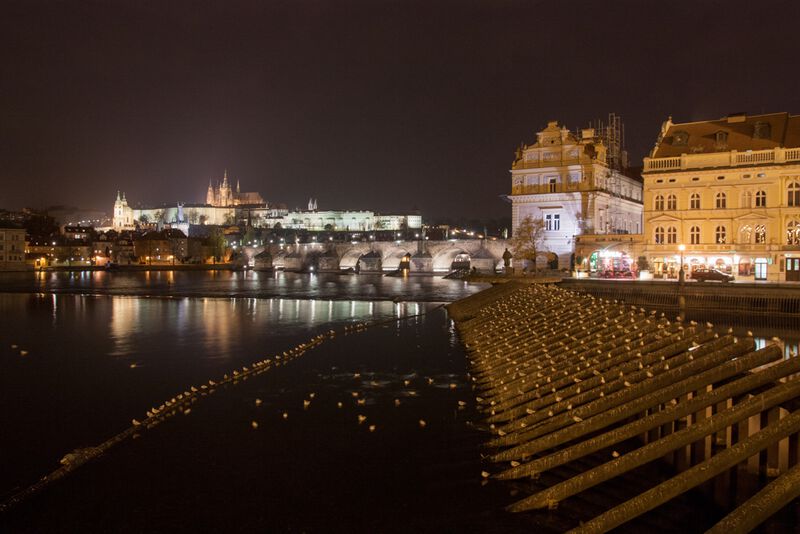}
\includegraphics[height=24mm,width=4cm]{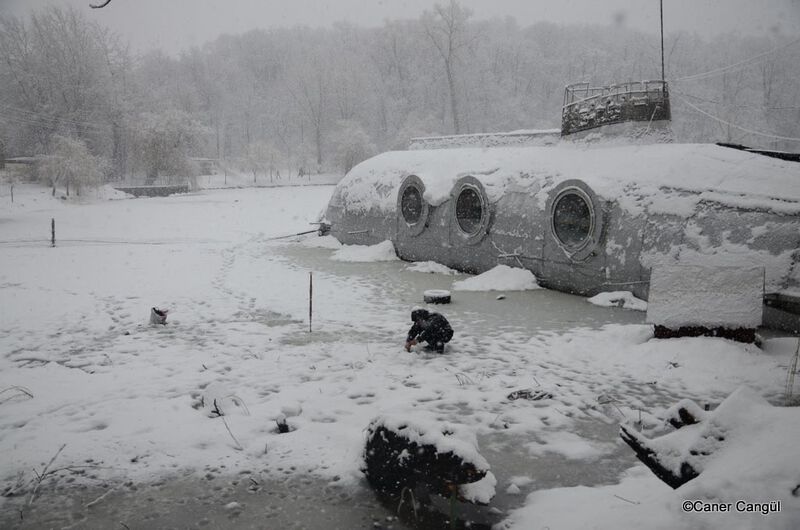}\\
\includegraphics[height=24mm,width=4cm]{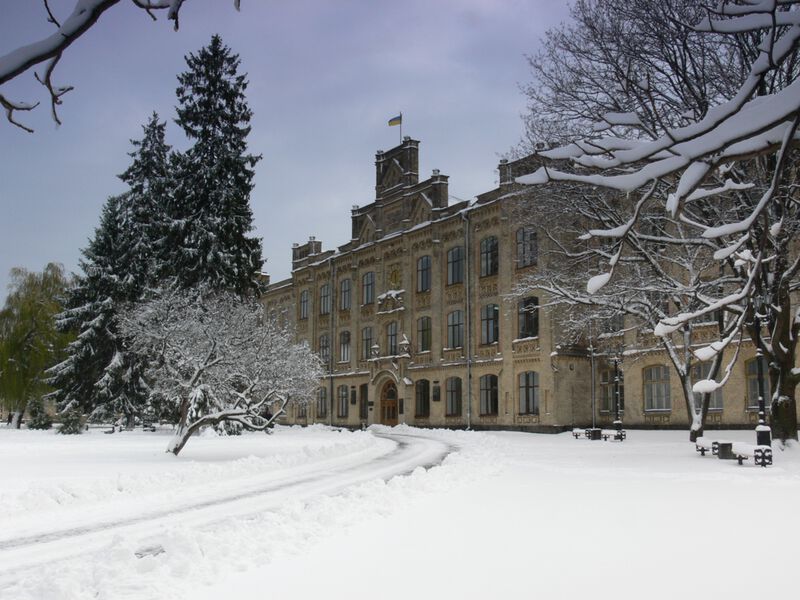}
\includegraphics[height=24mm,width=4cm]{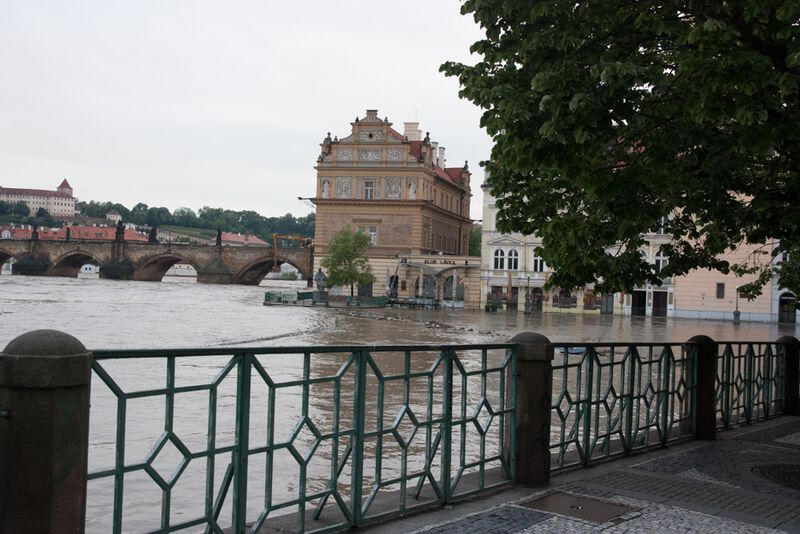}
\includegraphics[height=24mm,width=4cm]{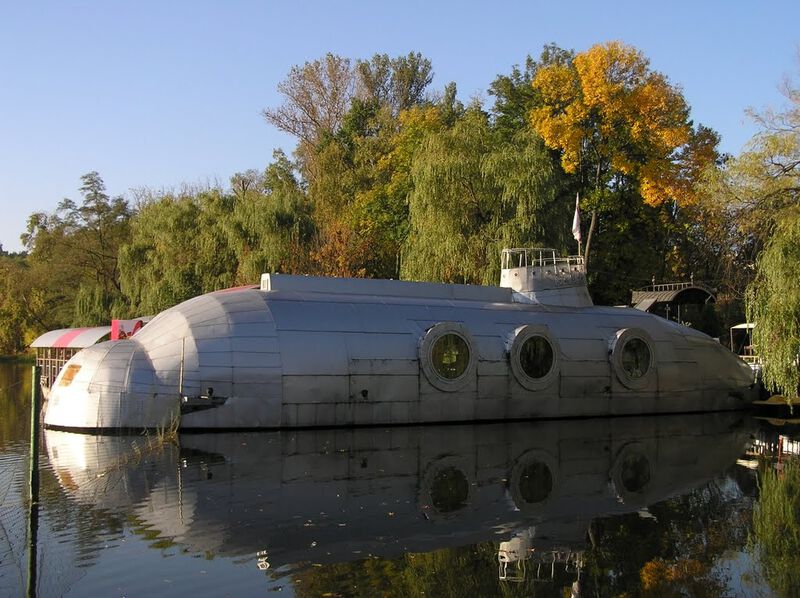}}
\label{ch:wxbs}
\section{Definition}
\label{wxbs:definition}
\begin{figure*}[tb]
\centering
\includegraphics[width=0.95\linewidth]{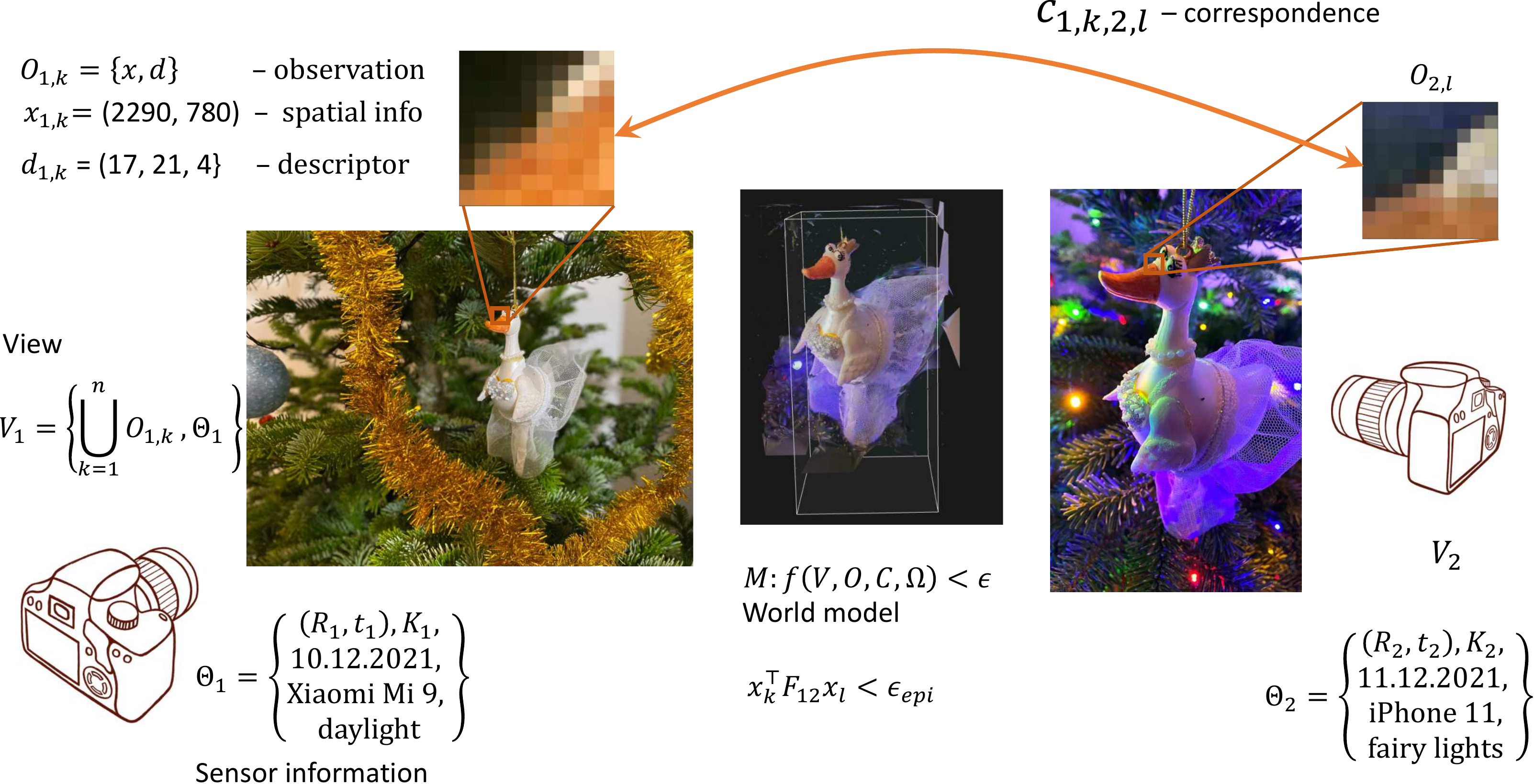}
 \caption{Wide multiple baseline stereo as a joint recovering the world model parameters $
\Omega$ and correspondences $c_{i,k,j,l}$}
  \label{fig:wxbs-slide}
\end{figure*}

Let us denote observations $O_{j,i}, i=1..n$, each of which belongs to one of the views $V_{j}, j=1..m$, $m \leq n$.  Observation $O_{j,i} = \{x;d\}$ consist of spatial information $x$ and the descriptor $d$; index $i$ is a unique index among all observations. View $V_j = \{\bigcup O_{j,\cdot} ; \theta_j\}$ consists of union of the observations and the additional information $\theta$, which is shared for all observations, which belong to view $V_j$. Check Figure~\ref{fig:wxbs-slide} for the illustration.

For example, a single observation can be an RGB pixel. Its spatial information is the pixel coordinates $x=(u,v)  \in R^2$ and the descriptor $d = (r,g,b) \in \{(0..255)^3\} $ is its color intensity value. The view $V$ then is the image, with information $\theta$ containing the camera pose, camera intrinsics, sensor type, illumination conditions, timestamp of the acquisition, etc. Some of this information can be unknown to the user, i.e., a hidden variable.  Another example could be an event camera~\cite{EventCameraSurvey2020}. In that case, the observation $O$ contains the pixel coordinates $x=(u,v) \in R^2 $ and the descriptor $d \in \{+, -, 0\}$ is the sign of the intensity change. The view $V$ will contain the information about the sensor, camera pose, timestamp and the single observation $O$, because every event has a unique timestamp. Observations and views can be of different nature and dimensionality. E.g., views $V_1$, $V_2$ might be RGB images, $V_3$ -- point cloud from a laser scaner, $V_4$ -- an image from a thermal camera, and so on. 

An unordered pair of observations $(O_{j,i},O_{k,l})$ forms a correspondence $c_{jikl}$ if they are belong to different views $V_{j} \ne V_{k}$. A set of observations is called multiview correspondence $C_o$, when there is no more than one observation $O_{j,i}$ per view $V_j$ $\forall{j} $. Some of observations $O_{j,i}$ can be empty $\varnothing$, i.e. not observed in the specific view $V_j$.

The world model $M$ is a system of contraints on views, observations and correspondences $M: f(V, O, C, \Omega) = 0$, where $\Omega$ stands for the world model parameters. For example, one of the popular models are epipolar geometry and rigid motion assumption: $x^{'T}Fx = 0$, where $\Omega$ is a fundamental matrix~\cite{Hartley2004} $F$ for this case. In practice, however, the exact satisfaction of the ground truth model is unlikely due to the measurement errors and noise. That is why we also incorporate an acceptable noise level $\epsilon$ into our model $M: f(V, O, C, \Omega) < \epsilon $. The correspondence is called ground truth or veridical if it satisfies the constraints posed by the ground truth world model. 

We can now define a wide baseline stereo. By the wide baseline stereo we understand the estimation of the unknown parameters $\Omega$ of the world model $M^{*}$ by establishing the correspondences which satisfy this world model, given the views and observations.

By the baseline we understand the difference between views information $\theta_j$. If views $V_j$ are significantly different in a single aspect, e.g., camera pose, we will call the task wide single baseline stereo. If more than one aspect is different, e.g., both camera pose and sensor type are different, we will call the task wide multiple baseline stereo, or WxBS.

Moreover, we limit our definition of the WxBS by setting the restriction on the possible world models. Most often in this thesis we will be using the the following world model. The scene is rigid and static. The observations are 2D projections of the 3D scene to the camera plane by the pinhole or rectilinear camera. The relationship between observations in different views is either epipolar geometry or projective transform~\cite{Hartley2004}. Any moving object does not satisfy the world model and therefore is considered an occlusion.

Let us define the areas similar to WxBS, mentioned in Chapter~\ref{ch:intro} using the framework we have just developed. The wide baseline stereo process, which in addition estimates the (latent) scene structure, e.g., in the form of 3d point world coordinates we would call Structure-from-Motion (SfM) or 3D reconstruction. We do not consider SfM in this thesis.

The optical flow is the process of establishing correspondences $c_{jikl}$ between the short temporal baseline views without recovering the world model parameters $\Omega$. The image registration problem is identical to the wide multiple baseline stereo in the definition, but the family of considered world models is different (wider). The visual localization is the estimation of the camera pose $\theta$ from the given view $V$.

\paragraph{Since the paper publication in 2015.} The rest of the chapter below is presented as was originally published in 2015. Since then, the interest to the topic, in the form of day-night~\cite{R2D22019, radenovic2016dusk, D2Net2019} image matching, visual relocalization in changing environments~\cite{6DOFPlace2017,Lynen19, melekhov2020image, Torii247} has been significant. A series of datasets and  benchmarks~\cite{hpatches2017, 6DOFPlace2017, Bian19} and local features~\cite{LIFT2016, R2D22019, D2Net2019, SuperPoint2017} have been proposed. We believe that we have only started to explore the area of wide multiple baseline stereo. 

\section{Wide multiple baseline stereo: evaluation}
\label{wxbs:wxbs}
We consider the generalization of Wide (geometric) Baseline Stereo to \textsc{WxBS}, a multiview image matching problem where two or more of the image formation and acquisition properties \emph{significantly} change, i.e. they have a \emph{  wide baseline}. The  "significant change" distinguishes the problem from image registration, where a dense correspondence is routinely established between multi-modal images and various complex transformations have been considered, see Zitov{\'a} and Flusser~\cite{Zitova2003}. %
For example, the standard physical models of image formation and acquisition consider, besides geometry,  the effects of illumination, the properties of the transparent medium where light rays pass through in the scene, the surface properties of objects and the properties of the imaging sensors.

The following single wide baseline stereo, or correspondence, problems and their combinations are considered: illumination (\textsc{WlBS}) -- differences in position, direction, number, intensity, and wavelength of light sources; geometry  (\textsc{WgBS}) -- difference in camera and object pose, scale and resolution - the ``classical'' WBS;
sensor  (\textsc{WsBS}) -- change in sensor type: visible, IR, MR; noise, image preprocessing algorithms inside the camera, etc;
appearance (\textsc{WaBS}) --  difference in the object appearance because of time or seasonal changes, occlusions, turbulent air,  etc.
We denote matching problems, or, equivalently, image pairs, with a significant change in only one of the groups listed  as {W{\sc1}BS}; if a combination of effects is present, as \textsc{WxBS}. To our knowledge, the most published to date image datasets and algorithms are in the \textsc{W1BS} class\cite{Mikolajczyk2005}, \cite{ASIFT2009}, \cite{Vonikakis2013},\cite{Aguilera2012},\cite{Hauagge2012}, \cite{Jacobs07}. 

We present a new public dataset with ground truth, which combines the above-mentioned challenges and contains both \textsc{W2BS} image pairs  including viewpoint and appearance, viewpoint and illumination, viewpoint and sensor, illumination and appearance change and \textsc{W3BS} -- problems where viewpoint, appearance and lighting differ significantly.

\begin{figure}[t]
\begin{center}
\def\svgwidth{1\textwidth}
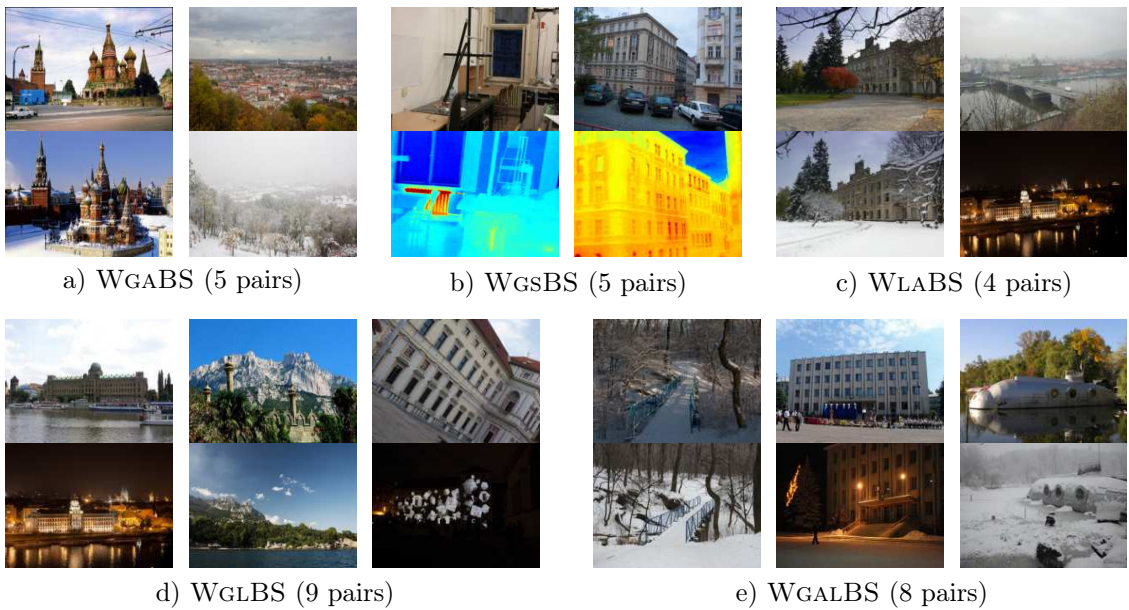
\end{center}
   \caption{Examples of image pairs from the \textsc{WxBS} dataset.}
\label{fig:wxbs-examples}
\vspace{-1em}
\end{figure}

\paragraph{WxBS dataset and evaluation protocol.} A set of 37 image pairs has been collected from Flickr, taken by the authors themselves and other sources. The dataset is divided into 6 categories based on the combinations of nuisance factors present, see Table~\ref{tab:wxbs-categories}. For every image, a set of approximately 20 ground-truth correspondences has been annotated. Selected examples are presented in Figure \ref{fig:wxbs-examples}.
The resolution of the majority of the images is $800 \times 600$ with the exception of LWIR images from the \textsc{WgsBS} dataset which were captured by a thermal camera with a resolution of $250 \times 250$ pixels.
The selected image pairs contain both urban and natural scenes.
\begin{table}[htb]
\caption{The WxBS datasets categories}
\vspace{0.5em}
\label{tab:wxbs-categories}
\footnotesize
\centering
\setlength{\tabcolsep}{.3em}
\begin{tabular}{llrr}
\hline
Short name & Nuisance & \#images & Avg. \# GT Corr. \\
\hline
\textsc{map2ph} & appearance (map to photo) & 6 pairs & homography provided\\
\textsc{WgaBS} & viewpoint, appearance & 5 pairs & 22 per img.\\
\textsc{WglBS} & viewpoint, lighting & 9 pairs & 21 per img.\\
\textsc{WgsBS} & viewpoint, modality & 5 pairs & 18 per img.\\
\textsc{WlaBS} & lighting, appearance& 4 pairs & 25 per img.\\
\textsc{WgalBS} & viewpoint, appearance, lighting& 8 pairs & 17 per img.\\
\end{tabular}
\end{table}

\paragraph{Ground truth and evaluation protocol.} In image registration tasks, it is often sufficient to define the ground truth as a homography between an image pair. However, the WxBS dataset contains significant viewpoint changes. In the case of a non-planar scene, a homography can, at best, cover the dominant plane. 

We assume that an ideal algorithm matches the majority of the scene content, thus our ground truth is a set of manually selected correspondences which evenly cover the part of the scene visible in both images. The average number of correspondences per image pair is shown in Table \ref{tab:wxbs-categories}. 

\paragraph{The evaluation protocol for the WxBS dataset.} For each image pair indexed with  $i \in \mathbb{Z}$  we have manually annotated a set of correspondences $(\mathbf{u}_i,\mathbf{v}_i) \in C_i$ where $\mathbf{u}$ and $\mathbf{v}$ are positions in the 1\textsuperscript{st} and the 2\textsuperscript{nd} image respectively. 
For epipolar geometry we use the \emph{symmetric epipolar distance}  and the \emph{symmetric reprojection error} for homography \cite{Hartley2004}.

Recall on ground truth correspondences $C_i$ of image pair $i$ and for geometry model $\mathbf{M}_i$ is computed as a function of a threshold~$\theta$
\begin{equation}
 \mathrm{r}_{i,\mathbf{M}_i}(\theta) = \frac{| (\mathbf{u}_i,\mathbf{v}_i) : (\mathbf{u}_i,\mathbf{v}_i) \in C_i,  e(\mathbf{M}_i,\mathbf{u},\mathbf{v}) < \theta|}{| C_i |}
\end{equation}
using appropriate error functions. For all pairs of each category $W$ we define an overall recall per category as:
\begin{equation}
 \mathrm{r}_{W}(\theta) = \frac{1}{| W | \sum_{i \in W} \mathrm{r}_{i,M_i}(\theta)}
\end{equation}
This measure is as the fraction of the confirmed annotated correspondences for a given threshold in a nuisance category.

\section{Evaluation of local descriptors on single baseline datasets}
\label{wxbs:results}
Nowadays, there is a series of benchmarks for local feature descriptors and detectors, most notably HPatches\cite{hpatches2017}, Image Matching Challenge~\cite{IMW2020}, Brown dataset~\cite{Brown07} and so on. However, in 2015 many of them were not existed and we have proposed \textsc{W1BS} local patch dataset for the evaluation of the descriptors and detectors. It is also used in Chapters~\ref{ch:hardnet}, \ref{ch:local-affine-features}, \ref{ch:mods}, so we will describe it here together with the initial results obtained with it. 

Datasets used in the experiments are listed in Table~\ref{tab:all-datasets}. When evaluating detectors (Section~\ref{wxbs:results}) all dataset images are used. 
However, descriptor evaluation is performed only on a subset of the most challenging and prominent pairs (i.e., only pairs 1-6 from OxfordAffine) with provided homography of each \textsc{WxBS} category.

Most of the published datasets (with exception of the LostInPast dataset~\cite{LostInPast2015}) include only a single nuisance factor per image pair.

This is suitable for the evaluation of the robustness to a particular nuisance factor but fails to predict performance in more complex environments. One of the motivations of the proposed \text{WxBS} dataset is to address this issue.
\begin{table}[htb]
\caption{Datasets used for evaluation}
\label{tab:all-datasets}
\vspace{1em}
\scriptsize
\centering
\setlength{\tabcolsep}{.3em}
\begin{tabular}{llll}
\hline
Short name& Proposed by& \#images&Type\\
\hline
GDB&Kelman~\etal~\cite{Kelman2007}, 2007&22 pairs& \textsc{WlBS}, \textsc{WsBS}\\
SymB&Hauagge and Snavely~\cite{Hauagge2012}, 2012&46 pairs&\textsc{WaBS}, \textsc{WlBS}\\
MMS& Aguilera~\etal~\cite{Aguilera2012}, 2012&100 pairs&\textsc{WsBS}\\
EVD&Mishkin~\etal~\cite{Mishkin2013}, 2013&15 pairs&\textsc{WgBS}\\
OxAff&Mikolajczyk~\etal\cite{MikoDescEval2003},~\cite{Mikolajczyk2005}, 2013&8 sixplets&\textsc{WgBS}\\
EF&Zitnick and Ramnath~\etal\cite{Zitnick2011},2011&8 sixplets&\textsc{WgBS},\textsc{WlBS}\\
Amos&Jacobs~\etal\cite{Jacobs07},2007& $>$ 100K&\textsc{WlBS},\textsc{WaBS}\\
VPRiCE&VPRICE Challenge 2015~\cite{VPRICE2015}& 3K pairs&\textsc{WgaBS}, \textsc{WglBS},\textsc{WgsBS},\\
Past&Fernando~\etal\cite{LostInPast2015}, 2014& 502 images&\textsc{WgaBS}\\
WxBS&here& 37 pairs& \textsc{WaBS},\textsc{WgaBS},\textsc{WglBS}, \textsc{WgsBS},\textsc{WlaBS},\textsc{WgalBS} \\
\end{tabular}
\end{table}
\paragraph{ Descriptor evaluation}
The evaluation protocol is as follows. The dataset consists of 40 image pairs from the datasets listed in Table~\ref{tab:all-datasets} divided into 5 parts
by the nuisance factor. For all pairs, homography is the appropriate two-view relationship -- the images are either without significant relative depth, or taken from virtually identical viewpoints.  To minimize bias towards a specific detector, affine-covariant regions by Hessian-Affine~\cite{HesHarAff2004}, MSER~\cite{MSER2002} and FOCI~\cite{Zitnick2011} in the first -- least challenging image of the pair are used (visible in the case of IR-vis, day for day-night, frontal when view point changes, etc.). 
The affine-covariant regions have been detected with dominant orientation and then reprojected to the second image by the ground truth homography. Features which are not visible in the second image have been discarded. Therefore, the geometric repeatability of affine regions on the selected regions is always $100\%$ and the maximum possible recall is 1. Color-to-grayscale image transformation has been done via channel averaging, which gives the best matching performance~\cite{Kanan2012}.

Then the affine regions were normalized to patch size 41x41 (scale $\sigma = 3\sqrt{3}$) and described with given descriptors. An affine-normalization procedure is performed even for the fast binary descriptors, which is rarely used because of the significant additional processing time. However, the goal of our experiment is to explore the descriptor performance in challenging conditions, not their speed. The procedure helps -- the typical threshold of the Hamming distance for binary descriptors  on unnormalized patch is around 60-80, while on affine normalized patches similar performance is obtained with a threshold around 10-30. All descriptors clearly benefit from the affine-normalized process, e.g., the graffiti 1-6 pair from the OxfordAffine dataset could be matched with FREAK descriptor only when using a normalized patch.

The tested descriptors are: SIFT~\cite{SIFT2004}, rSIFT~\cite{RootSIFT2012}, hrSIFT (gradients in interval $\left[0;\pi\right)$)~\cite{Kelman2007}, InvSIFT (SIFT with reordered cells as for inverted image)~\cite{Hare2011}, LIOP\cite{LIOP2011}, AKAZE~\cite{AKAZE2013}, MROGH~\cite{MROGH2011}, FREAK~\cite{FREAK2012}, ORB~\cite{ORB2011}, SymFeat~\cite{Hauagge2012}, SSIM~\cite{Shechtman2007} (implementation \cite{Chatfield2009}), DAISY~\cite{Tola2010} and $L_2$-normalized raw grayscale pixel intensities. 
\newcommand{\smallImgWidth}{0.16}
\begin{figure}[tb]
\centering
\begin{subfloat}[\texttt{WGBS}]
{\includegraphics[width=\smallImgWidth\linewidth]{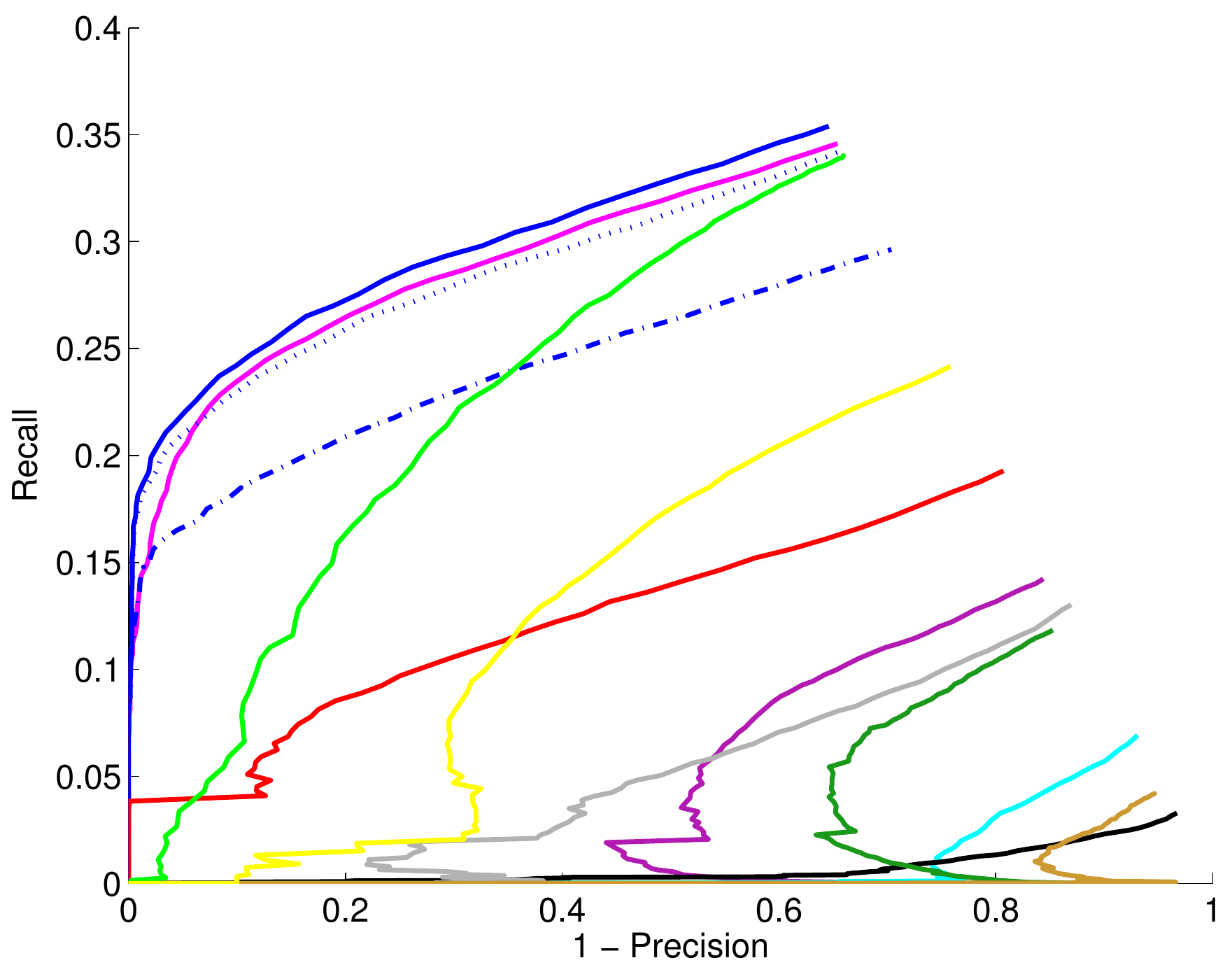}}
\end{subfloat}
\begin{subfloat}[\texttt{WABS}]
{\includegraphics[width=\smallImgWidth\linewidth]{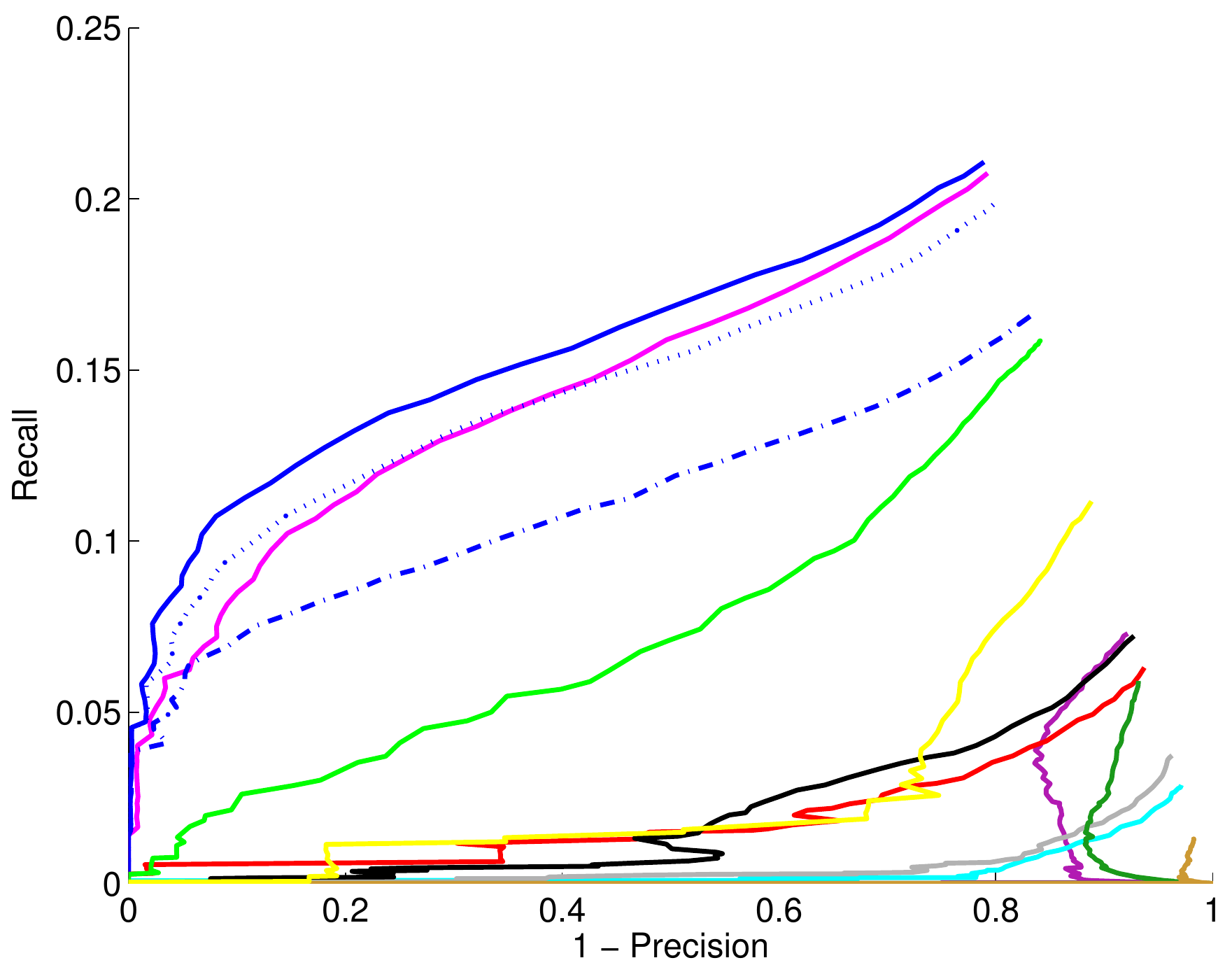}}
\end{subfloat}
\begin{subfloat}[\texttt{WLBS}]
{\includegraphics[width=\smallImgWidth\linewidth]{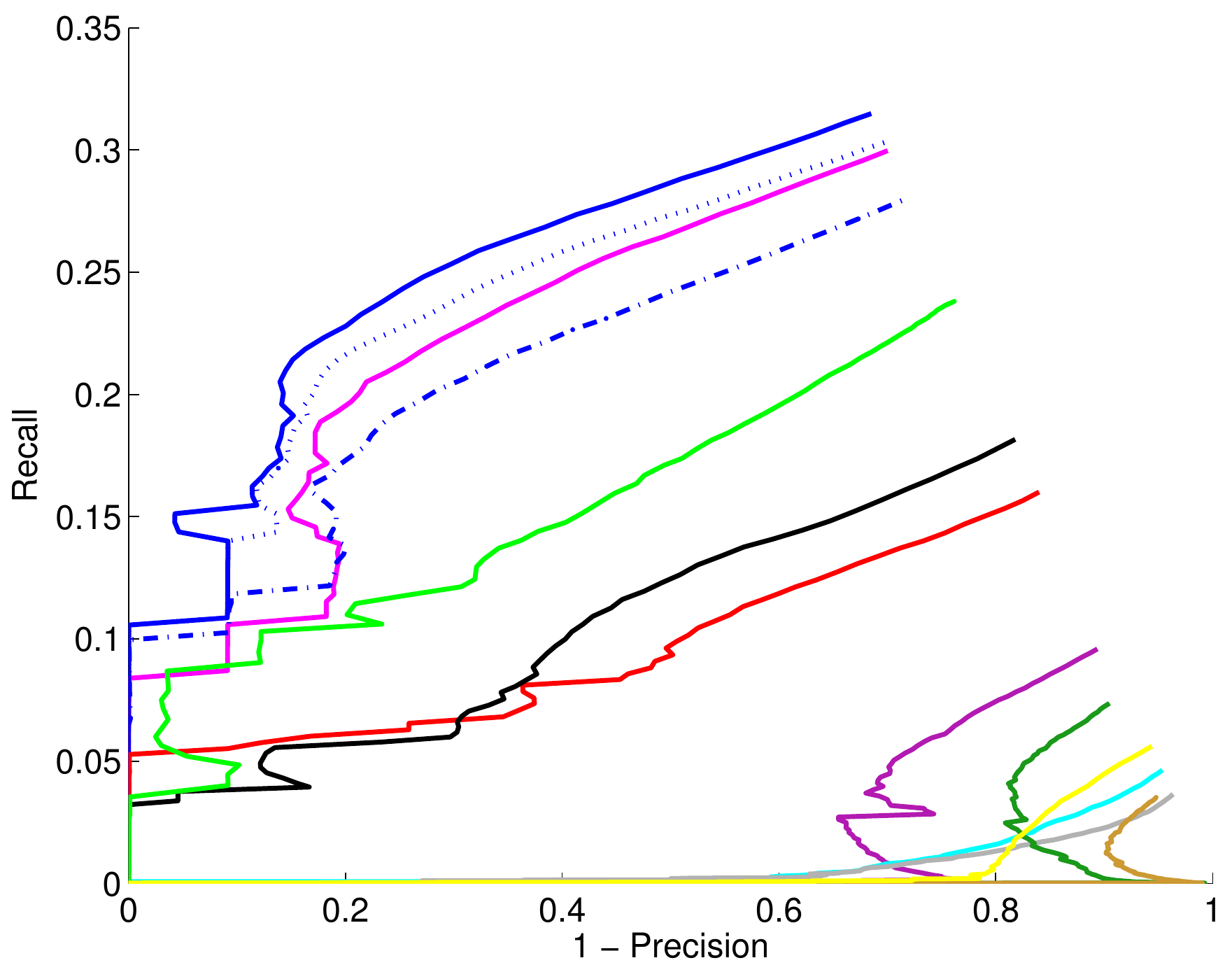}}
\end{subfloat}
\begin{subfloat}[\texttt{WSBS}]
{\includegraphics[width=\smallImgWidth\linewidth]{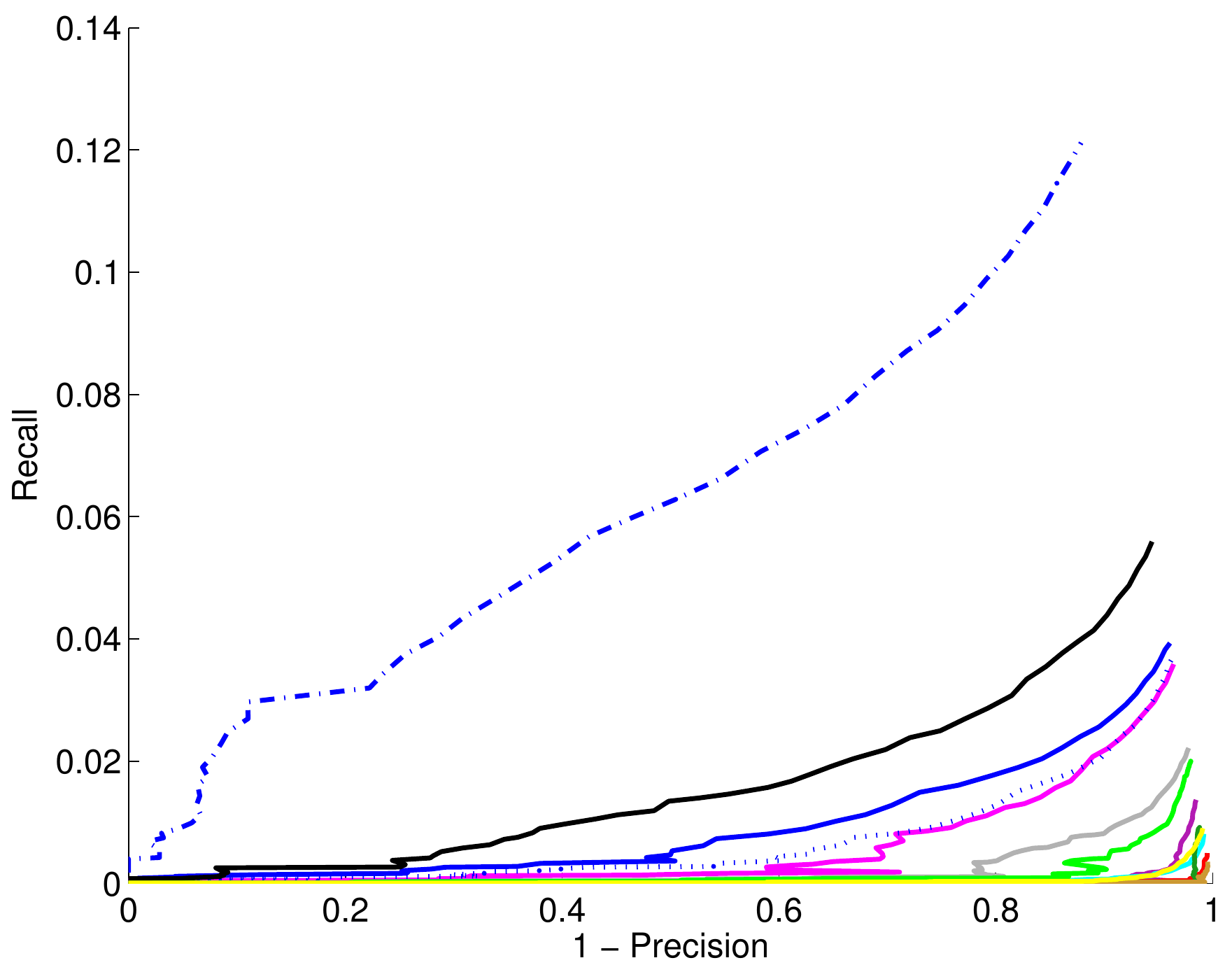}}
\end{subfloat}
\begin{subfloat}[\texttt{map2ph}]
{\includegraphics[width=\smallImgWidth\linewidth]{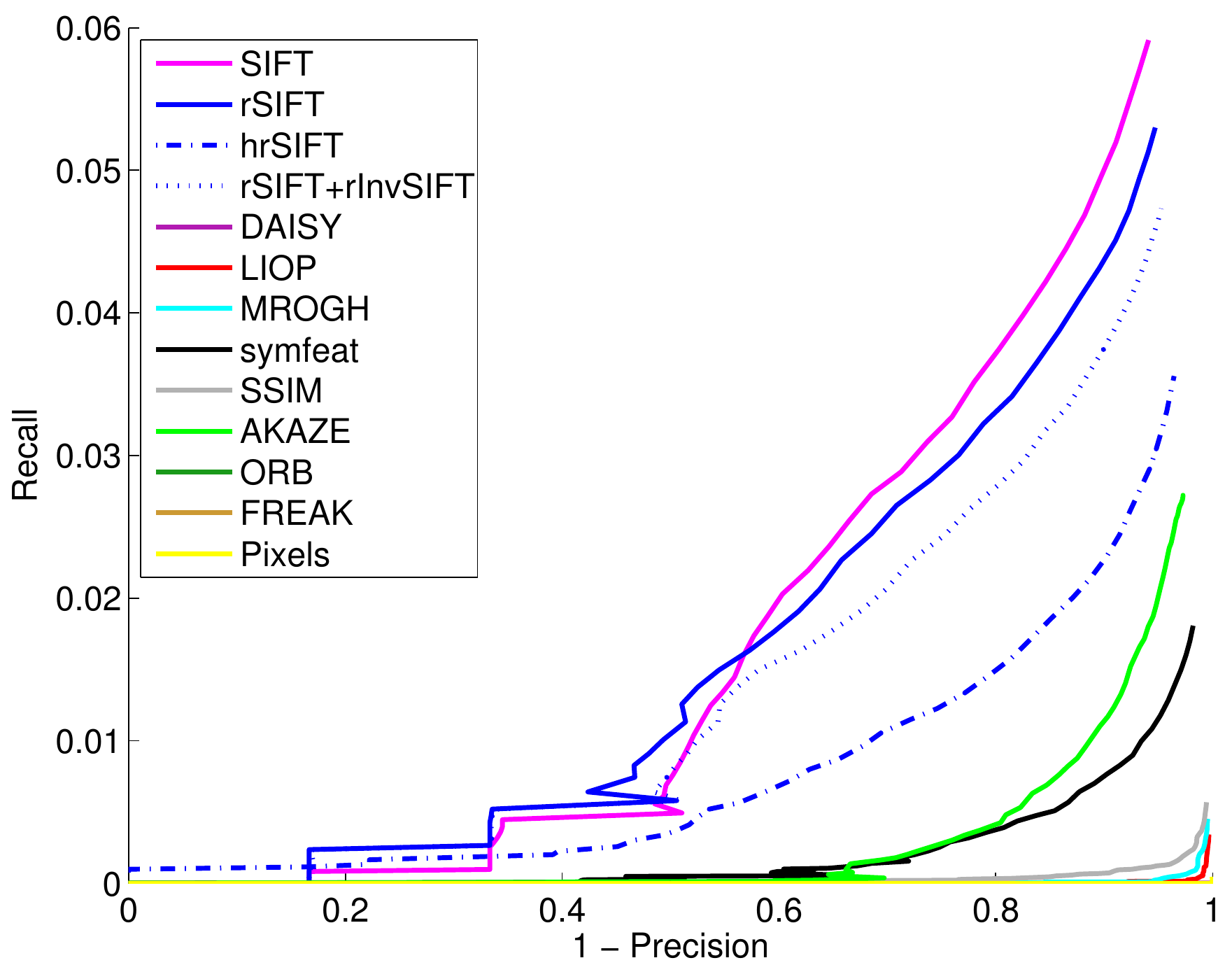}}
\end{subfloat}
\begin{subfloat}[\texttt{all}]
{\includegraphics[width=\smallImgWidth\linewidth]{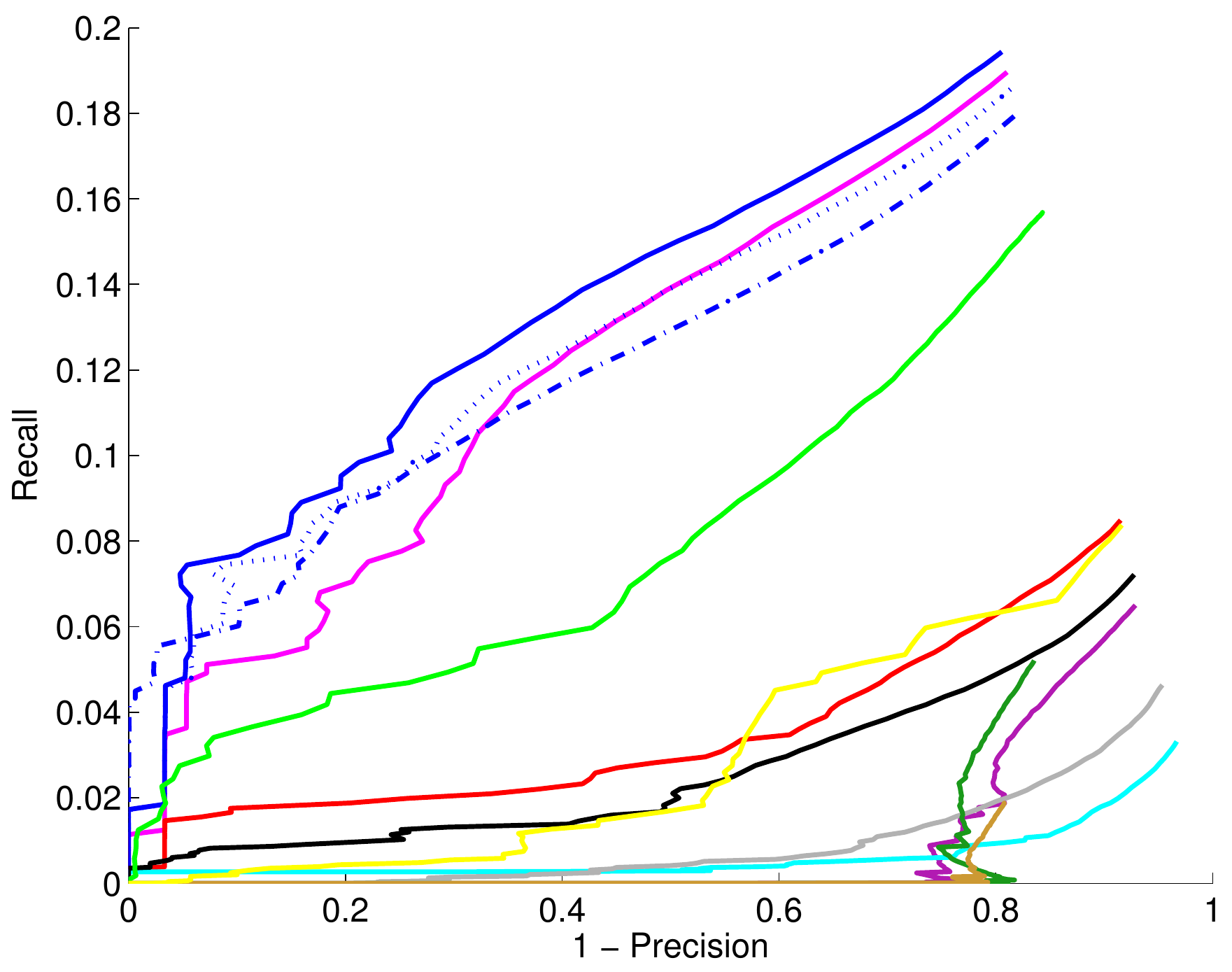}}
\end{subfloat}
\\
\begin{subfloat}[\texttt{WGBS}]
{\includegraphics[width=\smallImgWidth\linewidth]{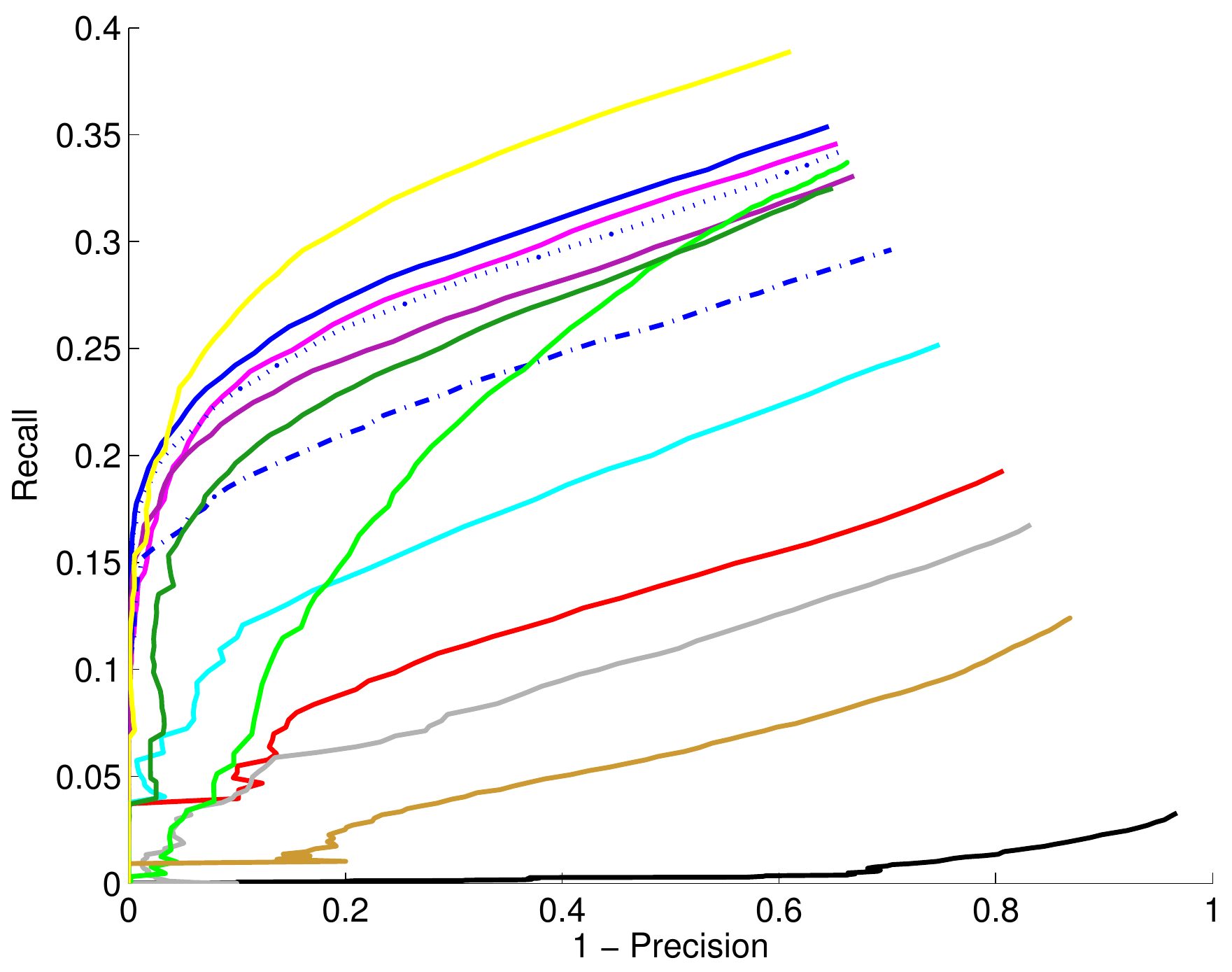}}
\end{subfloat}
\begin{subfloat}[\texttt{WABS}]
{\includegraphics[width=\smallImgWidth\linewidth]{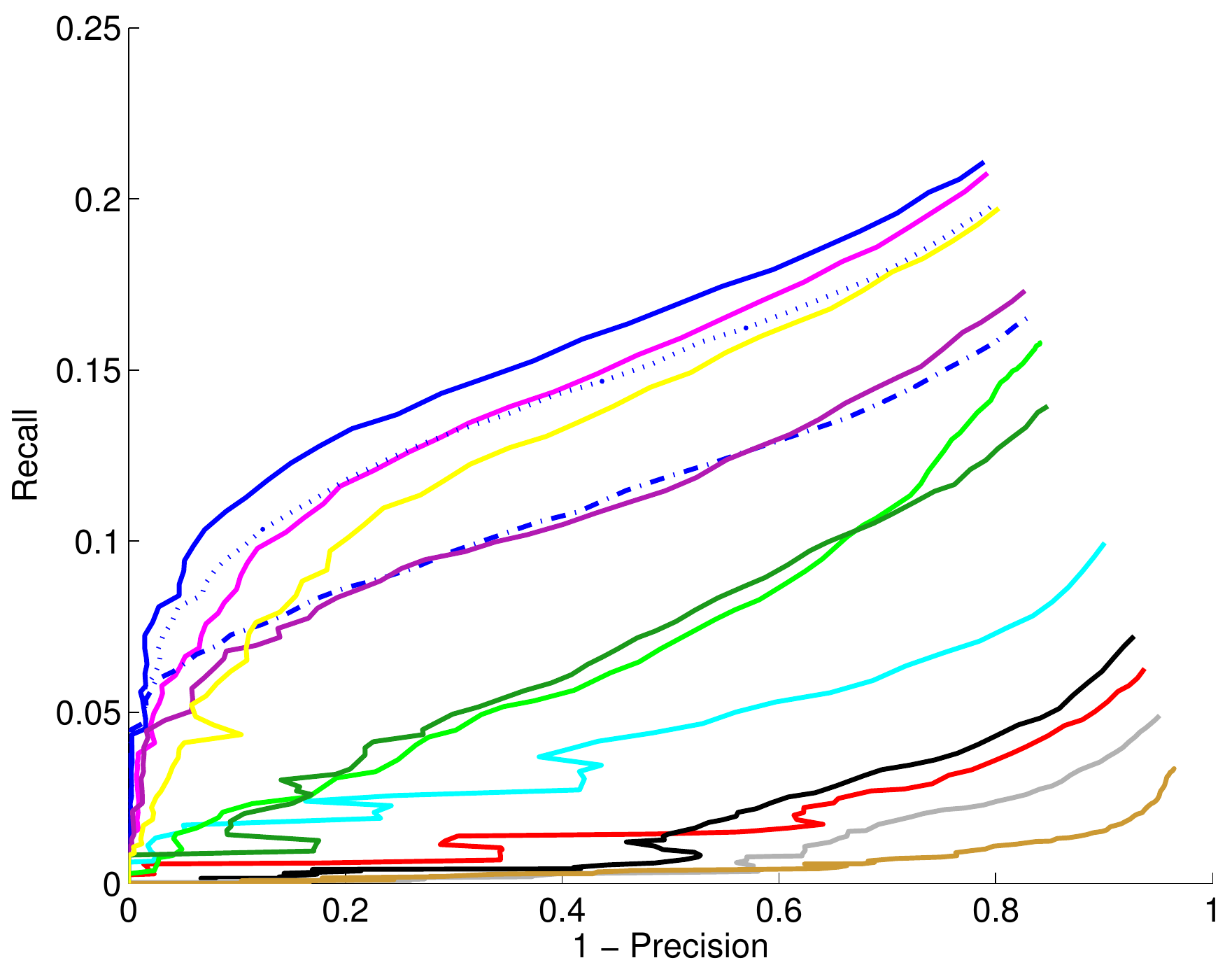}}
\end{subfloat}
\begin{subfloat}[\texttt{WLBS}]
{\includegraphics[width=\smallImgWidth\linewidth]{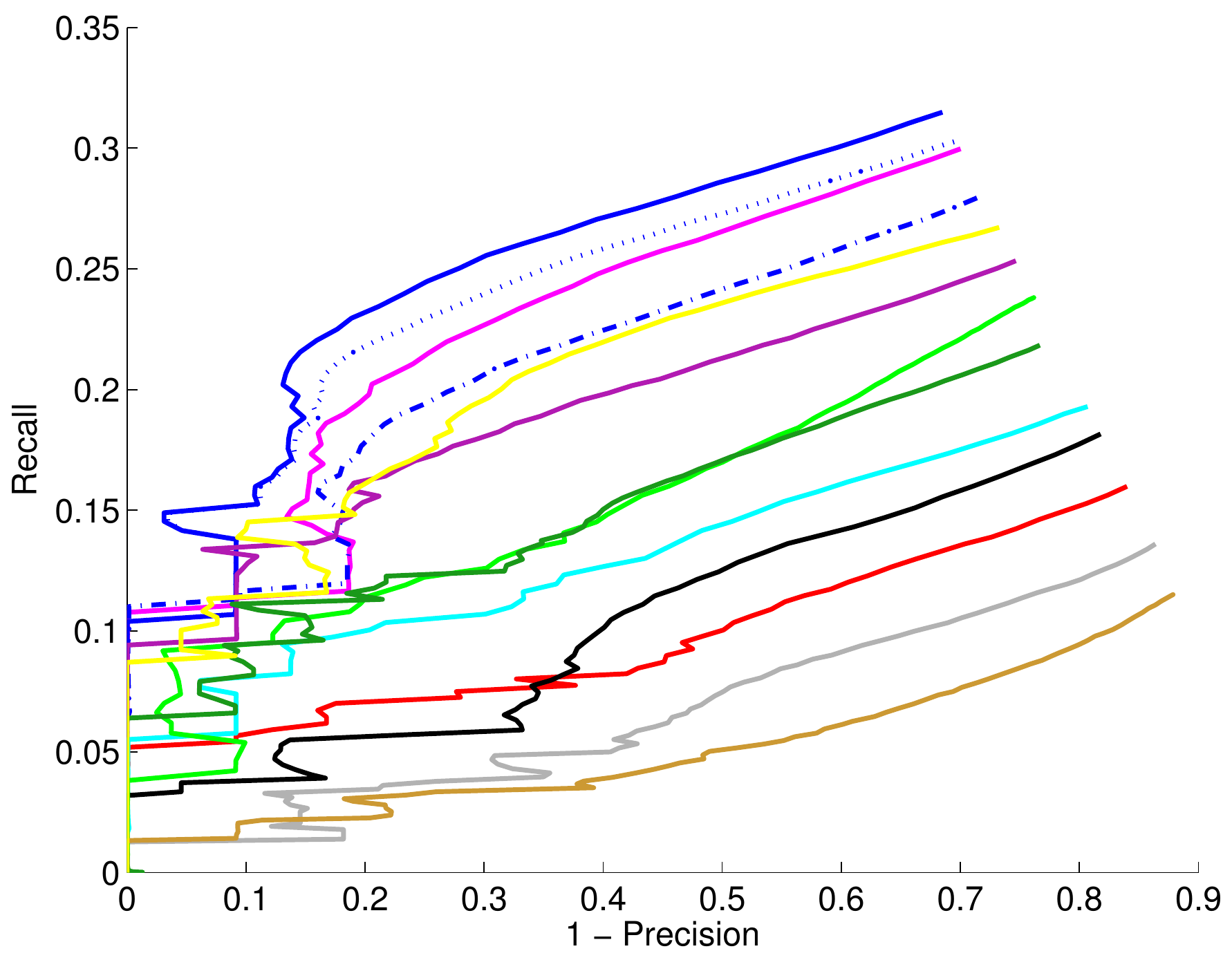}}
\end{subfloat}
\begin{subfloat}[\texttt{WSBS}]
{\includegraphics[width=\smallImgWidth\linewidth]{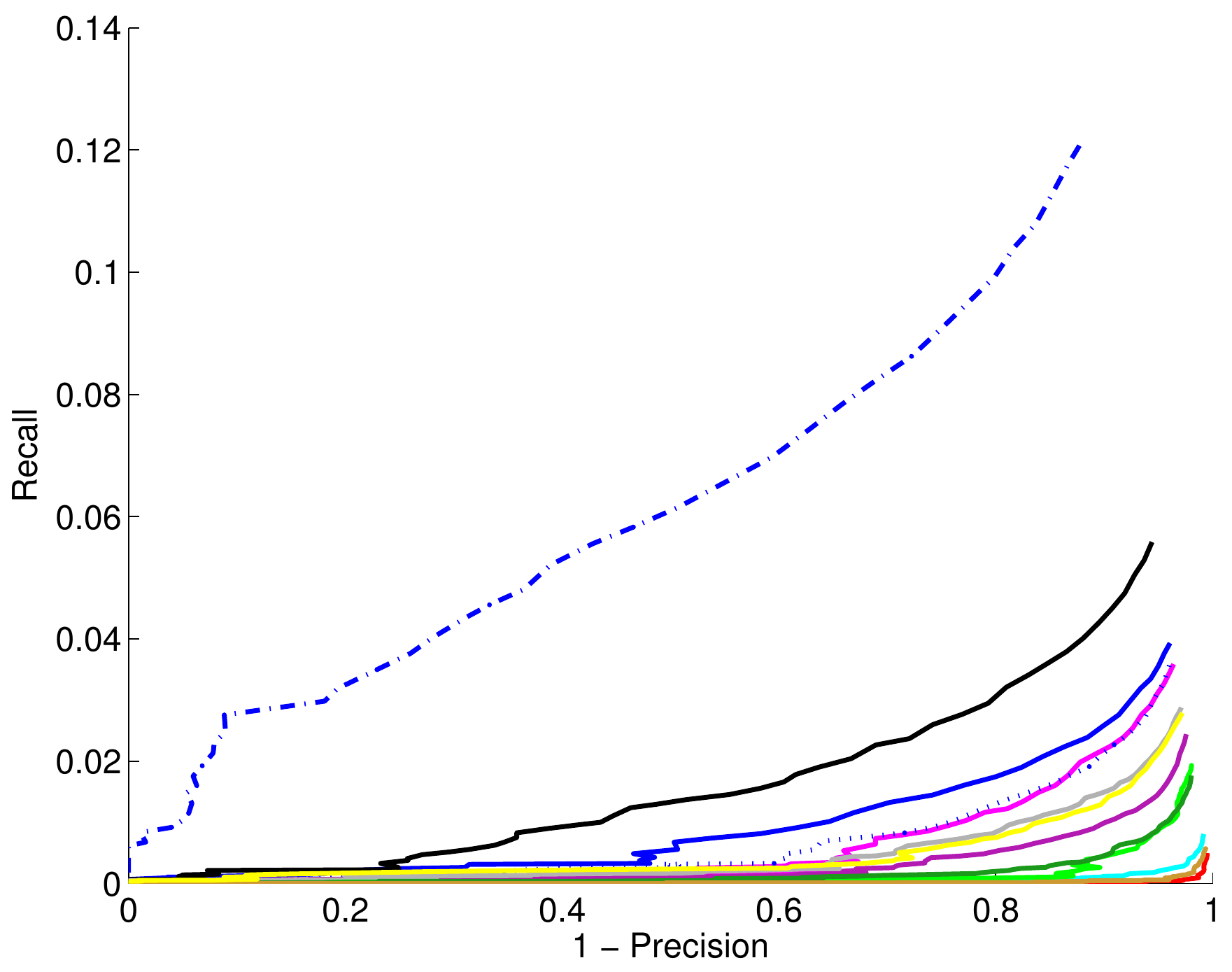}}
\end{subfloat}
\begin{subfloat}[\texttt{map2ph}]
{\includegraphics[width=\smallImgWidth\linewidth]{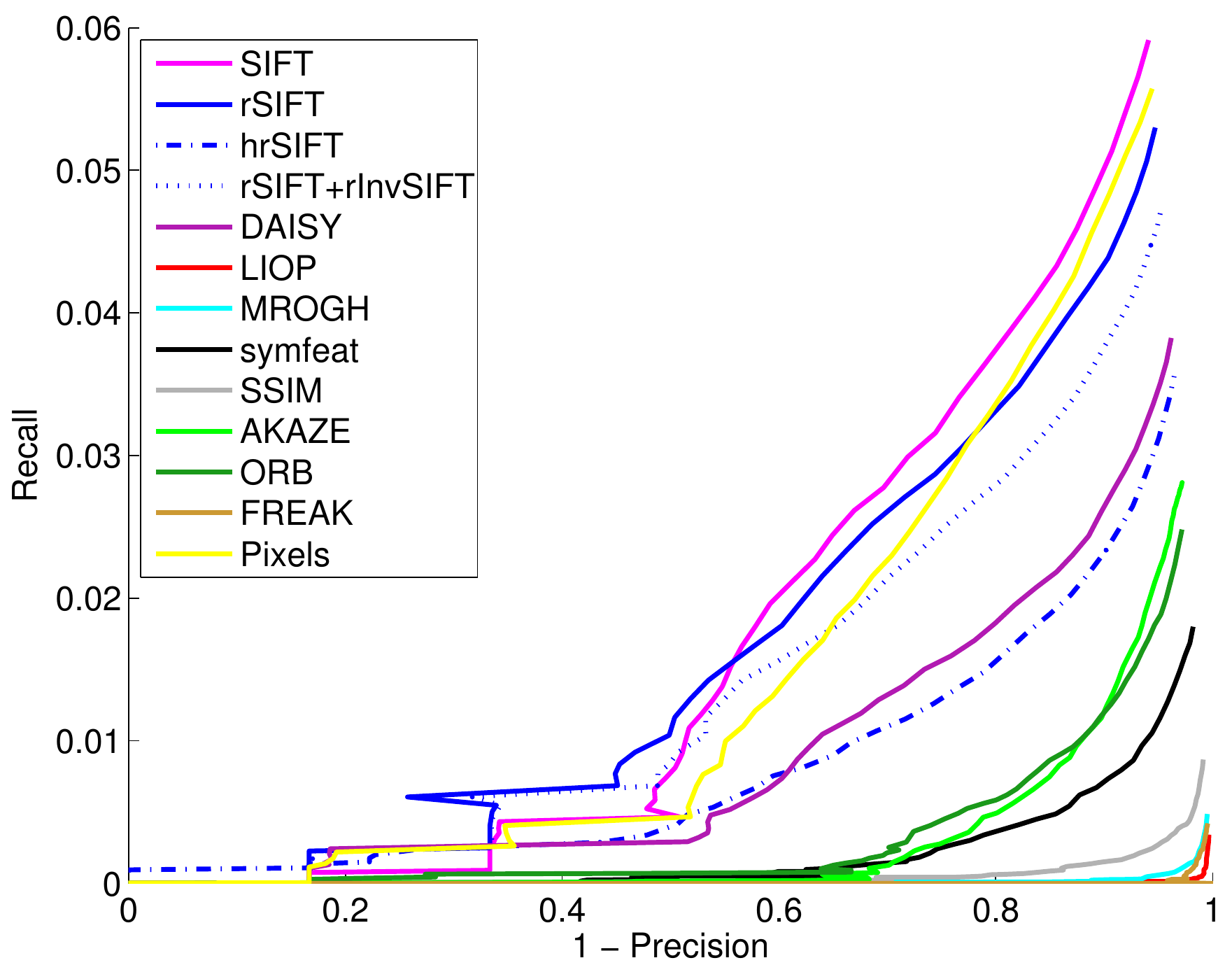}}
\end{subfloat}
\begin{subfloat}[\texttt{all}]
{\includegraphics[width=\smallImgWidth\linewidth]{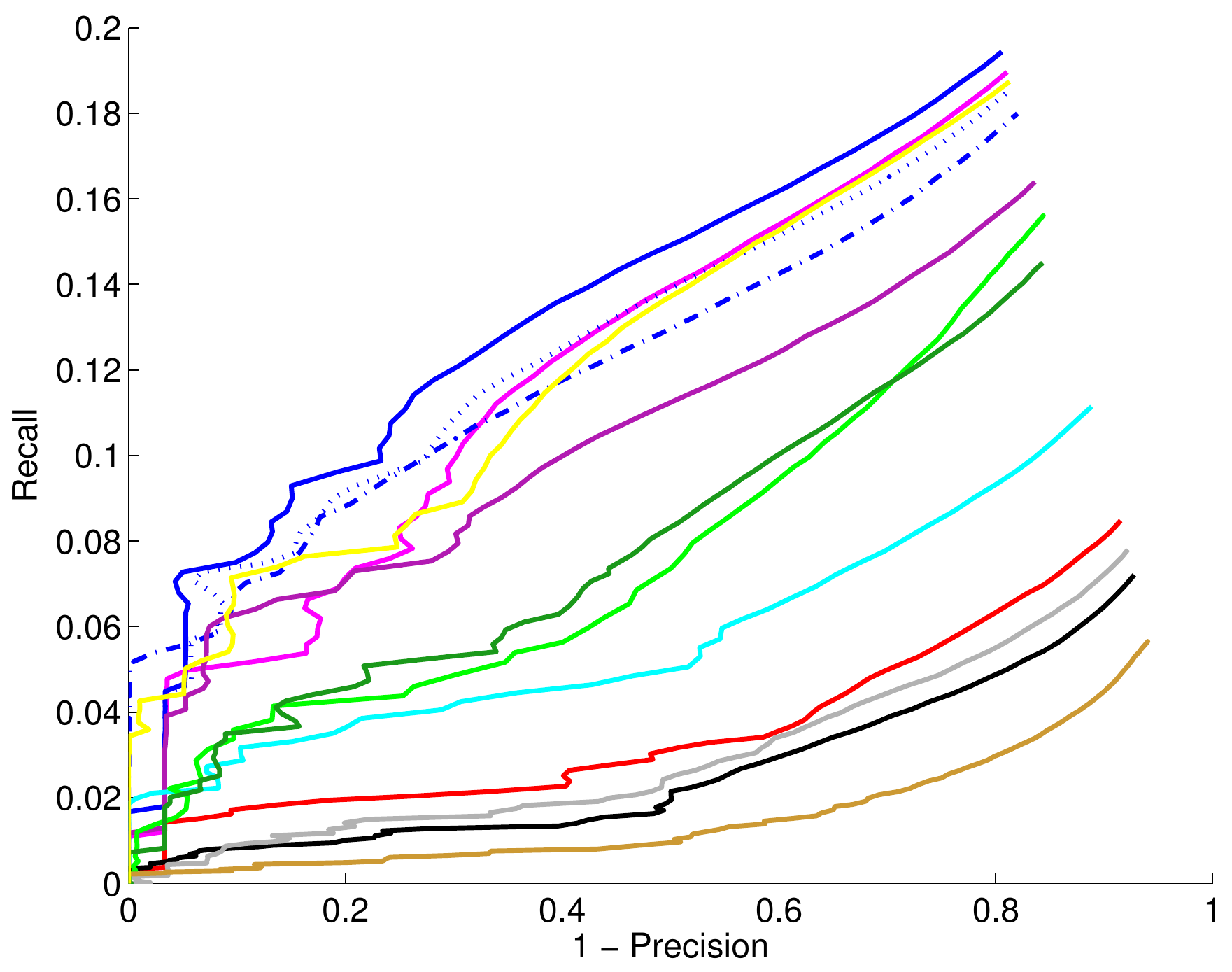}}
\end{subfloat}
\\

\begin{subfloat}[\texttt{WGBS}]
{\includegraphics[width=\smallImgWidth\linewidth]{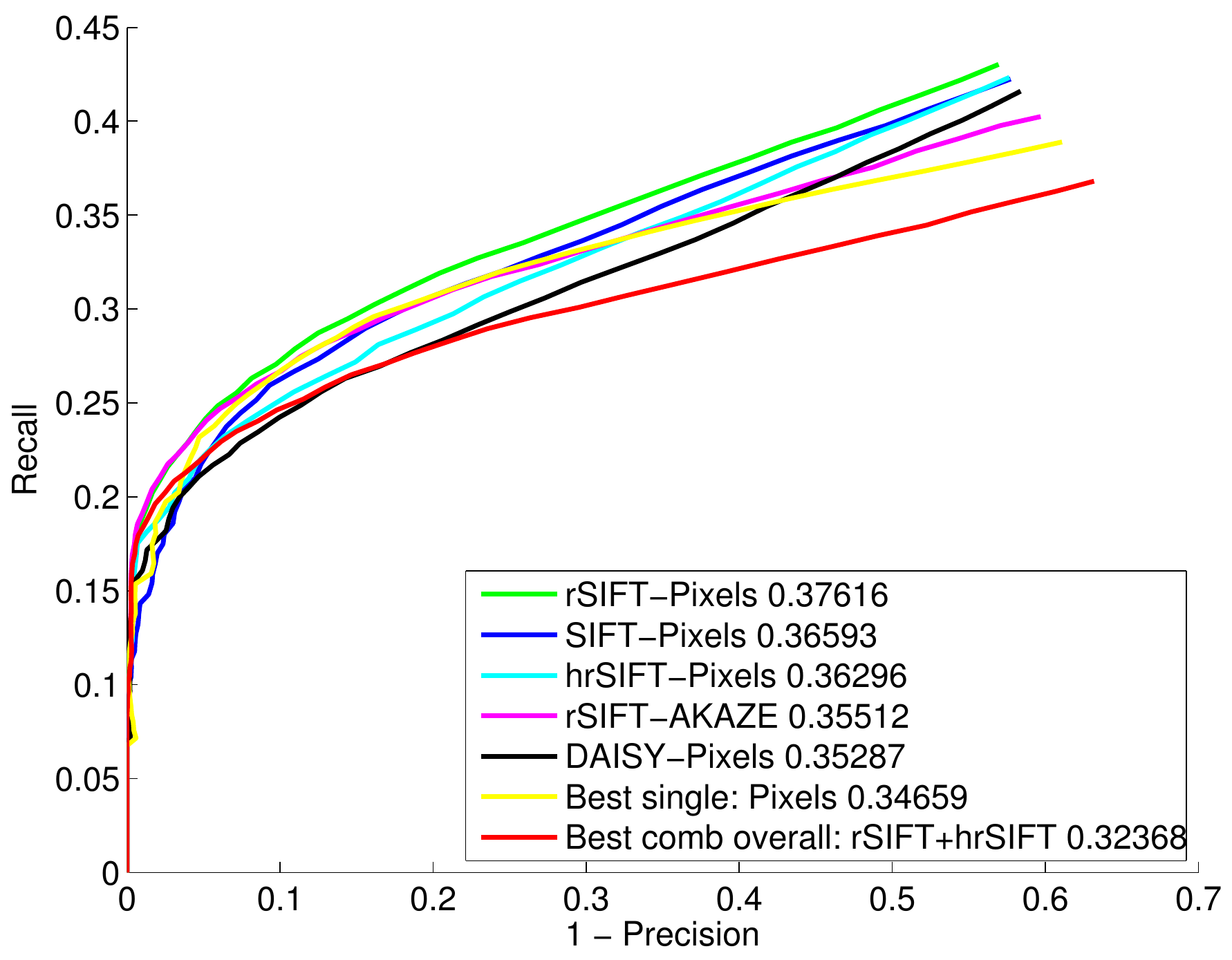}}
\end{subfloat}
\begin{subfloat}[\texttt{WABS}]
{\includegraphics[width=\smallImgWidth\linewidth]{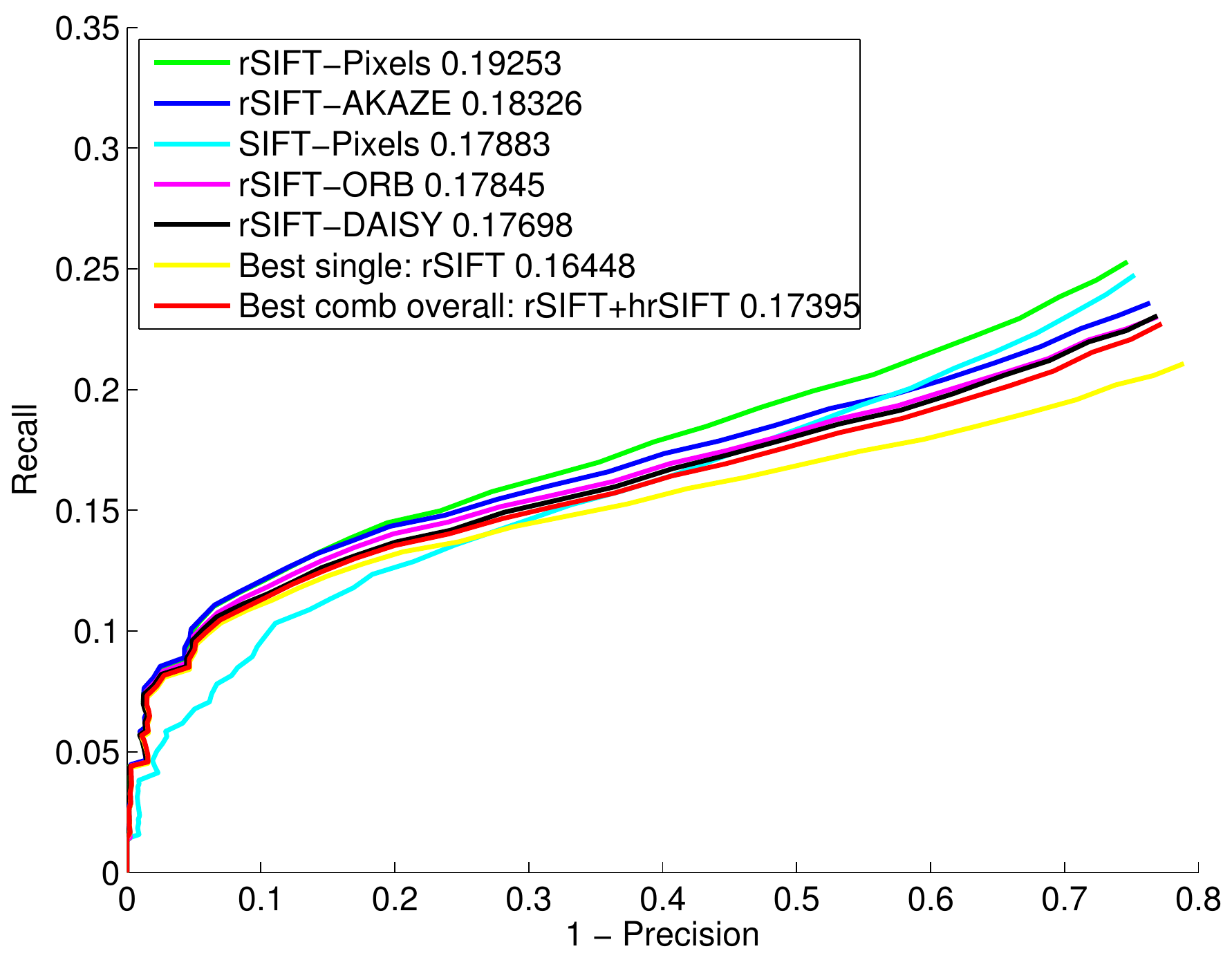}}
\end{subfloat}
\begin{subfloat}[\texttt{WLBS}]
{\includegraphics[width=\smallImgWidth\linewidth]{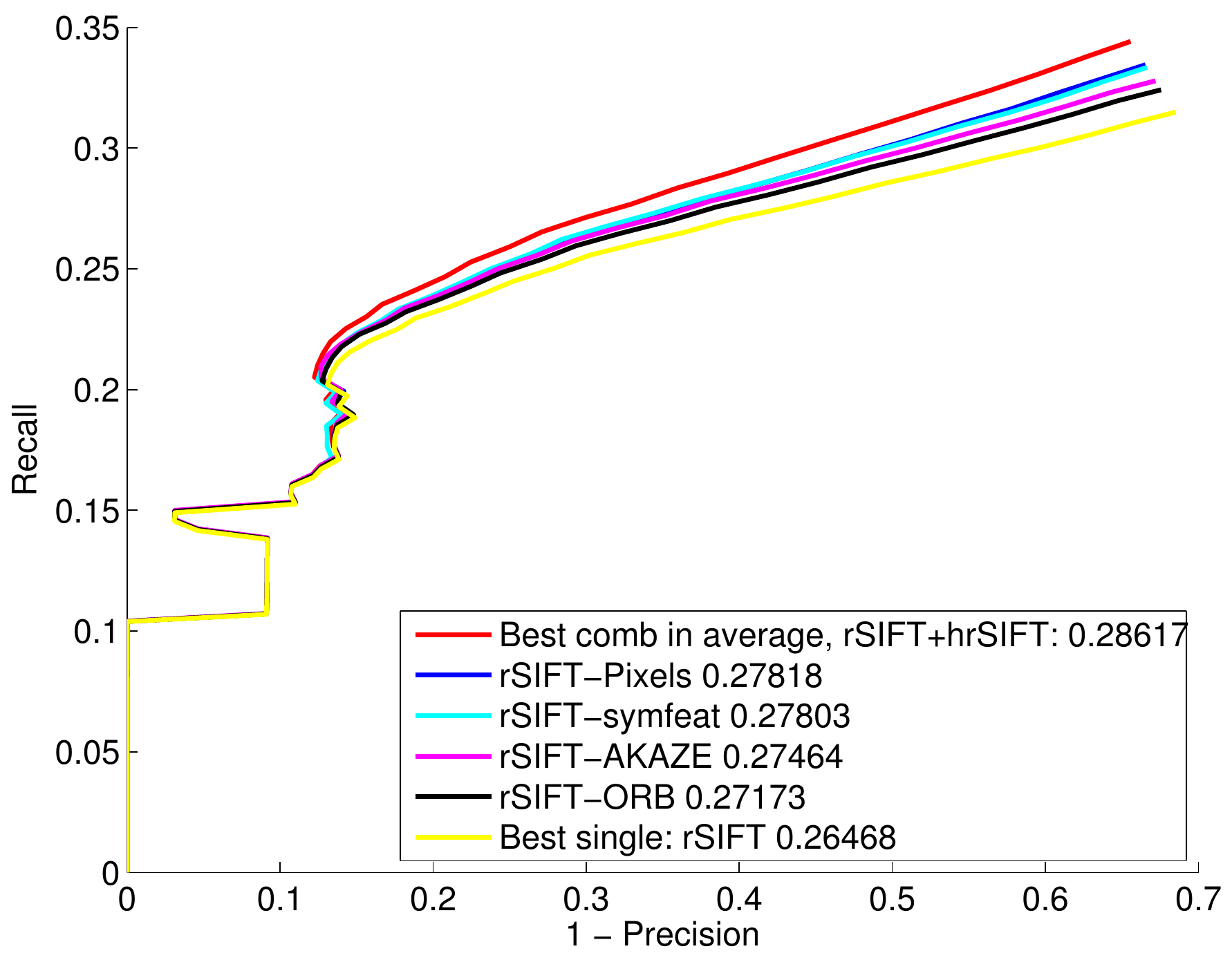}}
\end{subfloat}
\begin{subfloat}[\texttt{WSBS}]
{\includegraphics[width=\smallImgWidth\linewidth]{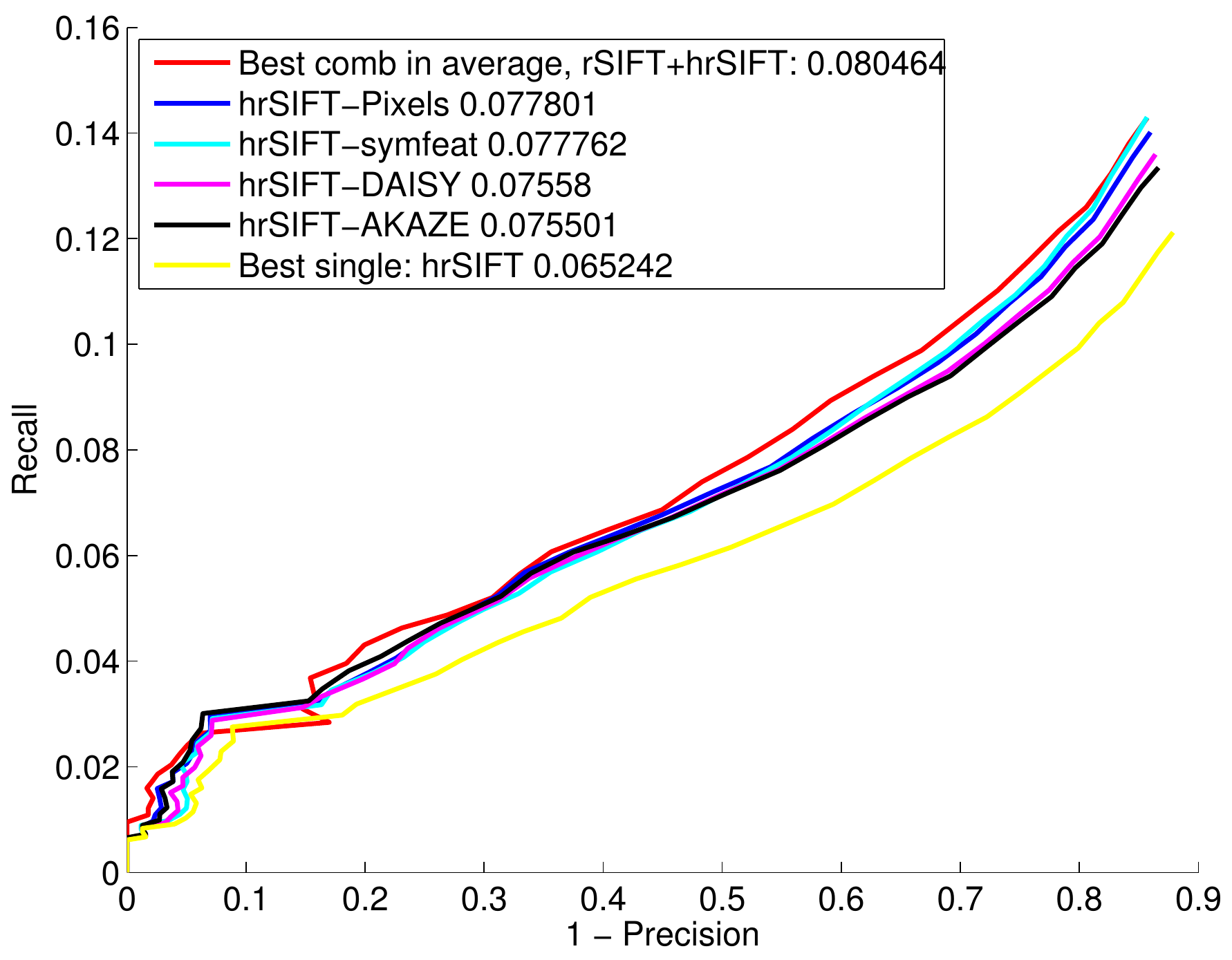}}
\end{subfloat}
\begin{subfloat}[\texttt{map2ph}]
{\includegraphics[width=\smallImgWidth\linewidth]{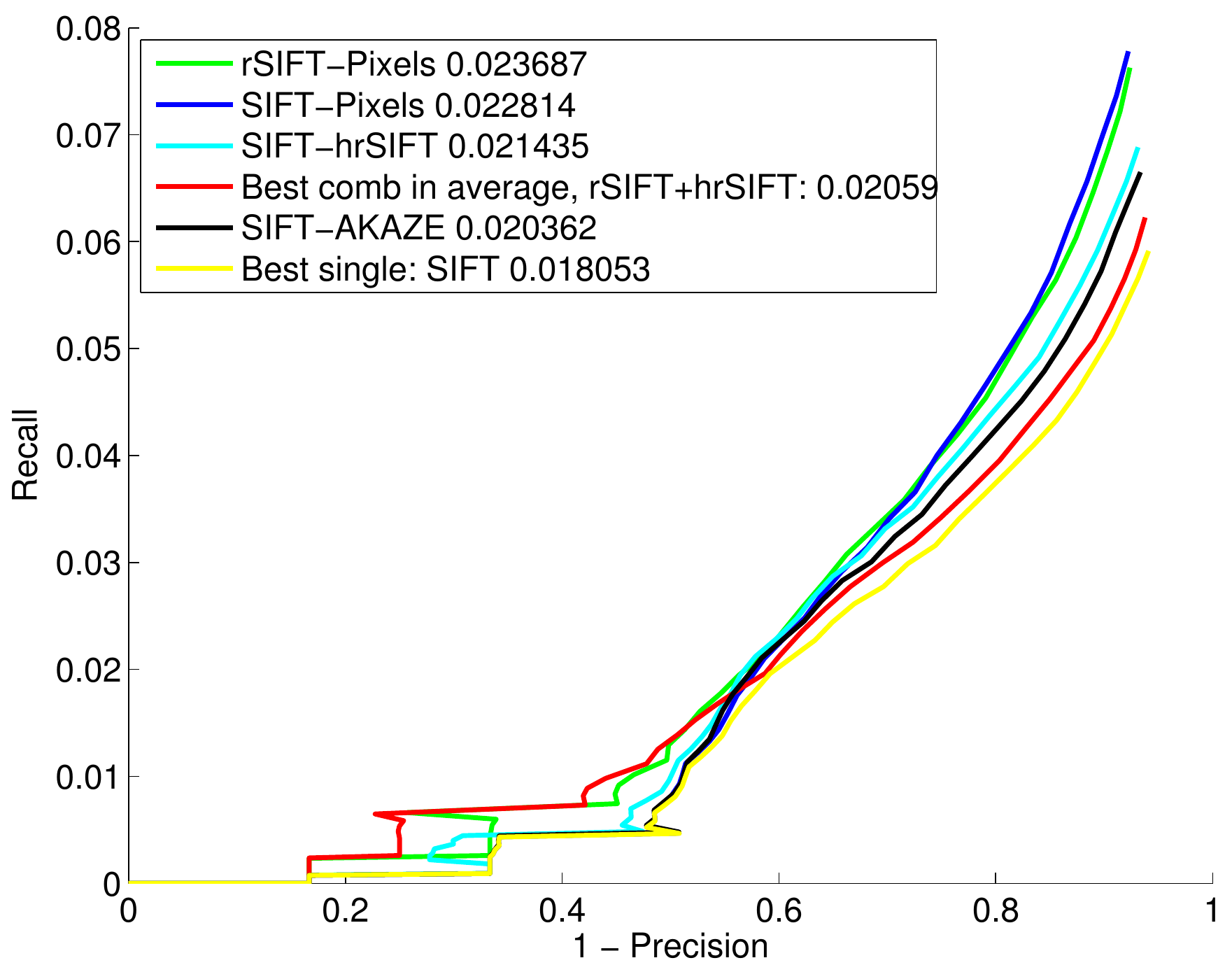}}
\end{subfloat}
\begin{subfloat}[\texttt{all}]
{\includegraphics[width=\smallImgWidth\linewidth]{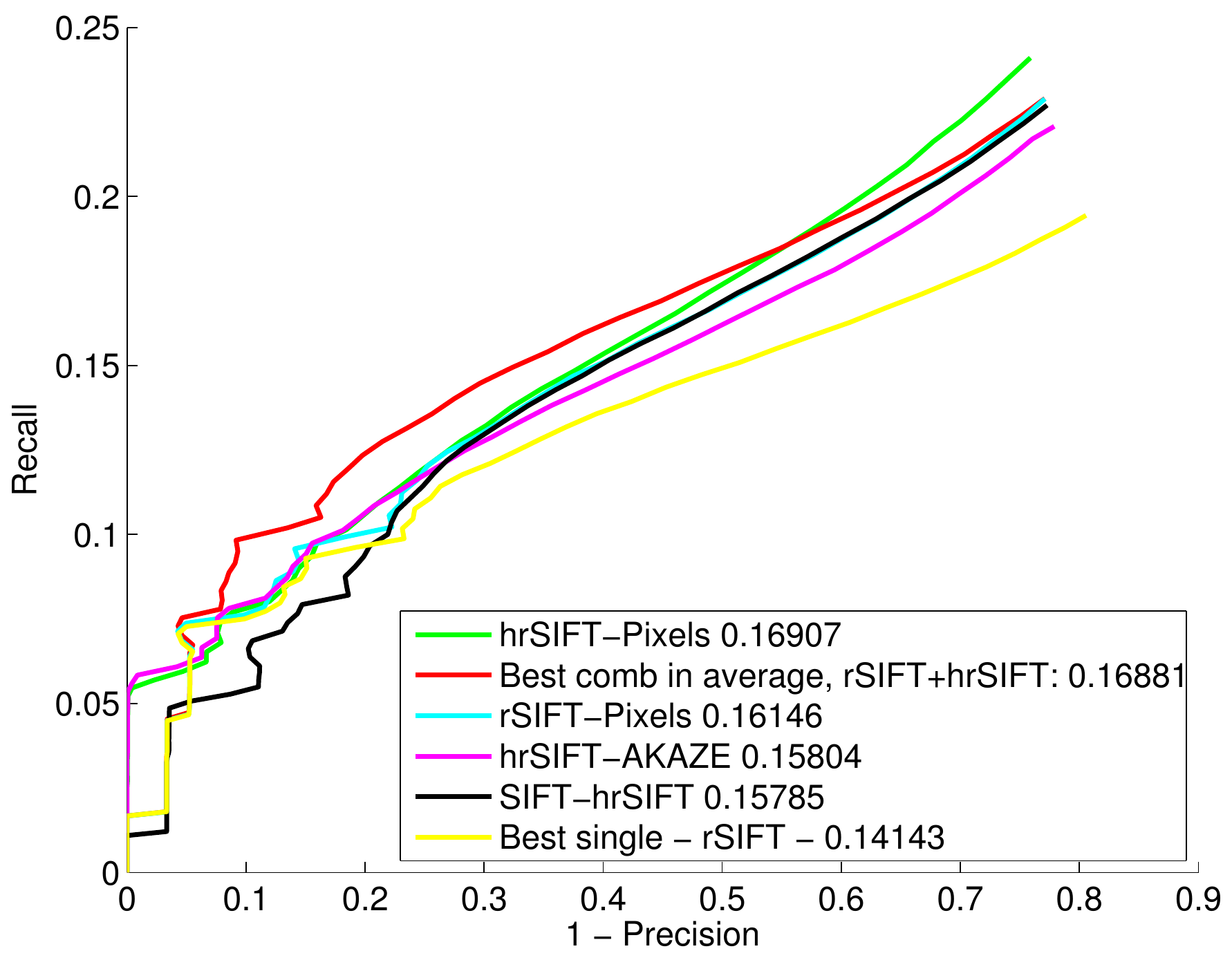}}
\end{subfloat}
\\

\def\svgwidth{1\linewidth}
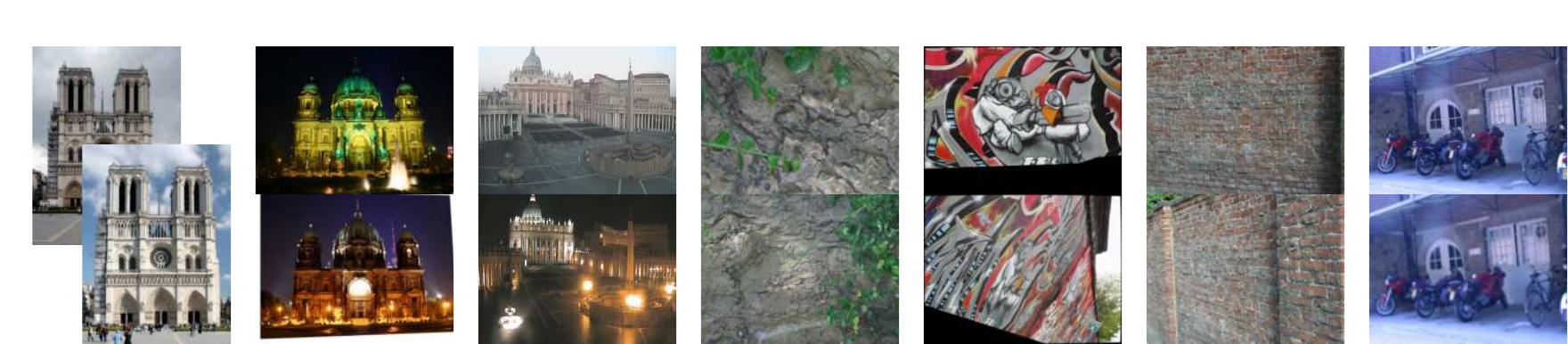
\caption{First row: descriptors computed using authors' implementation, second row - descriptors computed on photometrically normalized patches (mean = 0.5, var = 0.2) patches as done in SIFT. 
Third row: top 5 complementary pairs of descriptors (photometrically normalized). The numbers in legend are mean average precision. Bottom row: examples of the image pairs from each subset. Note that axis scales differs in each column, i.e. for different WxBS problems.}
\label{fig:desc_matching_perf}
\end{figure}
Floating point descriptors have been compared using $L_2$ distance, binary using Hamming distance. The Recall-Precision curves are shown in Figure~\ref{fig:desc_matching_perf}. The second nearest distance ratio is used to parameter the curve for floating point descriptors, the Hamming distance for binary ones. 

Note that most of the descriptors gain significantly from photometric normalization, cf. the first two rows of
Figure~\ref{fig:desc_matching_perf}. The published implementations are clearly sensitivity to contrast variations. 
    
The results show that gradient-histogram based SIFT and its variants including DAISY are the best performing descriptors by a big margin in the presence of any (geometric, illumination, etc) nuisance factors
despite the fact that some of the competitors -- LIOP, MROGH -- have been specifically designed to deal with illumination changes. The second best descriptor is -- surprisingly -- the patch with contrast-$L_2$-normalized  pixels, which beats all other descriptors. It has huge memory footprint -- 1681 floats, but the affine-photo-$L_2$-normed grayscale pixel intensities are a strong descriptor baseline.

Most of the descriptors, despite their different underlying assumptions and algorithmic structure, successfully match almost the same patches (see the third row in Figure~\ref{fig:desc_matching_perf}) -- and the most complementary descriptor to the leading rSIFT is its gradient-reversal-insensitive version -- hrSIFT.

The results confirming the domination of SIFT-based methods are in agreement with \cite{Stylianou2015} and \cite{LostInPast2015} despite the fact that they adopted a rather different evaluation methodology. However, we could not confirm the clear superiority of the SSIM over SymFeat descriptors, which could be explained by the fact that the SSIM descriptor was designed for use only with the SSIM detector.
\noindent

Tentative correspondences are generated using kD-tree~\cite{FLANN2009} and the 1st geometrically inconsistent rule with radius equal 10 pixels as a threshold is applied\cite{Mishkin2013}. Descriptors from different detector types (Hessian, MSER+, MSER-) as well as for different descriptors are put in seperate kD-trees. 
After matching, all tentative correspondences are put into a single list and duplicates, which appears due to view synthesis, are filtered if the features in both images are within a 3 pixel radius.

{\bf Detectors evaluation. } The following detectors are compared:  MSER~\cite{MSER2002}, DoG~\cite{SIFT2004}, Hessian-Affine~\cite{HesHarAff2004} (implementation~\cite{PerdochRetrieval2009}), FOCI~\cite{Zitnick2011}, IIDOG~\cite{Vonikakis2013}, WADE~\cite{Salti2013}, W$\alpha$SH~\cite{Varytimidis2012}, SURF~\cite{Bay2006}, SFOP~\cite{foerstner09}, AKAZE\cite{AKAZE2013}.
We focus on getting a reliable answer to the  "match/non-match" question in real image pairs. Therefore, the performance criterion is the number of successfully matched pairs using the best combination of descriptors (see Section {\bf Descriptor evaluation} ) -- ~rSIFT and hrSIFT. %
Image pairs are considered matched if $\geq$15 correct inliers to a homography are found. Since the Lost-in-past dataset contains 2300 matchable image pairs, which is unfeasible for direct matching, we have selected a subset of 172 medium-challenging image pairs. Other datasets are used fully.  

\noindent
{\bf Adaptive threshold of the detector response.}
One of the main problems in the matching of day to night and infrared images is the low number of detected features. The problem is acute in dark low contrast images  in the \textsc{WgsBS} and MMS~\cite{Aguilera2012} datasets. 
A possible approach addressing the problem is iiDoG~\cite{Vonikakis2013} where the difference of Gaussians is normalized by the sum of Gaussians. It works well, but cannot be easily applied for other types of detectors, e.g., MSER. 

\begin{table}[tb]
\centering
\caption{Detector evaluation results. The number of matched image pairs (left) and the average running time (right). The FOCI detector is run through MS Windows simulator wine, the time includes a big overhead.}
\label{tab:detector-results}
\scriptsize
\setlength{\tabcolsep}{.25em}
\vspace{0.5em}
\begin{tabular}{lrrrrrrrrrrrrrrrrrrrrrrrr}
\toprule
Alg.
& \multicolumn{2}{c}{EF}
& \multicolumn{2}{c}{EVD}
& \multicolumn{2}{c}{MMS}
& \multicolumn{2}{c}{\textsc{WgaBS}}
& \multicolumn{2}{c}{\textsc{WgalBS}}
& \multicolumn{2}{c}{\textsc{WglBS}}
& \multicolumn{2}{c}{\textsc{WgsBS}}
& \multicolumn{2}{c}{\textsc{WlaBS}}
& \multicolumn{2}{c}{Past}
& \multicolumn{2}{c}{OxAff}
& \multicolumn{2}{c}{SymB}
& \multicolumn{2}{c}{GDB}
\\
 & \shortstack{\# \\33}& \shortstack{time \\ $[s]$}
& \shortstack{\# \\ 15}& \shortstack{time \\ $[s]$}
& \shortstack{\# \\100}& \shortstack{time \\ $[s]$}
& \shortstack{\# \\ 5}& \shortstack{time \\ $[s]$}
& \shortstack{\# \\8}& \shortstack{time \\ $[s]$}
& \shortstack{\# \\ 9}& \shortstack{time \\ $[s]$}
& \shortstack{\# \\ 5}& \shortstack{time \\ $[s]$}
& \shortstack{\#\\ 4}& \shortstack{time \\ $[s]$}
& \shortstack{\#\\172}& \shortstack{time \\ $[s]$}
& \shortstack{\#\\40}& \shortstack{time \\ $[s]$}
& \shortstack{\#\\46}& \shortstack{time \\ $[s]$}
& \shortstack{\# \\22}& \shortstack{time \\ $[s]$}
\\
\midrule
\multicolumn{25}{c}{Threshold adaptation}\\
\midrule
MSER & \ccaEF{16} & 1.4 & \ccaEVD{3} & 1.4 & \ccaMMS{1} & 0.3 & \ccaWGABS{0} & 2.0 & \ccaWGALBS{0} & 1.3 & \ccaWGLBS{0} & 1.3 & \ccaWGSBS{0} & 0.8 & \ccaWLABS{1} & 1.2 & \ccalostinpast{8} & 1.3 & \ccaoxford{40} & 3.5 & \ccasnavely{23} & 2.4 & \ccastewart{9} & 2.4\\
AdMSER & \ccaEF{25} & 3.4 & \ccaEVD{8} & 4.0 & \ccaMMS{6} & 1.0 & \ccaWGABS{0} & 4.0 & \ccaWGALBS{0} & 3.2 & \ccaWGLBS{0} & 3.3 & \ccaWGSBS{0} & 1.4 & \ccaWLABS{1} & 2.6 & \ccalostinpast{11} & 2.9 & \ccaoxford{40} & 5.7 & \ccasnavely{26} & 4.6 & \ccastewart{13} & 6.9\\
DoG & \ccaEF{29} & 2.3 & \ccaEVD{0} & 2.8 & \ccaMMS{10} & 0.8 & \ccaWGABS{0} & 2.7 & \ccaWGALBS{0} & 2.3 & \ccaWGLBS{0} & 2.1 & \ccaWGSBS{0} & 1.0 & \ccaWLABS{1} & 2.4 & \ccalostinpast{13} & 2.0 & \ccaoxford{38} & 4.8 & \ccasnavely{29} & 2.7 & \ccastewart{12} & 4.7\\
iiDoG & \ccaEF{29} & 3.1 & \ccaEVD{0} & 3.0 & \ccaMMS{11} & 1.2 & \ccaWGABS{0} & 3.2 & \ccaWGALBS{0} & 2.9 & \ccaWGLBS{0} & 2.8 & \ccaWGSBS{0} & 1.2 & \ccaWLABS{1} & 2.5 & \ccalostinpast{13} & 2.2 & \ccaoxford{38} & 8.0 & \ccasnavely{29} & 2.9 & \ccastewart{12} & 6.1\\
AdDoG & \ccaEF{29} & 2.6 & \ccaEVD{0} & 3.4 & \ccaMMS{11} & 1.2 & \ccaWGABS{0} & 3.3 & \ccaWGALBS{0} & 3.0 & \ccaWGLBS{0} & 3.0 & \ccaWGSBS{0} & 1.5 & \ccaWLABS{1} & 2.7 & \ccalostinpast{13} & 2.7 & \ccaoxford{38} & 4.1 & \ccasnavely{30} & 3.0 & \ccastewart{12} & 4.8\\
HesAf & \ccaEF{32} & 4.6 & \ccaEVD{1} & 5.2 & \ccaMMS{15} & 1.2 & \ccaWGABS{0} & 5.5 & \ccaWGALBS{0} & 3.8 & \ccaWGLBS{0} & 4.2 & \ccaWGSBS{0} & 2.0 & \ccaWLABS{1} & 3.6 & \ccalostinpast{24} & 4.0 & \ccaoxford{40} & 11. & \ccasnavely{35} & 5.8 & \ccastewart{17} & 9.1\\
AdHesAf & \ccaEF{33} & 5.7 & \ccaEVD{2} & 7.6 & \ccaMMS{35} & 2.9 & \ccaWGABS{0} & 7.2 & \ccaWGALBS{1} & 6.5 & \ccaWGLBS{0} & 6.0 & \ccaWGSBS{0} & 3.2 & \ccaWLABS{1} & 4.9 & \ccalostinpast{25} & 5.4 & \ccaoxford{40} & 10. & \ccasnavely{35} & 7.2 & \ccastewart{18} & 13.\\
\hline
\multicolumn{25}{c}{Other detectors}\\
\hline
W$\alpha$SH & \ccaEF{0} & 1.8 & \ccaEVD{0} & 5.4 & \ccaMMS{0} & 0.6 & \ccaWGABS{0} & 2.8 & \ccaWGALBS{0} & 2.5 & \ccaWGLBS{0} & 1.4 & \ccaWGSBS{0} & 1.8 & \ccaWLABS{0} & 1.2 & \ccalostinpast{0} & 1.9 & \ccaoxford{24} & 4.1 & \ccasnavely{3} & 2.8 & \ccastewart{3} & 6.9\\
ORB & \ccaEF{3} & 4.1 & \ccaEVD{0} & 3.6 & \ccaMMS{1} & 0.8 & \ccaWGABS{0} & 2.8 & \ccaWGALBS{0} & 2.7 & \ccaWGLBS{0} & 3.6 & \ccaWGSBS{0} & 1.6 & \ccaWLABS{0} & 2.8 & \ccalostinpast{1} & 2.3 & \ccaoxford{28} & 8.7 & \ccasnavely{5} & 3.0 & \ccastewart{3} & 6.1\\
SURF & \ccaEF{27} & 2.3 & \ccaEVD{0} & 2.4 & \ccaMMS{7} & 1.0 & \ccaWGABS{0} & 2.5 & \ccaWGALBS{0} & 1.9 & \ccaWGLBS{0} & 2.1 & \ccaWGSBS{0} & 0.9 & \ccaWLABS{1} & 1.4 & \ccalostinpast{10} & 1.9 & \ccaoxford{38} & 5.8 & \ccasnavely{31} & 2.9 & \ccastewart{15} & 4.0\\
AKAZE & \ccaEF{28} & 4.3 & \ccaEVD{0} & 3.6 & \ccaMMS{10} & 0.8 & \ccaWGABS{1} & 4.7 & \ccaWGALBS{0} & 3.4 & \ccaWGLBS{0} & 4.0 & \ccaWGSBS{0} & 1.3 & \ccaWLABS{1} & 2.7 & \ccalostinpast{25} & 3.6 & \ccaoxford{38} & 13. & \ccasnavely{35} & 5.6 & \ccastewart{17} & 6.4\\
FOCI & \ccaEF{29} & 12. & \ccaEVD{0} & 39. & \ccaMMS{14} & 11. & \ccaWGABS{1} & 32. & \ccaWGALBS{0} & 29. & \ccaWGLBS{0} & 29. & \ccaWGSBS{0} & 20. & \ccaWLABS{1} & 29. & \ccalostinpast{21} & 13. & \ccaoxford{38} & 35. & \ccasnavely{35} & 27. & \ccastewart{17} & 45.\\
SFOP & \ccaEF{25} & 11. & \ccaEVD{0} & 16. & \ccaMMS{12} & 4.7 & \ccaWGABS{0} & 12. & \ccaWGALBS{0} & 10. & \ccaWGLBS{0} & 10. & \ccaWGSBS{0} & 9.2 & \ccaWLABS{0} & 7.5 & \ccalostinpast{11} & 12. & \ccaoxford{36} & 15. & \ccasnavely{24} & 11. & \ccastewart{8} & 17.\\
WADE & \ccaEF{16} & 14. & \ccaEVD{0} & 20. & \ccaMMS{0} & 3.4 & \ccaWGABS{0} & 58. & \ccaWGALBS{0} & 11. & \ccaWGLBS{0} & 14. & \ccaWGSBS{0} & 7.9 & \ccaWLABS{1} & 8.3 & \ccalostinpast{20} & 23. & \ccaoxford{34} & 60. & \ccasnavely{34} & 46. & \ccastewart{13} & 77.\\
\bottomrule

\end{tabular}
\end{table}

Instead, we propose to use the following adaptive thresholding for all feature detectors. 
First, all local extrema of the response function are detected (i.e., no thresholding occurs). Next, the detected features are sorted according to the response magnitude. If the number of detected features with response magnitude $\geq$ $\Theta$ is greater than a given threshold $R_{\text{min}}$, these are  output and the algorithm terminates (this is the standard approach).
If there are not enough features above the threshold, top $R_{\text{min}}$ features are outputed.

The results are summarized in Table~\ref{tab:detector-results}. Note that none of the methods was able to match almost any image pair which combines more nuisance factors. %

Results in Table~\ref{tab:detector-results} confirm that the proposed adaptive thresholding strategy works as well
as, or even better, than iiDoG for DoG, but it is ~1.5 times faster. It also significantly improves the results of the MSER and Hessian-Affine, even when the main nuisance is in the viewing geometry (EVD dataset).

\chapter[HardNet local feature descriptor]{HardNet local feature descriptor \\
\includegraphics[width=0.95\linewidth]{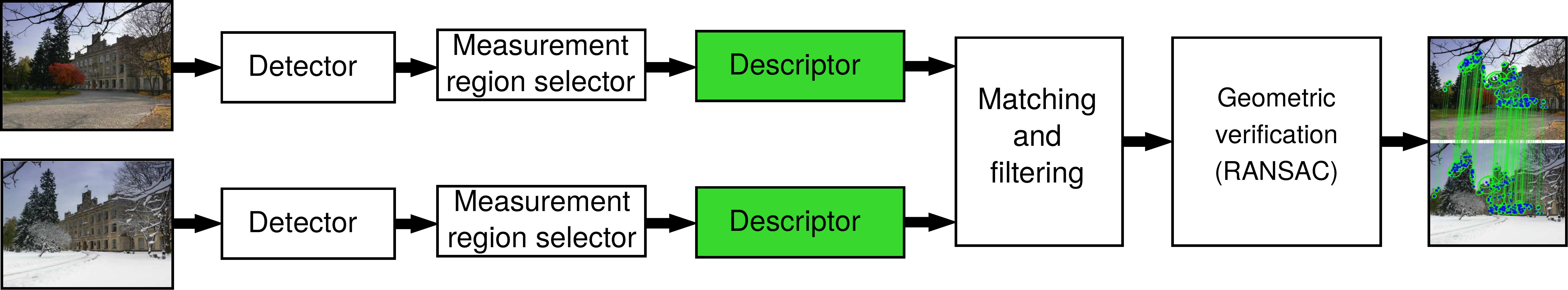}}
\label{ch:hardnet}

In this Chapter we focus on descriptor learning and, using a novel method, train a convolutional neural network (CNN), called HardNet. We additionally show that our learned descriptor outperforms both hand-crafted and learned descriptors in real-world tasks like image retrieval and two-view matching under extreme conditions. We explore possible architectural, training and data choices in Section~\ref{hardnet:pultar} further improving HardNet descriptor.

\section{Related work}
\label{hardnet:relatedwork}

\paragraph{Classical local feature descriptors.}
A local feature descriptor is a mapping from a (small) image to a metric space, which is robust to changes in acquisition conditions so that descriptors related to the same 3D points are similar and dissimilar otherwise~\cite{Pultar2020improving}.

Early local descriptors include color pixel histograms\cite{Swain1991}, eigenvalues of the patch~\cite{Turk1991}, grayscale histograms~\cite{Schiele1996}, later replaced by grayscale invariants~\cite{Schmid1997} which were inspired by geometry representation in biological systems~\cite{Koenderink1987}. 

The next big step in local patch description was the SIFT~\cite{SIFT2004} -- a histogram of gradient orientations. Most of the hand-crafted descriptors after this exploited and extended the same idea of aggregating image gradients -- RootSIFT~\cite{RootSIFT2012}, PCA-SIFT~\cite{Ke2004}, HoG~\cite{HoG2005}, RootSIFT-PCA~\cite{Bursuc15}, DSP-SIFT~\cite{DSPSIFT2015}, ConvOpt~\cite{Simonyan2012}. Another, less popular family is the intensity-order based descriptors -- LIOP~\cite{LIOP2011}, MROGH~\cite{MROGH2011}, etc. 
For real-time applications, a series of binary descriptors -- local binary pattern~\cite{LBP1994} -- based on pairwise intensity comparisons was developed. The most popular are BRIEF~\cite{BRIEF2012}, its version ORB~
\cite{ORB2011}, FREAK~\cite{FREAK2012} and so on. They are orders of magnitude faster than SIFT, but worse in terms of accuracy and robustness. 

\paragraph{Learned local feature descriptors.} Simonyan and Zisserman~\cite{Simonyan2012} proposed a simple filter plus pooling scheme learned with convex optimization to replace the hand-crafted filters and poolings in SIFT.
Han~\etal~\cite{MatchNet2015} proposed a two-stage Siamese architecture -- for embedding and for two-patch similarity. The latter network improved the matching performance, but prevented the use of fast approximate nearest neighbor algorithms like kd-tree~\cite{FLANN2009}. Zagoruyko and Komodakis~\cite{DeepComp2015} have independently presented a similar Siamese-based method which explored different convolutional architectures. Simo-Serra~\etal~\cite{simo2015deepdesc} harnessed hard-negative mining with a relative shallow architecture that exploited pair-based similarity.

The next three papers have followed the classical SIFT matching scheme in the training procesure.
Balntas~\etal~\cite{TFeat2016} used a triplet margin loss and a triplet distance loss, with random sampling of the patch triplets. They show the superiority of the triplet-based architecture over a pair-based. Although, unlike SIFT matching or our work, they sampled negatives randomly. Choy~\etal~\cite{UCN2016} calculate the distance matrix for mining positive as well as negative examples, followed by pairwise contrastive loss. 

Tian~et~al~\cite{L2Net2017} proposed a descriptor called L2Net. L2Net uses $n$ matching pairs in batch for generating $n^2 - n$ negative samples and requires that the distance to the ground truth matches is minimum in each row and column. No other constraint on the distance or distance ratio is enforced. Instead, they propose a penalty for the correlation of the descriptor dimensions and adopt deep supervision~\cite{DeepSupervision2014} by using intermediate feature maps for matching. 
Given the good performance of the L2Net, we have adopted its architecture as a base for our descriptor. We show that it is possible to learn an even more powerful descriptor with a significantly simpler learning objective without the need of two auxiliary loss terms.

\begin{figure}[htb]
\centering
 \includegraphics[width=0.98\linewidth]{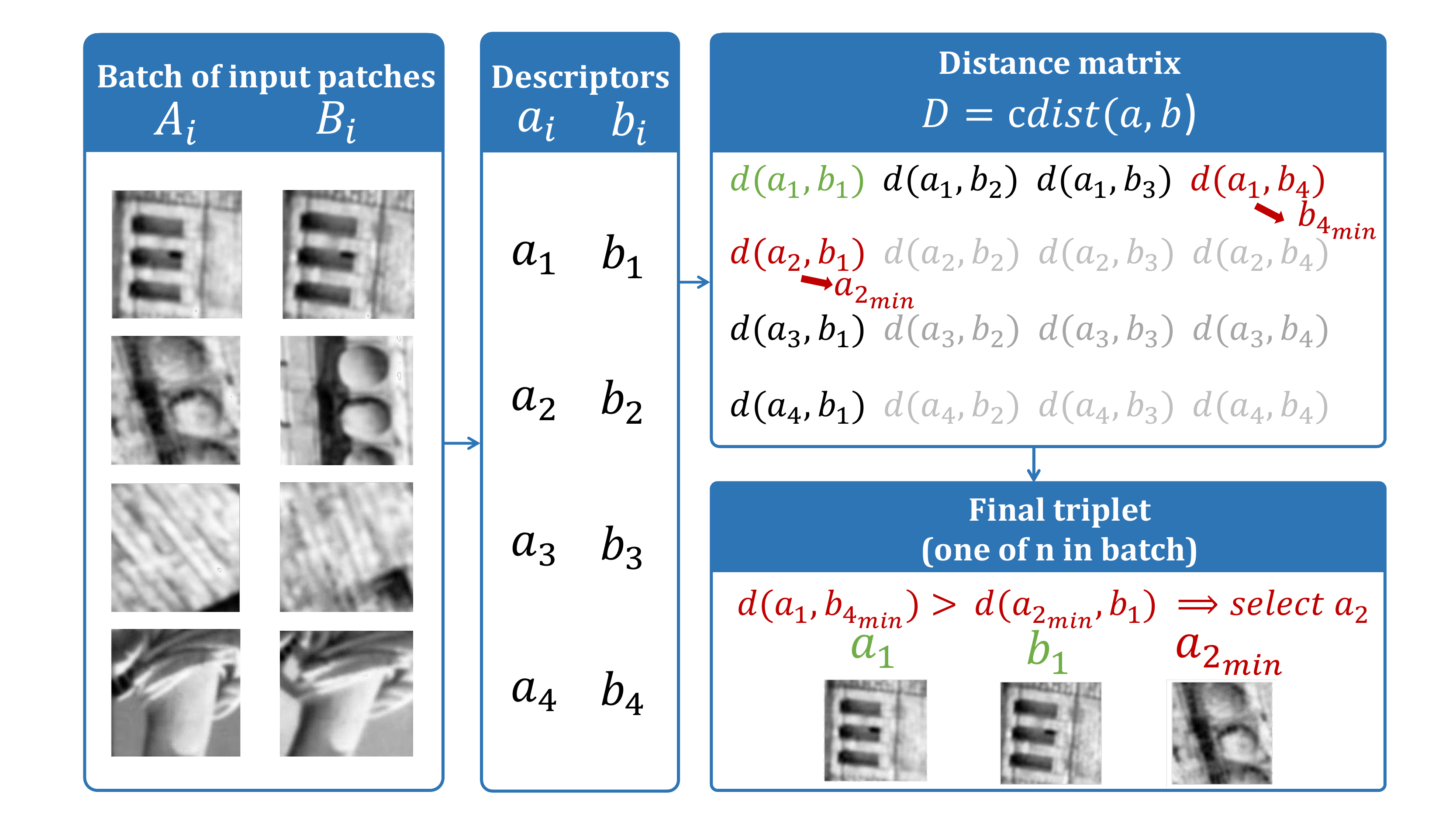}
 \caption{HardNet sampling procedure. First, patches are described by the descriptor network, then a distance matrix is calculated. The closest non-matching descriptor -- shown in red -- is selected for each $A_i$ and $B_i$ patch from positive pair (green) respectively. Finally, among two negative candidates the hardest one is chosen. All operations are done in a single forward pass.}
 \label{fig:sampling}
\end{figure}
\section{The HardNet descriptor}
\label{hardnet:main}
\subsection{Sampling and loss}
\label{main:loss}
Our learning objective mimics SIFT matching criterion. The process is shown in Figure~\ref{fig:sampling}.

First, a batch $\mathcal{X} = (A_i, B_i)_{i=1..n}$ of matching local patch pairs is generated. Both patches from a pair $(A_i, B_i)$ correspond to the same point on the 3D surface. One could sample not pairs, but bigger tuples of the matching patches, however, we have not observed the benefits of doing so. We make sure that in batch $\mathcal{X}$, there is exactly one pair originating from a given 3D point. 

Second, the $2n$ patches in $\mathcal{X}$ are passed through the network as shown in Figure~\ref{fig:L2Net}.

L2 pairwise distance matrix $D = \text{cdist}(a,b)$, where, $d(a_i, b_j) =2\sqrt{1 - a_i b_j}$, $i=1..n, j=1..n$ of size $ n\times n$ is calculated, where $a_i$ and $b_j$ denote the descriptors of patches $A_i$ and $b_j$ respectively.

Next, for each descriptor from the matching pair $(a_i, b_i)$ the closest non-matching descriptors %
are found.%

\begin{equation}
\begin{multlined}
j_{min} = \argmin_{j = 1..n, j \neq i}{d(a_i,b_j)},\\
k_{min} = \argmin_{k = 1..n, k \neq i}{d(a_k,b_i)}
\end{multlined}
\end{equation}

Then from each quadruplet of descriptors $(a_i, b_i, b_{j_{min}}, a_{k_{min}})$, a triplet is formed:
$(a_i, b_i, b_{j_{min}})$, if $d(a_i, b_{j_{min}}) < d(a_{k_{min}}, b_i)$ and $(b_i, a_i, a_{k_{min}})$ otherwise. 

Our goal is to minimize the distance between the matching descriptor and the closest non-matching descriptor. These $n$ triplet distances are fed into the triplet margin loss: 
\begin{equation}
L = \frac{1}{n}\sum_{i = 1,n}{\max{(0, 1 + d(a_i, b_i) - \min{(d(a_i, b_{j_{min}}), d(a_{k_{min}}, b_i))})}}
\end{equation}
where $\min{(d(a_i, b_{j_{min}}), d(a_{k_{min}}, b_i)}$ is precomputed during the triplet construction. 

The distance matrix calculation is done on GPU and the only overhead compared to the random triplet sampling is the distance matrix calculation and calculating the minimum over rows and columns. Moreover, compared to usual learning with triplets, our scheme needs only two-stream CNN, not three, which results in 30\% less memory consumption and computations. After the paper publication we have discovered the proposing the similar method to the HardNet in the domain of metric learning for re-ID~\cite{hermans2017defense}.

We experienced no significant overfitting with such procedure, unlike L2Net~\cite{L2Net2017}. This allowed us to get rid of and a constraint on the correlation of descriptor dimensions and simplify the training and implementation.
\subsection{HardNet architecture}
\label{main:model_architexture}
\begin{figure}[htpb]
\centering
 \includegraphics[width=0.98\linewidth]{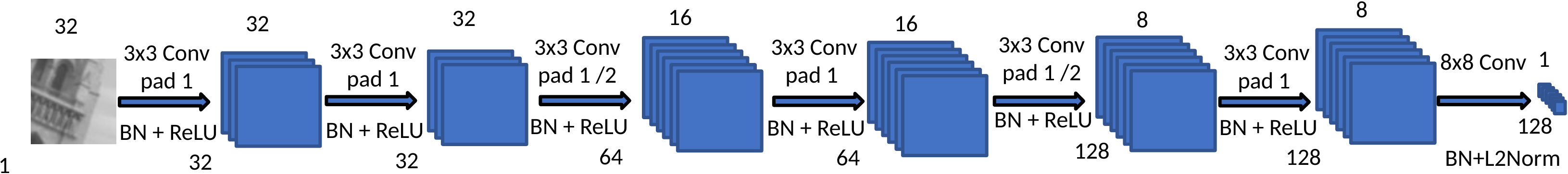}
 \caption{HardNet architecture, adopted from L2Net~\cite{L2Net2017}. Each convolutional layer is followed by batch normalization + ReLU, except the last one, which is followed by batch normalization and L2 normalization. Dropout regularization is used before the last convolution layer. Pad 1 -- symmetrical zero padding, /2 stands for stride 2. Numbers on the top denote number of channels, in the bottom -- spatial size.}
 \label{fig:L2Net}
\end{figure}
 
The HardNet architecture, shown in Figure~\ref{fig:L2Net}, is identical to L2Net~\cite{L2Net2017}. Padding with zeros is applied to all convolutional layers, to preserve the spatial size, except for the final one. There are no pooling layers, since we found that they decrease performance of the descriptor. That is why the spatial size is reduced by strided convolution. Batch normalization~\cite{BatchNorm2015} layer followed by ReLU~\cite{Nair2010RectifiedLinearUnits} non-linearity is added after each layer, except the last one, where ReLU is replaced with L2 normalization. Thus, the output of the network produces 128-D descriptor with a unit length. Dropout~\cite{Dropout2014} regularization with 0.3 dropout rate is applied before the last convolution layer. 
Grayscale input patches with size $32 \times 32$ pixels are normalized by subtracting the per-patch mean and dividing by the per-patch standard deviation. 
 
Optimization is done by stochastic gradient descent with a learning rate of 10.0 per batch size 1024, momentum of 0.9, and weight decay of 0.0001. Learning rate was linearly decayed to zero within 10 epochs for most of the experiments in this Chapter, as recommended in~\cite{Systematic2017}. Weights were initialized to orthogonally~\cite{OrthoNorm2013} with a gain equal to 0.6, biases set to 0.01. Training is done with PyTorch library~\cite{pytorch}. 

The original version of HardNet was trained with slightly different hyperparameters: learning rate 0.1 and dropout rate 0.1. We present the results for both versions in Figure~\ref{fig:hpatches-all} and Table~\ref{tab:verification_brown_performance}. 
\subsection{Model training}
\label{experiments:brown}
UBC Phototour~\cite{Brown2007}, also known as Brown dataset, is used for the training of the base HardNet. UBC  Phototour consists of three subsets: \emph{Liberty}, \emph{Notre Dame} and \emph{Yosemite} with about 400k normalized 64x64 patches in each. Keypoints were detected by DoG detector and verified by the 3D model. 

Test set consists of 100k matching and non-matching pairs for each sequence. Common setup is to train the descriptor on one subset and test on two others. Metric is the false positive rate (FPR) at a point of 0.95 true positive recall. It was found out by Michel Keller that~\cite{MatchNet2015} and~\cite{TFeat2016} evaluation procedure reports FDR (false discovery rate) instead of FPR (false positive rate). To avoid the incomprehension of results we’ve decided to provide both FPR and FDR rates and re-estimated the scores for straight comparison. Results are shown in Table~\ref{tab:verification_brown_performance}. The proposed descriptor outperforms competitors, with training augmentation, or without it. %

In the rest of the Chapter we use a descriptor trained on \emph{Liberty} sequence, unless stated otherwise. %

\npdecimalsign{.}
\nprounddigits{3}
\renewcommand{\ra}[1]{\renewcommand{\arraystretch}{#1}}
\begin{table*}[htb]
\ra{1.2}
\centering
\caption{Patch correspondence verification performance on the Brown dataset. We report a false positive rate at the true positive rate equal to 95\% (FPR95). \textbf{Some papers report false discovery rate (FDR) instead of FPR due to a bug in the source code}. For consistency, we provide FPR, either obtained from the original article or re-estimated from the given FDR (marked with *). The best results are in \textbf{bold}.}
\label{tab:verification_brown_performance}
\small
\begin{tabular}{@{}lcccccccc@{}} 
\toprule
Training  & Notredame & Yosemite & Liberty  & Yosemite & Liberty & Notredame  & \multicolumn{2}{c}{Mean}    \\
\cmidrule(r){2-3} \cmidrule(r){4-5} \cmidrule(r){6-7} \cmidrule(r){8-9} 
Test    & \multicolumn{2}{c}{Liberty} & \multicolumn{2}{c}{Notredame} & \multicolumn{2}{c}{Yosemite} & \multicolumn{1}{l}{FDR} & \multicolumn{1}{l}{FPR} \\
\cmidrule(r){1-9}
SIFT \cite{SIFT2004}     & \multicolumn{2}{c}{29.84}  & \multicolumn{2}{c}{22.53}   & \multicolumn{2}{c}{27.29}  &  & 26.55          \\
MatchNet*\cite{MatchNet2015} & 7.04     & 11.47         & 3.82       &  5.65     & 11.6     & 8.7     & 7.74 & 8.05 \\
TFeat-M* \cite{TFeat2016}  & 7.39     & 10.31         & 3.06       &  3.8     & 8.06     & 7.24     & 6.47& 6.64   \\
PCW \cite{PCW2017}  & 7.44 &9.84 &3.48 &3.54& 6.56 &5.02& &5.98\\
L2Net \cite{L2Net2017}    & 3.64     & 5.29        & 1.15         & 1.62     & 4.43     & 3.30     &  & 3.24  \\
HardNetNeurIPS          & 3.06 & 4.27    & 0.96     & 1.4     & 3.04     & 2.53     & 3.00   &2.54 \\
HardNet          & \textbf{1.47} & \textbf{2.67}    & \textbf{0.62}     & \textbf{0.88}     & \textbf{2.14}     & \textbf{1.65}     & \textbf{ }  & \textbf{1.57} \\

\cmidrule(r){1-9}
\multicolumn{9}{c}{Augmentation: flip, $90\degree$ random rotation } \\
\cmidrule(r){1-9}
GLoss+\cite{GlobalLoss} & 3.69     & 4.91         &0.77       &  1.14     & 3.09    & 2.67    &  & 2.71    \\
DC2ch2st+\cite{DeepComp2015} & 4.85     & 7.2         & 1.9       &  2.11     & 5.00     & 4.10     &  & 4.19  \\
L2Net+ \cite{L2Net2017} +   & 2.36     & 4.7     & 0.72     & 1.29     & 2.57     & \textbf{1.71}     &  & 2.23           \\
HardNet+NeurIPS     & 2.28 &  3.25   & 0.57 & 0.96 & 2.13     & 2.22 &   1.97 &   1.9\\
HardNet+     & \textbf{1.49} &  \textbf{2.51}   & \textbf{0.53} & \textbf{0.78} & \textbf{1.96}     & 1.84 &   \textbf{} &   \textbf{1.51}  \\       
\bottomrule
\end{tabular}
\end{table*}
\section{Empirical evaluation}
\label{hardnet:experiments}
We have extensively evaluated the learned descriptors on real-world tasks like two-view matching and image retrieval, as it is known~\cite{TFeat2016} that good performance on the patch verification task on Brown dataset does not always mean good performance in the nearest neighbor setup and vice versa.

We have selected RootSIFT~\cite{RootSIFT2012}, PCW~\cite{PCW2017}, TFeat-M*~\cite{TFeat2016}, and L2Net~\cite{L2Net2017} for direct comparison with our descriptor, as they show the best results on a variety of datasets.
\subsection{Patch descriptor evaluation}
\begin{figure}[htb]
\begin{center}
 \includegraphics[width=0.3\linewidth]{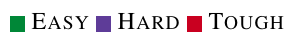}\\
 \end{center}
 \vspace{-0.3cm} 
 \includegraphics[width=0.25\linewidth]{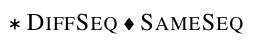}
 \hspace{1cm}
 \includegraphics[width=0.2\linewidth]{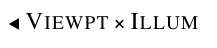} \hspace{5cm} \\
 \centering
 \includegraphics[width=0.32\linewidth]{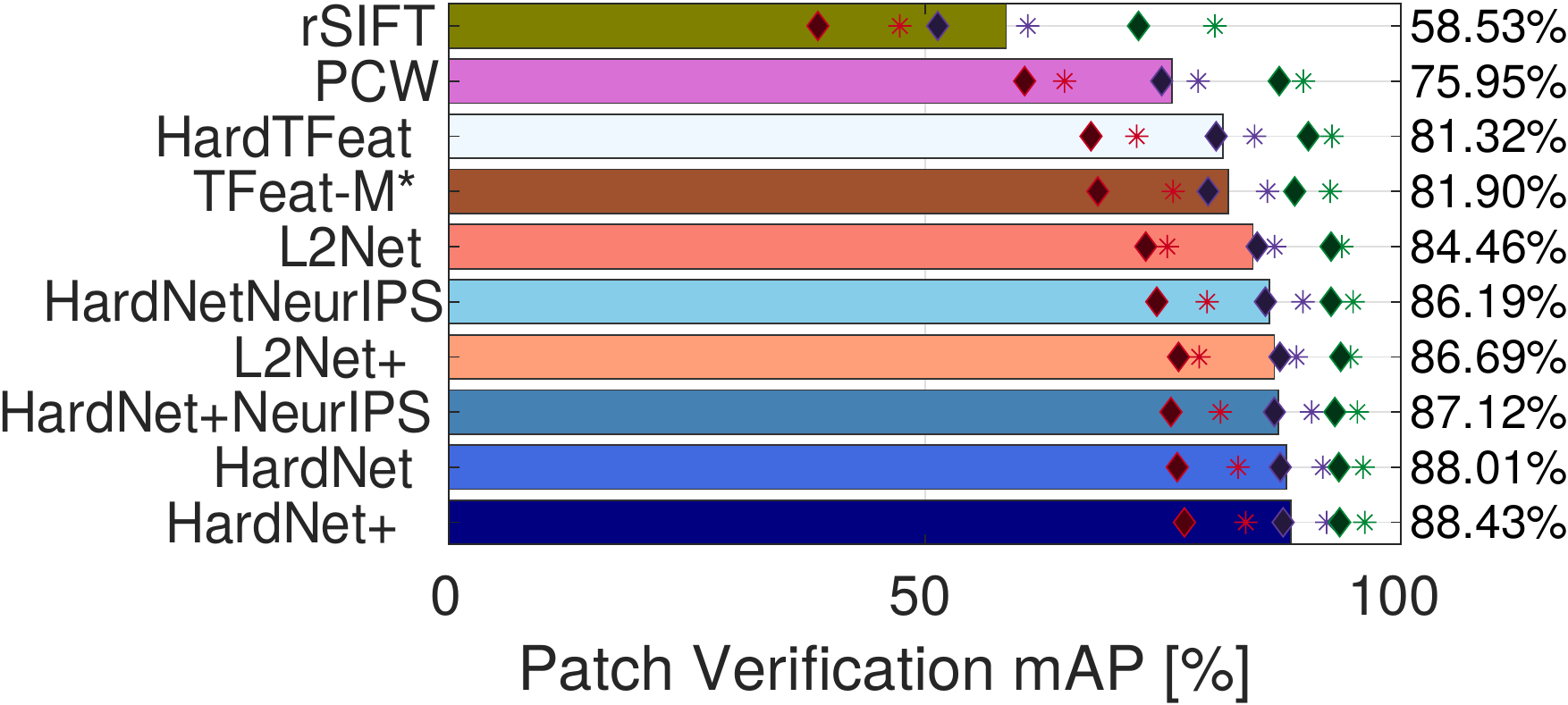}
 \includegraphics[width=0.32\linewidth]{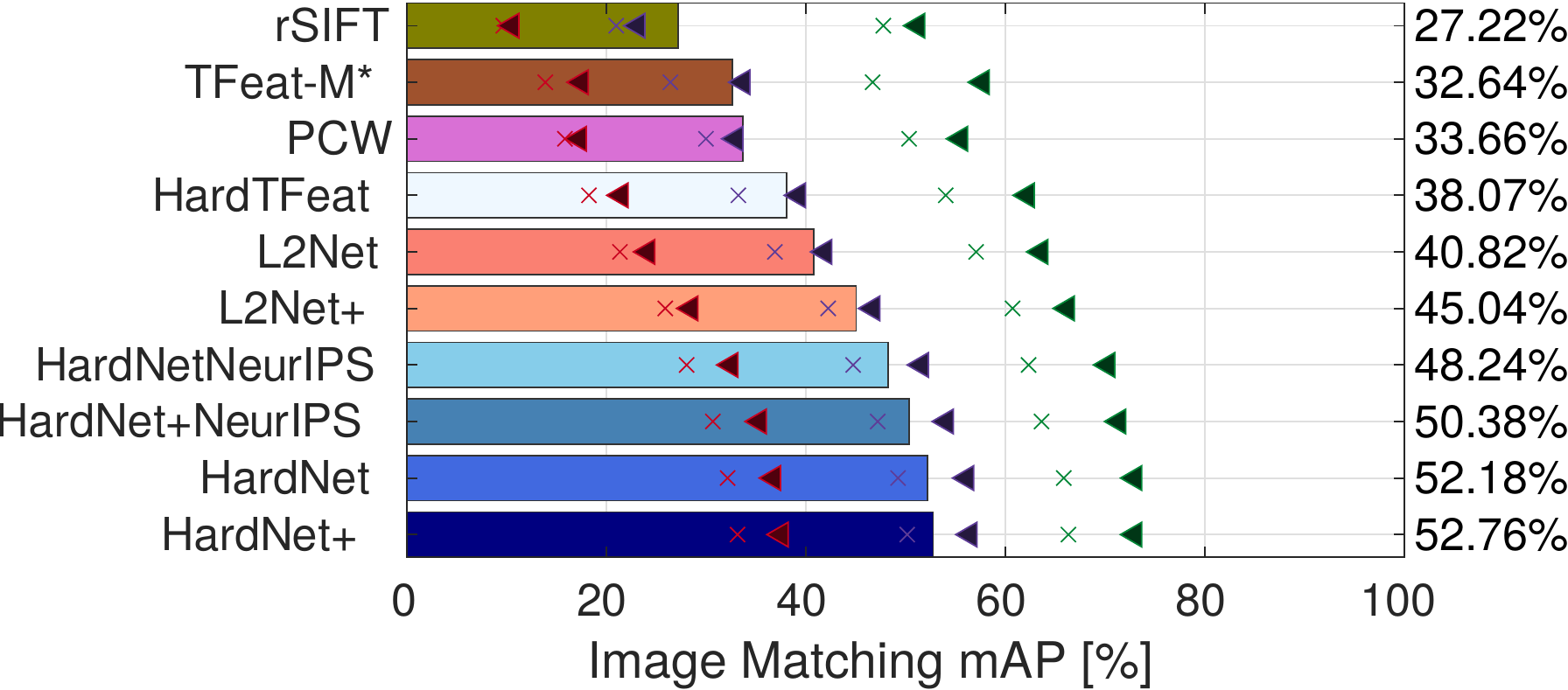}
 \includegraphics[width=0.32\linewidth]{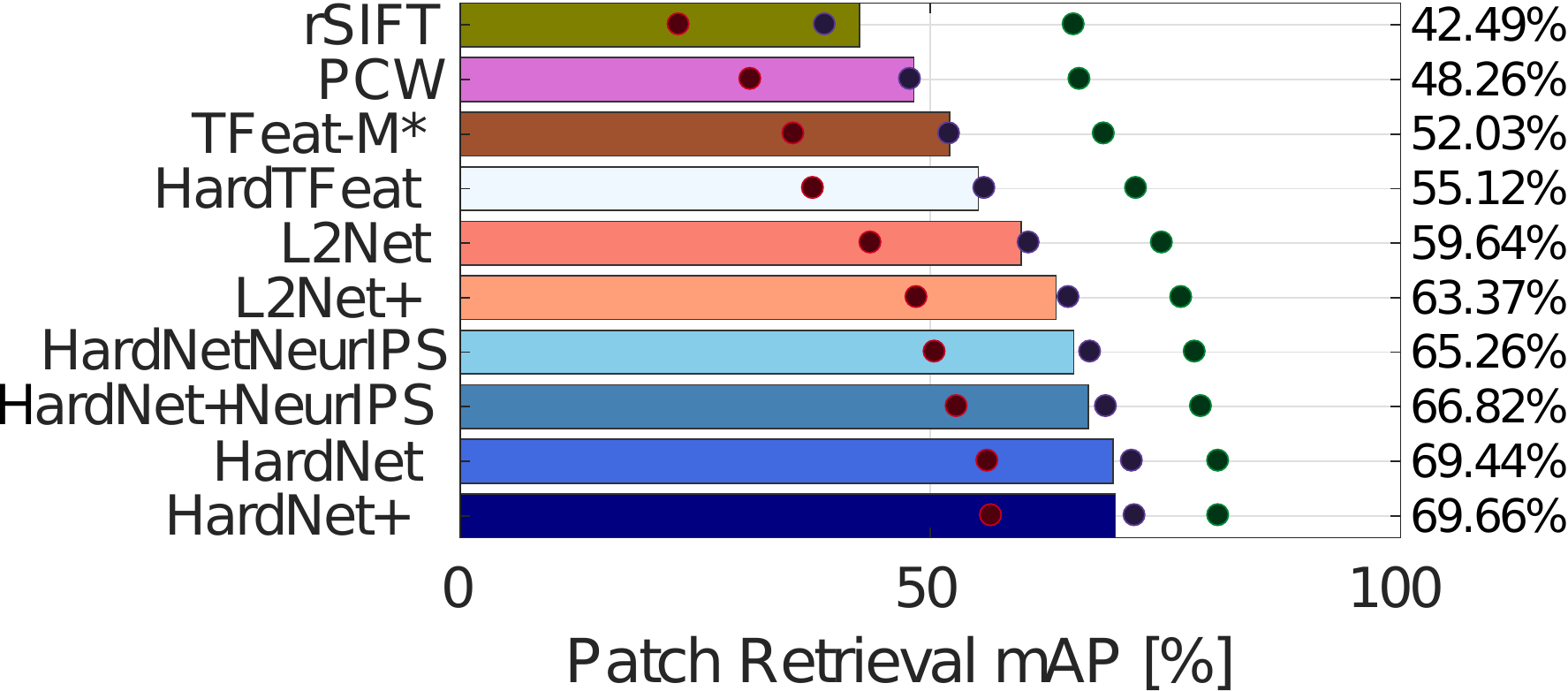}\\
  \caption{Left to right: Verification, matching and retrieval results on HPatches dataset. All the descriptors are trained on \emph{Liberty} subset of Brown~\cite{Brown2007} dataset, or are handcrafted. Comparison of descriptors trained on other datasets are in Figure~\ref{fig:hpatches-other}. 
 Marker color indicates the level of geometrical noise in: \textsc{Easy}, \textsc{Hard} and \textsc{Tough}. Marker type indicates the experimental setup. 
\textsc{DiffSeq} and \textsc{SameSeq} shows the source of negative examples for the verification task. \textsc{Viewpt} and \textsc{Illum} indicate the type of sequences for matching.%
The HardNet and HardNet+ were trained after the HardNet paper~\cite{HardNet2017} published with optimized hyperparameters learning rate.}
\label{fig:hpatches-all}
\end{figure}
HPatches~\cite{hpatches2017} is a dataset for local patch descriptor evaluation. It consists of 116 sequences of 6 images. The dataset is split into two parts: \textit{viewpoint} -- 59 sequences with significant viewpoint change and \textit{illumination} -- 57 sequences with significant illumination changes, both natural and artificial. 
Keypoints are detected by DoG, Hessian, and Harris detectors in the reference image and reprojected to the rest of the images in each sequence with 3 levels of geometric noise: \textit{Easy}, \textit{Hard}, and \textit{Tough} variants. The HPatches benchmark defines three tasks: patch correspondence verification, image matching, and small-scale patch retrieval. Patch correspondence verification checks how good descriptors in determination if two (random) patches are in correspondence or not. Image matching and patch retrieval are closer to the real-world applications and check how good is the descriptor in the retrieval of the correct match as a nearest neighbor. The difference between image matching and patch retrieval protocols is that the matching protocol considers only the patches from the single pair of images, while retrieval -- from multiple images. We refer the reader to the HPatches paper~\cite{hpatches2017} for a detailed protocol for each task. 

Results are shown in Figure~\ref{fig:hpatches-all}. L2Net and HardNet have shown similar performance on the patch verification task with a small advantage of HardNet. On the matching task, even the non-augmented version of HardNet outperforms the augmented version of L2Net+ by a noticeable margin. The difference is larger in the \textsc{Tough} and \textsc{Hard} setups.
Illumination sequences are more challenging than the geometric ones for all descriptors. 
We have trained a network with TFeat architecture, but with the proposed HardNet loss function -- it is denoted as HardTFeat. It outperforms the original version in matching and retrieval, while being on par with it on the patch verification task.
 
In patch retrieval, the relative performance of the descriptors is similar to the matching problem: HardNet beats L2Net+. Both descriptors significantly outperform the previous state-of-the-art, showing the superiority of the selected deep CNN architecture over the shallow TFeat model.

\begin{figure}[htb]
\begin{center}
 \includegraphics[width=0.3\linewidth]{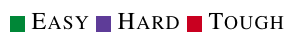}\\
 \end{center}
 \vspace{-0.3cm} 
 \includegraphics[width=0.25\linewidth]{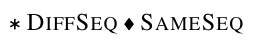}
 \hspace{1cm}
 \includegraphics[width=0.2\linewidth]{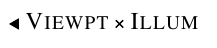} \hspace{5cm} \\
 \centering
 \includegraphics[width=0.31\linewidth]{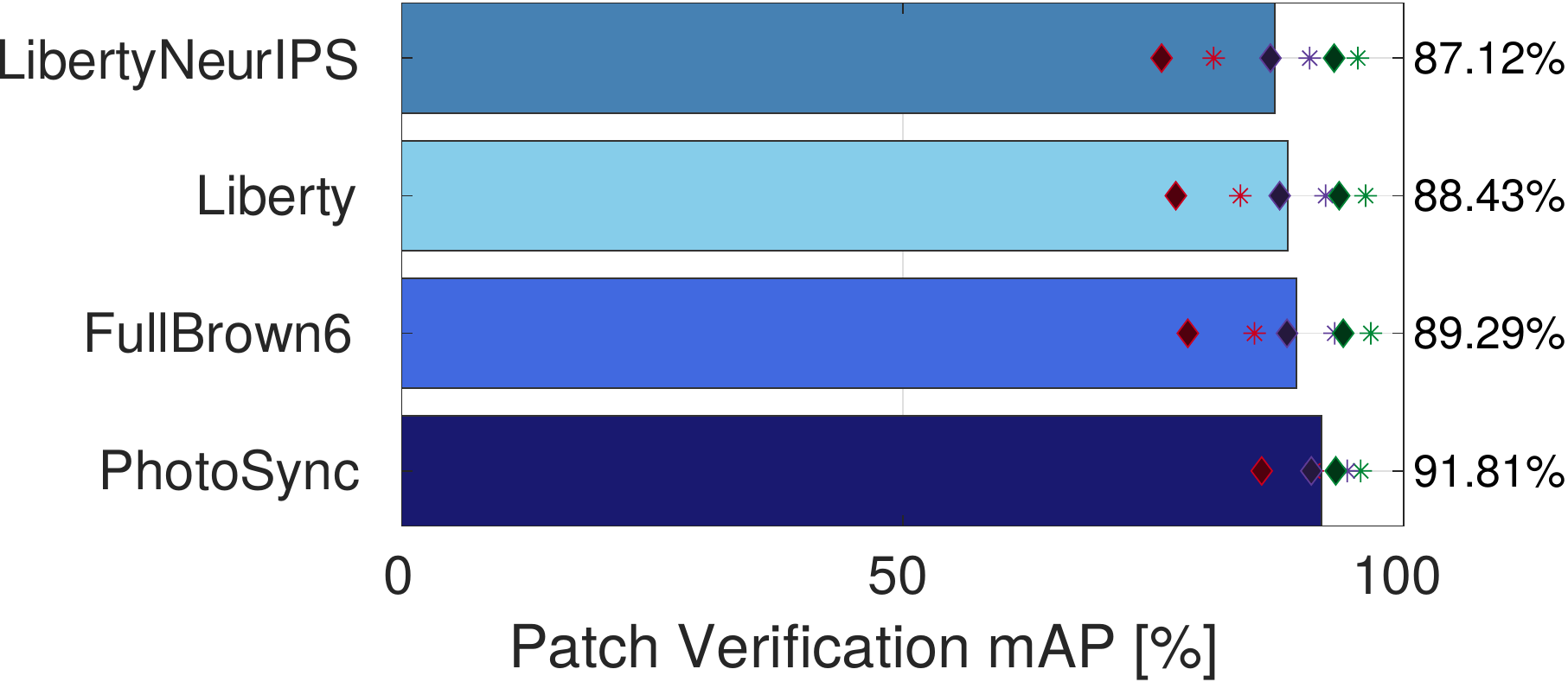}
 \includegraphics[width=0.30\linewidth]{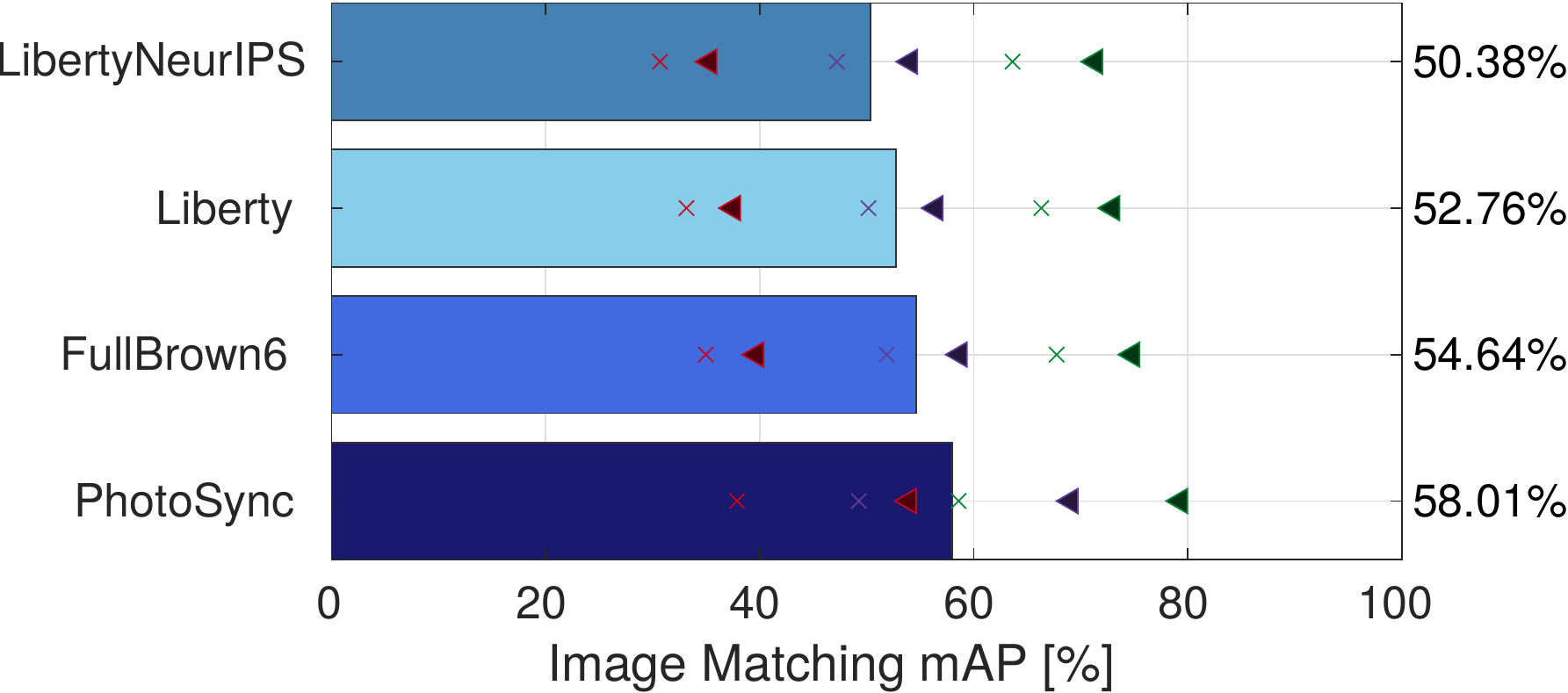}
 \includegraphics[width=0.31\linewidth]{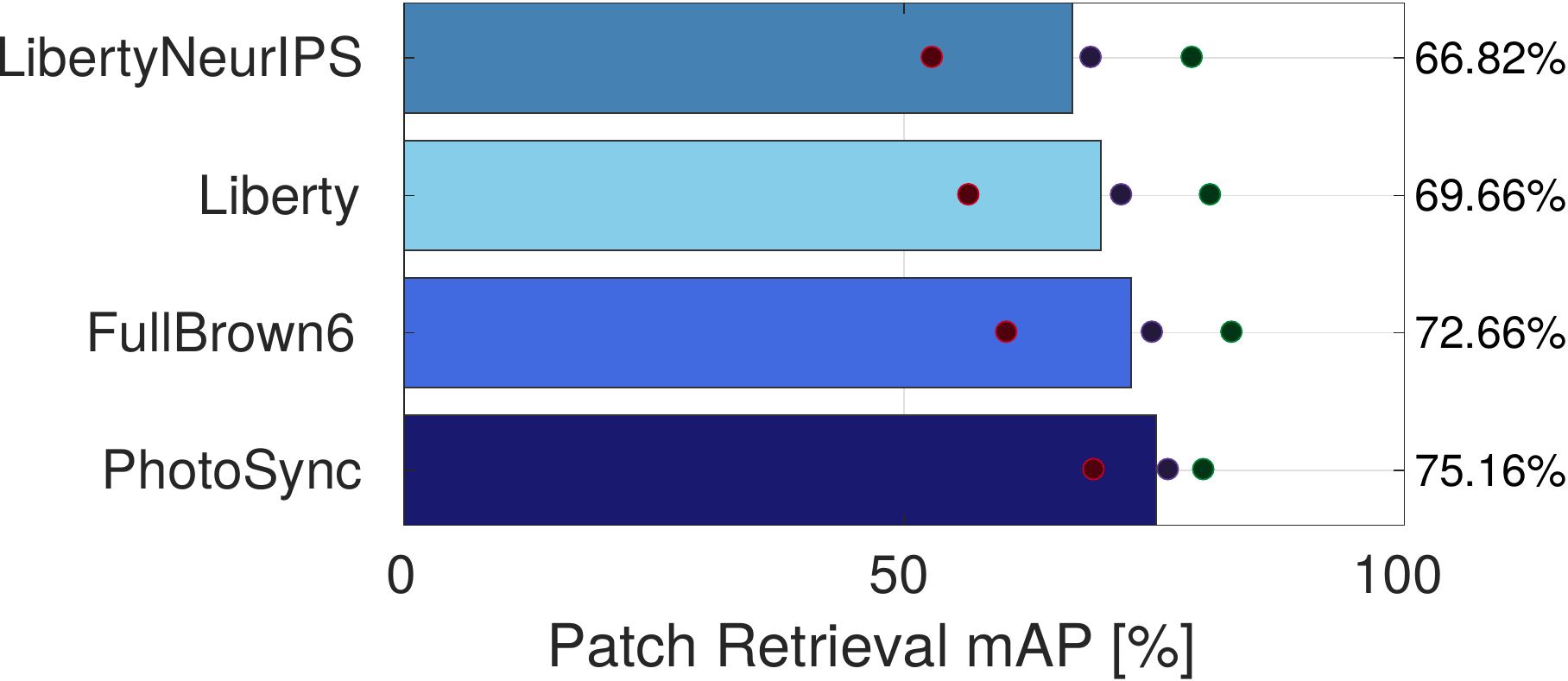}\\
  \caption{Comparison of HardNet versions, trained on different datasets. LibertyNeurIPS, Liberty -- \emph{Liberty}\cite{Brown2007}, FullBrown6 -- all six subsets of Brown dataset~\cite{Brown2007}, PhotoSync -- dataset from Mitra~\etal~\cite{HardNetPS2018}. 
 Left to right: Verification, matching and retrieval results on HPatches dataset. Marker color indicates the level of geometrical noise in: \textsc{Easy}, \textsc{Hard} and \textsc{Tough}. Marker type indicates the experimental setup. 
\textsc{DiffSeq} and \textsc{SameSeq} shows the source of negative examples for the verification task. \textsc{Viewpt} and \textsc{Illum} indicate the type of sequences for matching. 
\hspace{\linewidth}}
\label{fig:hpatches-other}
\end{figure}

We have studied the impact of the training dataset on descriptor performance. In particular, we compared the commonly used \emph{Liberty} subset with patches extracted by DoG detector, the full Brown dataset -- subsets \emph{Liberty}, \emph{Notredame} and \emph{Yosemite} with patches, extracted by DoG and Harris detectors. We also included results by Mitra~\etal, who trained HardNet on a new large-scale PhotoSync~\cite{HardNetPS2018} dataset. Results are shown in Figure~\ref{fig:hpatches-other}.
\subsection{Ablation study}
\label{hardnet:ablation}
\begin{table}[htp]
\ra{1.0}
\centering
\caption{Comparison of the loss functions and sampling strategies on the HPatches matching task, the mean mAP is reported. CPR stands for the regularization penalty of the correlation between descriptor channels, as proposed in~\cite{L2Net2017}. Hard negative mining is performed once per epoch. Best results are in \textbf{bold}. HardNet uses the hardest-in-batch sampling and the triplet margin loss.}
\label{tab:ablation}
\setlength{\tabcolsep}{1mm}
\begin{tabular}{lcccc}
\toprule
 Sampling / Loss& Softmin & Triplet margin & \multicolumn{2}{c}{Contrastive}\\
 \cmidrule(r){4-5}
 & & $m = 1$ & $m = 1$ & $m = 2$\\
\cmidrule(r){1-5}
Random & \multicolumn{4}{c}{overfit} \\
Hard negative & \multicolumn{4}{c}{overfit} \\
Random + CPR & 0.349 & 0.286 & 0.007 & 0.083\\
Hard negative + CPR &0.391 & 0.346 & 0.055&0.279 \\
\cmidrule(r){1-5}
Hardest in batch (ours) &0.474 & \textbf{0.482} & 0.444 & \textbf{0.482} \\
\bottomrule
\end{tabular}
\end{table}
\begin{figure}[htb]
\centering
 \includegraphics[width=0.18\linewidth]{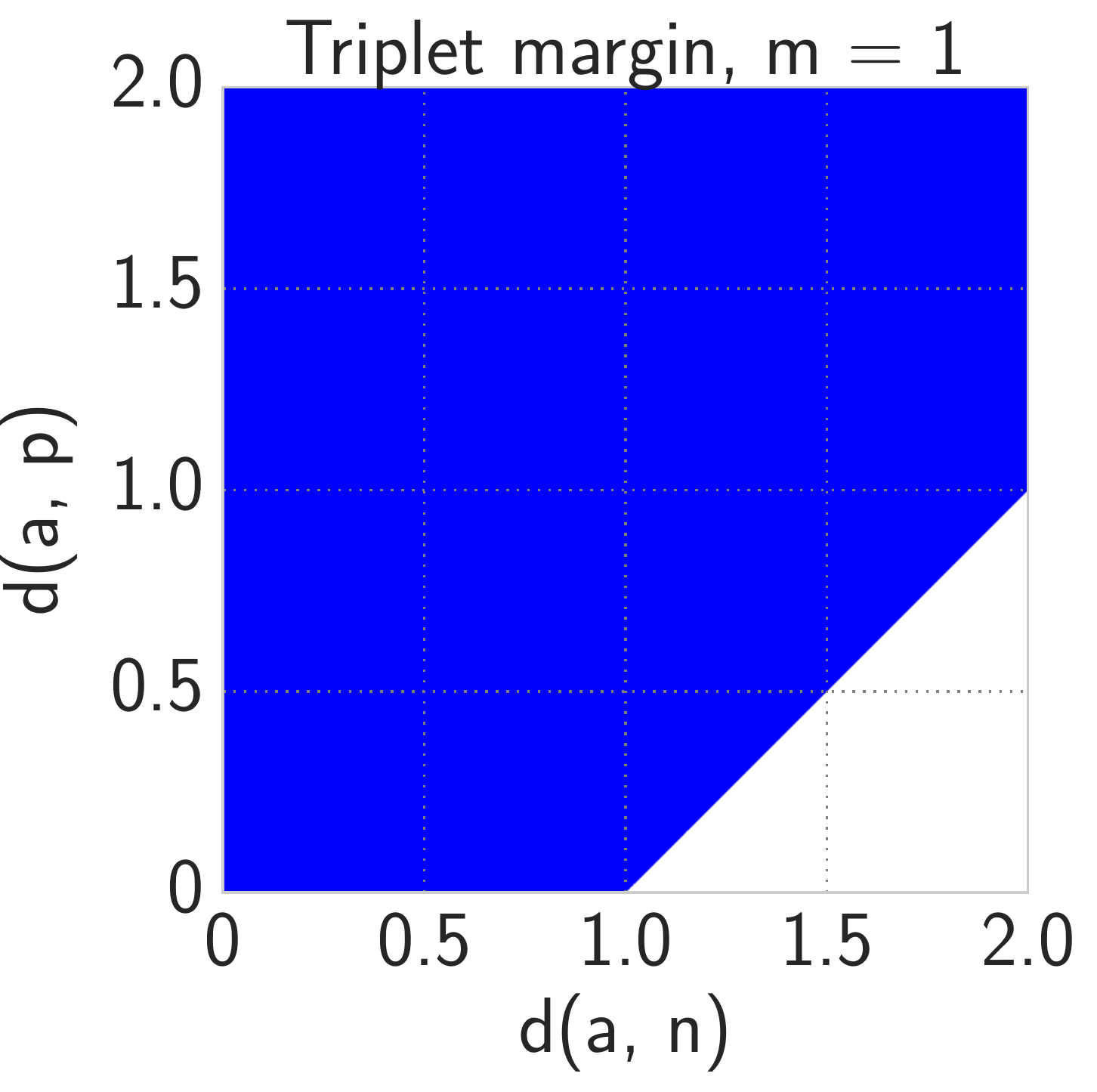}
 \includegraphics[width=0.18\linewidth]{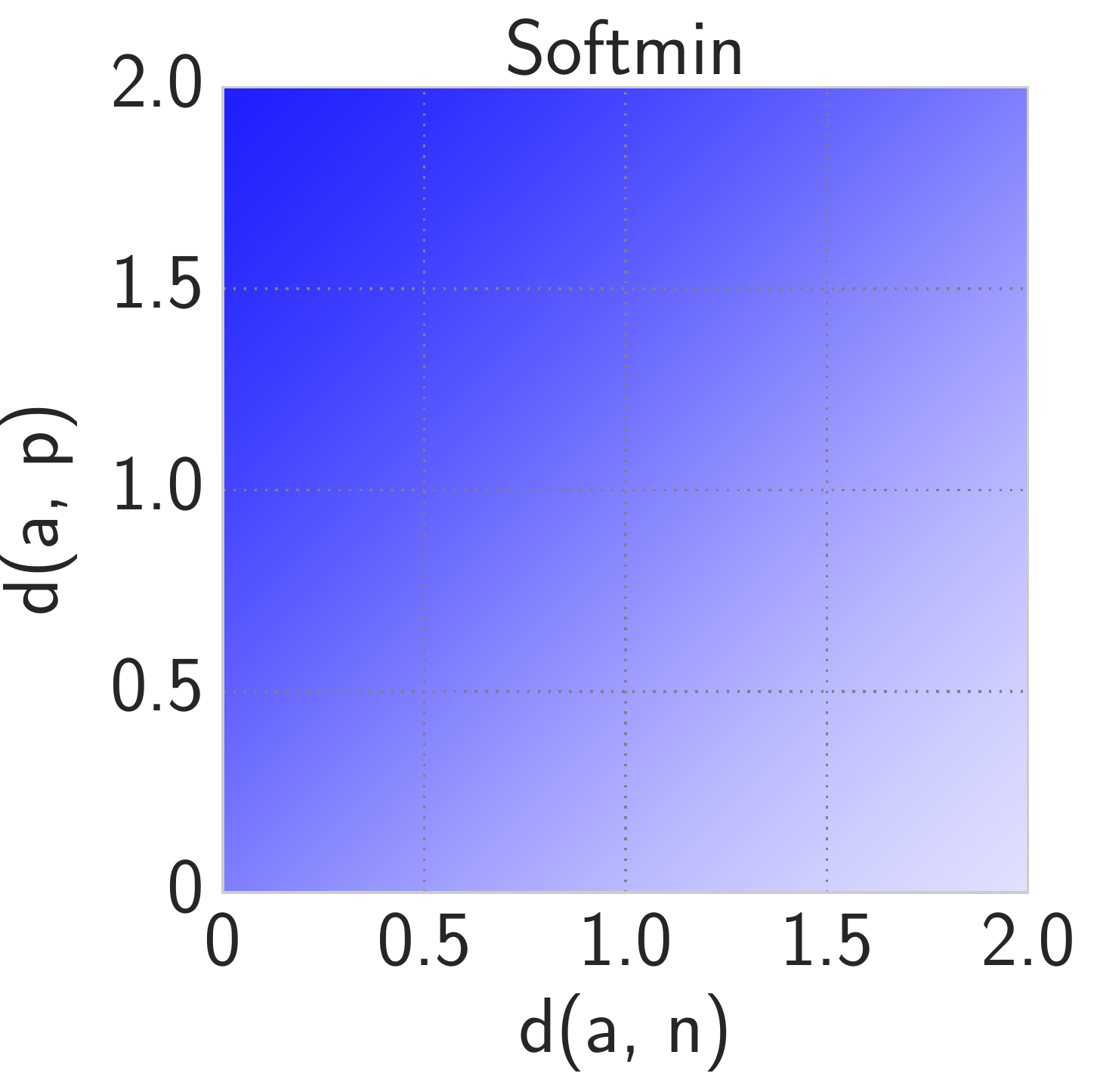}
 \includegraphics[width=0.18\linewidth]{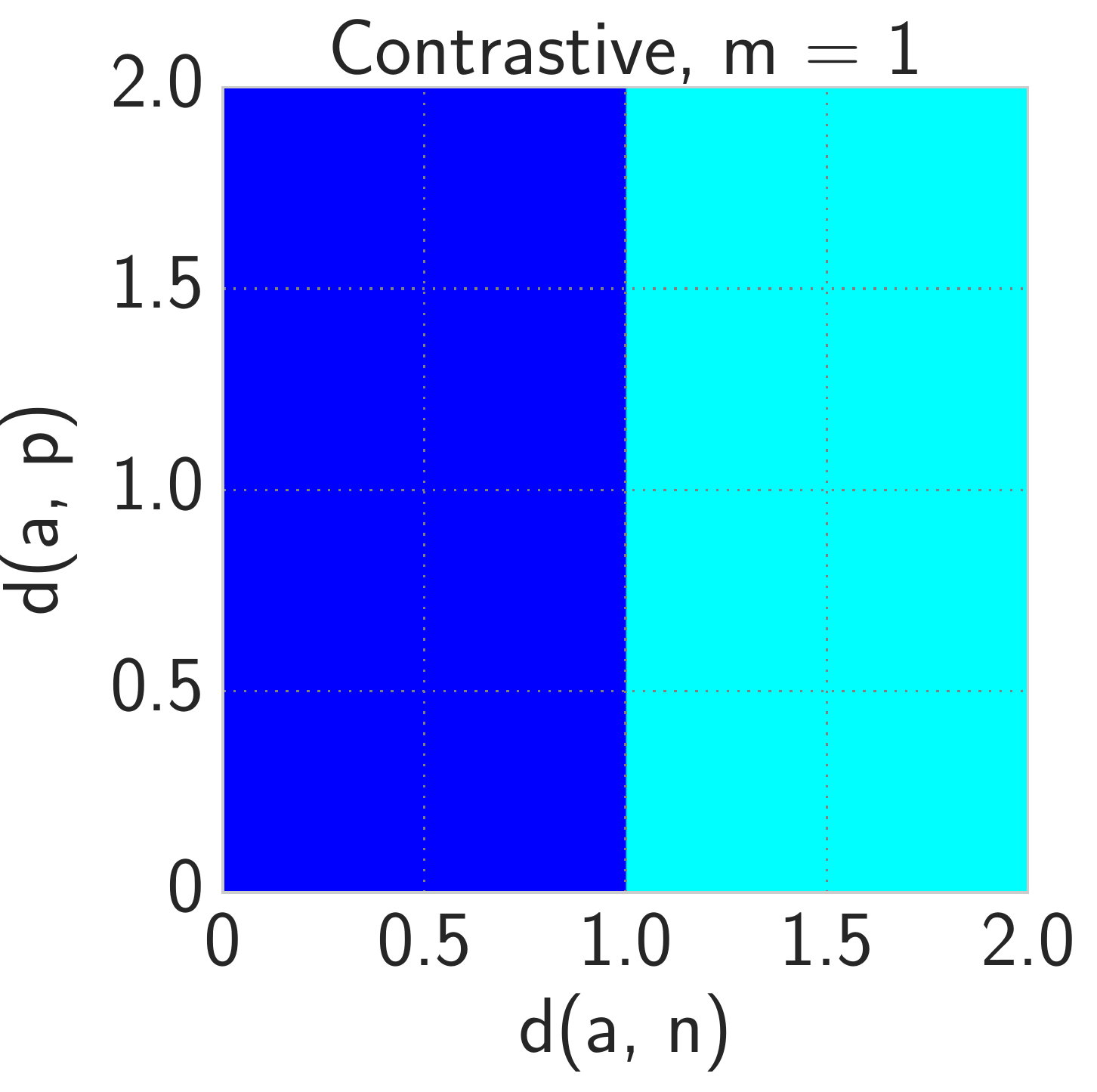}
 \includegraphics[width=0.18\linewidth]{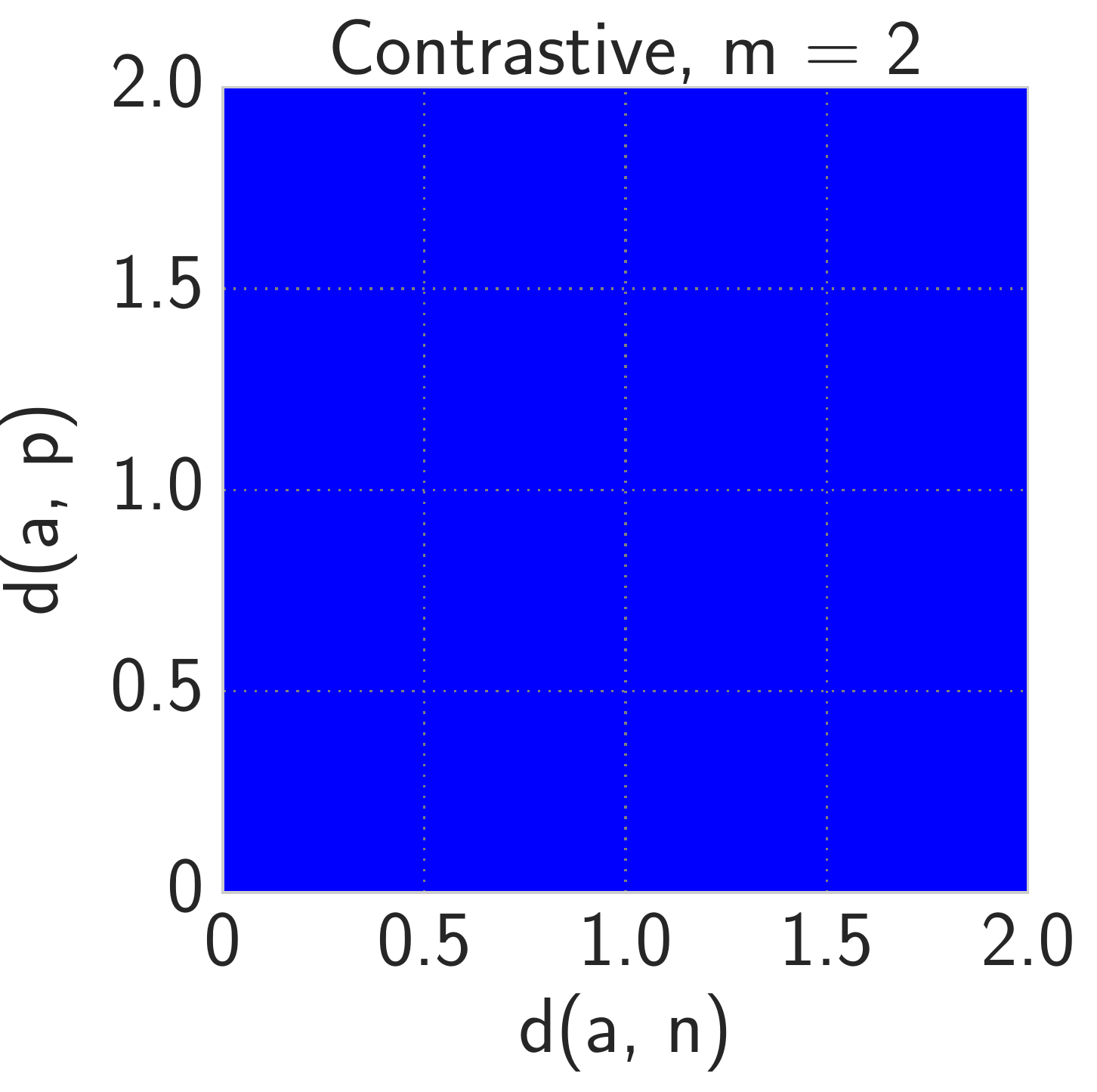}
 \raisebox{0.3\height}{\includegraphics[width=0.03\linewidth]{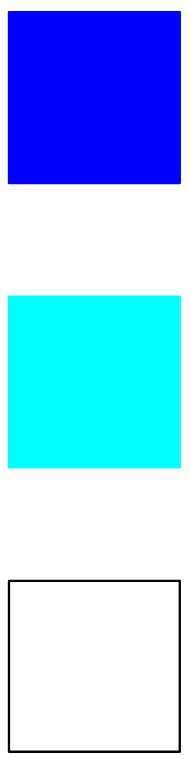}}
  \raisebox{1.25\height}{{\renewcommand{\arraystretch}{1.6} \begin{tabular}{l} $\frac{\partial L}{\partial n} = \frac{\partial L}{\partial p} = 1$\\
 $\frac{\partial L}{\partial n} = 0, \frac{\partial L}{\partial p} = 1$\\
 $\frac{\partial L}{\partial n} = \frac{\partial L}{\partial p} = 0$\end{tabular}}} \\
 \caption{Contribution to the gradient magnitude from the positive and negative examples. Horizontal and vertical axes show the distance from the anchor (a) to the negative (n) and positive (p) examples respectively. 
Softmin loss gradient quickly decreases when $d(a,n) > d(a,p)$, unlike the triplet margin loss. For the contrastive loss, negative examples with $d(a,n) > m$ contribute zero to the gradient. The triplet margin loss and the contrastive loss with a big margin behave very similarly.}
 \label{fig:loss-comparison}
\end{figure}
For a better understanding of the significance of the sampling strategy and the loss function, we conduct the experiments summarized in Table~\ref{tab:ablation}. We train our HardNet model (architecture is the same as L2Net model), change one parameter at a time, and evaluate its impact.

The following sampling strategies are compared: random, the proposed ``hardest-in-batch'', and ``classical'' hard negative mining, i.e., selecting in each epoch the closest negatives from the full training set.
The following loss functions are tested: softmin on distances, triplet margin with margin $m = 1$, contrastive with margins $m = 1, m = 2$. The last is the maximum possible distance for unit-normed descriptors. 
Mean mAP for HPatches Matching task is shown in Table~\ref{tab:ablation}. 

The proposed ``hardest-in-batch'' clearly outperforms all other sampling strategies for all loss functions and it is the main reason for HardNet good performance. The random sampling and “classical” hard negative mining led to huge overfit, when the training loss was high, but the test performance was low and varied several times from run to run. This behavior was observed with all loss functions. Similar results for random sampling were reported in~\cite{L2Net2017}. 

The poor results of hard negative mining (``hardest-in-the-training-set'') are surprising. We guess that this is due to dataset label noise, the mined ``hard negatives'' are actually positives. Visual inspection confirms this. We were able to get reasonable results with random and hard negative mining sampling only with an additional correlation penalty on descriptor channels (CPR), as proposed in~\cite{L2Net2017}.

Regarding the loss functions, softmin gave the most stable results across all sampling strategies, but it is marginally outperformed by contrastive and triplet margin loss for our strategy. One possible explanation is that the triplet margin loss and contrastive loss with a large margin have constant non-zero derivative w.r.t both positive and negative samples, see Figure~\ref{fig:loss-comparison}. In the case of contrastive loss with a small margin, many negative examples are not used in the optimization (zero derivatives), while the softmin derivatives become small once the distance to the positive example is smaller than to the negative one.
\subsection{Wide baseline stereo}
\label{experiments:wbs}
\begin{figure}[htb]
\centering
\includegraphics[width=0.98\linewidth]{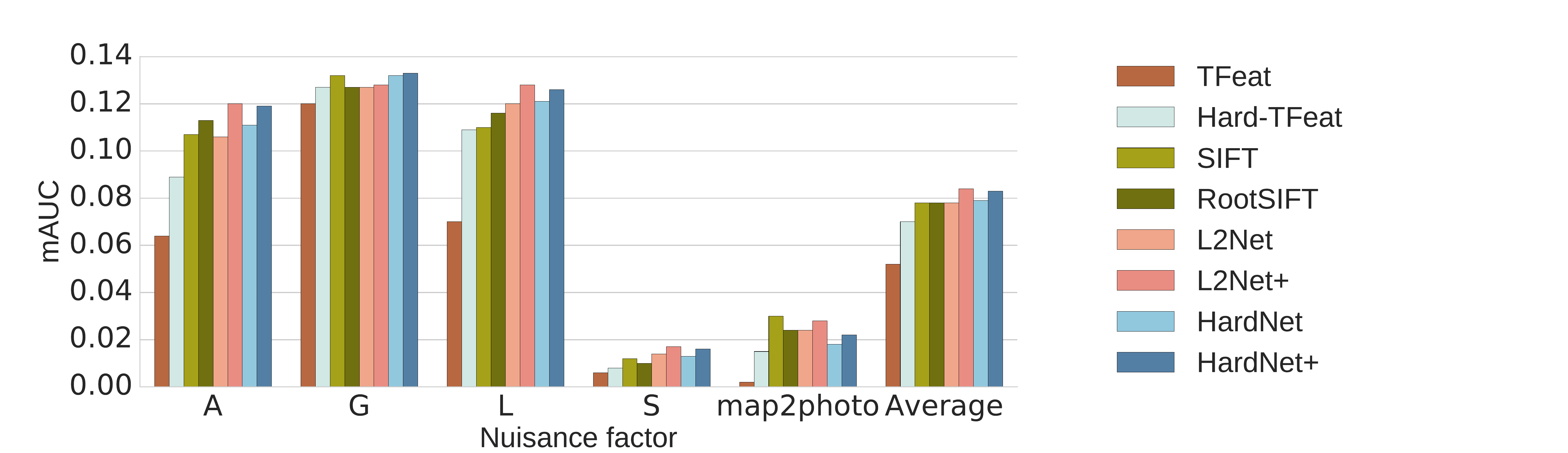}
\vspace{-1em}
\caption{Descriptor evaluation on the W1BS patch dataset, mean area under precision-recall curve is reported. Letters denote nuisance factor, A: appearance; G: viewpoint/geometry; L: illumination; S: sensor; map2photo: satellite photo vs. map.}
\label{fig:w1bs-matching}
\end{figure}

\begin{table}[htb]
\footnotesize
\ra{1.0}
\centering
\caption{Comparison of the descriptors on wide baseline stereo within MODS matcher\cite{WXBS2015} on wide baseline stereo datasets. Number of matched image pairs and average number of inliers are reported. Numbers is the header corresponds to the number of image pairs in dataset.}
\begin{tabular}{lrrrrrrrrrrrrrr}
\toprule
& \multicolumn{2}{c}{EF}
& \multicolumn{2}{c}{EVD}
& \multicolumn{2}{c}{OxAff}
& \multicolumn{2}{c}{SymB}
& \multicolumn{2}{c}{GDB}
& \multicolumn{2}{c}{WxBS}
& \multicolumn{2}{c}{LTLL}
\\
\cmidrule(r){2-3}
\cmidrule(r){4-5}
\cmidrule(r){6-7}
\cmidrule(r){8-9}
\cmidrule(r){10-11}
\cmidrule(r){12-13}
\cmidrule(r){14-15}
Descriptor
 & 33& inl.
 & 15& inl. 
 & 40& inl.
 & 46& inl.
 & 22& inl.
 & 37& inl.
 & 172& inl.\\
\cmidrule(r){1-15}
RootSIFT & \textbf{33} &  32 &  \textbf{15}   &  34 &   \textbf{40}   & 169 &  \textbf{45}   & 43 & \textbf{21}   & 52 & \textbf{11}  & \textbf{93} & 123 &27 \\
TFeat-M* & 32 & 30 &   \textbf{15}   & 37 & \textbf{40}   & 265 &  40   & 45 & 16  & 72 & 10& 62 &96 & 29\\
L2Net+ & \textbf{33}  & 34 &   \textbf{15}    & 34 & \textbf{40}    & 304 & 43 & 46 & 19  & \textbf{78} & 9 & 51 & \textbf{127} & 26\\
HardNet+ & \textbf{33} & \textbf{35} &  \textbf{15}  &   \textbf{41}   & \textbf{40}    & \textbf{316}   & 44 & \textbf{47}   & \textbf{21}   & 75 & \textbf{11}& 54 & \textbf{127} & \textbf{31}\\
\bottomrule
\end{tabular}
\label{tab:wxbs-table}
\end{table}

To validate the descriptors generalization and their ability to operate in extreme conditions, we tested them on the W1BS dataset~\cite{WXBS2015}, which is described in Chapter~\ref{ch:wxbs}. It consists of 40 image pairs with one particular extreme change between the images:
\begin{itemize*}[leftmargin=*, label={}]
\item \textbf{Appearance (A):} difference in appearance due to seasonal or weather change, occlusions, etc;
\item \textbf{Geometry (G):} difference in scale, camera and object position;
\item \textbf{Illumination (L):} significant difference in intensity, wavelength of light source;
\item \textbf{Sensor (S):} difference in sensor data (IR, MRI).
\end{itemize*}

Local features in W1BS dataset are detected with MSER~\cite{MSER2002}, Hessian-Affine~\cite{HesHarAff2004} (in implementation from \cite{PerdochRetrieval2009}) and FOCI~\cite{Zitnick2011} detectors. They fire on different local structures than DoG. Note that DoG patches were used for the training of the descriptors. Another significant difference to the HPatches setup is the absence of geometrical noise: all patches are perfectly reprojected to the target image in pair.
The testing protocol is the same as for the HPatches matching task. 

Results are shown in Figure~\ref{fig:w1bs-matching}. HardNet and L2Net perform comparably, the former is performing better on images with geometrical and appearance changes, while the latter works a bit better in map2photo and visible-vs-infrared pairs. Both outperform SIFT, but only by a small margin. However, considering the significant amount of domain shift, the descriptors perform very well, while TFeat loses badly to SIFT. 
HardTFeat significantly outperforms the original TFeat descriptor on the W1BS dataset, showing the superiority of the proposed loss. 

Good performance on the patch matching and verification task does not automatically lead to the better performance in practice, e.g., to more images registered. Therefore, we also compared the descriptors in the wide baseline stereo setup with two metrics: the number of successfully matched image pairs and the average number of inliers per matched pair, following the matcher comparison protocol from~\cite{WXBS2015}. The only change to the original protocol is that the first fast matching step with ORB detector and descriptor was removed, as we are comparing ``SIFT-replacement'' descriptors. 

The results are shown in Table~\ref{tab:wxbs-table}. Results on Edge Foci (EF)~\cite{Zitnick2011}, Extreme view~\cite{MODS2015} and Oxford Affine~\cite{Mikolajczyk2004} datasets are saturated and all descriptors are good enough for matching all image pairs. HardNet has an a slight advantage in a number of inliers per image. 
The rest of datasets: SymB~\cite{Hauagge2012}, GDB~\cite{Yang2007}, WxBS~\cite{WXBS2015} and LTLL~\cite{LostInPast2015} have one thing in common: image pairs are or from a different domain than the photo (e.g. drawing to drawing) or cross-domain (e.g., drawing to photo). Here HardNet outperforms learned descriptors and is on par with hand-crafted RootSIFT. We would like to note that HardNet was not learned to match in different domains, nor cross-domain scenarios, therefore such results show the generalization ability.
\subsection{Image retrieval}
\label{hardnet:imretrieval}
\begin{table}[tb]
\ra{1.0}
\centering
\caption{Performance (mAP) evaluation on bag-of-words (BoW) image retrieval. Vocabulary consisting of 1M visual words is learned on independent dataset, that is, when evaluating on Oxford5k, the vocabulary is learned with features of Paris6k and \emph{vice versa}. SV: spatial verification. QE: query expansion. The best results are highlighted in \textbf{bold}. All the descriptors except SIFT and HardNet++ were learned on \emph{Liberty} sequence of Brown dataset~\cite{Brown2007}. HardNet++ is trained on union of Brown and HPatches~\cite{hpatches2017} datasets.}
\label{tab:ox5kpar6k}
\setlength{\tabcolsep}{1mm}
\begin{tabular}{lc@{\hskip 7mm}ccc@{\hskip 7mm}cc}
\toprule
 & \multicolumn{3}{c}{Oxford5k} & \multicolumn{3}{c}{Paris6k} \\
 \cmidrule(r){2-4} \cmidrule(r){5-7}
Descriptor & BoW & +SV & +QE & BoW & +SV & +QE\\
\cmidrule(r){1-7}
TFeat-M*~\cite{TFeat2016} & 46.7 & 55.6 & 72.2 & 43.8 & 51.8 & 65.3\\
RootSIFT~\cite{RootSIFT2012} & 55.1	& 63.0	& 78.4 & 59.3 & 63.7 & 76.4 \\
L2Net+~\cite{L2Net2017} & \textbf{59.8} & 67.7 & 80.4 & \textbf{63.0} & 66.6 & 77.2 \\
HardNet & 59.0 & 67.6	& \textbf{83.2} & 61.4 & \textbf{67.4} & \textbf{77.5}\\
HardNet+ & \textbf{59.8} & \textbf{68.8} & 83.0 & 61.0 & 67.0 & \textbf{77.5}	\\
\cmidrule(r){1-7}
HardNet++ & \textbf{60.8} & \textbf{69.6} & \textbf{84.5} & \textbf{65.0} & \textbf{70.3} & \textbf{79.1} \\
\bottomrule
\end{tabular}
\end{table}

We evaluate our method and compare it against the related ones, on the practical application of image retrieval with local features.
Standard image retrieval datasets are used for the evaluation, \ie, Oxford5k~\cite{Philbin07} and Paris6k~\cite{Philbin08} datasets.
Both datasets contain a set of images (5062 for Oxford5k and 6300 for Paris6k) depicting 11 different landmarks together with distractors. 
For each of the 11 landmarks, there are 5 different query regions defined by a bounding box, constituting 55 query regions per dataset.
The performance is reported as mean average precision (mAP)~\cite{Philbin07}.

In the first experiment, for each image in the dataset, multi-scale Hessian-affine features~\cite{Mikolajczyk2005} are extracted.
Exactly the same features are described by ours and all related methods, each of them producing a 128-D descriptor per feature.
Then, k-means with approximate nearest neighbor~\cite{FLANN2009} is used to learn a 1 million visual vocabulary on an independent dataset, that is, when evaluated on Oxford5k, the vocabulary is learned with descriptors of Paris6k and \emph{vice versa}.
All descriptors of the testing dataset are assigned to the corresponding vocabulary, so finally, an image is represented by the histogram of visual word occurrences, \ie, the bag-of-words~(BoW)~\cite{Sivic-ICCV2003VideoGoogle} representation, and an inverted file is used for an efficient search.
Additionally, spatial verification~(SV)~\cite{Philbin07}, and standard query expansion~(QE)~\cite{Philbin08} are used to re-rank and refine the search results.
Comparison with the related work on patch description is presented in Table~\ref{tab:ox5kpar6k}. 
HardNet+ and L2Net+ perform comparably across both datasets and all settings, with slightly better performance of HardNet+ on average across all results (average mAP 69.5 vs. 69.1). 
RootSIFT, which was the best performing descriptor in image retrieval for a long time, falls behind with average mAP 66.0 across all results.
\begin{table}[tb]
\ra{1.0}
\centering
\caption{Performance (mAP) comparison with the state-of-the-art image retrieval with local features. Vocabulary is learned on independent dataset, that is, when evaluating on Oxford5k, the vocabulary is learned with features of Paris6k and \emph{vice versa}. All presented results are with spatial verification and query expansion. VS: vocabulary size. SA: single assignment. MA: multiple assignments. The best results are highlighted in \textbf{bold}.}
\label{tab:ox5kpar6kSOTA}
\setlength{\tabcolsep}{3mm}
\begin{tabular}{lccccc}
\toprule
 & & \multicolumn{2}{c}{Oxford5k} & \multicolumn{2}{c}{Paris6k} \\
 \cmidrule(r){3-4} \cmidrule(r){5-6}
Method & VS & SA & MA & SA & MA\\
\cmidrule(r){1-6}
SIFT--BoW~\cite{PerdochRetrieval2009} & 1M & 78.4 & 82.2 & -- & -- \\
SIFT--BoW-fVocab~\cite{Mikulik-IJCV2013FineVocab} & 16M & 74.0 & 84.9 & 73.6 & 82.4 \\
RootSIFT--HQE~\cite{Tolias-PR2014HQE} & 65k & 85.3 & 88.0 & 81.3 & 82.8 \\
HardNet++--HQE & 65k & \textbf{86.8} & \textbf{88.3} & \textbf{82.8} & \textbf{84.9} \\
\bottomrule
\end{tabular}
\end{table}
We also trained HardNet++ version -- with all available training data at the moment: the union of Brown and HPatches datasets, instead of just \emph{Liberty} sequence from Brown for the HardNet+. It shows the benefits of having more training data and is performing best for all setups.

Finally, we compare our descriptor with the state-of-the-art image retrieval approaches that use local features. 
For fairness, all methods presented in Table~\ref{tab:ox5kpar6kSOTA} use the same local feature detector as described before, learn the vocabulary on an independent dataset, and use spatial verification (SV) and query expansion (QE).
In our case (HardNet++--HQE), a visual vocabulary of 65k visual words is learned with an additional Hamming embedding (HE)~\cite{Jegou-IJCV2010HE} technique that further refines descriptor assignments with a 128 bit binary signature.
We follow the same procedure as RootSIFT--HQE~\cite{Tolias-PR2014HQE} method, by replacing RootSIFT with our learned HardNet++ descriptor.
Specifically, we use: (i) weighting of the votes as a decreasing function of the Hamming distance~\cite{Jegou-CVPR2009burstiness}; (ii) burstiness suppression~\cite{Jegou-CVPR2009burstiness}; (iii) multiple assignments of features to visual words~\cite{Philbin08,Jegou-PAMI2010accurate}; and (iv) QE with feature aggregation~\cite{Tolias-PR2014HQE}.
All parameters are set as in~\cite{Tolias-PR2014HQE}. The performance of our method is the best reported on both Oxford5k and Paris6k when learning the vocabulary on an independent dataset (mAP 89.1 was reported~\cite{RootSIFT2012} on Oxford5k by learning it on the same dataset comprising the relevant images), and using the same amount of features (mAP 89.4 was reported~\cite{Tolias-PR2014HQE} on Oxford5k when using twice as many local features, \ie, 22M compared to 12.5M used here).
\section{Exploring HardNet design choices~\cite{Pultar2020improving}}
\label{hardnet:pultar}
This Section contains various experiments with HardNet design choices, performed by Milan Pultar during his master studies~\cite{Pultar2020improving} under my supervision. The IMC benchmark, extensively used here, is described in Chapter~\ref{ch:benchmark}. AMOS patches dataset is described in~\cite{HardNetAMOS2019}.

\subsection{Architecture}
Architecture of a model is one of the paramount factors influencing the performance of the descriptor. A lot of work in the field of computer vision is focused on discovering architectures that achieve higher performance, are faster or more compact. However, there is a paucity of literature available describing new architectures for a local feature descriptor. Here we explore different architectural choices beyong original VGG-style L2Net architecture. 
\hiddensubsection{Automated Search}
\begin{table}[tb]
\centering
\caption[Evaluation of models found by the DARTS algorithm.]{Evaluation of models found by the DARTS algorithm, one model per row. Several trainings were run due to the recommendation in \cite{liu2018darts}. HPatches and AMOS Patches -- Mean average precision (mAP). IMC -- mean average accuracy (mAA) at 10\degree.}
\begin{tabular}{lccc}
\toprule
Architecture & HPatches & AMOS Patches & IMW PT \\
\midrule
DARTS 1 & 36.56 & 26.42 & 57.38 \\
DARTS 2 & 28.48 & 17.41 & 46.53 \\
DARTS 3 & 24.54 & 13.36 & 40.76 \\
\midrule
HardNet & \textbf{52.96} & \textbf{44.21} & \textbf{68.17} \\
\bottomrule
\end{tabular}
\label{tab:dartstab}
\end{table}

\begin{figure}[tb]
\centering
\begin{subfloat}[Reduction module.]
{\includegraphics[width=0.49\linewidth]{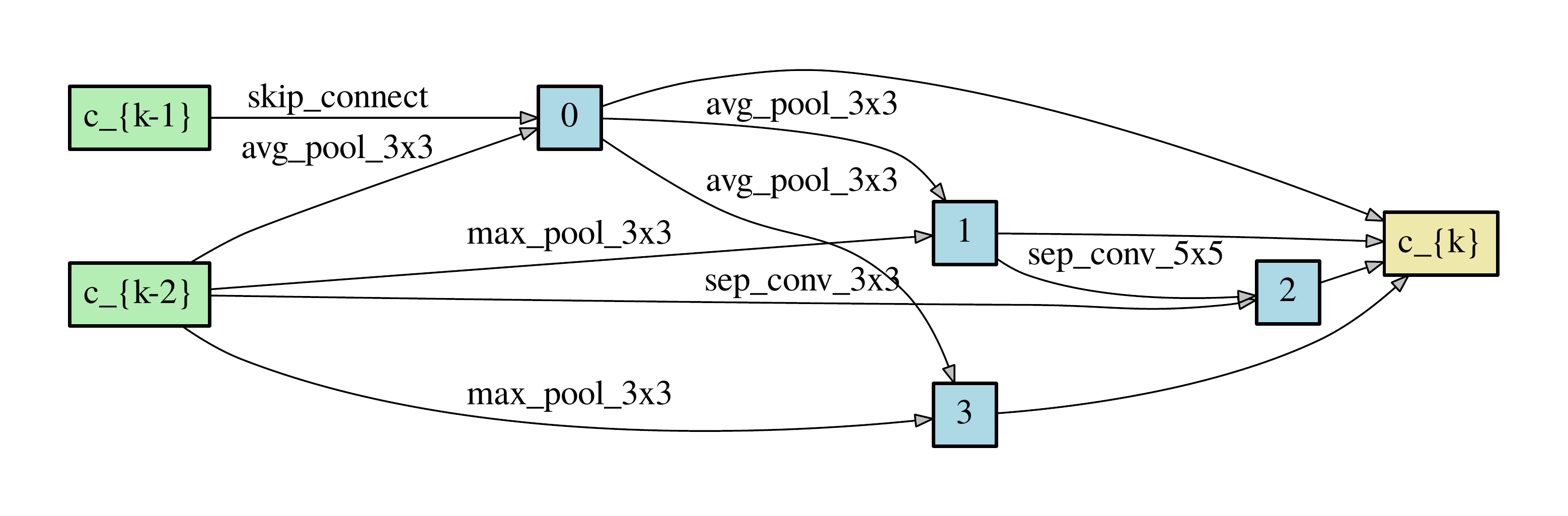}}
\end{subfloat}
\begin{subfloat}[Normal module.]
{\includegraphics[width=0.49\linewidth]{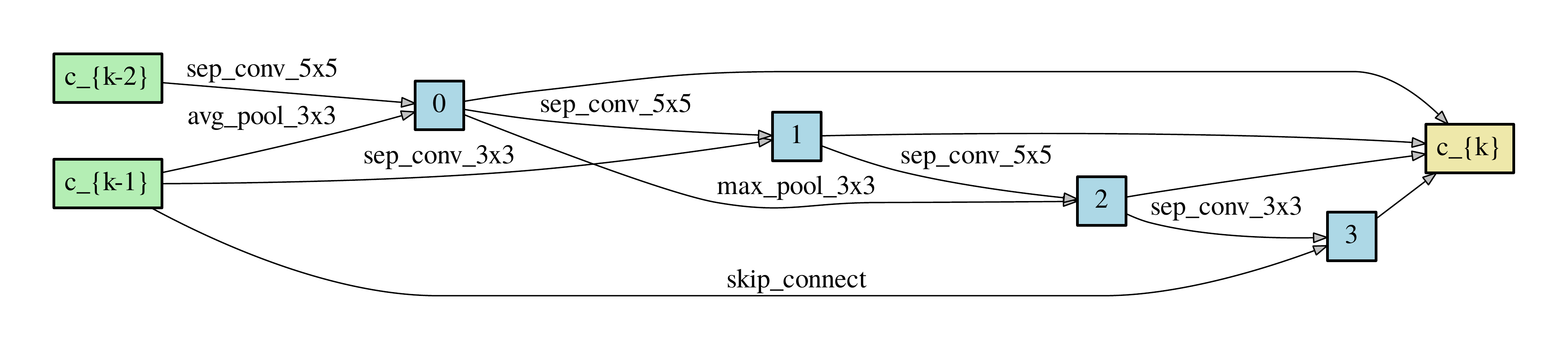}}
\end{subfloat}
\hfill\par
\caption[Modules found by the DARTS algorithm.]{Modules found by the DARTS algorithm visualized using the provided tool~\cite{liu2018darts}.}
\label{fig:darts}
\end{figure}

Differentiable Architecture Search (DARTS)~\cite{liu2018darts} is an algorithm which searches a space of architectures formulated in a continuous manner by using gradient descent. The bilevel optimization problem is expressed in~\cite{liu2018darts} as follows:
\begin{align}
&\min\limits_{\alpha} \mathcal{L}_{val}(w^*(\alpha), \alpha),\\
&\textrm{s.t.}\ w^*(\alpha)= \arg \min\limits_w \mathcal{L}_{train}(w, \alpha), \nonumber
\end{align}

where $\mathcal{L}_{train}$ and $\mathcal{L}_{val}$ is the loss function on the training and validation set, $\alpha$ is the architecture and $w$ are the associated weights. The search space is represented by a sequence of cells, which are composed of $k$ ordered nodes. The input to each node (called a connection or edge) is equal to the sum of its two predecessors. The output of a cell is defined as a concatenation of all its nodes. The edges between the nodes correspond to the operations that are to be learned. During the architecture search each of these are assigned parameters that are weighted using the softmax operator.

After the search procedure is finished, the operation corresponding to the highest weight is assigned to each edge. It is recommended to use a larger number of cells at this point due to smaller resource requirements.

In our experiments we use the publicly available implementation\footnote{Available at https://github.com/quark0/darts} and plug in our data loaders, so that Liberty and Notredame from UBC Phototour~\cite{PhotoTourism} is the training and validation dataset respectively. We run the architecture search for 40 epochs, 100 000 samples each, learning rate = 2.5, batch size = 64. Other settings are kept default.

Subsequent training is performed for 40 epochs, 500 000 tuples each, with batch size = 128 and learning rate = 0.025. In Table \ref{tab:dartstab} we list the results for several runs and compare them with HardNet. In Figure \ref{fig:darts} we present the found cells constituting the architecture DARTS 1. Reduction module is used in two positions in the architecture (1/3 and 2/3 of the length), all other cells are normal. The results of the DARTS models are significantly worse than those of HardNet. It might be, for example, due to the small batch size -- higher is not feasible due to GPU memory. Also, the method could be improved by changing the search space. We leave such modifications for future work.
\hiddensubsection{Manual Search}
\newcommand\cols{5}
\begin{table}[tb]
\centering
\caption[Variants of the HardNet architecture]{Variants of the HardNet. Each convolutional layer is followed by batch normalization and ReLU.}
\footnotesize
\begin{tabular}{ccccc}
\toprule
 HardNet & HardNet7x2 & HardNet8 & HardNet8x2 & HardNet9 \\
\midrule
 1.3M param & 5.3M param & 4.7M param & 19M param & 5.3 param \\
\midrule
\multicolumn{\cols}{c}{Block 1, spatial size 32x32}\\
\midrule
Conv 3x3x32 & Conv 3x3x64 & Conv 3x3x32 & Conv 3x3x64 & Conv 3x3x32 \\
Conv 3x3x32 & Conv 3x3x64 & Conv 3x3x32 & Conv 3x3x64 & Conv 3x3x32 \\
\midrule
\multicolumn{\cols}{c}{Block 2, spatial size 16x16}\\
\midrule
Conv 3x3x64/2 & Conv 3x3x128/2 & Conv 3x3x64/2 & Conv 3x3x128/2 & Conv 3x3x64/2 \\
Conv 3x3x64 & Conv 3x3x128 & Conv 3x3x64 & Conv 3x3x128 & Conv 3x3x64 \\
- & - & - & - & Conv 3x3x128 \\
\midrule
\multicolumn{\cols}{c}{Block 3, spatial size 8x8}\\
\midrule
Conv 3x3x128/2 & Conv 3x3x256/2 & Conv 3x3x128/2 & Conv 3x3x256/2 & Conv 3x3x128/2 \\
Conv 3x3x128 & Conv 3x3x256 & Conv 3x3x128 & Conv 3x3x256 & Conv 3x3x256 \\
- & - & Conv 3x3x256 & Conv 3x3x512 & Conv 3x3x256\\
\midrule
\multicolumn{\cols}{c}{Global pooling block}\\
\midrule
\multicolumn{\cols}{c}{Dropout(0.3)}\\
\midrule
Conv 8x8x128 & Conv 8x8x256 & Conv 8x8x256 & Conv 8x8x256 & Conv 8x8x256 \\
\midrule
\multicolumn{\cols}{c}{Flatten(), L2Norm()}\\
\bottomrule
\end{tabular}
\label{fig:archhard}
\end{table}

\begin{table}[tb]
\centering
\caption[Evaluation of the variants of the HardNet architecture]{Evaluation of the variants of the HardNet architecture. HPatches and AMOS Patches -- Mean average precision (mAP). IMC -- mean average accuracy (mAA) at 10\degree.}
\begin{tabular}{lccc}
\toprule
Architecture & HPatches & AMOS Patches & IMC \\
\midrule
HardNet & 52.96 & 44.21 & 68.17 \\
HardNet7x2 & \textbf{54.56} & \textbf{44.39} & 68.67 \\
HardNet8 & 53.67 & 43.77 & \textbf{69.43} \\
HardNet8x2 & 54.55 & 42.96 & 69.18 \\
HardNet9 & 54.21 & 42.85 & 69.34 \\
\bottomrule
\end{tabular}
\label{tab:manualhard}
\end{table}

\paragraph{VGG~\cite{VGGNet2015} Style}
HardNet is a VGG style network with blocks of convolutional layers of decreasing spatial size and an increasing number of channels. Here we introduce and evaluate several modifications to the architecture. With respect to the original version, HardNet7x2 contains two times more channels in each layer, HardNet8 has one convolutional layer added in the third block, and HardNet9 contains two more layers, one in the second and one om the third block. HardNet8x2 has two times more channels in each layer than HardNet8. See Table~\ref{fig:archhard} for a detailed architecture description.
In Table~\ref{tab:manualhard} we list the results for each architecture.
\renewcommand\cols{3}
\begin{table}[tb]
\centering
\caption[ResNet architectures]{ResNet architectures. Each convolutional layer is followed by batch normalization and ReLU.}
\footnotesize
\begin{tabular}{ccc}
\toprule
ResNet2 & ResNet3 & ResNet3x \\
\midrule
1.7 param & 2M params & 2.1M params \\
\midrule
\multicolumn{\cols}{c}{Block 1, spatial size 32x32}\\
\midrule
Conv 3x3x64 & Conv 3x3x32 & Conv 3x3x32 \\
Conv 3x3x64 & Conv 3x3x32 & Conv 3x3x64 \\
\midrule
\multicolumn{\cols}{c}{Block 2, spatial size 16x16}\\
\midrule
ResBlock2/2 32 & ResBlock3/2 16 & ResBlock3/2 32 \\
ResBlock2 64 & ResBlock3 32 & ResBlock3 64 \\
\midrule
\multicolumn{\cols}{c}{Block 3, spatial size 8x8}\\
\midrule
ResBlock2/2 64 & ResBlock3/2 64 & ResBlock3/2 64 \\
ResBlock2 128 & ResBlock3 128 & ResBlock3 128 \\
\midrule
\multicolumn{\cols}{c}{Global pooling block}\\
\midrule
\multicolumn{\cols}{c}{Dropout(0.3)}\\
\midrule
Conv 8x8x256 & Conv 8x8x256 & Conv 8x8x128 \\
\midrule
\multicolumn{\cols}{c}{Flatten(), L2Norm()}\\
\bottomrule
\end{tabular}
\label{fig:arch_res}
\end{table}

\begin{figure}[tb]
\centering
\begin{subfloat}[ResBlock2]{\includegraphics[width=0.15\linewidth]{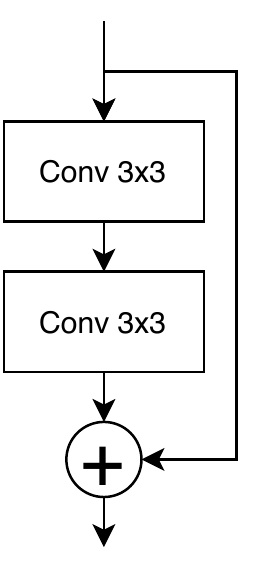}}
\end{subfloat}
\begin{subfloat}[ResBlock2/2]{\includegraphics[width=0.15\linewidth]{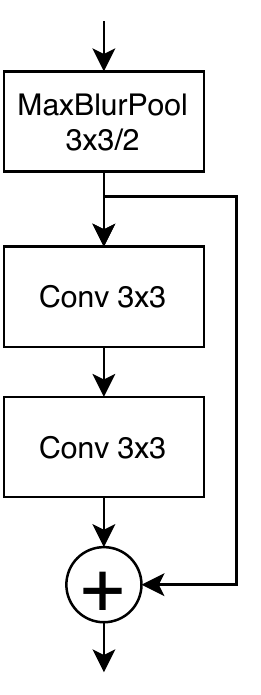}
}\end{subfloat}
\begin{subfloat}[ResBlock3]{\includegraphics[width=0.15\linewidth]{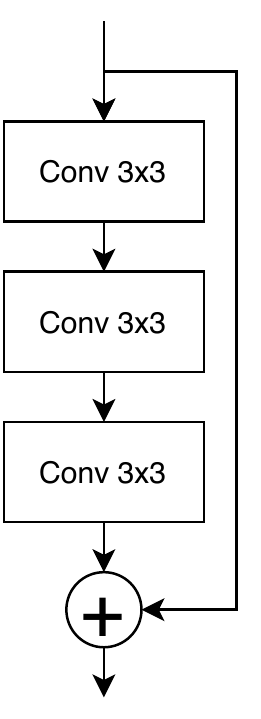}}
\end{subfloat}
\begin{subfloat}[ResBlock3/2]{\includegraphics[width=0.15\linewidth]{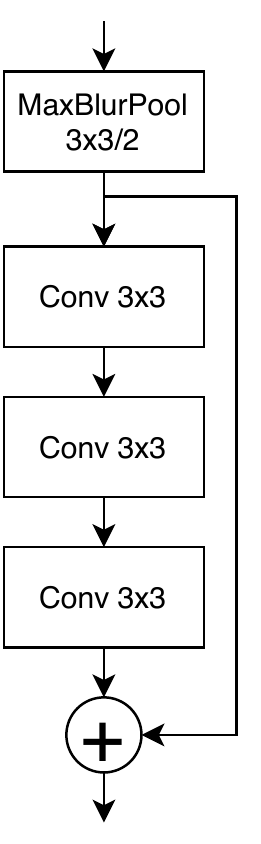}}
\end{subfloat}
\hfill\par
\caption{ResNet modules.}
\label{fig:resblock}
\end{figure}

\begin{table}[tb]
\centering
\caption[Evaluation of the ResNet architectures.]{Evaluation of the ResNet architectures together with the baseline HardNet. HPatches and AMOS Patches -- Mean average precision (mAP). IMC -- mean average accuracy (mAA) at 10\degree. }
\begin{tabular}{lccc}
\toprule
Architecture & HPatches & AMOS Patches & IMW PT \\
\midrule
ResNet2 & 52.63 & 40.57 & 68.31 \\
ResNet3 & 52.96 & 40.72 & 69.01 \\
ResNet3x & 52.25 & 38.50 & 68.27 \\
\midrule
HardNet8 & \textbf{53.67} & \textbf{43.77} & \textbf{69.43} \\
\bottomrule
\end{tabular}
\label{tab:evalres}
\end{table}

\paragraph{ResNet~\cite{He2016ResNet} Style}
Another popular architecture type is ResNet. It consists of blocks of convolutional layers called ResBlock. The output from each block is summed with its input, so the network learns an incremental change instead of a direct transformation. In Table~\ref{fig:arch_res} we list the architectures we tested. Evaluation of these models is shown in Table \ref{tab:evalres}. Each model was trained on the Liberty dataset for 20 epochs, 5 million samples each. The results are almost on par with those of HardNet models. However, in none of our experiments it was superior and it has a higher GPU memory requirement, so we can not advise to use this architecture type for the learning of a local feature descriptor.

\subsection{Batch size}
Batch size is an important factor in the setup of the training procedure. See in Figure~\ref{fig:bsize} the influence of the batch size on the performance of the trained descriptor. Notice that there is an increase in performance, measured by the mAP score on AMOS Patches and HPatches benchmarks, if we increase the batch size. It is likely that a higher batch size would be even more beneficial, but we are limited by the GPU memory. On a 32 GB machine, the maximum batch size is approximately 8192. The best performance on the IMC benchmark is achieved for the batch size around 4096. The IMC benchmark is described in detail in Chapter~\ref{ch:benchmark}. Interestingly, a further increase worsens the results on both HPatches and IMC.

\paragraph{Input Size}
\begin{table}[tb]
\centering
\caption[Evaluation of HardNet variants with different input size.]{Evaluation of HardNet variants with different input size. HPatches and AMOS Patches -- Mean average precision (mAP). IMC benchmark -- mean average accuracy (mAA) at 10\degree.}
\begin{tabular}{lccc}
\toprule
Model & HPatches & AMOS Patches & IMC \\
\midrule
HardNet8 & 52.37 & 42.42 & \textbf{69.06} \\
HardNet8, input size = 48 & 53.34 & 43.12 & 68.41 \\
HardNet8, input size = 64 & \textbf{54.19} & \textbf{43.98} & 68.25\\
\bottomrule
\end{tabular}
\label{tab:isize}
\end{table}

The input to the HardNet architecture is of size 32$\times$32 pixels. Due to the observation that enlarging the model architecture further from HardNet8 does not lead to better performance, we perform the following experiment. We make minor adjustments to the HardNet8 architecture so that it expects inputs of a bigger size. First, we change the stride to 2 in the penultimate layer, then the input size is 64$\times$64 pixels. If we also set the kernel size of the last convolutional layer to 6$\times$6 pixels, the input size is then 48$\times$48 pixels. We compare these three models in Table~\ref{tab:isize}. Notice the two opposite trends: bigger input size leads to better performance on HPatches and AMOS Patches, but such a model gives inferior results on the IMC benchmark.

\subsection{Margin value}
\begin{figure}[tb]
\begin{minipage}[c]{0.49\linewidth}
\begin{center}
\includegraphics[width=0.49\linewidth]{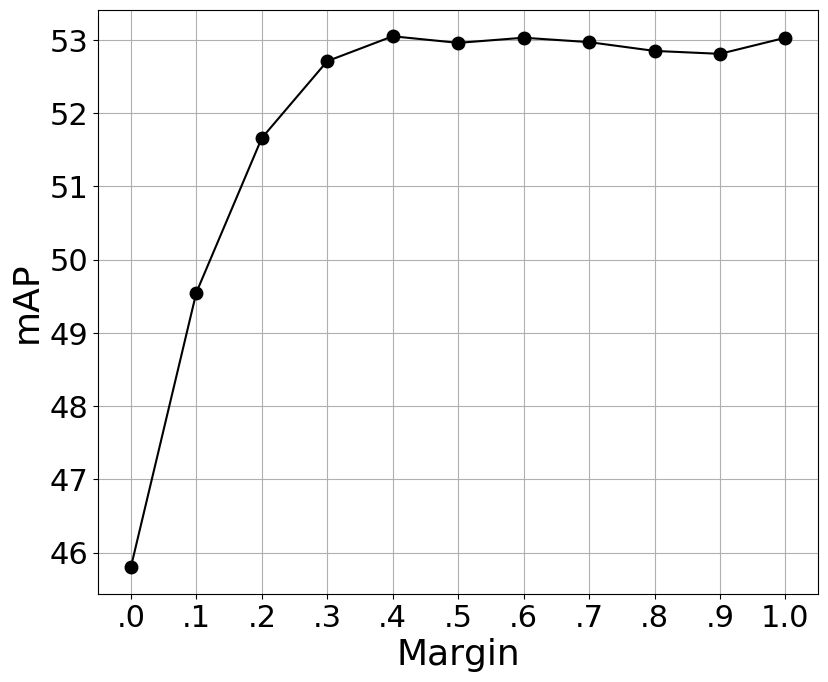}
\includegraphics[width=0.49\linewidth]{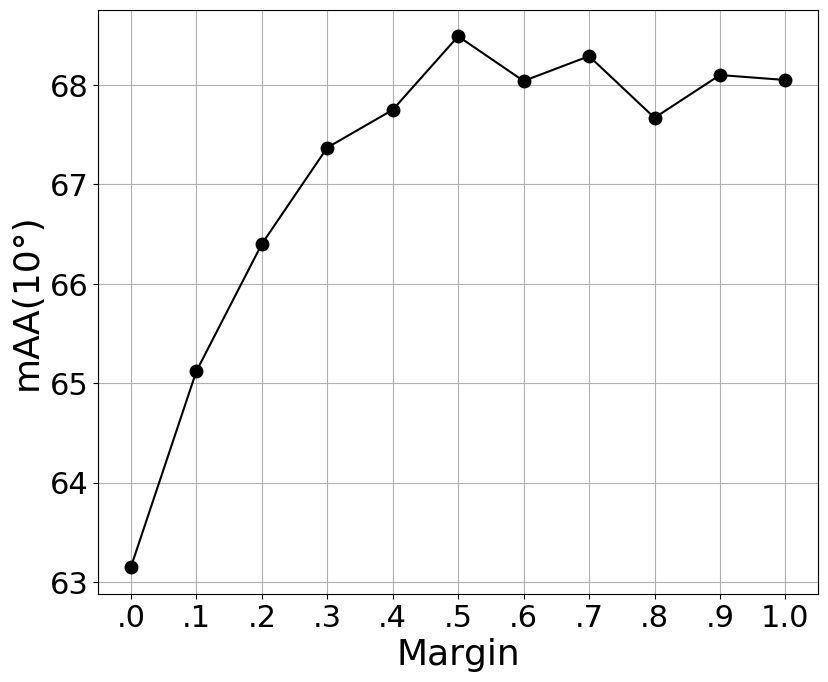}
\captionof{figure}{HardNet performance depending on the margin applied during training. Left: mAP on HPatches, right: mean average accuracy (mAA) with the threshold of 10\degree on IMC.}\label{fig:margin}
\end{center}
\end{minipage}
\begin{minipage}[c]{0.49\linewidth}
\begin{center}
\includegraphics[width=0.49\linewidth]{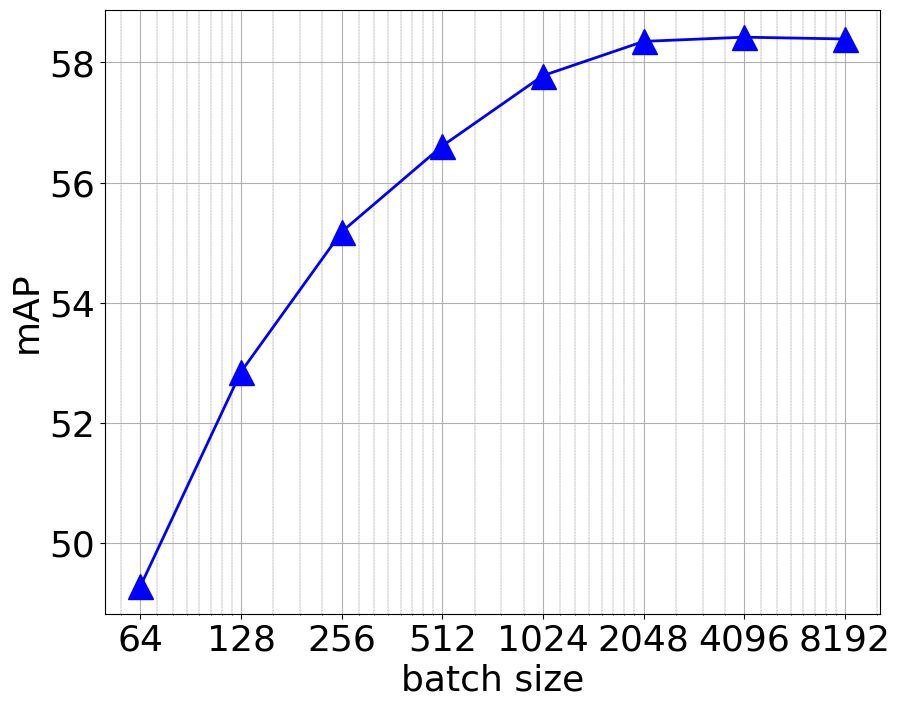}
\includegraphics[width=0.49\linewidth]{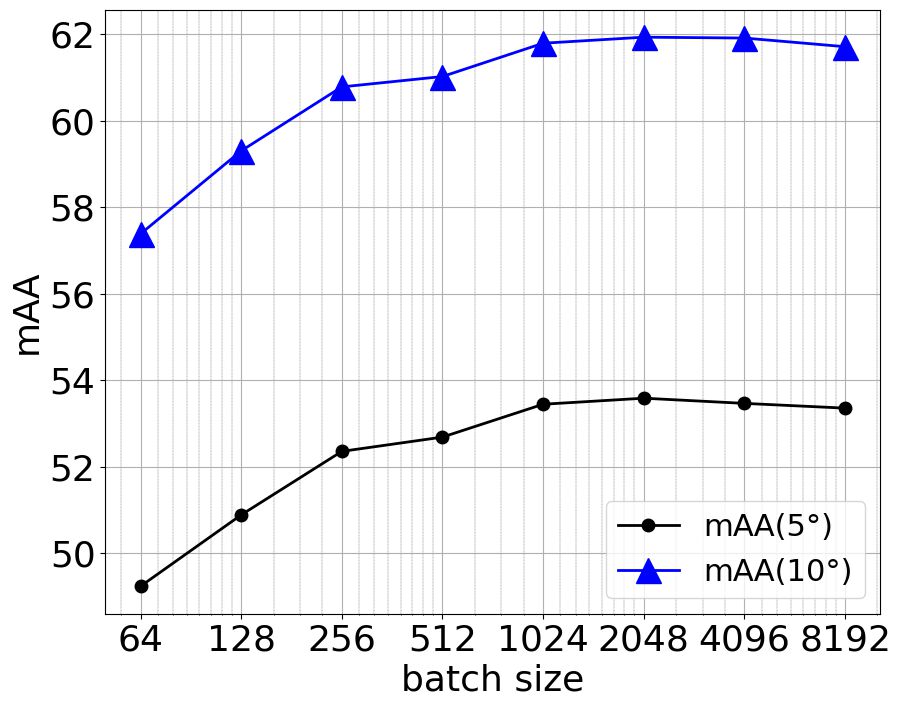}
\captionof{figure}{HardNet performance depending on the training batch size. Left: mAP on HPatches, right: mean average accuracy (mAA) with the threshold of 10\degree on IMC.}\label{fig:bsize}
\end{center}
\end{minipage}
\end{figure}

The purpose of this experiment is to determine the best value for the margin in the loss function. In Figure \ref{fig:margin} we can see that the performance roughly increases with the margin on both the CVPR and HPatches benchmarks. We can observe that the optimal value for the margin is around 0.5 in case of the IMC benchmark. In HPatches the difference in evaluation between margins above 0.4 is almost negligible.

\hiddensubsection{Final Pooling}

\begin{table}[htb]
\centering
\caption[Evaluation of final pooling variants.]{Final pooling variants. Modifications to the HardNet8 architecture by substituting the last convolutional layer are listed. The last row represents no modification. X$^y$ means the layer X is stacked $y$ times. We write n/c for models which failed to learn.HPatches and AMOS Patches -- Mean average precision (mAP). IMC -- mean average accuracy (mAA) at 10\degree.}
\begin{tabular}{lccc}
\toprule
Final Layer(s) & HPatches & AMOS Patches & IMC \\
\midrule
\multicolumn{4}{c}{Local Pooling}\\
\midrule
Conv 4x4x16/4 & 53.28 & 44.50 & 68.50 \\
Conv 3x3x16/2, p=1 & 52.76 & \textbf{44.78} & 68.36 \\
\midrule
\multicolumn{4}{c}{Increased Receptive Field}\\
\midrule
(Conv 3x3/2, p=1)$^1$ $\rightarrow$ Conv 4x4, p=0  & 52.31 & 42.00 & 68.29 \\
(Conv 3x3/2, p=1)$^2$ $\rightarrow$ Conv 2x2, p=0 & 50.25 & 39.14 & 67.62 \\
(Conv 3x3/1, p=0)$^3$ $\rightarrow$ Conv 2x2, p=0 & n/c & n/c & n/c \\
\midrule
\multicolumn{4}{c}{Non-learned}\\
\midrule
MaxPool 8x8 & 44.23 & 38.58& 67.11 \\
AvgPool 8x8  & n/c & n/c & n/c \\
\midrule
\multicolumn{4}{c}{Global Pooling (original)}\\
\midrule
Conv 8x8x256 & \textbf{53.67} & 43.77 & \textbf{69.43} \\
\bottomrule
\end{tabular}
\label{tab:finalpool}
\end{table}

The HardNet architecture uses global pooling implemented by a 8x8x128 convolution as the last layer in the model -- i.e. in fact a fully connected linear layer -- unlike SIFT, which uses local pooling, where the output vector is composed of 16 blocks based on the relative position in the input patch. In this section we compare different architectures and try several pooling approaches. All models are trained on the Liberty dataset for 20 epochs, 5 million samples each.

\paragraph{Local pooling}
In this experiment we try to mimic the mechanism of SIFT and replace the last convolutional layer in order to increase its stride so that the convolution is performed over separate blocks. In Table \ref{tab:finalpool} we can observe that the performance of these models is worse than of the baseline.
\paragraph{Increased Receptive Field}
We also test modifications to the HardNet8 architecture which increase its receptive field. In Table \ref{tab:finalpool} we can observe that such architectures do not outperform the baseline. Enlarging the architecture by 3 layers even fails to learn.

\paragraph{Non-learned Pooling}
Another possibility is to compute the maximum or average per input channel. MaxPool achieves a worse result and AvgPool fails to learn, see Table~\ref{tab:finalpool}.

\begin{figure}[htb]
\centering
\begin{subfloat}[HPatches full split.]
{\includegraphics[width=0.38\linewidth]{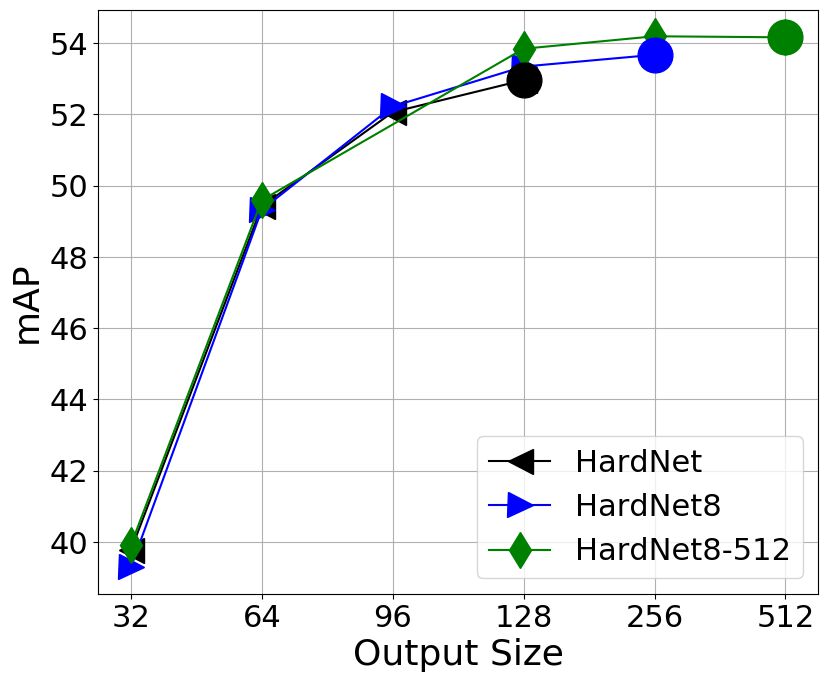}
}\end{subfloat}
\begin{subfloat}[IMC, validation set.]{
  \includegraphics[width=0.38\linewidth]{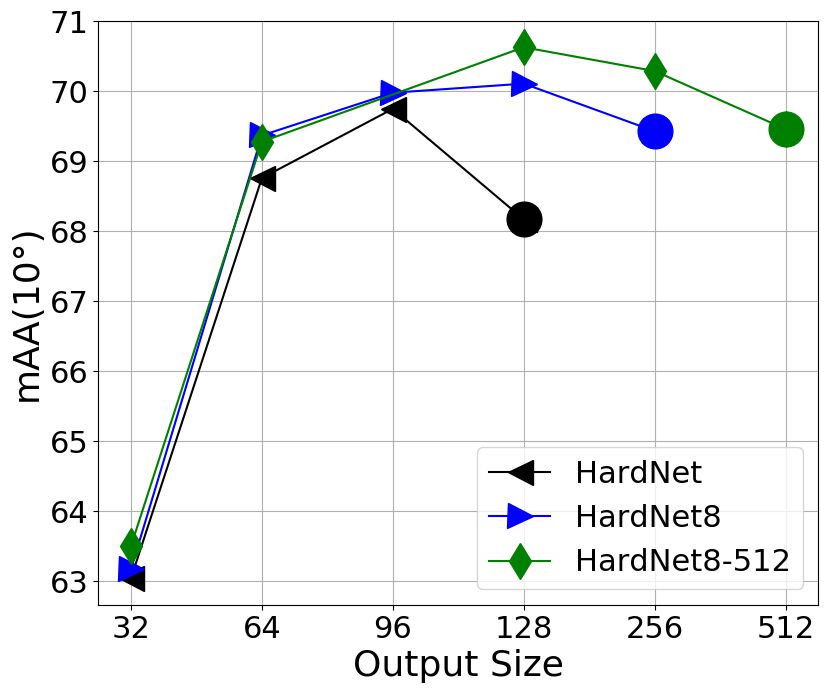}
}\end{subfloat}
\caption[Impact of the embeddings compression.]{Impact of the embeddings compression with PCA. HardNet8-512 denotes HardNet8 with the number of channels in the last layer changed to 512. HPatches and AMOS Patches -- Mean average precision (mAP). IMC -- mean average accuracy (mAA) at 10\degree. Original, non-compressed dimension is marked with a circle.}
\label{fig:pca}
\end{figure}

\subsection{Compression of Embeddings}
During the manual architecture search, we found that some of the larger models, e.g. HardNet8, achieve better performance. These architectures, however, also output a longer vector, which brings disadvantages such as higher memory usage and slower nearest-neighbour search. If we reduce the output size of HardNet8 to 128 to match the vector length of SIFT, we get inferior results - 68.75 mAA(10\degree) vs 69.43 mAA(10\degree) on the IMC benchmark. Another way to decrease the output size is to use a  dimensionality reduction technique. Here we use principal component analysis (PCA) to compress the feature embeddings. See in Figure~\ref{fig:pca} that dimensionality reduction is beneficial. We can observe that the best combination is HardNet8 with output size 512 followed by compression to size 128. In this experiment, we train on the Liberty dataset from UBC Phototour and we run PCA for the model outputs from the training data.

\subsection{Combining all together}
We evaluate two HardNet8 variants which use the improvements described above. The first, HardNet8-Univ, was trained on Liberty and AMOS Patches~\cite{HardNetAMOS2019} datasets for 20 epochs, 5 million samples each, batch size~=~3072.

The second model, HardNet8-PT, has 512 output channels and was trained on Liberty, Notredame and Colosseum Exterior~\cite{IMW2020} datasets for 40 epochs, 5 million samples each, batch size~=~9000. The embeddings from the model are compressed by PCA -- which is fitted on Liberty -- to dimensionality of 128.

Evaluation is performed on the test set of IMC -- the previous experiments were evaluated on the validation set. HPatches and AMOS Patches comprise of a single test split which is used both in the experiments and final evaluation.

Results are shown in Table \ref{tab:pultar-results}. HardNet8-Univ  works reasonably well on a wide range of conditions, be it viewpoint or illumination changes, and improves state-of-the-art on standard benchmarks. It is only outperformed by GeoDesc in the viewpoint split of HPatches. HardNet8-PT further improves the performance on the IMC benchmark in the Stereo 8k task.

Apart from benchmarking, we also conjecture that significantly better results can be achieved for a real-world task if HardNet8 is retrained on a specific combination of datasets, suitable for the use case -- e.g. AMOS Patches for illumination changes with no viewpoint change, while Liberty, Notredame, Colosseum Exterior and others for purely viewpoint changes.
\begin{table}[htb]
\noindent %
\centering
\caption[Final evaluation of the proposed descriptors.]{Final evaluation of the proposed HardNet8 descriptors. HPatches and AMOS Patches -- Mean average precision (mAP). IMC -- mean average accuracy (mAA) at 10\degree.}
\makebox[\textwidth]{ %
\begin{tabular}{lcccccc}
\toprule
Model & Trained on &\multicolumn{3}{c}{HPatches subset} & AMOS &IMC  \\
& &illum & view & full &Patches& Stereo 8k  \\
\midrule
SIFT &-- & 23.33 & 28.88 & 26.15 & 37.08 & 45.84\\
HardNet & Liberty & 49.86&	55.63&	52.79&	43.32&55.43 \\
SOSNet & Liberty &50.66&	56.73&	53.75 & 43.26&55.87 \\
HardNetPS & PS Dataset & 48.55&	67.43&	58.16 & 31.83 & 50.51 \\
GeoDesc & GL3d~\cite{shen2018mirror} & 50.52 & \textbf{67.48} & 59.15 & 25.50&51.11 \\
\midrule
HardNet8-Univ&AMOS+Liberty &\textbf{58.20}&63.60&\textbf{60.95}&\textbf{47.20}&56.22 \\
HardNet8-PT& Lib+Coloss+Notre & 51.04&55.81&53.46 & 45.01&\bf{57.58}  \\
\bottomrule
\end{tabular}
} %
\label{tab:pultar-results}
\end{table}

\clearpage
\section{Summary}
\label{hardnet:conclusions}
We proposed a novel loss function for learning a local image descriptor that relies on the hard negative mining within a mini-batch and the maximization of the distance between the closest positive and closest negative patches. The proposed sampling strategy outperforms classical hard-negative mining and random sampling for softmin, triplet margin, and contrastive losses. We have also evaluated a series of design choices influencing the performance of the HardNet. 

The resulting descriptor is compact -- it has the same dimensionality as SIFT (128), it shows state-of-art performance on standard matching, patch verification and retrieval benchmarks, and it is fast to compute on a GPU. The training source code and the trained convnets are available at \url{https://github.com/DagnyT/hardnet}.

\chapter[Local affine features]{Local affine features  \\ 
\includegraphics[width=0.95\linewidth]{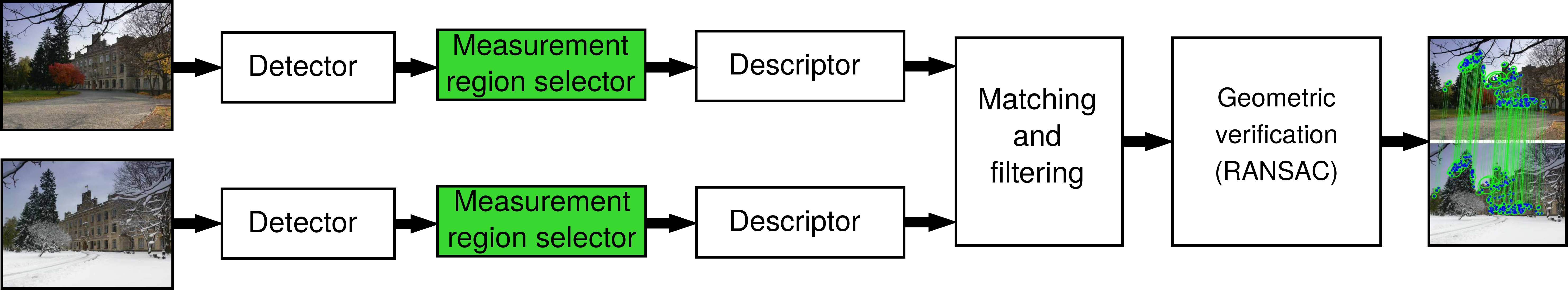}}
\label{ch:local-affine-features}

This Chapter makes four contributions towards the robust estimation of the local feature affine shape. 
First, we experimentally show that the geometric repeatability of a local feature is not a sufficient condition for successful matching. The learning of the affine shape increases the number of corrected matches if it steers the estimators towards discriminative regions and therefore must involve the optimization of a descriptor-related loss. Second, we propose a novel loss function for descriptor-based registration and learning, named the {\it hard negative-constant loss}. It combines the advantages of the triplet and contrastive positive losses. Third, we propose a method for learning the affine shape, orientation, and potentially other parameters related to  geometric and appearance properties of local features. The learning method  does not require a precise ground truth, which reduces the need for manual annotation.  Finally, the learned AffNet itself significantly outperforms prior methods for affine shape estimation and improves the state-of-the-art in BoW-based image retrieval by a large margin. Importantly, unlike the de facto standard~\cite{Baumberg2000}, AffNet does not significantly reduce the number of detected features, it is thus suitable even for pipelines where affine invariance is needed only occasionally.

\section{Introduction}

\paragraph{Local feature detector} finds "image patterns which differ from its immediate neighborhoods"~\cite{Tuytelaars2008}. Local detectors, compared to dense image sampling, serve two purposes: reduce the amount of computation needed for descriptor matching and increase the robustness to occlusions. Similar to the local descriptors, detectors can be divided into handcrafted-vs-learned groups and float-vs-binary kinds. 

The earliest detectors are still in use: the Hessian~\cite{Hessian78} blob detector and the Harris~\cite{Harris88} corner detector. Both detectors are based on image gradients. Later, they were extended to multi-scale~\cite{Lindeberg1993,Lindeberg1998,Multiscale2000} and affine-covariant~\cite{Baumberg2000,HesHarAff2004} versions. Other popular blob detector is Difference-of-Gaussians(DoG), proposed in SIFT paper~\cite{SIFT2004}. 

Binary detectors that rely on intensity comparisons for finding corner-like structures started with the SUSAN~\cite{Smith1997} detector. It was then accelerated by the learned order of comparisons for fast rejection -- FAST~\cite{FAST2006} and improved by adding the orientation estimation -- ORB~\cite{ORB2011}. Binary detectors for other than corner structures also exist, e.g. for the detection of saddle points~\cite{Saddle2016, saddle2019}.

There are other detectors that do not belong to the abovementioned groups. Most popular of them are the following. MSER~\cite{MSER2002} uses a segmentation algorithm for detecting regions with boundaries sharing similar pixel intensity. FOCI~\cite{Zitnick2011} looks for normalized intensity edge focal points and WASH~\cite{Varytimidis2012} is based on the Delaunay triangulation of the edge map. 

The popular deep learning-based detectors are the following. 
Yi~\etal\cite{OriNet2016} proposed to learn feature orientation by minimizing descriptor distance between positive patches which correspond to the same point on the 3D surface. This allows to avoid hand-picking a "canonical" orientation, thus learning the one which is most suited for descriptor matching.
Yi~\etal\cite{LIFT2016} proposed a multistage framework for learning the descriptor, orientation and translation-covariant detector. The detector was learned by maximizing the intersection-over-union and the reprojection error between the corresponding regions. Lenc and Vedaldi~\cite{Lenc2016} introduced the ``covariant constraint'' for learning various types of local feature detectors. Zhang~et.al.~\cite{NewTilde2017} proposed to ``anchor'' the detected features to some predefined features with known good discriminability like TILDE~\cite{TIlde2015}. 
Savinov~et.al.~\cite{QuadNets2017} proposed a ranking approach for unsupervised learning of a feature detector. Choy~et.al.~\cite{UCN2016} trained a ``Universal correspondence network'' (UCN) for a direct correspondence estimation with contrastive loss on a patch descriptor distance, avoiding the detection stage. 
Superpoint detector and descriptor~\cite{SuperPoint2017} share a similar idea for training the descriptor, while the detector is trained in a supervised way to find corners and line junctions. R2D2~\cite{R2D22019} learn a detector to optimize both the repeatability and ``reliablity''. Reliability is proposed and called matchability in the AffNet paper~\cite{AffNet2018}, which is described in Section~\ref{affnet:affnet-main}.

\section{Keypoints are not just points}
\label{sec:keypoints-are-not-just-points}
Local feature detectors are often referred as keypoint detectors and wide baseline stereo matching often is perceived as establishing (key-)point correspondences between images. While this might be true for some local features like SuperPoint~\cite{SuperPoint2017}, typically it is more than that.

Specifically, detectors like DoG~\cite{SIFT2004}, Harris~\cite{Harris88}, Hessian~\cite{Hessian78}, KeyNet~\cite{KeyNet2019}, ~ORB\cite{ORB2011}, and many others run on scale-space provide at least 3 parameters: $x$, $y$, and scale.

Most of the local descriptors -- SIFT~\cite{SIFT2004}, HardNet~\cite{HardNet2017} and so on -- are not rotation invariant, so the patch orientation often has to be estimated anyway to match reliably. While this stage can be avoided by designing the descriptor in the rotation-invariant way~\cite{lbp2009, sGLOH2}, this often requires a complex matching function~\cite{RIFT2005, sGLOH2} and results in less discrimination ability~\cite{bellavia2020NewAboutSIFT}.

The rotation estimation is done by various methods: corner center of mass (ORB~\cite{ORB2011}, dominant gradient orientation (SIFT)~\cite{SIFT2004} or by some learned estimator (OriNets~\cite{OriNet2016, AffNet2018}). Sometimes it is possible to rely on the smartphone`s IMU or photographer, and assume that the images are upright\cite{PerdochRetrieval2009}.

\begin{figure*}
\centering
\includegraphics[width=0.45\linewidth]{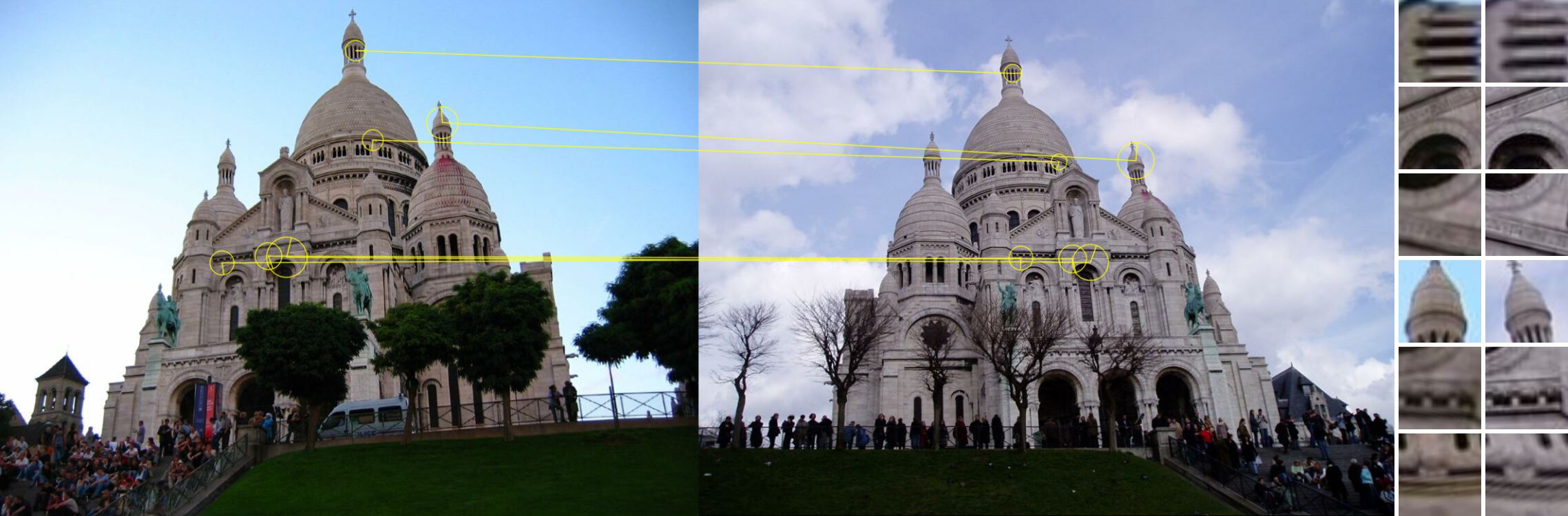}
\includegraphics[width=0.45\linewidth]{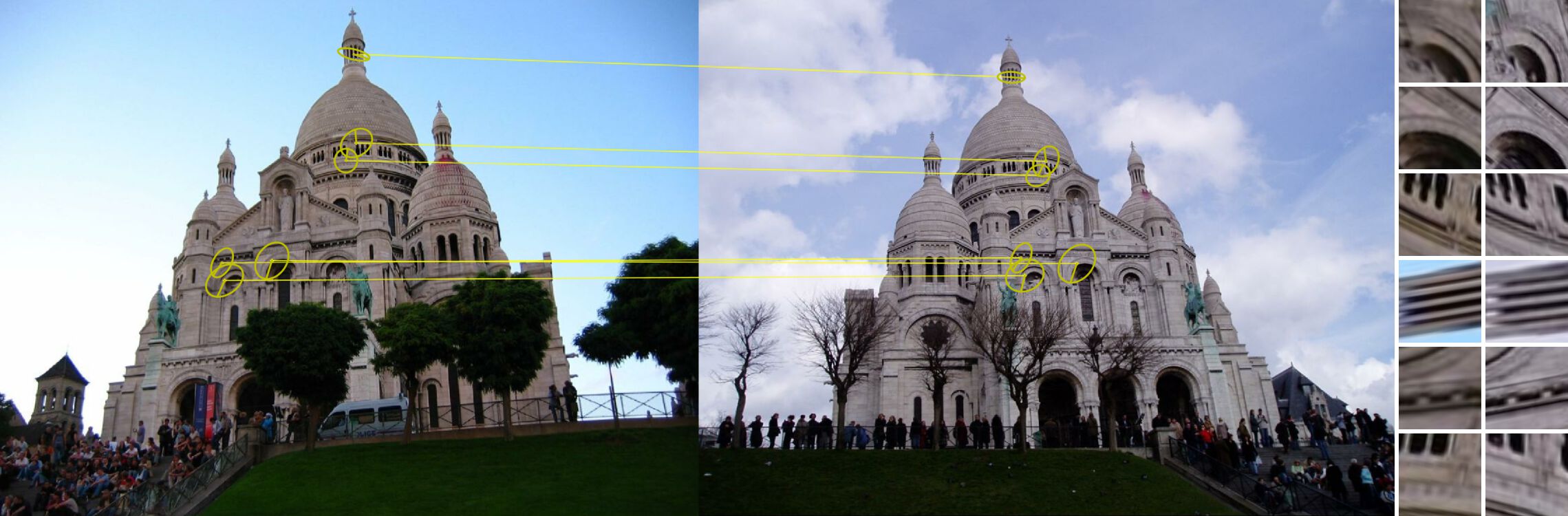}
\caption{Selected matching SIFT (left) and SIFT-AffNet(right) keypoints and corresponding patches. One can see that not only patch centers correspond to each other, but also other pixels, although less precise.}
\label{fig:matching-patches-similarity}
\end{figure*}

Thus, we can assume that if the local descriptors match, this means the local feature scale and orientation also match at least approximately, as illustrated in Figure~\ref{fig:matching-patches-similarity}. Possible exceptions are cases when the patch is symmetrical and the orientation is ambiguous up to some symmetry group.

\begin{figure}
\centering
\includegraphics[width=0.95\linewidth]{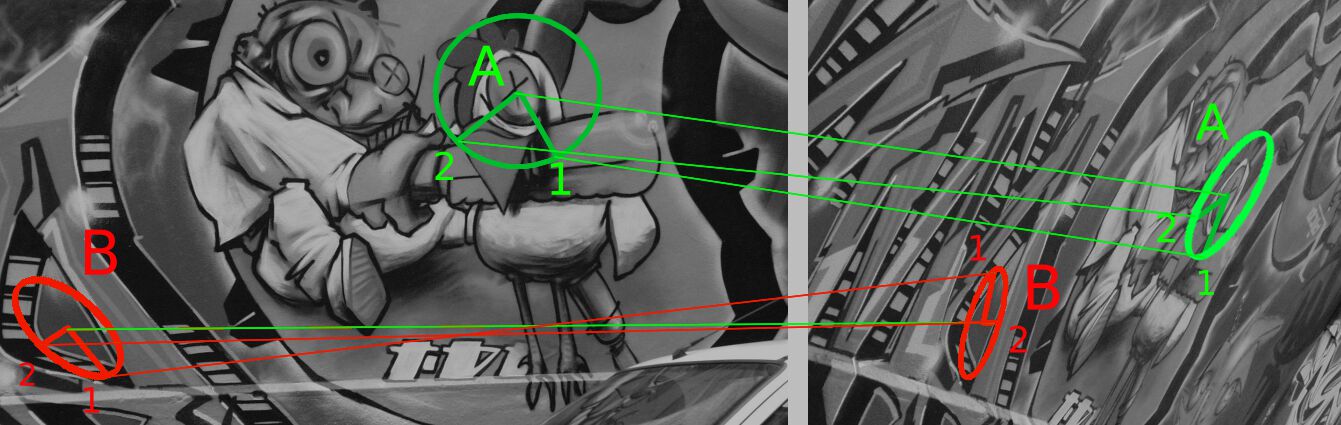}
\caption{Local affine correspondences. While centers of both regions A and B are correct point matches, only A is a correct affine correspondence.}
\label{fig:laf-check-ill}
\end{figure}

In addition, one could assume that we observe the patch not from the fronto-parallel position and try to estimate the local normal, or, more precisely, the affine shape of the feature point, modeling it as an ellipse instead of a circle, as illustrated in Figure~\ref{fig:matching-patches-similarity}. One could also think of affine shape estimation as finding the camera position, from where the patch is seen in some "canonical" view. This gives us 3 point correspondences from a single local feature match, see an example in Figure~\ref{fig:laf-check-ill}.

\section{Benefits of local affine features}
\label{benefits-of-local-affine-features}
\subsection{Making descriptor job easier}
\label{making-descriptor-job-easier}

The most straightforward benefit of using local affine features is that they increase the repeatability of the detector and potentially reduce the appearance changes of a local patch caused by viewpoint difference~\cite{Mikolajczyk2005}. This makes possible matching more challenging image pairs.

The practice is a little bit more complicated. While the affine-covariance helps the matching under the presence of the large viewpoint change, see Chapter~\ref{ch:mods}, it reduces the local patch discriminativity for the simpler cases. In other words, one needs to be as much invariant or covariant, as needed, but not more. For example, affine-convariant detector helps to match the circle under different angles, but fails when one needs to distinguish between a circle and an ellipse.  

Our benchmark~\cite{IMW2020}, see in Chapter~\ref{ch:benchmark}, which measures the accuracy of the output fundamental matrix, shows that the benefit of using affine instead of similarity-covariant features for the phototourism data might be quite unstable.  On the one hand, using AffNet consistently improves multiview camera pose accuracy estimation by 3-5\% relatively. On the other hand, in the large number of features setup, DoG-HardNet performs better in stereo than DoG-AffNet-HardNet.

Therefore, if the benefit of local features would be to only improve the descriptor extraction stage for mild viewpoint changes, it might be arguably not worth it. Luckily, there are more benefits, which are more pronounced.

\subsection{Making RANSAC job easier}
\label{making-ransac-job-easier}

Let us recall how RANSAC\cite{RANSAC1981} works.

\begin{enumerate}

\item
  Randomly sample a minimally required number of tentative
  correspondences to fit the geometrical model of the scene: 4 for
  homography, 7 for epipolar geometry, etc. and estimate the model.
\item
  Calculate "support": other correspondeces, which are consistent with the model.
\item
  Repeat steps (1), (2) and output the model which is supported with the
  most of correspondences. If you were lucky and have sampled all-inlier
  sample, meaning that all correspondences used to estimate the model
  were correct, you would have a correct model.
\end{enumerate}

Modern RANSACs\cite{gcransac2018,magsac2019} are more complicated than we have just described, but the principle is the same. The most important part is the sampling and it is sensitive to the inlier ratio \(\nu\) - the percentage of the correct
correspondences in the set. Let us denote the minimal number of
correspondences required to estimate the model as \textbf{m}. To recover
the correct model with the confidence \textbf{p} one needs to sample the number of correspondences, which is described by formula:

\begin{equation}
N = \frac{\log{(1 - p)}}{\log{(1 - \nu^{m})}}
\end{equation}

\begin{figure}
\centering
\includegraphics[width=0.95\linewidth]{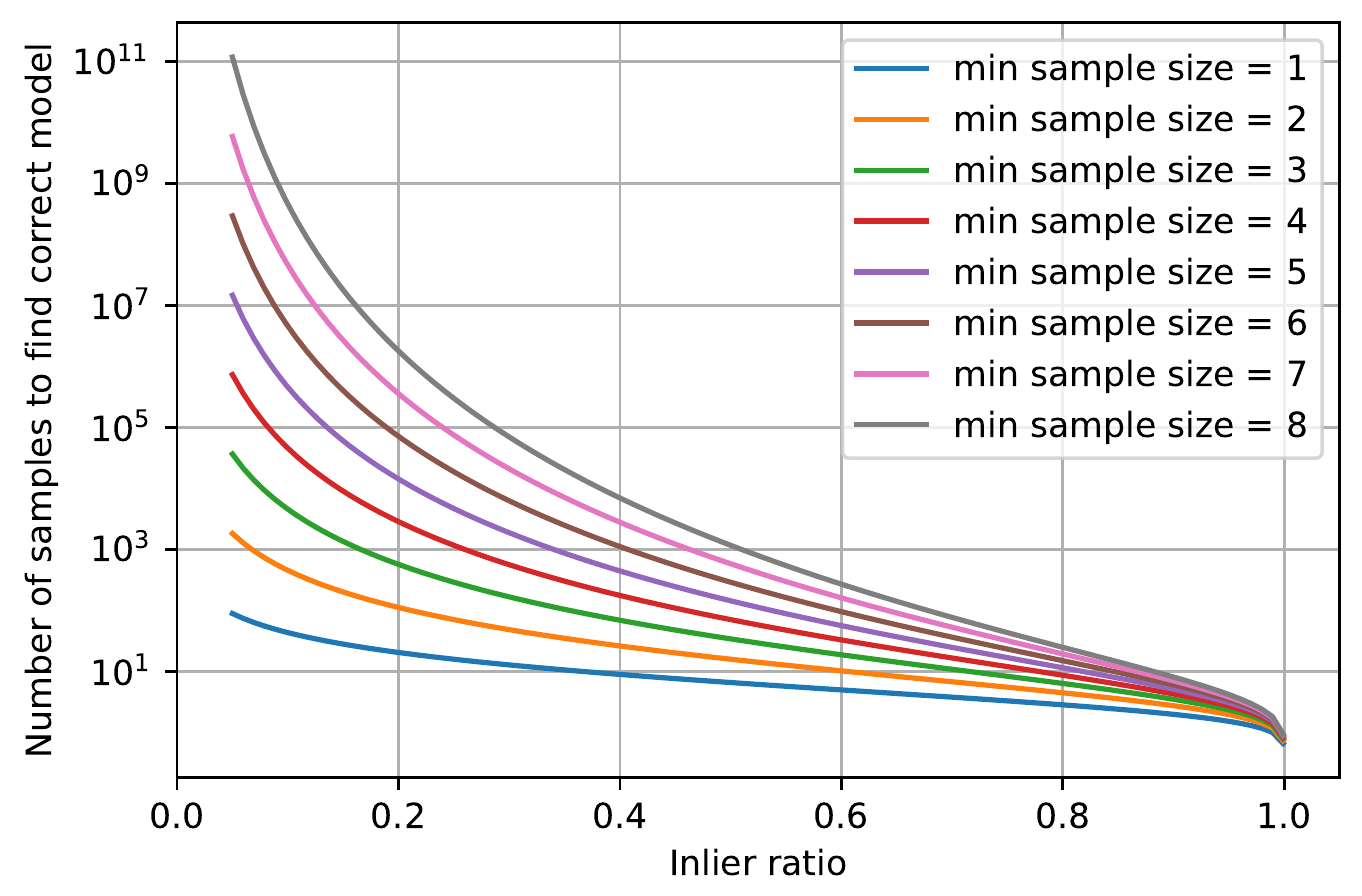}
\caption{Number of samples RANSAC needs to draw to find the correct model with the given probability as function of inlier rate, when confidence rate is 99\%.}
\label{fig:ransac-min-samples}
\end{figure}

Lets plot how the number of required samples changes with the inlier ratio for confidence equal 99\%, see Figure~\ref{fig:ransac-min-samples}. Note the log scale on Y axis. Reducing the minimal sample size required for the model estimation even by 1 saves an order of magnitude
of computation. In reality, the benefit is smaller, as modern RANSACs like GC-RANSAC\cite{gcransac2018} and MAGSAC\cite{magsac2019} could
estimate the correct model from the sample containing outliers, but it is still substantial, especially for low inlier rate cases. %

\subsubsection{Image retrieval}
\label{image-retrieval}
The ideal case would be to estimate a model from just a single sample
and that is exactly what is done in the spatial reranking paper by Philbin~\etal\cite{Philbin07}.

Specifically, they are solving a particular object retrieval problem:
given an image containing some object, return all images from the
database, which also contain the same object.

The initial list of images is formed by the descriptor distance and then
is reranked. The authors propose to approximate the perspective change
between two images as an affine image transformation, and count the number
of feature points, which are reprojected inside the second image. This number produces a better ranking that the original short-list.

\begin{figure}
\centering
\includegraphics[width=0.95\linewidth]{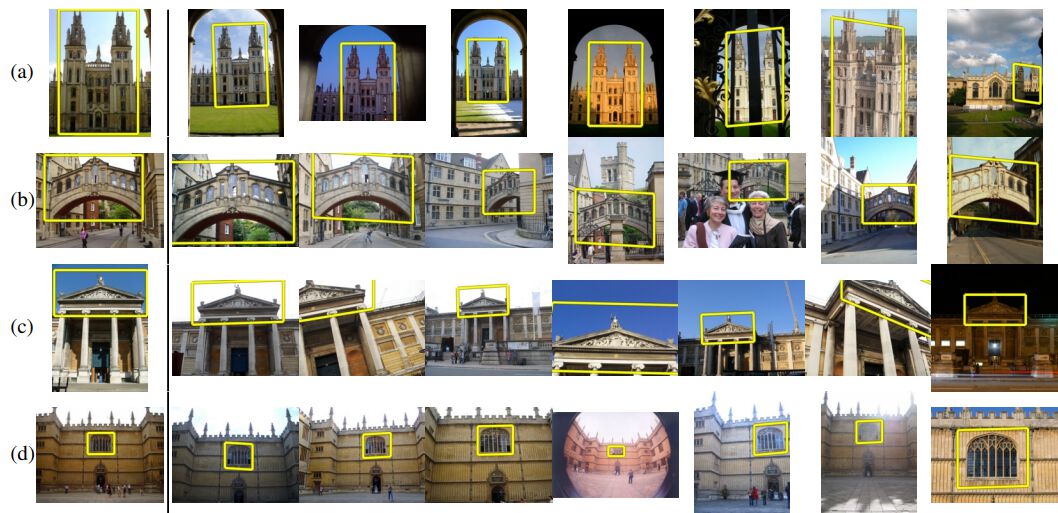}
\caption{From~\cite{Philbin07}. Affine local features are used for the fast spatial verification.}
\label{fig:spatial-verification}
\end{figure}

\subsubsection{Wide baseline stereo}
\label{back-to-wide-baseline-stereo}

While working for spatial re-ranking, 3-degrees of freedom camera model is too rough for the wide baseline stereo. Yet, going from 4 point
correspondences (PC) to 2 affine correspondences (AC) for homogaphy and from 7 PC to 3 AC for the fundamental matrix would be substantial benefit
anyway for the robust model estimation.

Various variants of RANSAC working with local affine features were proposed in the last 15 years: Perdoch~\etal\cite{perd2006epipolar}, Pritts
et.al.~\cite{PrittsRANSAC2013}, Barath and Kukelova~\cite{Barath2019ICCV}, Rodríguez~\etal\cite{RANSACAffine2020}.

Finally, the systematic study of using is presented by Barath~\etal\cite{barath2020making}. Authors show that if used naively,
the affine correspondence may lead to worse results, because they are more noisy than point correspondences. However, there is a bag of tricks presented
in the paper, which allow to solve the noise issue and make the affine RANSAC working in practice, resulting in orders of magnitude faster
computation.

Moreover, for special cases like autonomous driving, where the motion is mostly horizontal, one could even use 2 affine correspondes for both
motion estimation and consistency check, significantly improving the efficiency of the outliers removal compared to the standard RANSAC
loop~\cite{guan2020relative}.

Besides the special case considerations, additional constraints can also come from running other algorithms, like monocular depth estimation.
Such a constraint could reduce the required number of matches from two affine correspondences to a single one for the calibrated camera case~\cite{OneACMonoDepth2020}.

\subsection{Application-specific
benefits}
\label{application-specific-benefits}

Besides the wide baseline stereo, local affine features and correspondences have other applications. Here are some examples. %

\subsubsection{Image rectification}
\label{image-rectification}

Instead of matching local features between two images, one might match them within a single image. Why would someone do it? This allows finding
a repeated pattern: think about windows, doors, and so on. Typically they have the same physical size, therefore the difference in local features
around them could tell us about the geometry of the scene and lens distortion. 
This is the idea of the series of works by Pritts\etal~\cite{RepPat2018,RepPatRect2018a,pritts2019minimal,pritts2020}.

\begin{figure}
\centering
\includegraphics[width=0.98\linewidth]{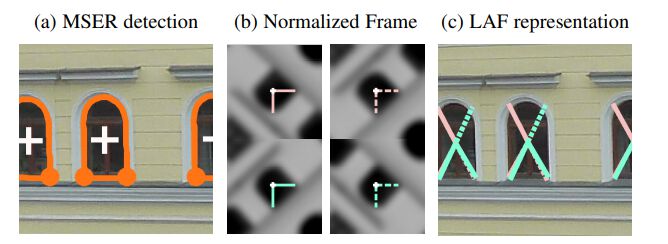}\\
\includegraphics[width=0.49\linewidth]{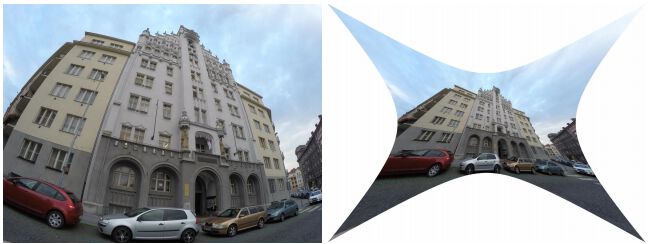}
\includegraphics[width=0.47\linewidth]{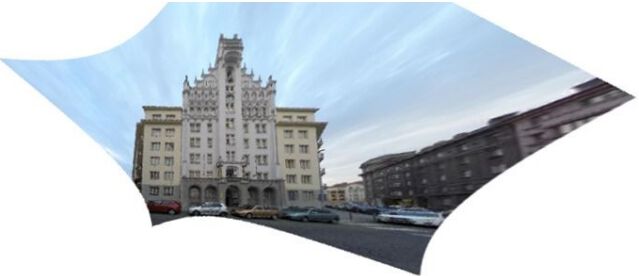}

\caption{Local affine features can be used for image rectification. First, repeated patterns are detected in the image, then their geometry is used for distortion estimation. From~\cite{pritts2019minimal}}
\label{fig:affine-rectification}
\end{figure}

\subsubsection{Surface normals
estimation}
\label{surface-normals-estimation}
Ivan Eichhardt and Levente Hajder have a series of works exploiting the local affine correspondences for surface normals estimation~\cite{SurfaceNormals2019}
\begin{figure}
\centering
\includegraphics[width=0.95\linewidth]{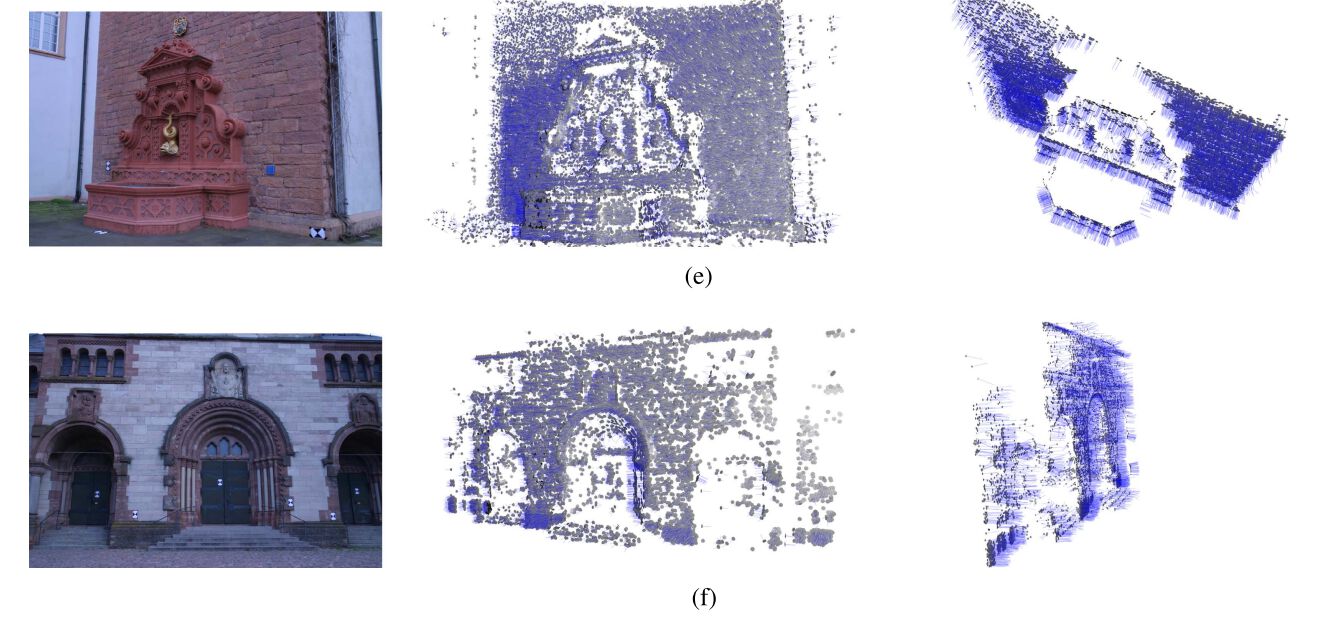}
\caption{From~\cite{SurfaceNormals2019}. Local affine features are used for estimation of surface normals (blue).}
\label{fig:surface-normal}
\end{figure}

\subsection{Related work}
The area of learning local features has been active recently, but the attention has focused dominantly on  learning descriptors~\cite{DeepComp2015,MatchNet2015,TFeat2016,L2Net2017,HardNet2017,SpreadDesc2017,ExemplarNet2016} and translation-covariant detectors~\cite{TIlde2015,NewTilde2017,Lenc2016,QuadNets2017}. The authors are not aware of any recent work on learning or improvement of local feature affine shape estimation. 
The most closely related work is thus the following.

Hartmann~\etal~\cite{Matchability2014} train random forest classifier for predicting feature matchability based on a local descriptor. "Bad" points are discarded, thus speeding up the matching process in a 3D reconstruction  pipeline. Yi~\etal\cite{OriNet2016} proposed to learn feature orientation by minimizing descriptor distance between positive patches, i.e. those corresponding to the same point on the 3D surface. This allows to avoid hand-picking a "canonical" orientation, thus learning the one which is the most suitable for descriptor matching.  We have observed that direct application of the method~\cite{OriNet2016}  for affine shape estimation leads to learning degenerate shapes collapsed to single line. 
Yi~\etal\cite{LIFT2016} proposed a multi-stage framework for learning  the descriptor, orientation and translation-covariant detector. The detector was trained by maximizing the intersection-over-union and the reprojection error between corresponding regions.

Lenc and Vedaldi~\cite{Lenc2016} introduced the ``covariant constraint'' for learning various types of local feature detectors. The proposed covariant loss is the Frobenius norm of the difference between  the local affine frames. The disadvantage of such approach is that it could lead to features that are, while being repeatable, not necessarily suited for the matching task (see Section~\ref{affnet:loss}).
On top of that, the common drawback of the Yi~\etal~\cite{LIFT2016} and Lenc and Vedaldi~\cite{Lenc2016} methods is that they require to know the exact geometric relationship between patches which increases the amount of work needed to prepare the training dataset. 
Zhang~\etal~\cite{NewTilde2017} proposed to ``anchor'' the detected features to some pre-defined features with known good discriminability like TILDE~\cite{TIlde2015}. We remark that despite showing images of affine-covariant features, the results presented in the paper are for translation-covariant features only. 
Savinov~\etal~\cite{QuadNets2017} proposed a ranking approach for unsupervised learning of a feature detector. While this is natural and efficient for learning the coordinates of the center of the feature, it is problematic to apply it for the affine shape estimation. The reason is that it requires sampling and scoring of many possible shapes.

Finally, Choy~\etal~\cite{UCN2016} trained a ``Universal correspondence network'' (UCN) for a direct correspondence estimation with contrastive loss on a patch descriptor distance. This approach is related to the current work, yet the two methods differ in several important aspects. First, UCN used an ImageNet-pretrained network which is subsequently fine-tuned. We learn the affine shape estimation from scratch. Second, UCN uses  dense feature extraction and negative examples extracted from the same image. While this could be a good setup for short baseline stereo, it does not work well for wide baseline, where affine features are usually sought. Finally, we propose the hard negative-constant loss instead of the contrastive one.  

\section{Learning affine shape and orientation}%
\label{affnet:affnet-main}
\subsection{Affine shape parametrization}%
A local affine frame is defined by 6 parameters of the affine matrix. Two form a translation vector $(x,y)$ which is given by the keypoint detector and in the rest of the Chapter we omit it and focus on the \emph{affine transformation} matrix $A$,
\begin{equation}
\label{eq:A}
A =\left(%
\begin{array}
{@{\hspace{0pt}}c@{\hspace{2pt}}c@{\hspace{0pt}}}
a_{11} & a_{12}\\
a_{21} & a_{22}\\ 
\end{array}\right).\end{equation}
Among many possible decompositions of matrix $A$, we use the following  
\begin{equation}
\label{eq:A'}
A = \lambda R(\alpha) A'  =  \det{A} 
\left(
\begin{array}{@{\hspace{0pt}}c@{\hspace{6pt}}c@{\hspace{0pt}}}
\cos{\alpha} & \sin{\alpha} \\
-\sin{\alpha} & \cos{\alpha}\\
\end{array}\right) \left(
\begin{array}
{@{\hspace{0pt}}c@{\hspace{2pt}}c@{\hspace{0pt}}}
a'_{11} & 0\\
a'_{21} & a'_{22}\\ 
\end{array}\right),
\end{equation}
where $\lambda = \det{A}$ is the scale, $R(\alpha)$ the \emph{orientation} matrix and $A'$~\footnote{$A'$ has a (0,1) eigenvector, preserving the vertical direction.} is the
\emph{affine shape} matrix with $\det{A'}$ = 1.
$A'$ is decomposed into identity matrix $I$ and \emph{residual shape}~$A''$:
\begin{equation}
\label{eq:A''}
A' = I + A'' = %
 \left(
\begin{array}{@{\hspace{0pt}}c@{\hspace{2pt}}c@{\hspace{0pt}}}
a'_{11} & 0 \\
a'_{21} & a'_{22}\\
\end{array}\right) =
 \left(\begin{array}{@{\hspace{0pt}}c@{\hspace{6pt}}c@{\hspace{0pt}}}
1 & 0\\
0 & 1\\ 
\end{array}\right)  + 
\left(
\begin{array}{@{\hspace{0pt}}c@{\hspace{2pt}}c@{\hspace{0pt}}}
a''_{11} & 0\\
a''_{21} & a''_{22}\\
\end{array}\right) 
\end{equation}
We show that the different parameterizations of the affine transformation significantly influence the performance of CNN-based estimators of local geometry, see Table~\ref{tab:aff-shape-parametrization}.
 \begin{figure}[tb]%
 \centering
 Initial positions\\
 \includegraphics[width=0.30\linewidth]{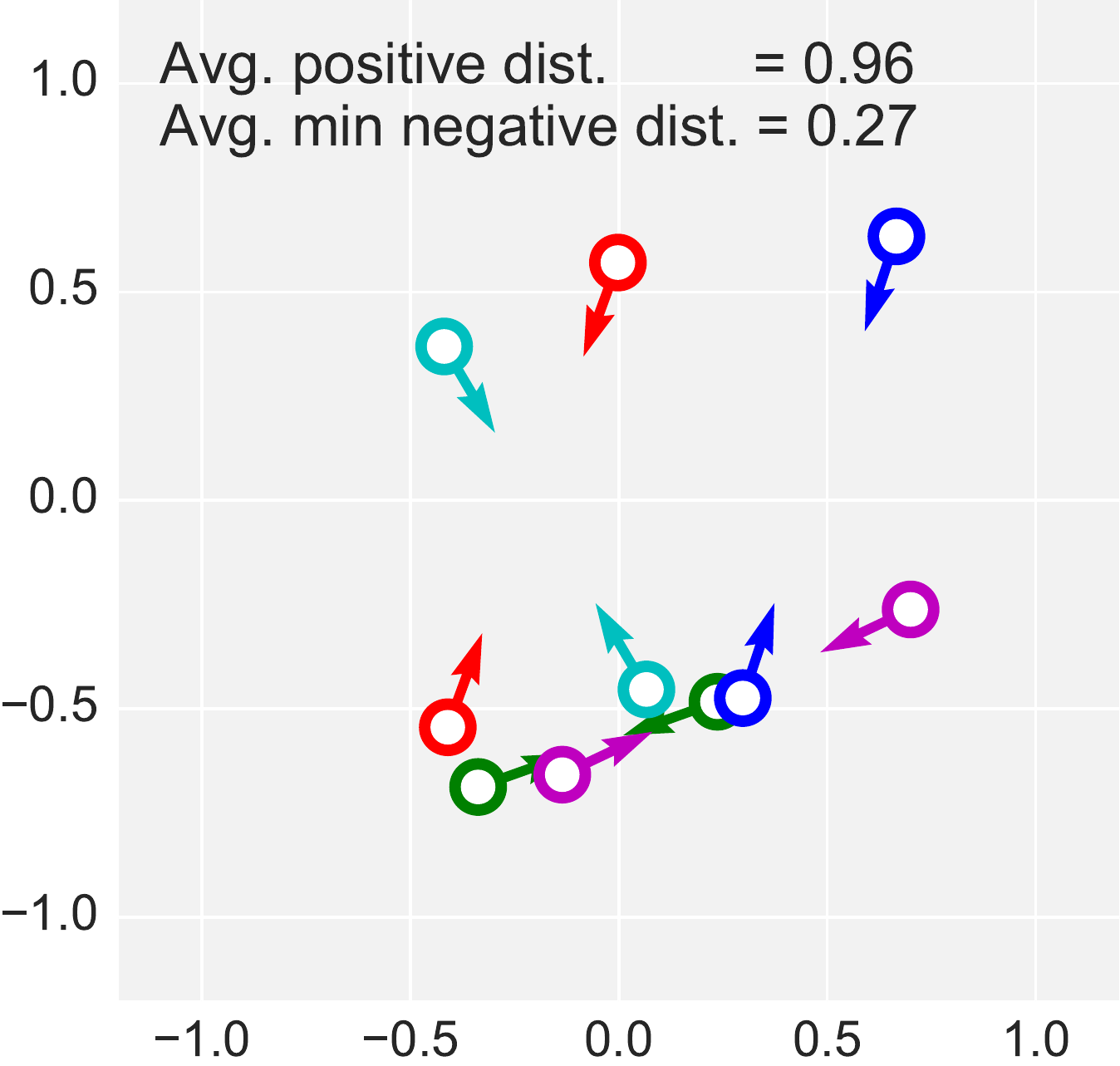}\hfill
 \includegraphics[width=0.30\linewidth]{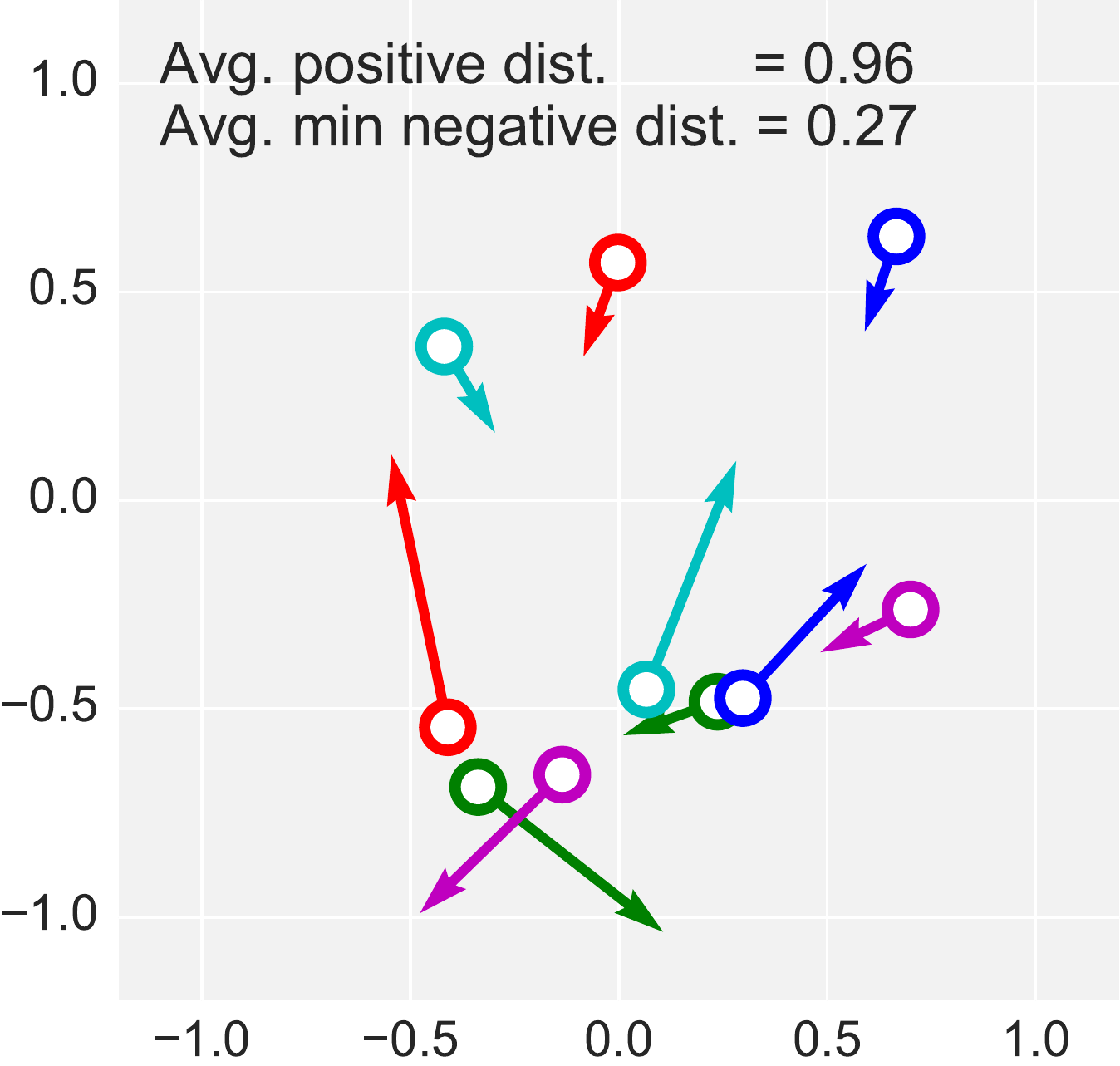}\hfill
 \includegraphics[width=0.30\linewidth]{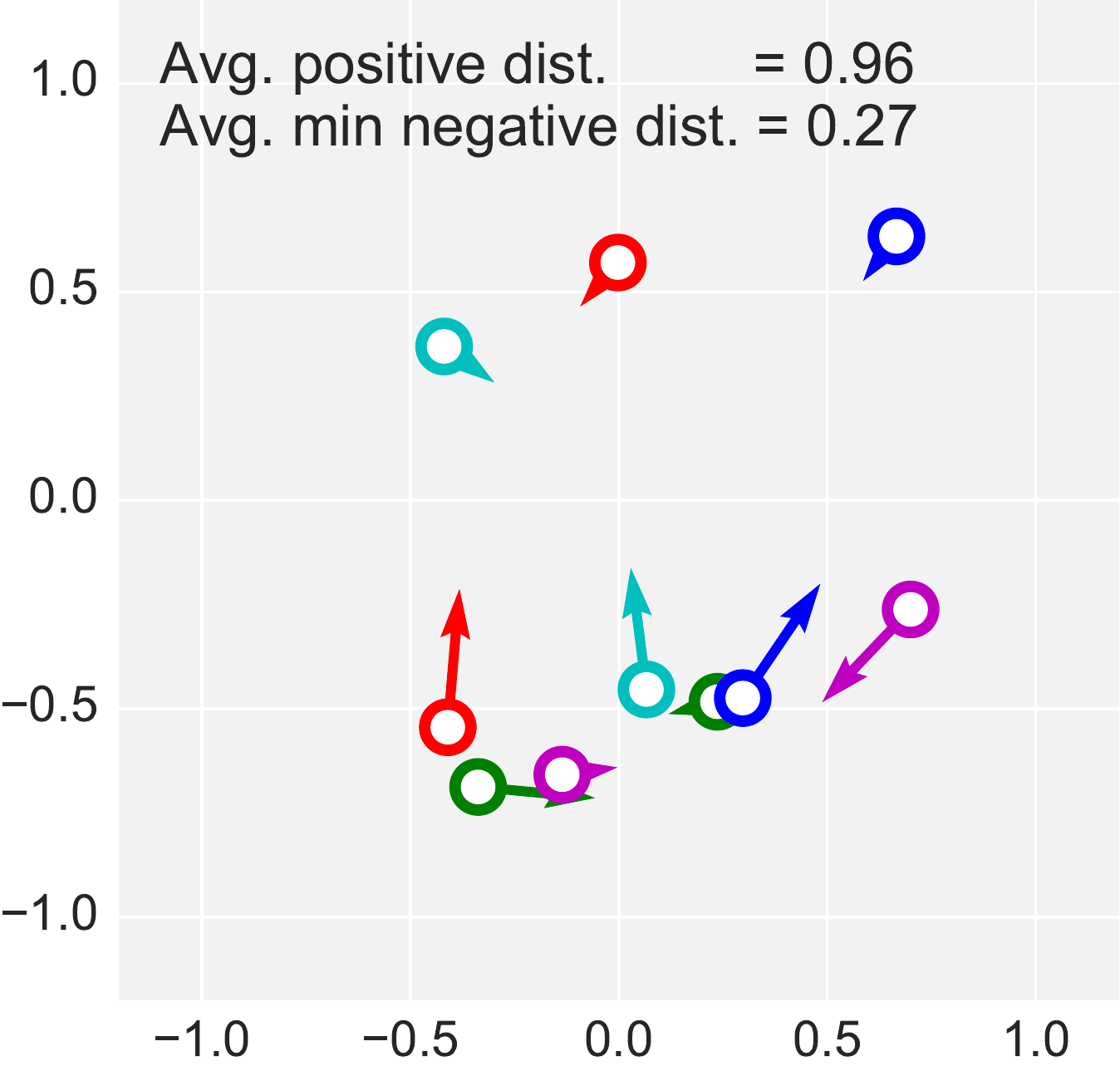}\\
After 150 Adam steps\\
\begin{minipage}[h]{0.30\linewidth}
\center{\includegraphics[width=1\linewidth]{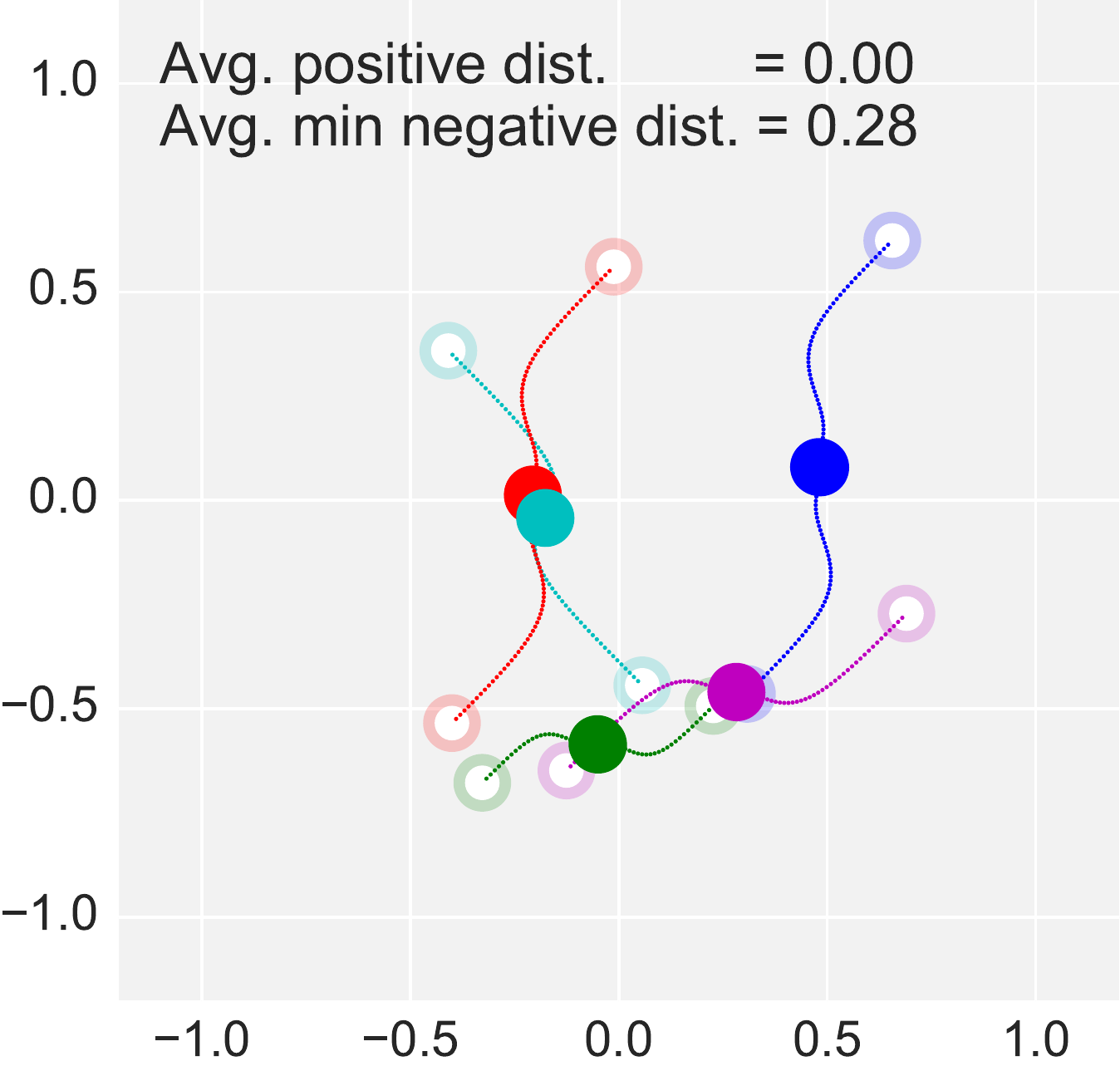}} PosDist loss\\
\end{minipage}
\hfill
\begin{minipage}[h]{0.30\linewidth}
\center{\includegraphics[width=1\linewidth]{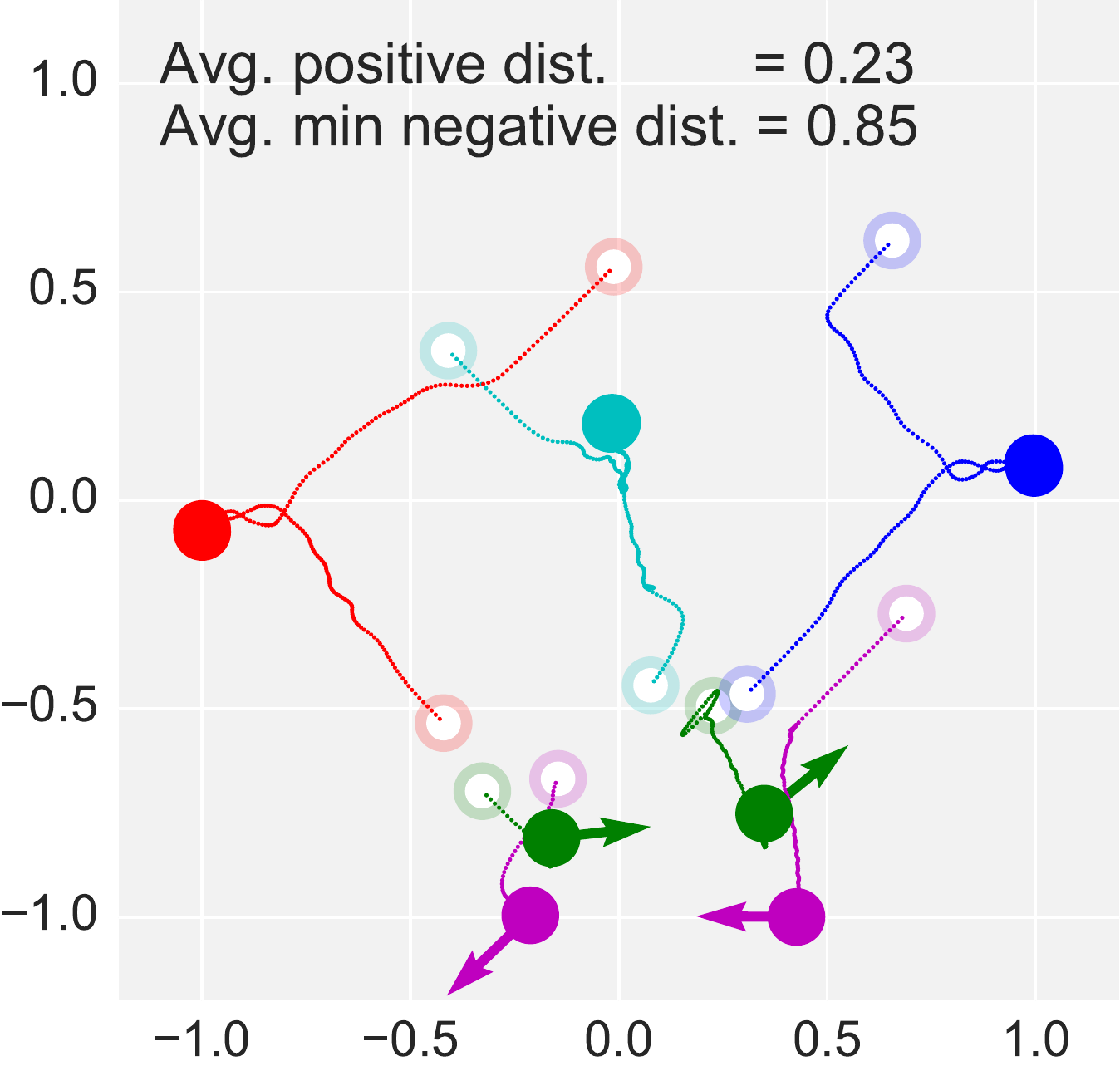}} HardNeg loss\\
\end{minipage}
\hfill
\begin{minipage}[h]{0.30\linewidth}
\center{\includegraphics[width=1\linewidth]{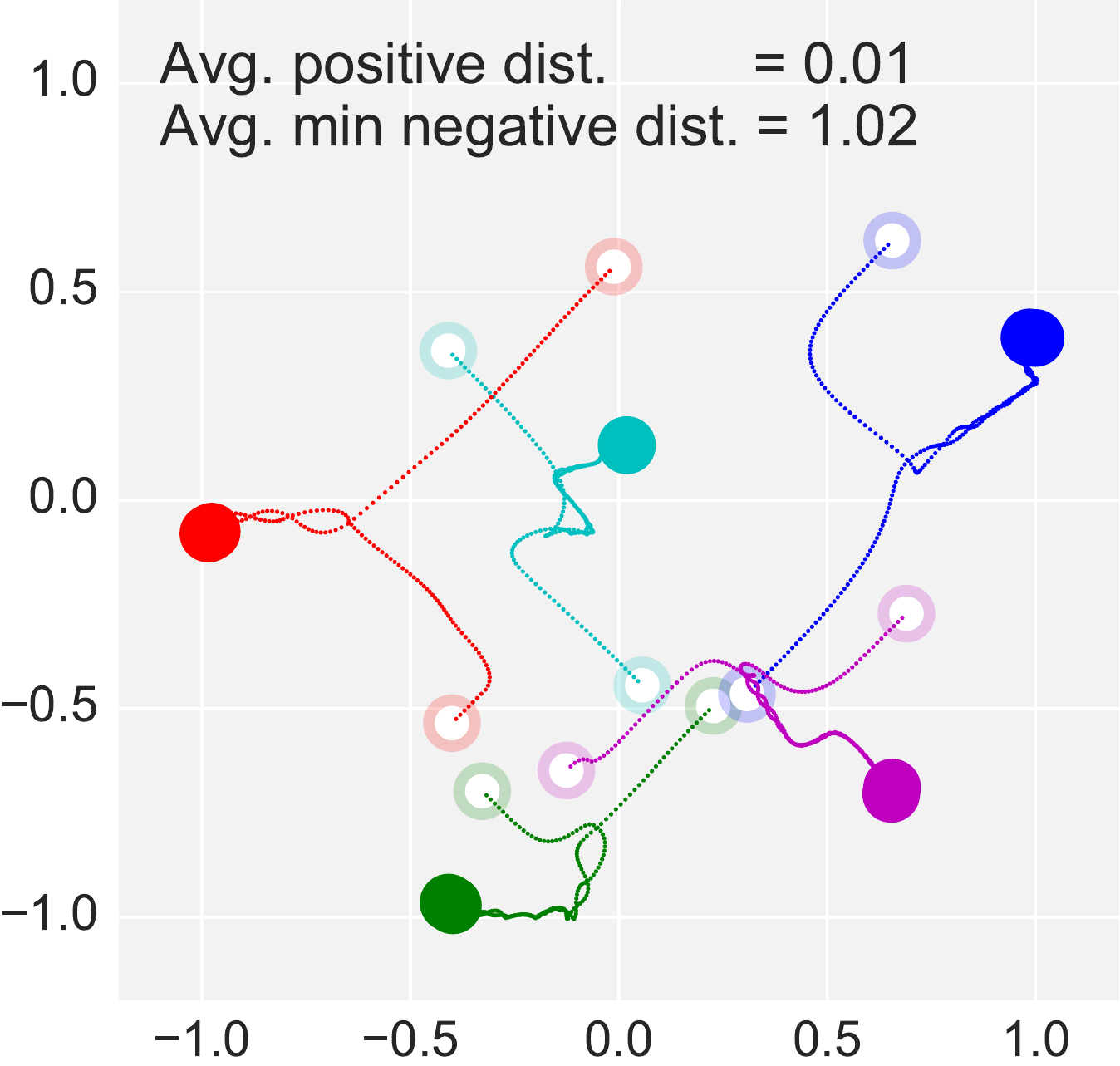}} HardNegC loss\\
\end{minipage}
 \caption{A toy example optimization problem illustrating the proposed hard negative-constant (HardNegC) loss. Five pairs of points, representing 2D descriptors, are generated and the losses are minimized by Adam~\cite{Adam2014}: the positive descriptor distance (PosDist)~\cite{OriNet2016} -- left, the hard negative (HardNeg) margin loss~\cite{HardNet2017} -- center, HardNegC-- right. 
Top row: identical initial positions of five pairs of matching points. Arrows show the gradient direction and relative magnitude. 
Bottom row: points after 150 steps of Adam optimization, trajectories are shown by dots. HardNeg loss has a difficulty with the green and magenta point pairs, because the negative example lies between two positives. Minimization of the positive distance only leads to a small distance to the negative examples. The proposed HardNegC loss first pushes same class points close to each other and then distributes them to increase distance to the negative pairs.}
\label{fig:toy}
\end{figure}%
\subsection{Hard negative-constant loss}%
\label{affnet:loss}
We propose a loss function called hard negative-constant loss (HardNegC). It is based on the hard negative triplet margin loss~\cite{HardNet2017} (HardNeg), but the distance to the hardest (i.e. closest) negative example is treated as constant and the respective derivative of $L$  is set to zero:
\begin{equation}
L = \frac{1}{n}\sum_{i = 1,n}{\max{(0, 1 + d(s_i, \dot{s}_i) - d(s_i, N))}}, \quad
\frac{\partial L}{\partial N} \coloneqq  0,
\label{eq:hardnegc}
\end{equation}
where $d(s_i, \dot{s}_i)$ is the distance between the matching descriptors, $d(s_i, N)$ is a distance to the hardest negative example $N$ in the mini-batch for $i^{th}$ pair.
\begin{equation*}
d(s_i, N) = \min{( \min_{j \ne i}{d(s_i, \dot{s}_j)}, \min_{j\ne i}{d(s_j, \dot{s}_i)})}
\end{equation*}
The difference between the Positive descriptor distance loss (PosDist) used for learning local feature orientation in~\cite{OriNet2016} and the HardNegC and HardNeg losses is shown on a toy example in Figure~\ref{fig:toy}. Five pairs of points in the 2D space are generated and their positions are updated by the Adam optimizer~\cite{Adam2014} for the three loss functions. PosDist converges first, but the different class points end up near each other, because the distance to the negative classes is not incorporated in the loss. The HardNeg margin loss has trouble when the points from different classes lie between each other. The HardNegC loss behavior first resembles the PosDist loss, bringing positive points together and then distributes them in the space, satisfying the triplet margin criterion.
\subsection{Descriptor losses for shape registration}%
\begin{figure}[tb]
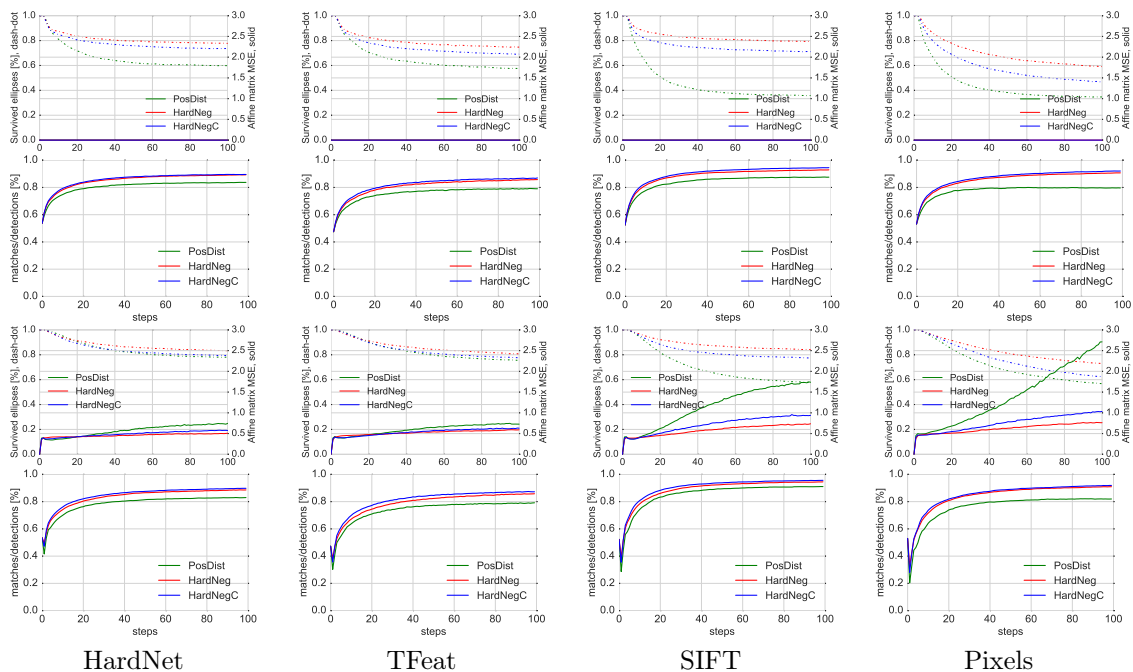
%
\input{chapters/affnet/fig-direct-opt-ideal-joint.tex}
\input{chapters/affnet/fig-direct-opt-ideal.tex}
\caption{Matching score versus geometric repeatability experiment. Affine shape registration by a minimization of descriptor losses of corresponding features. Descriptor losses: green -- L2- descriptor distance (PosDist)~\cite{OriNet2016}, red -- hard triplet margin HardNeg~\cite{HardNet2017}, blue -- proposed HardNegC. Average over HSequences, illumination subset.  
\emph{All features are initially perfectly registered}. 
First two rows: single feature geometry for both images, second two rows: feature geometries are independent in each image.
Top row: geometric error of corresponding features (solid) and percentage of non-collapsed, i.e. elongation $\leq$ 6, features (dashed). Bottom row: the percentage of correct matches.
This experiment shows that even perfectly initially registered feature might not be matched with any of descriptors -- initial matching score is roughly $\approx 30..50\%$. But it is possibly to find measurement region, which offers both discriminativity and repeatability.
 PosDist loss squashes  most of the features, leading to the largest geometrical error. HardNeg loss produces the best results in the number of survived feature and geometrical error. HardNegC performs slightly worse than HardNeg, slightly outperforming it on matching score. However, HardNegC is easier to optimize for AffNet learning -- see Table~\ref{tab:desc-and-loss}.}
\label{fig:direct-opt-ideal}
\end{figure}

Exploring how local feature repeatability is connected with descriptor similarity, we conducted an shape registration experiment (Figure~\ref{fig:direct-opt-ideal}). Hessian features are detected in reference HSequences~\cite{hpatches2017} illumination images and reprojected by (identity) homography to another image in the sequence. Thus, the repeatability is 1 and the reprojection error is 0. Then, the local descriptors (HardNet~\cite{HardNet2017}, SIFT~\cite{SIFT2004}, TFeat~\cite{TFeat2016} and raw pixels) are extracted and the features are matched by first-to-second-nearest neighbor ratio~\cite{SIFT2004} with threshold 0.8. This threshold was suggested by Lowe~\cite{SIFT2004} as a good trade-off between false positives and false negatives. For SIFT, 22\% of the geometrically correct correspondences are not the nearest SIFTs and they cannot be matched, regardless of the the threshold. In our experiments, the 0.8 threshold worked well for all descriptors and we used it, in line with previous papers, in all experiments.

Notice that for all descriptors, the percentage of correct matches even for the \emph{perfect} geometrical registration is only about 50\%. 

Adam optimizer is used to update the affine region $A$ to minimize the descriptor-based losses: PosDist, HardNeg and HardNegC. 
The top two rows show the results for $A$ matrices coupled for both images, bottom -- the descriptor difference optimization is allowed to deform $A$ and $\dot{A}$ in both images independently, which leads to a pair of affine regions that are not in perfect geometric correspondence, yet they are more matchable. Note that no training of any kind is involved. %

Such descriptor-driven optimization, not maintaining perfect registration, produces a descriptor that is matched successfully up to 90\% of the detections under illumination changes. %

For most of the unmatched regions, the affine shapes become degenerate lines -- shown in the top graphs, and the number of degenerate ellipses is high for PosDist loss; HardNeg and HardNegC perform better.

\begin{figure}[tb]%
\input{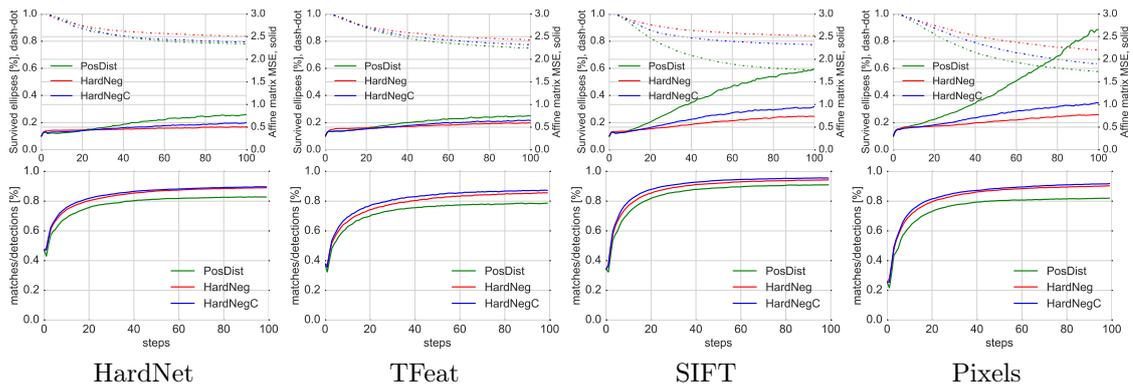}
\caption{Minimization of descriptor loss by optimization of affine parameters of corresponding features. Average over HPatchesSeq, illumination subset. Top row: geometric error of corresponding features (full line)  and percentage of non-collapsed, \ie elongation $\leq$ 6, features (dashed line). Bottom row: the fraction correct matches.
 All features initially have the same medium amount of reprojection noise.  Left to right: HardNet, SIFT, TFeat, mean-normalized pixels descriptors.}
 \label{fig:direct-opt-random}
 \end{figure}

The bottom row of Figure~\ref{fig:direct-opt-ideal} shows the results of experiments where affine shape pairs are independent in each image. Optimization of descriptor losses leads to an increase of the geometric error on the affine shape. Error $E$ is defined as the mean square error on A matrix difference:
\begin{equation}
E = \sum_{i=1}^{n} \frac{2(A_i -\dot{A}_i)^2}{\det{A} +\det{\dot{A}}}
\end{equation}
Again, PosDist loss leads to a larger error. CNN-based descriptors -- HardNet and TFeat -- lead to a relative small geometric error when reaching the matchability plateau, while for SIFT and raw pixels the shapes diverge. Figure~\ref{fig:direct-opt-random} shows the case when the initialized shapes include a small amount of reprojection error. 
\begin{figure}[tb]%
\centering
 \includegraphics[width=0.95\linewidth]{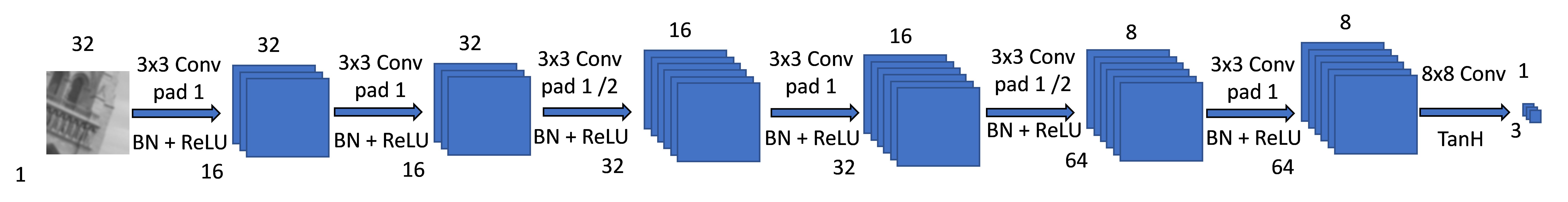}\\
 \caption{AffNet. Feature map spatial size -- top, \# channels -- bottom. /2 stands for stride 2.}
 \label{fig:architecture}
\end{figure}%
\begin{figure}[tb]%
\centering
\includegraphics[width=0.95\linewidth]{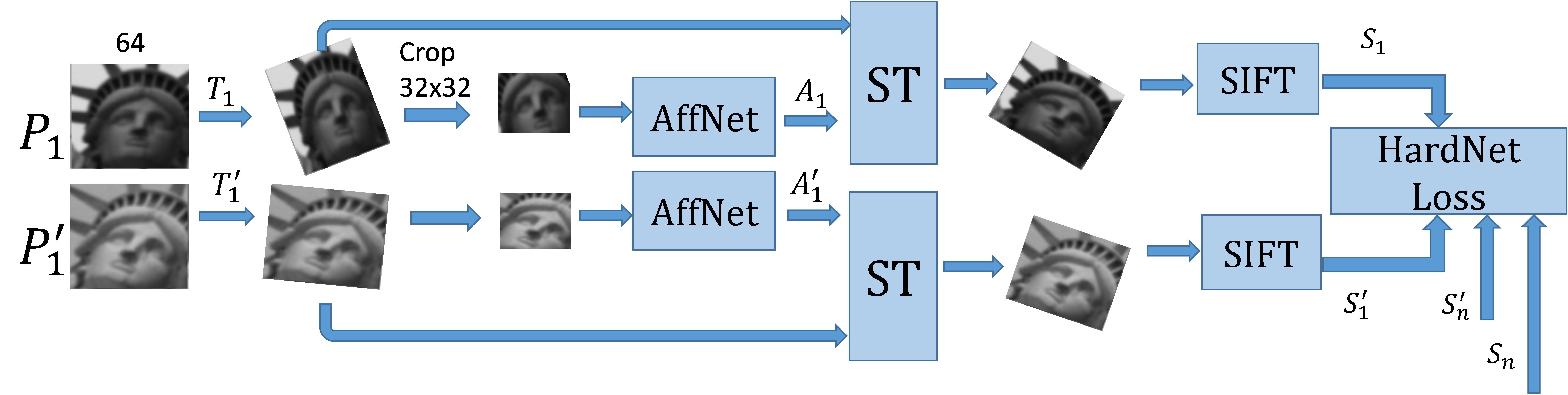}
 \caption{AffNet training. Corresponding patches undergo random affine transformation $T_i, \dot{T}_i$, are cropped and fed into AffNet, which outputs affine transformation $A_i, \dot{A}_i$ to an unknown canonical shape. ST -- the spatial transformer warps the patch into an estimated canonical shape. The patch is described by a differentiable CNN descriptor. $n \times n$ descriptor distance matrix is calculated and used to form triplets, according to the HardNegC loss.}
\label{fig:training-scheme}
\end{figure}%
\subsection{AffNet training procedure}%
The main blocks of the proposed training procedure are shown in Figure~\ref{fig:training-scheme}. 
First, a batch of matching patch pairs $(P_i, \dot{P}_i)_{i=1..n}$ is generated, where $P_i$ and $\dot{P}_i$ correspond to the same point on a 3D surface. Rotation and skew transformation matrices $(T_i, T_i')$ are randomly and independently generated. The patches $P_i$ and $\dot{P}_i$ are warped by  $(T_i, \dot{P}_i)$ respectively into $A$-transformed patches. Then, a $32\times 32$ center patch is cropped and a pair of transformed patches is fed into the convolutional neural network AffNet, which predicts a pair of affine transformations $A_i$, $\dot{P}_i$, that are applied to the $T_i$-transformed patches via spatial transformers ST~\cite{SpatialTransformers2015}. 

Thus, geometrically normalized patches are cropped to $32\times 32$ pixels and fed into the descriptor network, e.g. HardNet, SIFT or raw patch pixels, obtaining descriptors $(s_i, \dot{s}_i)$. 
Descriptors $(s_i, \dot{s}_i)$ are then used to form triplets by the procedure proposed in~\cite{HardNet2017}, followed by our newly proposed hard negative-constant loss (Eq.~\ref{eq:hardnegc}).

More formally, we are finding affine transformation model parameters $\theta$ such that estimated affine transformation $A$ minimizes the descriptor HardNegC loss:
\begin{equation}
A(\theta | (P,\dot{P})) = \argmin_{\theta}{L(s,\dot{s})}
\end{equation}
\subsection{Training dataset and data preprocessing}%
UBC Phototour~\cite{Brown2007} dataset is used for training. It consists of three subsets: \emph{Liberty}, \emph{Notre Dame} and \emph{Yosemite} with about 2 $\times$ 400k normalized 64x64 patches in each, detected by DoG and Harris detectors. Patches are verified by the 3D reconstruction model. We randomly sample 10M pairs for training.

Although positive points correspond to roughly the same point on the 3D surface, they are not perfectly aligned, having position, scale, rotation, and affine noise.
We have randomly generated affine transformations, which consist in a random rotation -- tied for the pair of corresponding patches, and anisotropic scaling $t$ in random direction by magnitude $t_m$, which is gradually increased during the training from the initial value of 3 to 5.8 at the middle of the training. The tilt is uniformly sampled from range $[0,t_m]$. %
\subsection{Implementation details}%
The CNN architecture is adopted from HardNet\cite{HardNet2017}, see Fig.~\ref{fig:architecture}, 
with the number of channels in all layers reduced  2x and the last 128D output  replaced by  a 3D output predicting the ellipse shape. 
The network formula is 16C3-16C3-32C3/2-32C3-64C3/2-64C3-3C8, where 32C3/2 stands for 3x3 kernel with 32 filters and stride 2. Zero-padding is applied in all convolutional layers to preserve the size, except the last one. BatchNorm~\cite{BatchNorm2015} layer followed by ReLU~\cite{Nair2010RectifiedLinearUnits} is added after each convolutional layer, except the last one, which is followed by hyperbolic tangent activation.
Dropout~\cite{Dropout2014} with 0.25 rate is applied before the last convolution layer. Grayscale input patches $32 \times 32$ pixels are normalized by subtracting the per-patch mean and dividing by the per-patch standard deviation. 

Optimization is done by SGD with learning rate 0.005, momentum 0.9, weight decay 0.0001. The learning rate decayed linearly~\cite{Systematic2017} to zero within 20 epochs. The training was done with PyTorch~\cite{pytorch} and took 24 hours on Titan X GPU; the bottleneck is the data augmentation procedure. The inference time is 0.1~ms per patch on Titan X, including patch sampling done on CPU and Baumberg iteration -- 0.05~ms per patch on CPU.
\section{Empirical evaluation}%
\label{affnet:exp}
\begin{figure}[tb]%
\centering
  \includegraphics[width=0.24\linewidth]{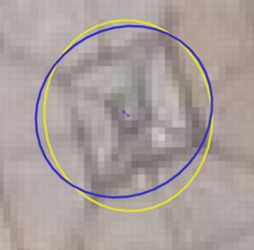}
   \includegraphics[width=0.24\linewidth]{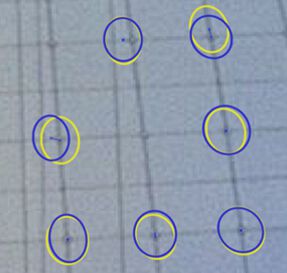}
  \includegraphics[width=0.24\linewidth]{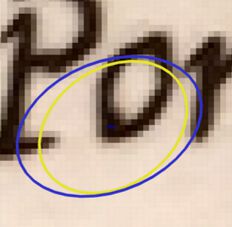}
   \includegraphics[width=0.24\linewidth]{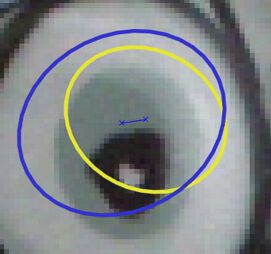}\\
     \includegraphics[width=0.24\linewidth]{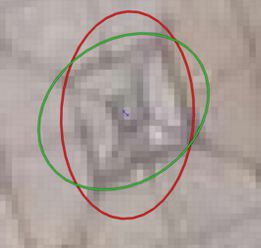}
   \includegraphics[width=0.24\linewidth]{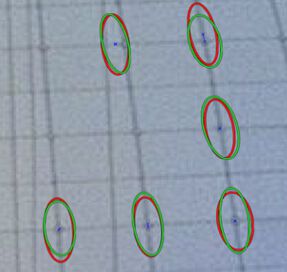}
  \includegraphics[width=0.24\linewidth]{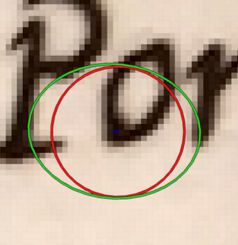}
   \includegraphics[width=0.24\linewidth]{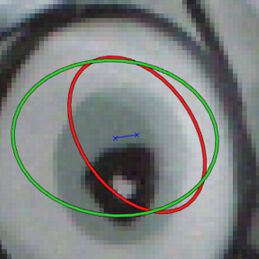}\\
 \caption{ AffNet (top) and Baumberg (bottom) estimated affine shape. One ellipse is detected in the reference image, the other is a reprojected closest match from the second image. Baumberg ellipses tend to be more elongated, average axis ratio is 1.99 vs. 1.63 for AffNet, median: Baumberg 1.72 vs 1.39 AffNet. The statistics are calculated over 16M features on Oxford5k.}
 \label{fig:patch-examples}
\end{figure}%
\subsection{Loss functions and descriptors for learning measurement region}%
We trained different versions of the AffNet and orientation networks with different combinations, affine transformation parameterizations, and descriptors with the procedure described above. 
The results of the comparison based on the number of correct matches (reprojection error $\leq$ 3 pixels) on the hardest pair for each of the 116 sequences from the HSequences~\cite{hpatches2017} dataset are shown in Tables~\ref{tab:desc-and-loss},\ref{tab:aff-shape-parametrization}.

The proposed HardNetC loss is the only loss function with no "not converged" results. In the case of convergence, all tested descriptors and loss functions lead to comparable performance, unlike the registration experiments in the previous section. We believe it is because now the CNN always outputs the same affine transformation for a patch, unlike in the previous experiment, where repeated features may end up with different shapes.

Affine transformation parameterizations are compared in Table~\ref{tab:aff-shape-parametrization}. All attempts to learn affine shape and orientation jointly in one network fail completely, or perform significantly worse than the two-stage procedure, when the affine shape is learned first and orientation is estimated on an affine-shape-normalized patch. 
Learning the residual shape $A''$ (Eq.~\ref{eq:A''}) leads to the best results overall. Note that such parameterization does not contain enough parameters to include fthe eature orientation, thus "joint" learning is not possible. Slightly worse performance is obtained by using an identity matrix prior for learnable biases in the output layer. %
\begin{table}[tb]%
\ra{1.0}
\centering
\caption{Learning the affine transform: loss functions and descriptor comparison. The median of average number of correct matches on the HSequences~\cite{hpatches2017} hardest image pairs 1-6 for the Hessian detector and the HardNet descriptor. The match considered correct for reprojection error $\leq$ 3 pixels. Affine shape is parametrized as in Eq.~\ref{eq:A''}. n/c -- did not converge.}
\label{tab:desc-and-loss}
\setlength{\tabcolsep}{1mm}
\begin{tabular}{cccc}
\toprule
Training descriptor/loss & PosDist& HardNeg & HardNegC  \\
\cmidrule(r){1-4}
\multicolumn{4}{c}{Affine shape}\\
\cmidrule(r){1-4}
SIFT & n/c & 385 &  386 \\
HardNet & n/c & n/c &  \textbf{388} \\
\cmidrule(r){1-4}
Baumberg~\cite{Baumberg2000} &\multicolumn{3}{c}{298}  \\
\cmidrule(r){1-4}
\multicolumn{4}{c}{Orientation}\\
\cmidrule(r){1-4}
SIFT &  \textbf{387} & 379 &  382 \\
HardNet & 386 & 383 & 380 \\
\cmidrule(r){1-4}
Dominant orientation~\cite{SIFT2004} &\multicolumn{3}{c}{339}  \\
\bottomrule
\end{tabular}
\end{table}
\begin{table}[tb]%
\centering
\caption{Learning the affine transform: parameterization comparison. The average number of correct matches on the HPatchesSeq~\cite{hpatches2017} hardest image pairs 1-6 for the Hessian detector and the HardNet descriptor.
Cases compared, affine shape combined with the de-facto handcrafted standard dominant orientation, affine shape and orientation learnt separately or jointly. The match considered correct for reprojection error $\leq$ 3 pixels. The HardNegC loss and HardNet descriptor used for learning. n/c -- did not converge.}
\label{tab:aff-shape-parametrization}
\setlength{\tabcolsep}{2mm}
\begin{tabular}{lcllccc}
\toprule
&&&& \multicolumn{3}{c}{Orientation} \\
\cmidrule(r){5-7}
& & Estimated& biases & \multicolumn{2}{c}{Learned} & Dominant \\ 
\cmidrule(r){5-6}
Eq. &Matrix& parameters &  init & jointly & separately & gradient~\cite{SIFT2004} \\ 
\cmidrule(r){1-7}
(\ref{eq:A})&$A$ & $(a_{11},a_{12}, a_{21},a_{22})$ & 0 & n/c & n/c  & n/c \\
(\ref{eq:A})&$A$  & $(a_{11},a_{12}, a_{21},a_{22})$ & 1 & n/c & 360  & 320 \\
\cmidrule(r){1-7}
(\ref{eq:A'}) &$A'$,& $(a'_{11},0 , a'_{21},a'_{22})$, & 1 & 250 & 327  & 286 \\
&  $R(\alpha)$  & $(\sin{\alpha}, \cos{\alpha}$) & & & \\
\cmidrule(r){1-7}
(\ref{eq:A''})& $A''$ & $(a''_{11},a''_{21},a''_{22})$&  1 & - &  370 & 340  \\
(\ref{eq:A''})&$A''$ & $(1 + a''_{11},a''_{21}, 1 + a''_{22})$&  0 & - &  \textbf{388} & 349  \\
\bottomrule
\end{tabular}
\end{table}
\begin{figure}[tb]
\centering
 \includegraphics[width=0.24\linewidth]{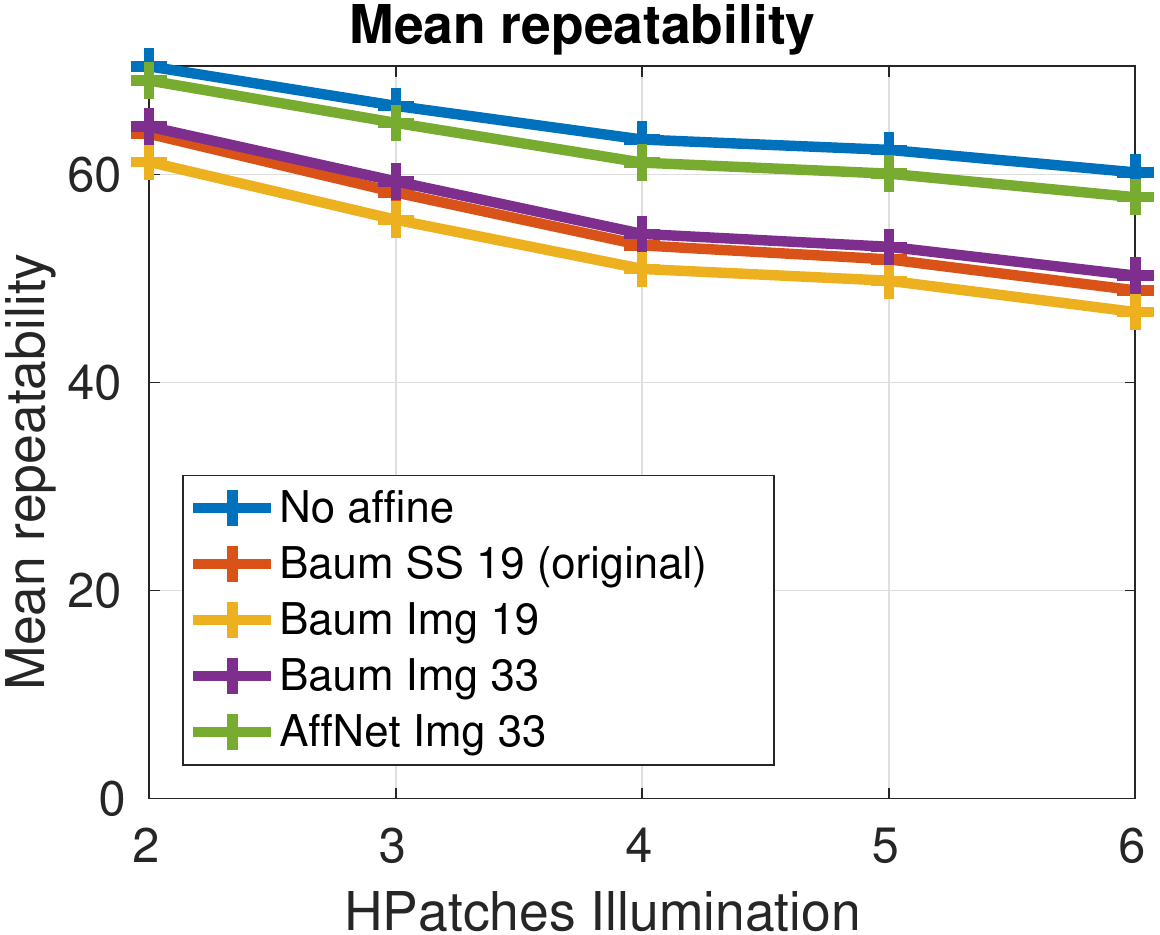}
 \includegraphics[width=0.24\linewidth]{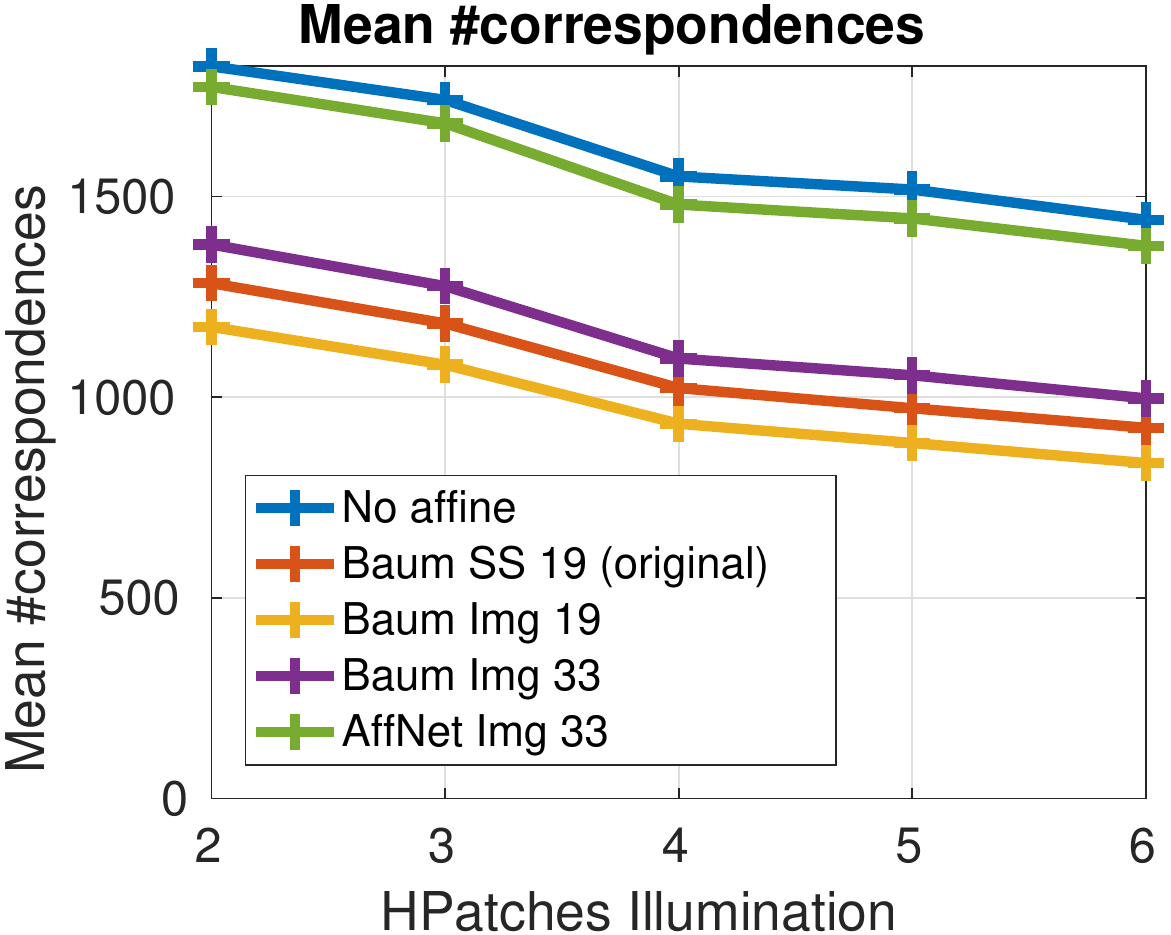}
 \includegraphics[width=0.24\linewidth]{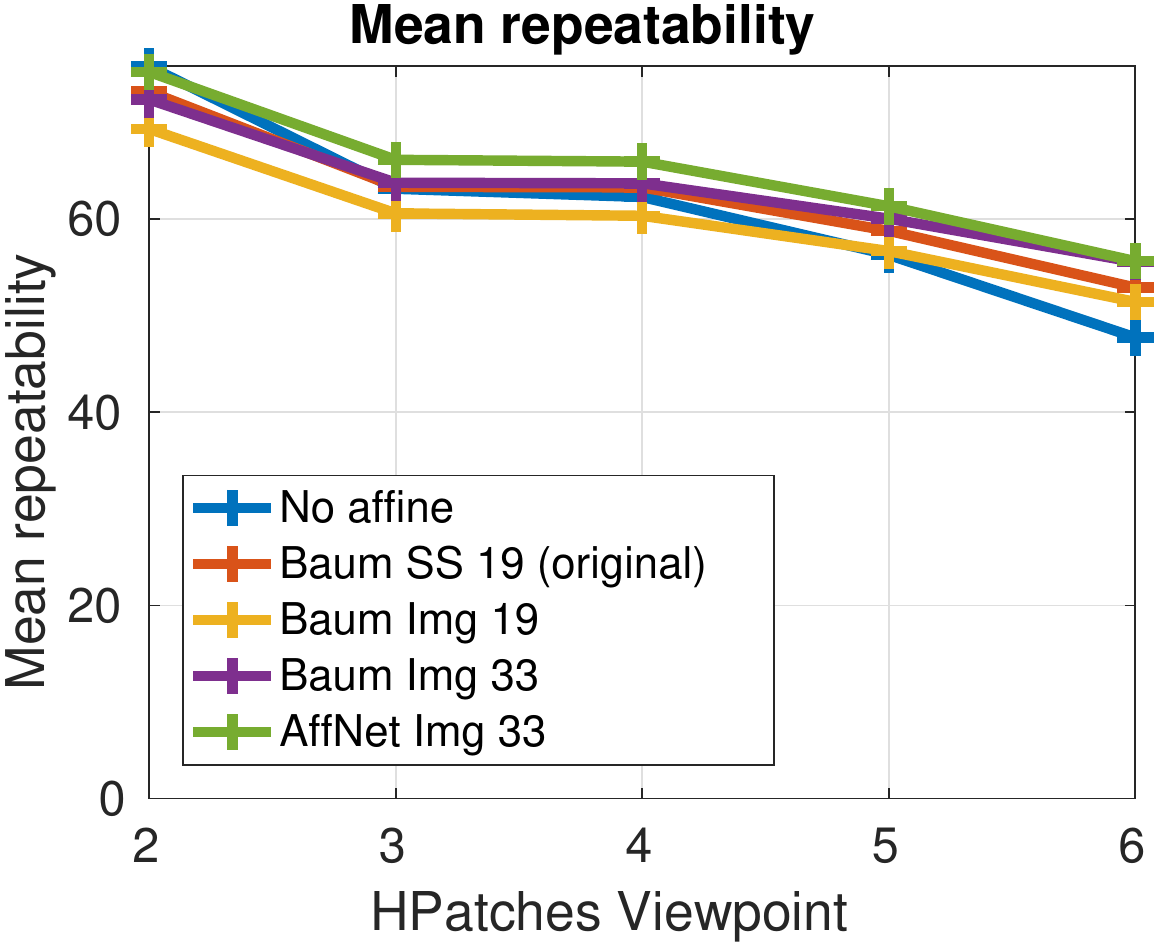}
 \includegraphics[width=0.24\linewidth]{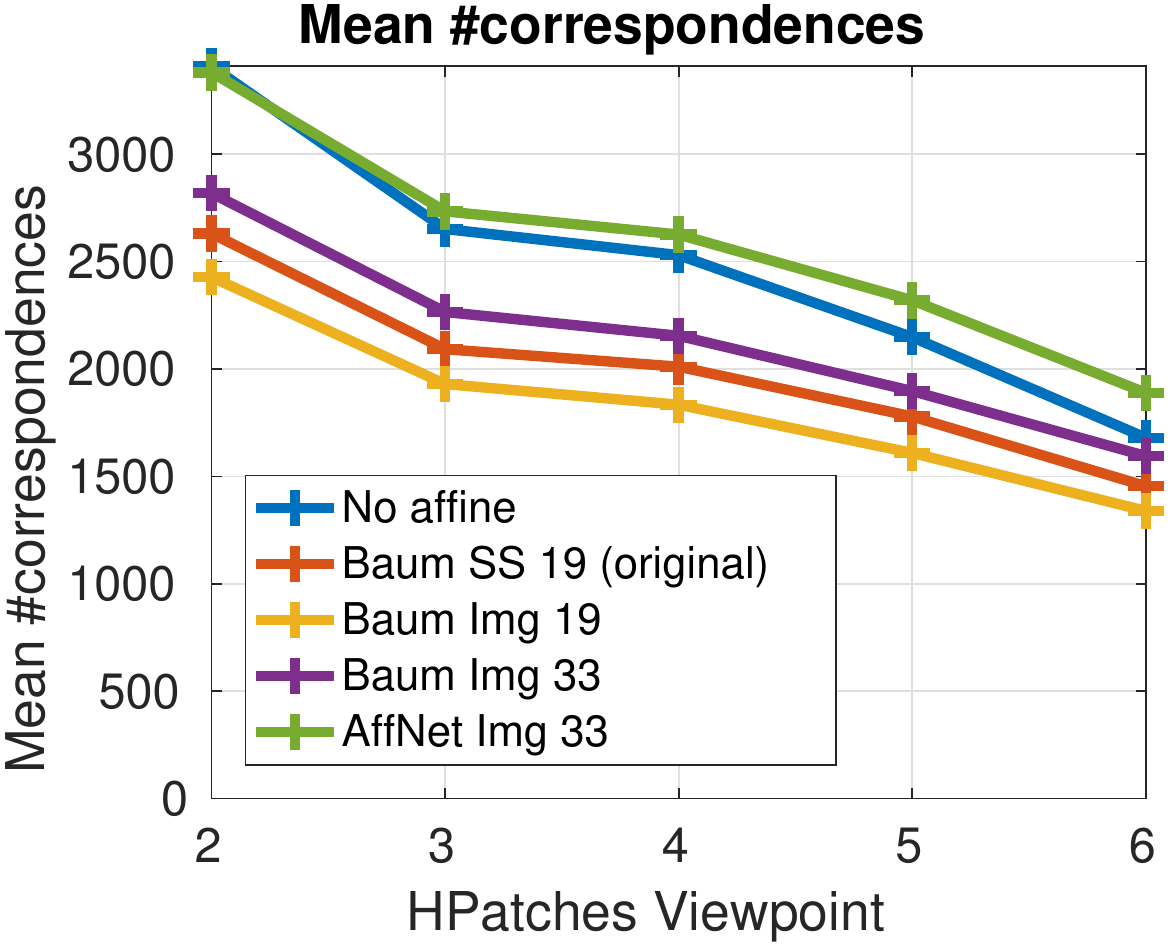}\\
 \includegraphics[width=0.24\linewidth]{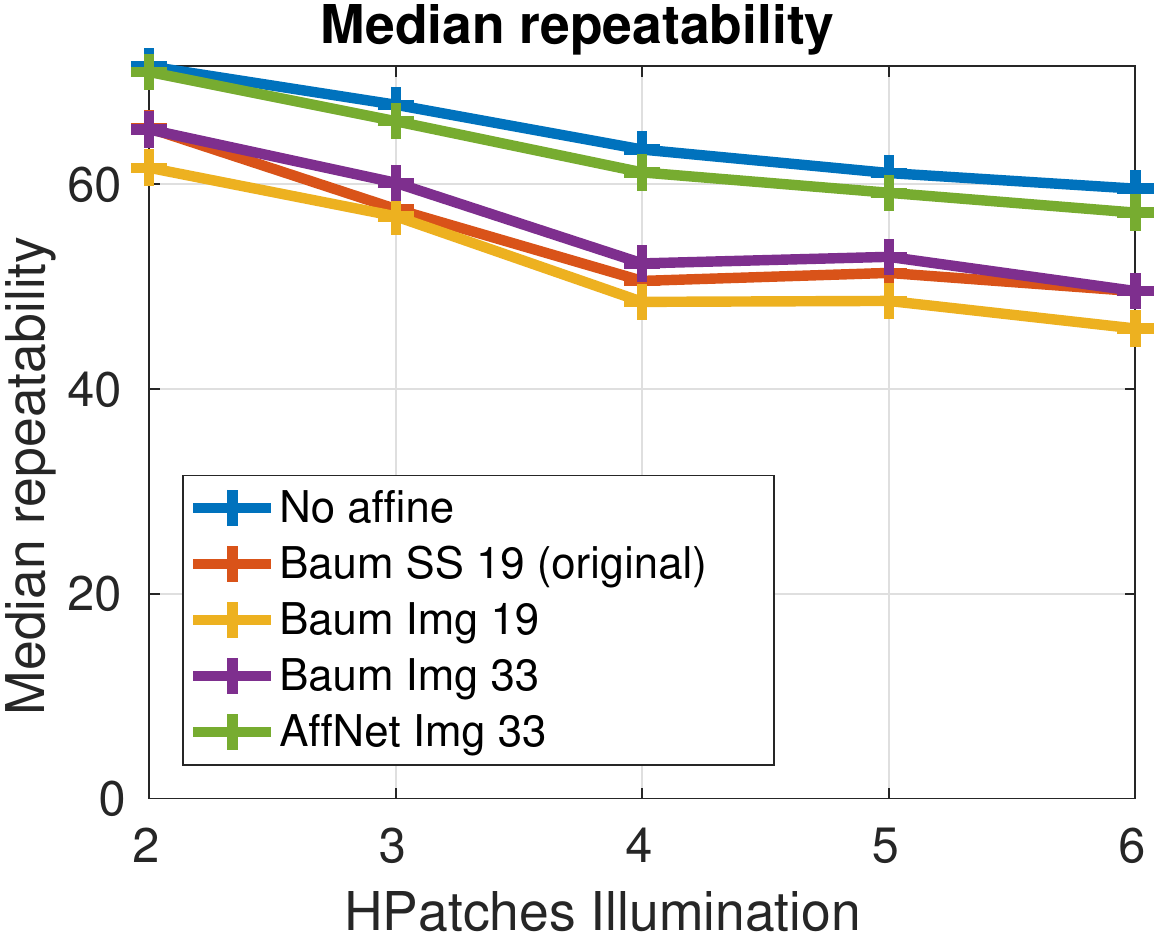}
 \includegraphics[width=0.24\linewidth]{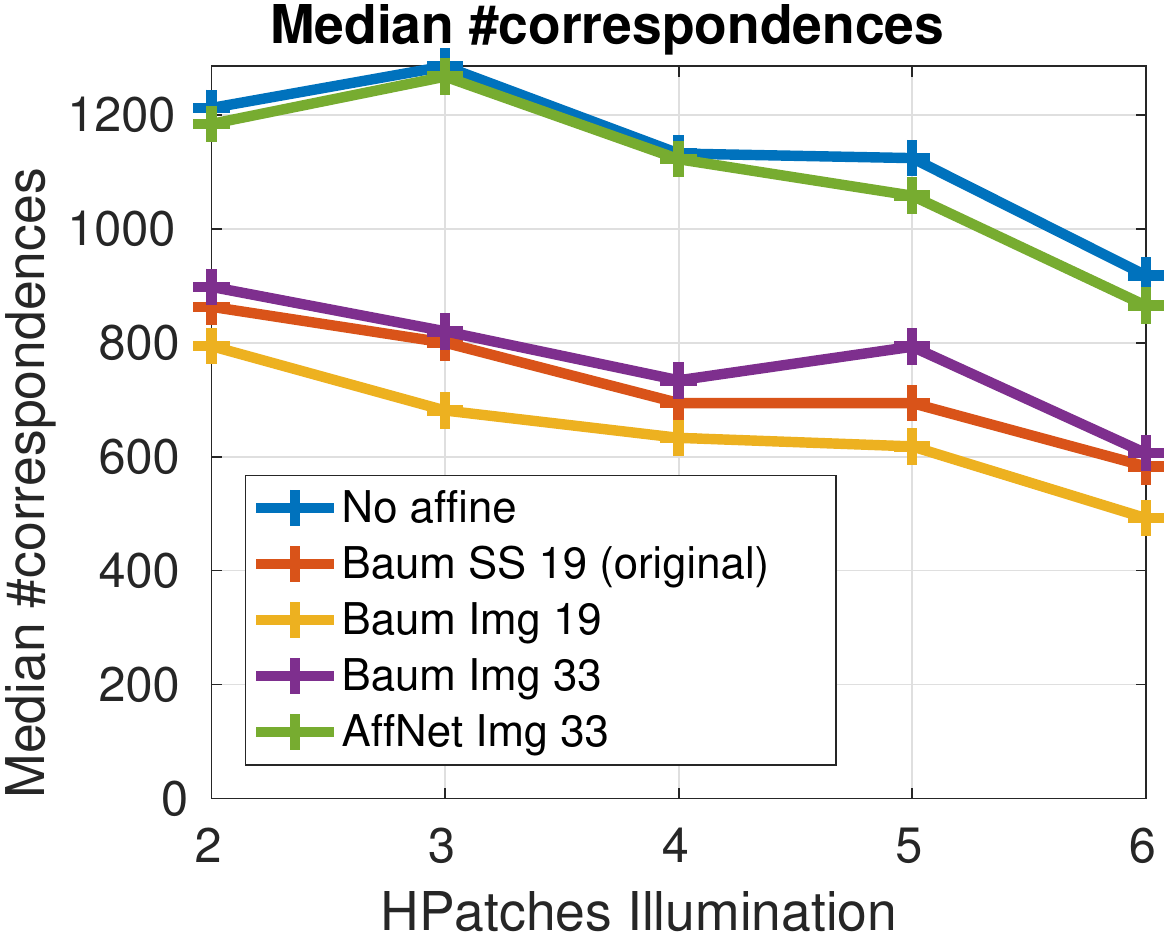}
 \includegraphics[width=0.24\linewidth]{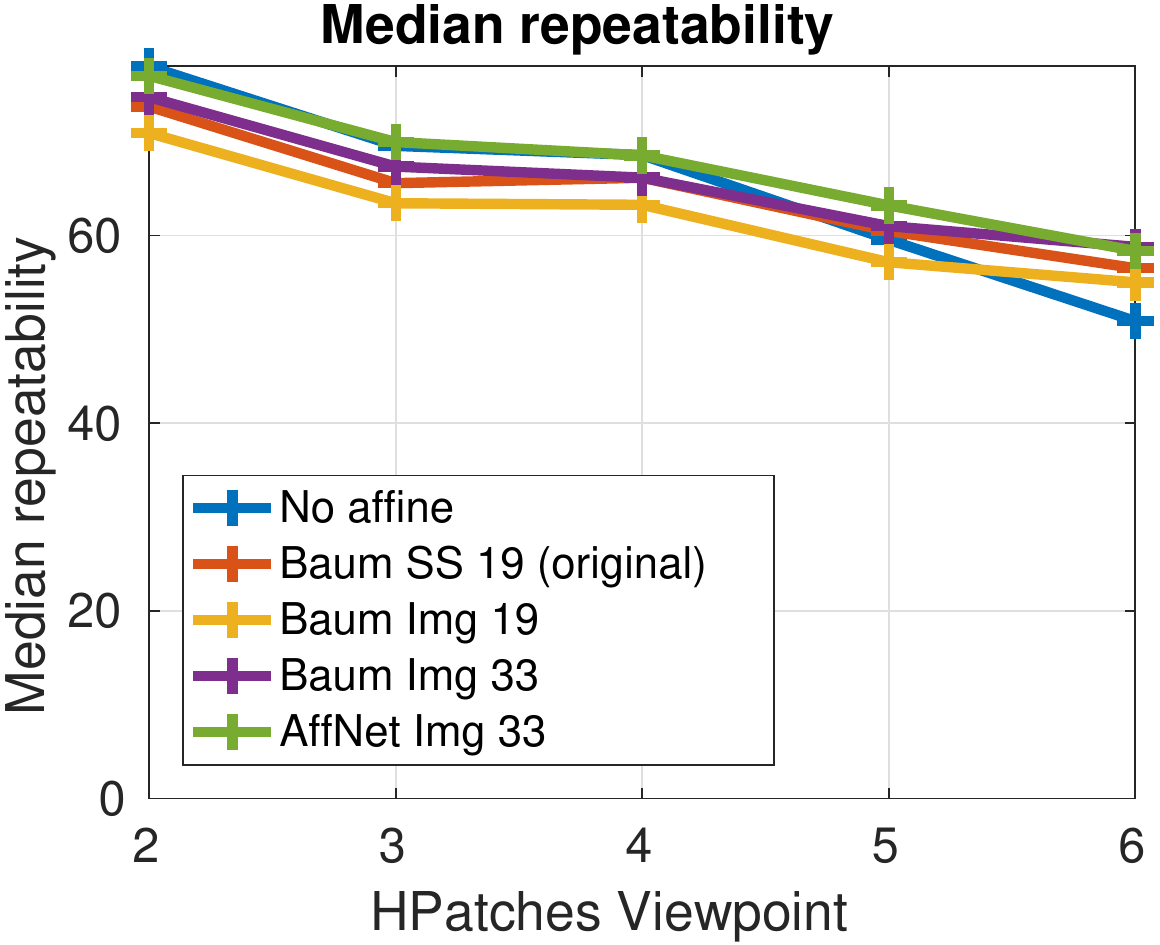}
 \includegraphics[width=0.24\linewidth]{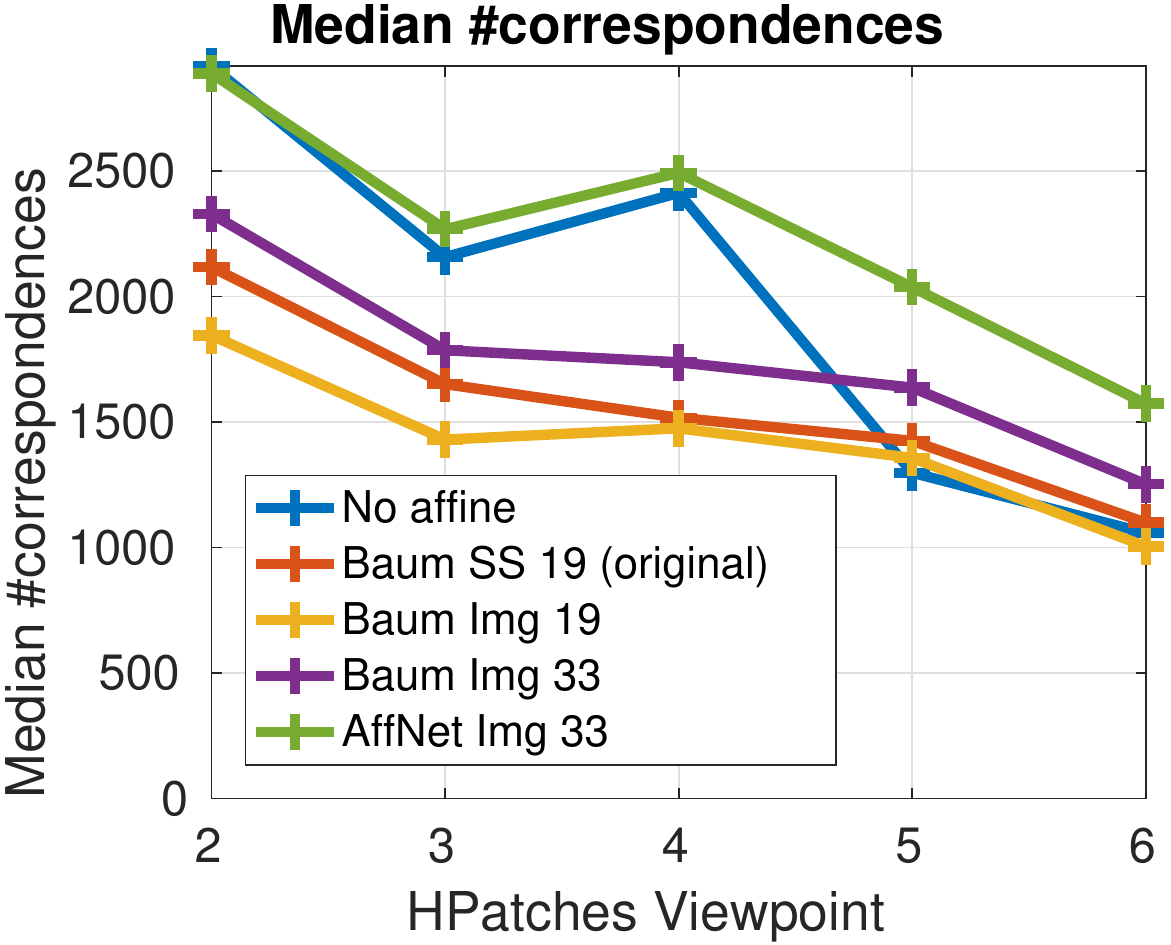}\\
 \caption{Repeatability and the number of correspondences (mean top, median bottom row) on the HSequences~\cite{hpatches2017}. AffNet is compared with the de facto standard Baumberg iteration~\cite{Baumberg2000} according to the Mikolajczyk protocol~\cite{Mikolajczyk2005}. Left -- images with illumination differences, right -- with viewpoint and scale changes. SS -- patch is sampled from the scale-space pyramid at the level of the detection, image -- from the original image; 19 and 33 -- patch sizes. Hessian-Affine is from\cite{PerdochRetrieval2009}. For illumination subset, performance of Hessian with no adaptation is an upper bound, and AffNet performs close to it.}
 \label{fig:repeatability}
\end{figure}
\subsection{Repeatability}%
\label{exp:rep}
Repeatability of affine detectors: Hessian detector + affine shape estimator was benchmarked, following the classical work by Mikolajczyk~\etal~\cite{Mikolajczyk2005}, but with the recently introduced larger HSequences~\cite{hpatches2017} dataset by VLBenchmarks toolbox~\cite{lenc2012vlbenchmarks}.

HSequences consists of two subsets. \textit{Illumination} part contains 57 image sixplets with illumination changes, both natural and artificial. There is no difference is viewpoint in this subset, the geometrical relation between images in sixplets is identity.Second part is \textit{Viewpoint}, where 59 image sixplets vary in scale, rotation, but mostly in horizontal tilt. The average viewpoint change is a bit smaller than in well-known \emph{graffiti} sequence from Oxford-Affine dataset~\cite{Mikolajczyk2005}.

Local features are detected in pairs of images, reprojected by ground truth homography to the reference image and the closest reprojected region is found for each region in the reference image. The correspondence is considered correct when the overlap error of the pair is less than 40\%. The repeatability score for a given pair of images is a ratio between the number of correct correspondences and the smaller number of detected regions in the common part of a scene among two images. 

Results are shown in Figure~\ref{fig:repeatability}. The original affine shape estimation procedure, implemented in~\cite{PerdochRetrieval2009} is denoted Baum SS 19, as $19\times 19$ patches are sampled from the scale space. AffNet takes $32\times 32$ patches, which are sampled from the original image. Therefore, for a fair comparison, we also tested Baum versions, where patches are sampled from the original image, with 19 and 33 pixels patch size.

AffNet slightly outperforms all variants of Baumberg procedure for images with viewpoint changes in terms of repeatability and more significantly -- in the number of correspondences.
The difference is even bigger for the image with illumination change only, where AffNet performs almost the same as the plain Hessian, which is the upper bound here, as this part of the dataset has no viewpoint changes. 

We have also tested AffNet with other detectors on the Viewpoint subset of the HPatches. The repeatabilities are the following (no affine adaptation/Baumberg/AffNet): DoG: 0.46/0.51/0.52, Harris: 0.41/0.44/0.47, Hessian: 0.47/0.52/0.56
The proposed method outperforms the standard (Baumberg) for all detectors. 

One reason for such difference is the feature rejection strategy.
Baumberg iterative procedure rejects the feature in one of three cases. First, elongated ellipses with long-to-short axis ratio more than six are rejected.
Second, features touching boundary of the image are rejected. This is true for the AffNet postprocessing procedure as well, but AffNet produces less elongated shapes: the average axis ratio for Oxford5k 16M features is 1.63 vs. 1.99 for Baumberg. Both cases happen less often for AffNet, increasing the number of surviving features by 25\%. We compared the performance of the Baumberg vs. AffNet with the same number of features in Section~\ref{exp:retrieval}. 
Finally, features whose shape did not converge within sixteen iterations are removed. This is quite rare, it happens in approximately 1\% of cases. 
Examples of shapes estimated by AffNet and the Baumberg procedure are shown in Fig.~\ref{fig:patch-examples}. 
\begin{table*}[htb]
\ra{1}
\centering
\caption{AffNet vs. Baumberg affine shape estimators on wide baseline stereo datasets, with Hessian and adaptive Hessian detectors, following the protocol~\cite{WXBS2015}. The number of matched image pairs and the average number of inliers. The \fbox{numbers} of image pairs in a dataset are boxed. Best results are in \textbf{bold}.
}
\setlength{\tabcolsep}{4pt}
\begin{tabular}{lrrrrrrrrrrrr}
\toprule
& \multicolumn{2}{c}{EF}
& \multicolumn{2}{c}{EVD}
& \multicolumn{2}{c}{OxAff}
& \multicolumn{2}{c}{SymB}
& \multicolumn{2}{c}{GDB}
& \multicolumn{2}{c}{LTLL}
\\
& \multicolumn{2}{c}{\cite{Zitnick2011}}
& \multicolumn{2}{c}{\cite{MODS2015}}
& \multicolumn{2}{c}{\cite{Mikolajczyk2005}}
& \multicolumn{2}{c}{\cite{Hauagge2012}}
& \multicolumn{2}{c}{\cite{Yang2007}}
& \multicolumn{2}{c}{\cite{LostInPast2015}}
\\
\cmidrule(r){2-3}
\cmidrule(r){4-5}
\cmidrule(r){6-7}
\cmidrule(r){8-9}
\cmidrule(r){10-11}
\cmidrule(r){12-13}
Detector
 & \fbox{33}& inl.
 & \fbox{15}& inl. 
 & \fbox{40}& inl.
 & \fbox{46}& inl.
 & \fbox{22}& inl.
 & \fbox{172}& inl.\\
\cmidrule(r){1-13}
HesAff~\cite{Mikolajczyk2004} & \textbf{33} & 78 & \textbf{2} & 38 & \textbf{40} & 1008 & 34 & 153 & 17 & 199 &26 & 34  \\
HesAffNet & \textbf{33} & \textbf{112} & \textbf{2} & \textbf{48} & \textbf{40} & \textbf{1181}& \textbf{37} & \textbf{203} & \textbf{19} & \textbf{222}& \textbf{46} & \textbf{36}  \\
\cmidrule(r){1-13}
AdHesAff~\cite{WXBS2015} & \textbf{33} & 111 & 3 & 33 & \textbf{40} & 1330 & 35 & 190 & 19 & 286 & 28 & 35 \\
AdHesAffNet&   \textbf{33} & \textbf{165} & \textbf{4} & \textbf{42} & \textbf{40} & \textbf{1567} & \textbf{37} & \textbf{275} & \textbf{21} & \textbf{336} &  \textbf{48} & \textbf{39}\\
\bottomrule
\end{tabular}
\label{tab:wxbs-table}
\end{table*}
\subsection{Wide baseline stereo}%
\label{exp:wbs}
We conducted an experiment on wide baseline stereo, following the local feature detector benchmark protocol, defined in~\cite{WXBS2015} on a set of two-view matching datasets~\cite{Hauagge2012,Yang2007,Zitnick2011,LostInPast2015}. The local features are detected by a benchmarked detector, described by HardNet++~\cite{HardNet2017} and HalfRootSIFT~\cite{HalfSIFT2007} and geometrically verified by RANSAC~\cite{Lebeda2012}. Two following metrics are reported: the number of successfully matched image pairs and the average number of correct inliers per matched pair. We have replaced the original affine shape estimator in Hessian-Affine with AffNet in Hessian and Adaptive threshold Hessian (AdHess) 

The results are shown in Table~\ref{tab:wxbs-table}. AffNet outperforms Baumberg in both the number of registered image pairs and/or the number of correct inliers in all datasets, including painting-to-photo pairs in SymB~\cite{Hauagge2012} and multimodal pairs in GDB~\cite{Yang2007}, despite it was not trained for that domains.

The total runtimes per image are the following (average for 800x600 images). Baseline HesAff + dominant gradient orientation + SIFT: no CNN components – 0.4 sec. HesAffNet (CNN) + dominant gradient orientation + SIFT – 0.8s, 3 CNN components: HesAffNet + OriNet + HardNet – 1.2 s. Now the data is naively transferred from CPU to GPU and back during each of the stages, which generates the major bottleneck.

\subsection{Image retrieval}%
\label{exp:retrieval}
\begin{table*}[tb]%
\ra{1}
\centering
\caption{Performance (mAP) evaluation of bag-of-words (BoW) image retrieval on the Oxford5k and Paris6k benchmarks. Vocabulary consisting of 1M visual words is learned on the independent dataset: Oxford5k vocabulary for Paris6k evaluation and \emph{vice versa}. SV: spatial verification. QE($t$): query expansion with $t$ inliers threshold. The best results are in \textbf{bold}.}
\label{tab:ox5kpar6k}
\setlength{\tabcolsep}{1mm}
\resizebox{\textwidth}{!}{%
\begin{tabular}{lc@{\hskip 5mm}c@{\hskip 3mm}ccc@{\hskip 5mm}c@{\hskip 3mm}cc}
\toprule
 & \multicolumn{4}{c}{Oxford5k} & \multicolumn{4}{c}{Paris6k} \\
 \cmidrule(r){2-5} \cmidrule(r){6-9}
Detector--Descriptor & \tiny{BoW} & \tiny{+SV} & \tiny{+SV+QE(15)} & \tiny{+SV+QE(8)} & \tiny{BoW} & \tiny{+SV} & \tiny{+SV+QE(15)} & \tiny{+SV+QE(8)} \\
\cmidrule(r){1-9}
HesAff--RootSIFT~\cite{RootSIFT2012} & 55.1	& 63.0 & 78.4 & 80.1 & 59.3 & 63.7 & 76.4 & 77.4 \\
HesAffNet--RootSIFT & 61.6 & 72.8 & 86.5 & 88.0 & 63.5 & 71.2 & 81.7 & 83.5 \\
\cmidrule(r){1-9}
HesAff--TFeat-M*~\cite{TFeat2016} & 46.7 & 55.6 & 72.2 & 73.8 & 43.8 & 51.8 & 65.3 & 69.7\\
HesAffNet--TFeat-M* & 45.5 & 57.3 & 75.2 & 77.5 & 50.6 & 58.1 & 72.0 & 74.8\\
\cmidrule(r){1-9}
HesAff--HardNet++~\cite{HardNet2017} & 60.8 & 69.6 & 84.5 & 85.1 & 65.0 & 70.3 & 79.1 & 79.9\\
HesAffNetLess--HardNet++ & 64.3 & 73.3 & 86.1 & 87.3 & 62.0 & 68.7 & 79.1 & 79.2\\
HesAffNet--HardNet++ & \textbf{68.3} & \textbf{77.8} & \textbf{89.0} & \textbf{91.1} & \textbf{65.7} & \textbf{73.4} & \textbf{83.3} & \textbf{83.3}\\
\bottomrule
\end{tabular}%
}
\end{table*}
\begin{table*}[tb] %
\ra{1}
\centering
\caption{Performance (mAP) comparison with the state-of-the-art in local feature-based image retrieval. Vocabulary is learned on the independent dataset: Oxford5k vocabulary for Paris6k evaluation and \emph{vice versa}. All results are with spatial verification and query expansion. VS: vocabulary size. SA: single assignment. MA: multiple assignments. The best results are in \textbf{bold}.}
\label{tab:ox5kpar6kSOTA}
\setlength{\tabcolsep}{3mm}
\begin{tabular}{lccccc}
\toprule
 & & \multicolumn{2}{c}{Oxford5k} & \multicolumn{2}{c}{Paris6k} \\
 \cmidrule(r){3-4} \cmidrule(r){5-6}
Method & VS & SA & MA & SA & MA\\
\cmidrule(r){1-6}
HesAff--SIFT--BoW-fVocab~\cite{Mikulik-IJCV2013FineVocab} & 16M & 74.0 & 84.9 & 73.6 & 82.4 \\
HesAff--RootSIFT--HQE~\cite{Tolias-PR2014HQE} & 65k & 85.3 & 88.0 & 81.3 & 82.8 \\
HesAff--HardNet++--HQE~\cite{HardNet2017} & 65k & 86.8 & 88.3 & 82.8 & 84.9 \\
\cmidrule(r){1-6}
HesAffNet--HardNet++--HQE & 65k & \textbf{87.9} & \textbf{89.5} & \textbf{84.2} & \textbf{85.9} \\
\bottomrule
\end{tabular}
\end{table*}%
\begin{table*}[tb] %
\ra{1.0}
\centering
\caption{Performance (mAP, mP@10) comparison with the state-of-the-art in image retrieval on the R-Oxford and R-Paris benchmarks~\cite{RevisitOxford2018}. SV: spatial verification. HQE: hamming query expansion. $\alpha$QE: $\alpha$ query expansion.  DFS: global diffusion. 
The best results are in \textbf{bold}.}
\label{tab:Rox5kpar6kSOTA}
\setlength{\tabcolsep}{1mm}
\resizebox{\textwidth}{!}{%
\begin{tabular}{lcccccccc}
\toprule
 & \multicolumn{4}{c}{Medium} & \multicolumn{4}{c}{Hard} \\
 \cmidrule(r){2-5} \cmidrule(r){6-9}
 & \multicolumn{2}{c}{R-Oxford} & \multicolumn{2}{c}{R-Paris} & \multicolumn{2}{c}{R-Oxford} & \multicolumn{2}{c}{R-Paris} \\
\cmidrule(r){2-3} \cmidrule(r){4-5} \cmidrule(r){6-7} \cmidrule(r){8-9}
Method & mAP & mP@10 & mAP & mP@10 & mAP & mP@10 & mAP & mP@10 \\
\cmidrule(r){1-9}
ResNet101--GeM+$\alpha$QE~\cite{Radenovic-arXiv17} & 67.2 & 86.0 & 80.7 & \textbf{98.9} & 40.7 & 54.9 & 61.8 & 90.6 \\
ResNet101--GeM\cite{Radenovic-arXiv17}+DFS~\cite{Iscen2017CVPR} & 69.8 & 84.0 & 88.9 & 96.9 & 40.5 & 54.4 & 78.5 & \textbf{94.6} \\
ResNet101--R-MAC\cite{Gordo-IJCV17}+DFS~\cite{Iscen2017CVPR} & 69.0 & 82.3 & \textbf{89.5} & 96.7 & 44.7 & 60.5 & \textbf{80.0} & 94.1 \\
ResNet50--DELF\cite{DELF2017}--HQE+SV & 73.4 & 88.2 & 84.0 & 98.3 & 50.3 & 67.2 & 69.3 & 93.7 \\
\cmidrule(r){1-9}
HesAff--RootSIFT--HQE~\cite{Tolias-PR2014HQE} & 66.3 & 85.6 & 68.9 & 97.3 & 41.3 & 60.0 & 44.7 & 79.9 \\
HesAff--RootSIFT--HQE+SV~\cite{Tolias-PR2014HQE} & 71.3 & 88.1 & 70.2 & 98.6 & 49.7 & 69.6 & 45.1 & 83.9 \\
\cmidrule(r){1-9}
HesAffNet--HardNet++--HQE & 71.7 & 89.4 & 72.6 & 98.1 & 47.5 & 66.3 & 48.9 & 85.9 \\
HesAffNet--HardNet++--HQE+SV & \textbf{75.2} & \textbf{90.9} & 73.1 & 98.1 & \textbf{53.3} &
\textbf{72.6} & 48.9 & 89.1 \\
\bottomrule
\end{tabular}
}
\end{table*}%
We evaluate the proposed approach on standard image retrieval datasets Oxford5k~\cite{Philbin07} and Paris6k~\cite{Philbin08}.
Each dataset contains images (5062 for Oxford5k and 6391 for Paris6k) depicting 11 different landmarks and distractors. The performance is reported as mean average precision (mAP)~\cite{Philbin07}.
Recently, these benchmarks have been revisited, annotation errors fixed, and new, more challenging sets of queries added~\cite{RevisitOxford2018}. The revisited datasets define new test protocols: \textit{Easy}, \textit{Medium}, and \textit{Hard}.

We use the multi-scale Hessian-affine detector~\cite{Mikolajczyk2005} with the Baumberg method for affine shape estimation. The proposed AffNet replaces Baumberg, which we denote HessAffNet. The use of HessAffNet increased the number of used features from 12.5M to 17.5M for Oxford5k and from 15.6M to 21.2M for Paris6k, because more features survive the affine shape adaptions, as explained in Section~\ref{exp:rep}. We also performed an additional experiment by restricting the number of AffNet features to the same as in Baumberg -- HesAffNetLess in Table~\ref{tab:ox5kpar6k}. We evaluated HesAffNet with both hand-crafted descriptor RootSIFT~\cite{RootSIFT2012} and state-of-the-art learned descriptors~\cite{TFeat2016,HardNet2017}.%

First, HesAffNet is tested within the traditional bag-of-words~(BoW)~\cite{Sivic-ICCV2003VideoGoogle} image retrieval pipeline. A flat vocabulary with 1M centroids is created with the k-means algorithm and approximate nearest neighbor search~\cite{FLANN2009}. All descriptors of an image are assigned to a respective centroid of the vocabulary, and then they are aggregated with a histogram of occurrences into a BoW image representation.%

We also apply spatial verification~(SV)~\cite{Philbin07} and the standard query expansion~(QE)~\cite{Philbin08}. QE is performed on images that have either 15 (typically used) or 8 inliers after the spatial verification. %
 The results of the comparison are presented in Table~\ref{tab:ox5kpar6k}.

AffNet achieves the best results on both Oxford5k and Paris6k datasets, in most of the cases it outperforms the second best approach by a large margin. This experiment clearly shows the benefit of using AffNet in the local feature detection pipeline.

Additionally, we compare with state-of-the-art local-feature-based image retrieval methods.
A visual vocabulary of 65k words is learned, with Hamming embedding (HE)~\cite{Jegou-IJCV2010HE} technique added that further refines the descriptor assignments with a 128 bits binary signature. We follow the same procedure as HesAff--RootSIFT--HQE~\cite{Tolias-PR2014HQE} method.
All parameters are set as in~\cite{Tolias-PR2014HQE}. 
The performance of AffNet methods is the best, reported on both Oxford5k and Paris6k for local features. 

Finally, on the revisited R-Oxford and R-Paris, we compare with state-of-the-art methods in image retrieval, both local and global feature based: the best-performing fine-tuned networks~\cite{He2016ResNet}, ResNet101 with generalized mean pooling (ResNet101--GeM)~\cite{Radenovic-arXiv17} and ResNet101 with regional maximum activations pooling (ResNet101--R-MAC)~\cite{Gordo-IJCV17}. Deep methods use re-ranking methods: $\alpha$ query expansion ($\alpha$QE)~\cite{Radenovic-arXiv17}, and global diffusion (DFS)~\cite{Iscen2017CVPR}. Results are in Table~\ref{tab:Rox5kpar6kSOTA}. 

HesAffNet performs best on the R-Oxford. It is consistently the best performing local feature method, yet is worse than deep methods on R-Paris. A  possible explanation is that deep networks (ResNet and DELF) were finetuned from ImageNet, which contains Paris-related images, e.g., Sacre Coeur and Notre Dame Basilica in the ``church'' category. Therefore global deep nets are partially evaluated on the training set. %

\section{Summary}
\label{affnet:conclusions}
We presented a method for learning the affine shape of local features in a weakly supervised manner. The proposed HardNegC loss function might find other application domains as well. Our intuition is that the distance to the hard-negative estimates the local density of all points and provides a scale for the positive distance. %
The resulting AffNet regressor bridges the gap between the performance of the similarity-covariant and affine-covariant detectors on images with short baseline and big illumination differences and it improves the performance of affine-covariant detectors in the wide baseline setup. AffNet applied to the output of the Hessian detector improves the state-of-the art in wide baseline matching, affine detector repeatability and image retrieval.

We experimentally show that descriptor matchability, not only repeatability, should be taken into account when learning a feature detector.

\chapter[MODS: Matching with On-Demand View Synthesis]{MODS: Matching with On-Demand View Synthesis \\
\includegraphics[width=0.98\linewidth]{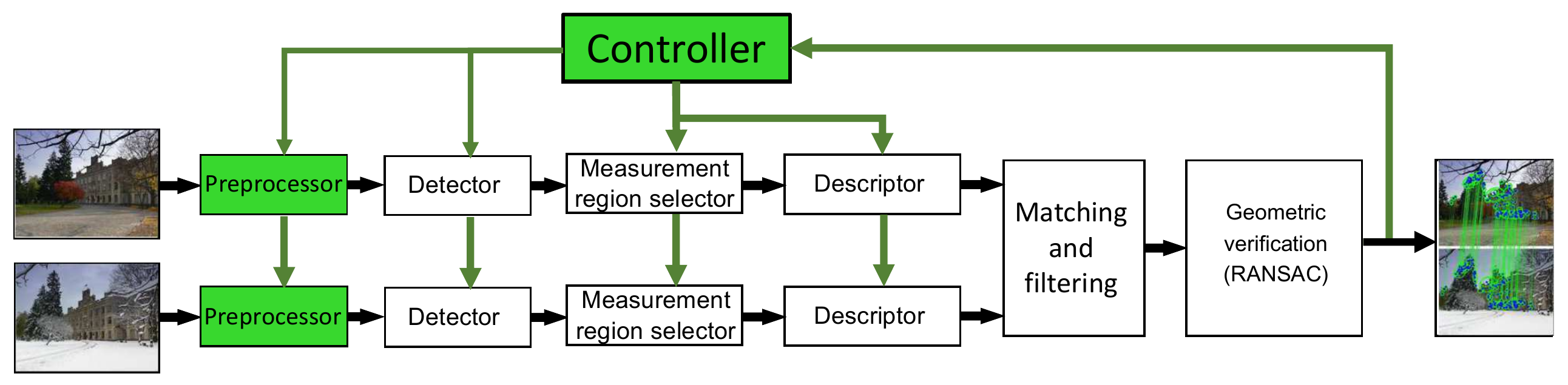}}
\label{ch:mods}
We have discussed in the previous Chapters how difficult can be the WxBS problem. In this Chapter we study a problem of matching image pairs, taken from extreme viewpoints. We introduce a benchmark and the Extreme View Dataset (EVD) and show that the standard WxBS pipeline performs poorly on this dataset. We present a solution -- by transforming a standard feedforward pipeline into a controlled system by adding a feedback loop and on-demand view synthesis. The resulting algorithm called MODS -- matching with on-demand view synthesis -- is able to match not only images acquired under oblique camera angles, but improves on other WxBS problems as well, like matching historical photographies. 

\section{Introduction}
\begin{figure}[htb]
\centering
\includegraphics[width=0.95\linewidth]{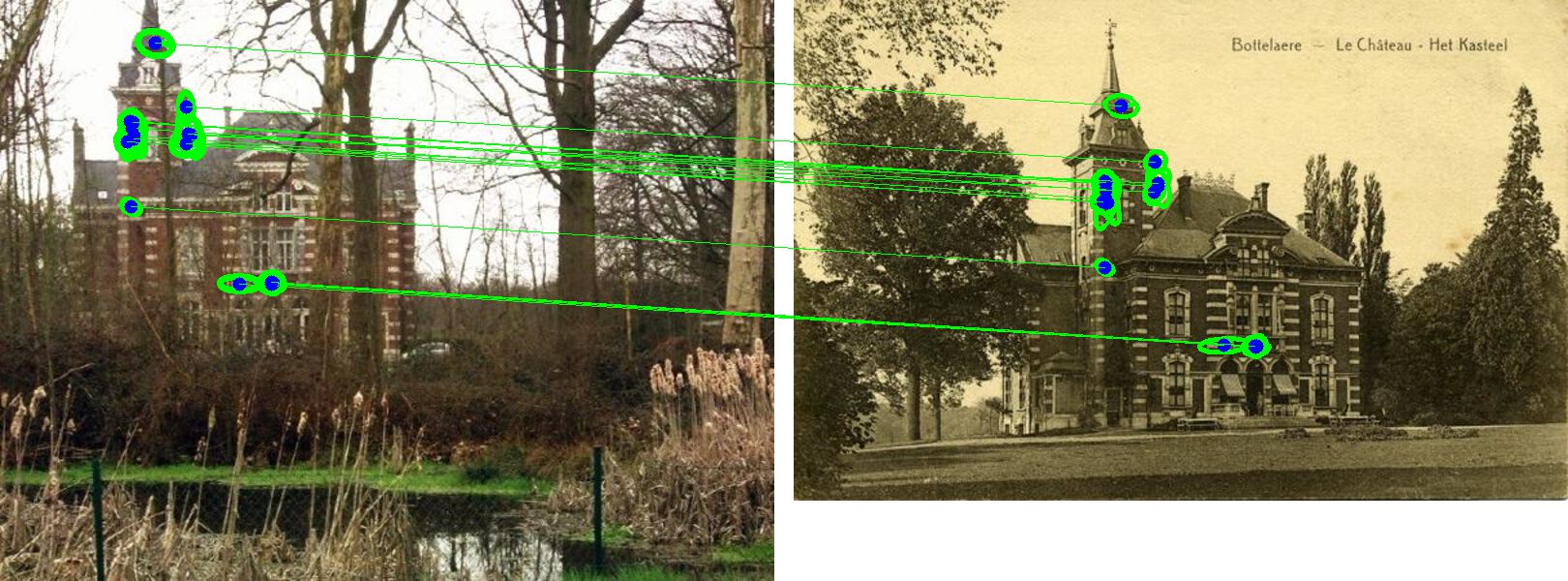}
\caption{A pair of historical photographies, that is not matchable by standard methods. Matched only with help of affine view synthesis in the MODS\cite{MODS2015} framework. Recent historical virtual reality application uses MODS features~\cite{MODS-User2020}.}
\label{fig:mods-match-lost-in-past}
\end{figure}

\begin{figure}
\centering
\includegraphics[width=0.9\linewidth]{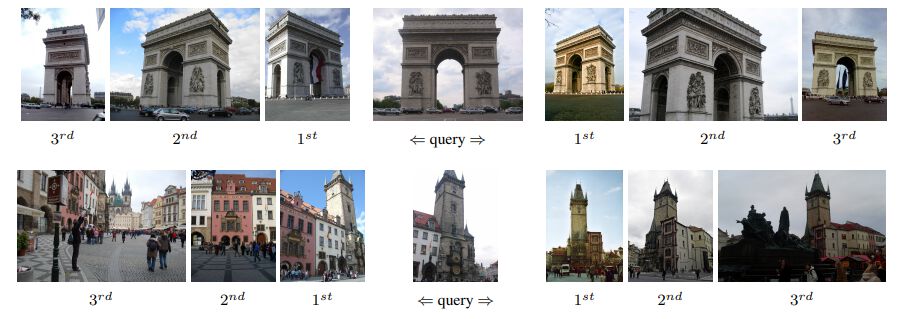}
\caption{Frontal and side views of the Arc de Triomphe (top) and Astronomical Clock (bottom). One cannot match leftmost and rightmost views, but can register together those images if there are enough proxy images in between them. Figure from ~\cite{SingleImage3dRec2015}.}
\label{fig:seeing-around-corner}
\end{figure}

\begin{figure}
\centering
\begin{subfloat}[]{
\includegraphics[width=0.39\linewidth]{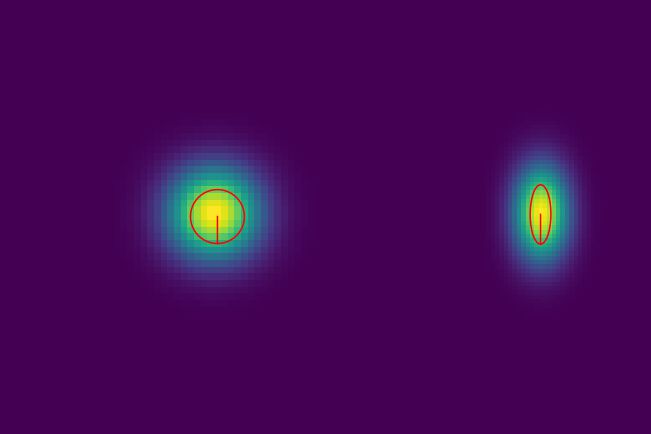}}
\end{subfloat}%
\begin{subfloat}[]{
\includegraphics[width=0.55\linewidth]{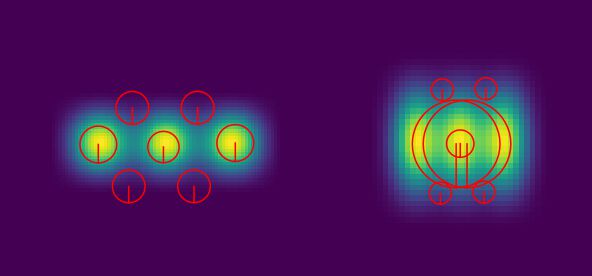}}
\end{subfloat}%
\caption{Hessian-Affine detections of the synthetic blobs. (a) Detection of the single isolated blog, face-on (left) and from the oblique angle (right). The same support region is detected. (b) Detection of the several blobs near to each other, face-on (left) and from the oblique angle (right). The detections are merging, when the blobs are pictured from a side viewpoint. This results in a failure of the blob detector invariance.}
\label{fig:hessian-fail}
\end{figure}

Standard wide-baseline stereo or 3D reconstruction pipelines work well in many situations. Even if an image pair is not matched, it is usually not a problem. For example, one could still register images from very different viewpoints, if there is a sequence of images in between, as shown in Figure~\ref{fig:seeing-around-corner}, from "From Single Image Query to Detailed 3D Reconstruction" paper~\cite{SingleImage3dRec2015}.
However, that might not always be possible. For example, the number of pictures is limited because they are historical~\cite{MODS-User2019, MODS-User2020} and there is no way how one could go and take more without inventing a time machine. What to do? One way would be to use affine features like Hessian-AffNet\cite{AffNet2018} or MSER\cite{MSER2002}. However, they help only up to some extent and what if the views we need to match are more extreme? See an example in Figure~\ref{fig:mods-match-lost-in-past}.  Let us consider an image which consists of three blobs, which are situated close to each other. Under the tilt transform they merge into a single blob, as shown in Figure~\ref{fig:hessian-fail}. It is not the shape of the region, which is detected incorrectly, but the centers of the features themselves. 
One way to address the problem is to simulate the real camera viewpoint change by affine or perspective warps of the current image. This idea was first proposed by Lepetit and Fua in 2006\cite{AffineTree2006}. They synthesized views to find distinctive keypoints repeatedly detectable under affine deformations. Synthetic views provided a training set for learning a random forest classifier that labeled individual feature points. Feature points in different images with the same label were assumed to be in correspondence. The simple keypoint detector of Lepetit and Fua is very fast, but invariant only to translation and rotation and thus the number of views necessary to achieve acceptable repeatability was high. The method was tested on pairs undergoing significant affine transformations, but the final representation did not scale and can not be easily used for indexing.

\begin{algorithm}[tbp]
\small
\caption{{\bf ASIFT}}
\label{alg:ASIFT}
\begin{algorithmic}
\Require{$I_1$, $I_2$ -- two images.}
\Ensure{List of corresponding points; fundamental matrix F.}\\
\algrule
\For {$I_1$ and $I_2$ independently}
\State \textbf{1.1 Generate synthetic views} according to the tilt-rotation-detector setup.
\State \textbf{1.2 Detect and describe local features}.
\EndFor
\State \textbf{2 Generate tentative correspondences} for each pair of the views synthesized from $I_1$ and $I_2$ \\
\hskip1em $I_1$ and $I_2$ independently using the 2nd closest ratio. 
\State \textbf{3 Add correspondences to the global list}. \\
\hskip1em Reproject the corresponding features to the original images. 
\State \textbf{4 Filter duplicate, ``one-to-many'' and ``many-to-one'' matches}. 
\State \textbf{5 Geometrically verify tentative correspondences} using ORSA~\cite{Moisan2004}\\ 
\hskip1em while estimating F.
\end{algorithmic}
\end{algorithm}

Later, Morel~\etal~\cite{ASIFT2009} proposed a new matching pipeline -- see Alg.~\ref{alg:ASIFT}. The authors showed that view synthesis extends the handled range of viewpoint differences.
The ASIFT algorithm starts by generating synthetic views (described in Section~\ref{mods:synthetic-view-generation}) for both images. Next, feature detection and description are performed using standard SIFT~\cite{SIFT2004} in each synthesized view. Tentative correspondences are formed for all pairs of views synthesized  from the first and second image. The matching stage thus entails $n^2$ independent matching problems, where $n$ is the number of synthesized views per image. The set of correspondences between the images is the union of results for all synthesized pairs.
The \emph{duplicate filtering} stage of ASIFT prunes correspondences with small spatial distance (2~pixels) of local features in both images -- all such correspondences except one (random) are eliminated from the final correspondence set.  ``\emph{One-to-many}'' correspondences -- correspondences of features which are close to each other (are situated in the radius of $\sqrt{2}$ pixels) in one image while spread in other synthetic views are also eliminated, despite the fact that some of the pairs can be correct.
Finally, geometric verification is performed by ORSA~\cite{Moisan2004}. ORSA is a RANSAC-based method, which exploits an a-contrario approach to detect incorrect epipolar geometries. Instead of having a constant error threshold, ORSA looks for matches that have the highest ``diameter'', i.e., matches which cover a large image area. ASIFT was shown to match images of a scene with a viewpoint difference up to $80^\circ$ for planar objects \cite{ASIFT2009}. Computational costs are in the order of tens of seconds to a few minutes.

The latest extensions of wide-baseline matching pipeline are limited to modifications of the ASIFT algorithm. Liu~\etal~\cite{Liu2012} synthesized perspective warps rather than affine. Pang~\etal~\cite{Pang2012} replaced SIFT by SURF~\cite{Bay2006} in the ASIFT algorithm to reduce the computation time. 

\section{The MODS algorithm}
\label{mods:algorithm}
The main idea of the proposed iterative MODS algorithm (see Alg.~\ref{alg:MODS}) is to repeat a sequence of two-view matching procedures, until a required number of geometrically verified correspondences is found. In each iteration, a different and potentially complementary detector is used and a different set of views is synthesized.
The algorithm starts with fast detectors with limited invariance, proceeding progressively to more complex, robust, but computationally costly ones. MODS is thus capable of solving simple matching problems fast without losing the ability to deal with very difficult cases where a combination of detectors is employed to extend the state-of-the-art.

The adopted sequence of detectors and view synthesis  parameters is an outcome of an extensive experimental search.  The objective was to solve the most challenging problems in  the development set, i.e., to correctly recover their two-view geometry,  while keeping the speed comparable to standard single-detector wide-baseline matchers for simple problems.  Details about the
selected configuration and the optimization process are given in Section~\ref{mods:parameters}. The rest of the section describes the steps involved in the iterations of the MODS algorithm, which is compared to the standard two-view matching and ASIFT pipelines.

\begin{algorithm}[tb]
\small
\caption{{\bf MODS}}
\label{alg:MODS}
\begin{algorithmic}
\Require{$I_1$, $I_2$ -- two images; $\theta_m$ -- minimum required number of matches; \\$S_{\text{max}}$ -- maximum number of iterations. }
\Ensure{Fundamental or homography matrix F or H; list of corresponding points.}\\
\hskip-1em\textbf{Variables}: $N_{\text{matches}}$ -- detected correspondences, $\text{Iter}$ -- current iteration.
\algrule
\While {$(N_{\text{matches}} < \theta_{m})$ \textbf{and} $(\text{Iter} < S_{\text{max}})$} 
\For {$I_1$ and $I_2$ independently}
\State \textbf{1.1 Generate synthetic views} according to the \\
\hskip3em scale-tilt-rotation-detector-descriptor setup for the Iter (Tables~\ref{tab:MODS-configuration}, \ref{tab:view-synth-configuration}).
\State \textbf{1.2 Detect and describe local features}.
\State \textbf{1.3 Reproject local features} to the original image. \\ 
\hskip3em Add the described features to the global list.
\EndFor
\State \textbf{2 Generate tentative correspondences} using the first geom. inconsistent rule. 
\State \textbf{3 Filter duplicate matches}. 
\State \textbf{4 Geometrically verify tentative correspondences} with DEGENSAC~\cite{Degensac2005}\\
\hskip2em while estimating F or H.
\EndWhile
\end{algorithmic}
\end{algorithm}
\subsection{Synthetic views generation} %
\begin{figure}[tb]%
\centering
\includegraphics[width=0.49\linewidth]{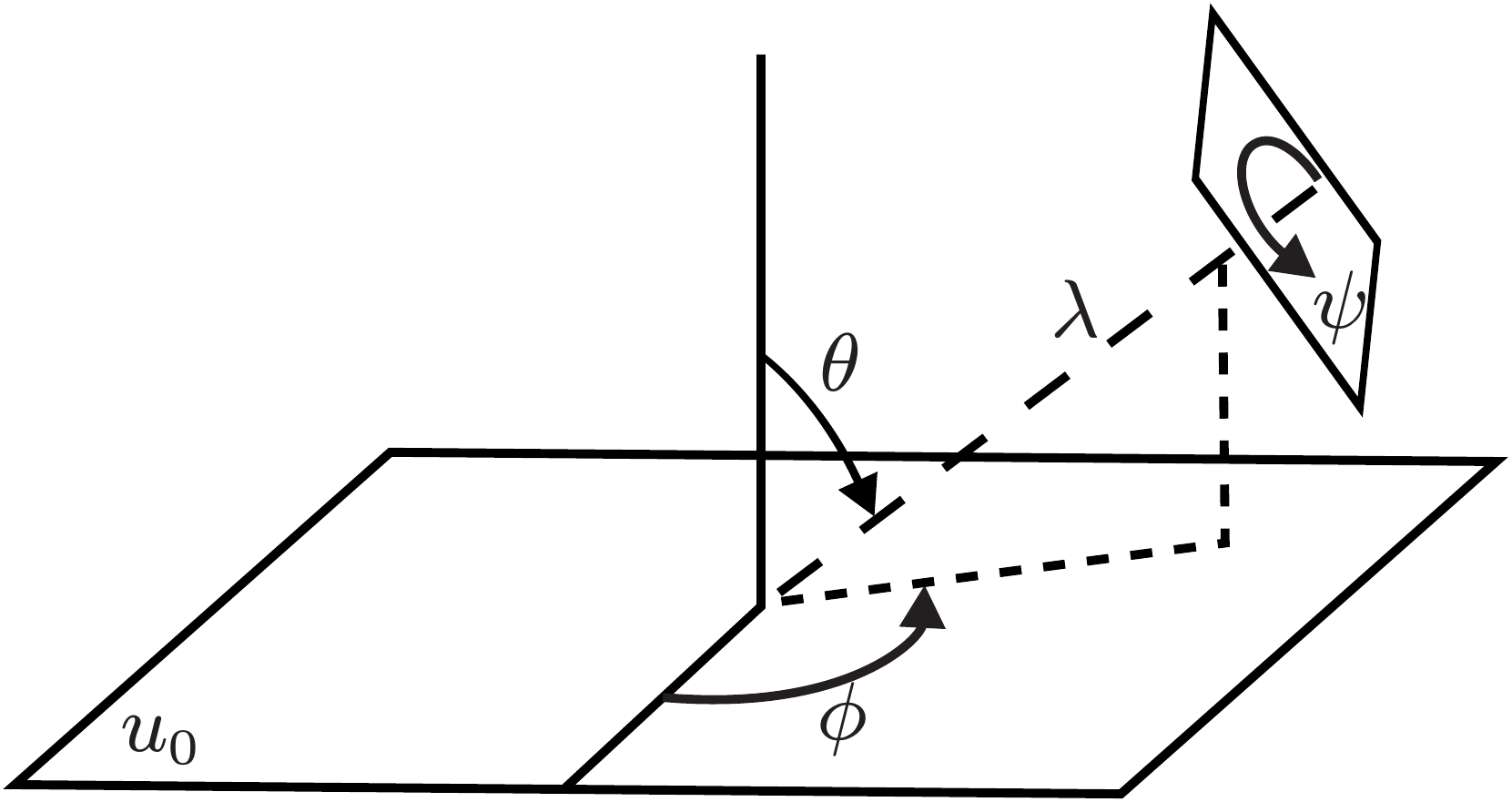}
\includegraphics[width=0.49\linewidth]{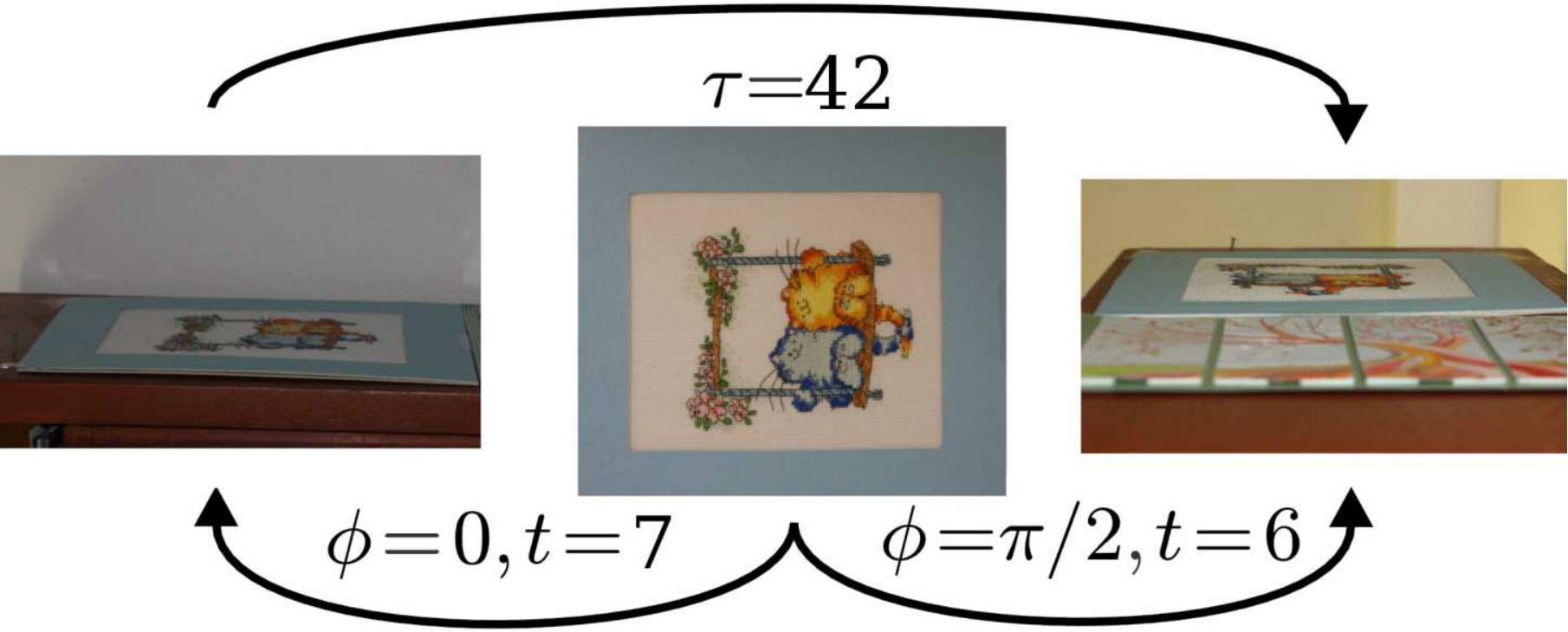}
\caption{(left) the affine camera model~(\ref{eq:affine-matrix-decomposition}). Latitude $\theta = \arccos{1/t}$ -- latitude, longitude  $\phi$, scale $\lambda$ scale. (right) Transitional tilt $\tau$ for absolute tilt $t$ and rotation $\phi$.}
\label{fig:affine-camera}
\end{figure}

\label{mods:synthetic-view-generation}
MODS (Alg.~\ref{alg:MODS}) starts by synthetic view generation.  It is well known that a homography $H$ can be approximated by an affine transformation $A$ at a point using the first order Taylor expansion. The affine transformation can be uniquely decomposed by SVD into a rotation, skew, scale and rotation around the optical axis~\cite{Hartley2004}. In \cite{ASIFT2009}, the authors proposed to decompose the affine transformation $A$ as 
\begin{equation}
\begin{array}{rcl}
\label{eq:affine-matrix-decomposition}
\scriptstyle
A&=&H_\lambda{}R_1(\psi)T_t R_2 (\phi) = \\[3pt]
&=&\lambda{}%
\left(
\scriptstyle
\begin{array}{@{\hspace{0pt}}c@{\hspace{6pt}}c@{\hspace{0pt}}}
\cos{\psi} & -\sin{\psi} \\
\sin{\psi} & \cos{\psi}\\
\end{array}\right)\!\left(%
\scriptstyle
\begin{array}{
@{\hspace{0pt}}c@{\hspace{6pt}}c@{\hspace{0pt}}}
t & 0  \\
0 & 1 \\        
\end{array}\right)\!\left(%
\scriptstyle
\begin{array}{
@{\hspace{0pt}}c@{\hspace{6pt}}c@{\hspace{0pt}}}
\cos{\phi} & -\sin{\phi}\\
\sin{\phi} & \cos{\phi}\\
\end{array}\right)
\end{array}
\end{equation}
where $\lambda{} > 0$, $R_1$ and $R_2$ are rotations, and $T_t$ is a diagonal matrix with $t > 1$. Parameter $t$ is called the absolute tilt, $\phi \in \langle 0,\pi)$ is the optical axis longitude and $\psi \in \langle 0 ,2\pi)$ is the rotation of the camera around the optical axis (see Figure~\ref{fig:affine-camera}). Each synthesized view is thus parametrized by the tilt, longitude and optionally the scale and represents a sample of the view-sphere resp. view-volume around the original image.

The view synthesis proceeds in the following steps: at first, a scale synthesis is performed by building a Gaussian scale-space with Gaussian $\sigma={\sigma_{\text{base}}}\cdot{S}$ and downsampling factor $S$ ($S<1$). Then, each image in the scale-space is in-plane rotated by longitude $\phi$ with step $\Delta\phi={\Delta\phi_{\text{base}}}/{t}$. In the third step, all rotated images are convolved with a Gaussian filter with $\sigma = \sigma_{\text{base}}$ along the vertical  and  $\sigma = t\cdot{}\sigma_{\text{base}}$ along the horizontal direction to eliminate aliasing in a final tilting step. The final tilt is applied by shrinking the image along the horizontal direction by factor $t$. The synthesis parameters are: the set of scales \{S\}, $\Delta\phi_{\text{base}}$ -- the longitude sampling step at tilt $t=1$, the set of simulated tilts~\{t\}.
\subsection{Local feature detection and description} %
The second step of MODS is the detection and description of local features.
It is known that different local feature detectors are suitable for different types of images~\cite{Mikolajczyk05} and that some detectors are complementary in the image structures they respond to~\cite{Aanaes2012}. Our experiments show  that combining detectors improves the overall robustness and speed of the matching procedure.

MODS combines a fast similarity covariant FAST (in ORB implementation) detector and affine covariant detectors MSER and Hessian-Affine. 
The normalized patches are described by the binary descriptor BRIEF~\cite{ORB2011} (in ORB implementation) and a recent modification of SIFT~\cite{SIFT2004} -- the RootSIFT~\cite{RootSIFT2012}. The local feature frames computed on the synthesized views are backprojected to the coordinate system of the original image by the known affine matrix $A$ and associated with the descriptor and the originating synthetic view. MODS step configurations are specified in Table~\ref{tab:MODS-configuration}. 

For the MSER and Hessian-Affine detectors, the fast affine feature extraction process from~\cite{PatchExtraction2014} was applied.

\begin{table}[tb]
\caption{MODS step configurations are defined by a detector, descriptor and the set of the synthesized views. RootSIFT is used for all detectors but ORB which is described by BRIEF.}
\label{tab:MODS-configuration}
\centering
\bgroup
\begin{tabular}{ll}%
\toprule
Iter. & Setup \\  
\cmidrule(r){1-2}
1 & ORB,$\{S\} = \{1\}$, $\{t\} = \{1\}$, $\Delta\phi=360\degree{}/t$ \\    %
2 & ORB,$\{S\} = \{1\}$, $\{t\} = \{1;5;9\}$, $\Delta\phi=360\degree{}/t$ \\ %
3 & MSER,$\{S\} = \{1;0.25;0.125\}$, $\{t\} = \{1\}$, $\Delta\phi=360\degree{}/t$\\ %
4 & MSER,$\{S\} = \{1;0.25;0.125\}$, $\{t\} = \{1;3;6;9\}$, $\Delta\phi=360\degree{}/t$\\
5 & HessAff, $\{S\} = \{1\}$, $\{t\} = \{1;2;4;6;8\}$, $\Delta\phi=360\degree{}/t$\\     %
6 & HessAff, $\{S\} = \{1\}$, $\{t\} = \{1;2;4;6;8\}$, $\Delta\phi=120\degree{}/t$\\  %
7 & HessAff, $\{S\} = \{1\}$, $\{t\} = \{1;2;4;6;8;10\}$, $\Delta\phi=60\degree{}/t$ \\
\bottomrule
\end{tabular}
\egroup
\end{table}
\begin{table}[tb]
\caption{MODS variants tested. The corresponding steps are the same as in Table~\ref{tab:MODS-configuration}. The starred MODS:$3^{*}$-$6^{*}$ configuration slightly differs from Table~\ref{tab:MODS-configuration} and corresponds to~\cite{Mishkin2013}.}
\label{tab:MODS-variants}
\centering
\bgroup
\tabulinesep=0.4mm
\begin{tabular}{llllllll}
\toprule
Name & \multicolumn{7}{c}{Steps}\\
\cmidrule(r){1-8}
MODS == MODS:1-7 & 1 & 2 & 3 & 4 & 5 &6 &7\\
MODS:2-7 & - & 2 & 3 & 4 & 5 &6 &7\\
MODS:$3^{*}$-6& - & - & $3^{*}$ & $4^{*}$ & 5 &6 &-\\
MODS:3-7& - & - & 3 & 4 & 5 &6 &7\\
MODS:3,2-7& 3 & - & 2 & 4 & 5 &6 &7\\
MODS:2,4,7& - & 2 & - & 4 & - &- &7\\
MODS:5,1-7& 5 & 1 & 2 & 3 & 4 &6 &7\\
\bottomrule
\end{tabular}
\egroup
\end{table}
\subsection{Tentative correspondence generation}
\begin{figure}[tb]%
\centering
\includegraphics[width=1\linewidth]{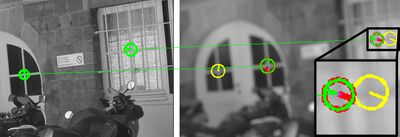}\\
\vspace{1 mm}
\includegraphics[width=1\linewidth]{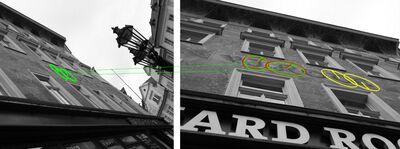}
\caption{Green regions -- correct correspondences rejected by the standard {first to second closest ratio} test (second closest region is in red), but recovered  by the {first to first geometrically inconsistent ratio} (first geometrically inconsistent region is in yellow) matching strategy. DoG regions (top), MSERs (bottom).}
\label{fig:1st-inconsistent-illustration}
\end{figure}

The next step of the MODS algorithm is the generation of tentative correspondences.
Different strategies for the computation of tentative correspondences in wide-baseline matching were proposed. The standard method for matching SIFT(-like) descriptors is based on the ratio of the distances to the closest and the second closest descriptors in the other image~\cite{SIFT2004}.
While the performance of this test is in general very good, it degrades when multiple observations of the same feature are present. 
In this case, the presence of similar descriptors will lead to the first to second SIFT ratio to be close to 1 and the correspondences will "annihilate" each other, despite the fact that they represent the same geometric constraints and are therefore not mutually contradictory (see Figure~\ref{fig:1st-inconsistent-illustration}). 
The problem of multiple detections is amplified in matching by view synthesis since covariantly detected local features are often repeatedly discovered in multiple synthetic views. 

To address this problem, we propose a modified matching strategy denoted \emph{first to first geometrically inconsistent -- FGINN} . Instead of comparing the first to the second closest descriptor distance, the distance of the first descriptor and the closest descriptor that is geometrically inconsistent with the first one is used. We call descriptors in one image {geometrically inconsistent} if the Euclidean distance between the centers of the regions is $\geq n$ pixels (default: $n=10$). 
The difference of the first-to-second closest ratio strategy and the FGINN strategy is illustrated in~Figure~\ref{fig:1st-inconsistent-illustration}.

\subsection{Geometric verification}
The last step of the MODS is the geometric verification. It consists of three substeps. 

\subsubsection{Duplicate filtering} 
The redetection of covariant features in synthetic views results in duplicates in tentative correspondences. The \emph{duplicate filtering} prunes correspondences with close spatial distance ($\approx$ 10~pixels) of local features in both images -- all these correspondences except one -- with the smallest descriptor distance ratio -- are eliminated from the final correspondences list.
The number of pruned correspondences can be used later for evaluating the quality (probability of being correct) in PROSAC-like~\cite{PROSAC2005} geometric verification.
\subsubsection{RANSAC}
The LO-RANSAC~\cite{LOransac2003} algorithm searches for the maximal
set of geometrically consistent tentative correspondences. The model of the
transformation is set either to homography or epipolar geometry, or
automatically determined by a DegenSAC~\cite{Degensac2005} procedure.

\subsubsection{Local affine frame check}
\begin{figure}[hptb]]%
\centering
\includegraphics[width=0.99\linewidth]{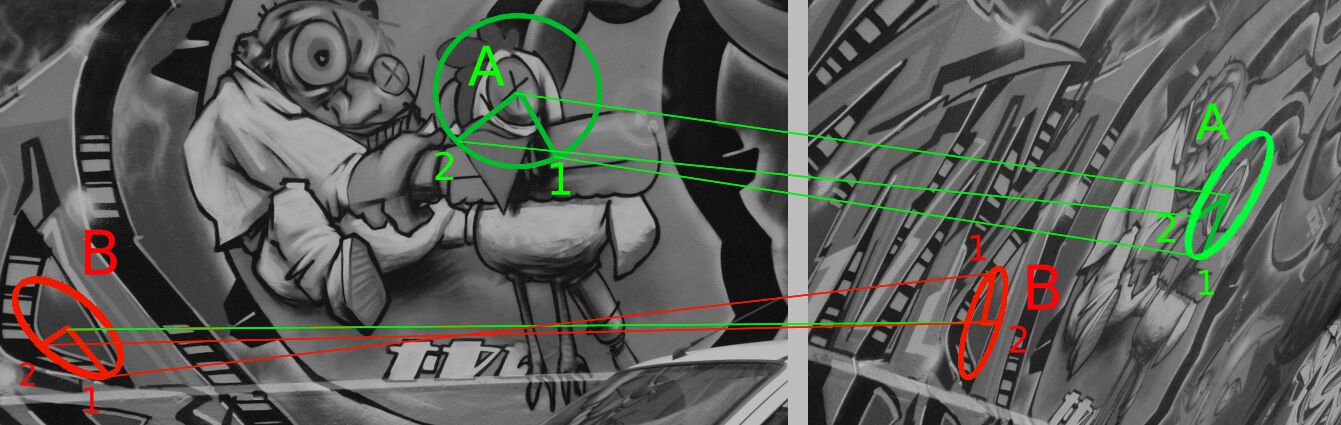}
\caption{The LAF-check. While centers of both regions A and B are consistent with found homography, farthest (1) and closest (2) points of the ellipse pass the check only for region A.}
\label{fig:laf-check}
\end{figure}%

Since the epipolar geometry constraint is much less restrictive than a homography, wrong correspondences consistent with some (random) fundamental matrix appear. The local affine frame consistency check (LAF-check) eliminates virtually all incorrect correspondences. The procedure uses coordinates of the closest and furthest ellipse points from the ellipse center of both matched local affine frames to check whether the whole local feature is consistent with the estimated geometry model (see Figure~\ref{fig:laf-check}). 
The check is performed with the geometric model obtained by RANSAC. Regions which do not pass the check are discarded from the list of inliers. If the number of correspondences after the LAF-check is fewer than the user defined minimum, the matcher continues with the next step of view synthesis. 

\section{Implementation and parameter setup}
\label{mods:parameters}
In this subsection, we discuss the tilt-rotation-detector setups of the MODS algorithm, and threshold selection for the \emph{first to first geometrically inconsistent -- FGINN} matching strategy validation.

\subsection{View synthesis for different detectors and descriptors}
The two main parameters of the view synthesis, tilt \{t\} sampling and the latitude step $\Delta\phi_{\text{base}}$, were explored in the following synthetic experiment. 

\begin{figure}[tb]
\centering
\includegraphics[height=1.65cm]{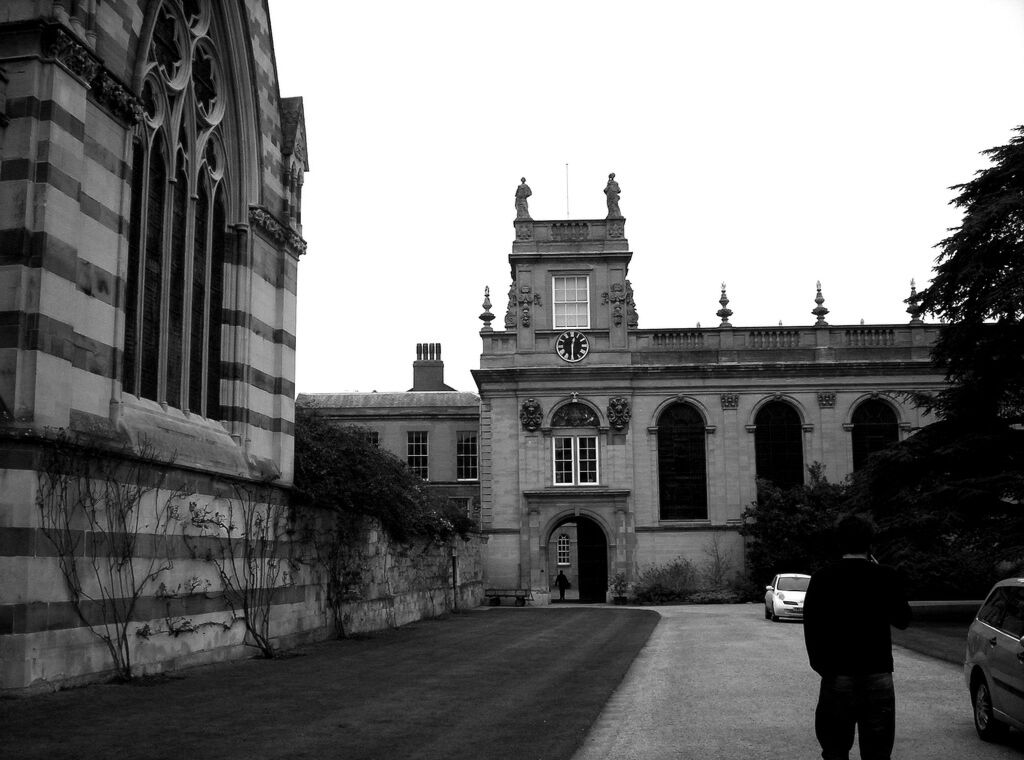}
\includegraphics[height=1.65cm]{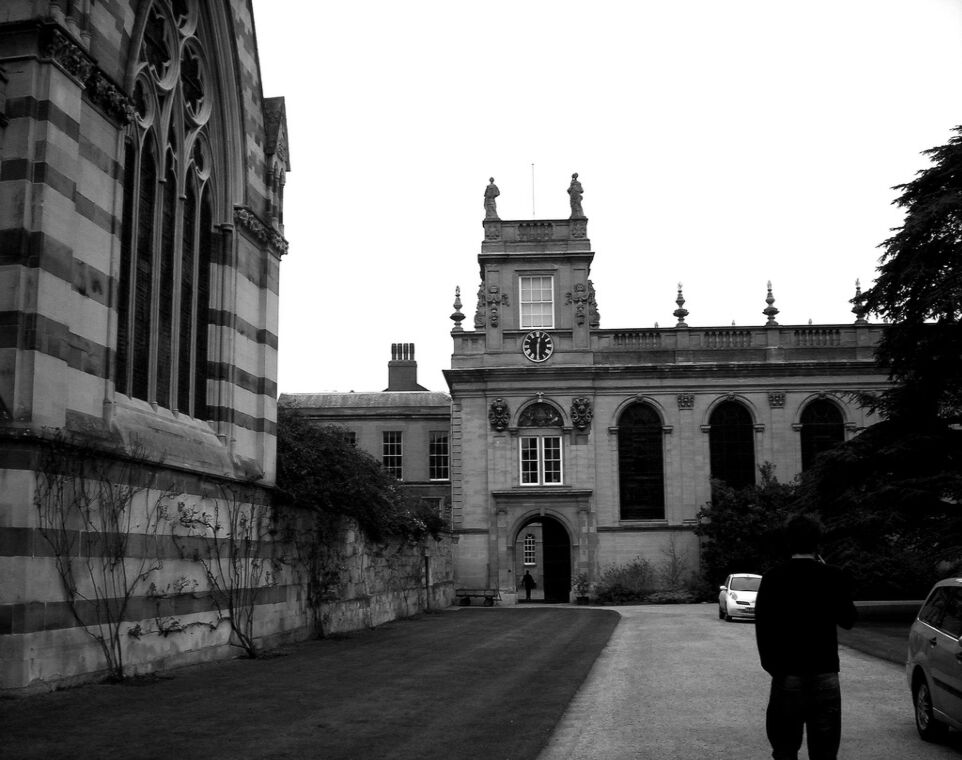}
\includegraphics[height=1.65cm]{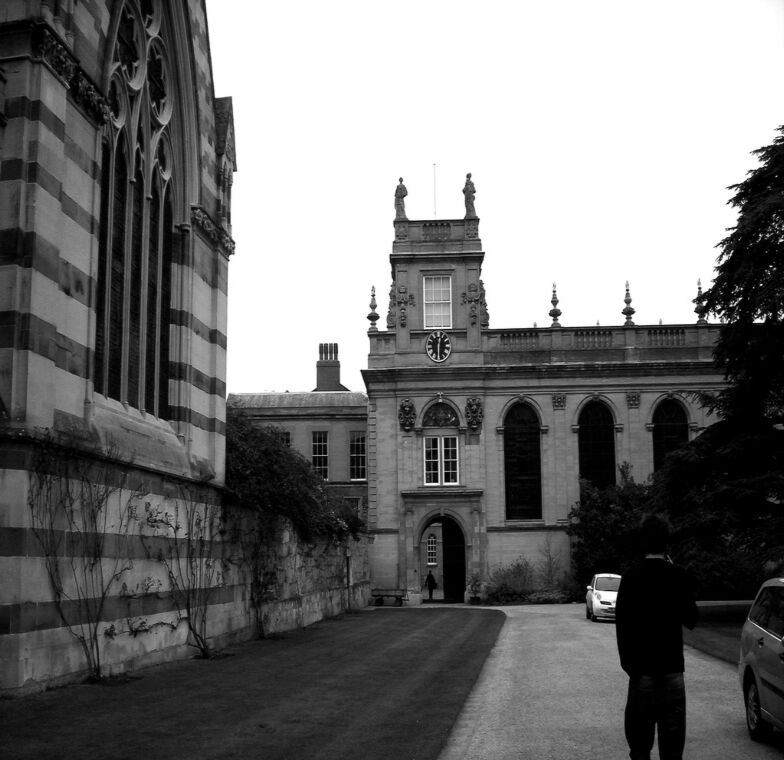}
\includegraphics[height=1.65cm]{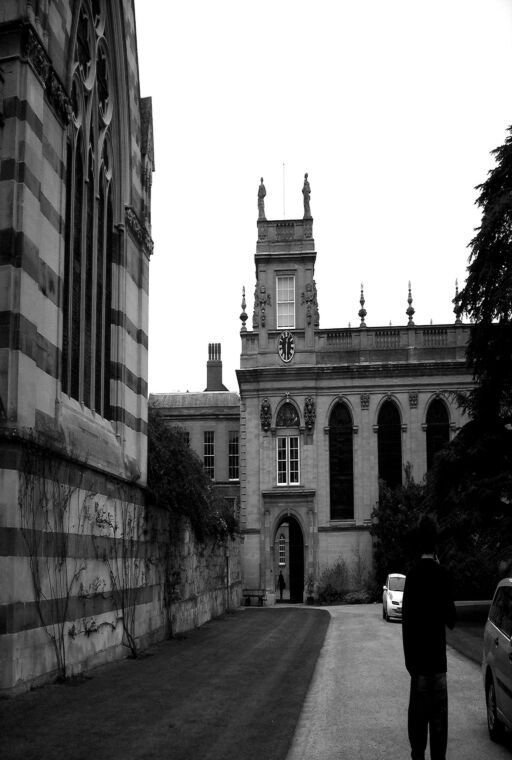}
\includegraphics[height=1.65cm]{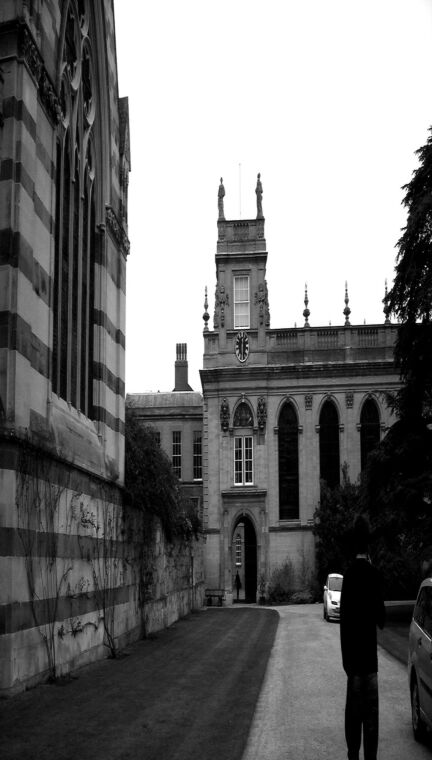}
\includegraphics[height=1.65cm]{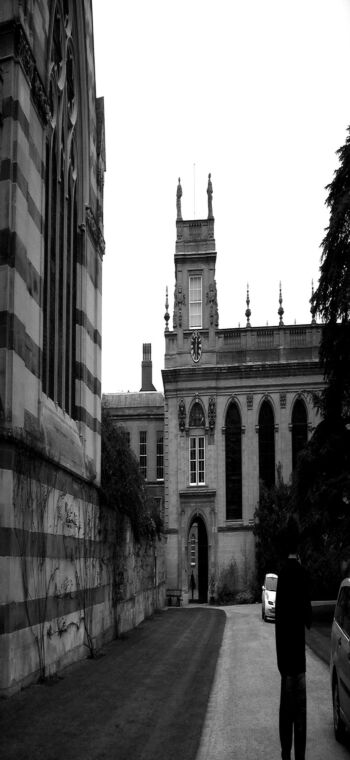}
\includegraphics[height=1.65cm]{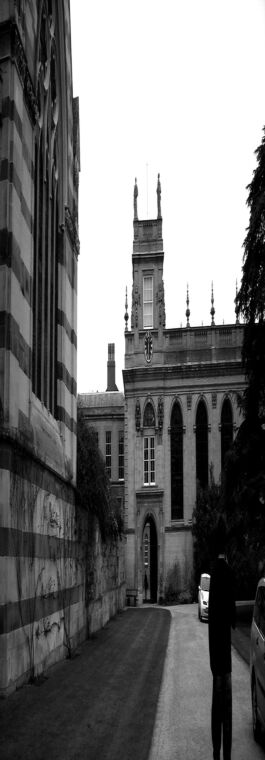}
\includegraphics[height=1.65cm]{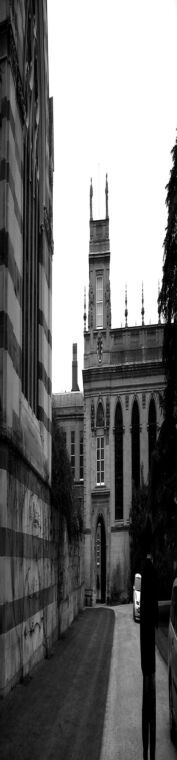}
\includegraphics[height=1.65cm]{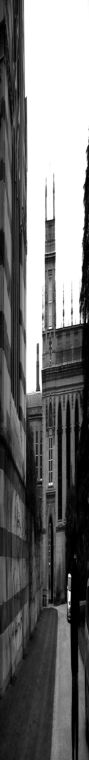}\\
\includegraphics[height=1.49cm]{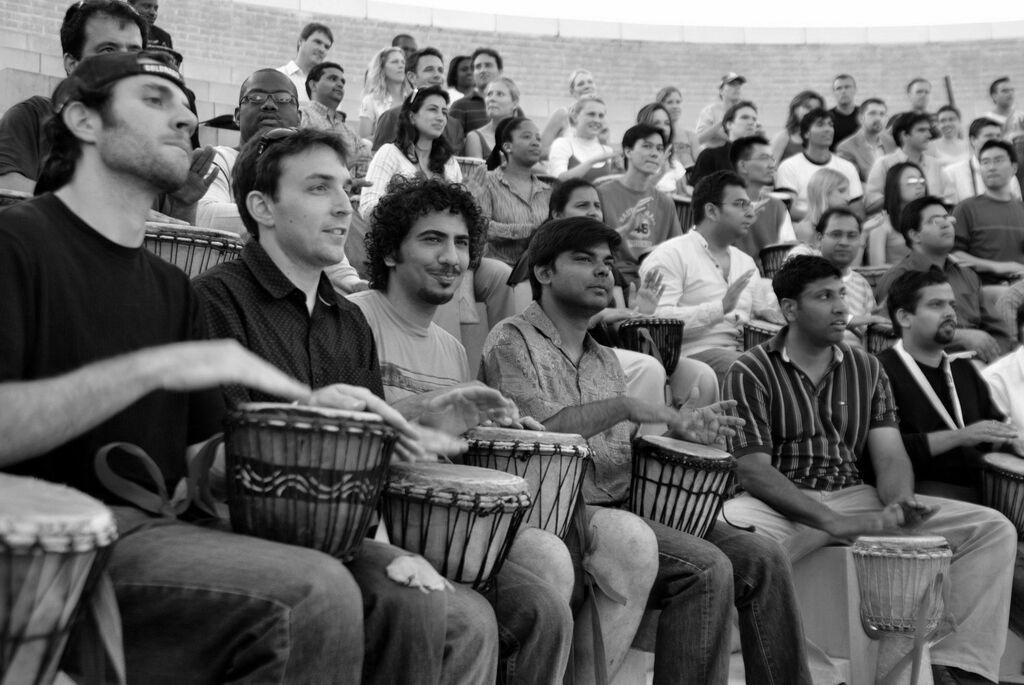}
\includegraphics[height=1.49cm]{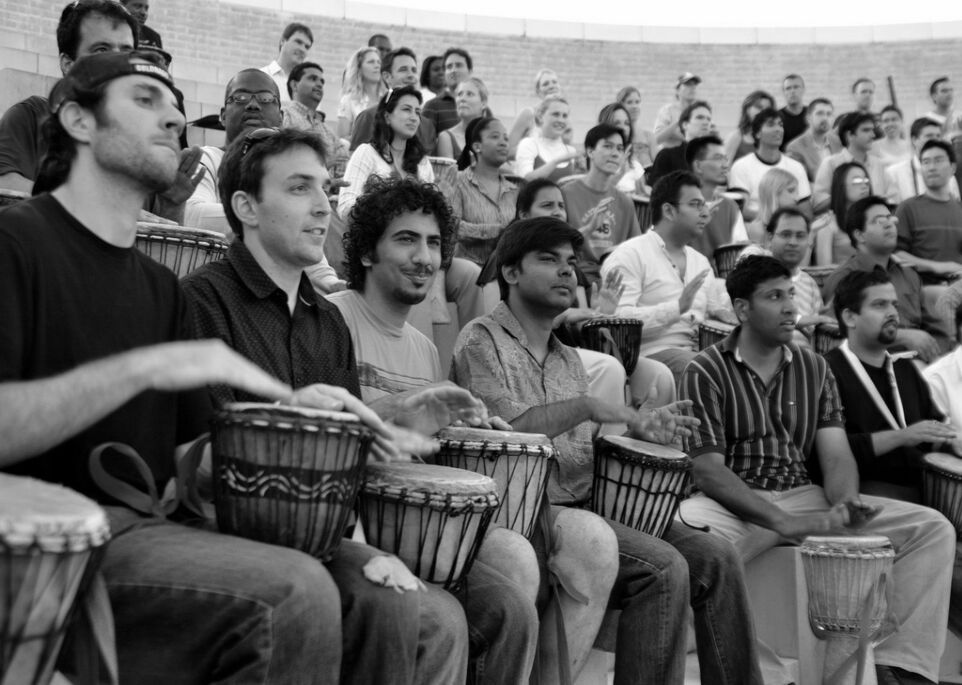}
\includegraphics[height=1.49cm]{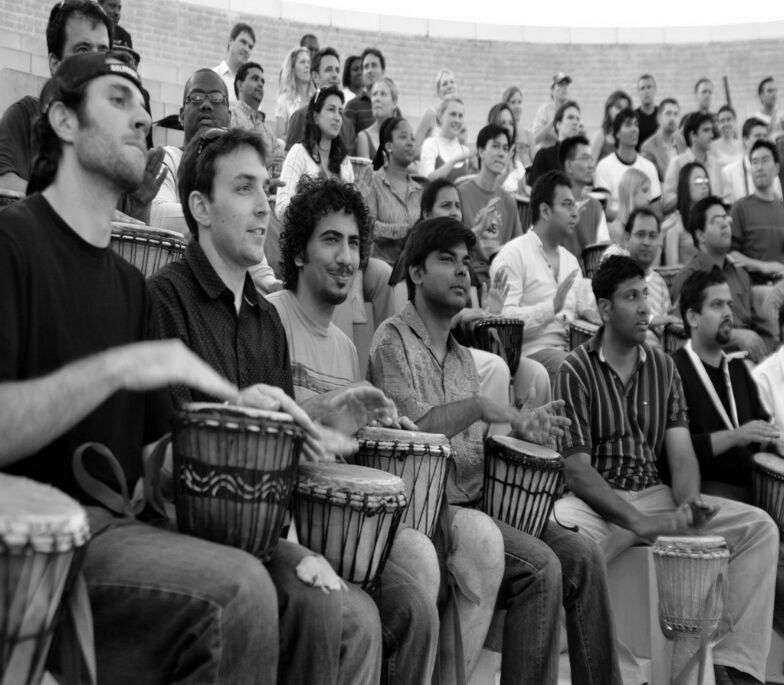}
\includegraphics[height=1.49cm]{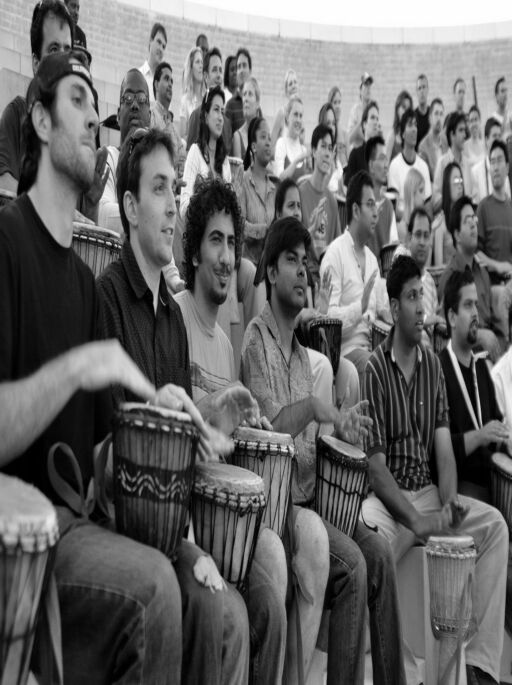}
\includegraphics[height=1.49cm]{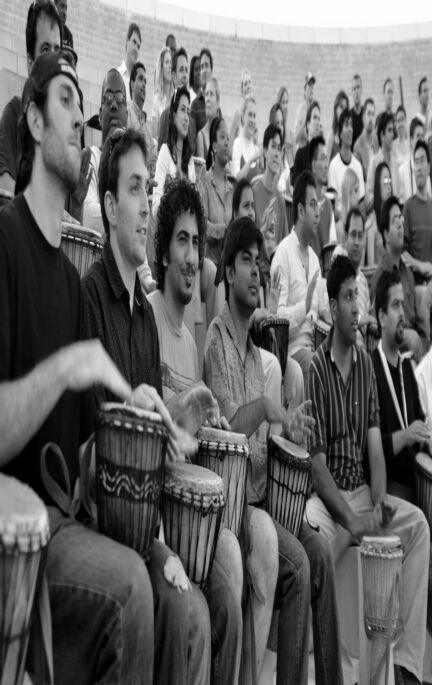}
\includegraphics[height=1.49cm]{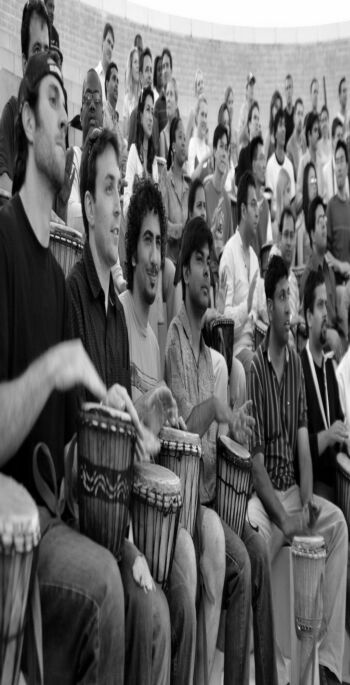}
\includegraphics[height=1.49cm]{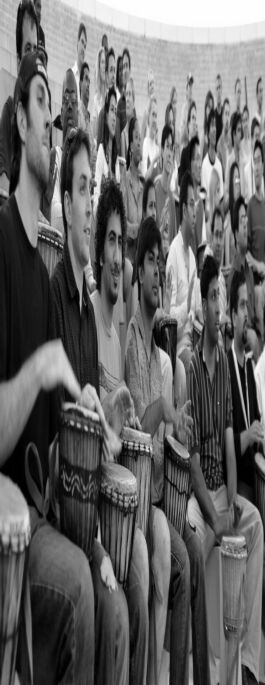}
\includegraphics[height=1.49cm]{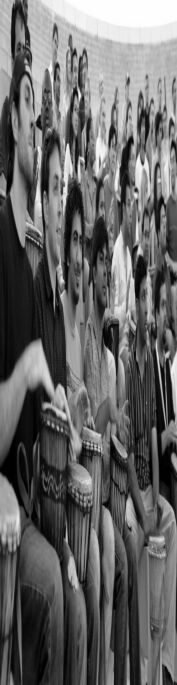}
\includegraphics[height=1.49cm]{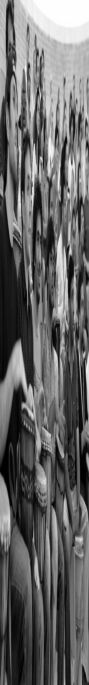}\\
\caption{Two examples of image sets from the synthetic dataset. Original unwarped images are from the Oxford buildings dataset~\cite{Philbin07}.}
\label{fig:synthetic-dataset}
\end{figure}

\begin{figure}[tbp]
\includegraphics[width=0.49\linewidth]{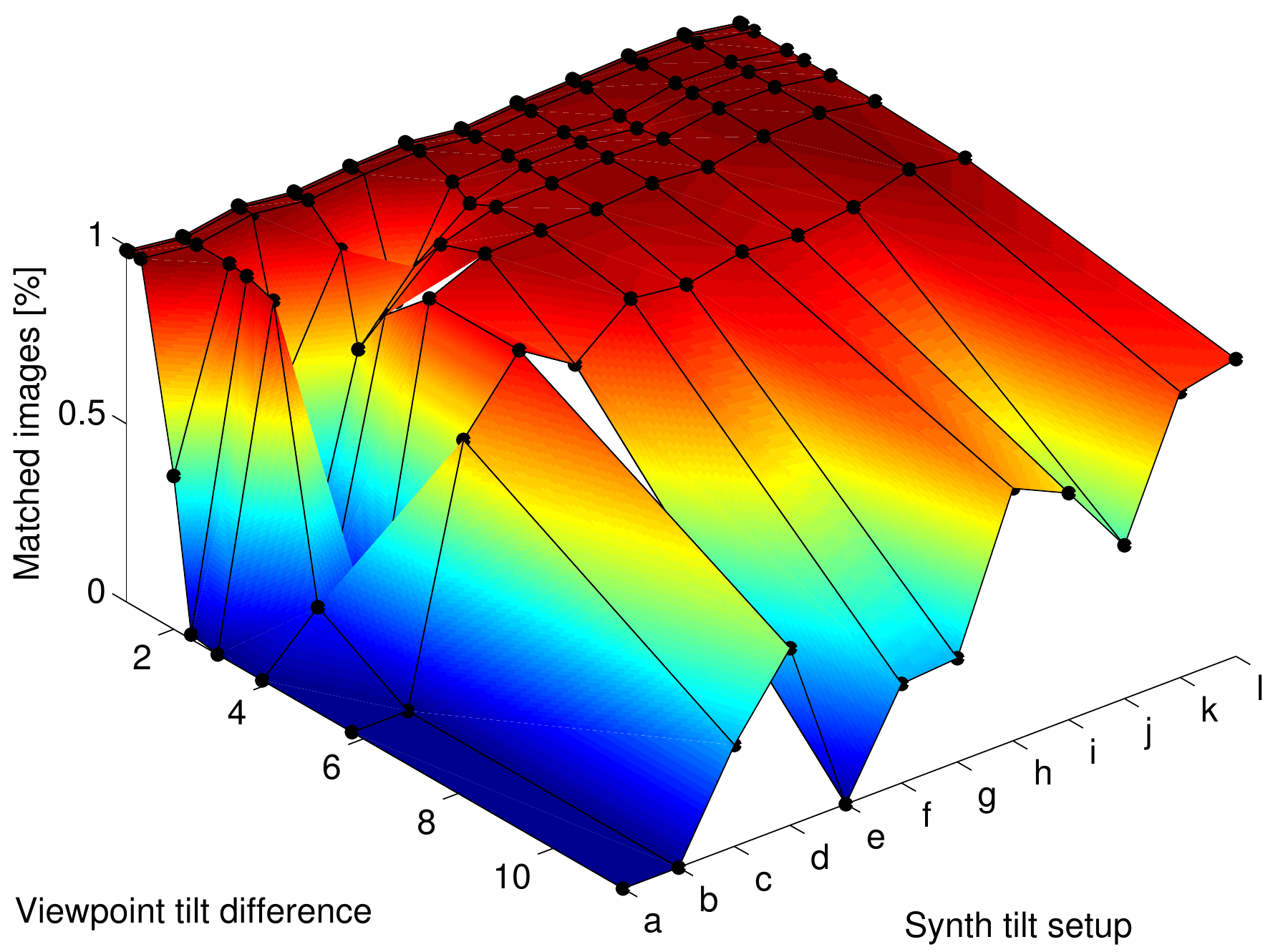}
\includegraphics[width=0.49\linewidth]{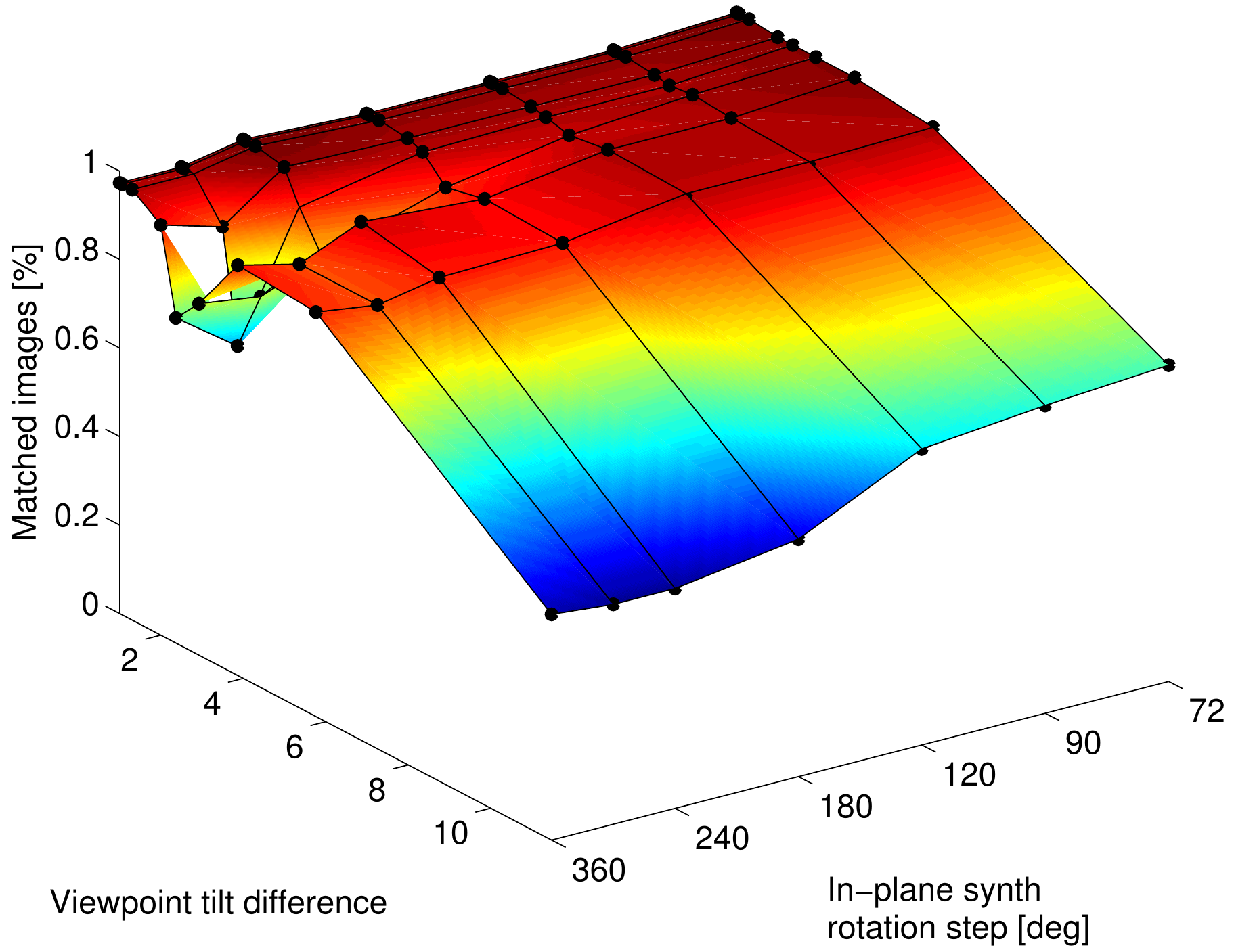}\\
\includegraphics[width=0.49\linewidth]{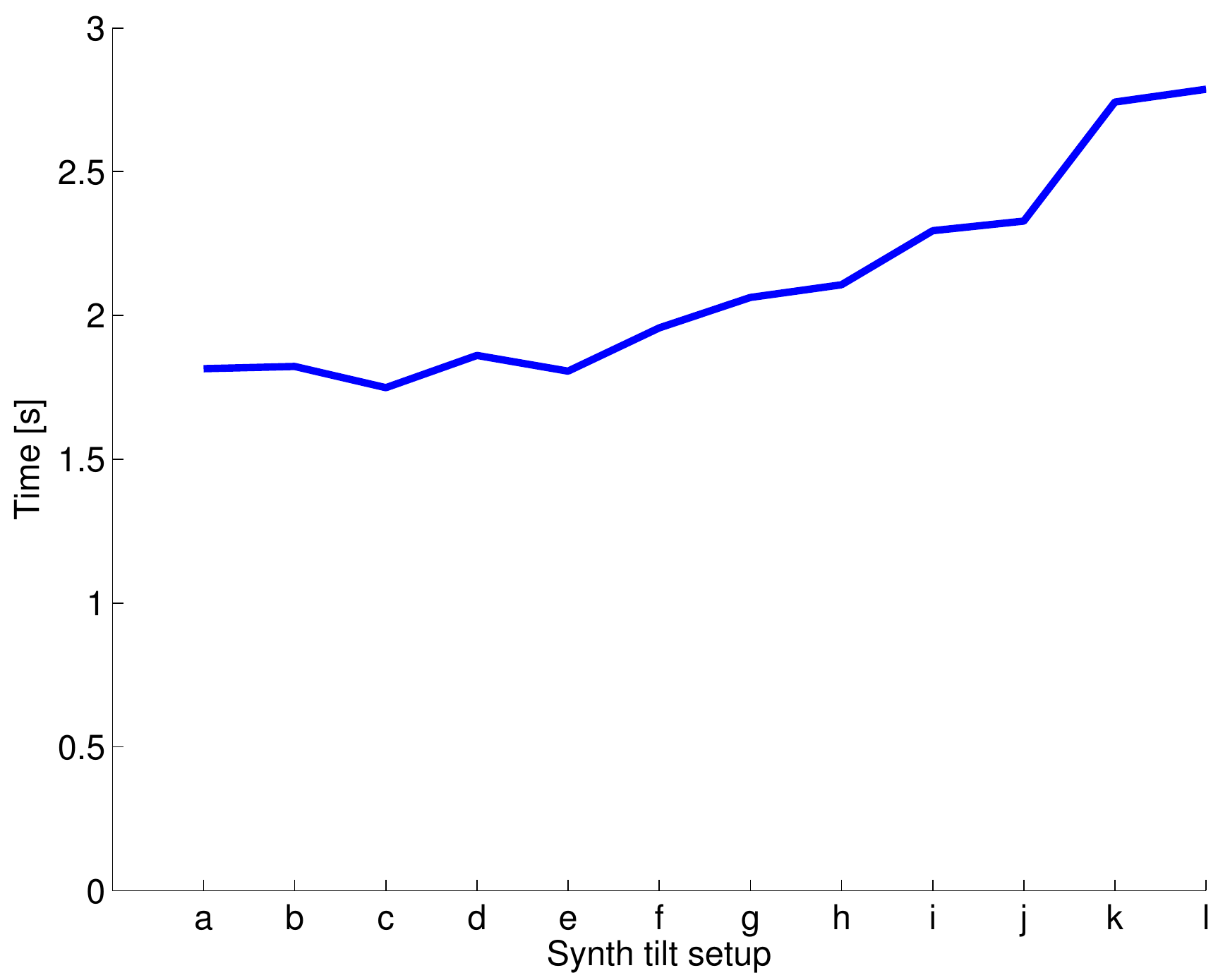}
\includegraphics[width=0.49\linewidth]{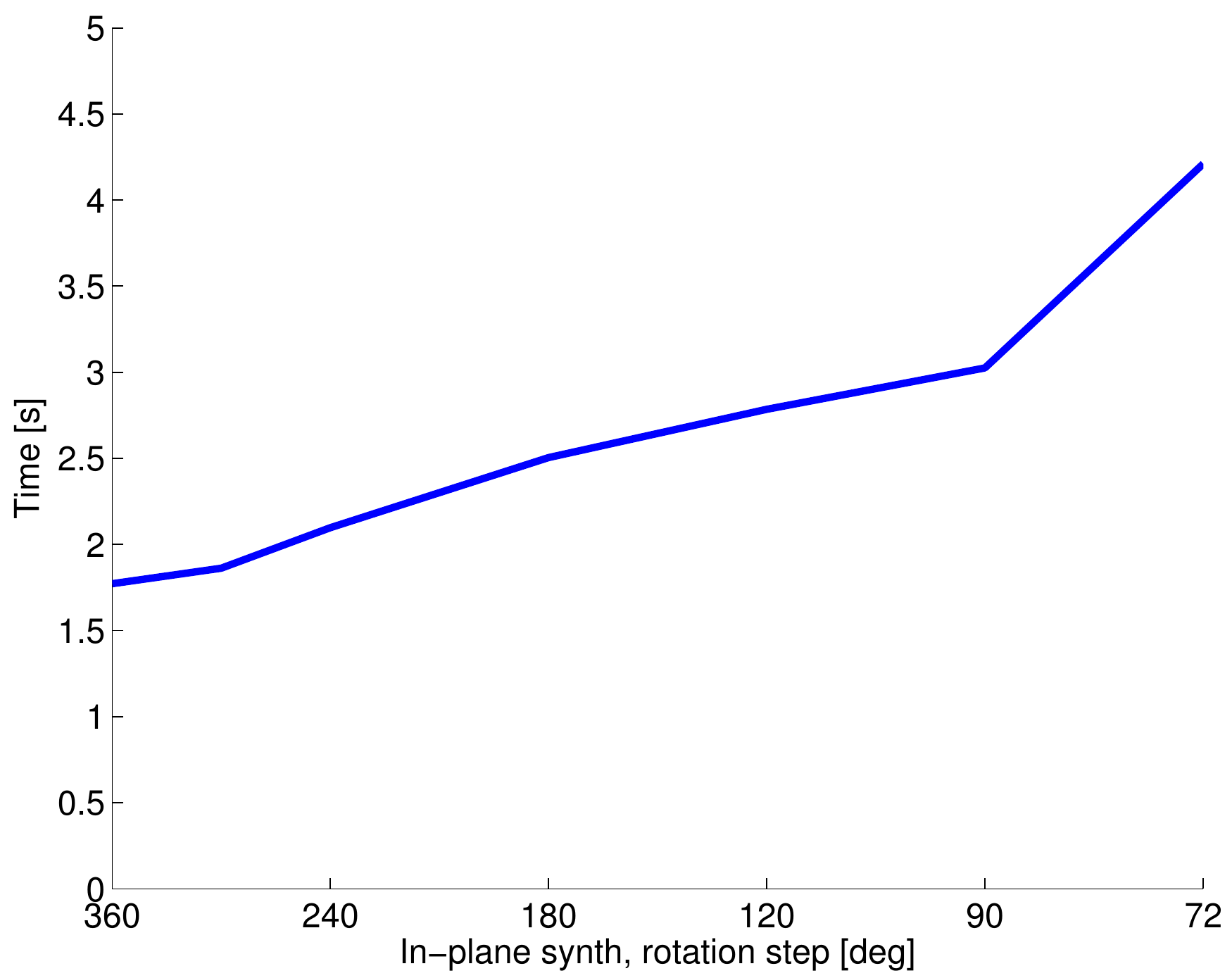}\\
\caption{Top: the percentage of images matched  depending the synthetic viewpoint difference for the DoG detector with tilt configurations given in Table~\ref{tab:view-synth-configs}.
Left: different tilt synthesis configurations (for $\Delta\phi = 240\degree$), right: different rotation synthesis configurations (for tilt set (d) = \{1,5,9\}).   
Bottom: running time per image pair for the respective configuration.}
\label{fig:dog-tilt-rot-setup}
\end{figure}

A set of simulated views with latitude angles $\theta=($0, 20, 40, 60, 65, 70, 75, 80, 85$\degree$), corresponding to tilt series $t=($1.00, 1.06, 1.30, 2.00, 2.36, 2.92, 3.86, 5.75, 11.47$)$\footnote{assuming that the original image is the fronto-parallel view} was generated for each of 150 random images from the  Oxford Building Dataset\footnote{available at http://www.robots.ox.ac.uk/$\sim$vgg/data/oxbuildings/}~\cite{Philbin07}. Example images are shown in Figure~\ref{fig:synthetic-dataset}. The ground truth affine matrix $A$ was computed for each simulated view using equation~(\ref{eq:affine-matrix-decomposition}) and used in the final verification step. 
The original image was matched against its warped version, and  the running time and number of inliers for each combination of the detector, tilt and rotation (see Table~\ref{tab:view-synth-configs}) were computed. In all, 84 setups for each of the 8 detectors on the 150 image pairs were evaluated. 
As an example, we show the relation between the density of the view-sphere sampling and the number of images matched with the DoG detector in Figure~\ref{fig:dog-tilt-rot-setup}.  

Since our goal is to find a variety of detector-tilt-rotation configurations operating with different matching ability -- run-time trade-offs, we defined ``easy", ``medium" and ``hard" problems on the synthetic dataset. Successful two-view matching was defined as recovering $n \geq 50$ ground truth correspondences on a synthetically warped image. The threshold is set high -- synthetic warping of an image is underestimating the reduction of the number of matchable features induced by the effects of a corresponding viewpoint change, e.g., due to non-planarity of the scene or illumination changes.
The matcher is considered to solve an ``easy'' problem if the percentage of the matched images $f\geq 50\%$ of total images, ``medium'' if $f \geq 90\%$ of images matched and solved ``hard'' if  $f \geq 99\%$ of the images are matched.

The experiment with the synthetically warped dataset gives a hint about the limits of configurations. Three configurations that solved the maximum tilt difference for each case fastest for a given detector were selected for evaluation. The configurations are specified in Table~\ref{tab:view-synth-configuration}. 

\begin{figure}[tb]
\includegraphics[width=0.34\linewidth]{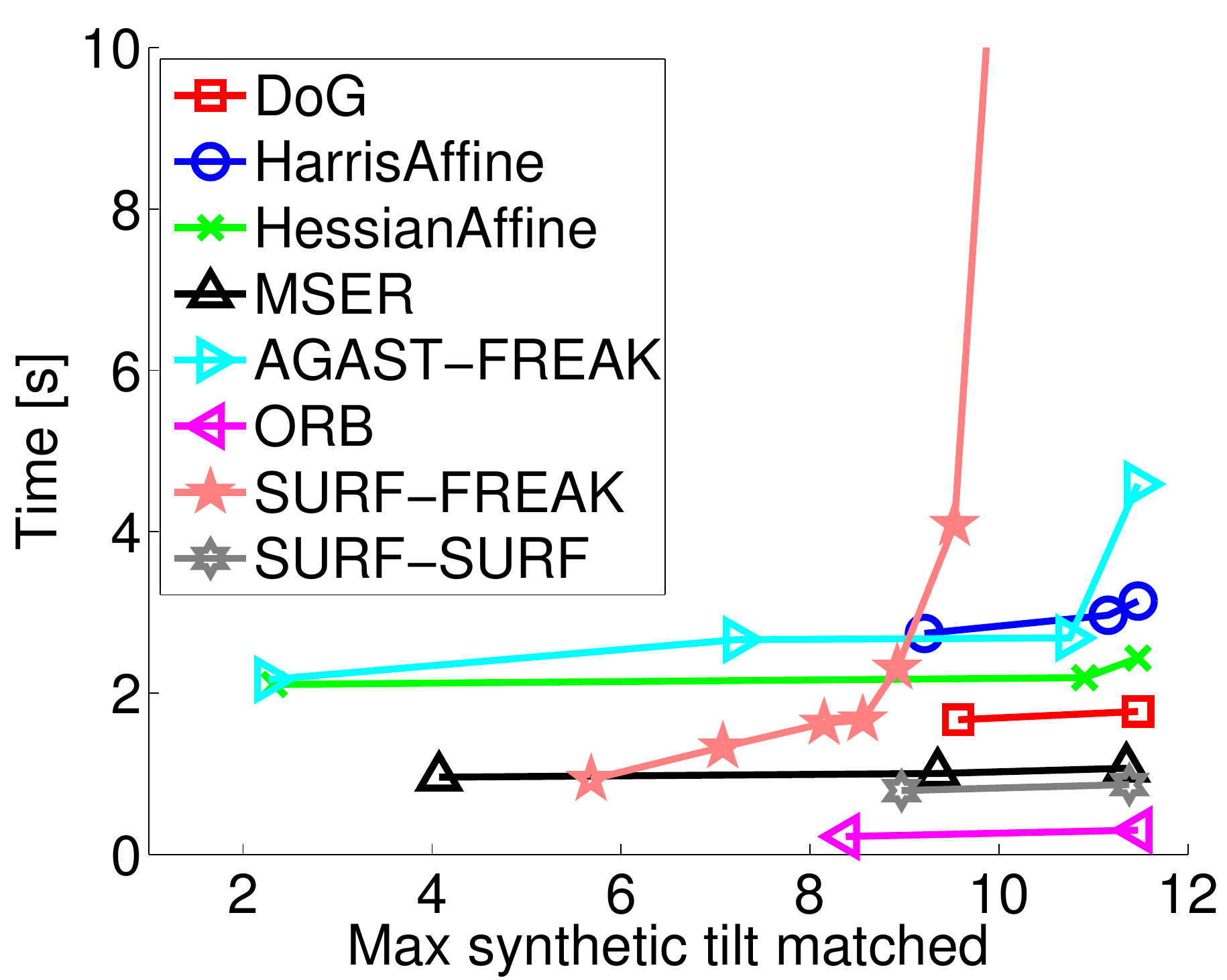}
\includegraphics[width=0.31\linewidth]{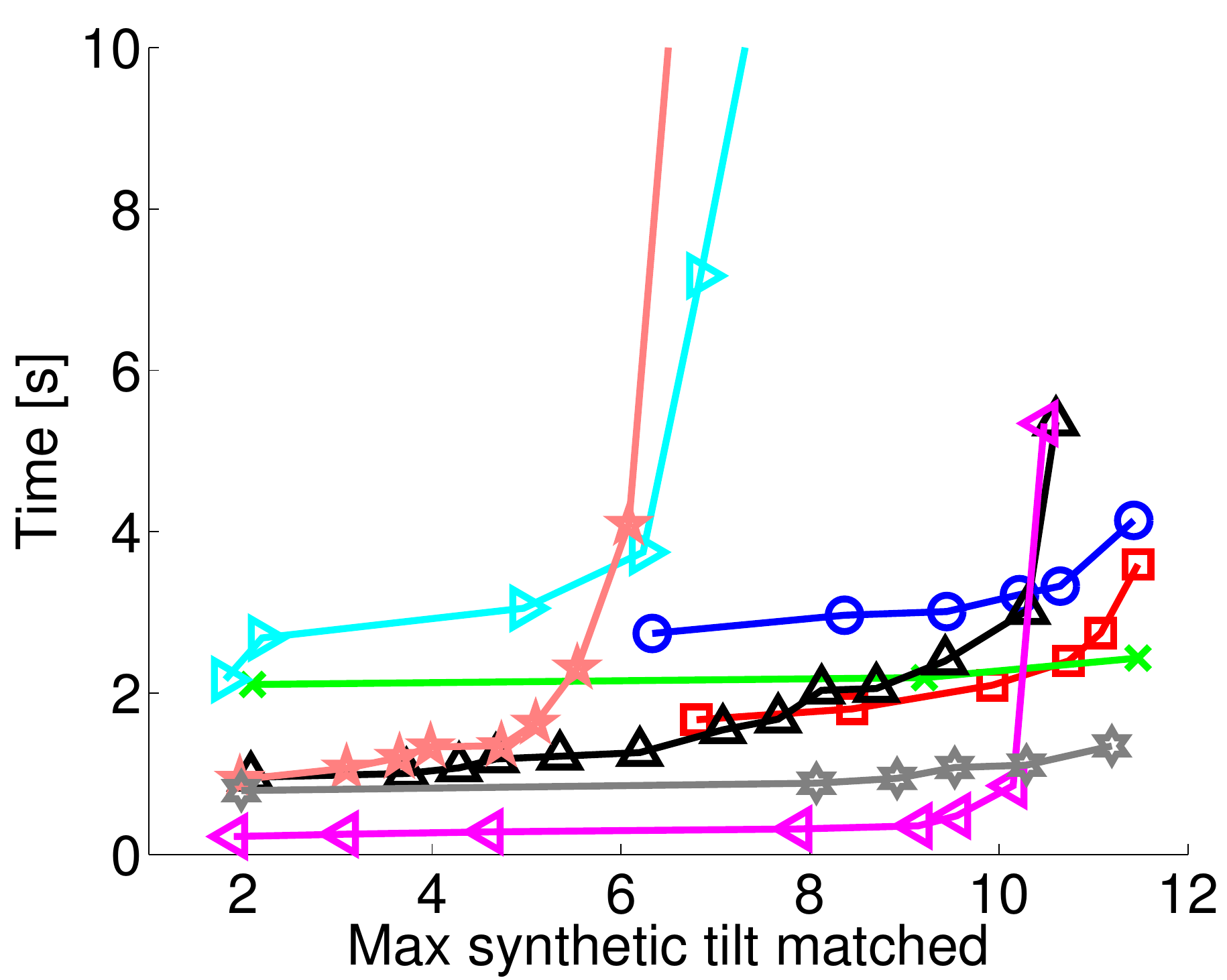}
\includegraphics[width=0.31\linewidth]{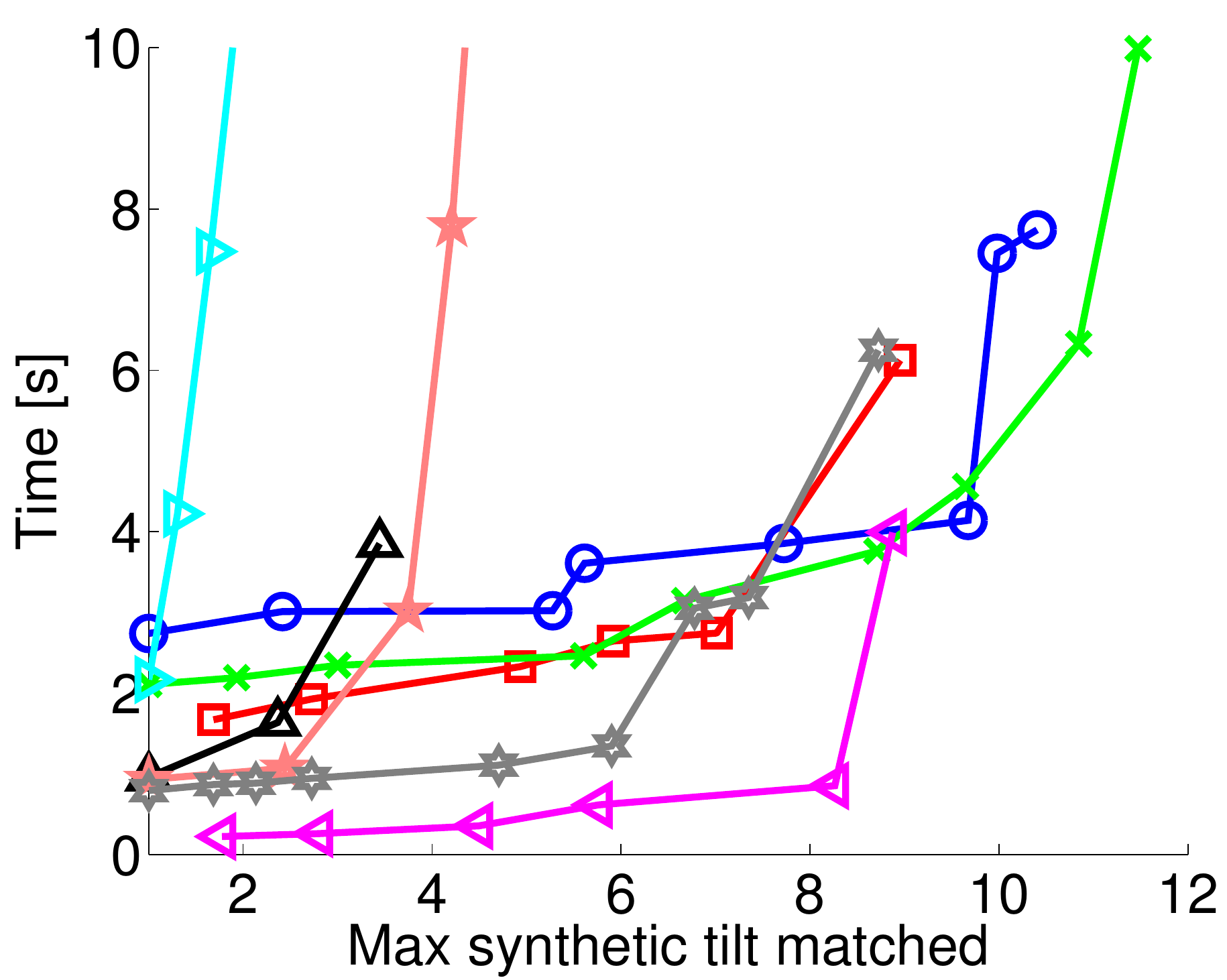}\\
\caption{Performance of view synthesis configurations  on the synthetic dataset. Average time needed  to match 
(left)``easy'', (center) ``medium'' and (right)``hard'' problems. The time for the fastest detector configuration that solved corresponding problems for a given tilt is shown.}
\label{fig:view-synth-configs}
\end{figure}

The average time necessary to match a given synthetic tilt difference for different detectors with the optimal configuration is shown in Figure~\ref{fig:view-synth-configs}. The computations were performed on the Intel i7 3.9GHz (8 cores) desktop with 8Gb RAM with parallel processing.

Note that view synthesis significantly increases the matching performance of all detectors, but not uniformly. The left plot of Figure~\ref{fig:view-synth-configs} shows that a very sparse viewsphere sampling greatly improves matching at almost no computational cost for all detectors. However, after reaching a certain density,  additional views do not add correspondences in the hardest cases -- see the right graph of Figure~\ref{fig:view-synth-configs}. The ORB detector-descriptor clearly outperforms other detectors in terms of speed, but fails to match all images with the maximum tilt difference. The Hessian-Affine shows the best performance and it matched all pairs. 

\begin{table}[tb]
\caption{View synthesis configuration evaluation. Configuration is a triplet - the detector, descriptor and the set of the synthesized views.}
\label{tab:view-synth-configs}
\centering
\begin{tabu}spread 0pt{*{2}{X[-1m,l]}}
\toprule
Detector-descriptor combination& DoG-SIFT, Hessian-Affine-SIFT, Harris-Affine-SIFT, MSER-SIFT, SURF-SURF, SURF-FREAK, AGAST-FREAK, ORB\\
\cmidrule(r){1-2}
Tilt set&  a = \{1\},  b = \{1,2\}, c = \{1,8\}, d = \{1,5,9\}, e = \{1,4\},f = \{1,4,8\}, g = \{1,2,4,8\}, h = \{1,3,6,9\}, i = \{1,4,8,10\}, j = \{1,2,4,6,8\}, 
 k = \{1,$\sqrt{2}$, 2, 2$\sqrt{2}$, 4, 4$\sqrt{2}$, 8\}, l = \{1,2,3,4,5,6,7,8,9\}.\\
\cmidrule(r){1-2}
$\phi_{base} [\degree]$ & 360, 240, 180, 120, 90, 72, 60\\
\bottomrule
\end{tabu}
\end{table}

\begin{table}[tb]
\caption{View synthesis configurations with best synthetic dataset performance.}
\label{tab:view-synth-configuration} 
\scriptsize 
\centering 
\bgroup
\begin{tabu}spread 0pt{*{4}{X[-1m,l]}}
\toprule
 & \multicolumn{3}{c}{Configurations}\\
 \cmidrule(r){2-4}
Local feature & {\centering {\sc Easy}} &{\centering {\sc Medium}} & {\centering{\sc Hard}} \\
\cmidrule(r){1-4}
DoG-SIFT & $\{t\} = \{1;5;9\}$, $\Delta\phi=360\degree{}/t$  & $\{t\} = \{1;2;3;4;5;6;7;8;9\}$, $\Delta\phi=180\degree{}/t$ & $\{t\} = \{1;2;4;6;8\}$, $\Delta\phi=60\degree{}/t$  \\
\cmidrule(r){1-4}
HarrAff-SIFT &  $\{t\} = \{1;3;6;9\}$, $\Delta\phi=360\degree{}/t$ & $\{t\} = \{1;2;3;4;5;6;7;8;9\}$, $\Delta\phi=360\degree{}/t$ & $\{t\} = \{1;2;3;4;5;6;7;8;9\}$, $\Delta\phi=120\degree{}/t$ \\
\cmidrule(r){1-4}
HessAff-SIFT &  $\{t\} = \{1;5;9\}$, $\Delta\phi=360\degree{}/t$ & $\{t\} = \{1;5;9\}$, $\Delta\phi=360\degree{}/t$ & $\{t\} = \{1;2;4;6;8\}$, $\Delta\phi=60\degree{}/t$  \\
\cmidrule(r){1-4}
MSER-SIFT  &  $\{t\} = \{1;8\}$, $\Delta\phi=360\degree{}/t$ &  $\{t\} =  \{1;\sqrt{2}; 2; 2\sqrt{2}; 4;$ $4\sqrt{2}; 8\}$, $\Delta\phi=60\degree{}/t$ & $\{t\} = \{1;3;6;9\}$, $\Delta\phi=60\degree{}/t$  \\
\cmidrule(r){1-4}
AGAST-FREAK  &  $\{t\} = \{1;4;8;10\}$, $\Delta\phi=360\degree{}/t$ & $\{t\} = \{1;3;6;9\}$, $\Delta\phi=60\degree{}/t$ & $\{t\} = \{1;3;6;9\}$, $\Delta\phi=72\degree{}/t$  \\
\cmidrule(r){1-4}
ORB &  $\{t\} = \{1;5;9\}$, $\Delta\phi=360\degree{}/t$& $\{t\} = \{1;2;3;4;5;6;7;8;9\}$, $\Delta\phi=90\degree{}/t$ & $\{t\} =  \{1;\sqrt{2}; 2; 2\sqrt{2}; 4;$ $4\sqrt{2}; 8\}$, $\Delta\phi=72\degree{}/t$ \\
\cmidrule(r){1-4}
SURF-FREAK &  $\{t\} =  \{1;\sqrt{2}; 2; 2\sqrt{2}; 4;$ $4\sqrt{2}; 8\}$, $\Delta\phi=60\degree{}/t$ & $\{t\} =  \{1;\sqrt{2}; 2; 2\sqrt{2}; 4;$ $4\sqrt{2}; 8\}$, $\Delta\phi=90\degree{}/t$& $\{t\} =  \{1;\sqrt{2}; 2; 2\sqrt{2}; 4;$ $4\sqrt{2}; 8\}$, $\Delta\phi=72\degree{}/t$ \\
\cmidrule(r){1-4}
SURF-SURF &  $\{t\} = \{1;5;9\}$, $\Delta\phi=360\degree{}/t$ & $\{t\} = \{1;2;3;4;5;6;7;8;9\}$, $\Delta\phi=360\degree{}/t$& $\{t\} = \{1;2;3;4;5;6;7;8;9\}$, $\Delta\phi=72\degree{}/t$ \\
\bottomrule
\end{tabu}
\egroup
\end{table}

\subsection{First geometrically inconsistent nearest neighbor ratio correspondence selection strategy}
\begin{figure}[tb]
\includegraphics[width=0.49\linewidth]{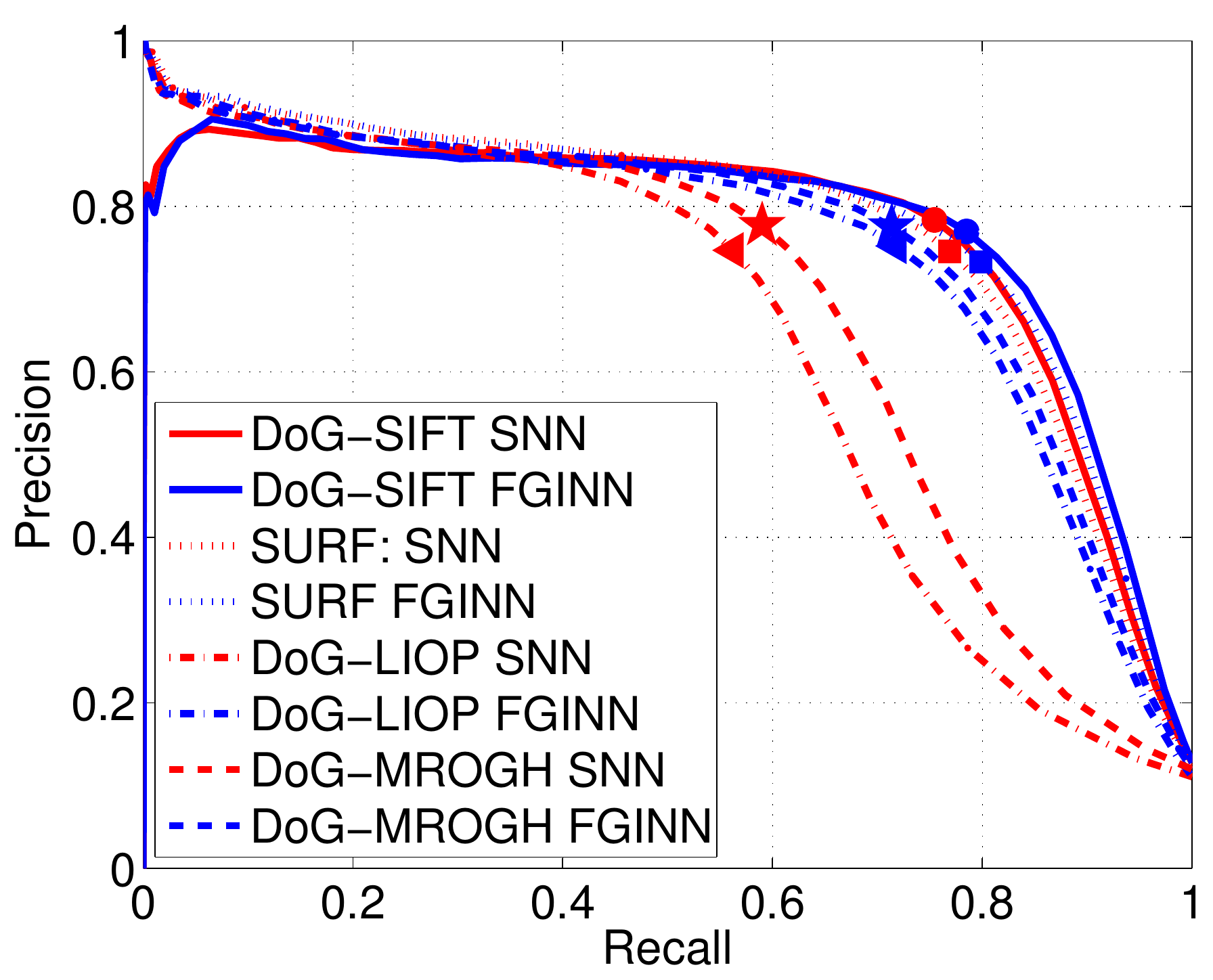}
\includegraphics[width=0.49\linewidth]{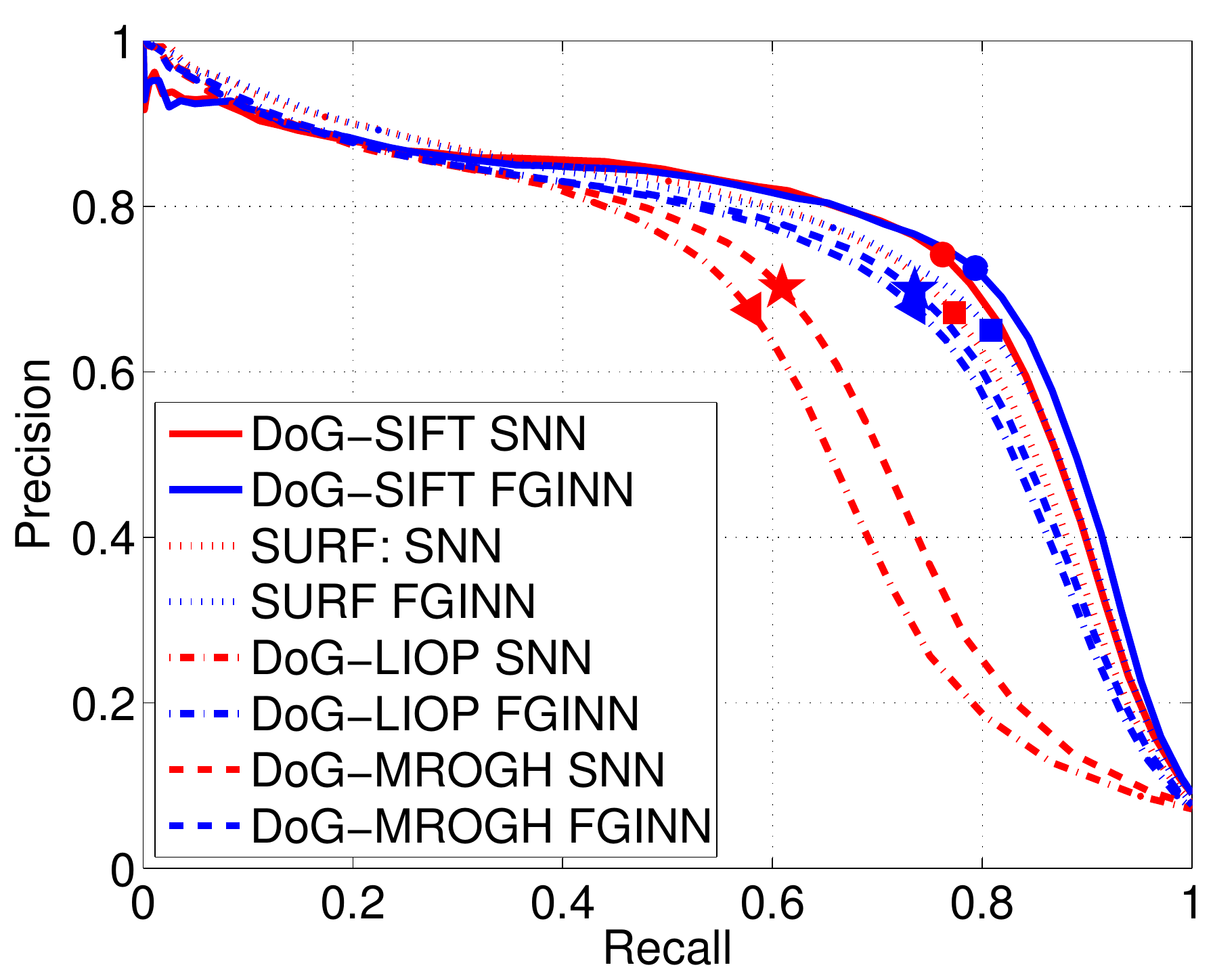}\\
 \caption{Comparison of the FGINN and SNN matching strategies for SIFT, SURF, LIOP and MROGH on images from the Oxford~\cite{Mikolajczyk2005} and Cordes~\etal~\cite{Cordes2011} datasets. Markers show operationg points for the common distance ratio threshold = 0.8.  Matching without (left) view synthesis an using view synthesis (right) with parameters  $\{t\} = \{1;5;9\}$, $\Delta\phi=360\degree{}/t$. The recall and precision of the correspondence filtering step, therefore the maximum recall is 1 when all correspondences are kept.}
 \label{fig:1st-to-2nd-synth}
 \end{figure} 
The following protocol was used to find the thresholds and to evaluate the performance of the proposed First Geometrically Inconsistent Nearest Neighbor \emph{FGINN} strategy. First, similarity covariant regions were detected using the DoG detector (we also tried Hessian-Affine, MSER and SURF, with very similar results) and described using four popular descriptors -- RootSIFT, SURF, LIOP~\cite{LIOP2011} and MROGH~\cite{MROGH2011} which are typically matched with the second-nearest region SNN strategy.  Then for each keypoint descriptor, the  first, second, and first geometrically inconsistent descriptors in the other image were found. The matching keypoints were then labeled as correct if their Sampson error was within 1 pixel of the ground truth location given by homography for the image pair, and incorrect otherwise. 

The experiment  was performed on 26 image pairs of the publicly available  datasets~\cite{Mikolajczyk2005},\cite{Cordes2011} (image pairs 1-3, precise homography provided) and ~\cite{Lebeda2012} (the homography was estimated using the provided precise ground truth correspondences).
The recall-precision curves for correspondences from all images were plotted with a varying  ratio threshold from 0 to 1 in Figure~\ref{fig:1st-to-2nd-synth}.
The FGINN curves for SIFT and SURF slightly outperform standard SNNs, while for LIOP and MROGH the difference is much more significant. The significantly higher benefit of  the FGINN rule for LIOP and MROGH can be explained by their lower sensitivity to keypoint shift, which in turn means that undesirable suppression of keypoints happens in a larger neighborhood. 
The lower sensitivity to shifts was experimentally verified.

\section{Experiments}       
\label{mods:main}     %
We have tested MODS and, as a baseline, ASIFT\footnote{Reference code from \url{
http://demo.ipol.im/demo/my\_affine\_sift}} and single detector configurations specified in Table~\ref{tab:view-synth-configuration} on seven public. datasets~\cite{Moreels2007},\cite{Mishkin2013},\cite{Altwaijry2013}, \cite{Hauagge2012}, \cite{Kelman2007}, \cite{Aguilera2012}, \cite{Mikulik2014}.
\paragraph{Implementation details of the MODS algorithm and parameter setting}
The kd-tree algorithm from FLANN library~\cite{FLANN2009} was used to efficiently find the N-closest descriptors.  The distance ratio thresholds of the FGINN matching strategy were experimentally selected based on the CDFs of matching and non-matching descriptors.%

The MODS algorithm allows to set the minimum desired number of inliers which have a very low probability to be a random result as a stopping criterion. The recommended value -- 15 inliers for the homography -- did not produce false positive results in experiments. Computations were performed on Intel i3 CPU @ 2.6GHz with 4Gb RAM with 4 cores.
\subsection{MODS variants testing on Extreme Viewpoint and Oxford Dataset}
\label{main:real-world-testing}
To evaluate the performance of matching algorithms, we introduce a two-view matching evaluation dataset\footnote{Available at http://cmp.felk.cvut.cz/wbs/index.html} with extreme viewpoint changes, see Table~\ref{tab:image-dataset}. The dataset includes image pairs from publicly available datasets: {\sc adam} and {\sc mag}~\cite{ASIFT2009}, {\sc graf}~\cite{Mikolajczyk05} and {\sc there}~\cite{Cordes2011}. The ground truth homography matrices were estimated by LO-RANSAC using correspondences from all detectors in view synthesis configuration $\{t\} = \{1;\sqrt{2};2;2\sqrt{2};4;4\sqrt{2};8\}$, $\Delta\phi=72\degree{}/t$. The number of inliers for each image pair was $\geq$
50 and the homographies were manually inspected. For the image pairs {\sc graf} and {\sc there} precise homographies are provided by
Cordes~\etal~\cite{Cordes2011}. Transition tilts $\tau$ were computed using equation~(\ref{eq:affine-matrix-decomposition}) with SVD decomposition of the linearized homography at center of the first image of the pair (see Table ~\ref{tab:image-dataset}). Oxford~\cite{Mikolajczyk05} dataset with 42 image pairs (1-2, ..., 1-6) was used for easier wide baseline problems.

\paragraph{Experimental protocol} The evaluated algorithms matched image pairs and the output keypoints correspondences were checked with ground truth homographies. The image pair is considered as solved, when at least 10 output correspondences are correct.

\begin{table}[tb]
\caption{The Extreme View Dataset -- EVD. Image sources: C -- \cite{Cordes2011}, Ox -- ~\cite{Mikolajczyk05}, M -- \cite{ASIFT2009}. }
\label{tab:image-dataset}
\centering
\footnotesize
\bgroup
\def\arraystretch{1.0}
\setlength{\tabcolsep}{.54em}
\tabulinesep=0.5mm
\begin{tabu}spread 0pt{*{16}{X[-1m,c]}}
\toprule
\# & 1 & 2& 3&4&5&6&7&8&9&10&11&12&13&14&15\\
\cmidrule(l){1-16}
Name & {\sc there} &{\sc graf}&{\sc adam}&{\sc mag}&{\sc grand}&{\sc pkk}&{\sc face}&{\sc girl}&{\sc shop}&{\sc dum}&{\sc index}&{\sc cafe}&{\sc fox}&{\sc cat}&{\sc vin}\\
Ref. & C & Ox &M & M& EVD &EVD & EVD& EVD& EVD & EVD &EVD&EVD &EVD & EVD&EVD \\
$\tau$ -- tilt &6.3 &3.6 &4.8& 20& 2.9 & 7.1&6.9 &8.0&9.1 & 6.9&8.5&11.9&22.5&47&49.8\\
\cmidrule(l){1-16}
\shortstack{Size \\$[$px$]$ }&\shortstack{1536\\x\\1024} &\shortstack{800\\x\\640} & \shortstack{600\\x\\450}& \shortstack{600\\x\\450}& \shortstack{1000\\x\\667} & \shortstack{1000\\x\\750}&\shortstack{1000\\x\\750} &\shortstack{1000\\x\\750}&\shortstack{1000\\x\\562}&\shortstack{1000\\x\\729} & \shortstack{1000\\x\\750}&\shortstack{800\\x\\533}&\shortstack{1000\\x\\563}&\shortstack{1000\\x\\598}&\shortstack{1000\\x\\715}\\
\bottomrule
\end{tabu}
\egroup
\\
\bgroup
\def\arraystretch{1.2}%
\setlength{\tabcolsep}{.16667em}
\tabulinesep=0.2mm
\begin{tabu*}spread 0pt{*{9}{|X[-1m,c]}|}
\toprule
\# & Image 1 & Image 2 &\# & Image 1 & Image 2&\# & Image 1 & Image 2 \\
\cmidrule(r){1-9}
1
& \includegraphics[height=1.4cm]{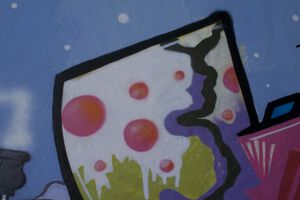} 
& \includegraphics[height=1.4cm]{chapters/mods/there2.jpg}
& 6 
& \includegraphics[height=1.4cm]{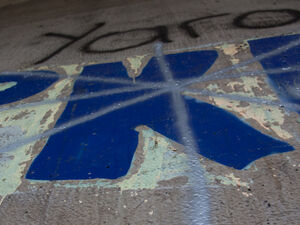} 
& \includegraphics[height=1.4cm]{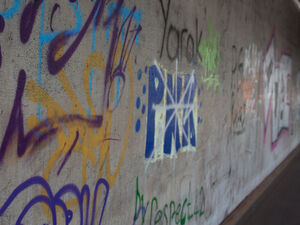} 
&11
& \includegraphics[height=1.4cm]{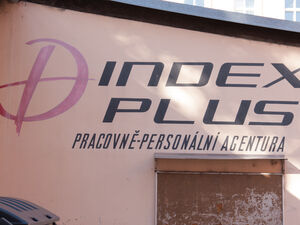} 
& \includegraphics[height=1.4cm]{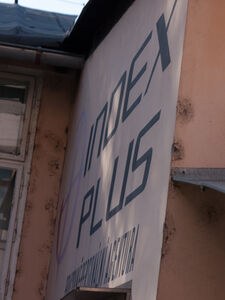} \\ 
\cmidrule(r){1-9}
2
& \includegraphics[height=1.4cm]{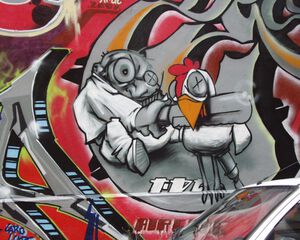} 
& \includegraphics[height=1.4cm]{chapters/mods/graf2.jpg}
&7
& \includegraphics[height=1.4cm]{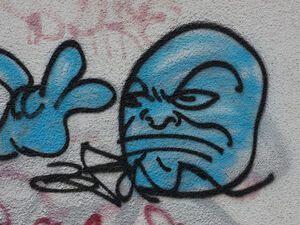} 
& \includegraphics[height=1.4cm]{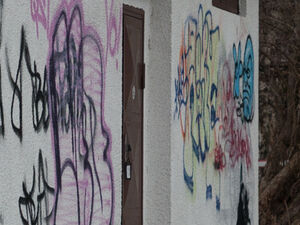}
&
12& \includegraphics[height=1.4cm]{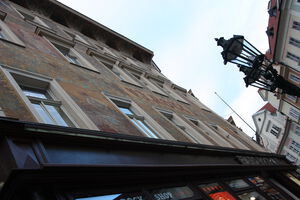} 
& \includegraphics[height=1.4cm]{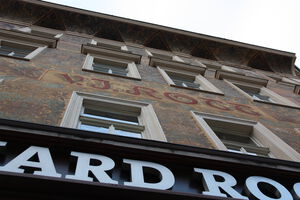} \\ 
\cmidrule(r){1-9}
3 
& \includegraphics[height=1.4cm]{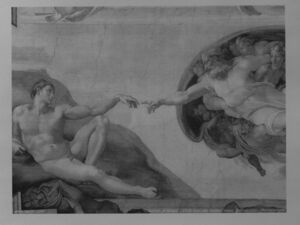} 
& \includegraphics[height=1.4cm]{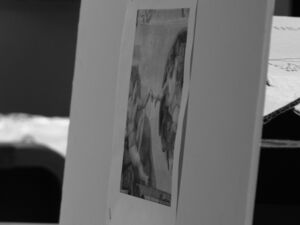} 
&8
& \includegraphics[height=1.4cm]{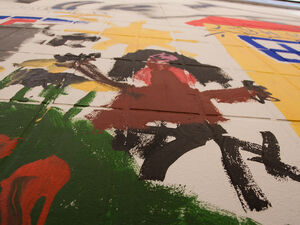} 
& \includegraphics[height=1.4cm]{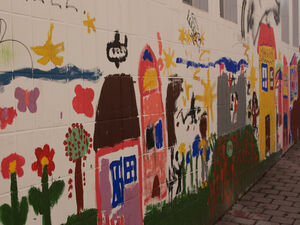}
& 13
& \includegraphics[height=1.4cm]{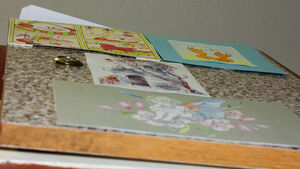} 
& \includegraphics[height=1.4cm]{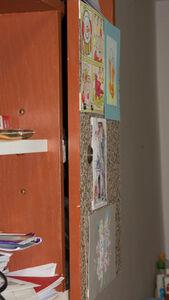}\\ 
\cmidrule(r){1-9}
4
& \includegraphics[height=1.4cm]{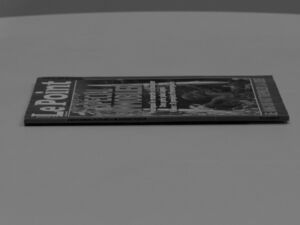} 
& \includegraphics[height=1.4cm]{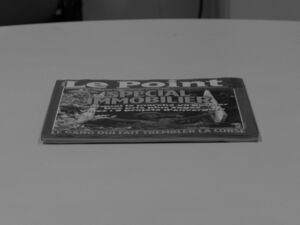}
&9 
& \includegraphics[height=1.4cm]{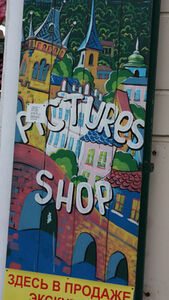} 
& \includegraphics[height=1.4cm]{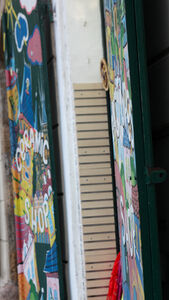}
&14 
&\includegraphics[height=1.4cm]{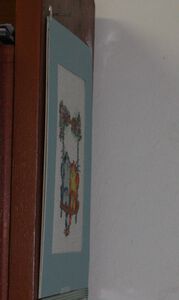} 
& \includegraphics[height=1.4cm]{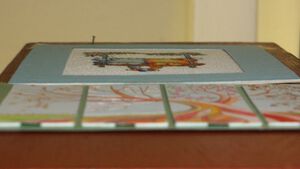} \\
\cmidrule(r){1-9}
5
&  \includegraphics[height=1.4cm]{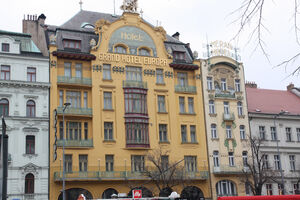} 
& \includegraphics[height=1.4cm]{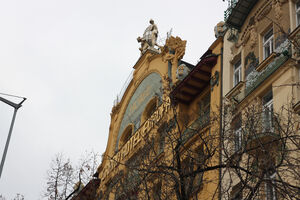}
&10
& \includegraphics[height=1.4cm]{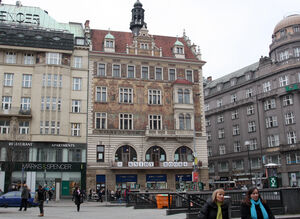} 
& \includegraphics[height=1.4cm]{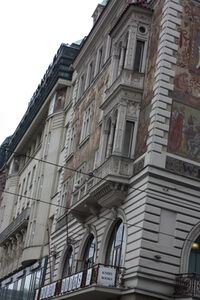}
&15
& \includegraphics[height=1.4cm]{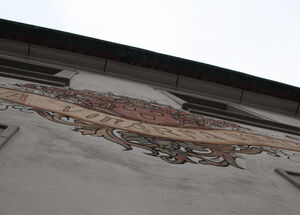}
& \includegraphics[height=1.4cm]{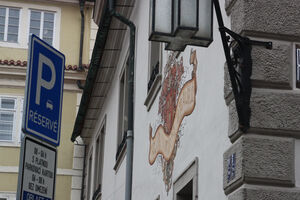} \\ 
\bottomrule
\end{tabu*}
\egroup
\end{table}
\begin{table}[tb]
\caption{ASIFT view synthesis configuration}
\label{tab:ASIFT-configuration}
\footnotesize 
\centering
\bgroup
\tabulinesep=0.4mm
\begin{tabu} spread 0pt{X[-1m,c]X[-1,m,l]}
\toprule
Detector& \centering View synthesis setup\\
\cmidrule(r){1-2}
DoG & $\{S\} = \{1\}$, $\{t\} = \{1;\sqrt{2};2;2\sqrt{2};4;4\sqrt{2};8\}$, $\Delta\phi=72\degree{}/t$\\
\bottomrule
\end{tabu}
\egroup
\end{table}

\begin{figure}[tb]
\centering
\includegraphics[width=0.34\linewidth]{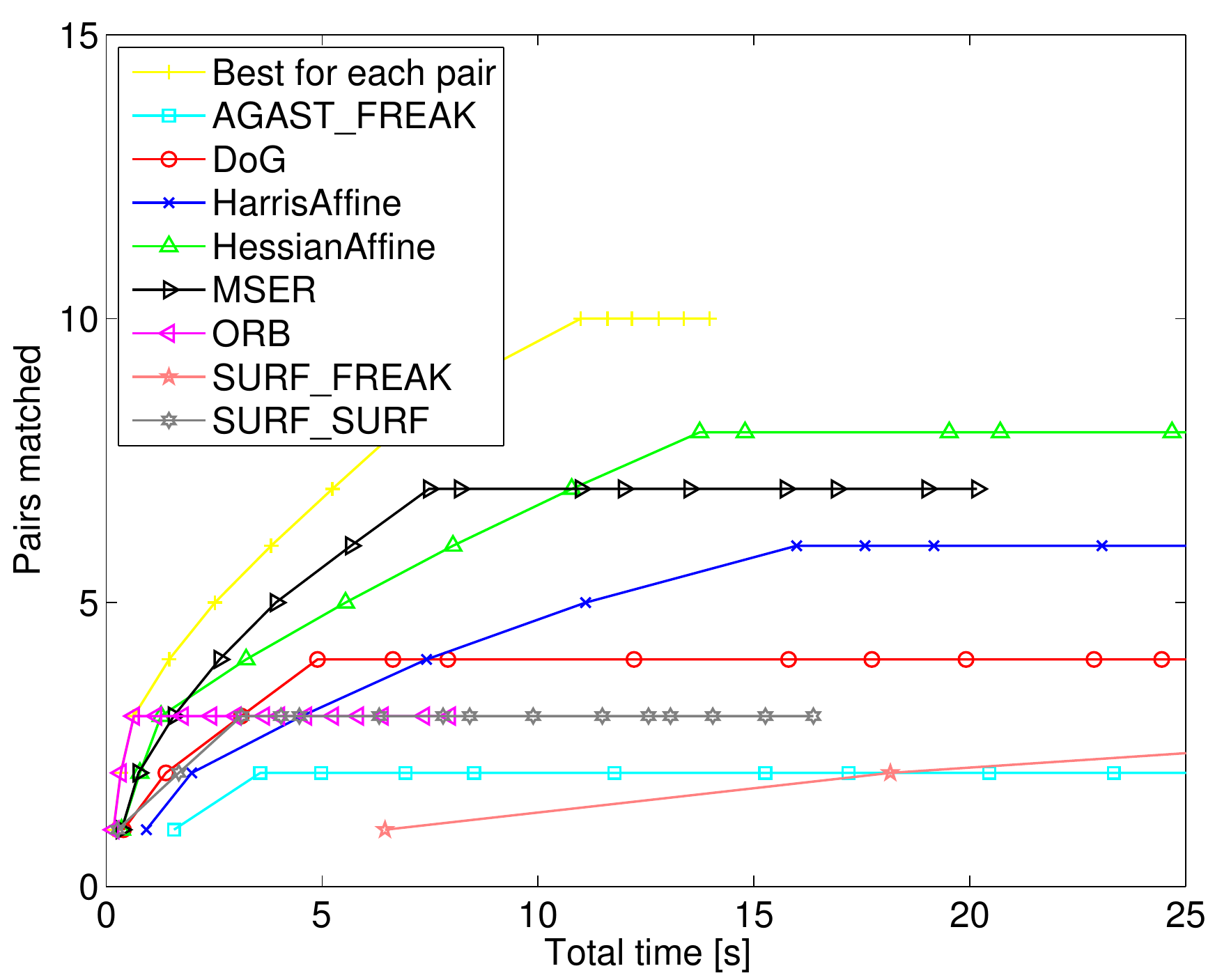}
\includegraphics[width=0.32\linewidth]{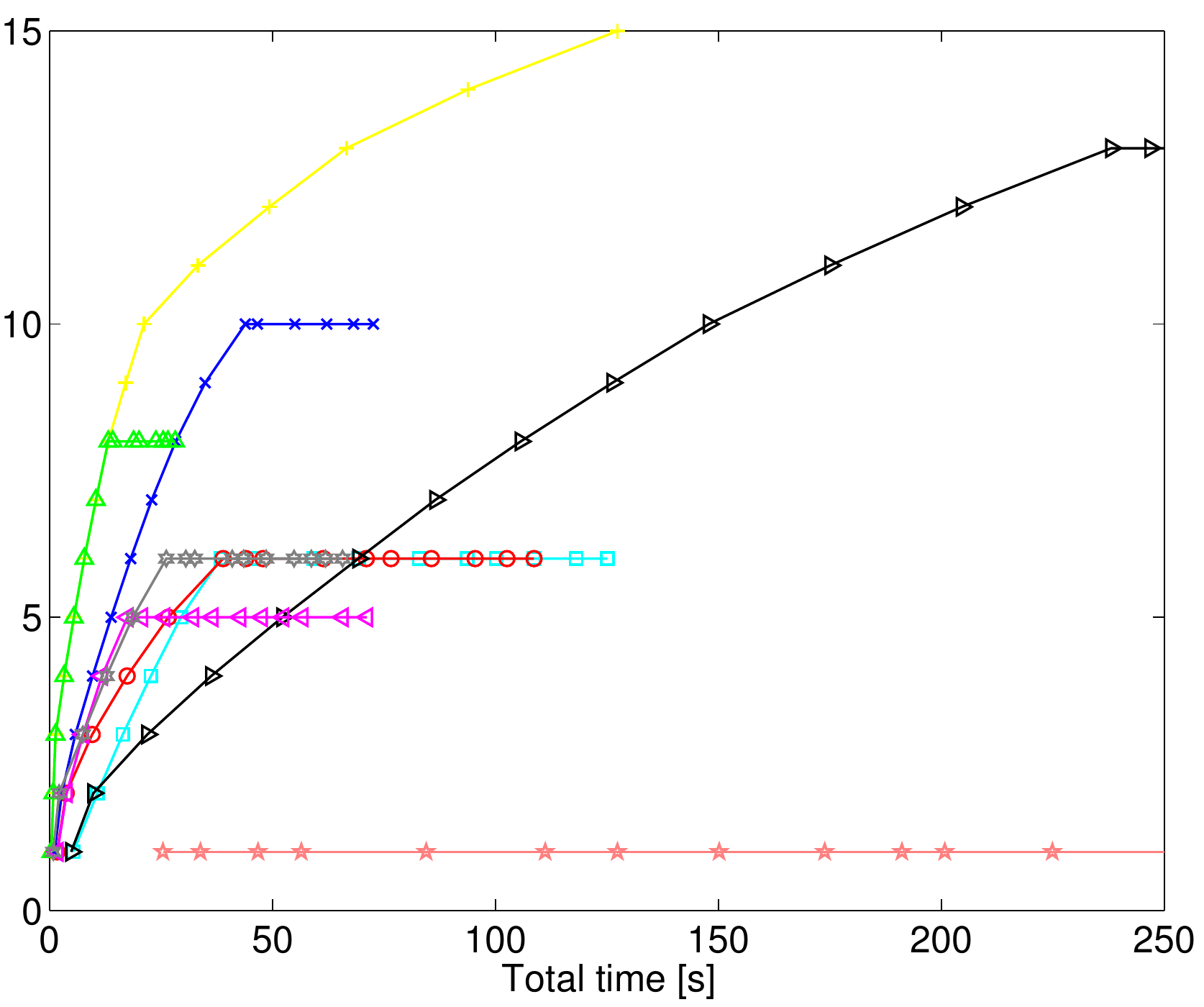}
\includegraphics[width=0.32\linewidth]{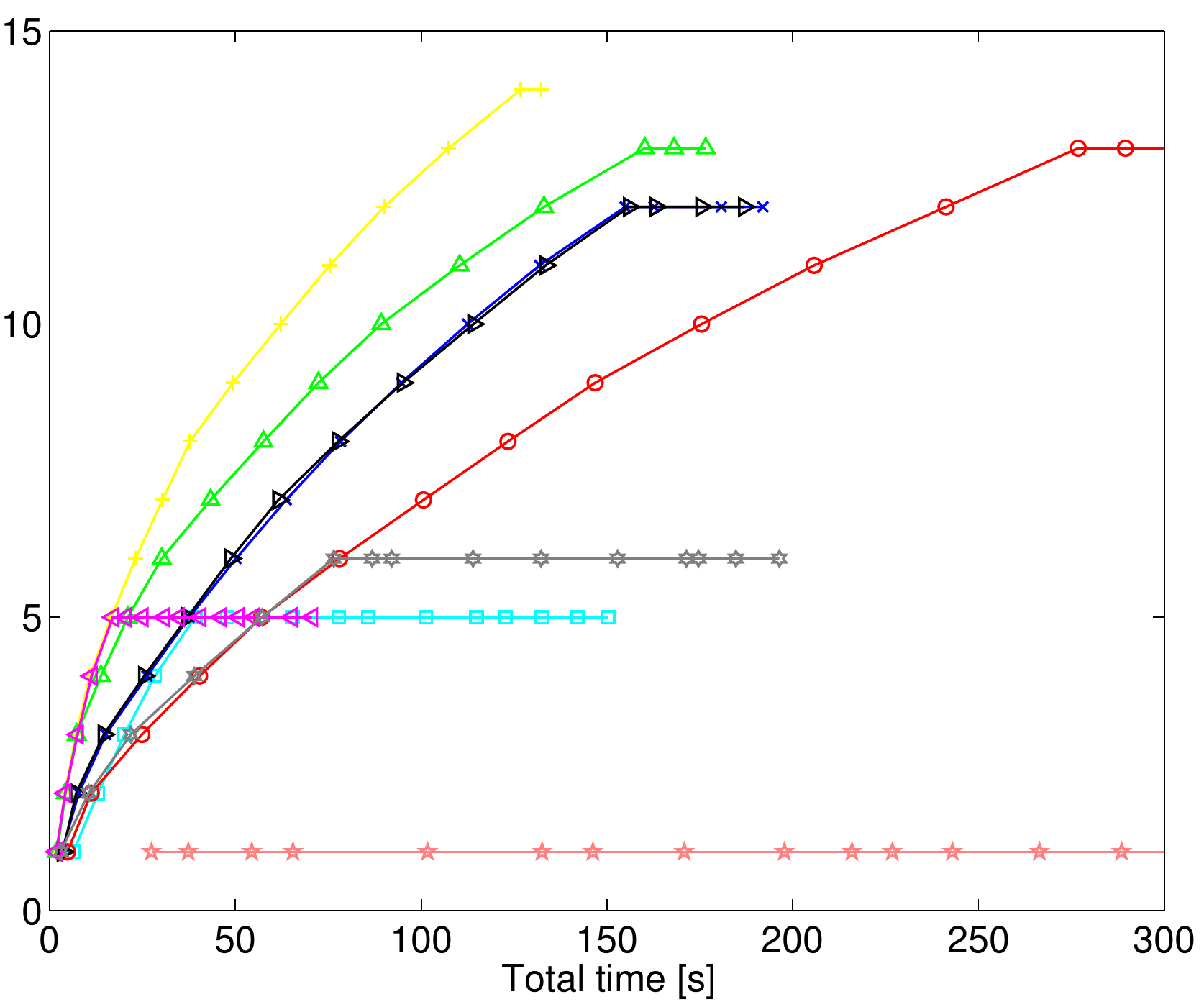}\\
\caption{Performance  of different configurations on the Extreme View Dataset. Cumulative percentage of image pairs successfully matched in a given time for configurations defined in
Table~\ref{tab:view-synth-configuration}. Each mark represents one image pair. The fastest detector configuration which was able to match each pair was selected. Left -- 'easy', middle -- 'medium', left -- 'hard' configurations. An images is considered matched if 10 correct inliers were found.}
\label{fig:evd-configuration-performance}
\end{figure}

Figure~\ref{fig:evd-configuration-performance} compares the different view synthesis configurations. Note that no single detector solved all image pairs. The Hessian-Affine, MSER, Harris-Affine and DoG successfully solved resp. 13, 13, 12, and 13 out of the 15 image pairs, however, at the expense of the high computational cost. We also noticed that if one would know the suitable detector and configuration for each image, it is possible to match all image pairs. 

\begin{figure}[tb]
\centering
\includegraphics[width=0.49\linewidth]{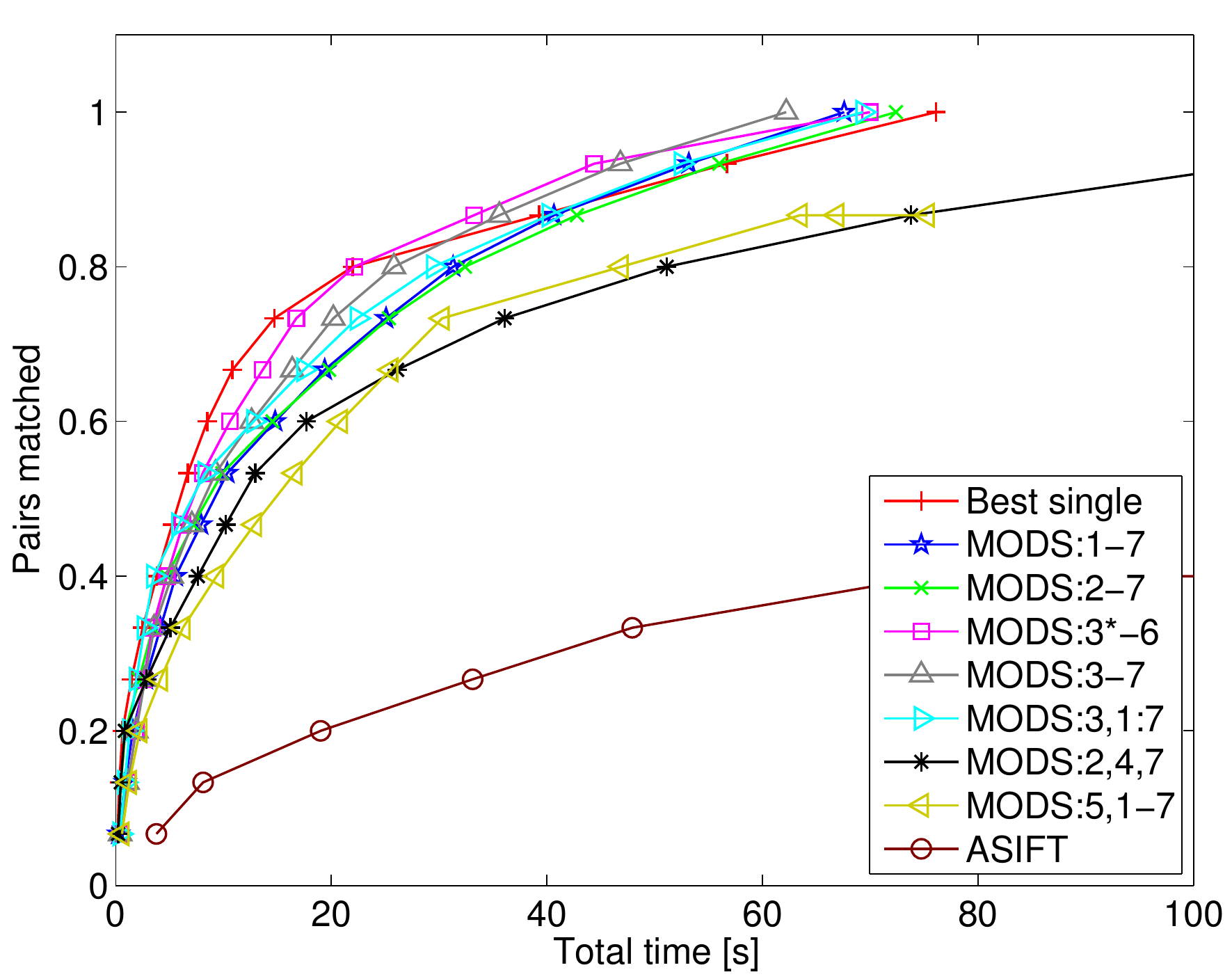}
\includegraphics[width=0.46\linewidth]{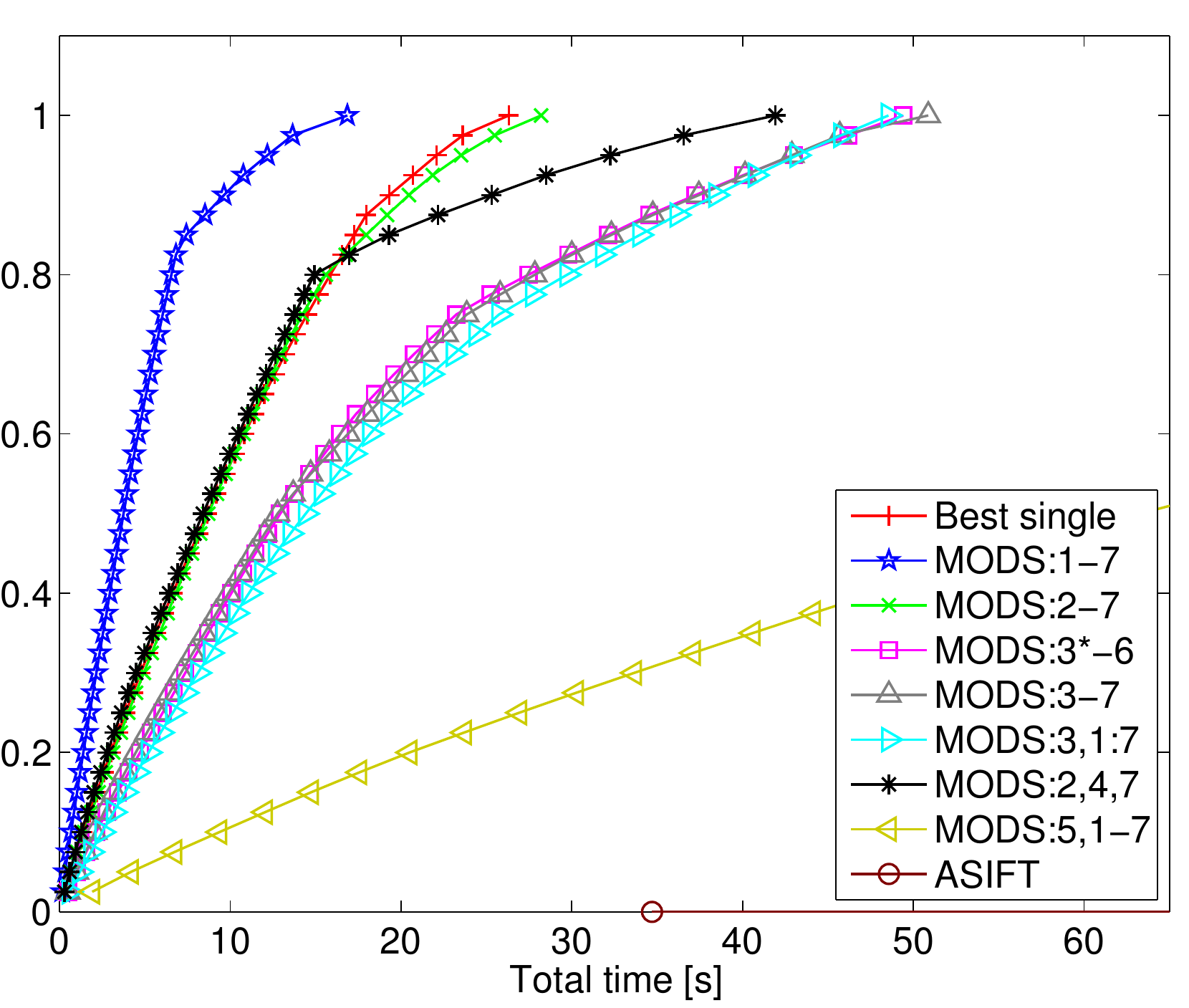}\\
\caption{Performance of MODS configurations specified in Table~\ref{tab:MODS-variants}. Cumulative percentage of image pairs successfully matched in time. Left - EVD dataset, right - Oxford dataset. The graphs are cropped. 
Each mark represents one image pair. The fastest single detector configuration which is able to match each pair was selected and plotted separately. Image considered matched if 10 correct inliers was found.}
\label{fig:MODS-variants-performance}
\end{figure}

The MODS algorithm with more time-consuming configurations solves all image pairs and does it faster than a suitable configuration for each image pair -- see Figure~\ref{fig:MODS-variants-performance}. We have tested several variants of the MODS configurations, stated in Table~\ref{tab:MODS-variants}.
Experiments shows that the proposed MODS configuration is very fast on the easy WBS problems as in Oxford dataset (see Figure~\ref{fig:MODS-variants-performance}, right graph) and has very little overhead on the harder EVD dataset -- it is the second best after the configuration without ORB steps. The results of the MODS medium configuration -- without the first sparse synthesis step -- shows the fruitfulness of the progressive view synthesis. 

ASIFT is able to match only 6 image pairs from the dataset. The ASIFT algorithm generates a lower number of correct inliers and works slower than our identical DoG configuration (which has the same tilt-rotation set). The main causes are the elimination of "one-to-many", including correct, correspondences, the inferiority of the standard second closest ratio matching strategy, and a simple brute-force algorithm of matching used in ASIFT.

\begin{figure}[htb]
\centering
\includegraphics[width=.99\linewidth]{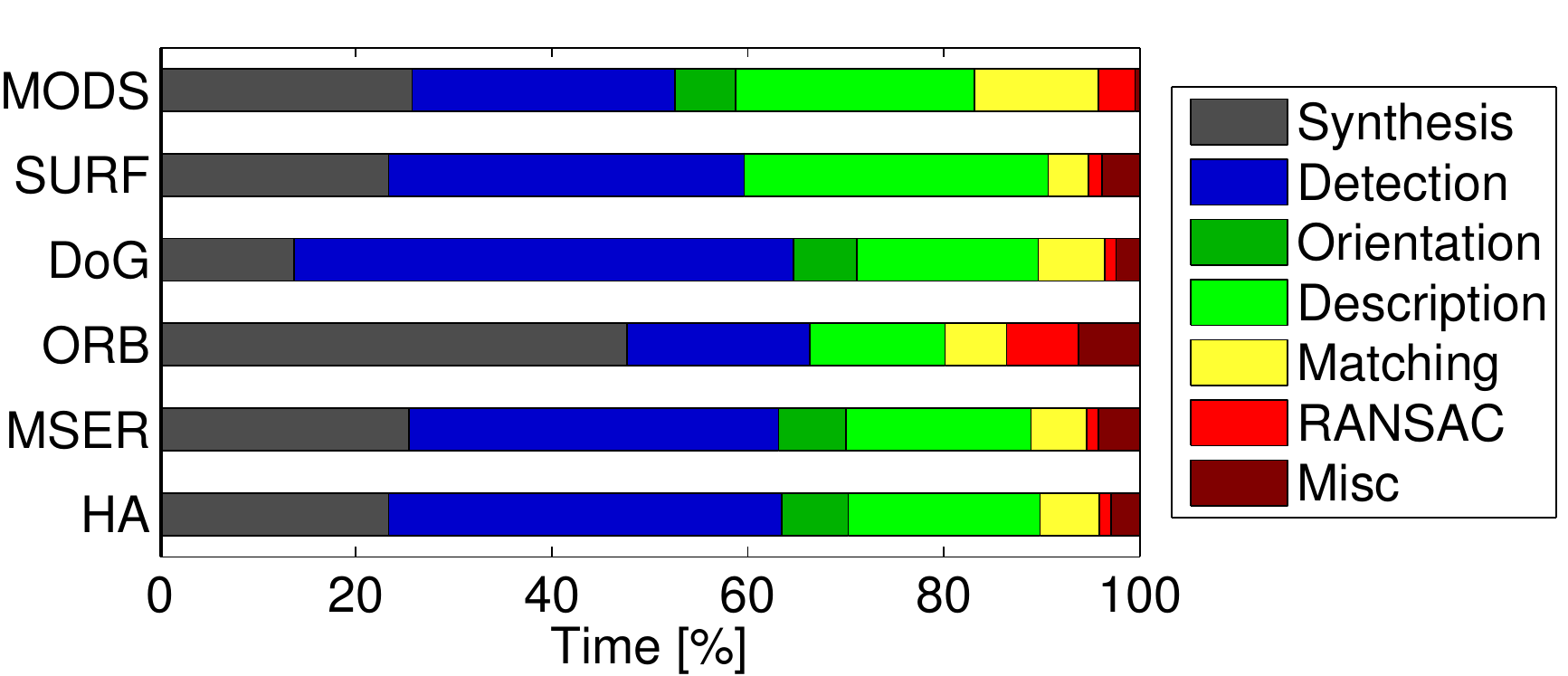} 
\caption{Percentages of time spent in the main stages of the matching with view synthesis process on a single core, {\sc easy} configuration. Detection and SIFT description, i.e. the dominant gradient estimation and the descriptor computation are the most time-consuming parts.}
 \label{fig:time-consist}
\end{figure}

Fig.~\ref{fig:time-consist} shows the breakdown of the computational time. The most time-consuming parts -- detection and description (including the dominant orientation estimation)  -- take 40\%  and 35\%  resp. of all time. Without applying the fast SIFT computation from ~\cite{PatchExtraction2014}, the SIFT description takes more than 50\% of the time. The ORB is an exception - the synthesis is not so profitable, since it takes more time than the detection and description itself. Note that the whole process is almost linear in the area of the synthesized views. The only super-linear part, matching, takes only 10\% of the time. 

\subsection{MODS testing on other datasets}
\label{main:other-datasets}
\subsubsection{Extreme zoom dataset}
\label{experiments:extrazoom}
\begin{figure}[tb]%
\includegraphics[height=0.33\linewidth]{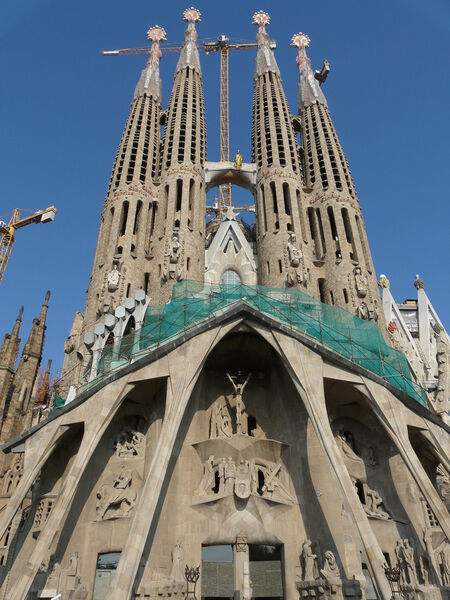}
\includegraphics[height=0.33\linewidth]{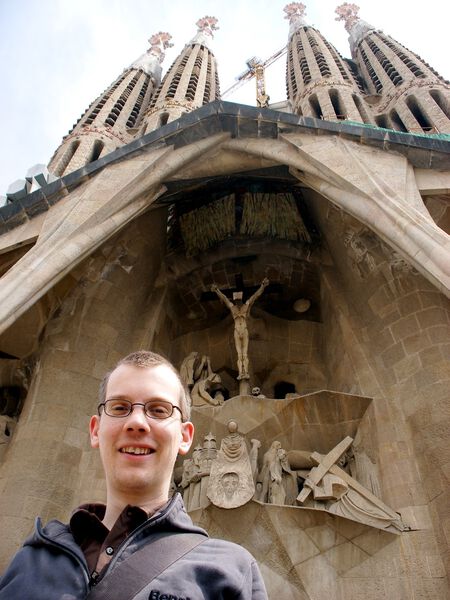}
\includegraphics[height=0.33\linewidth]{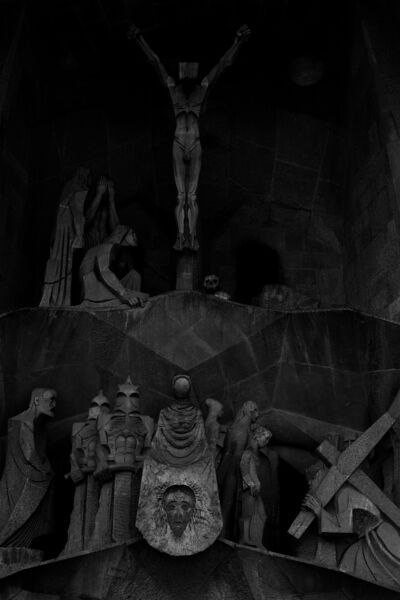}
\includegraphics[height=0.33\linewidth]{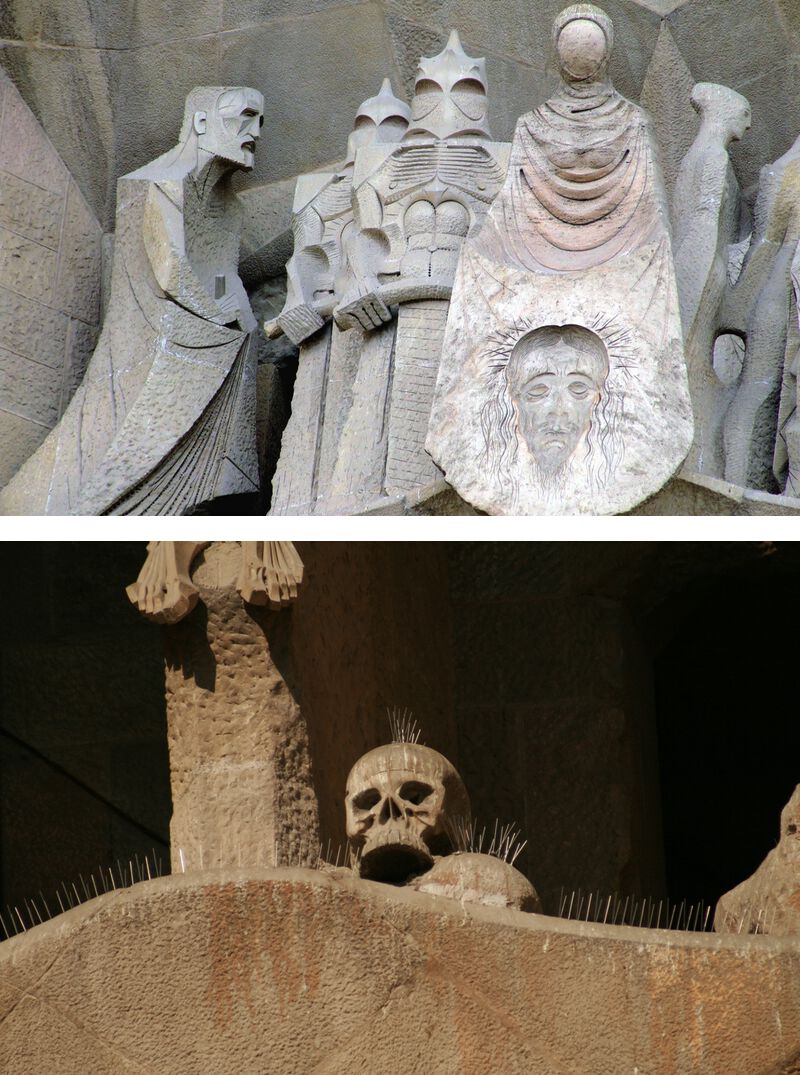}
\caption{Examples from the  Extreme Zoom Dataset}
\label{fig:extra-zoom-examples}
\centering
\end{figure}

We introduce the  Extreme Zoom dataset (EZD), which is a small subset of the retrieval dataset used in~\cite{Mikulik2014}. It consists of six sets of images with an increasing level of zoom (see examples in Figure~\ref{fig:extra-zoom-examples}).
The state-of-the art matcher -- ASIFT~\cite{ASIFT2009} and registration algorithm DualBootstrap~\cite{Yang2007} as well a results for MSER, ORB and Hessian-Affine matchers without view synthesis were compared to MODS. Image pairs are matched with the tested algorithms. An image pair is considered solved when at least 10 output correspondences are correct.  Results are shown in Table~\ref{tab:extreme-zoom-results}.
\begin{table}[htb]
\caption{Results on the Extreme Zoom Dataset.}
\label{tab:extreme-zoom-results}
\footnotesize
\centering
\setlength{\tabcolsep}{.3em}
\begin{tabular}{lcccc}
\toprule
& \multicolumn{4}{c}{Zoom level,  matched}\\
\cmidrule(r){2-5}
Matcher& I (max 6)&II (max 6) & III (max 6) & IV (max 4) \\ 
\cmidrule(r){1-5}
ORB & 0 & 0 & 0& 0\\
MSER & 2 & 2 & 1 & 0\\
DBstrap & 3 & 1 & 0 & 0\\
HessAff & 5& \textbf{3} & 2 & 0  \\
ASIFT & 3 & \textbf{3} & 2 & 0\\
\cmidrule(r){1-5}
MODS & \textbf{6} & \textbf{3} & \textbf{3} & 0\\
\bottomrule
\end{tabular}
\end{table}
\subsubsection{Ultra wide baseline dataset}
\label{experiments:ultrawide}
\begin{figure}[tb]%
\includegraphics[width=0.24\linewidth]{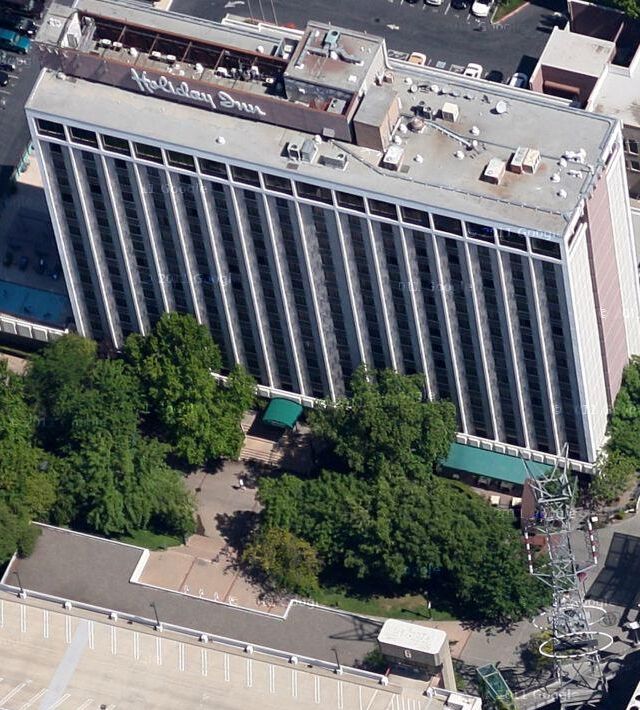}
\includegraphics[width=0.24\linewidth]{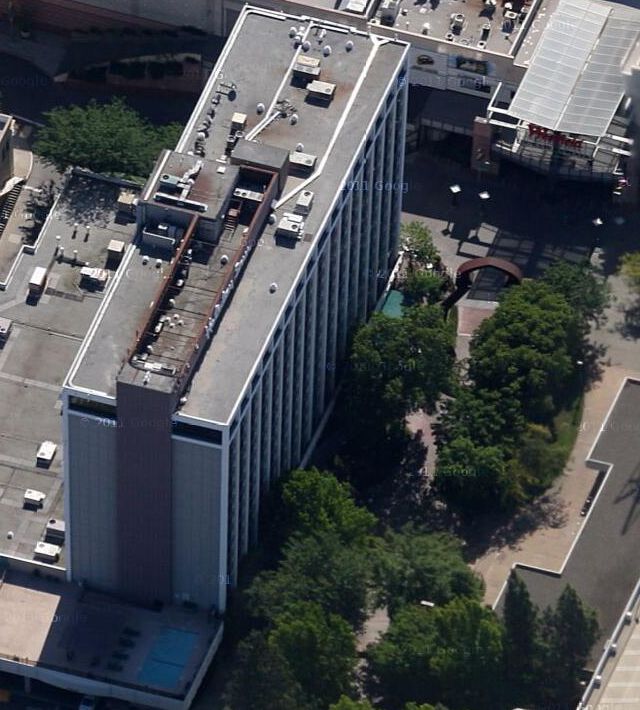}
\includegraphics[width=0.24\linewidth]{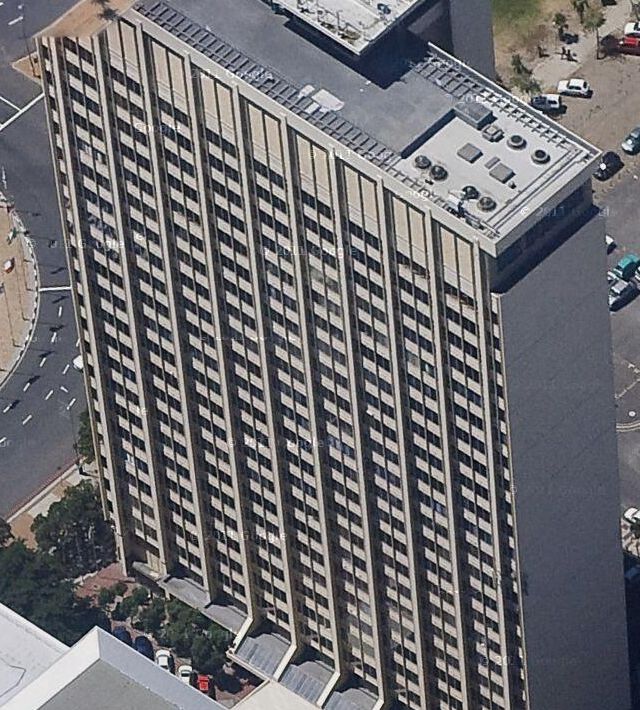}
\includegraphics[width=0.24\linewidth]{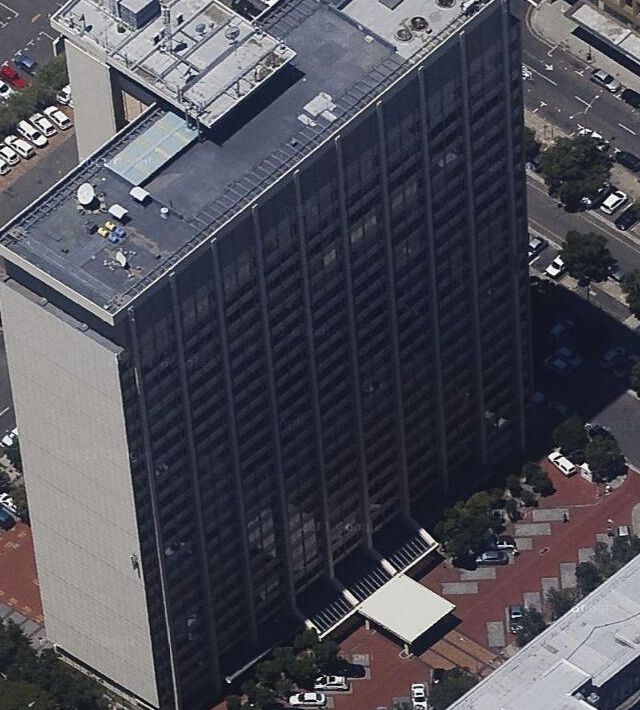}
\caption{Examples of image pairs from the Ultra wide-baseline dataset~\cite{Altwaijry2013}}
\label{fig:ultra-wide-examples}
\centering
\end{figure}

The MODS performance was evaluated on the city-from-air dataset from \cite{Altwaijry2013}. The dataset comprises 30 pairs (examples shown in Figure~\ref{fig:ultra-wide-examples}) of photographs of buildings taken from the air. The viewpoint difference is quite large, the images contain repeated structures, illumination differs. The authors proposed a matcher based on HoG~\cite{HoG2005} descriptor with a view synthesis and compare it to ASIFT and D-Nets~\cite{DNets2012} for skyscraper frontal face matching. We follow the evaluation protocol which considers a pair matched correct only if the facade plane is matched ($\geq 75\%$ correct inliers). If the output homography was ground/roof, it is considered incorrect. The results are shown in Table~\ref{tab:ultra-wide-results}.

Note that no special adjustment is done in MODS for homography selection, so the reported performance is a lower bound.
\begin{table}[htb]
\caption{Results on Ultra wide-baseline dataset (all results except MODS taken from ~\cite{Altwaijry2013})}
\label{tab:ultra-wide-results}
\footnotesize
\centering
\setlength{\tabcolsep}{.3em}
\begin{tabular}{llll}
\toprule
Method& Correct& Different & Failure/\\
&(or shifted)&  plane & ground plane\\
\cmidrule(r){1-4}
Altwaijry and Belongie~\cite{Altwaijry2013} & \textbf{9} &  1& 20 \\ 
ASIFT-homography & 1& 5 & 24 \\ 
D-Nets & 7& 2& 21 \\ 
\cmidrule(r){1-4}
MODS & 8& 1&21 \\ 
\bottomrule
\end{tabular}
\end{table}
\subsubsection{Datasets with other than geometric changes}
\label{experiments:other}
Despite being designed for (extreme) wide baseline stereo problems, MODS performance was evaluated on other datasets: GDB-ICP~\cite{Kelman2007} (modality, viewpoint and photometry changes), SymBench~\cite{Hauagge2012} (photometrical changes and photo-vs-painting pairs),  and MMS~\cite{Aguilera2012} (infrared-vs-visible pairs) -- see Table~\ref{tab:other-datasets}. The state-of-the-art matcher -- ASIFT~\cite{ASIFT2009} and registration algorithm DualBootstrap~\cite{Yang2007} as well,  as the results of MSER, ORB and Hessian-Affine matchers without view synthesis were compared to MODS.

\begin{table}[htb]
\caption{Evaluation Datasets}
\label{tab:other-datasets}
\footnotesize
\centering
\setlength{\tabcolsep}{.3em}
\begin{tabular}{lll}
\toprule
Short name& \#images & Nuisance\\
\cmidrule(r){1-3}
GDB-ICP~\cite{Kelman2007}, 2007&22 pairs& Illumination, modality\\
SymBench~\cite{Hauagge2012}, 2012&46 pairs&Illumination, modality\\
MMS~\cite{Aguilera2012}, 2012&100 pairs&Modality\\
EVD~\cite{Mishkin2013}, 2013&15 pairs&Viewpoint change\\
\bottomrule
\end{tabular}
\end{table}
Image pairs are matched with the tested algorithms. Output keypoint correspondences were checked against the ground truth homographies. An image pair is considered  solved when at least 10 output correspondences are correct
Our primary evaluation criterion is the ability to find sufficiently correct geometric transformations in a reasonable time;  accurate geometry can be found in a consecutive step.

The computations were performed on Intel i3 3.0GHz desktop (4 cores) with 4Gb RAM.

Results are shown in Table~\ref{tab:other-results}. MODS is the fastest method and it is able to match most image pairs in GDB-ICP and SymBench datasets without using  symmetrical parts or other problem-specific features. Images from the MMS dataset (as well as other thermal images) produce a small number of features as they do not contain many textured surfaces and have a very short geometrical baseline. Those are the main reasons why area-based method -- Dual-Bootstrap -- works significantly better than the feature-based methods.

After lowering the threshold for detectors allowing to detect more feature points and using the orientation restricted SIFT~\cite{HalfSIFT2007} in addition to the RootSIFT, MODS-IR solved 83 out of 100 image pairs from MMS dataset. 
\begin{table}[htb]
\caption{Performance on non-WBS datasets. For comparison, results of MSER, ORB and Hessian-Affine matchers without view synthesis are added. Best results are in \textbf{bold}, average time per image pair is shown.}
\label{tab:other-results}
\footnotesize
\centering
\setlength{\tabcolsep}{.35em}
\begin{tabular}{lrrrrrrrr}
\toprule

& \multicolumn{2}{c}{GDB-ICP}
& \multicolumn{2}{c}{SymBench}
& \multicolumn{2}{c}{MMS}
& \multicolumn{2}{c}{EVD}\\
\cmidrule(r){2-9}
Matcher
& \shortstack{pairs \\ solved}& \shortstack{time \\ $[s]$}
& \shortstack{pairs \\ solved}& \shortstack{time \\ $[s]$}
& \shortstack{pairs \\ solved}& \shortstack{time \\ $[s]$}
& \shortstack{pairs \\ solved}& \shortstack{time \\ $[s]$} \\
\cmidrule(r){1-9}
ASIFT &         15/22 &         41.5&          27/46&        14.7&          8/100&          3.2&           5/15& 12.4\\
MODS:1-7  & \textbf{17/22}& \textbf{2.8}&  \textbf{38/46}&\textbf{3.7}&        12/100&  \textbf{2.0}& \textbf{15/15}& \textbf{2.4}\\
DBstrap &       16/22 &         17.6&  \textbf{38/46}&        21.7&\textbf{79/100}&         9.3&           0/15 &1.9\\
ORB &            0/22 &         0.2&   0/46&         0.4&          0/100&          0.1&           0/15 &0.3\\
MSER &           8/22 &         1.7&          21/46 &         0.8&           0/100&         0.2&           4/15 &0.6\\
HessAff &        11/22 &        1.9&          29/46 &         1.5&          2/100&          0.4&           2/15 &1.1\\
MODS-IR:1-7  & \textbf{21/22}&      7.6&  \textbf{39/46}&\textbf{15.1}& \textbf{83/100}&  \textbf{9.1}& 14/15& 8.6\\
\bottomrule
\end{tabular}
\end{table}

\subsection{MODS testing on a non-planar dataset}
\label{main:3d-experiments}
\subsubsection{The dataset and evaluation protocol}
\label{experiments:protocol}
\begin{table}[tbp]%
\caption{Reference views of the image sequences used in the evaluation (from~\cite{Moreels2007}).}
\label{tab:dataset}
\centering
\bgroup
\def\arraystretch{0.9}%
\setlength{\tabcolsep}{.3em}
\begin{tabu}spread 0pt{|X[-1m,c]|X[-1m,c]|X[-1m,c]|X[-1m,c]|}
\hline
Abroller& Bannanas & Camera2 &Car \\
\includegraphics[width=0.8\linewidth]{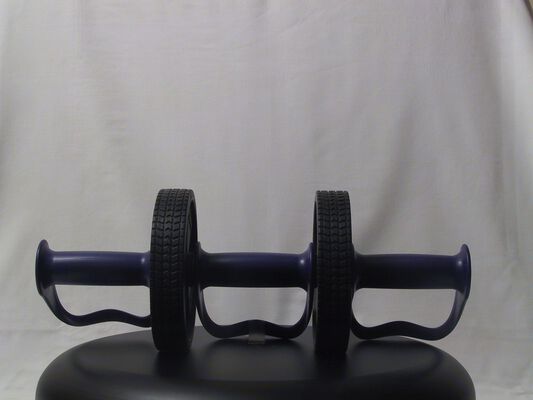} 
&\includegraphics[width=0.8\linewidth]{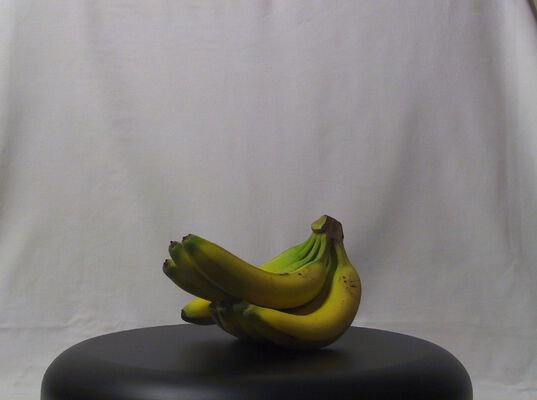} 
&\includegraphics[width=0.8\linewidth]{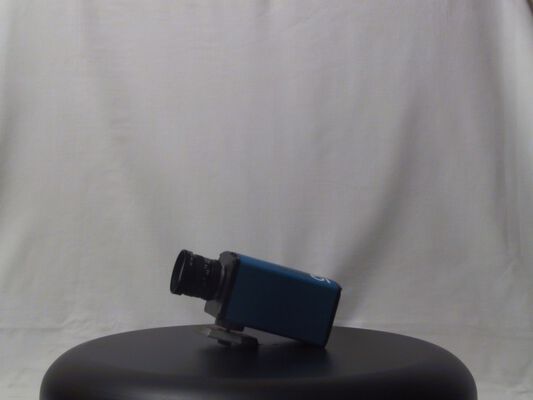} 
&\includegraphics[width=0.8\linewidth]{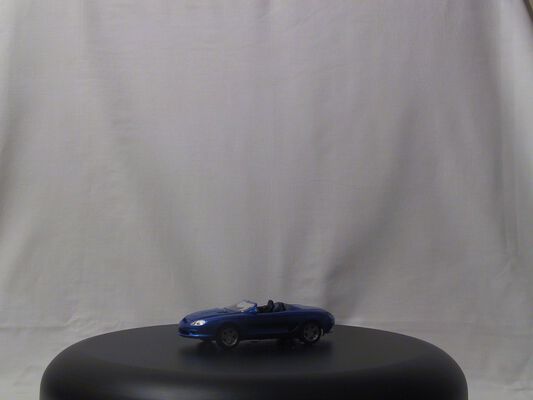}\\ 
\hline
Car2& CementBase & Cloth &Conch \\
\includegraphics[width=0.8\linewidth]{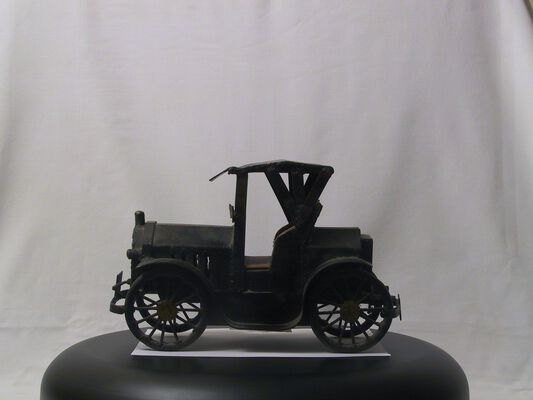} 
&\includegraphics[width=0.8\linewidth]{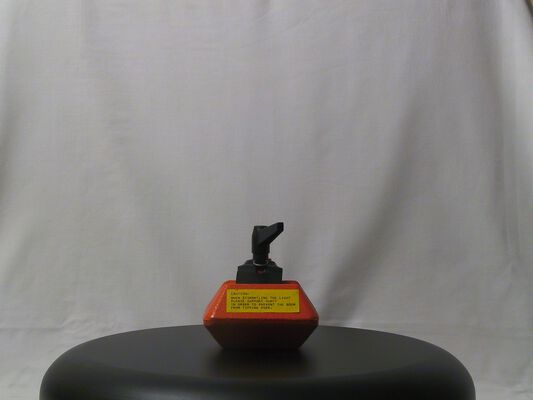} 
&\includegraphics[width=0.8\linewidth]{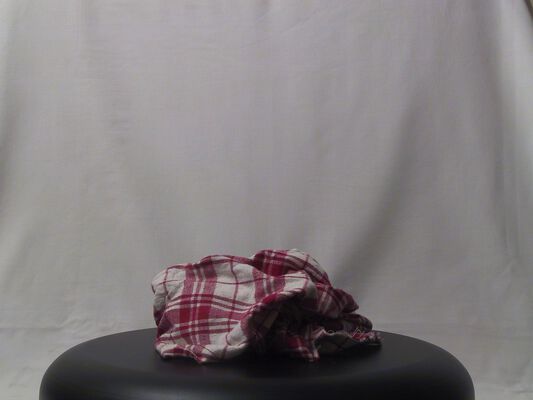} 
&\includegraphics[width=0.8\linewidth]{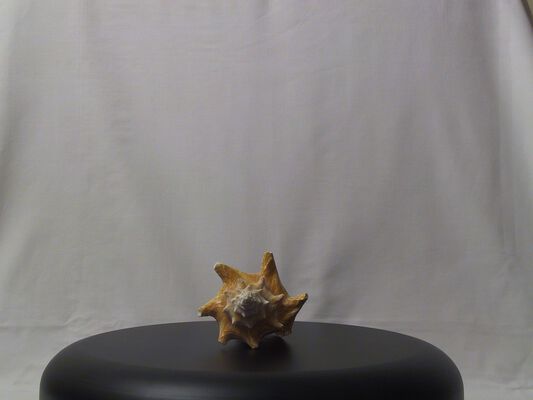}\\ 
\hline
Desk& Dinosaur & Dog &DVD \\
\includegraphics[width=0.8\linewidth]{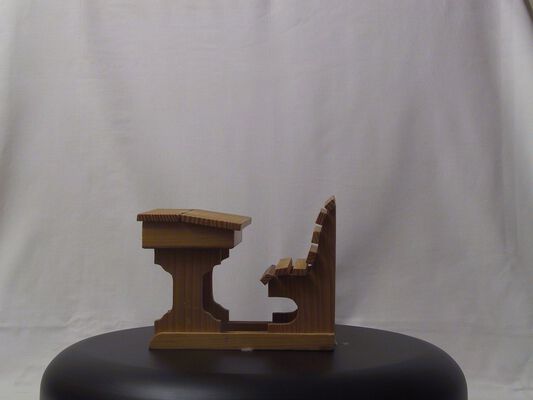} 
&\includegraphics[width=0.8\linewidth]{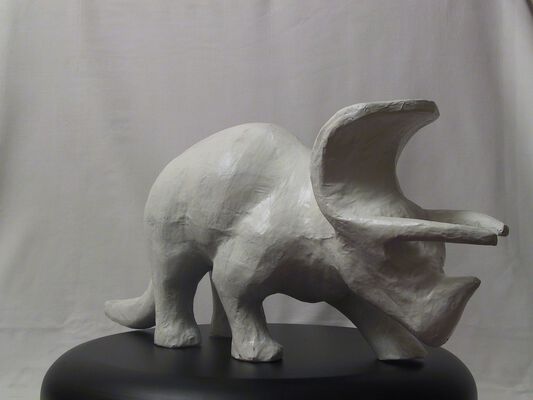} 
&\includegraphics[width=0.8\linewidth]{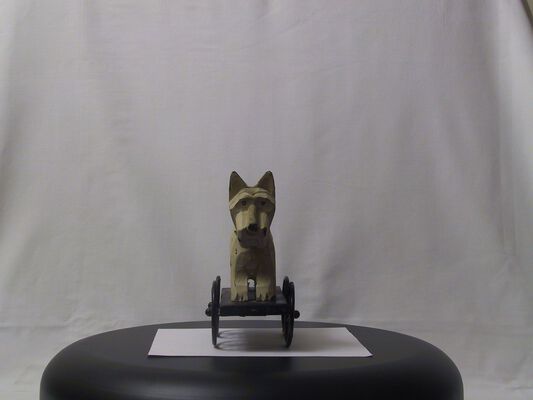} 
&\includegraphics[width=0.8\linewidth]{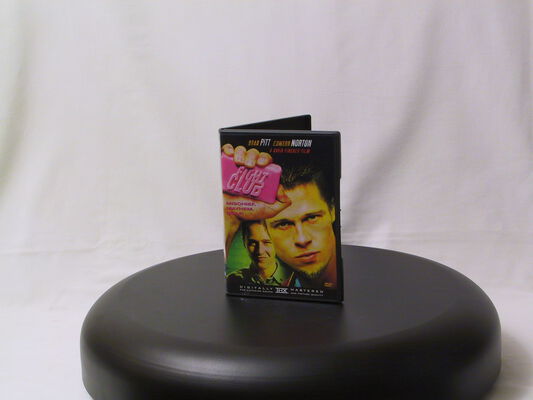}\\ 
\hline
FloppyBox& FlowerLamp & Gelsole &GrandfatherClock \\
\includegraphics[width=0.8\linewidth]{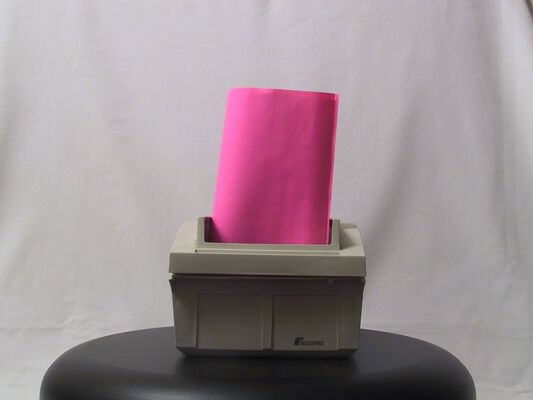} 
&\includegraphics[width=0.8\linewidth]{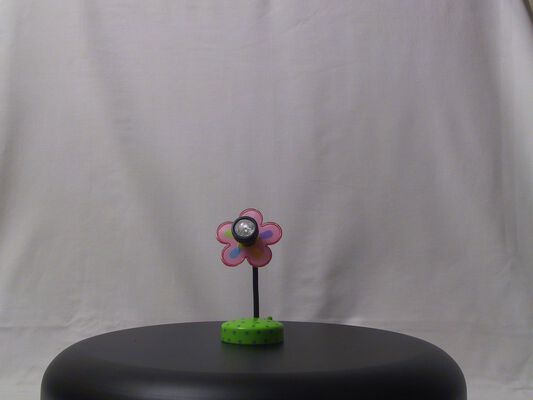} 
&\includegraphics[width=0.8\linewidth]{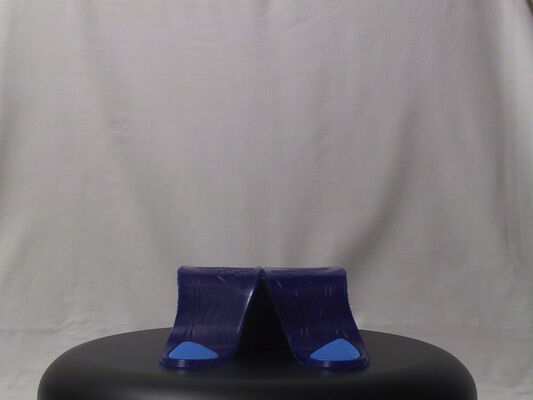} 
&\includegraphics[width=0.8\linewidth]{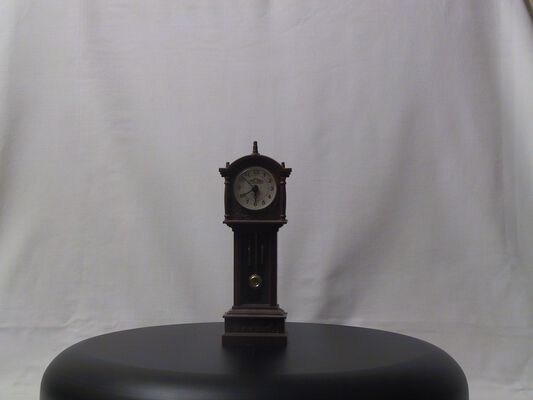}\\ 
\hline
Horse& Keyboard & Motorcycle &MouthGuard \\
\includegraphics[width=0.8\linewidth]{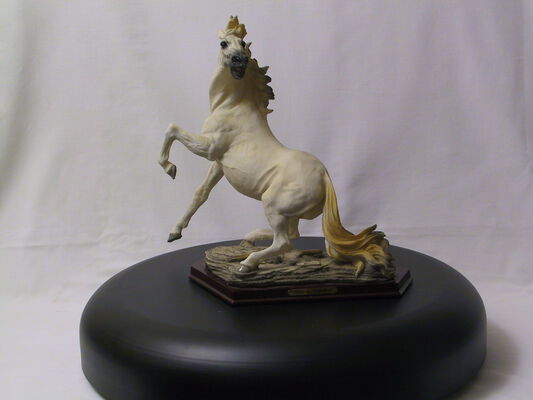} 
&\includegraphics[width=0.8\linewidth]{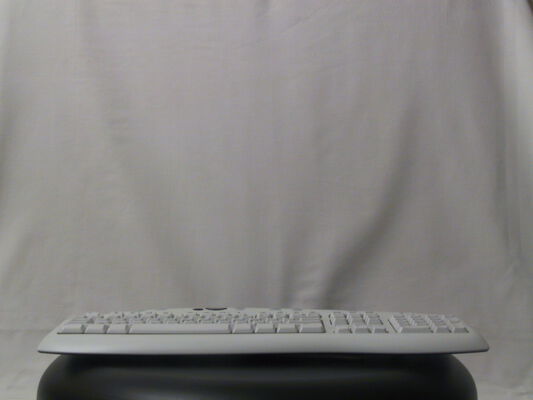} 
&\includegraphics[width=0.8\linewidth]{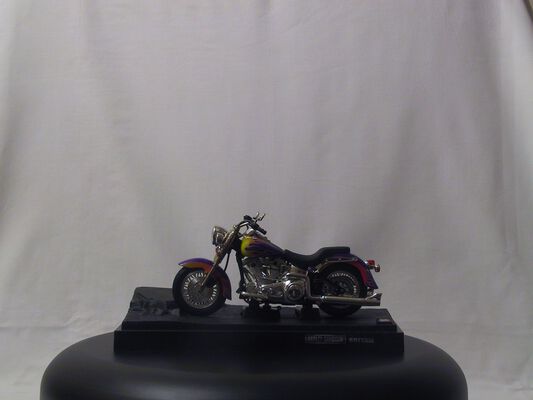} 
&\includegraphics[width=0.8\linewidth]{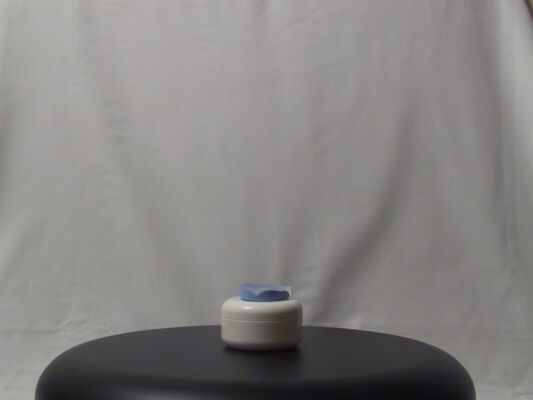}\\ 
\hline
PaperBin& PS2 & Razor &RiceCooker \\
\includegraphics[width=0.8\linewidth]{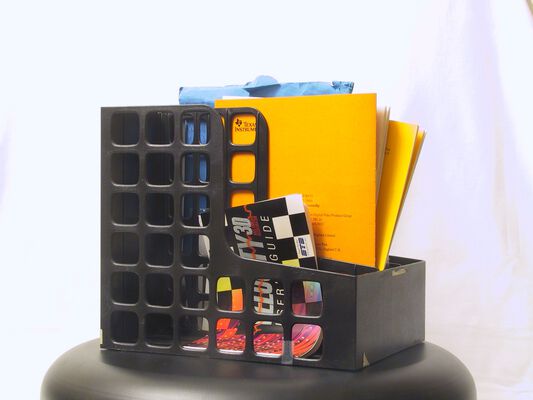} 
&\includegraphics[width=0.8\linewidth]{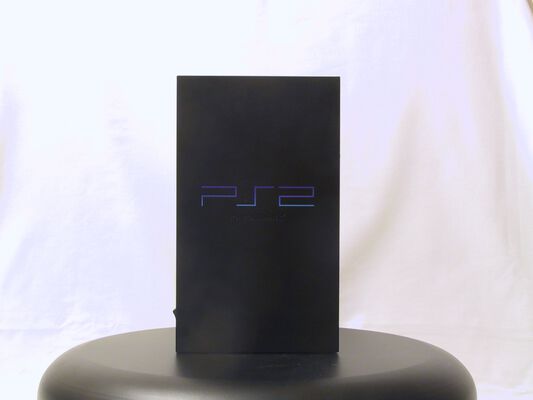} 
&\includegraphics[width=0.8\linewidth]{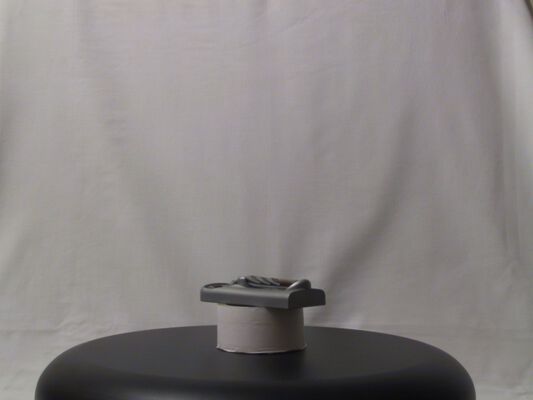} 
&\includegraphics[width=0.8\linewidth]{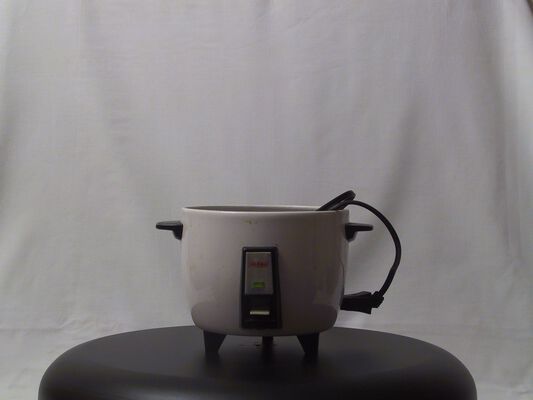}\\ 
\hline
Rock& RollerBlade & Spoons &TeddyBear \\
\includegraphics[width=0.8\linewidth]{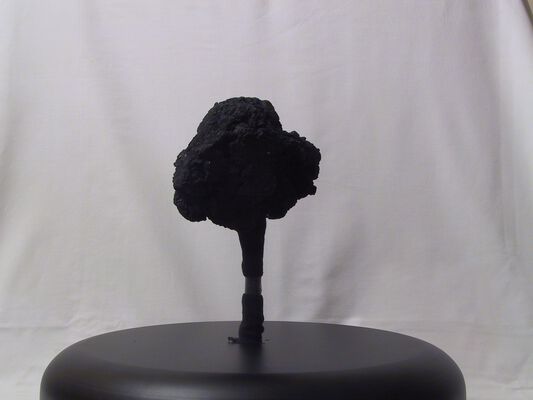} 
&\includegraphics[width=0.8\linewidth]{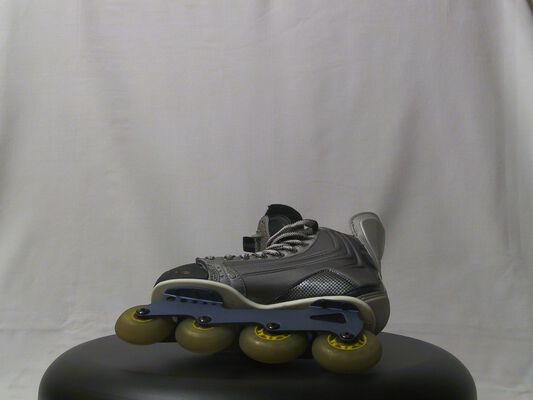} 
&\includegraphics[width=0.8\linewidth]{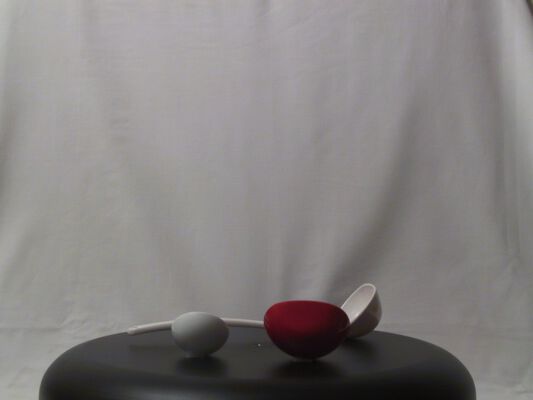} 
&\includegraphics[width=0.8\linewidth]{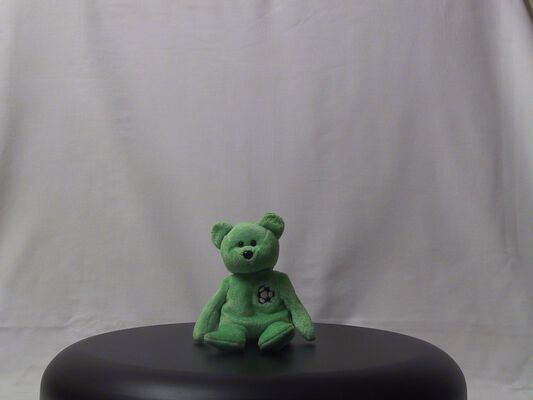}\\ 
\hline
Toothpaste& Tricycle & Tripod &VolleyBall \\
\includegraphics[width=0.8\linewidth]{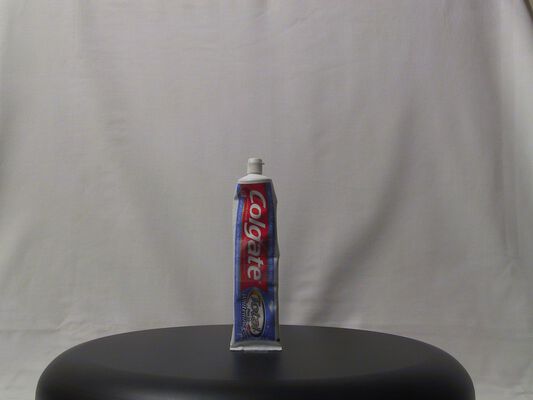} 
&\includegraphics[width=0.8\linewidth]{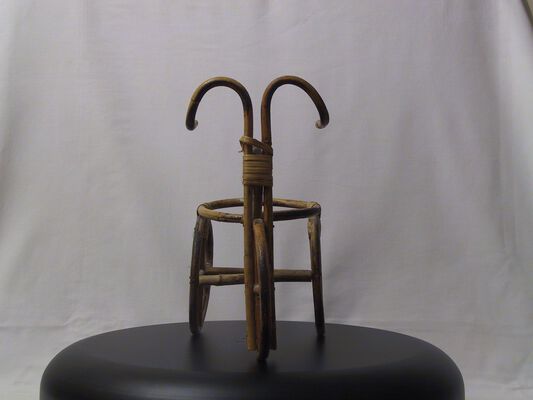} 
&\includegraphics[width=0.8\linewidth]{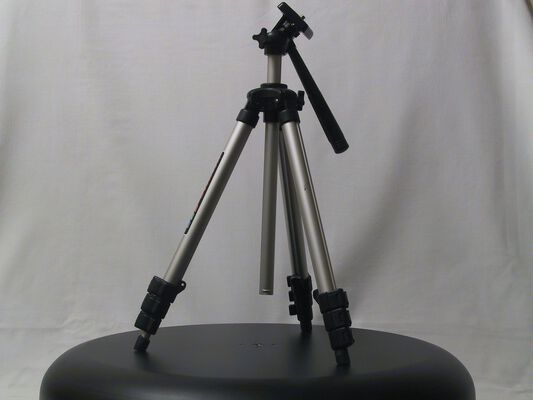} 
&\includegraphics[width=0.8\linewidth]{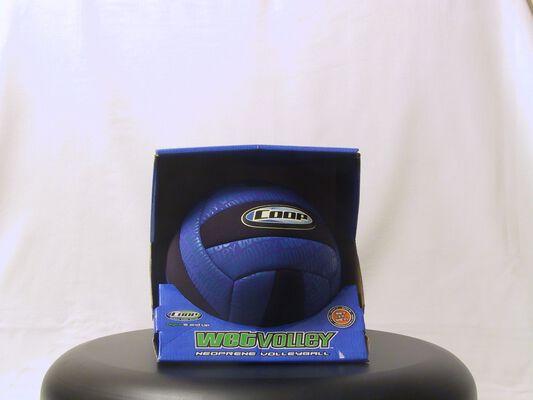}\\ 
\hline
\end{tabu}
\egroup
\end{table}%
The evaluation dataset consists of 35 image sequences taken from the Turntable dataset~\cite{Moreels2007} (``Bottom'' camera) shown in Table~\ref{tab:dataset}. Eight image sets contain objects with relatively large planar surfaces and the remaining ones are low-textured, ``general  3D'' objects.

The view marked as ``$0^{\circ}$'' in the Turntable dataset was used as a reference view and $0-90^\circ$ and 270-355~$\degree$ views  with a $5^\circ$ step were matched against it using the procedure described in Sec.~\ref{mods:main}, forming a $[-90^{\circ},90^{\circ}]$ sequence. Note that the reference view is not usually the ``frontal'' or ``side'' view, but rather some intermediate view which caused asymmetry in the results (see Fig.~\ref{fig:jet-angle}, Table~\ref{tab:results}).

The output of the matchers is a set of correspondences and the estimated geometrical transformation. The accuracy of the matched correspondences was chosen as the performance criterion, similarly to the protocol in~\cite{Dahl2011}.
For all output correspondences, the symmetrical epipolar error~\cite{Hartley2004} $e_{\text{SymEG}}$ was computed according to the following expression:
\begin{equation}
\label{eq:symmetric-error-F}
e_{\text{SymEG}}\left(\mathbf{F},\mathbf{u},\mathbf{v}\right)=\left(\mathbf{v}^{\top}\mathbf{F}\mathbf{u}\right)^2 \times \left(\frac{1}{(\mathbf{F}\mathbf{u})^2_1 + (\mathbf{F}\mathbf{u})^2_2} +
\frac{1}{(\mathbf{F}^{\top}\mathbf{v})^2_1 + (\mathbf{F}^{\top}\mathbf{v})^2_2}\right),
\end{equation}
where $\mathbf{F}$ -- fundamental matrix, $\mathbf{u}$, $\mathbf{v}$ -- corresponding points, $(\mathbf{F}\mathbf{u})^2_j$ -- the square of the $j$-th entry of the vector $\mathbf{F}\mathbf{u}$.

The ground truth fundamental matrix was obtained from the difference in camera positions~\cite{Hartley2004}, assuming that the turntable is fixed and the camera moves around the object, according to the following equation: 
\begin{multline}
\label{eq:fundamental-matrix}
F=K^{-\top}RK^{\top}[KR^{\top}t]_\times,\\
R=\left( \begin{array}{ccc}
\cos{\phi} & 0 &-\sin{\phi} \\
0&1&0\\
\sin{\phi} &0& \cos{\phi}
\end{array} \right), 
K=\left( \begin{array}{ccc}
\frac{mf}{\text{FR}_X} & 0 & \frac{m}{2} \\
0& \frac{nf}{\text{FR}_Y}  & \frac{n}{2} \\
0 &0& 1
\end{array} \right),
t=r\left( \begin{array}{ccc}
\sin{\phi} \\
0\\
1-\cos{\phi}
\end{array} \right) ,
\end{multline}
where $R$ is the orientation matrix of the second camera, $K$ -- the camera projection matrix, $t$ -- the virtual translation of the second camera, $r$ -- the distance from camera to the object, $\phi$ -- the viewpoint angle difference, $\text{FR}_X,\text{FR}_Y$ -- the focal plane resolution, $f$ -- the focal length, $m$, $n$ -- the sensor matrix width and height in pixels. The last five parameters were obtained from EXIF data. 

One of the evaluation problems is that background regions, i.e.~regions that are not on the object placed on the turntable, are often detected and matched, influencing the geometry transformation estimation. The matches are correct, but consistent with an identity transform of the (background of) the test images, not the fundamental matrix associated with the movement of the object on the turntable. To solve this problem, the median value of the correspondence errors was chosen as the measure of precision because of its tolerance to the low number of outliers (e.g.~the above-mentioned background correspondences), and its sensitivity to the incorrect geometric model estimated by RANSAC. 

An image pair is considered as \emph{correctly} matched if the median symmetrical epipolar error on the correspondences using the ground truth fundamental matrix is $\leq$~6~pixels.
\subsubsection{Results}
\label{experiments:results}
\begin{figure}[tb]%
\centering
\includegraphics[width=0.28\linewidth]{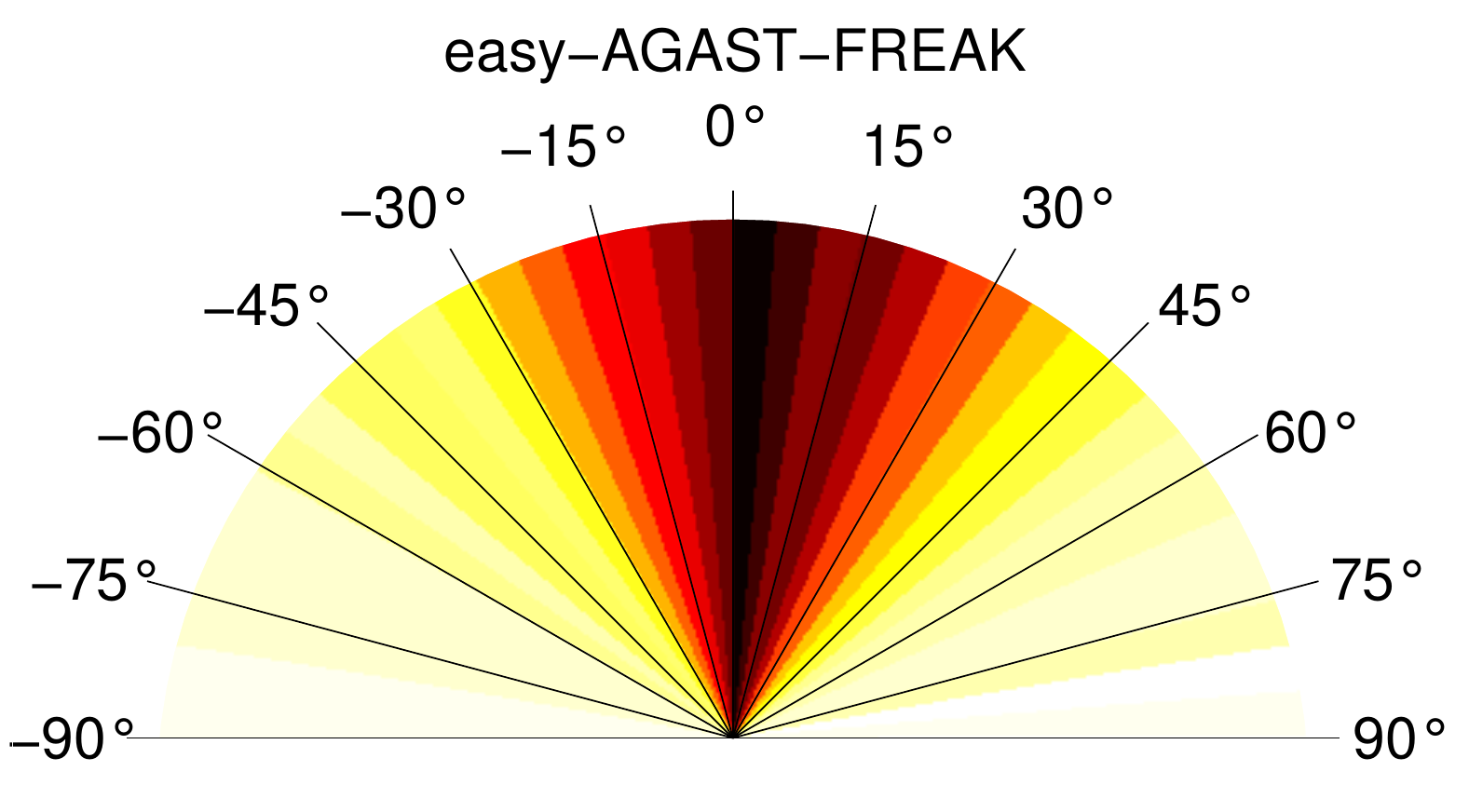}
\includegraphics[width=0.28\linewidth]{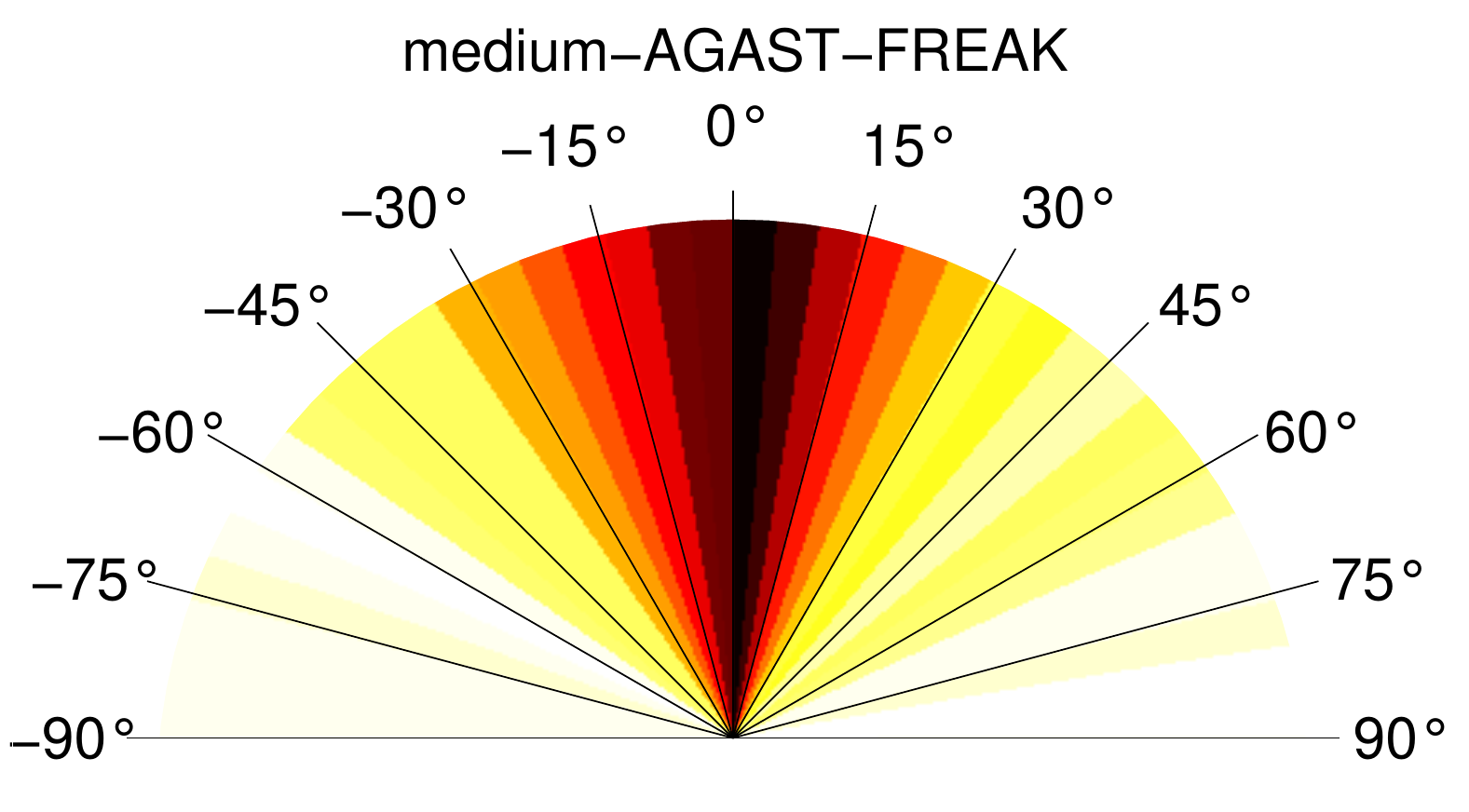}
\includegraphics[width=0.30\linewidth]{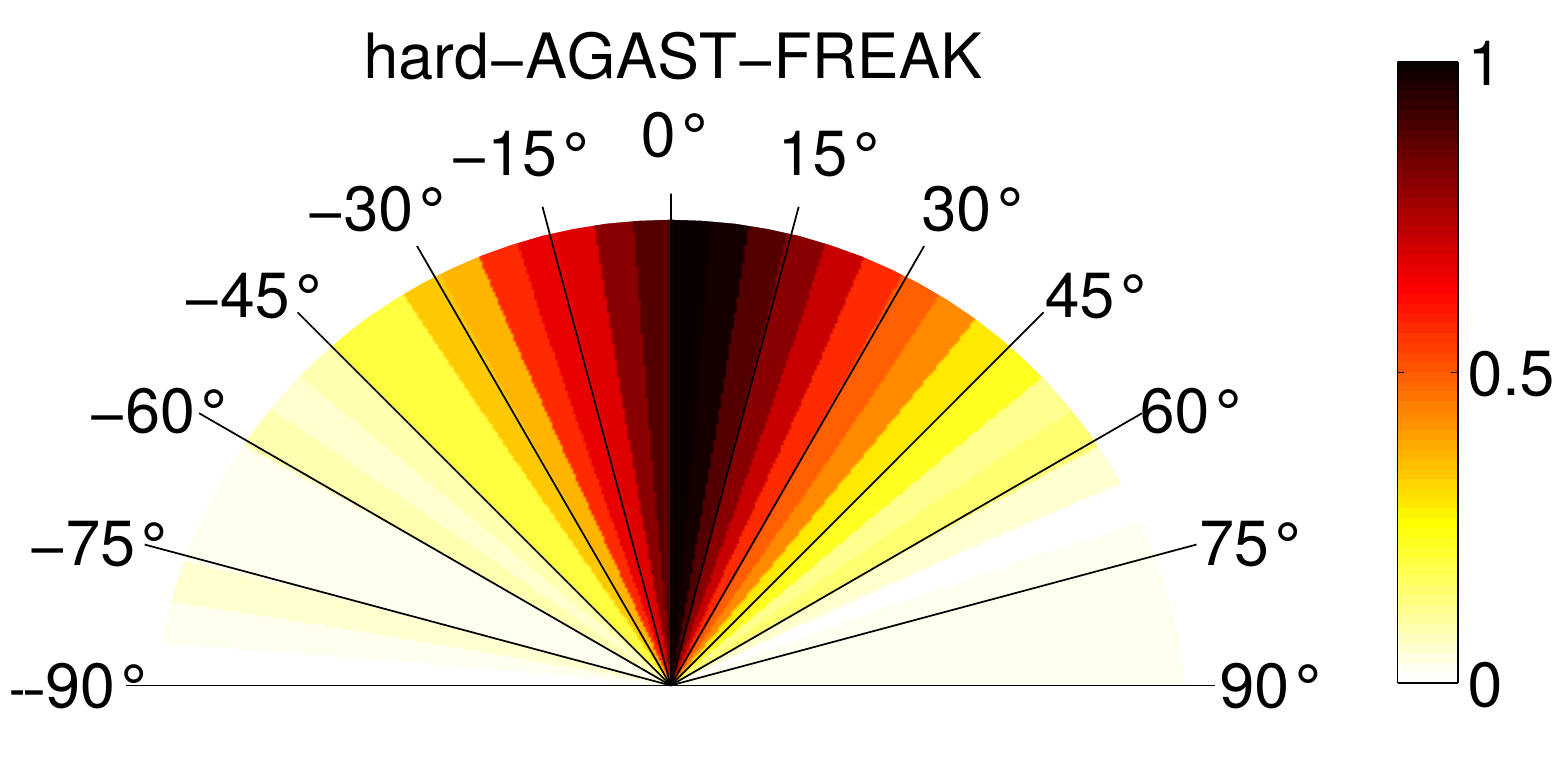}\\
\includegraphics[width=0.28\linewidth]{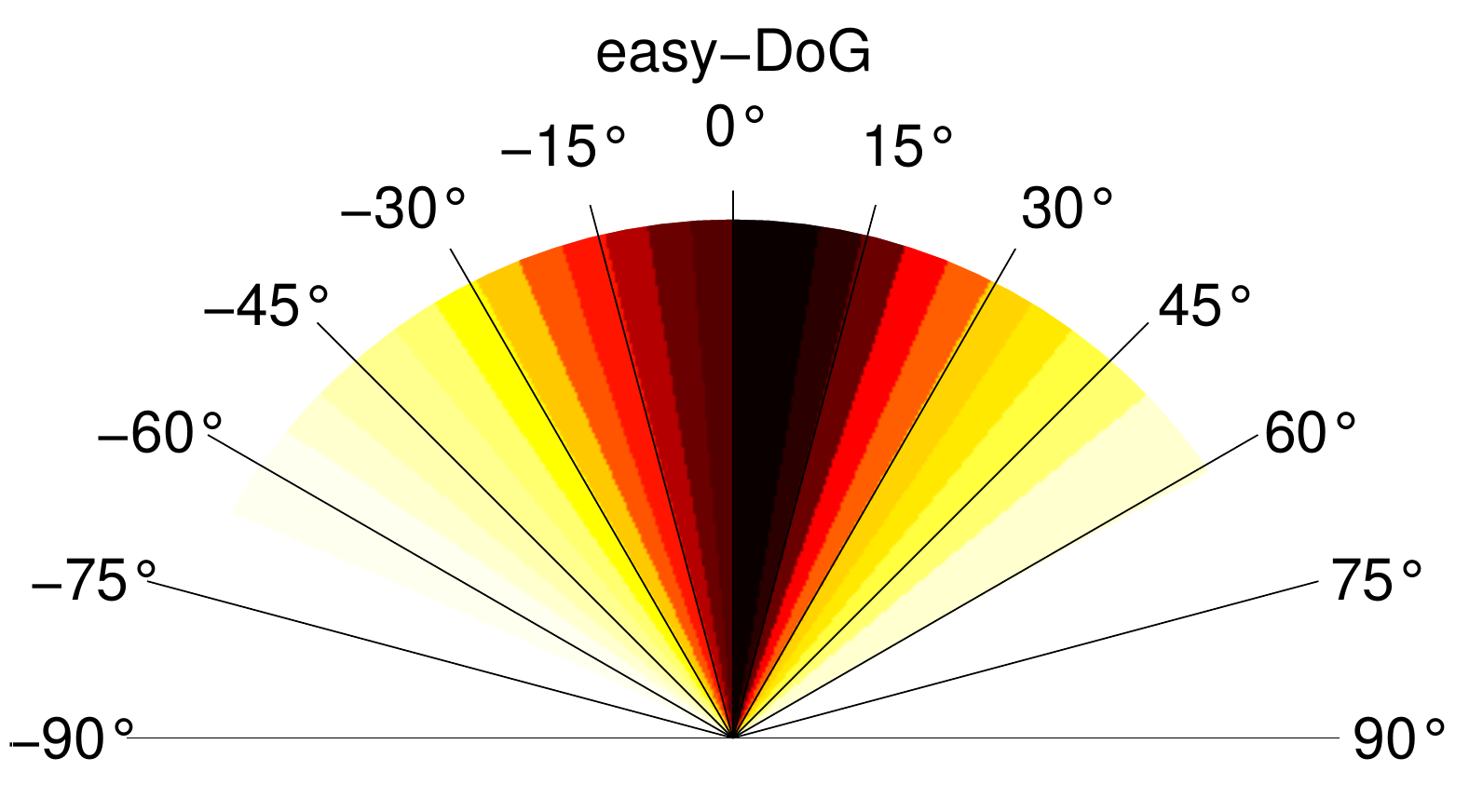}
\includegraphics[width=0.28\linewidth]{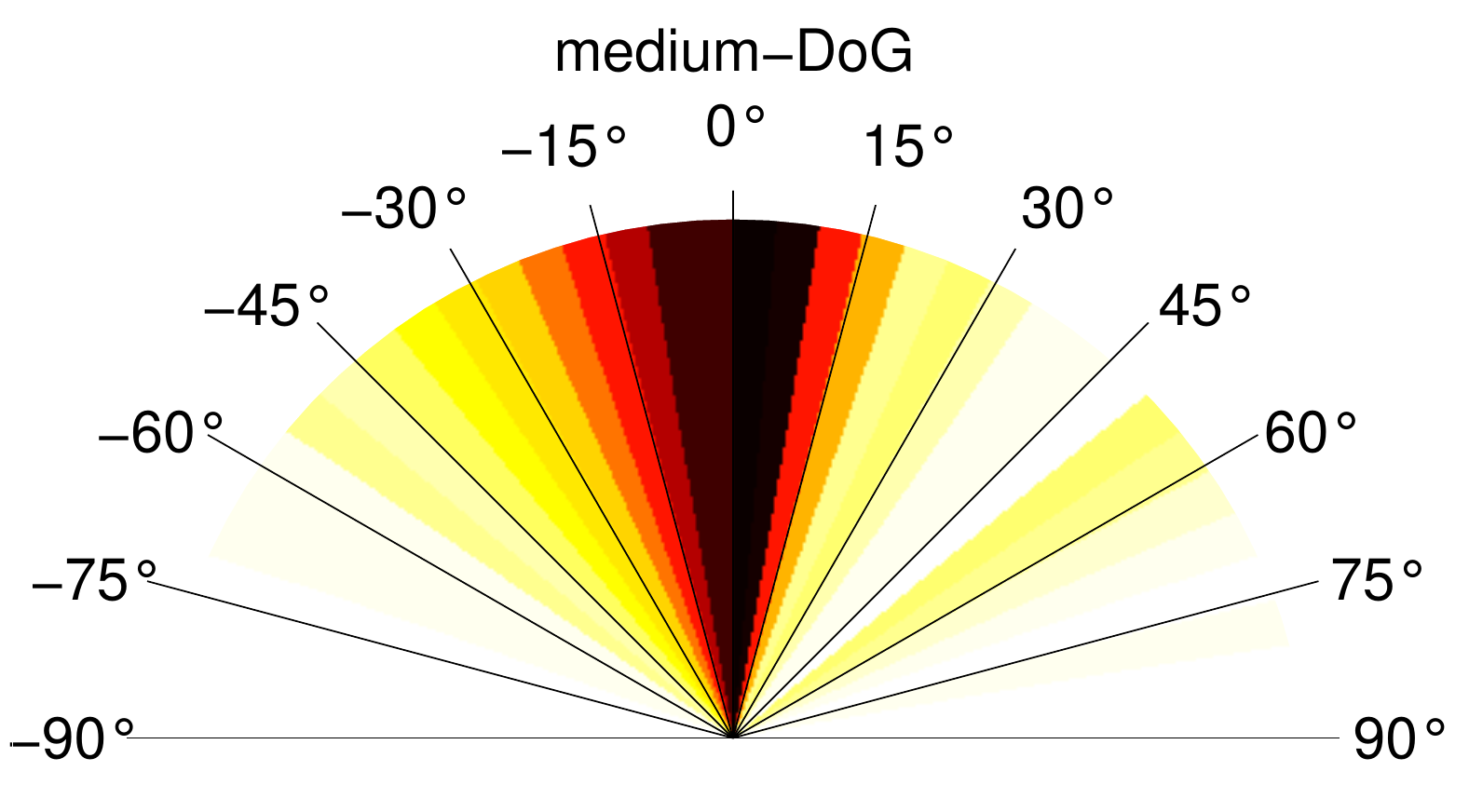}
\includegraphics[width=0.30\linewidth]{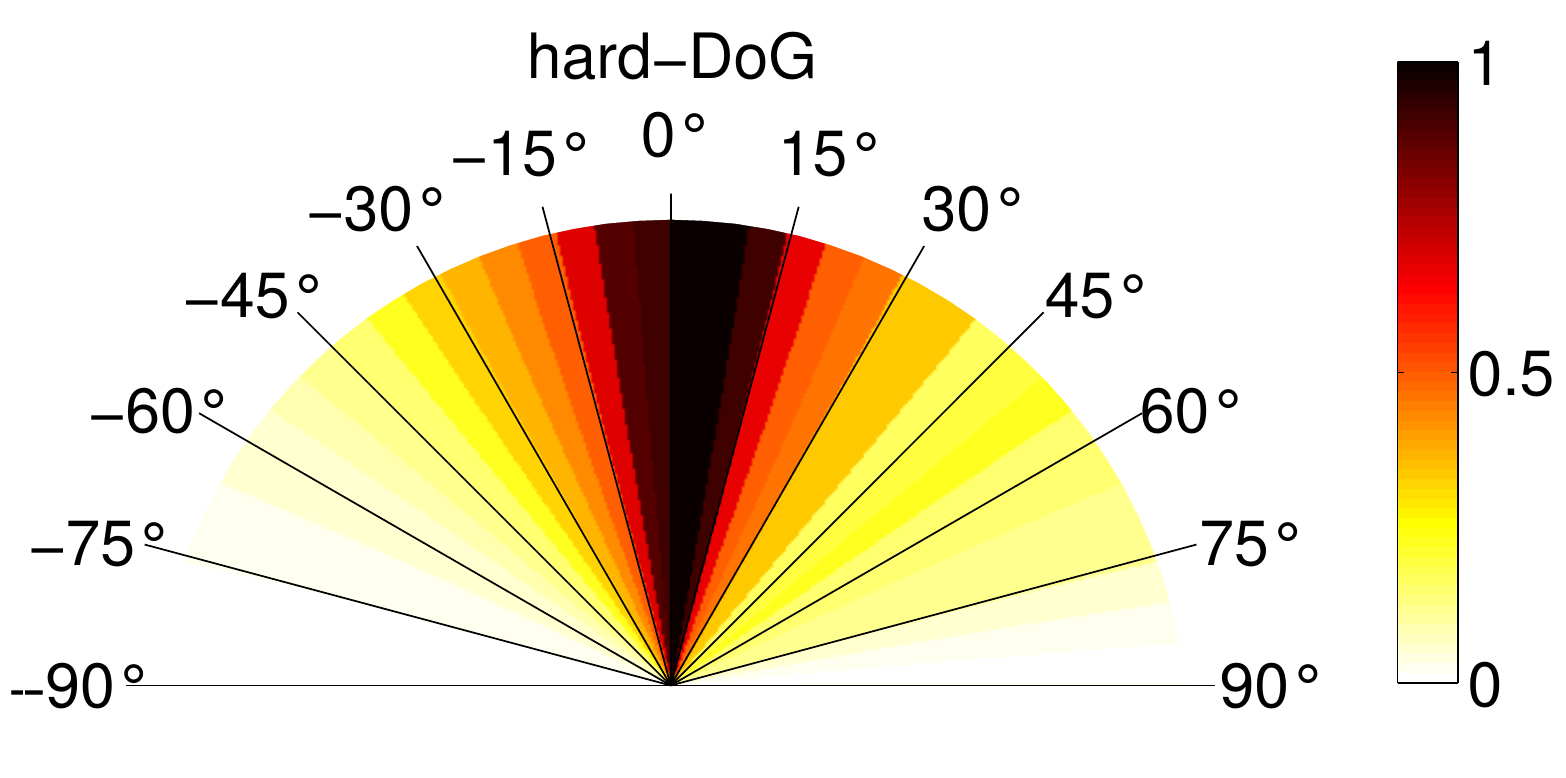}\\
\includegraphics[width=0.28\linewidth]{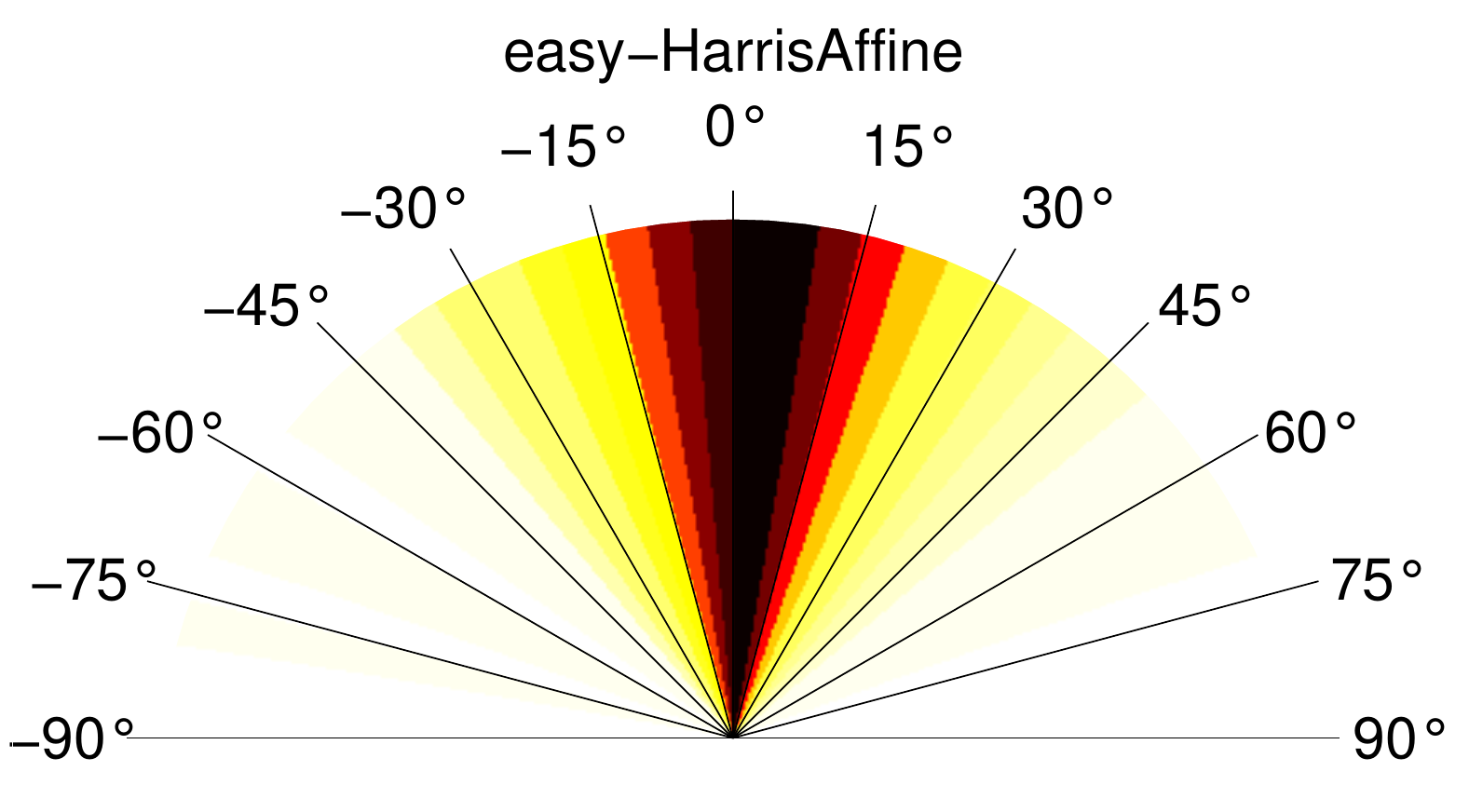}
\includegraphics[width=0.28\linewidth]{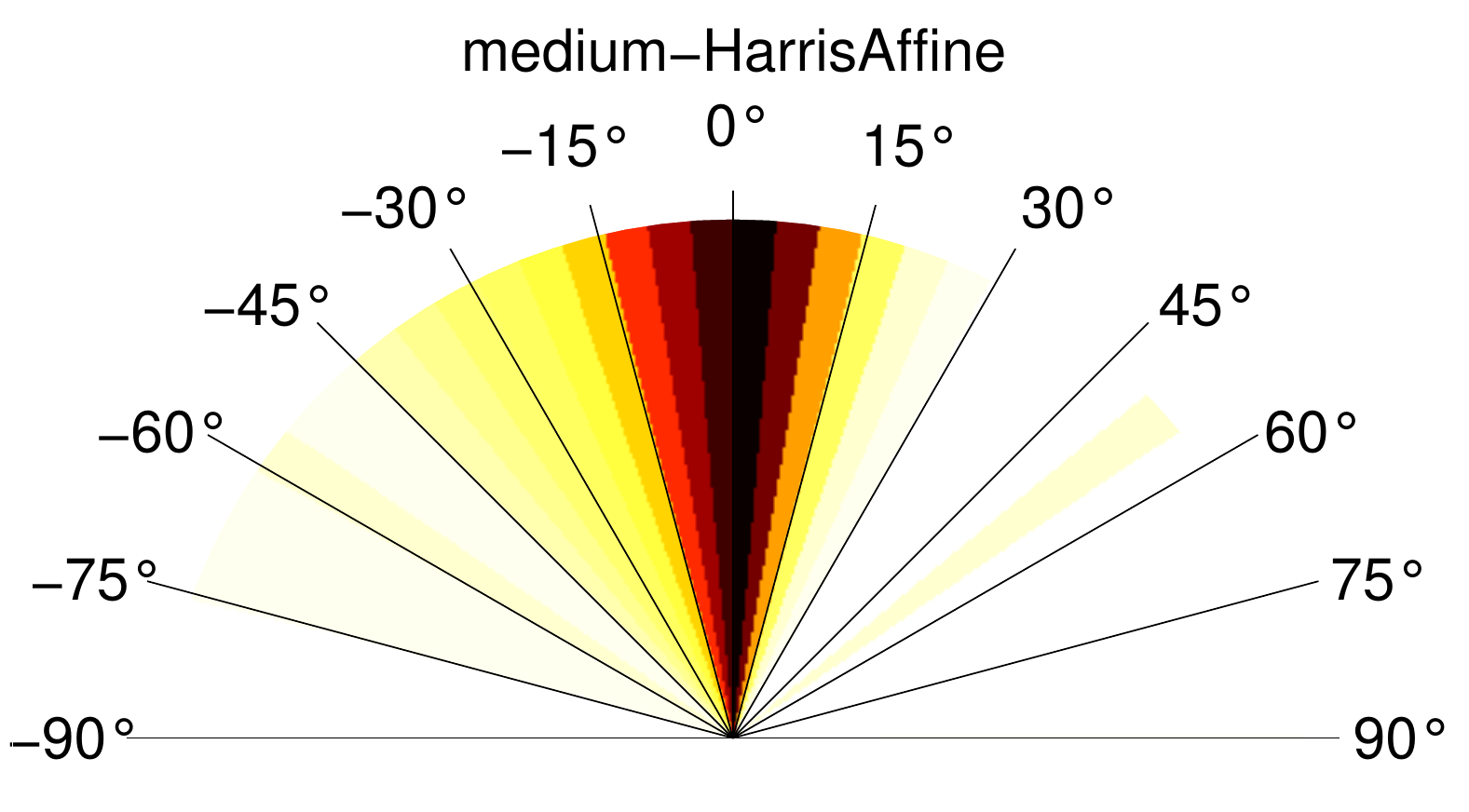}
\includegraphics[width=0.30\linewidth]{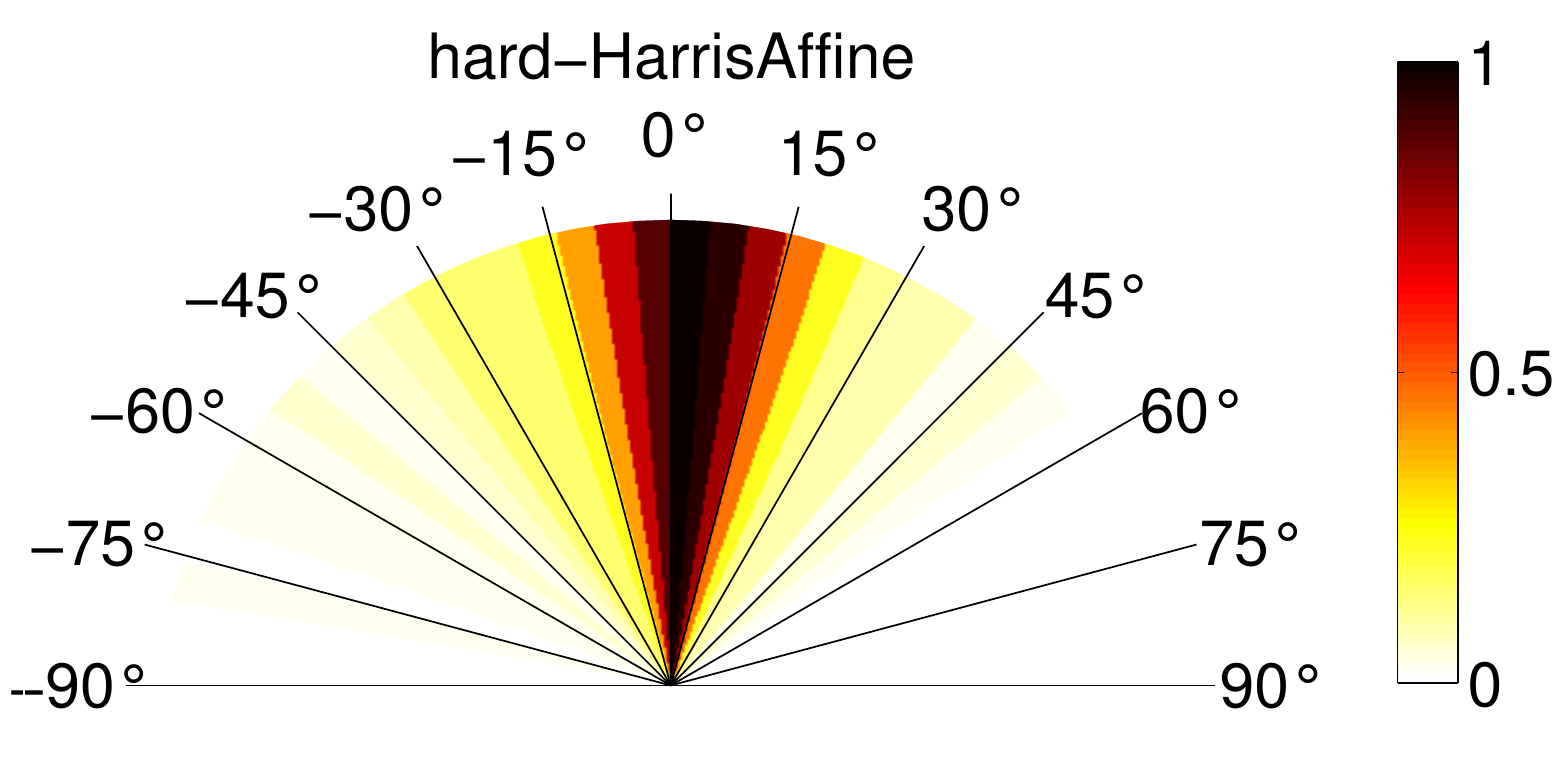}\\
\includegraphics[width=0.28\linewidth]{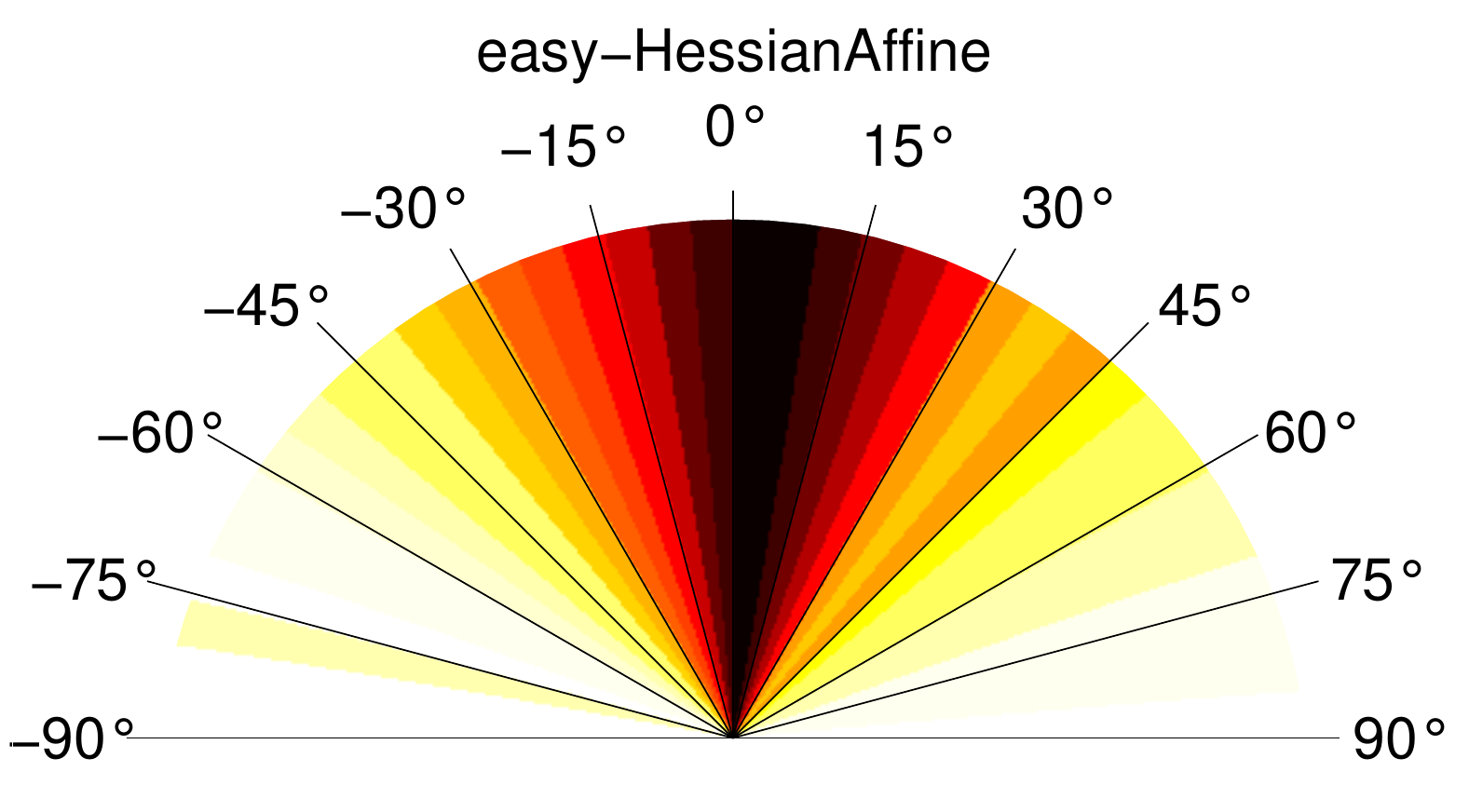}
\includegraphics[width=0.28\linewidth]{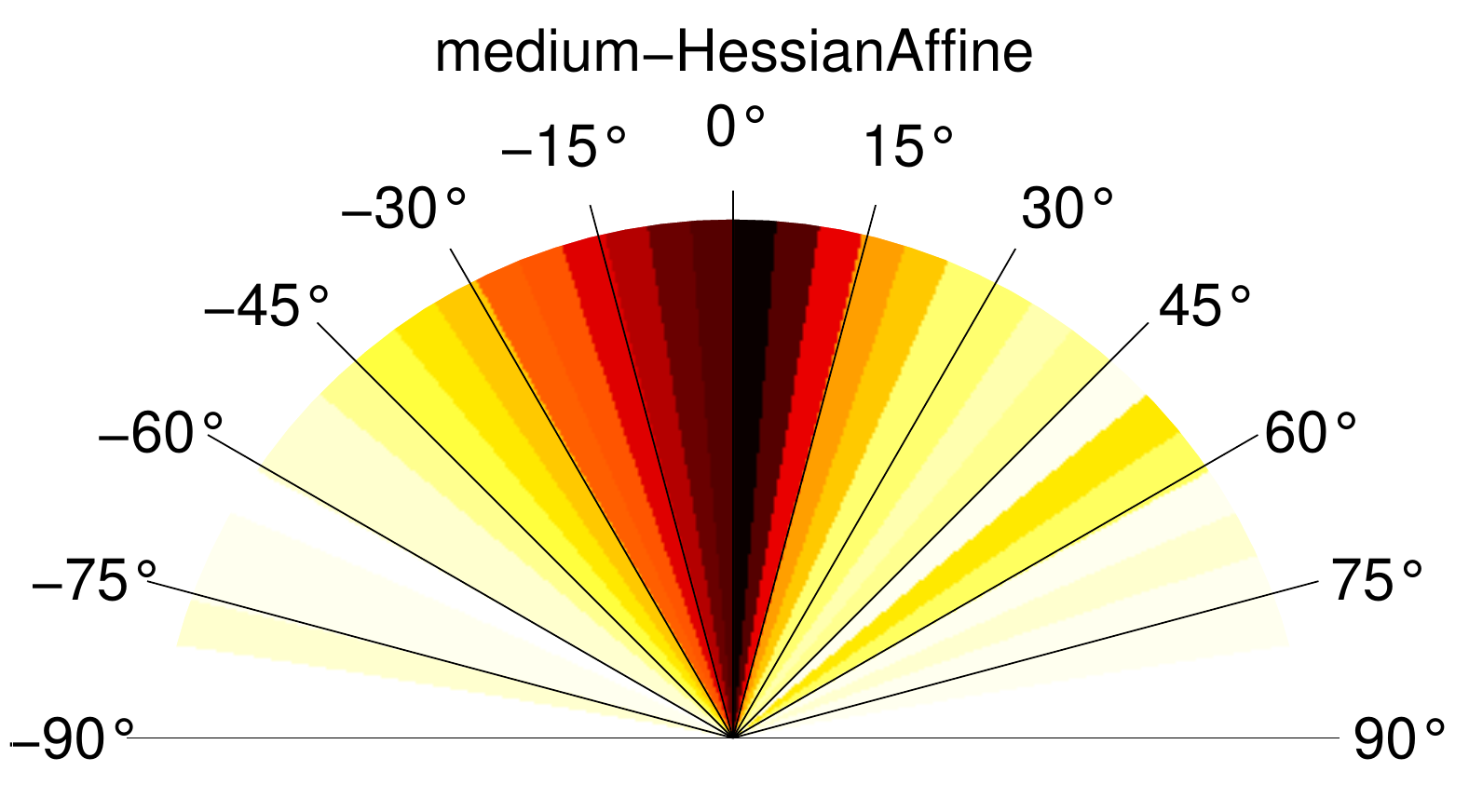}
\includegraphics[width=0.30\linewidth]{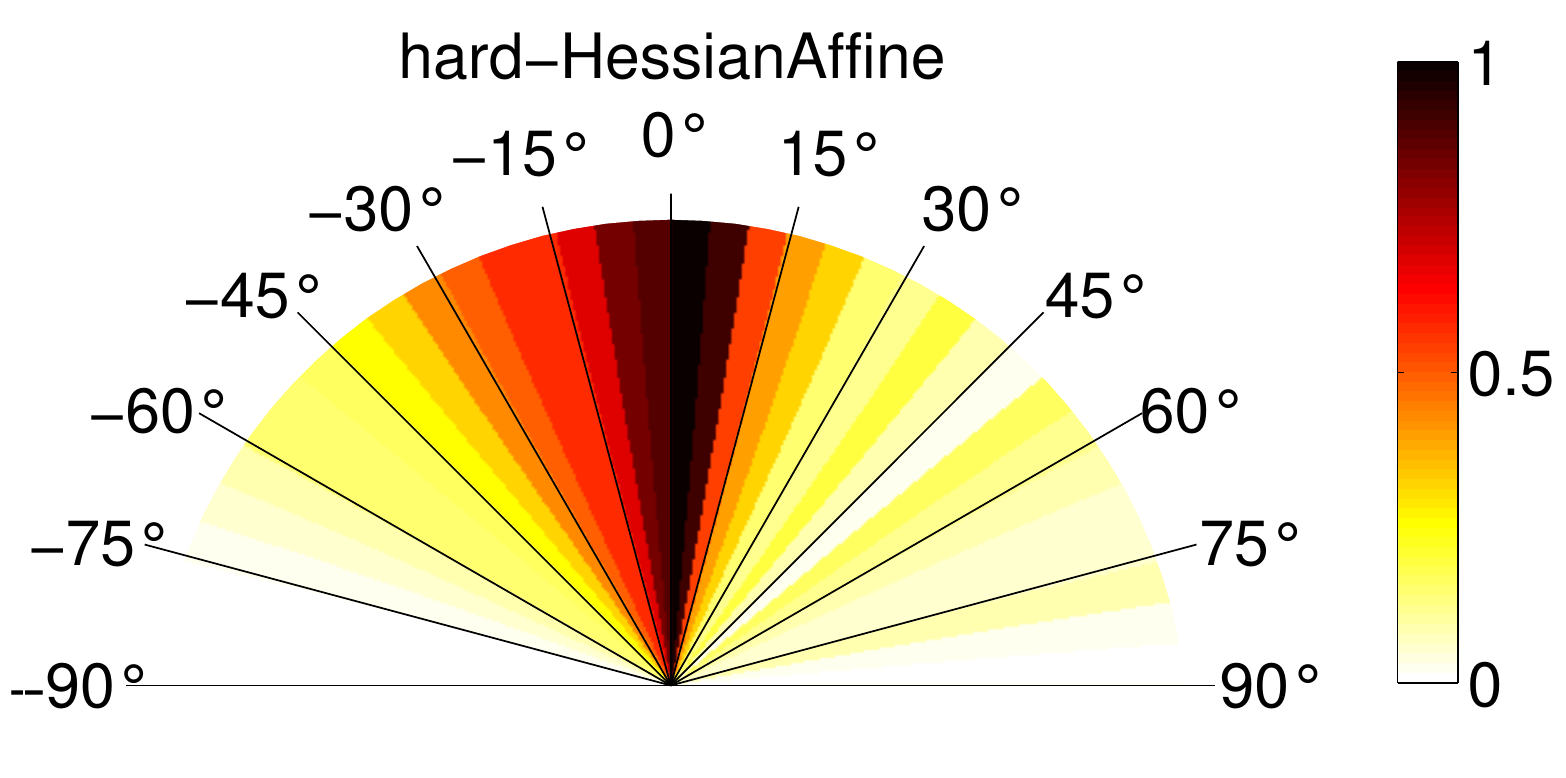}\\
\includegraphics[width=0.28\linewidth]{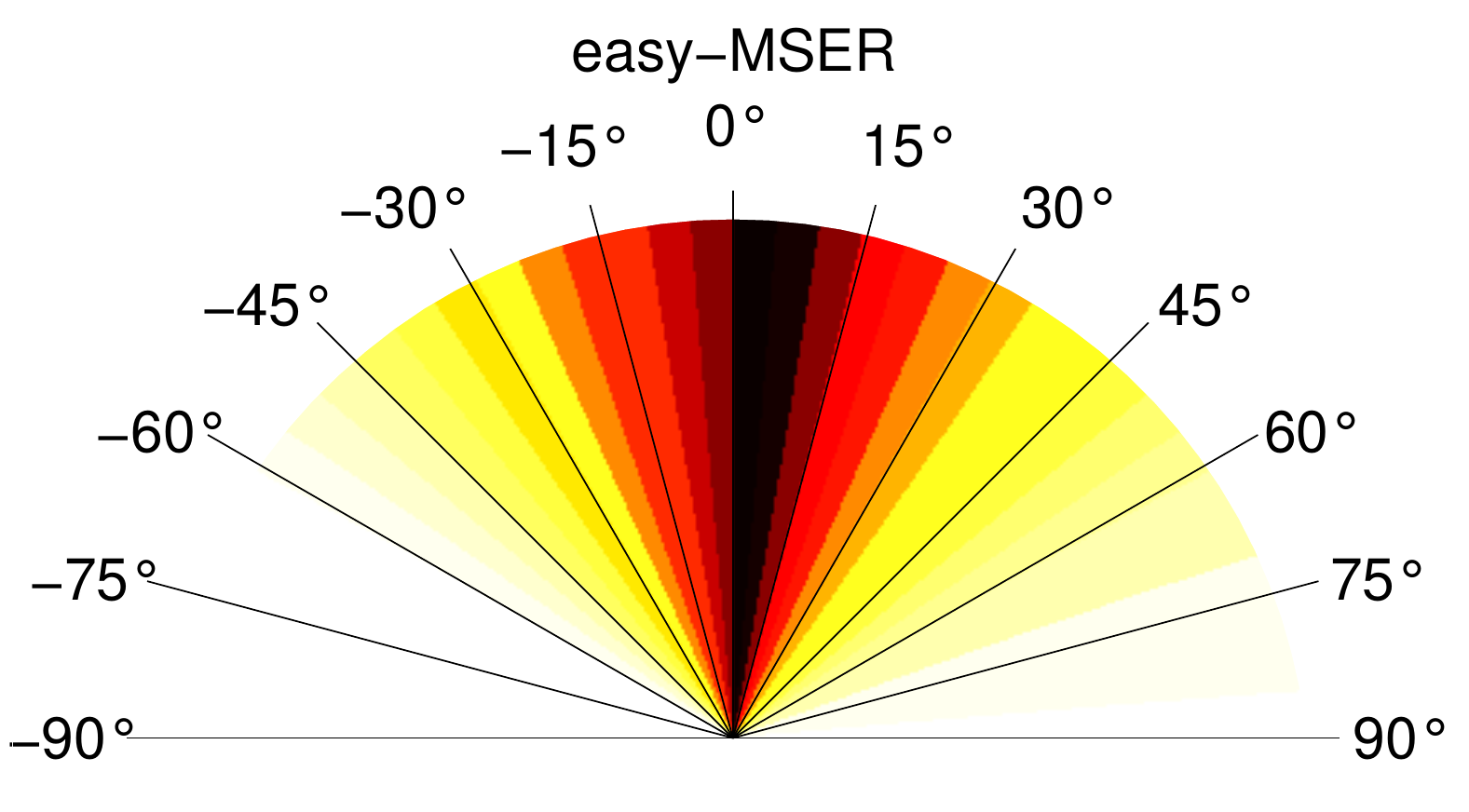}
\includegraphics[width=0.28\linewidth]{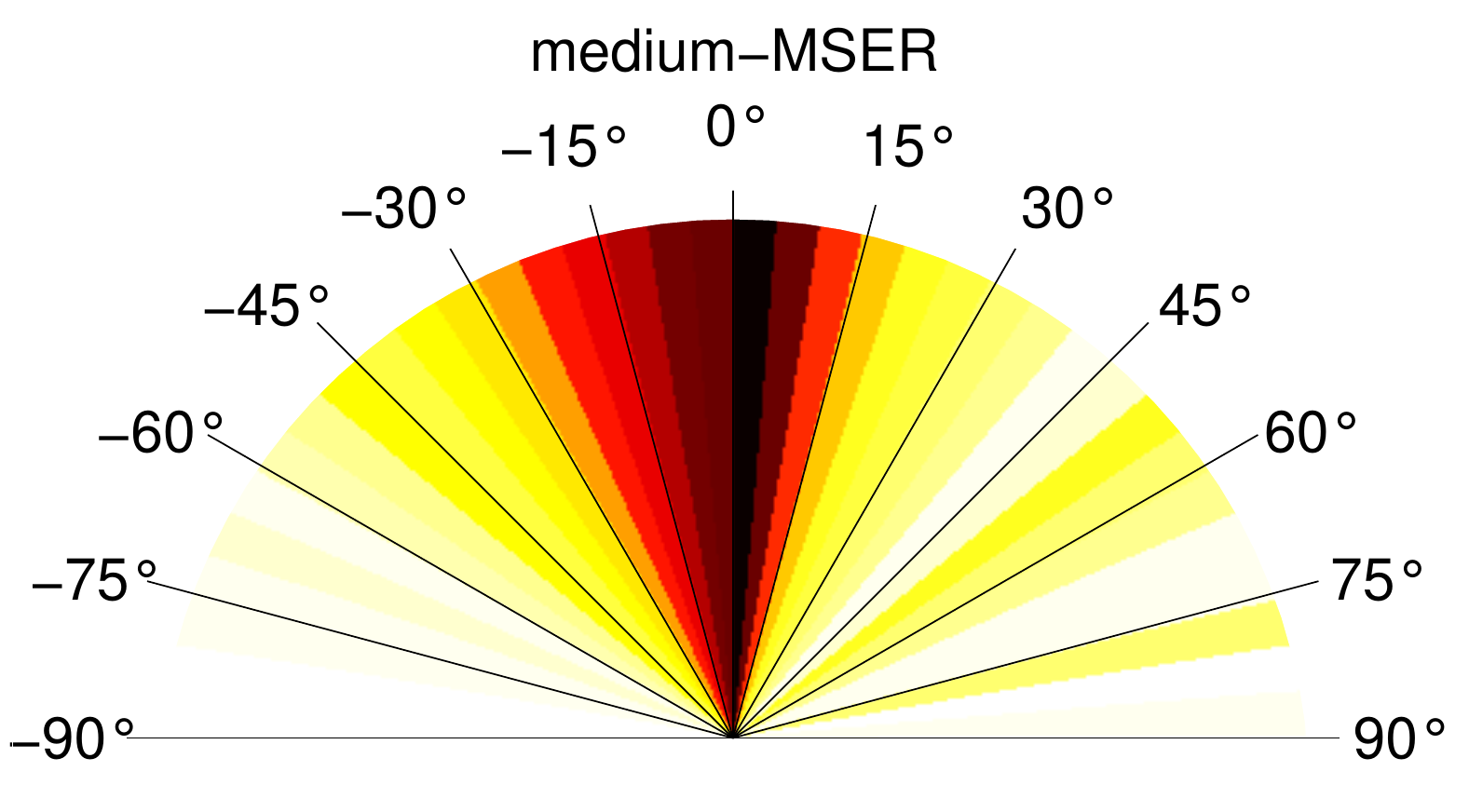}
\includegraphics[width=0.30\linewidth]{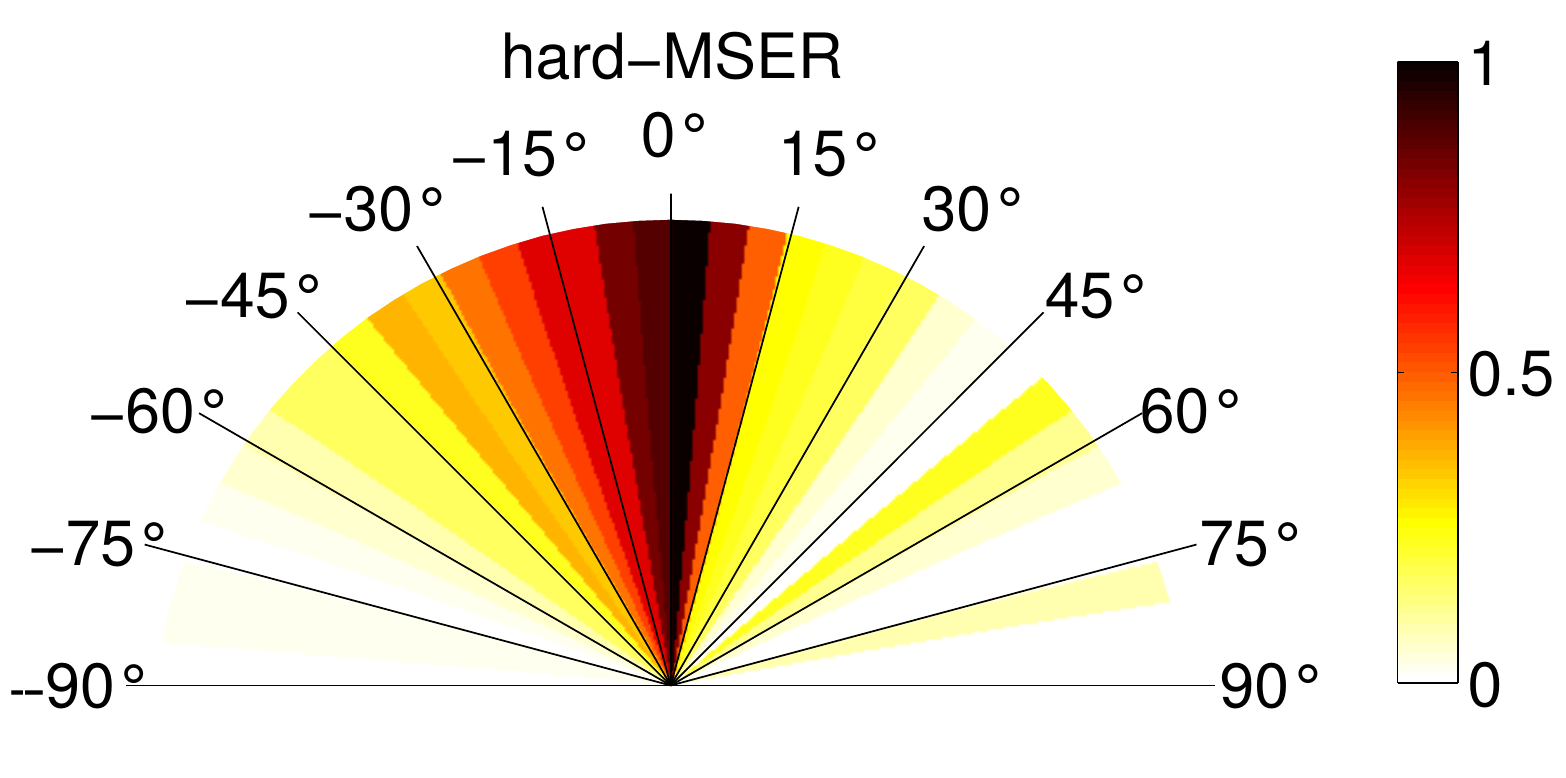}\\
\includegraphics[width=0.28\linewidth]{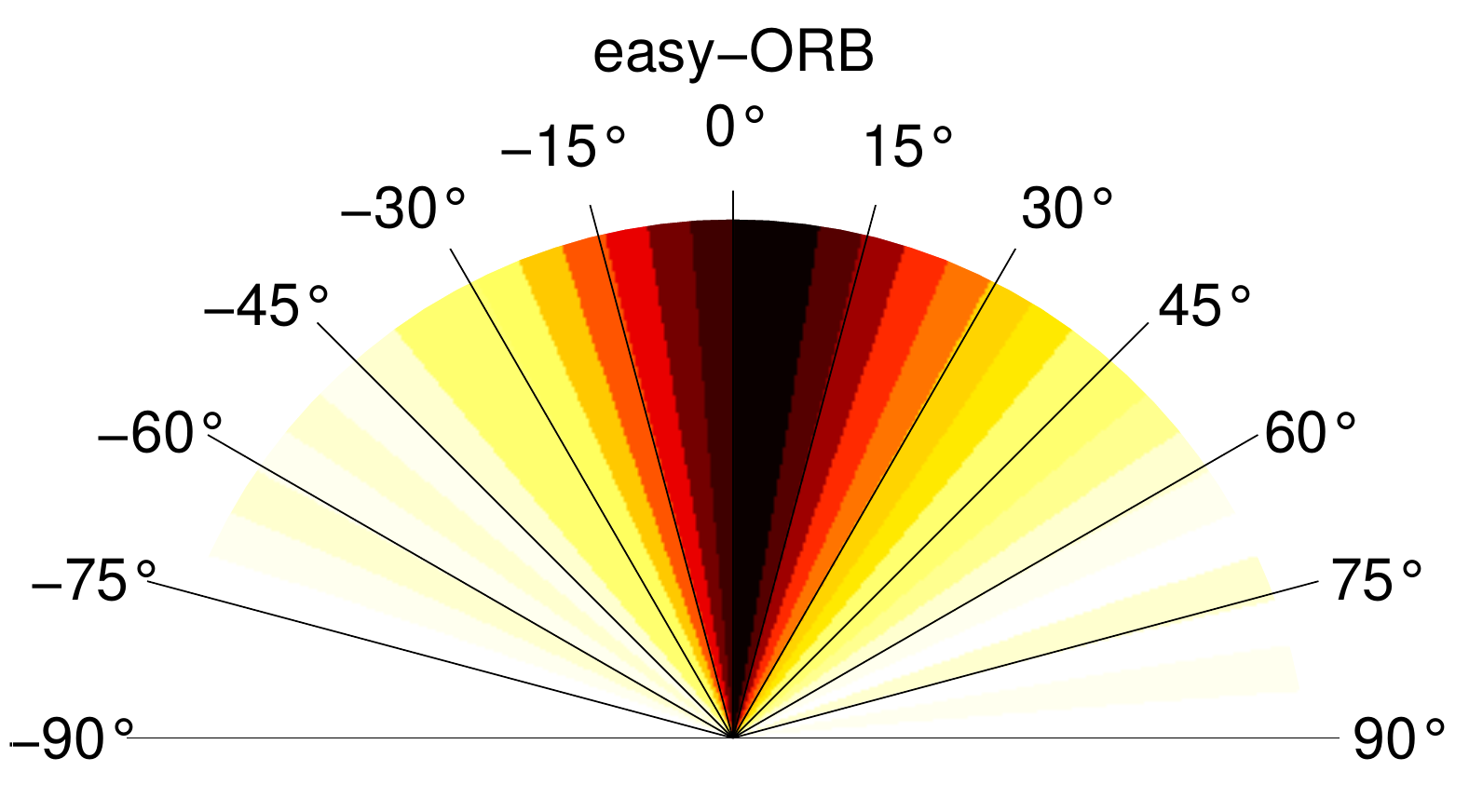}
\includegraphics[width=0.28\linewidth]{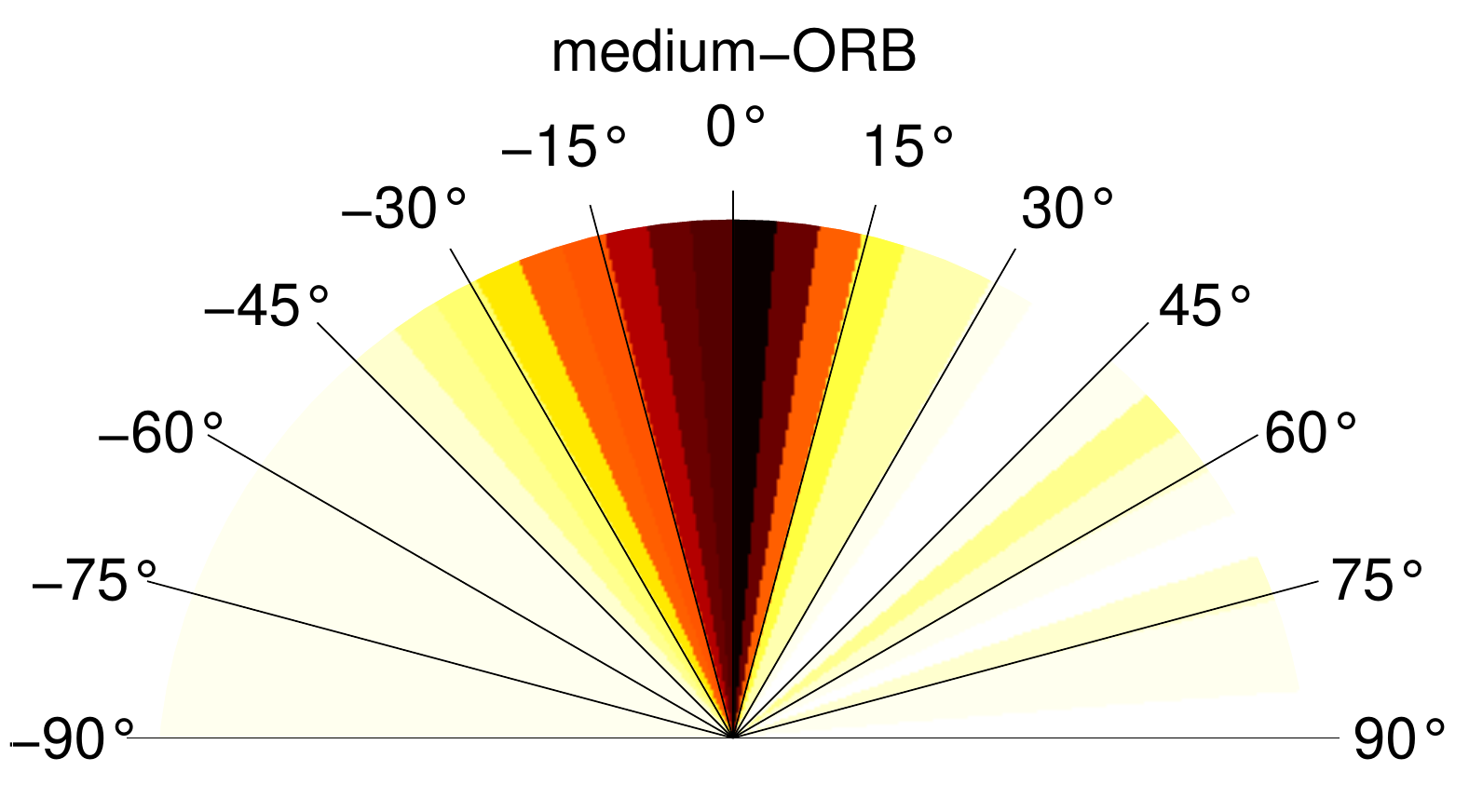}
\includegraphics[width=0.30\linewidth]{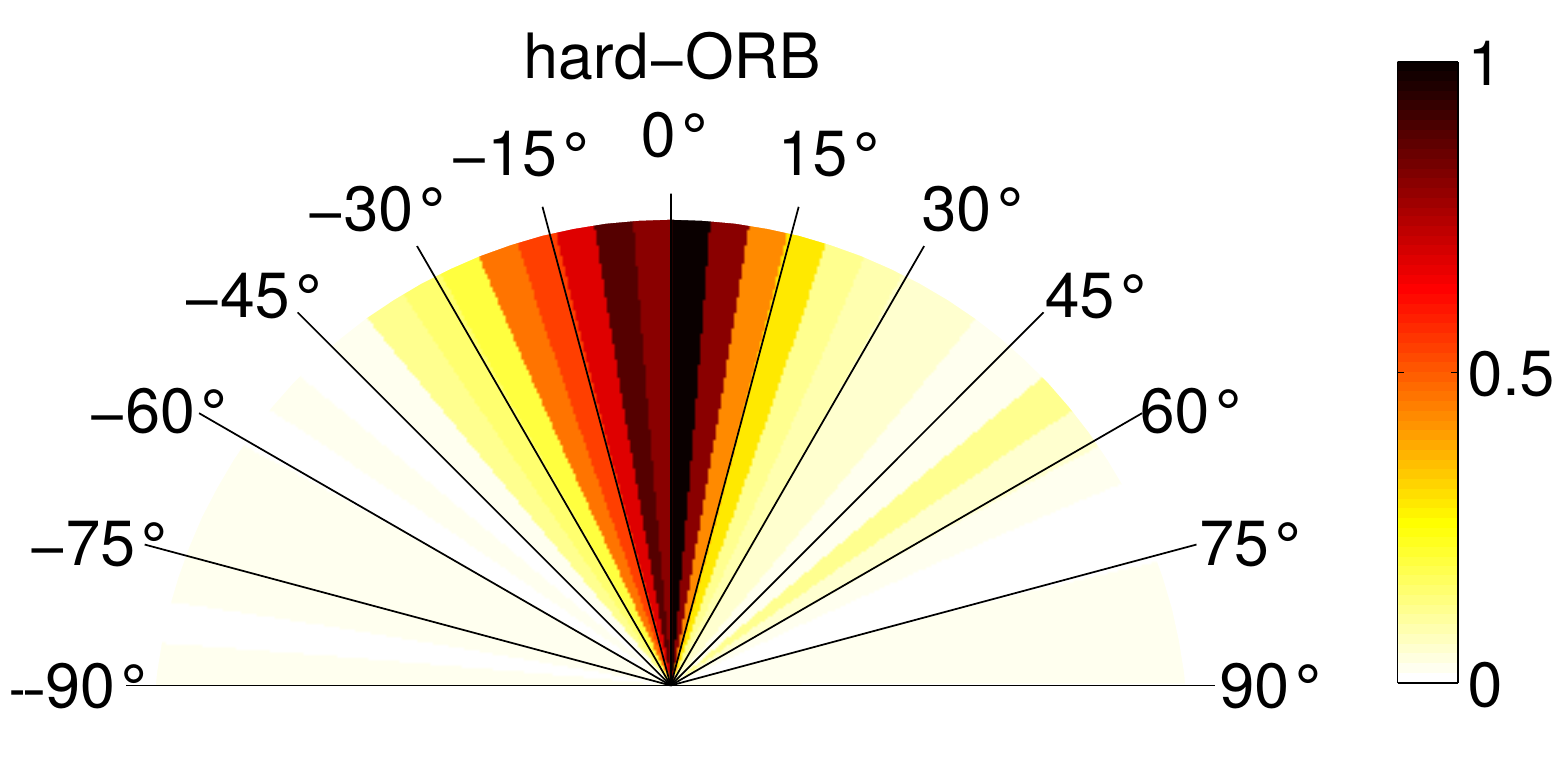}\\
\includegraphics[width=0.28\linewidth]{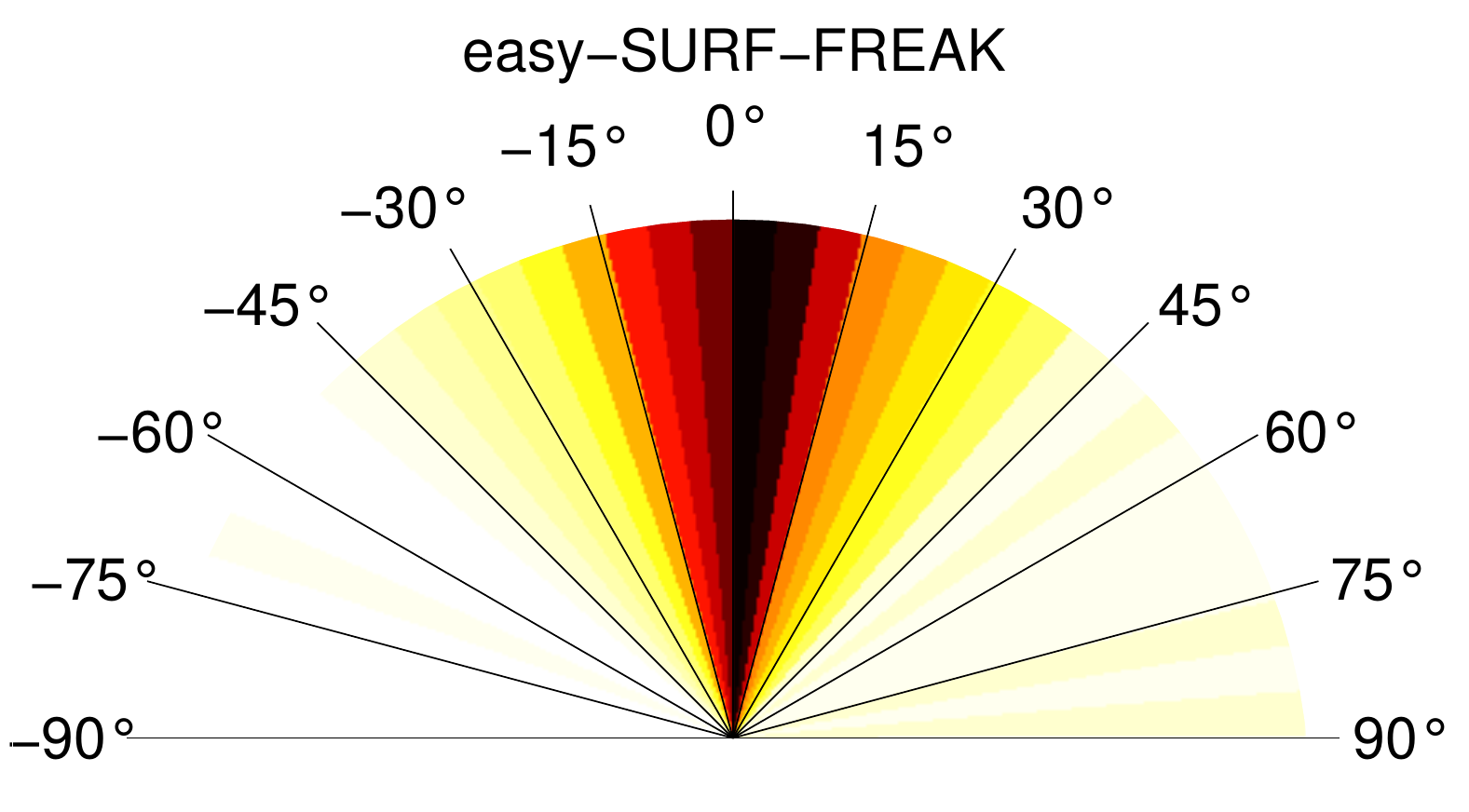}
\includegraphics[width=0.28\linewidth]{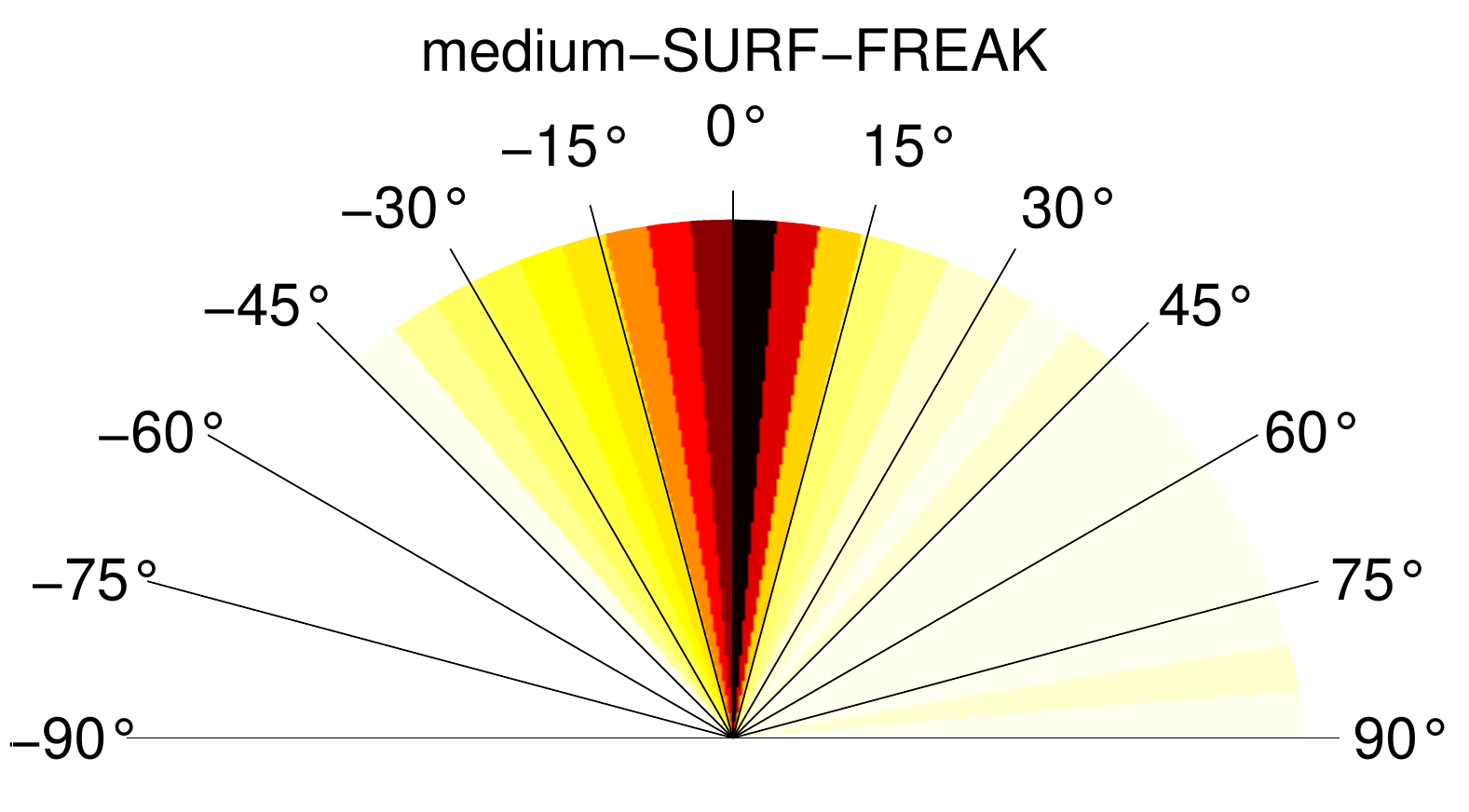}
\includegraphics[width=0.30\linewidth]{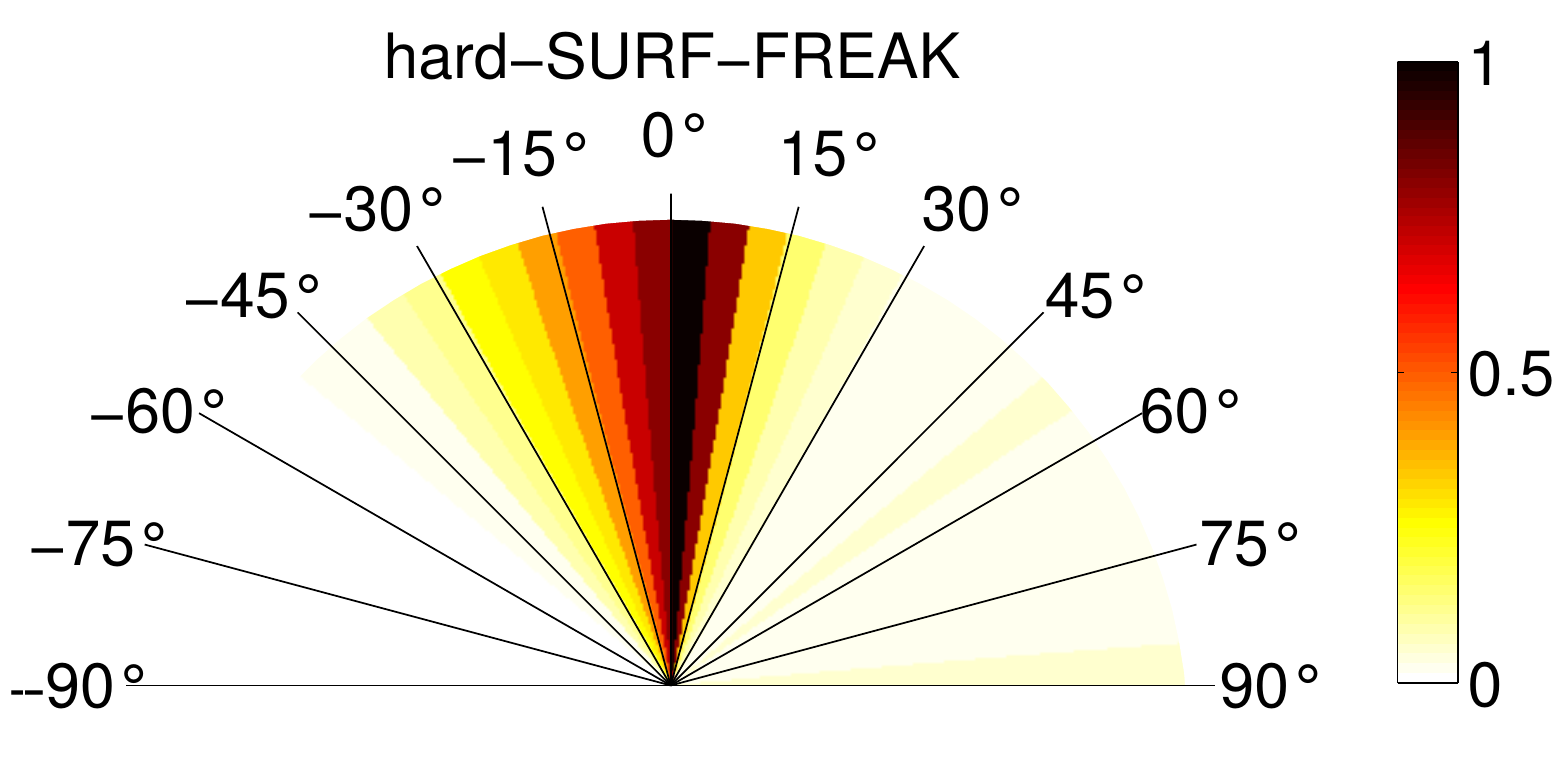}\\
\includegraphics[width=0.28\linewidth]{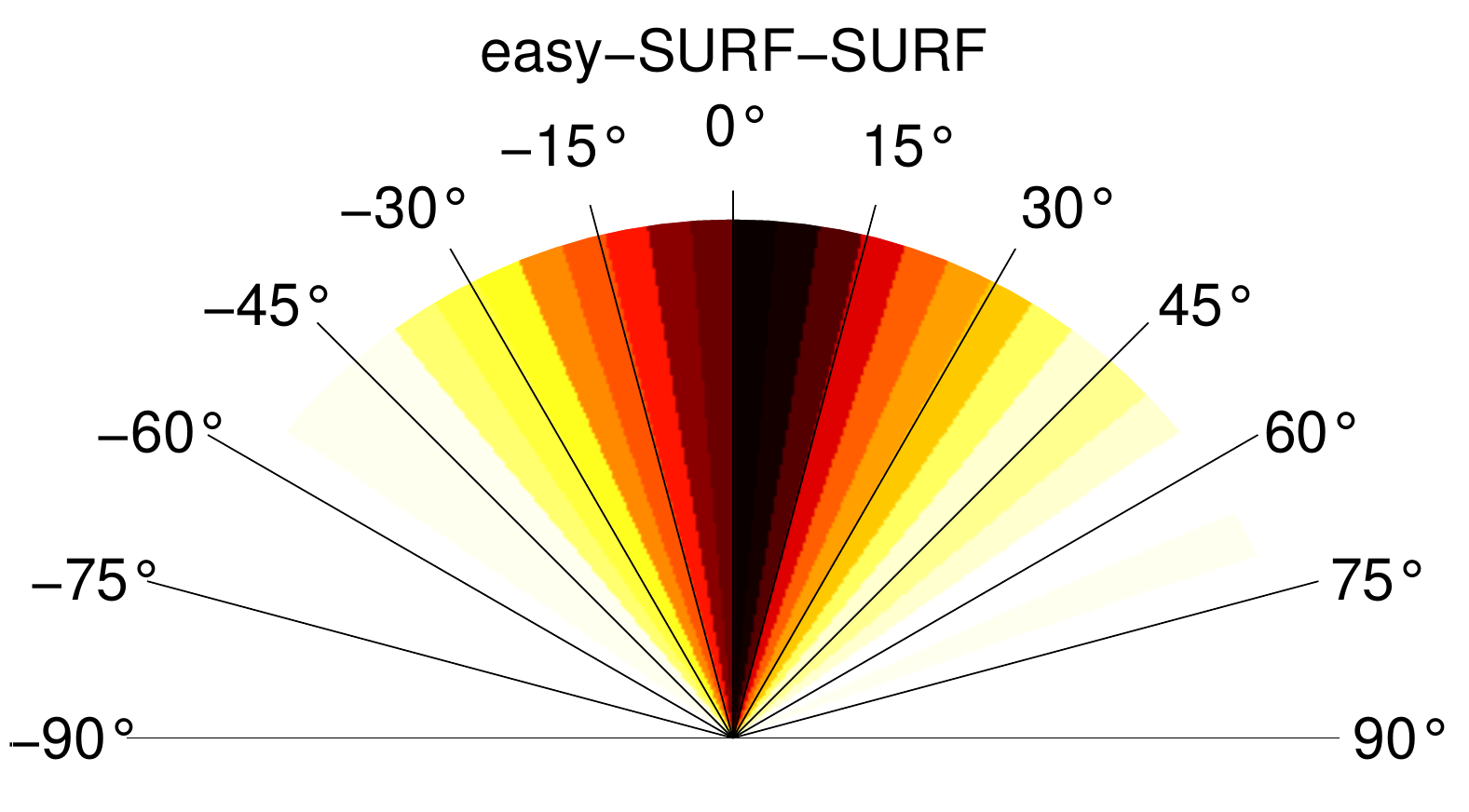}
\includegraphics[width=0.28\linewidth]{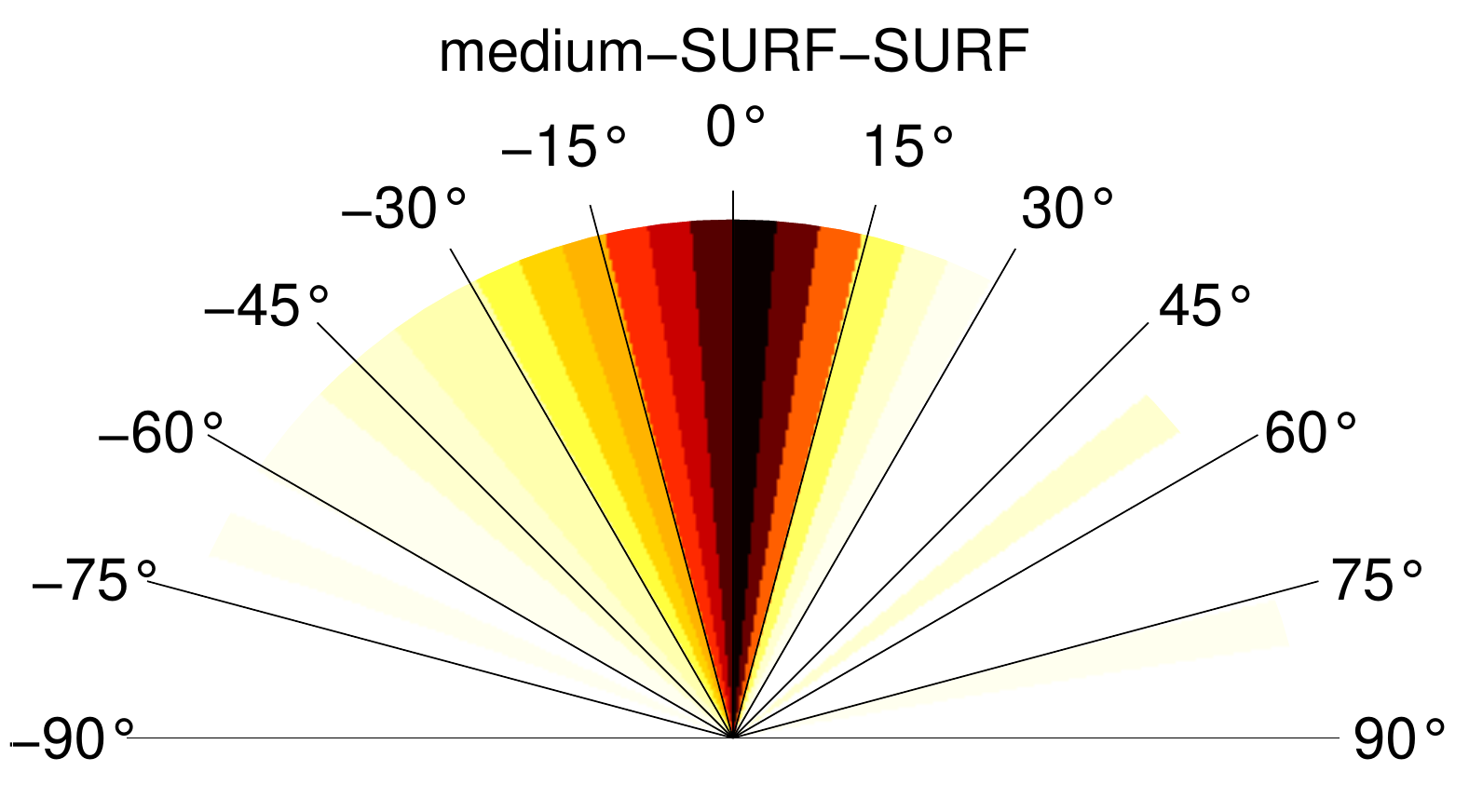}
\includegraphics[width=0.30\linewidth]{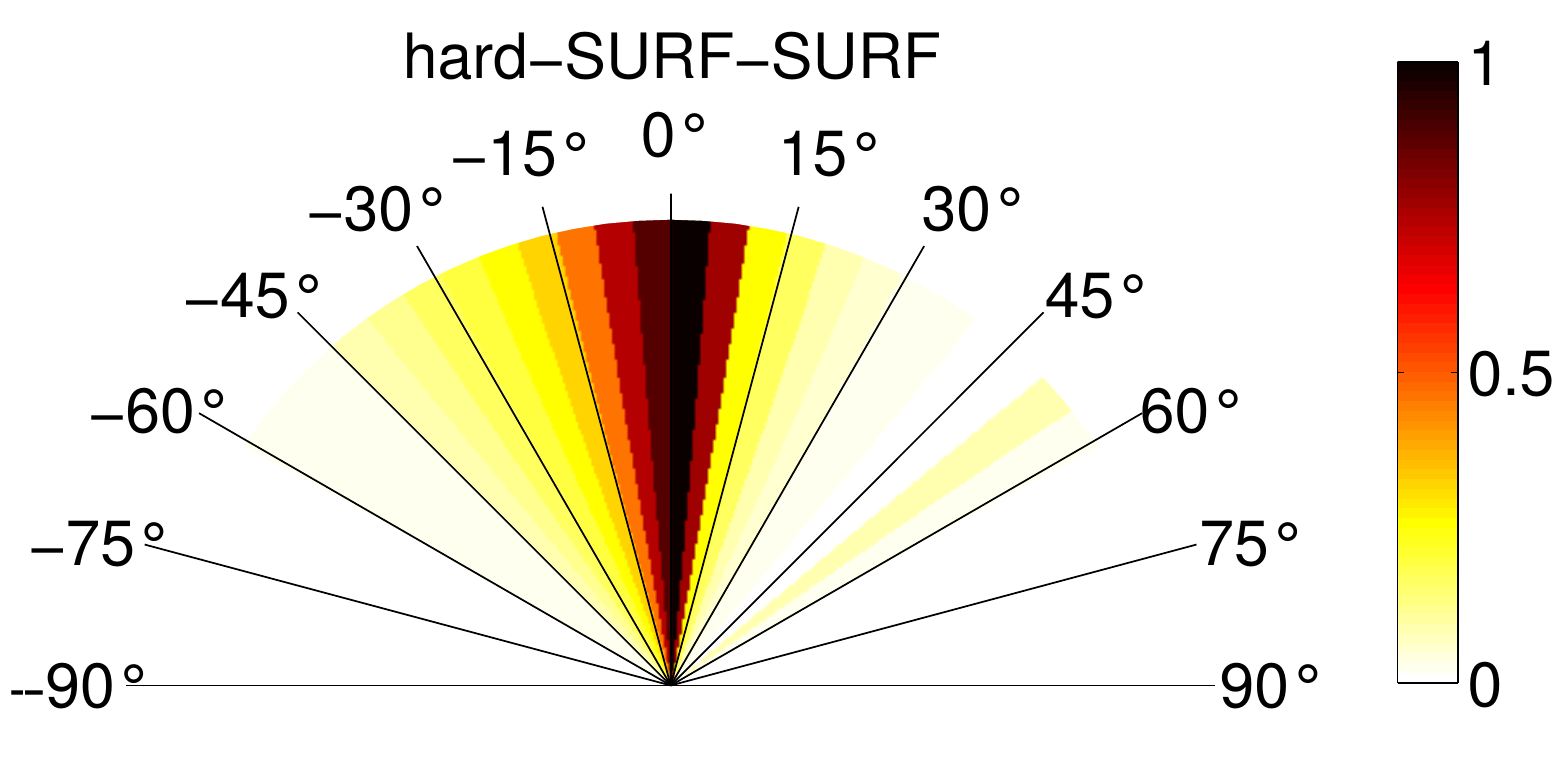}\\
\includegraphics[width=0.28\linewidth]{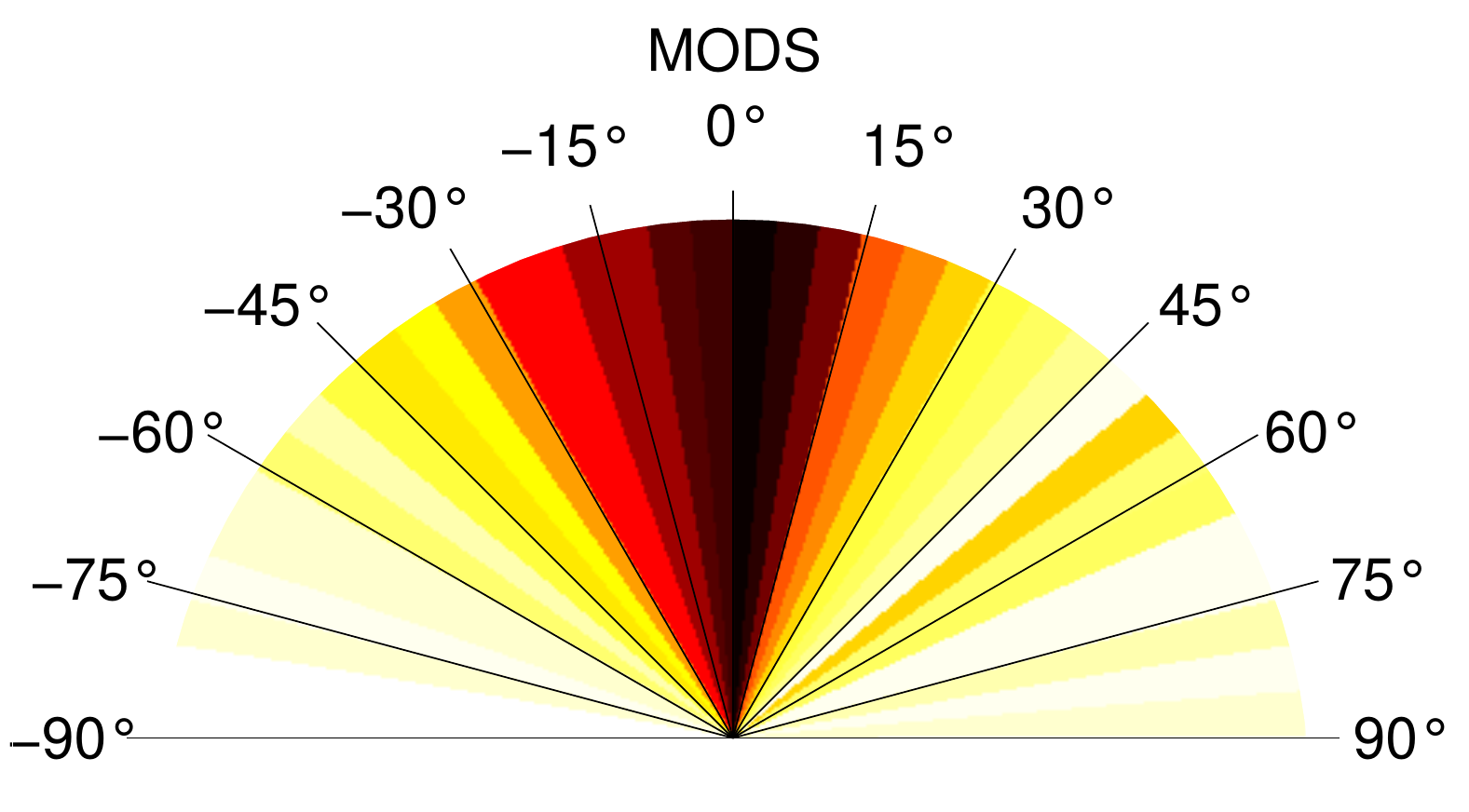}
\includegraphics[width=0.30\linewidth]{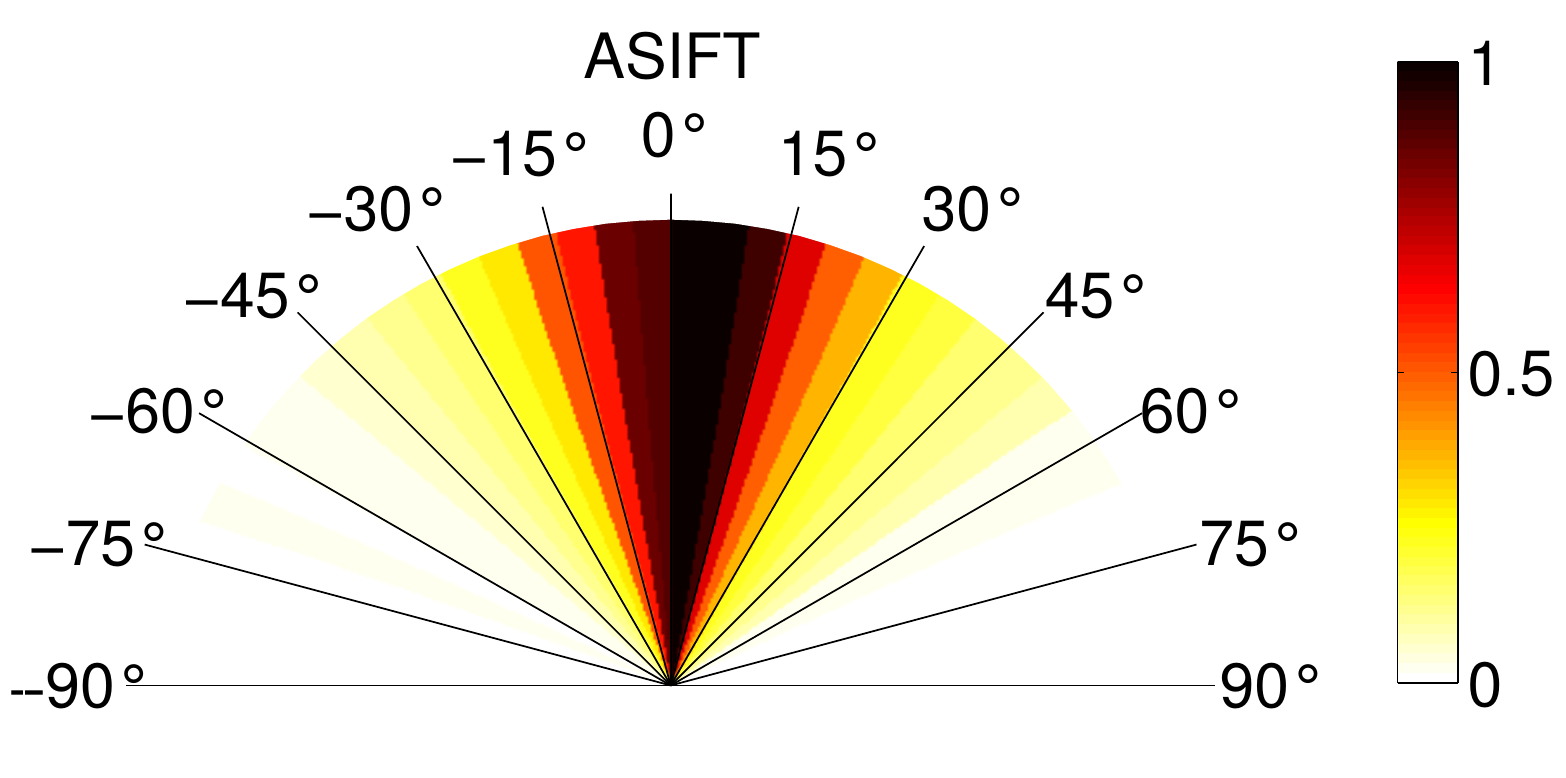}\\
\caption{A comparison of view synthesis configurations on the Turntable dataset~\cite{Moreels2007}. The fraction of correctly matched images for a given viewpoint difference.}
\label{fig:jet-angle}
\end{figure}

\begin{table}[tb]
\caption{Experiment on the 3D dataset. The configurations are defined in Table~\ref{tab:view-synth-configuration}. The number of correctly matched image pairs and the run-time per pair. Best results are \fbox{boxed}. Results within 90\% of the best are in \textbf{bold}. The configurations are sorted by average time for image pair.}
\label{tab:results}
\footnotesize
\centering
\setlength{\tabcolsep}{.35em}
\begin{tabular}{lrrrrrrrrrcr}
\toprule
&\multicolumn{10}{c}{Image sets solved (out of 35)} & \\
\cmidrule(r){2-12}
&\multicolumn{10}{c}{Viewpoint angular difference} & Time\\
Matcher& $0\degree$& $5\degree$& $10\degree$& $15\degree$& $20\degree$& $25\degree$& $30\degree$&$35\degree$ &$40\degree$ &$\geq45\degree$&  [s] \\
\cmidrule(r){1-12}
HessianAffine easy  &\textbf{35}& \textbf{32}&\textbf{29}&\textbf{26}&\fbox{22}  &17&13&\fbox{14}&  \textbf{9}     &6&0.9\\
SURF-SURF easy      &\textbf{35}& \textbf{31}&24&18&15&12&7&5&4&2&0.9\\
HessianAffine medium&\textbf{35}& \textbf{30}&26&24&18&17&12&10&7&\textbf{10} &1.0\\
ORB easy            &\textbf{35}& \textbf{31}&\textbf{27}&20&16&11&10&5&5&4&1.0\\
AGAST-FREAK easy    &\textbf{35}&28&\textbf{29}&\textbf{26}&19&17&12&9&7&6&1.3\\
DoG easy            &\textbf{35}& \fbox{33}  &\fbox{30}  &22&18&12&10&7&5&3&1.4\\
MSER easy           &\textbf{35}&28&22&21&15&13&10&8&7&5&1.5\\
SURF-SURF hard      &\textbf{35}&26&16&11&9&7&6&4&3&3&2.1\\
HarrisAffine easy   &\textbf{35}&29&22&12&8&6&5&3&2&1&2.7\\
AGAST-FREAK hard    &\textbf{35}& \textbf{31}&\textbf{28}&\textbf{25}&\textbf{20}&17&   \fbox{15}&10&8&5&4.7\\
HarrisAffine medium &\textbf{35}&27&20&11&7&6&5&4&3&2&4.7\\
HessianAffine hard  &\textbf{35}&29&24&20&\textbf{20}&17&   \fbox{15}&11&  \textbf{9}     &6&5.6\\
DoG medium          &\textbf{35}& \textbf{32}&26&21&16&11&10&9&6&5&6.8\\
AGAST-FREAK medium  &\textbf{35}&29&23&22&18&14&13&6&6&6&7.2\\
ORB hard            &\textbf{35}& \textbf{31}&24&19&16&7&5&4&1&4&7.3\\
SURF-FREAK easy     &\textbf{35}&25&21&13&10&8&6&3&2&2&7.3\\
ORB medium          &\textbf{35}& \textbf{30}&26&18&17&10&5&4&2&4&8.3\\
SURF-FREAK medium   &\textbf{35}&22&15&10&9&7&6&4&1&2&9.1\\
SURF-SURF medium    &\textbf{35}&25&20&13&11&7&3&3&2&2&10.7\\
SURF-FREAK hard     &\textbf{35}&25&17&14&10&9&4&3&1&2&10.9\\
MSER hard           &\textbf{35}&29&24&24&19&16&12&\textbf{13}&8&8&14.5\\
HarrisAffine hard   &\textbf{35}&27&16&8&5&5&5&3&2&2&15.2\\
DoG hard            &\textbf{35}& \textbf{32}&24&17&16&13&12&8&7&8&16.7\\
MSER medium         &\textbf{35}&29&26&23&\textbf{21}&14&10&9&7&9&17.5\\
\cmidrule(r){1-12}
ASIFT               &\textbf{35}& \textbf{32}&        24 &      18   &     13     &8  &    7      &5&     4     &3 & 27.6\\ 
\cmidrule(r){1-12}
MODS:1-7                &\textbf{35}& \textbf{31}&\textbf{27}&\fbox{27} &\fbox{22} &\fbox{22} &\textbf{14}& 9&\fbox{10}&\fbox{11} & 6.7\\
\bottomrule
\end{tabular}
\end{table}
Figure~\ref{fig:jet-angle} and Table~\ref{tab:results} show the percentage and the number of image sequences respectively for which the reference and tested views for the given viewing angle difference were matched correctly. 

The difference between {\sc easy}, {\sc medium} and {\sc hard} configurations is small for structured scenes --- unlike planar ones. Difficulties in matching are caused not by the inability to detect distorted regions but by object self-occlusions. Therefore the synthesis of additional views does not bring more correspondences.

Experiments with view synthesis confirmed results~\cite{Moreels2007} that the Hessian-Affine outperforms other detectors for matching of structured scenes and can be used alone in such scenes. MODS shows similar performance, but is slower than the Hessian-Affine configuration.

\begin{figure}[tbp] %
\centering
\includegraphics[width=0.32\linewidth]{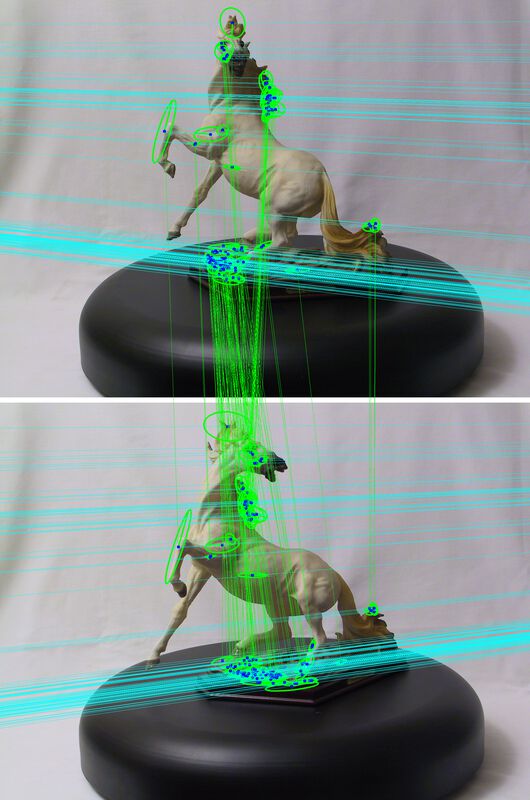}
\includegraphics[width=0.32\linewidth]{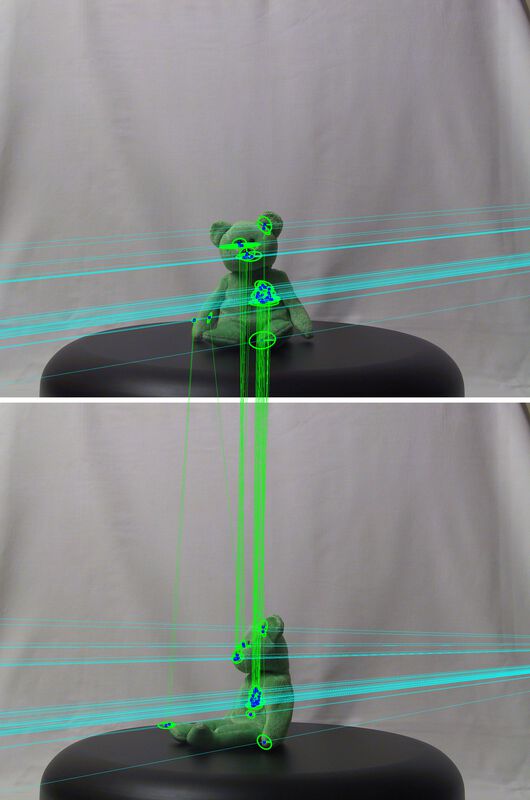}
\includegraphics[width=0.32\linewidth]{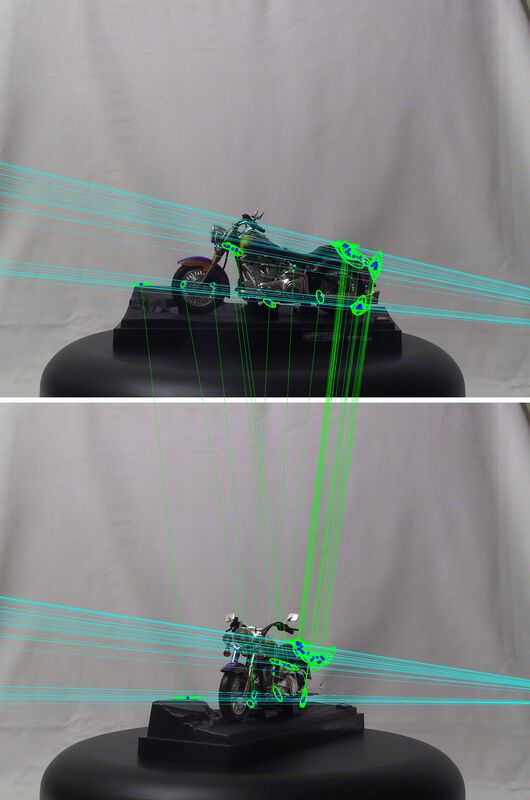}\\
\vspace{0.5ex}
\includegraphics[width=0.32\linewidth]{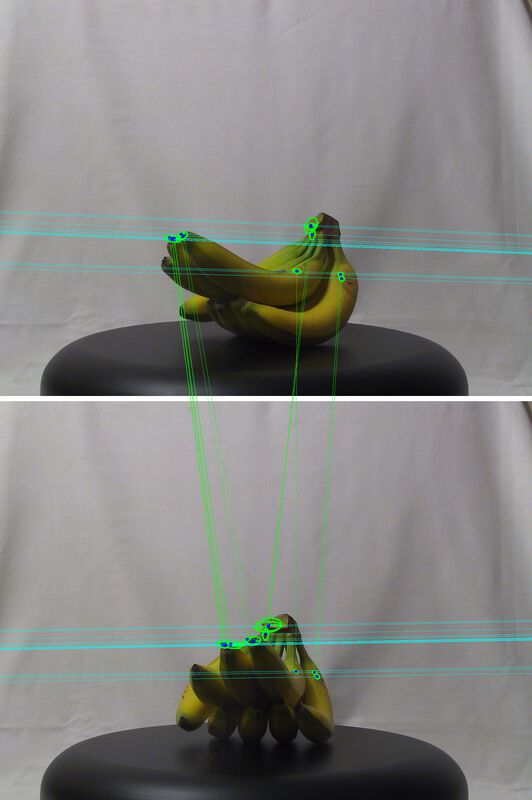}
\includegraphics[width=0.32\linewidth]{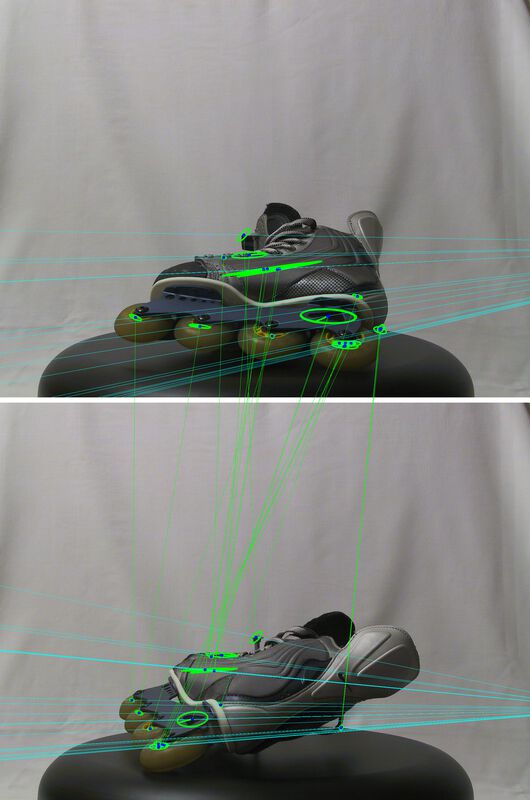}
\includegraphics[width=0.32\linewidth]{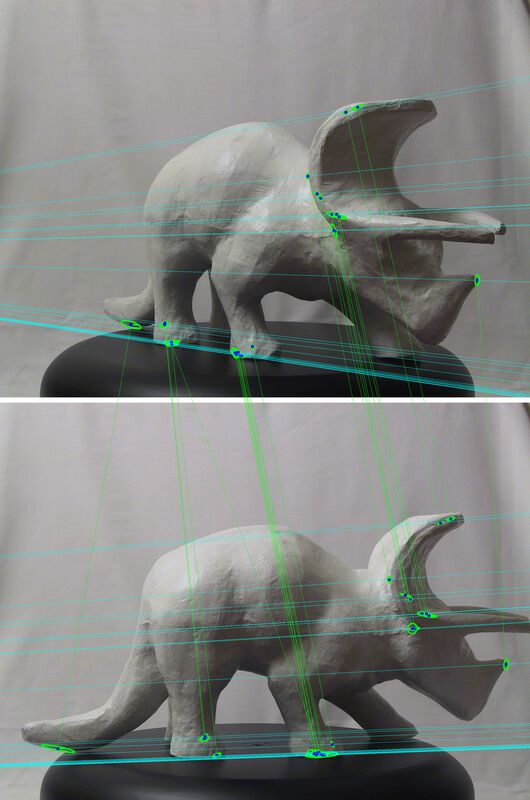}\\
\caption{Correspondences found by MODS. Green -- corresponding regions, cyan -- epipolar lines. Objects with significant self-occlusion and mostly homogenious texture were selected.}
\label{fig:mods-matches}
\end{figure}

The computations were performed on Intel i5 3.0GHz (4 cores) desktop with 16Gb RAM. Examples of the matched images are shown in Fig.~\ref{fig:mods-matches}.

\section{Why does affine synthesis help?}
\label{why-does-affine-synthesis-help}
\begin{figure}[tb]
\centering
\includegraphics[width=0.49\linewidth]{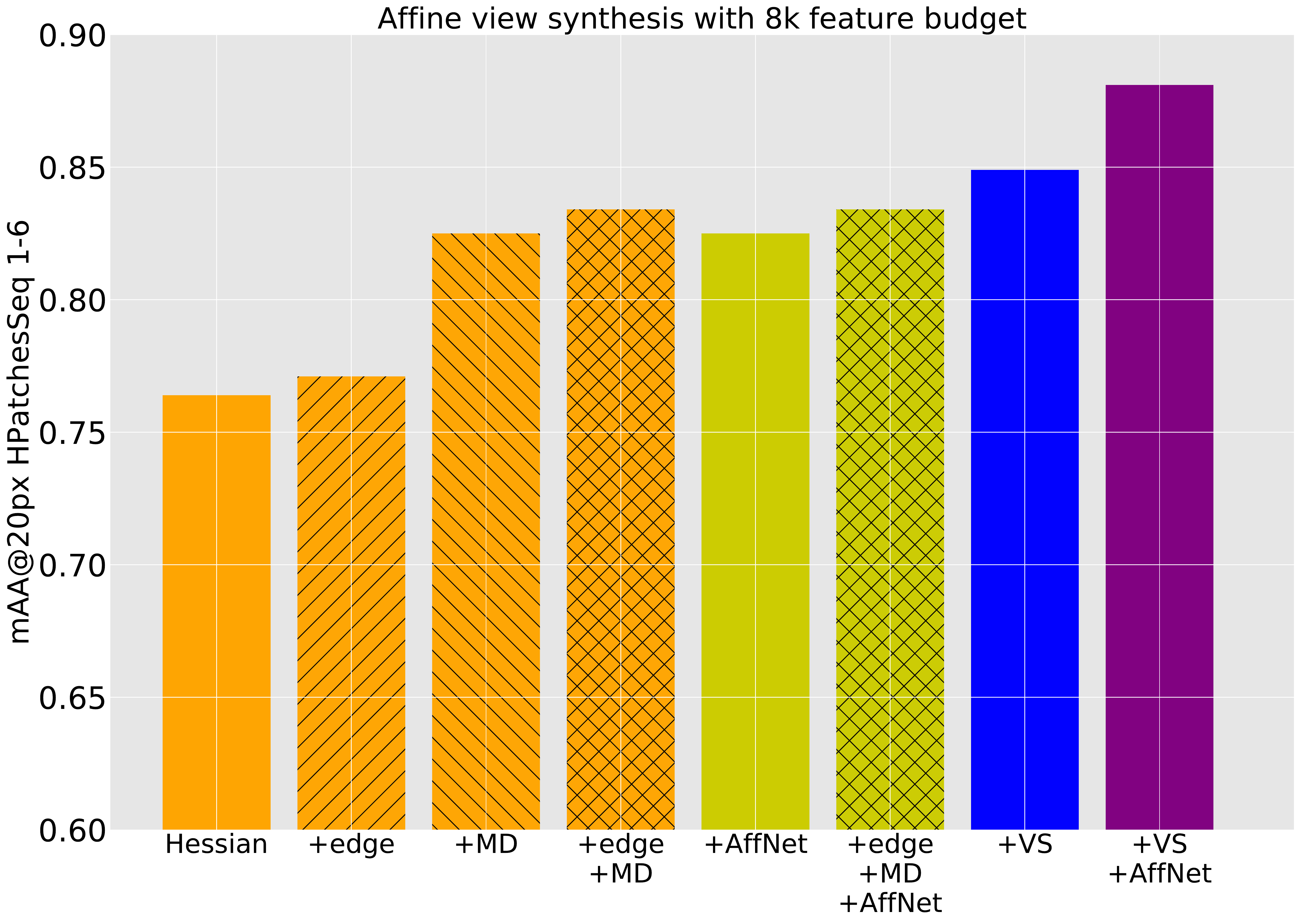}
\includegraphics[width=0.49\linewidth]{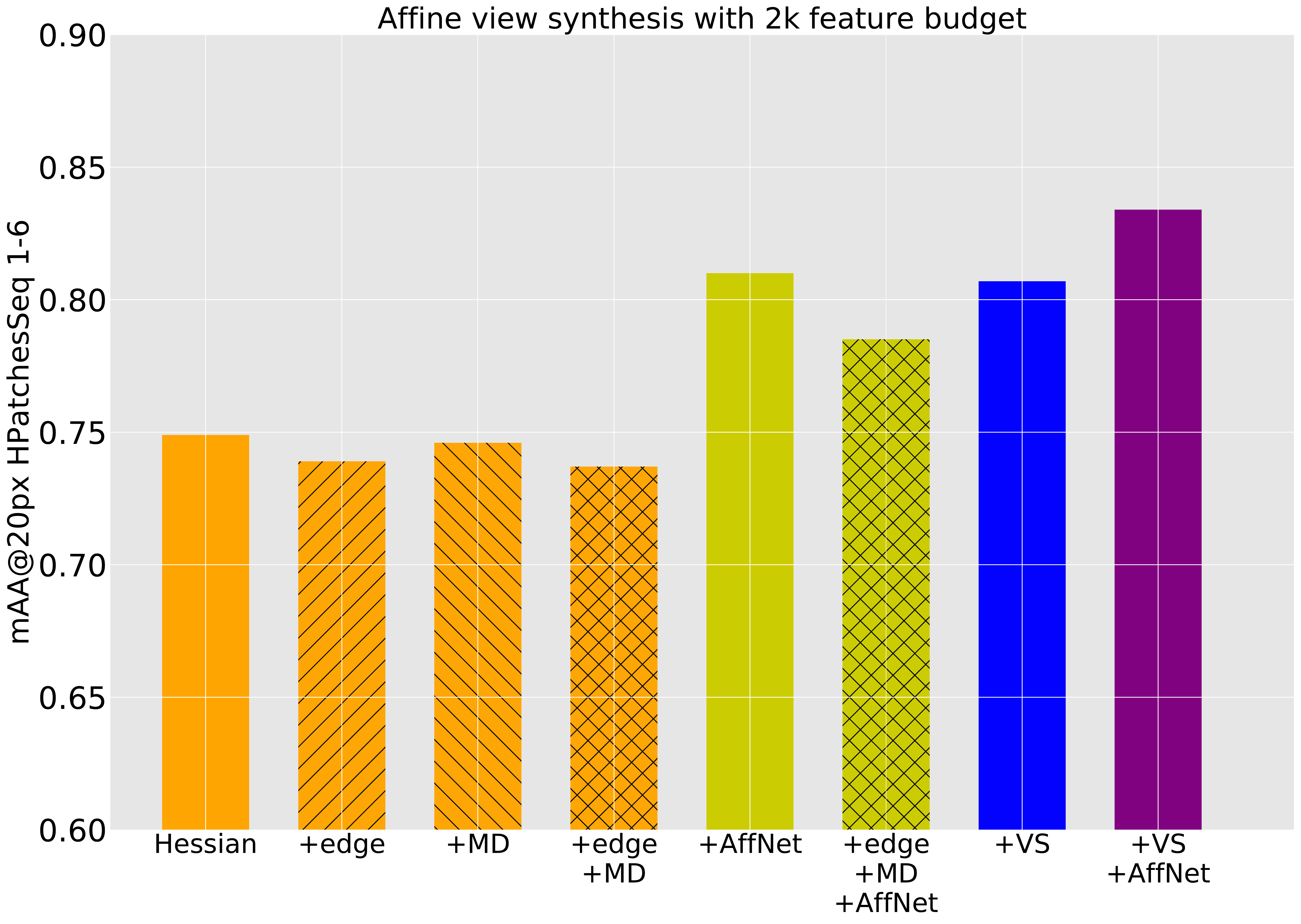}
\caption{Ablation study of the different benefit from the affine view synthesis: edge -- ability to detect elongated blobs by relaxed edge threshold, MD -- multiple affine decriptors, AffNet -- affine shape estimation by AffNet, VS -- view synthesis. Feature budget: left -- 8k, right -- 2k}
\label{fig:affine-ablation-study}
\end{figure}

ASIFT and other view-synthesis based approaches are known more than decade. However, we are not aware of a study explaining why affine synthesis helps in practice. Could one get a similar performance without view synthesis? To answer this question, we performed an ablation study to test the following hypothesis:

\begin{enumerate}
\def\labelenumi{\arabic{enumi}.}
\setcounter{enumi}{-1}

\item
  May it be that the most of improvements come from the fact that we
  have much more features? That is why we fix the number of features for
  all approaches.
\item
  Some regions from ASIFT, when reprojected to the original image, are
  quite narrow. Could we get them just by removing edge-like feature
  filtering, which is done in SIFT, Hessian and other detectors. Denoted
  \textbf{+edge}
\item
  Instead of doing affine view synthesis, one could directly use the
  same affine parameters to get the affine regions to describe, so the
  each keypoint would have several associated regions+descriptors.
  Denoted \textbf{+MD}
\item
  Using AffNet to directly estimated local affine shape without multiple
  descriptors. Denoted \textbf{+AffNet}
\item
  Combine (1), (2) and (3).
\end{enumerate}

The study was conducted on the HPatches Sequences dataset\cite{hpatches2017}, the hardest image
pairs (1-6) of the viewpoint subset. The metric is similar to the one used in~\cite{IMW2020}
-- the mean average accuracy. Instead of the camera pose we threshold the reprojection error of the estimated homography. First, the mean reprojection error for the part visible in both images is calculated. Second, a sequence of thresholds in $\log_2$ space from 1.0 to 20.0 is applied Finally, all per-threshold accuracies are averaged into the final score.

We run the Hessian detector with the RootSIFT descriptor, FLANN matching, and LO-RANSAC, as implemented in MODS.  Features are
sorted according to the detector response and their total number is clipped to 2048 and 8000 to ensure that the improvements do not come from just having more features.

Note that we do not study the view synthesis effect on regular image pairs. Instead, we were focusing on the case when view synthesis is required: matching oblique views of mostly planar scenes.

Results are shown in Figure~\ref{fig:affine-ablation-study}. Indeed, all of the factors: detecting more edge-like features, having multiple descriptors, or better affine shape
improve the results over the plain Hessian detector. However, even all combined are not enough to match performance of the affine view
synthesis + plain Hessian detector. Yet, the best setup is to use both Hessian-AffNet and view synthesis.
The picture is a bit different in a small feature budget: neither
multiple-(affine)-descriptors per keypoint, nor allowing edge-like
feature help. On the other hand, affine view synthesis still improves
results of the Hessian. And, again, the best performance is achieved
with a combination of view synthesis and AffNet shape estimation.

\section{Summary}
\label{mods:conclusions}
An algorithm for two-view matching called  Matching On Demand with view Synthesis algorithm (MODS) was introduced. 
The most important contributions of the algorithm are its ability to adjust its complexity to the problem at hand, and its robustness, i.e., the ability to solve a broader range of wide-baseline problems than the state of the art. This is achieved while being fast on simple problems. 

The apparent robustness vs. speed trade-off is finessed by the use of progressively more time-consuming feature detectors, and by on-demand generation of synthesized images that is performed until a reliable estimate of the geometry is obtained. The MODS method demonstrates that the answer to the question "which detector is the best?" depends on the problem at hand, and that it is fruitful to focus on the "how to combine detectors" problem.

We are the first to propose a view synthesis for two-view wide-baseline matching with affine-covariant detectors, which is superficially counter-intuitive, and we show that matching with the Hessian-Affine or MSER detectors outperforms the state-of-the-art ASIFT.
View synthesis performs well when used with simple and very fast detectors like ORB, which obtains results similar to ASIFT but in orders of magnitude shorter time.

Minor contributions include an improved method for tentative correspondence selection, applicable both with and without view synthesis and a modification of the standard first to second nearest distance due  that increases the number of correct matches by 5-20\% at no additional computational cost.

The evaluation of the MODS algorithm was carried out both on standard publicly available datasets as well as a new set of geometrically challenging wide baseline problems that we collected and will make public. The experiments show that the MODS algorithm solves matching problems beyond the state-of-the-art and yet is comparable in speed to standard wide-baseline matchers on easy problems. 
Moreover, MODS performs well on other classes of difficult two-view problems like matching of images from different modalities, with large difference of acquisition times or with  significant lighting changes.

\chapter[Image Matching Benchmark]{Image Matching Across Wide Baselines: From Paper to Practice\\
\includegraphics[width=0.8\linewidth, trim= 80 250 170 15, clip]{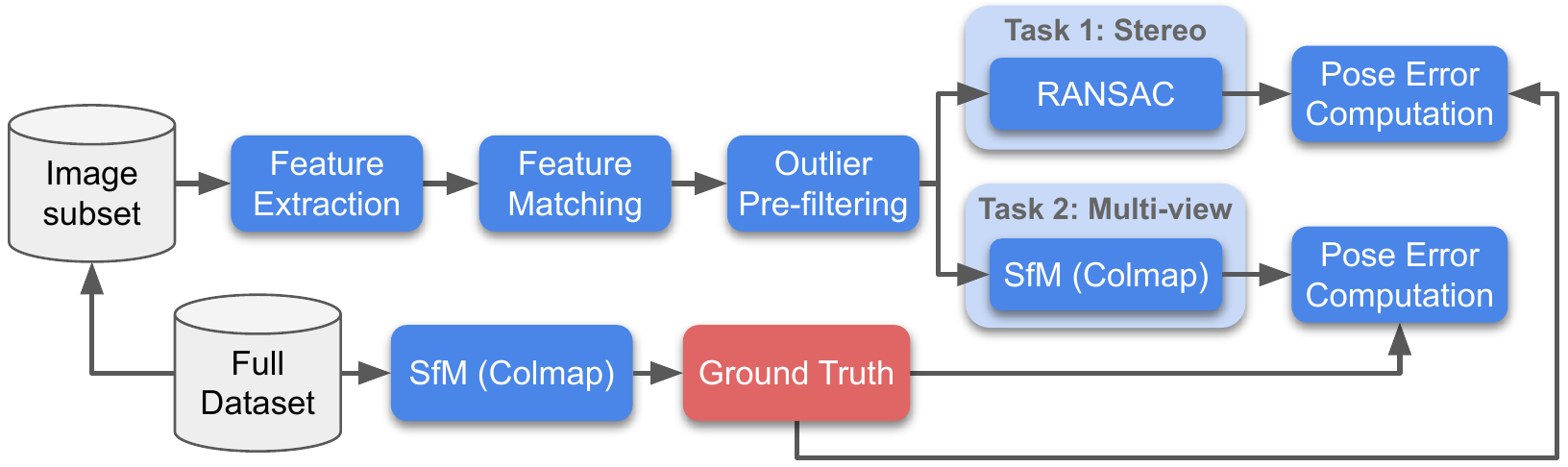}}
\label{ch:benchmark}
This Chapter presents a benchmark of the all components of the wide baseline stereo pipeline components. We carefully tune all the baselines and show importance of the match filtering strategies and modern RANSACs. 
\begin{figure}
\centering
\renewcommand{\imsize}{6.5cm}
\renewcommand{\hspacer}{0.3mm}
\renewcommand{\vspacer}{0.1mm}
\renewcommand{\imscl}{0.01}
\includegraphics[scale=\imscl,width=0.495\linewidth, trim = 0 0 0 80, clip]{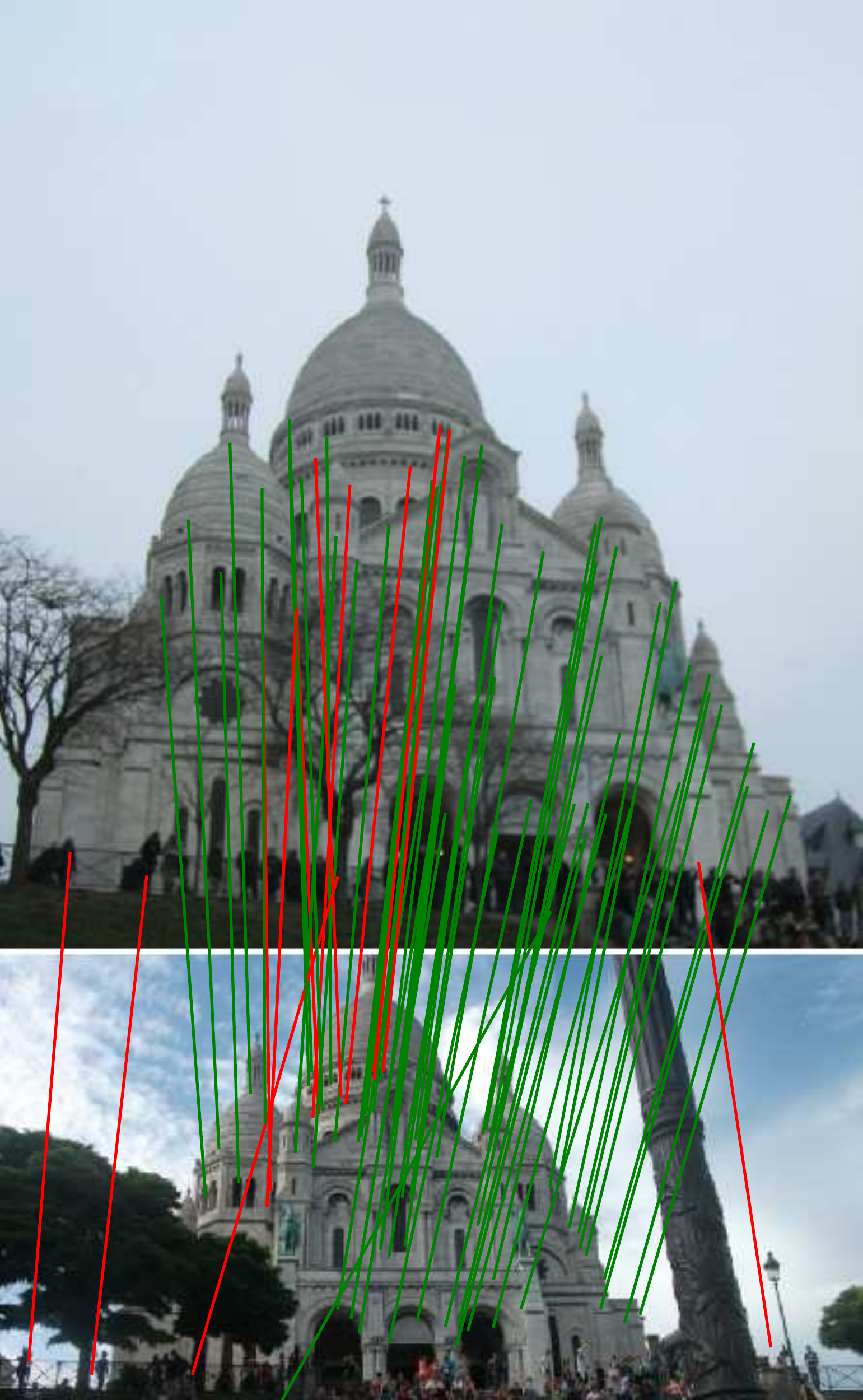}\hspace{\hspacer}%
\includegraphics[scale=\imscl,width=0.495\linewidth, trim = 0 0 0 80, clip]{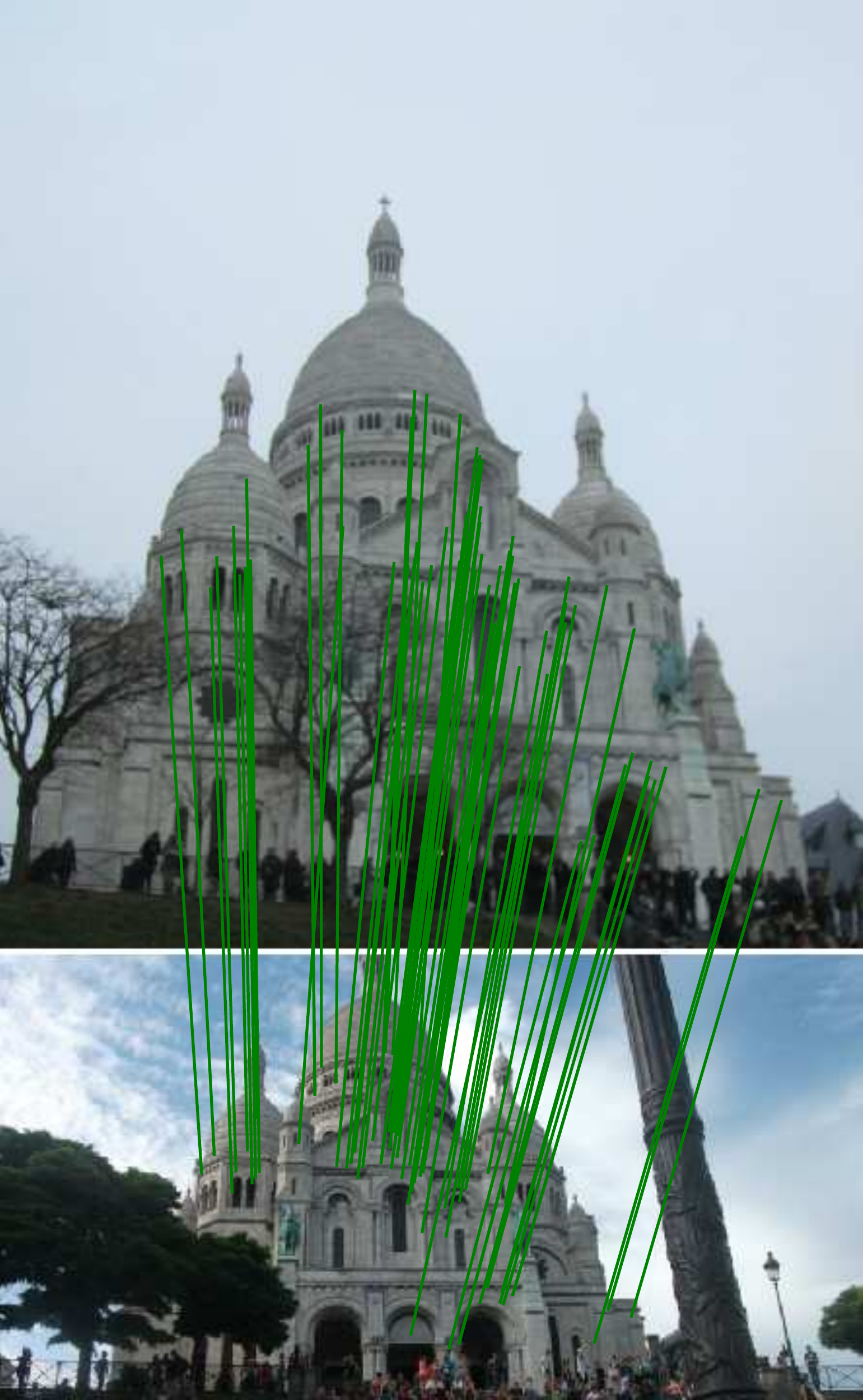}\hspace{\hspacer}%
\caption{
Every paper claims to outperform the state of the art. Is this possible, or an artifact of insufficient validation? On the left, we show stereo matches obtained with {\bf D2-Net} (2019)~\cite{D2Net2019}, a state-of-the-art local feature, using OpenCV RANSAC with its default settings. We color the inliers in green if they are correct and in red otherwise. On the right, we show
{\bf SIFT} (1999)~\cite{SIFT2004} with a carefully tuned MAGSAC~\cite{magsac2019} -- notice how the latter performs much better. This illustrates our take-home message: to correctly evaluate a method's performance, it needs to be embedded within the pipeline used to solve a given problem, and the different components in said pipeline need to be tuned carefully and jointly, which requires engineering and domain expertise.
We fill this need with a new, modular benchmark for sparse image matching,
incorporating dozens of built-in methods.}

\label{fig:teaser}
\vspace{-4mm}
\end{figure}

\vspace{-3mm}
\section{Introduction}
\label{imc:intro}

Matching two or more views of a scene is at the core of fundamental computer vision problems, including image retrieval~\cite{SIFT2004,NetVLAD2016,Radenovic-ECCV2016CNNfromBOW,ASMK2016,DELF2017}, 3D reconstruction~\cite{RomeInDay2009,Heinly15,COLMAP2016,Zhu_2018_CVPR}, re-localization~\cite{ActiveSearch2012,Sattler18,Lynen19}, and SLAM~\cite{Mur15,DeTone17a,SuperPoint2017}.
\edit{Despite decades of research, image matching remains unsolved in the general wide-baseline scenario.
Image matching is a challenging problem with many factors that need to be taken into account, e.g., viewpoint, illumination, occlusions, and camera properties.
It has therefore been traditionally approached with sparse methods -- that is, with local features.}

Recent efforts have moved towards \edit{\emph{holistic}}, end-to-end solutions \cite{PoseNet2015,Balntas_2018_ECCV,Bui_2019_ICCV}. 
\edit{Despite their promise},
they \edit{are yet to} outperform the \edit{\emph{separatists}}~\cite{sattler2019understanding,zhou2019learn} \edit{that are based on the classical paradigm of step-by-step solutions.}
For example, in \edit{a classical wide baseline stereo pipeline~\cite{Pritchett1998}, one} 
(1)~extracts local features, such as SIFT \cite{SIFT2004},
(2)~creates a list of tentative matches by nearest-neighbor search in descriptor space, and 
(3)~retrieves the pose with a minimal solver inside a robust estimator, such as the 7-point algorithm \cite{Hartley94_7pt} in a RANSAC loop \cite{RANSAC1981}.
To build a 3D reconstruction out of a set of images, same matches are fed to a bundle adjustment pipeline~\cite{Hartley2004,Triggs00} to jointly optimize the camera intrinsics, extrinsics, and 3D point locations.
This modular structure simplifies the engineering a solution to the problem and allows for incremental improvements, of which there have been hundreds, if not thousands.

New methods for each of the sub-problems, such as feature extraction and pose estimation, are typically studied in isolation, using intermediate metrics, which simplifies their evaluation.
However, there is no guarantee that gains in one part of the pipeline will translate to the final application, as these components interact in complex ways.
For example, patch descriptors, including very recent work~\cite{He18b,Wei18,SoSNet2019,Mukundan19}, are often evaluated on Brown's seminal patch retrieval database~\cite{Brown10},
\edit{introduced in 2007}.
They 
show dramatic improvements -- \edit{up to 39x relative~\cite{Wei18}} -- over handcrafted methods such as SIFT, but it is unclear whether this 
\edit{remains} true on real-world applications. In fact, we later demonstrate that the gap
narrows dramatically when 
\edit{decades-old}
baselines are properly tuned.

We posit that it is \edit{critical}
to look beyond intermediate metrics and focus on downstream performance.
This is particularly important {\em now}
\revision{as, while
deep networks are reported to outperform algorithmic solutions on classical, \edit{sparse} problems such as outlier filtering \cite{CNe2018,Ranftl18,Zhang19,Sun19,brachmann2019ngransac}, bundle adjustment \cite{Tang19,shi2019selfsupervised}, SfM \cite{Vijayna17,Deepsfm2019} and SLAM \cite{Tateno_2017_CVPR,gradslam2020},
our findings in this Chapter suggest that this may not always be the case.
}
To this end, we introduce a benchmark for wide-baseline image matching, including:
\begin{enumerate}[label=(\alph*)]
  \item \edit{A dataset with thousands of phototourism images of 25 landmarks, taken from diverse viewpoints, with different cameras, in varying illumination and weather conditions -- all of which are necessary for a comprehensive evaluation.
  We reconstruct the scenes 
  with 
  SfM, without the need for human intervention,
  providing depth maps and ground truth poses for 26k images, and reserve another 4k for a private test set.}
  \item A modular pipeline\footnote{\label{code}{\scriptsize \url{https://github.com/vcg-uvic/image-matching-benchmark}}} incorporating dozens of methods for feature extraction, matching, and pose estimation,
  both classical and state-of-the-art, as well as multiple heuristics, \edit{all of} which can be swapped out and tuned separately.
  \item Two downstream tasks -- stereo and multi-view reconstruction -- evaluated with both downstream and intermediate metrics, for comparison.
  \item A thorough study of dozens of methods and techniques, both hand-crafted and learned, and their combination, along with a \edit{recommended} procedure for \edit{effective} hyper-parameter selection.
\end{enumerate}

The framework
\edit{enables}
researchers to evaluate how a new approach performs in a standardized pipeline, both {\em against} its competitors\footnote{\label{website}{\scriptsize \url{https://vision.uvic.ca/image-matching-challenge}}}, and {\em alongside} state-of-the-art solutions for other components, from which it
\edit{simply cannot be detached.}
This is crucial, as the true performance can be easily hidden by sub-optimal hyperparameters.

\begin{figure*}
\centering
\renewcommand{\arraystretch}{0.2}
\renewcommand{\tabcolsep}{0mm}
\footnotesize
\centering
\renewcommand{\imsize}{1cm}
\begin{tabular}{@{}cccccccccc@{}}
\includegraphics[scale=\imscl,height=\imsize]{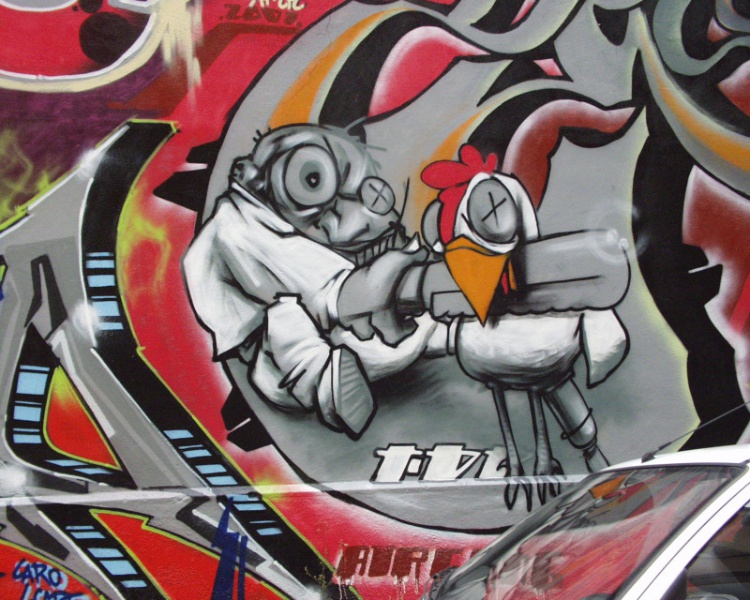}\hspace{\hspacer} &
\includegraphics[scale=\imscl,height=\imsize]{chapters/benchmark/fig/other-benchmarks/new/hpatches0-fit.jpg}\hspace{\hspacer} &
\includegraphics[scale=\imscl,height=\imsize]{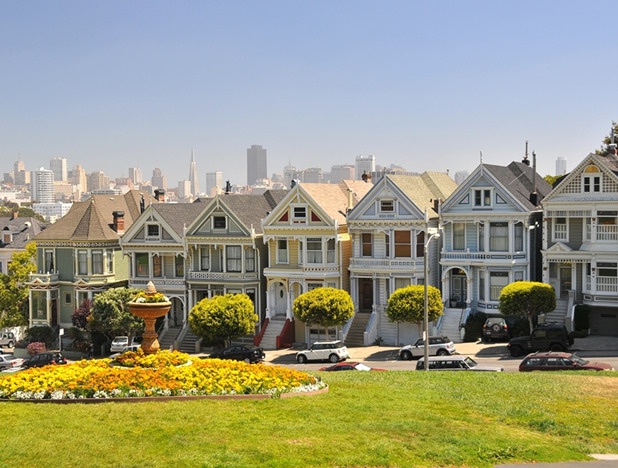}\hspace{\hspacer} &
\includegraphics[scale=\imscl,height=\imsize]{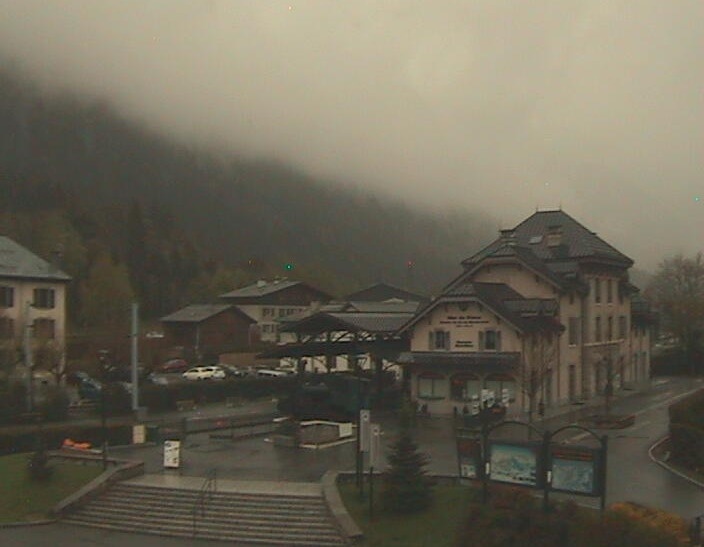}\hspace{\hspacer} &
\includegraphics[scale=\imscl,height=\imsize]{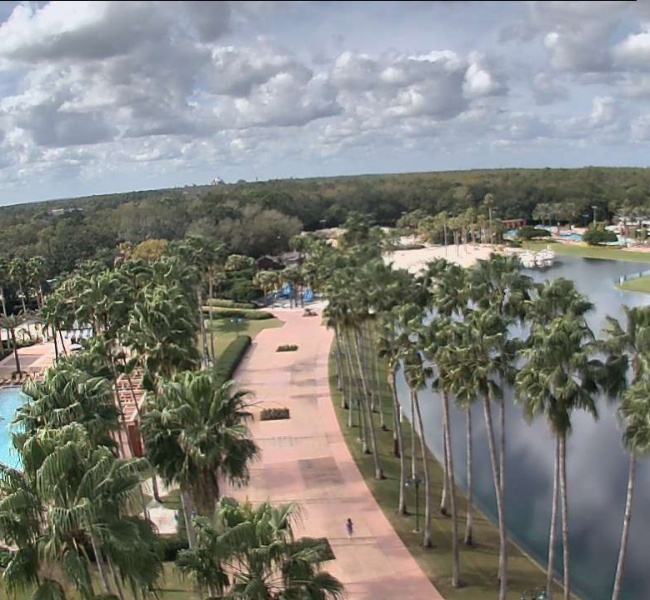}\hspace{\hspacer} &
\includegraphics[scale=\imscl,height=\imsize]{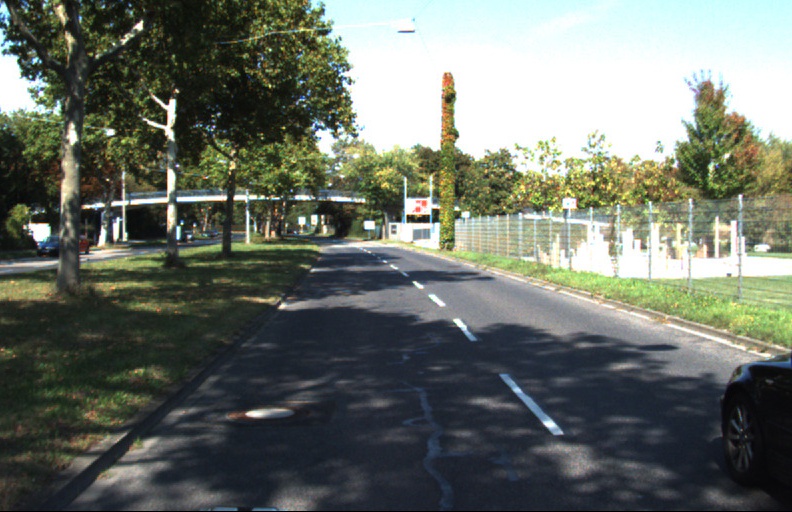}\hspace{\hspacer} &
\includegraphics[scale=\imscl,height=\imsize]{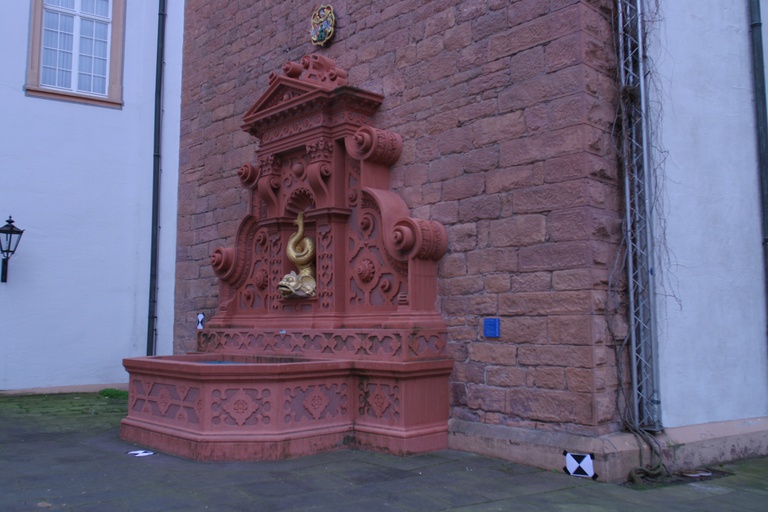}\hspace{\hspacer} &
\includegraphics[scale=\imscl,height=\imsize]{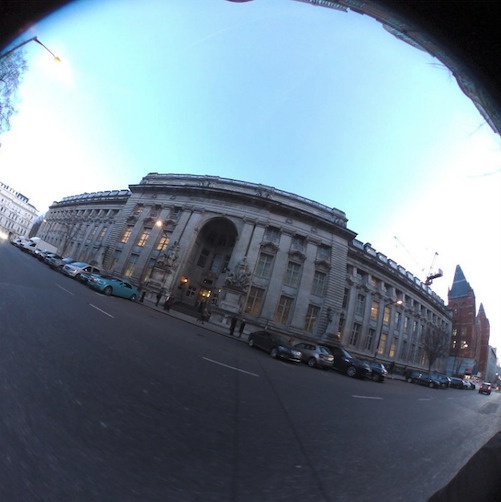}\hspace{\hspacer} &
\includegraphics[scale=\imscl,height=\imsize]{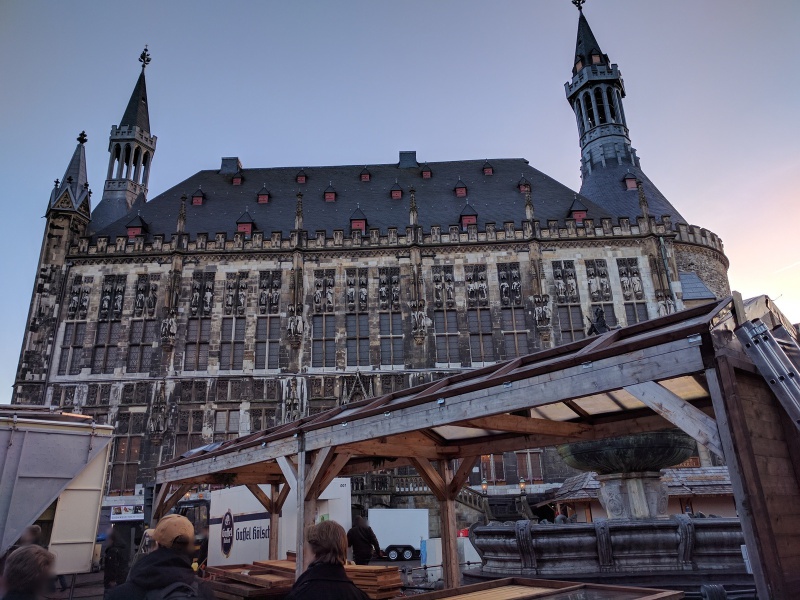}\hspace{\hspacer} &
\includegraphics[scale=\imscl,height=\imsize]{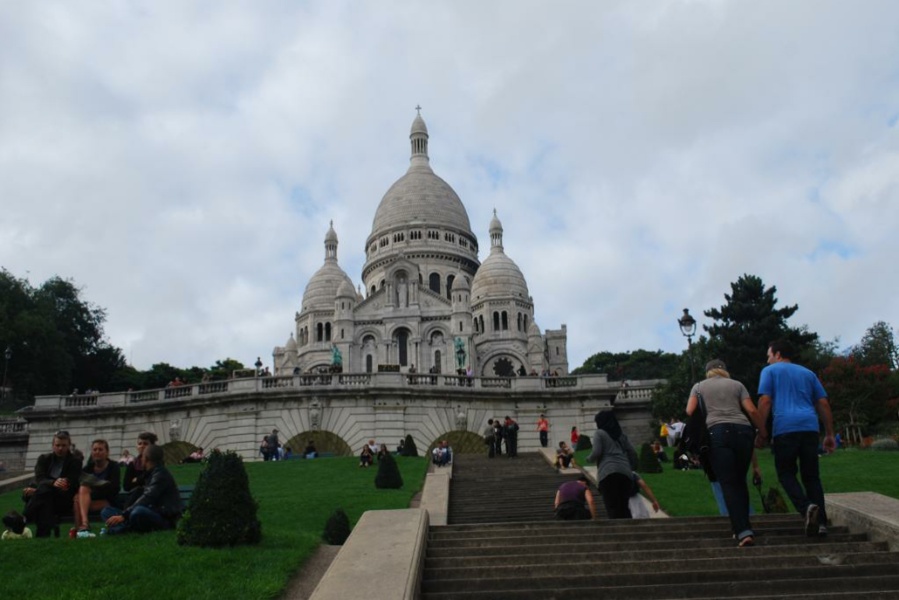}\hspace{\hspacer} \\
\includegraphics[scale=\imscl,height=\imsize]{chapters/benchmark/fig/other-benchmarks/new/oxford1-fit.jpg}\hspace{\hspacer} &
\includegraphics[scale=\imscl,height=\imsize]{chapters/benchmark/fig/other-benchmarks/new/hpatches1-fit.jpg}\hspace{\hspacer} &
\includegraphics[scale=\imscl,height=\imsize]{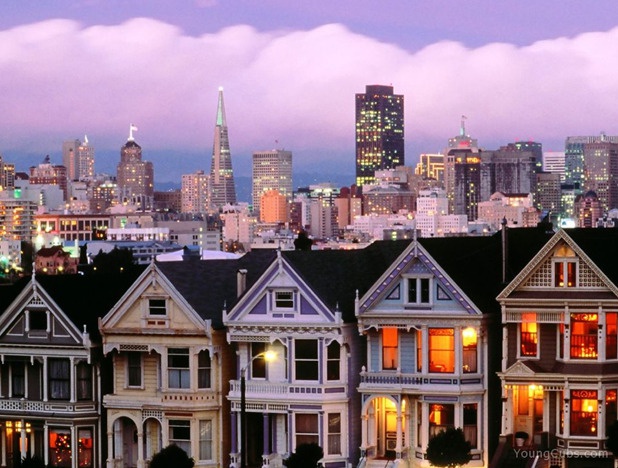}\hspace{\hspacer} &
\includegraphics[scale=\imscl,height=\imsize]{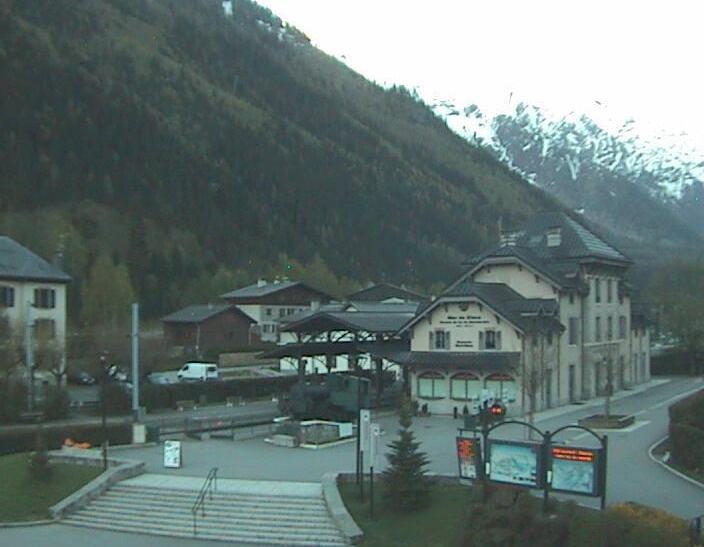}\hspace{\hspacer} &
\includegraphics[scale=\imscl,height=\imsize]{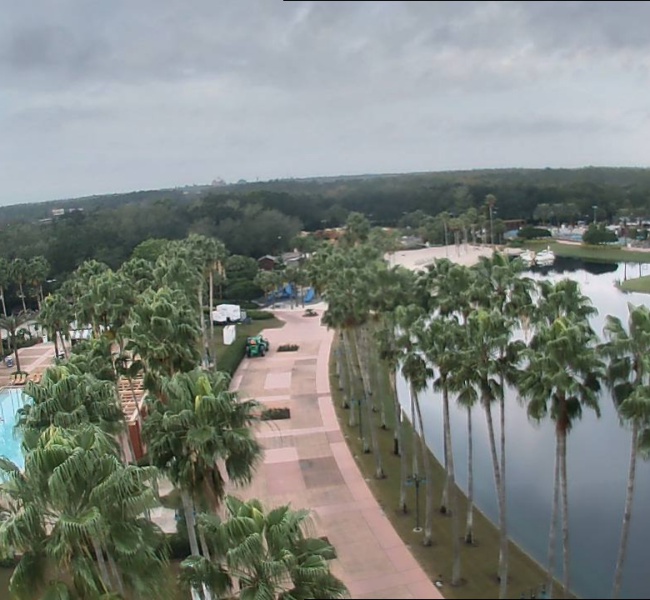}\hspace{\hspacer} &
\includegraphics[scale=\imscl,height=\imsize]{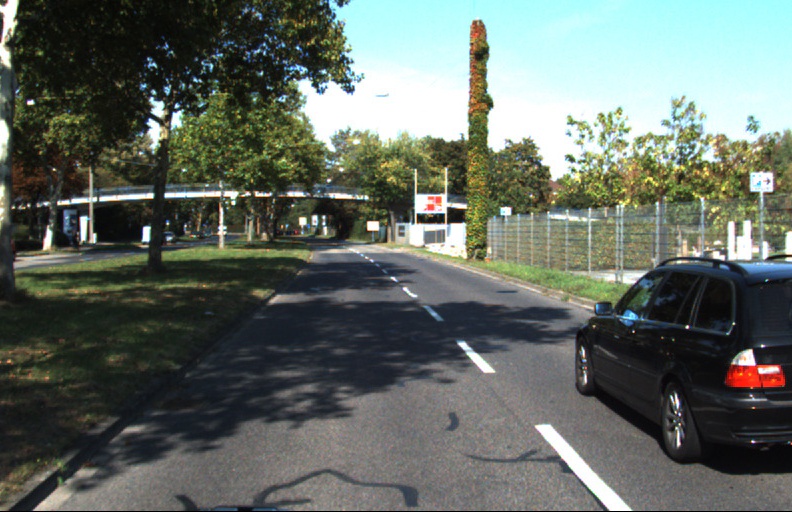}\hspace{\hspacer} &
\includegraphics[scale=\imscl,height=\imsize]{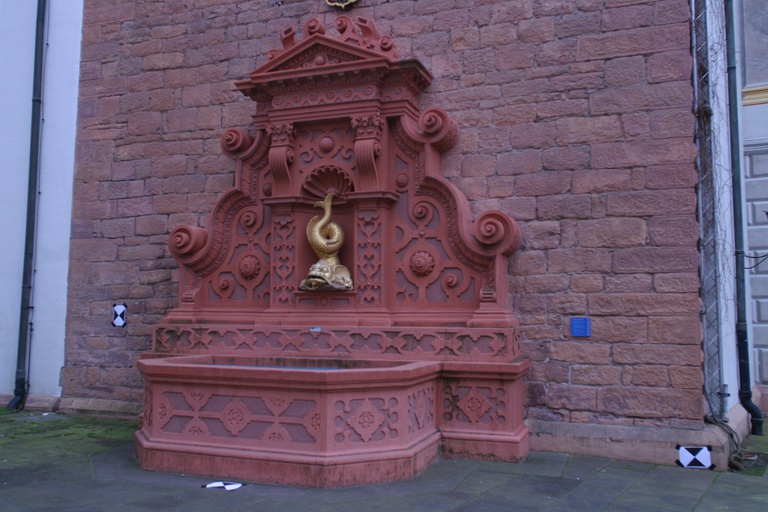}\hspace{\hspacer} &
\includegraphics[scale=\imscl,height=\imsize]{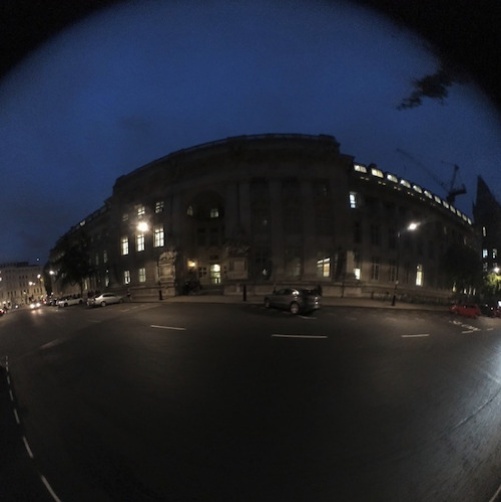}\hspace{\hspacer} &
\includegraphics[scale=\imscl,height=\imsize]{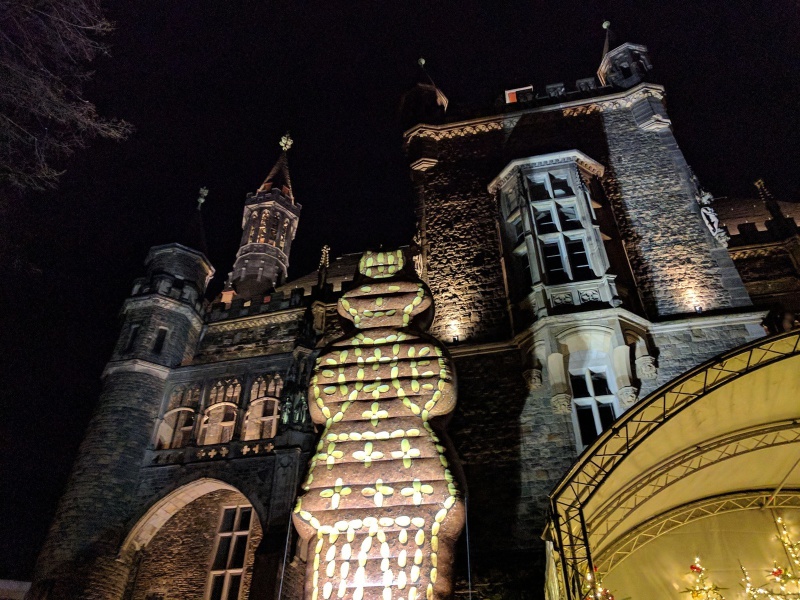}\hspace{\hspacer} &
\includegraphics[scale=\imscl,height=\imsize]{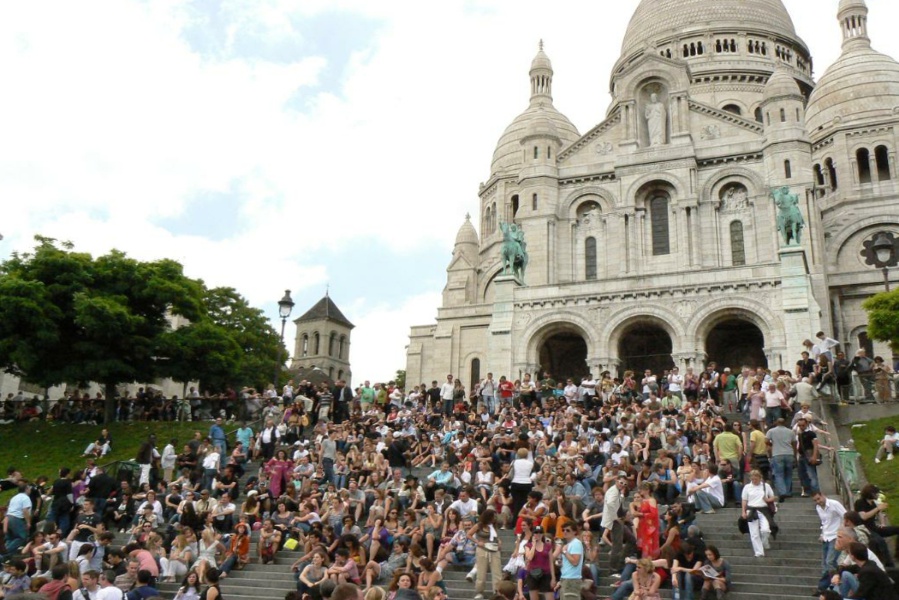}\hspace{\hspacer} \\
\includegraphics[scale=\imscl,height=\imsize]{chapters/benchmark/fig/other-benchmarks/new/oxford2-fit.jpg}\hspace{\hspacer} &
\includegraphics[scale=\imscl,height=\imsize]{chapters/benchmark/fig/other-benchmarks/new/hpatches2-fit.jpg}\hspace{\hspacer} &
\includegraphics[scale=\imscl,height=\imsize]{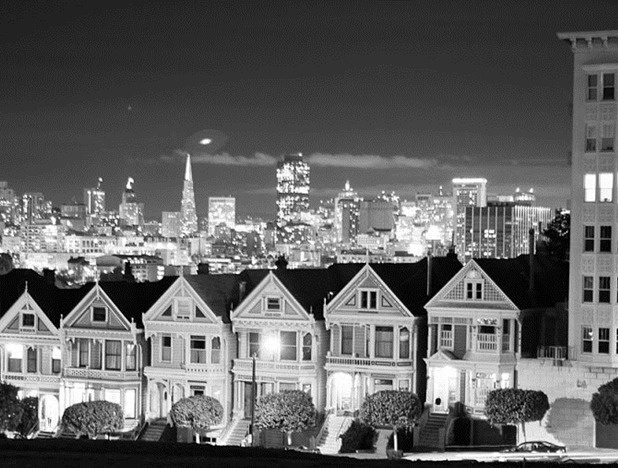}\hspace{\hspacer} &
\includegraphics[scale=\imscl,height=\imsize]{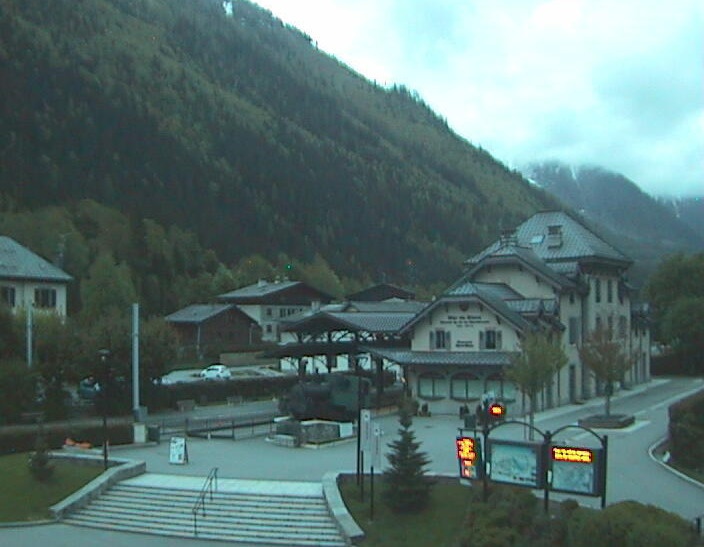}\hspace{\hspacer} &
\includegraphics[scale=\imscl,height=\imsize]{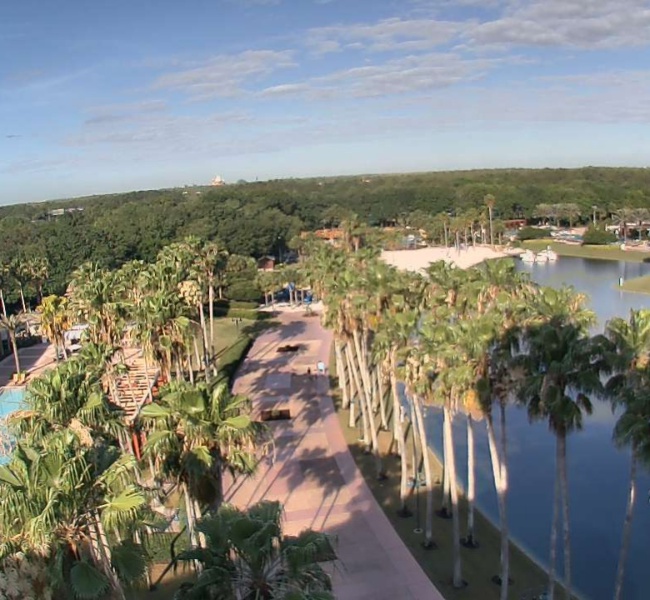}\hspace{\hspacer} &
\includegraphics[scale=\imscl,height=\imsize]{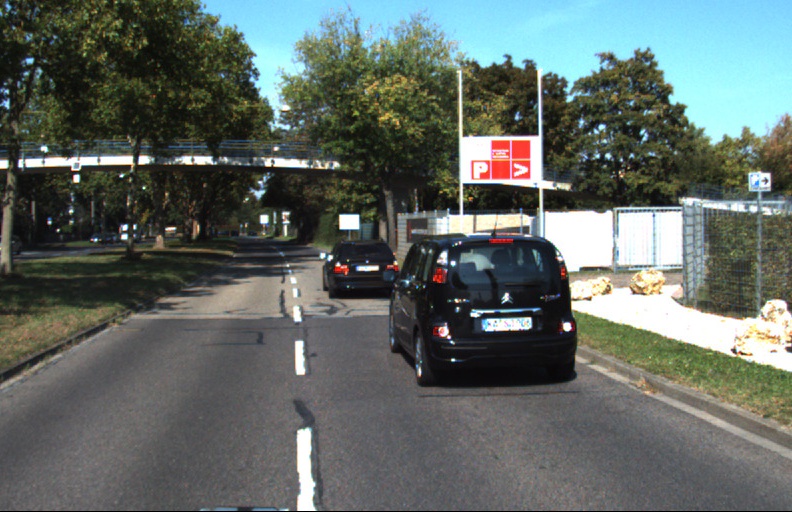}\hspace{\hspacer} &
\includegraphics[scale=\imscl,height=\imsize]{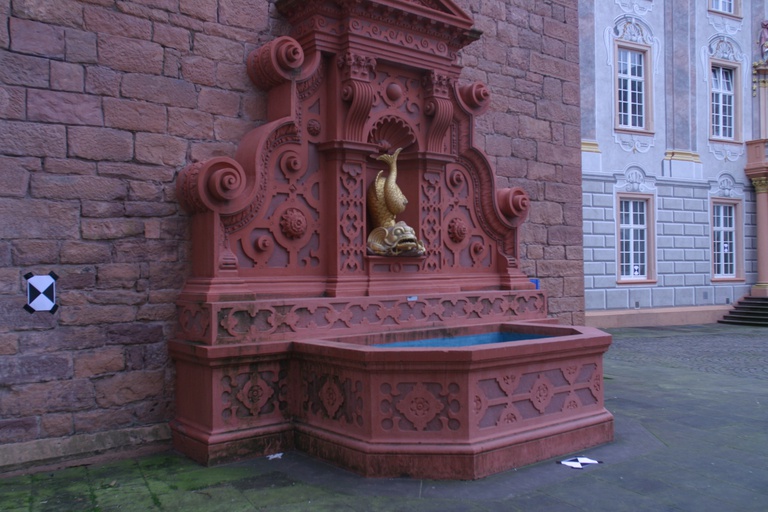}\hspace{\hspacer} &
\includegraphics[scale=\imscl,height=\imsize]{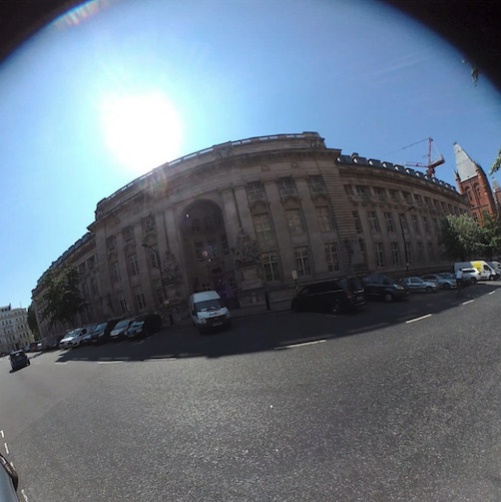}\hspace{\hspacer} &
\includegraphics[scale=\imscl,height=\imsize]{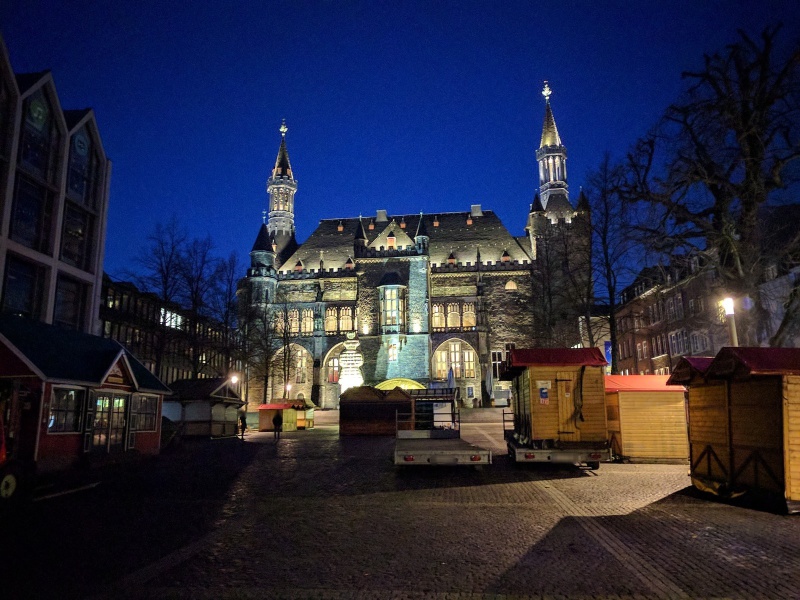}\hspace{\hspacer} &
\includegraphics[scale=\imscl,height=\imsize]{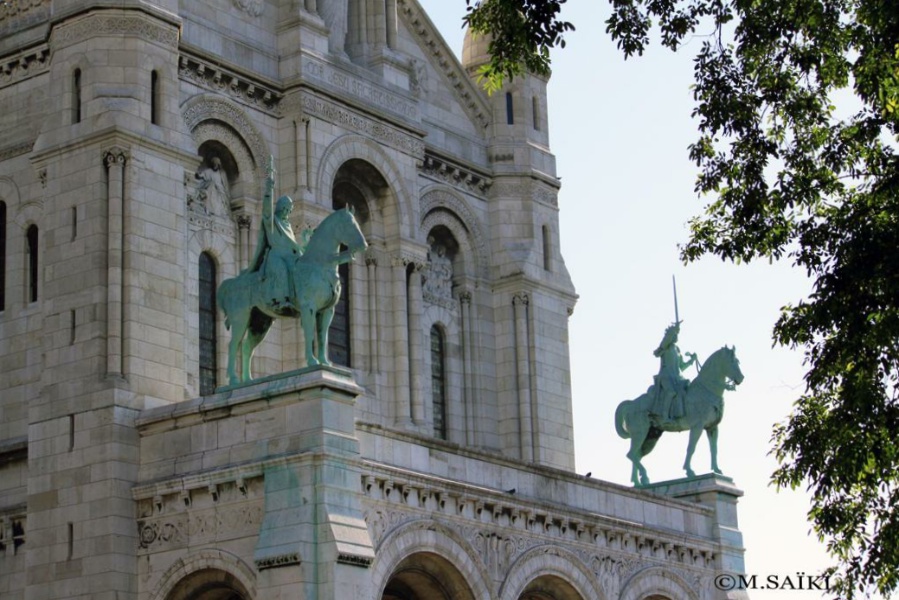}\hspace{\hspacer} \\
\vspace{1mm} \\
{\small (a)} & {\small (b)} & {\small (c)}  & {\small (d)} & {\small (e)} & {\small (f)} & {\small (g)} & {\small (h)} & {\small (i)} & {\small (j)} \\
\end{tabular}
\caption{\edit{
{\bf On the limitations of previous datasets.}
\edit{To highlight the need for a new benchmark,} we show examples from datasets and benchmarks featuring posed images that have been previously used to evaluate local features and robust matchers (some images are cropped for the sake of presentation).
From left to right: 
(a) VGG Oxford~\cite{MikoDescEval2003}, 
(b) HPatches~\cite{hpatches2017}, 
(c) Edge Foci~\cite{Zitnick2011}, 
(d) Webcam~\cite{TIlde2015}, 
(e) AMOS~\cite{Jacobs07}, 
(f) Kitti~\cite{Geiger12}, 
(g) Strecha~\cite{Strecha08b}, 
(h) SILDa~\cite{SILDa2018}, 
(i) Aachen~\cite{ActiveSearch2012}, 
(j) Ours. 
Notice that many (a-e) contain only planar structures or illumination changes, which makes it easy -- or trivial -- to obtain ground truth poses, encoded as homographies. 
Other datasets are small -- (g) contains very accurate depth maps, but only two scenes and 19 images total -- or do not contain a wide enough range of photometric and viewpoint transformations.} \edit{Aachen (i) is closer to ours (j), but relatively small, limited to one scene, and focused on re-localization on the day vs. night use-case.}}
\label{fig:other_benchmarks}
\end{figure*}

\section{Related Work}
\label{imc:related}
The literature on image matching is too vast for a thorough overview. We cover relevant methods for feature extraction and matching, pose estimation, 3D reconstruction, applicable datasets, and evaluation frameworks.

\subsection{Local features}
Local features became a staple in computer vision with the introduction of SIFT~\cite{SIFT2004}. They typically involve three distinct steps: keypoint detection, orientation estimation, and descriptor extraction. Other popular \edit{classical} solutions are SURF~\cite{Bay2006}, ORB~\cite{ORB2011}, and AKAZE~\cite{AKAZE2013}.

Modern descriptors often train deep networks on pre-cropped patches, typically from SIFT keypoints (\ie Difference of Gaussians or DoG). They include Deepdesc~\cite{simo2015deepdesc}, TFeat~\cite{TFeat2016}, L2-Net~\cite{L2Net2017}, HardNet~\cite{HardNet2017}, SOSNet~\cite{SoSNet2019}, and LogPolarDesc~\cite{LogPolar2019} -- 
most of them are trained on the same dataset~\cite{Brown10}. Recent works leverage additional cues, such as geometry or global context, including GeoDesc~\cite{geodesc2018} and ContextDesc~\cite{ContextDesc2019}.
There have been multiple attempts to learn keypoint detectors separately from the descriptor, including TILDE~\cite{TIlde2015}, TCDet~\cite{NewTilde2017}, QuadNet~\cite{QuadNets2017}, and Key.Net~\cite{KeyNet2019}.
An alternative is to treat this as an end-to-end learning problem, a trend that started with the introduction of LIFT~\cite{LIFT2016} and includes DELF~\cite{DELF2017}, SuperPoint~\cite{SuperPoint2017}, LF-Net~\cite{LFNet2018}, D2-Net~\cite{D2Net2019} and R2D2~\cite{R2D22019}.

\subsection{Robust matching}
Inlier ratios in wide-baseline stereo can be below 10\% -- and sometimes \edit{even} lower.
This is typically approached with iterative sampling schemes based on RANSAC~\cite{RANSAC1981}, relying on closed-form solutions for pose solving such as the 5-~\cite{Nister03}, 7-~\cite{Hartley94_7pt} or 8-point algorithm~\cite{Hartley97}.
\edit{Improvements to this classical framework include local optimization~\cite{LOransac2003}, using likelihood instead of reprojection (MLESAC)~\cite{MLESAC00}, speed-ups using probabilistic sampling of hypotheses (PROSAC)~\cite{PROSAC2005}, degeneracy check using homographies (DEGENSAC)~\cite{Degensac2005},
Graph Cut as a local optimization (GC-RANSAC)~\cite{gcransac2018}, and auto-tuning of thresholds using confidence margins (MAGSAC)~\cite{magsac2019}.

\edit{As an alternative} 
direction, recent works, starting with CNe (Context Networks) in~\cite{CNe2018}, train deep networks for outlier rejection taking correspondences as input, often followed by a RANSAC loop. 
Follow-ups include~\cite{Ranftl18,Zhao19,Sun19,Zhang19}.}
\edit{Differently from RANSAC solutions, they typically process all correspondences in a single forward pass, without the need to \edit{iteratively} sample hypotheses.}
Despite their promise, it remains unclear how well they perform in real\edit{-world} settings \edit{compared to a well-tuned RANSAC}.
\subsection{Structure from Motion (SfM)}
In \edit{SfM,} one jointly optimizes the location of the 3D points and the camera intrinsics and extrinsics.
Many improvements have been proposed over the years~\cite{RomeInDay2009,Heinly15,Cui_2017_CVPR_HSFM,PSfMO2017,Zhu_2018_CVPR}.
The most popular frameworks are VisualSFM~\cite{Wu13} and COLMAP~\cite{COLMAP2016} --
we rely on the latter to \edit{both} generate the ground truth and as the backbone of our multi-view task.
\begin{figure*}
\centering
\renewcommand{\imsize}{1.4cm}
\renewcommand{\hspacer}{0.3mm}
\renewcommand{\vspacer}{0.1mm}
\renewcommand{\imscl}{0.01}
\includegraphics[scale=\imscl,height=\imsize]{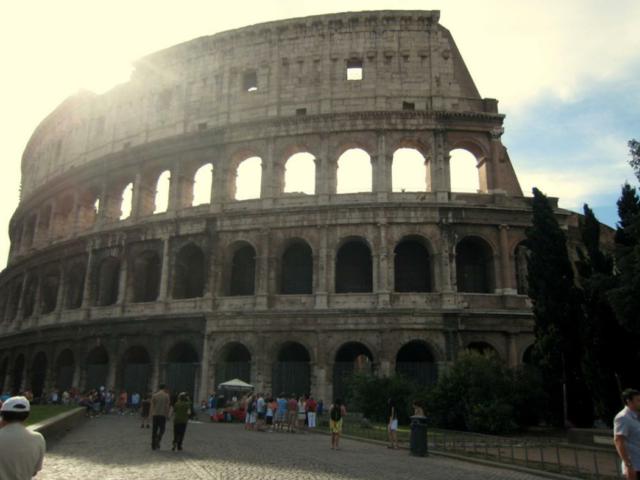}\hspace{\hspacer}%
\includegraphics[scale=\imscl,height=\imsize]{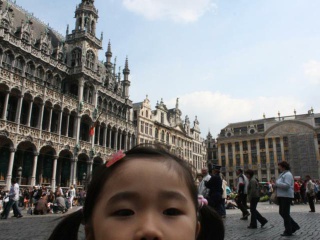}\hspace{\hspacer}%
\includegraphics[scale=\imscl,height=\imsize]{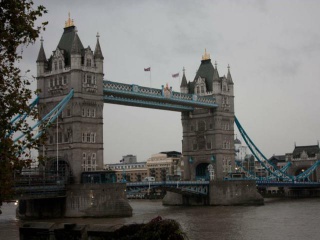}\hspace{\hspacer}%
\includegraphics[scale=\imscl,height=\imsize]{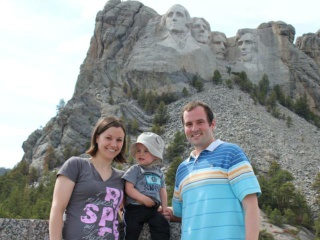}\hspace{\hspacer}%
\includegraphics[scale=\imscl,height=\imsize]{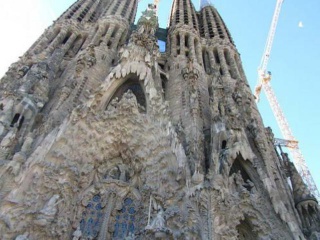}\hspace{\hspacer}%
\includegraphics[scale=\imscl,height=\imsize]{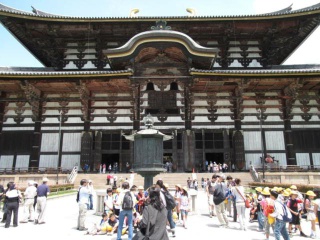}\hspace{\hspacer}%
\includegraphics[scale=\imscl,height=\imsize]{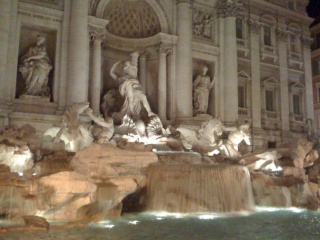}\hspace{\hspacer}%
\includegraphics[scale=\imscl,height=\imsize]{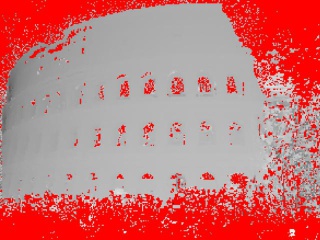}\hspace{\hspacer}%
\includegraphics[scale=\imscl,height=\imsize]{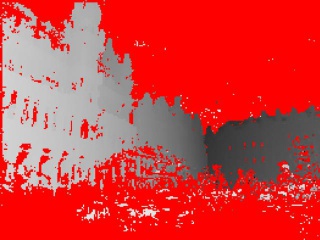}\hspace{\hspacer}%
\includegraphics[scale=\imscl,height=\imsize]{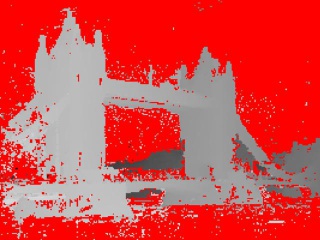}\hspace{\hspacer}%
\includegraphics[scale=\imscl,height=\imsize]{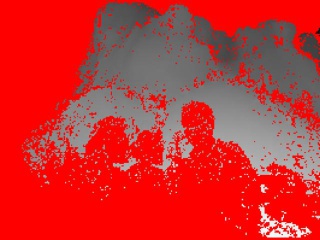}\hspace{\hspacer}%
\includegraphics[scale=\imscl,height=\imsize]{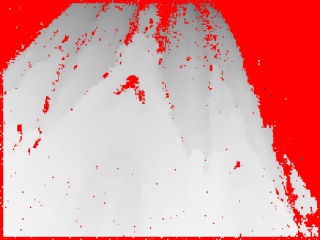}\hspace{\hspacer}%
\includegraphics[scale=\imscl,height=\imsize]{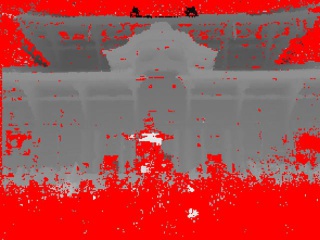}\hspace{\hspacer}%
\includegraphics[scale=\imscl,height=\imsize]{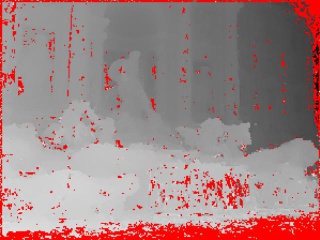}\hspace{\hspacer}%
\caption{
{\bf The Image Matching Challenge PhotoTourism (IMC-PT) dataset.}
We show a few selected images from our dataset and their corresponding depth maps, with occluded pixels marked in red. 
}
\label{fig:dataset_samples}
\vspace{-2mm}
\end{figure*}

\subsection{Datasets and benchmarks}
Early works \edit{on local features and robust matchers typically} relied on the Oxford dataset~\cite{MikoDescEval2003}, \edit{which contains} 48 images and ground truth homographies. 
It helped establish two common metrics \edit{\edit{for evaluating} local feature performance}: repeatability and matching score.
Repeatability evaluates the keypoint detector: given \edit{two} sets of keypoints over two images projected into each other, it is defined as the ratio of keypoints whose support regions' \edit{overlap} is above a threshold.
The matching score (MS) is similarly defined, but also requires their descriptors to be the nearest neighbours.
Both require pixel-to-pixel correspondences -- \ie, features outside valid areas are ignored.

A modern alternative to Oxford is HPatches~\cite{hpatches2017}, which contains 696 images with differences in illumination {\em or} viewpoint -- \edit{but not both}. However, the scenes are planar, without occlusions \edit{limiting its applicability}.

Other datasets 
\edit{that have been used to evaluate local features} 
include DTU~\cite{Aanaes2012}, Edge Foci~\cite{Zitnick2011}, Webcam~\cite{TIlde2015}, AMOS~\cite{HardNetAMOS2019}, and Strecha's~\cite{Strecha08b}. 
They all have limitations -- \edit{be it} narrow baselines, noisy ground truth, or a small number of images.
\edit{In fact, most learned descriptors have been trained and evaluated on~\cite{Brown07}, which provides a database of pre-cropped patches with correspondence labels, and measures the performance in terms of patch retrieval. 
While this seminal dataset and evaluation methodology helped develop many new methods, it is not clear how the results translate to different scenarios -- particularly since new methods outperform classical ones such as SIFT by orders of magnitude, which suggests overfitting.}

Datasets used for navigation, re-localization, or SLAM, in outdoor environments are also \edit{relevant to our problem.
These include} KITTI~\cite{Geiger12}, Aachen~\cite{ActiveSearch2012}, Robotcar~\cite{Maddern17}, and CMU seasons~\cite{Sattler18,Badino11}. 
However, they do not feature the wide range of transformations present in the phototourism data.
\edit{Megadepth~\cite{MegaDepth2018} is a more representative, phototourism-based dataset,
which using COLMAP, as in our case}
-- \edit{it could, in fact, easily be folded into ours.}

Benchmarks, by contrast, are few and far between -- they include VLBenchmark~\cite{lenc2012vlbenchmarks}, HPatches~\cite{hpatches2017}, and SILDa~\cite{SILDa2018} -- all limited in scope.
A large-scale benchmark for SfM was proposed in~\cite{Schonberger17},
\edit{which built 3D reconstructions with different local features. However, only a few scenes contain ground truth, so most of their metrics are qualitative -- \eg number of observations or average reprojection error.}
Yi~\etal~\cite{CNe2018} and Bian~\etal~\cite{Bian19}
evaluate different methods for pose estimation on several datasets -- however, few methods are considered and they are not carefully tuned. 

We highlight some of these datasets/benchmarks, and their limitations in \fig{other_benchmarks}.
We are, to the best of our knowledge, the first to introduce a public modular benchmark for 3D reconstruction with sparse methods
using downstream metrics, and featuring a comprehensive dataset with a large range of image transformations.

\section{\revision{The~Image~Matching~Challenge~PhotoTourism~Dataset}}
\label{imc:data}
While it is possible to obtain very accurate poses and depth maps under controlled scenarios with devices like LIDAR, this is costly and requires a specific set-up that
does not scale well.
For example, Strecha's dataset~\cite{Strecha08b}
follows that approach but contains only 19 images.
We argue that a truly representative dataset must contain a wider range of transformations -- including different imaging devices, time of day, weather, partial occlusions, etc.
Phototourism images satisfy this condition
and are readily available.

We thus build on \edit{25} collections of popular landmarks originally selected in~\cite{Heinly15, YFCC100M2016}, each with hundreds to thousands of images.
\revision{Images are downsampled with bilinear interpolation to a maximum size of 1024 pixels along the long-side and their poses were obtained with COLMAP \cite{COLMAP2016}, which provides the (pseudo) ground truth}.
We do exhaustive image matching before Bundle Adjustment -- unlike~\cite{Schoenberger16b}, which uses only 100 pairs for each image -- and thus provide enough matching images for any conventional SfM to return near-perfect results \revision{in standard conditions}.

\edit{%
\revision{Our approach is to obtain a ground truth signal using reliable, off-the-shelf technologies, while making the problem as easy as possible -- and then evaluate new technologies on a much harder problem, using only a subset of that data.}
\edit{For example, we reconstruct a scene with hundreds or thousands of images with vanilla COLMAP and then evaluate ``modern'' features and matchers against its poses using only two images (``stereo'') or up to 25 at a time (``multiview'' with SfM).}
For a discussion regarding the accuracy of our ground truth data, please refer to \secref{eval_gt}.}

In addition to point clouds, COLMAP provides dense depth estimates. 
\revision{These are noisy, and have no notion of occlusions -- a depth value is provided for every pixel.}
We remove occluded pixels \revision{from depth maps} using the reconstructed model \edit{from COLMAP}; see~\fig{dataset_samples} for examples.
We rely on these ``cleaned'' depth maps to compute \edit{classical,} pixel-wise metrics -- repeatability and matching score.
We find that some images are flipped 90$^\circ$, and use the \edit{reconstructed pose} to rotate them -- \edit{along with their poses} -- so they are roughly `upright', \edit{which is a reasonable assumption for this type of data.}

\subsection{Dataset details}
\begin{table}%
\renewcommand{\arraystretch}{1}
\renewcommand{\tabcolsep}{7pt}
\small
\begin{center}
\begin{tabular}{@{}lcrr@{}}
\toprule
Name & Group & Images & 3D points \\
\midrule
``Brandenburg Gate'' & T & 1363 & 100040 \\
``Buckingham Palace'' & T & 1676 & 234052 \\
``Colosseum Exterior'' & T & 2063 & 259807 \\
``Grand Place Brussels''& T & 1083 & 229788 \\
``Hagia Sophia Interior''& T & 888 & 235541 \\
``Notre Dame Front Facade'' &T & 3765 & 488895 \\
``Palace of Westminster''  &T & 983 & 115868 \\
``Pantheon Exterior'' & T& 1401 & 166923 \\
``Prague Old Town Square'' & T& 2316 & 558600 \\
``Reichstag'' & T& 75 & 17823 \\
``Taj Mahal'' &T & 1312 & 94121 \\
``Temple Nara Japan'' &T & 904 & 92131 \\
``Trevi Fountain'' & T & 3191 & 580673 \\
``Westminster Abbey'' & T & 1061 & 198222 \\
``Sacre Coeur'' (SC) & V& 1179 & 140659 \\
``St. Peter's Square'' (SPS) & V & 2504 & 232329 \\
\midrule
Total T + V & & 25.7k & 3.6M \\
\midrule
``British Museum'' (BM) & E & 660 & 73569 \\
``Florence Cathedral Side'' (FCS) & E & 108 & 44143 \\
``Lincoln Memorial Statue'' (LMS) &  E &850 & 58661 \\
``London (Tower) Bridge'' (LB) &  E &629 & 72235 \\
``Milan Cathedral'' (MC) &  E &124 & 33905 \\
``Mount Rushmore'' (MR) &  E &138 & 45350 \\
``Piazza San Marco'' (PSM) &  E &249 & 95895 \\
``Sagrada Familia'' (SF) &  E &401 & 120723 \\
``St. Paul's Cathedral'' (SPC) &  E &615 & 98872 \\
\midrule
Total E & & 3774 & 643k \\
\bottomrule
\end{tabular}
\end{center}
\caption{
\revision{{\bf Scenes in the IMC-PT dataset.} 
We provide some statistics for the training (T), validation (V), and test (E) scenes.}}
\label{tbl:sequences_trainval}
\end{table}
\edit{%
Out of the 25 scenes, containing almost 30k registered images in total, we select 2 for validation and 9 for testing. \revision{The remaining scenes can be used for training, if so desired, and are not used in this Chapter}.
We provide images, 3D reconstructions, camera poses, and depth maps, for every training and validation scene. 
For the test scenes we release only a subset of 100 images and keep the ground truth private, \revision{which is an integral part of the Image Matching Challenge. Results on the private test set can be obtained by sending submissions to the organizers, who process them.}

\revision{The scenes used for training, validation and test are listed in \tbl{sequences_trainval}},
along with the acronyms used in several figures.
For the validation experiments
-- Sections \ref{imc:ablation} and \ref{imc:appendix} -- 
we choose
two of the larger scenes, ``Sacre Coeur'' and ``St. Peter's Square'', which in our experience 
\edit{provide results that are quite representative of what should be expected on \revision{the Image Matching Challenge Dataset}.}
These two subsets have been released, so that the validation results are reproducible and comparable.
They can all be downloaded from the challenge website\footnoteref{website}.
}
\begin{figure}[htb]%
\centering
\renewcommand{\imscl}{0.01}
\includegraphics[scale=\imscl,width=0.5\linewidth]{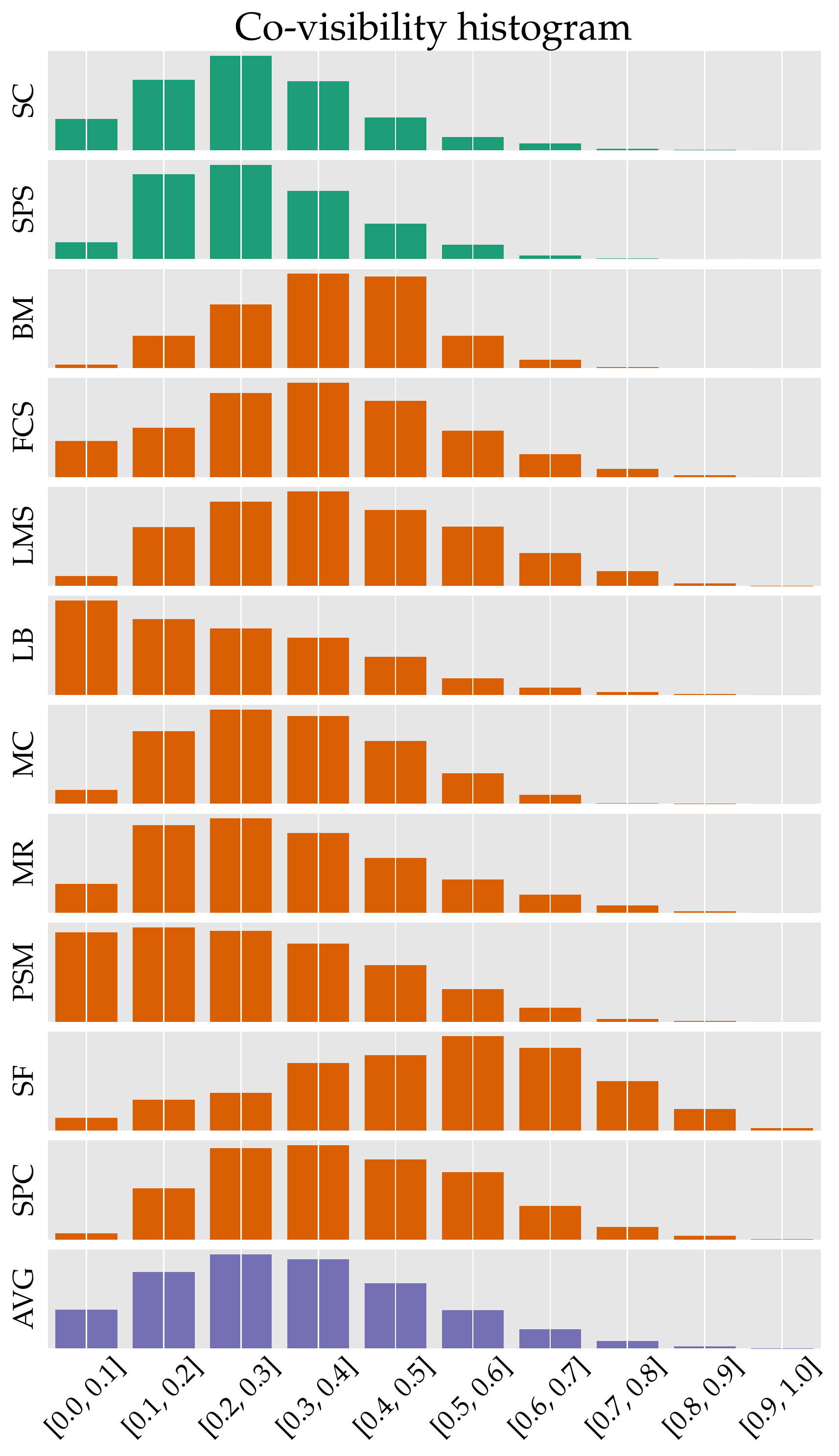}
\caption{{\bf Co-visibility histogram.} 
We break down the co-visibility measure for each scene in the validation (green) and test (red) sets, as well as the average (purple). 
Notice how the statistics may vary significantly from scene to scene.}
\label{fig:visibility_histograms}
\end{figure}

\begin{figure*}
\centering
\renewcommand{\imsize}{5.0cm}
\renewcommand{\hspacer}{0.3mm}
\renewcommand{\vspacer}{0.1mm}
\renewcommand{\imscl}{0.01}
\includegraphics[scale=\imscl,height=\imsize]{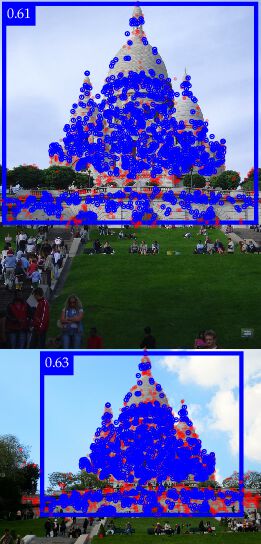}\hspace{\hspacer}%
\includegraphics[scale=\imscl,height=\imsize]{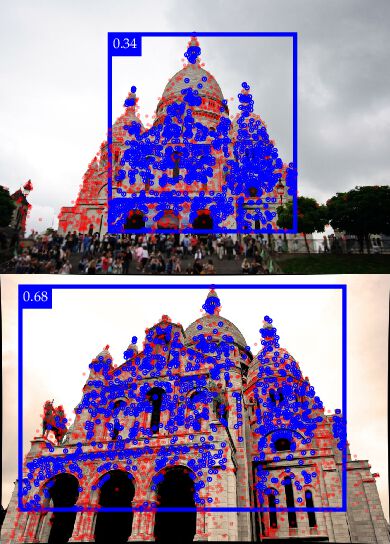}\hspace{\hspacer}%
\includegraphics[scale=\imscl,height=\imsize]{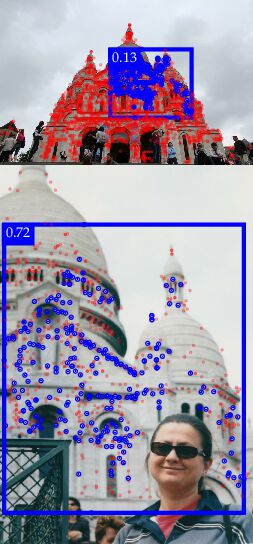}\hspace{\hspacer}%
\includegraphics[scale=\imscl,height=\imsize]{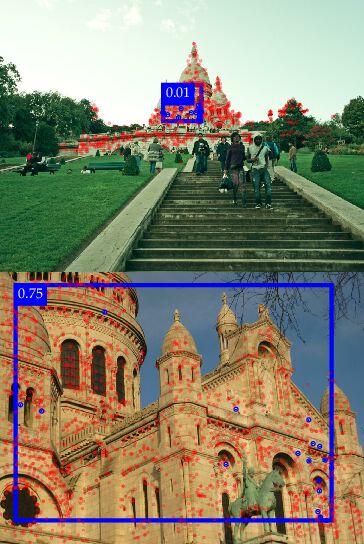}\hspace{\hspacer}\\
\vspace{\vspacer}%
\vspace{1ex}%
\renewcommand{\imsize}{4.5cm}%
\includegraphics[scale=\imscl,height=\imsize]{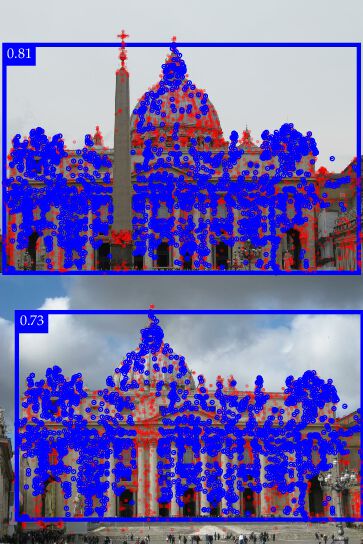}\hspace{\hspacer}%
\includegraphics[scale=\imscl,height=\imsize]{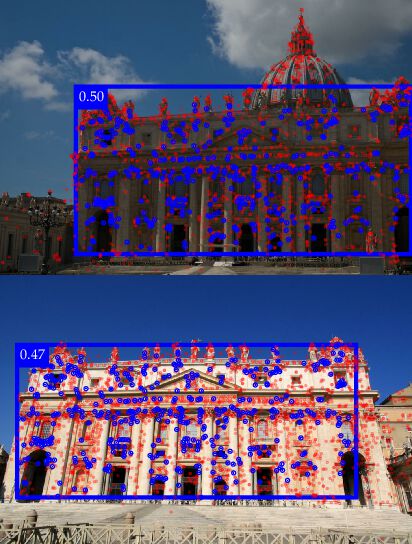}\hspace{\hspacer}%
\includegraphics[scale=\imscl,height=\imsize]{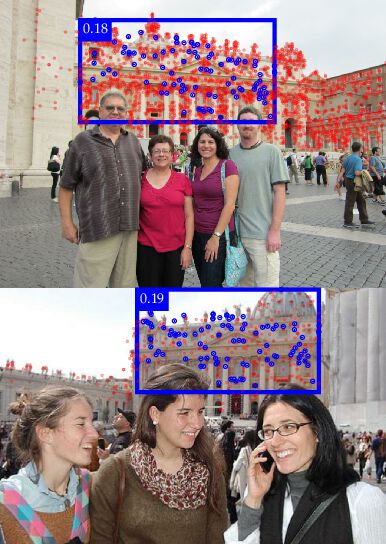}\hspace{\hspacer}%
\includegraphics[scale=\imscl,height=\imsize]{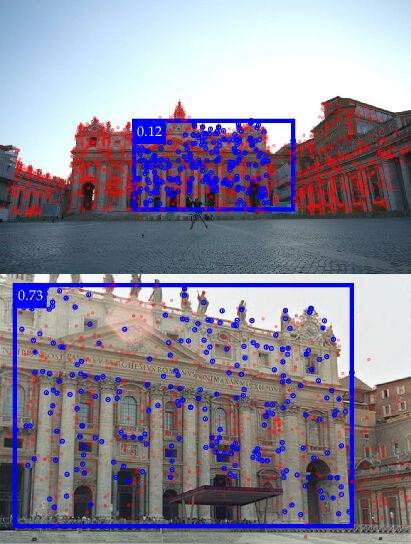}\hspace{\hspacer}%
\caption{{\bf Co-visibility examples }-- image pairs from validation scenes ``Sacre Coeur'' (top) and ``St. Peter's Square'' (bottom). Image keypoints that are part of the 3D reconstruction are blue if they are co-visible in both images, are red otherwise. The bounding box of the co-visible points, used to compute a per-image co-visibility ratio, is shown in blue. The co-visibility value for the image pair is the lower of these two values.
Examples include different `difficulty' levels. All of the pairs are used in the evaluation except the top-right one, as we set a cut-off at 0.1.
}
\label{fig:visibility_examples}
\vspace{-2mm}
\end{figure*}

\subsection{\edit{Estimating the co-visibility between two images}}%
\label{imc:covis}
\edit{%
\edit{When evaluating image pairs, one has to be sure that two given images share a common part of the scene} --
they may be registered by the SfM reconstruction without having any pixels in common as long as other images act as a `bridge' between them.}
Co-visible pairs of images are determined with a simple heuristic.
3D model points, detected in both the considered images, are localised in 2D in each image. The bounding box around the keypoints is estimated. The ratio of the bounding box area and the whole image is the ``visibility'' of image $i$ in $j$ and $j$ in $i$ respectively;  \edit{see~\fig{visibility_examples} for examples}.
The minimum of the two ``visibilities'' is the ``co-visibility'' ratio  $\covisibility_{i,j} \in [0,1]$, which characterizes how challenging matching of the particular image pair is.
The co-visibility varies significantly from scene to scene. The histogram of co-visibilities is shown in \fig{visibility_histograms}, providing insights into
how ``hard'' each scene is -- without accounting for some occlusions.

For stereo, the minimum co-visibility threshold is set to 0.1. 
For the multi-view task, subsets where at least 100 3D points are visible in each image, are selected, as in~\cite{CNe2018,Zhang19}.
We find that both criteria work well in practice.

\subsection{On the quality of the ``ground-truth''}
\label{imc:eval_gt}

Our core assumption is that accurate poses can be obtained from large sets of images without human intervention. Such poses are used as the ``ground truth'' for evaluation of image matching performance on pairs or small subsets of images -- a harder, proxy task.
Should this assumption hold, the (relative) poses retrieved with a large enough number of images would not change as more images are added, and these poses would be the same regardless of which local feature is used.
\begin{table}
\begin{center}
\scriptsize
\setlength\tabcolsep{4pt} %
\begin{tabular}{l ccccc}
\toprule
\multirow{2}{*}{Local featured type} & \multicolumn{4}{c}{Number of Images}\\
\cmidrule{2-6}
& 100 & 200 & 400 & 800 & all \\
\midrule
SIFT~\cite{SIFT2004}&0.06$^\circ$&0.09$^\circ$&0.06$^\circ$&0.07$^\circ$&0.09$^\circ$\\
SIFT (Upright)~\cite{SIFT2004}& 0.07$^\circ$& 0.07$^\circ$& 0.04$^\circ$& 0.06$^\circ$& 0.09$^\circ$\\
HardNet (Upright)~\cite{HardNet2017}& 0.06$^\circ$& 0.06$^\circ$& 0.06$^\circ$& 0.04$^\circ$& 0.05$^\circ$\\
SuperPoint~\cite{SuperPoint2017}& 0.31$^\circ$& 0.25$^\circ$& 0.33$^\circ$& 0.19$^\circ$& 0.32$^\circ$\\
R2D2~\cite{R2D22019}& 0.12$^\circ$& 0.08$^\circ$& 0.07$^\circ$& 0.08$^\circ$& 0.05$^\circ$\\
\bottomrule
\end{tabular}
\end{center}
\caption{
\revision{{\bf Standard deviation of the pose difference of three COLMAP runs with different number of images}. Most of them are below 0.1$^\circ$, except for SuperPoint.}
}
\label{tbl:colmap_std}
\end{table}
\begin{table}
\begin{center}
\scriptsize
\setlength\tabcolsep{2pt} %
\begin{tabular}{lcccc}
\toprule
\multirow{2}{*}{Local feature type} & \multicolumn{4}{c}{Number of images}\\
\cmidrule{2-5}
& 100 vs. all & 200 vs. all & 400 vs. all & 800 vs. all \\
\midrule
SIFT~\cite{SIFT2004} & 0.58$^\circ$ / 0.22$^\circ$ & 0.31$^\circ$ / 0.08$^\circ$ & 0.23$^\circ$ / 0.05$^\circ$ & 0.18$^\circ$ / 0.04$^\circ$\\
SIFT (Upright)~\cite{SIFT2004} & 0.52$^\circ$ / 0.16$^\circ$ & 0.29$^\circ$ / 0.08$^\circ$ & 0.22$^\circ$ / 0.05$^\circ$ & 0.16$^\circ$ / 0.03$^\circ$\\
HardNet (Upright)~\cite{HardNet2017} & 0.35$^\circ$ / 0.10$^\circ$ & 0.33$^\circ$ / 0.08$^\circ$ & 0.23$^\circ$ / 0.06$^\circ$ & 0.14$^\circ$ / 0.04$^\circ$\\
SuperPoint~\cite{SuperPoint2017} & 1.22$^\circ$ / 0.71$^\circ$ & 1.11$^\circ$ / 0.67$^\circ$ & 1.08$^\circ$ / 0.48$^\circ$ & 0.74$^\circ$ / 0.38$^\circ$\\
R2D2~\cite{R2D22019} & 0.49$^\circ$ / 0.14$^\circ$ & 0.32$^\circ$ / 0.10$^\circ$ & 0.25$^\circ$ / 0.08$^\circ$ & 0.18$^\circ$ / 0.05$^\circ$\\
\bottomrule
\end{tabular}
\end{center}
\caption{
\edit{{\bf \edit{Pose convergence in SfM.}}
\edit{We report the mean/median of the difference (in degrees) between the poses extracted with the full set of 1179 images for ``Sacre Coeur'', and different subsets of it, for \revision{four} local feature methods -- to keep the results comparable we only look at the 100 images in common across all subsets. 
We report the maximum among the angular difference between rotation matrices and translation vectors. 
The estimated poses are stable, with as little as 100 images.}
}}
\label{tbl:conv_gt}
\end{table}
\begin{table}
\begin{center}
\scriptsize
\setlength\tabcolsep{3pt} %
\begin{tabular}{c cccc}
\toprule
\multirow{2}{*}{Reference} & \multicolumn{4}{c}{Compared to}\\
\cmidrule{2-5}
& SIFT (Upright) & HardNet (Upright) & SuperPoint & R2D2 \\
\midrule
SIFT~\cite{SIFT2004} & 0.20$^\circ$ / 0.05$^\circ$ & 0.26$^\circ$ / 0.05$^\circ$ & 1.01$^\circ$ / 0.62$^\circ$ & 0.26$^\circ$ / 0.09$^\circ$ \\
\bottomrule
\end{tabular}
\end{center}
\caption{
{\bf Difference between poses obtained with different local features.}
We report the mean/median of the difference (in degrees) between the poses extracted with SIFT (Upright), HardNet (Upright), SuperPoint, or R2D2, and those extracted with SIFT.
We use the maximum of the angular difference between rotation matrices and translation vectors.
SIFT (Upright), HardNet (Upright), and R2D2 give near-identical results to SIFT.}
\label{tbl:val_gt}
\end{table}
To validate this, we pick the scene ``Sacre Coeur'' and compute SfM reconstructions with a varying number of images: 100, 200, 400, 800, and 1179 images (the entire ``Sacre Coeur'' dataset), where each set contains the previous one; new images are being added and no images are removed. 
\revision{We run each reconstruction three times, and report the average result of the three runs, to account for the variance inside COLMAP. The standard deviation among different runs is reported in \tbl{colmap_std}.}
Note that these reconstructions are only defined up to a scale factor -- we do not know the absolute scale that could be used to compare the reconstructions against each other. That is why we use a simple, pairwise metric instead. We 
\edit{pick all the pairs out of the 100 images present in the smallest subset, and compare how much their relative pose change with respect to their counterparts reconstructed using the entire set -- we do this for every subset, \ie, 100, 200, etc.}
\edit{Ideally, we would like the differences between the relative poses to approach zero as more images are added.}
\edit{We list the results in \tbl{conv_gt}, for different local feature methods.}
\edit{Notice how the poses converge, especially in terms of the median, as more images are used, for all methods -- and that the reconstructions using only 100 images are already very stable.
For SuperPoint we use a smaller number of features (2k per image), which is not enough to achieve pose convergence, but the error is still reduced as more images are used.}

\begin{figure}
\centering
\renewcommand{\arraystretch}{0.2}
\renewcommand{\tabcolsep}{1mm}
\footnotesize
\centering
\renewcommand{\imsize}{1.28cm}
\begin{tabular}{@{}ccc@{}}
\includegraphics[scale=\imscl,width=0.33\linewidth,height=0.5\linewidth]{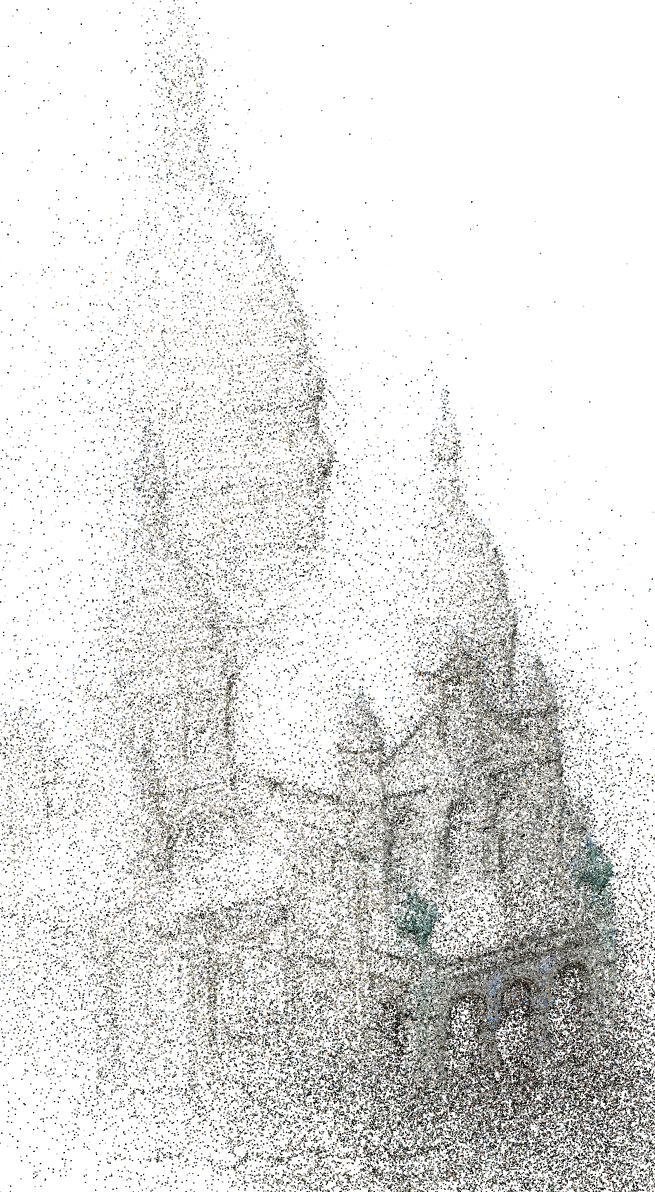} &
\includegraphics[scale=\imscl,width=0.33\linewidth,height=0.5\linewidth]{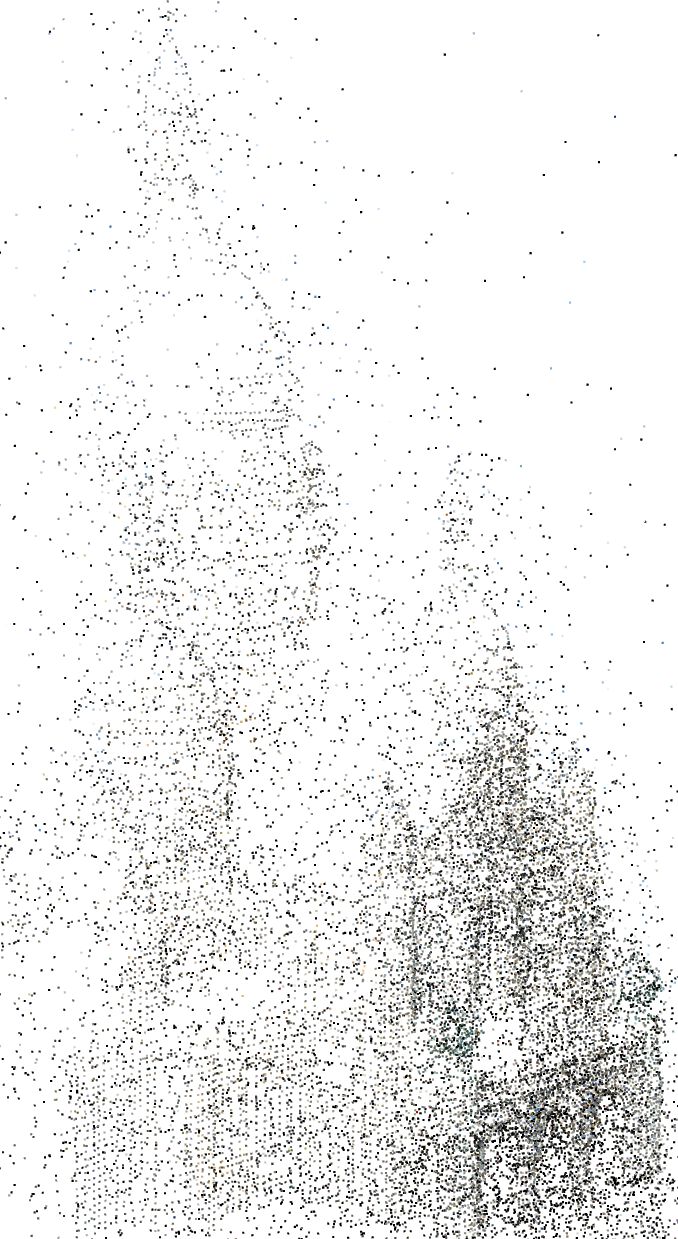} &
\includegraphics[scale=\imscl,width=0.33\linewidth,height=0.5\linewidth]{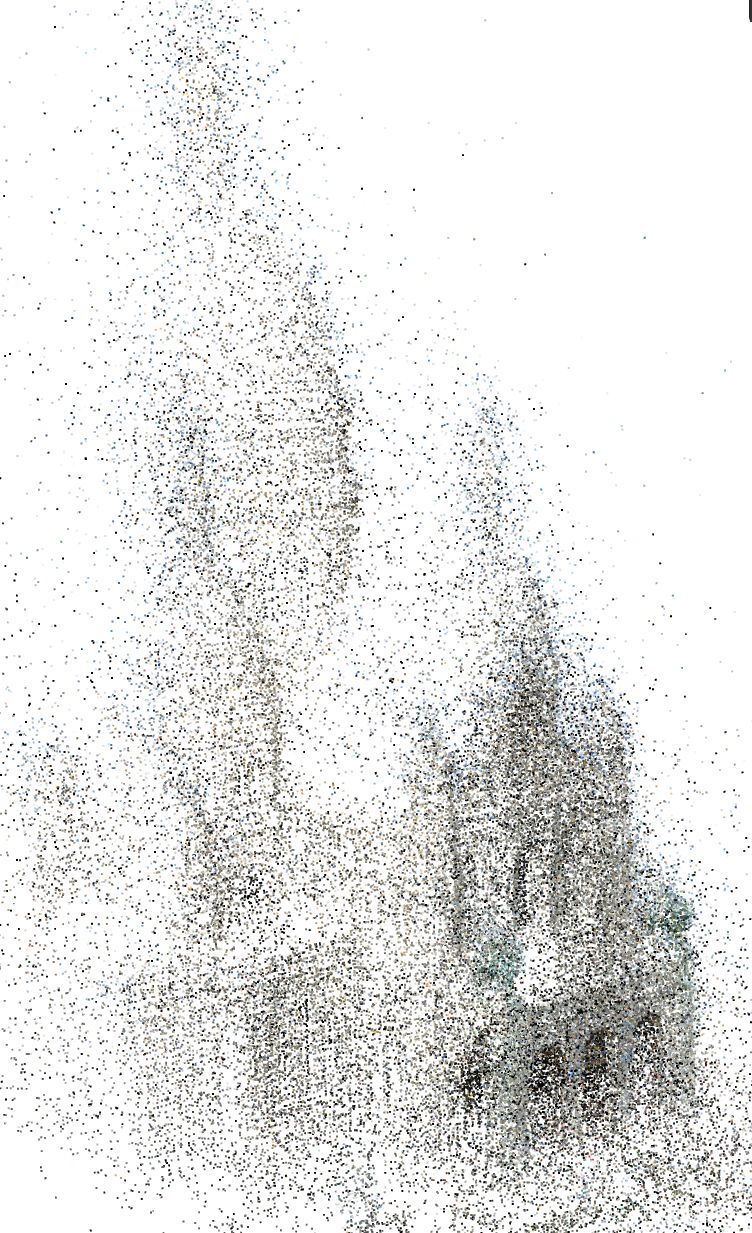} \\
\vspace{1mm} \\
(a) SIFT & (b) SuperPoint & (c) R2D2 \\
\end{tabular}
\caption{\edit{{\bf COLMAP with different local features.} We show the reconstructed point cloud for the scene ``Sacre Coeur'' using three different local features: SIFT, SuperPoint, and R2D2, using all images available (1179). The reconstructions with SIFT and R2D2 are both dense, albeit somewhat different. The reconstruction with SuperPoint is quite dense, considering it can only extract a much smaller number of features effectively, but its poses appear less accurate}.}
\label{fig:pointcloud}
\end{figure}

\begin{figure*}[bth]
\centering
\renewcommand{\imsize}{3.8cm}
\renewcommand{\hspacer}{0.3mm}
\renewcommand{\vspacer}{0.1mm}
\renewcommand{\imscl}{0.01}
\includegraphics[scale=\imscl,height=\imsize]{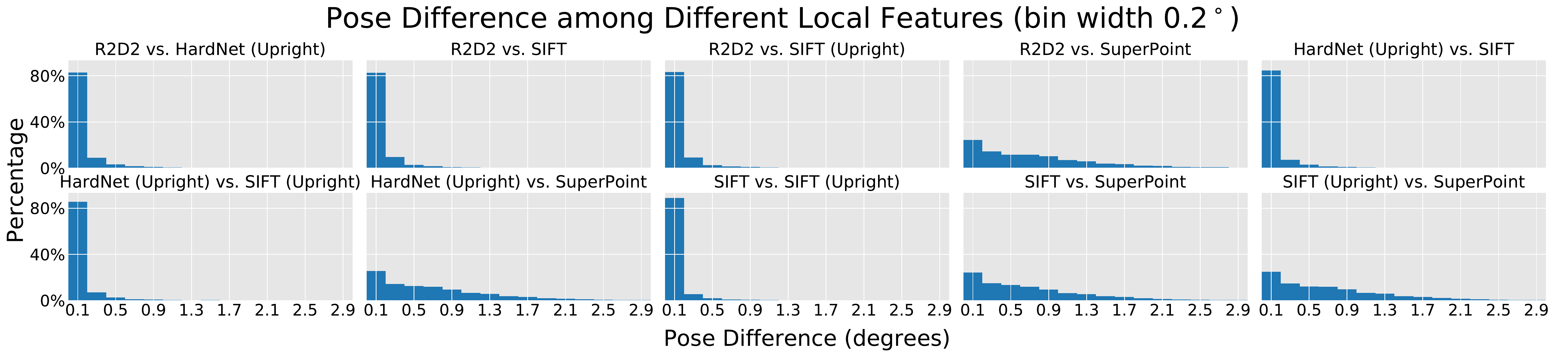}
\caption{\revision{
{\bf Histograms of pose differences between reconstructions with different local feature methods.}
We consider five different local features -- including rotation-sensitive and upright SIFT -- resulting in 10 combinations. The plots show that about 80\% percent of image pairs are within a 0.2$^o$ pose difference, with the exception of those involving SuperPoint. 
}}
\label{fig:colmap_hist}
\end{figure*}

\edit{%
We conduct a second experiment to verify that there is no bias towards using SIFT features for obtaining the ground truth.
We compare the poses obtained with SIFT to those obtained with other} local features -- note that our primary metric uses nothing but the {\em estimated poses} for evaluation.
We report results in \tbl{val_gt}. \revision{We also show the histogram of pose differences in ~\fig{colmap_hist}}.
The differences in pose %
\revision{due to the use of different local features have a median value below 0.1$^\circ$.
In fact, the pose variation of individual reconstructions with SuperPoint is of the same magnitude as the difference between reconstructions from SuperPoint and other local features: see \tbl{conv_gt}.
We conjecture that the reconstructions with SuperPoint, which does not extract a large number of keypoints, are less accurate and stable.}
This is further supported by the fact that the point cloud obtained with the entire scene generated with SuperPoint is less dense (125K 3D points) than the ones generated with SIFT (438K) or R2D2 (317k); see~\fig{pointcloud}.
\revision{In addition, we note that SuperPoint keypoints have been shown to be less accurate when it comes to precise alignment~\cite[Table~4, $\epsilon=1$]{SuperPoint2017}}. 
\edit{Note also that the poses from R2D2 are nearly identical to those from SIFT.}

\begin{figure*}[bthp]
\centering
\includegraphics[width=0.8\linewidth, trim= 80 250 170 15, clip]{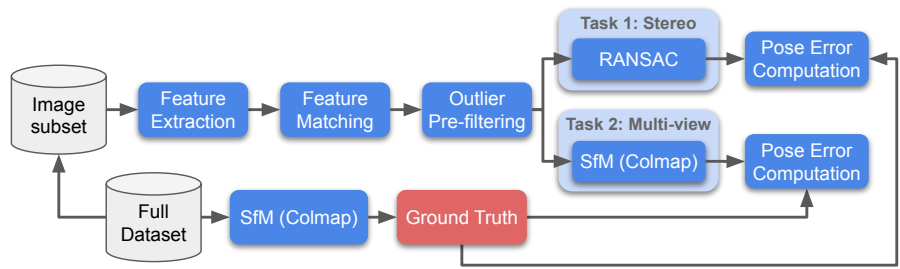}
\caption{
{\bf The benchmark pipeline} takes \edit{a subset of} $N$ images of a scene as input, extracts features for each,
and computes matches for all $M$ image pairs, $M=\frac{1}{2}N(N-1)$.
After an optional filtering step, the matches are fed to two different tasks.
Performance is measured {\em downstream}, by a
pose-based metric, common across tasks. \edit{The ground truth is extracted once, on the full set of images.}
}
\label{fig:pipeline}
\vspace{-3mm}
\end{figure*}

These observations reinforce
\edit{our trust}
in the accuracy of
\edit{our ground truth --
given a sufficient number of images, the choice of local feature is irrelevant, at least for the purpose of retrieving accurate poses.}
Our evaluation considers pose errors of up to 10$^\circ$, at a resolution of 1$^\circ$ -- significantly smaller than the fluctuations observed here, which we consider negligible.
Note that these conclusions may not hold on large-scale SfM requiring loop closures, but our dataset contains landmarks, which do not suffer from this problem.

In addition, we note that the use of dense, ground truth depth from SfM, \revision{which is arguably less accurate that camera poses}, has been verified by multiple parties, for training and evaluation, including:
CNe~\cite{CNe2018}, DFE~\cite{Ranftl18}, LF-Net~\cite{LFNet2018}, D2-Net~\cite{D2Net2019}, LogPolarDesc~\cite{LogPolar2019}, OANet~\cite{OANet2019}, and SuperGlue~\cite{sarlin2019superglue} among others, suggesting it is sufficiently accurate -- several of these rely on the data used in our work.

\edit{As a final observation, while the poses are {\em stable}, they could still be {\em incorrect}. 
This can happen on highly symmetric structures: for instance, a tower with a square or circular cross section. 
In order to prevent such errors from creeping into our evaluation, we visually inspected all the images in our test set. 
Out of 900 of them, we found 4 misregistered samples, all of them from the same scene, ``London Bridge'', which were removed from our data.
}

\section{Pipeline}
\label{imc:benchmark}
We outline our pipeline in \fig{pipeline}.
It takes as input $\numim{=}100$ images per scene.
The feature extraction module computes up to $\numfeat$ features from each image.
The
feature matching
module generates a list of putative matches for each image pair, \ie{} $\frac{1}{2}N(N-1)=4950$ combinations.
These matches can be optionally processed by an
outlier pre-filtering
module.
They are then fed to two tasks: stereo, and multiview reconstruction with SfM.
We now describe each of these components in detail.

\subsection{Feature extraction}
We consider three broad families of local features.
The first includes %
full, ``classical'' pipelines, most of them handcrafted:
SIFT~\cite{SIFT2004} (and RootSIFT~\cite{RootSIFT2012}), SURF~\cite{Bay2006}, ORB~\cite{ORB2011}, and AKAZE~\cite{AKAZE2013}.
We also consider FREAK~\cite{FREAK2012} descriptors with BRISK~\cite{Brisk2011} keypoints. We take these from OpenCV.
For all of them, except ORB, we lower the detection threshold to extract more features, which increases performance \edit{when operating with a large feature budget}.
We also consider DoG alternatives from VLFeat \cite{vedaldi2010vlfeat}: 
(VL-)DoG, Hessian~\cite{Hessian78}, Hessian-Laplace~\cite{HesHarAff2004}, Harris-Laplace~\cite{HesHarAff2004}, MSER~\cite{MSER2002}; and their affine-covariant versions: DoG-Affine, Hessian-Affine \cite{HesHarAff2004,Baumberg2000}, DoG-AffNet \cite{AffNet2018}, and Hessian-AffNet \cite{AffNet2018}.

The second group includes descriptors learned on DoG keypoints: L2-Net~\cite{L2Net2017},
Hardnet~\cite{HardNet2017}, Geodesc~\cite{geodesc2018}, SOSNet~\cite{SoSNet2019}, ContextDesc~\cite{ContextDesc2019}, and LogPolarDesc~\cite{LogPolar2019}.

The last group consists of pipelines learned end-to-end (e2e): Superpoint~\cite{SuperPoint2017}, LF-Net~\cite{LFNet2018}, D2-Net~\cite{D2Net2019} (with both single- (SS) and multi-scale (MS) variants), and R2D2~\cite{R2D22019}. Additionally, we consider Key.Net~\cite{KeyNet2019}, a learned detector paired with HardNet \revision{and SOSNet} descriptors -- we pair it with original implementation of HardNet instead than the one provided by the authors, as it performs better\footnote{In~\cite{KeyNet2019} the models are converted to TensorFlow -- we use the original PyTorch version.}.

\textbf{Post-IJCV update.} We have added 3 more local features to the benchmark after the official publication of the paper -- DoG-AffNet-HardNet~\cite{HardNet2017, AffNet2018}, DoG-TFeat~\cite{TFeat2016} and DoG-MKD-Concat~\cite{MKD2018}. We take them from the kornia~\cite{kornia2020} library. Results of this features are included into the Tables and Figures with a special mark $^{*}$.

\subsection{Feature matching}
We break this step into four stages.
Given images $\bI^i$ and $\bI^j$, $i \neq j$, we create an initial set of matches by nearest neighbor (NN) matching from $\bI^i$ to $\bI^j$, obtaining a set of matches $\mathbf{m}_{i \shortto j}$.
We optionally do the same in the opposite direction, $\mathbf{m}_{j \shortto i}$.
Lowe's ratio test \cite{SIFT2004} is applied to each list to filter out non-discriminative matches, with a threshold $\ratiotest \in [0,1]$, creating ``curated'' lists $\tilde{\mathbf{m}}_{i \shortto j}$ and $\tilde{\mathbf{m}}_{j \shortto i}$.
The final set of putative matches is the lists intersection, $\tilde{\mathbf{m}}_{i \shortto j} \cap \tilde{\mathbf{m}}_{j \shortto i} = \tilde{\mathbf{m}}^{\cap}_{i \leftrightarrow j}$ (known in the literature as one-to-one, mutual NN, bipartite, or cycle-consistent), or union $\tilde{\mathbf{m}}_{i \shortto j} \cup \tilde{\mathbf{m}}_{j \rightarrow i} = \tilde{\mathbf{m}}^{\cup}_{i \leftrightarrow j}$  (symmetric).
We refer to them as ``both'' and ``either'', respectively.
We also implement a simple unidirectional matching, \ie, $\tilde{\mathbf{m}}_{i \shortto j}$. 
Finally, the distance filter is optionally applied, removing matches whose distance is above a threshold.

The ``both'' strategy is similar to the ``symmetrical nearest neighbor ratio'' (sNNR)~\cite{bellavia2020NewAboutSIFT}, proposed concurrently --
SNNR combines the nearest neighbor ratio in both directions into a single number by taking the harmonic mean, while our test takes the maximum of the two values.

\subsection{Outlier pre-filtering}
Context Networks \cite{CNe2018}, or CNe for short, proposed a method to find sparse correspondences with a permutation-equivariant deep network based on PointNet \cite{PointNet2017}, sparking a number of follow-up works \cite{Ranftl18,Dang18a,Zhao19,Zhang19,Sun19}.
We embed CNe into our framework.
It often works best when paired with RANSAC \cite{CNe2018,Sun19}, so we consider it as an {\em optional} pre-filtering step before RANSAC -- and apply it to both stereo and multiview.
As the published model was trained on one of our validation scenes, we re-train it on ``Notre Dame Front Facade'' and ``Buckingham Palace'', following CNe training protocol, \ie, with 2000 SIFT features, unidirectional matching, and no ratio test.
We evaluated the new model on the test set and observed that its performance is better than the one that was released by the authors.
It could be further improved by using different matching schemes, such as bidirectional matching, but we have not explored this in this paper and leave as future work.

We perform one additional, but necessary, change: CNe (like most of its successors) was originally trained to estimate the Essential matrix instead of the Fundamental matrix~\cite{CNe2018}, \ie, it assumes known intrinsics.
In order to use it within our setup, we normalize the coordinates by the size of the image instead of using ground truth calibration matrices.
This strategy has also been used in \cite{Sun19}, and has been shown to work well in practice.

\subsection{Stereo task}
\label{imc:task_stereo}
The list of tentative matches is given to a robust estimator, which estimates $\mathbf{F}_{i,j}$, the Fundamental matrix between $\bI_i$ and $\bI_j$.
In addition to (locally-optimized) RANSAC~\cite{RANSAC1981,LOransac2003}, as implemented in OpenCV~\cite{OpenCV}, and sklearn~\cite{sklearn}, we consider more recent algorithms \edit{with publicly available implementations}: DEGENSAC~\cite{Degensac2005}, GC-RANSAC~\cite{gcransac2018} and MAGSAC~\cite{magsac2019}. \revision{We use the original GC-RANSAC implementation\footnote{\url{https://github.com/danini/graph-cut-ransac/tree/benchmark-version}}, and not the most up-to-date version, which incorporates MAGSAC, DEGENSAC, and later changes.}

For DEGENSAC we additionally consider disabling the degeneracy check, which theoretically should be equivalent to the OpenCV and sklearn implementations -- we call this variant ``PyRANSAC''.
Given $\mathbf{F}_{i,j}$, the known intrinsics $\bK_{\{i,j\}}$ were used to compute the Essential matrix $\bE_{i,j}$, as
$\bE_{i,j} = \bK_j^T \bF_{i,j} \bK_i$. Finally, the relative rotation and translation vectors were recovered with a cheirality check with OpenCV's \texttt{recoverPose}.

\subsection{Multiview task}
\label{imc:task_multiview}
Large-scale SfM is notoriously hard to evaluate, as it requires accurate ground truth.
Since our goal is to benchmark {\em local features} and {\em matching methods}, and not SfM \edit{algorithms}, we opt for a different strategy.
We reconstruct a scene from small image subsets, which we call ``bags''.
We consider bags of 5, 10, and 25 images, which are randomly sampled from the original set of 100 images per scene -- with a co-visibility check.
We create 100 bags for bag sizes 5, 50 for bag size 10, and 25 for bag size 25 -- \ie, 175 SfM runs in total.

We use COLMAP~\cite{COLMAP2016}, feeding it the matches computed by the previous module -- note that this comes before the robust estimation step, as COLMAP implements its own RANSAC.
If multiple reconstructions are obtained, we consider the largest one.
We also collect and report statistics such as the number of landmarks or the average track length. Both statistics and error metrics are averaged over the three bag sizes, each of which is in turn averaged over its individual bags.

\subsection{Error metrics}
Since the stereo problem is defined up to a scale factor~\cite{Hartley2004}, our main error metric is based on {\em angular errors}. 
We compute the difference, in degrees, between the {\em estimated} and {\em ground-truth} translation and rotation {\em vectors} between two cameras. 
We then threshold it over a given value for all possible -- \ie, co-visible -- pairs of images.
\edit{Doing so over different angular thresholds renders a curve.
We compute the mean Average Accuracy (mAA) by integrating this curve up to a maximum threshold, which we set to 10$^\circ$ -- this is necessary because large errors always indicate a bad pose: 30$^\circ$ is not necessarily better than 180$^\circ$, both estimates are wrong.}
Note that by computing the area under the curve we are giving more weight to methods which are more accurate at lower error thresholds, compared to using a single value at a certain designated threshold.

This metric was originally introduced in \cite{CNe2018}, where it was called mean Average Precision (mAP). 
We argue that ``accuracy'' is the correct terminology, since we are simply evaluating how many of the predicted poses are ``correct'', as determined by thresholding over a given value -- \ie, our problem does not have
``false positives''.

The same metric is used for multiview. 
Because we do not know the scale of the scene {\em a priori}, it is not possible to measure the translation error in metric terms. 
While we intend to explore this in the future, such a metric, while more interpretable, is not without problems -- for instance, the range of the distance between the camera and the scene can vary drastically from scene to scene and make it difficult to compare their results.
To compute the mAA in pose estimation for the multiview task, we
take the mean of the average accuracy for every pair of cameras -- setting the pose error to $\infty$ for pairs containing unregistered views.
If COLMAP returns multiple models which cannot be co-registered (which is rare) we consider only the largest of them for simplicity.

For the stereo task, we can report this value for different co-visibility thresholds: we use $\covisibility = 0.1$ by default, which preserves most of the ``hard'' pairs. 
Note that this is not applicable to the multiview task, as all images are registered at once via bundle adjustment in SfM.

\edit{Finally, we consider repeatability and matching score. 
Since many end-to-end methods do not report \edit{and often do not} have a clear 
measure of scale -- or support region -- we simply threshold by pixel distance, \edit{as in \cite{FAST2006}}.}
For the multiview task, we also compute the Absolute Trajectory Error (ATE)~\cite{Sturm12}, a metric widely used in SLAM. 
\edit{Since, once again,} the reconstructed model is
\edit{scale-agnostic}, we first scale the reconstructed model to that of the ground truth and then compute the ATE.
\edit{Note that ATE needs a minimum of three points to align the two models.}
\subsection{Implementation}
The benchmark code has been open-sourced\footnoteref{code}
along with every method used in the paper\footnote{{\footnotesize
\url{https://github.com/vcg-uvic/image-matching-benchmark-baselines}}}.
The implementation relies on SLURM~\cite{Slurm2003} for \edit{scalable} job scheduling, \edit{which is compatible with our supercomputer clusters} -- \edit{we also provide on-the-cloud, ready-to-go images}\footnote{{\footnotesize \url{https://github.com/etrulls/slurm-gcp}}}. 
The benchmark can also run on a standard computer, sequentially.
It is computationally expensive, as it requires matching about 45k image pairs.
The most costly step -- leaving aside feature extraction, which is very method-dependent -- is typically feature matching: 2--6 seconds per image pair\footnote{Time measured on ‘n1-standard-2’ VMs on Google Cloud Compute: 2 vCPUs with 7.5 GB of RAM and no GPU.}, depending on the descriptor size.
Outlier pre-filtering takes about 0.5--0.8 seconds per pair,
\edit{excluding some overhead to reformat the data into its expected format.}
RANSAC methods vary between 0.5--1 second -- as explained in \secref{ablation} we limit their number of iterations based on a compute budget, but the actual cost depends on the number of matches.
Note that these values are computed on the validation set -- for the test set experiments we increase the RANSAC budget, in order to remain compatible with the rules of \revision{the Image Matching Challenge\footnoteref{website}}.
We find COLMAP to vary drastically between set-ups.
New methods will be continuously added, and
\edit{we welcome contributions to the code base.}

\section{Details are Important}
\label{imc:ablation}

Our experiments indicate that each method needs to be carefully tuned.
In this section we outline the methodology we used to find the right hyperparameters on the validation set, and demonstrate why it is crucial to do so.

\begin{figure}
\centering
\renewcommand{\hspacer}{0.3mm}
\renewcommand{\vspacer}{0.1mm}
\renewcommand{\imscl}{0.01}
\includegraphics[scale=\imscl,width=\linewidth]{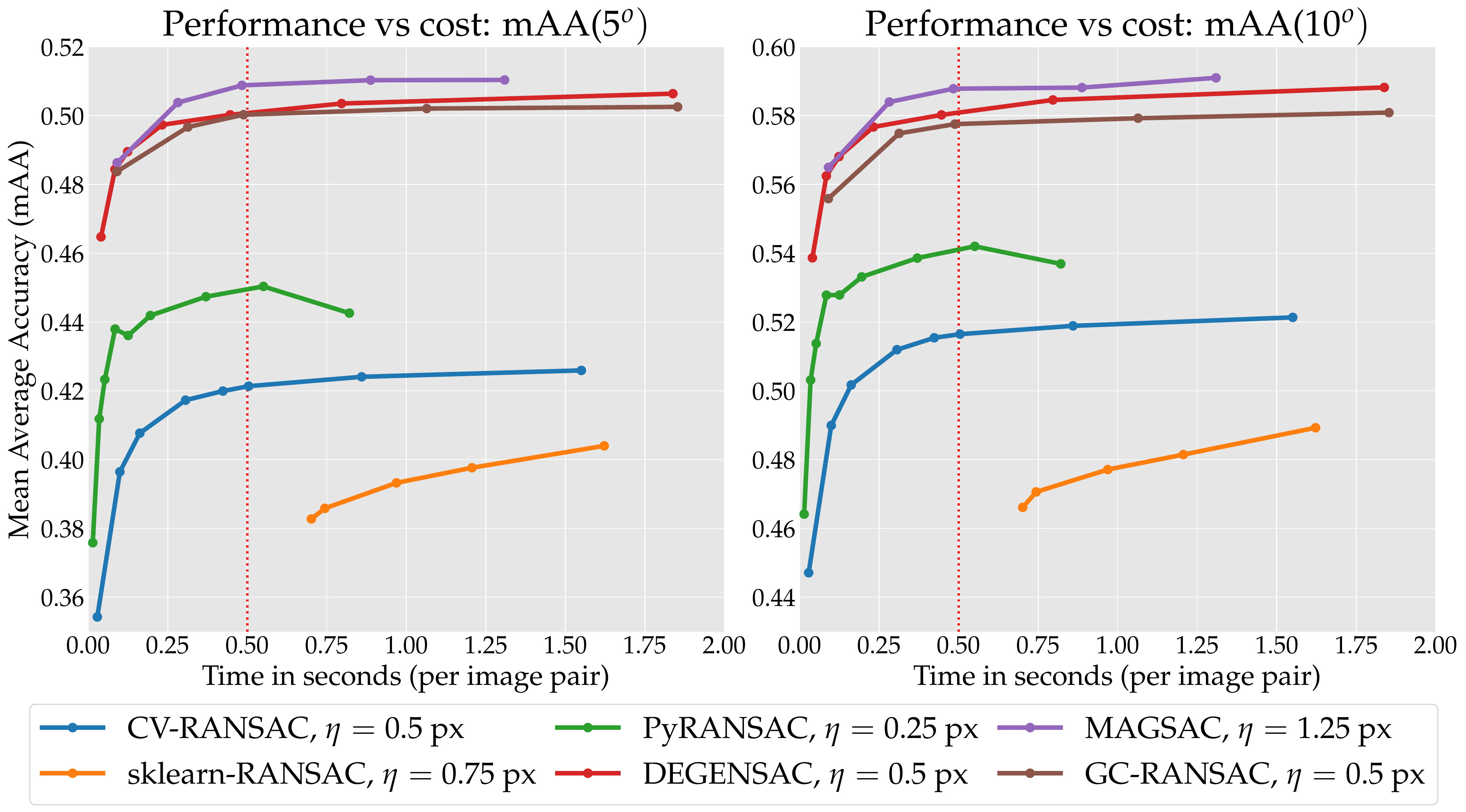}
\caption{{\bf Validation -- Performance vs. cost for RANSAC.}
\edit{We evaluate six RANSAC variants, using 8k SIFT features with ``both'' matching and a ratio test threshold of $\ratiotest{=}0.8$.
The inlier threshold $\ransacReprojTh$ and iterations limit $\ransacMaxIters$ are variables -- we plot only the best $\ransacReprojTh$ for each method, for clarity, and
set a budget of 0.5 seconds per image pair (dotted red line).
For each RANSAC variant, we pick the largest $\ransacMaxIters$ under this time ``limit'' and use it for all validation experiments.}
Computed on `n1-standard-2' VMs on Google Compute (2 vCPUs, 7.5 GB).
}
\label{fig:plots-ransac-cost-vs-map}
\end{figure}

\subsection{RANSAC: Leveling the field}
\label{imc:tuning_ransac}

Robust estimators are, \edit{in our experience,} the most sensitive part of the stereo pipeline, \edit{and thus the one we first turn to.}
All methods \edit{considered in this Section} have three parameters in common: the confidence level in their estimates, $\ransacConf$; the outlier (epipolar) threshold, $\ransacReprojTh$; and the maximum number of iterations, $\ransacMaxIters$.
\edit{We find the confidence value to be} the least sensitive, so we set it to $\ransacConf=0.999999$.

We evaluate each method with different values for $\ransacMaxIters$ and $\ransacReprojTh$, \edit{using reasonable defaults: 8k SIFT features with bidirectional matching with the ``both'' strategy and a ratio test threshold of 0.8.}
\edit{We plot the results in \fig{plots-ransac-cost-vs-map}, against their computational cost -- for the sake of clarity we only show the curve corresponding to the best reprojection threshold $\ransacReprojTh$ for each method}.

\edit{Our aim with this experiment} is to place all methods on an ``even ground'' by setting a common budget, \edit{as we need to find a way to compare them}.
We pick 0.5 seconds, where all methods have mostly converged. 
\edit{Note that these are different implementations and are obviously not directly comparable to each other, but this is a simple and reasonable approach}. 
\edit{We set this budget} by choosing
$\ransacMaxIters$ as per \fig{plots-ransac-cost-vs-map}, instead of actually {\em enforcing} a time limit, \edit{which would not be comparable across different set-ups.}
Optimal values for $\ransacMaxIters$ can vary drastically, from 10k for MAGSAC to 250k for PyRANSAC.
\edit{MAGSAC gives the best results} \edit{for this experiment, closely} followed by DEGENSAC.
We patch OpenCV
to increase the limit of iterations, which \edit{was} hardcoded to $\ransacMaxIters=1000$; this patch is now integrated into OpenCV.
This increases performance by 10-15\% relative, within our budget.
However, PyRANSAC is significantly better \edit{than OpenCV version even with this patch}, so we use it as our ``vanilla'' RANSAC instead.
The sklearn implementation is too slow for practical use.

We find that, in general, default settings can be woefully inadequate.
For example, OpenCV recommends $\ransacConf=0.99$ and $\ransacReprojTh=3$ pixels, which results in a mAA at 10$^\circ$ of \edit{0.3642} on the validation set -- a performance drop of \edit{29.3}\% relative.
\begin{figure}%
\centering
\renewcommand{\imsize}{5.2cm}
\renewcommand{\hspacer}{0.3mm}
\renewcommand{\vspacer}{0.1mm}
\renewcommand{\imscl}{0.01}
\includegraphics[scale=\imscl,width=\linewidth]{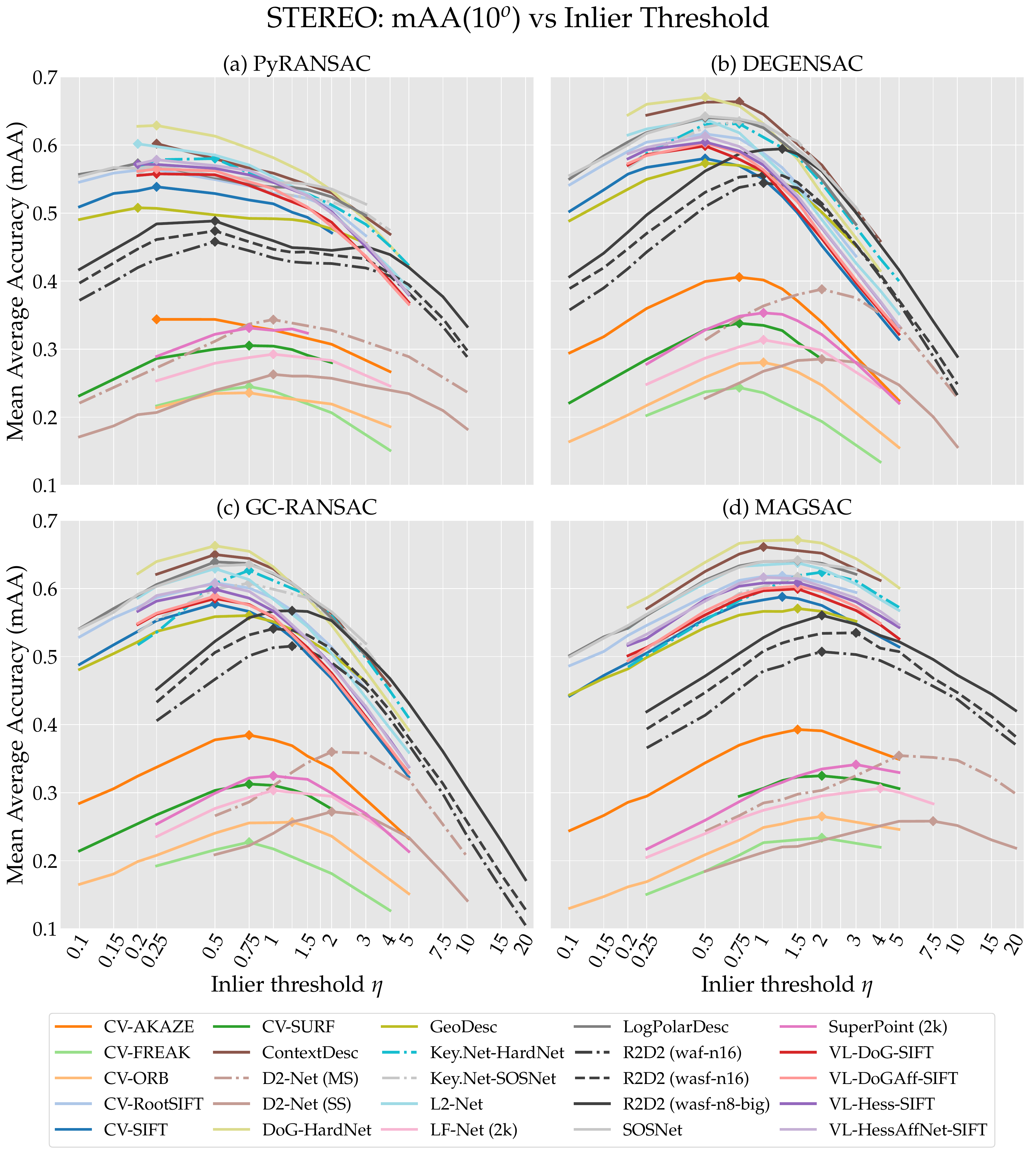}%
\caption{
{\bf Validation -- Inlier threshold for RANSAC, $\ransacReprojTh$.}
We determine $\ransacReprojTh$ for each combination, using 8k features
\edit{(2k for LF-Net and SuperPoint) 
with the ``both'' matching strategy and a reasonable value for the ratio test.}
Optimal parameters (diamonds) are listed in the \secref{appendix}.%
}
\label{fig:results_ransac_nort}
\vspace{-2mm}
\end{figure}

\subsection{RANSAC: One method at a time}
The last free parameter is the inlier threshold $\ransacReprojTh$.
We expect the optimal value for this parameter to
be different for each local feature, with looser thresholds required for methods operating on higher recall/lower precision, \edit{and end-to-end methods trained on lower resolutions}.

We report a wide array of experiments in~\fig{results_ransac_nort} that confirm our intuition: descriptors learned on DoG keypoints are clustered, while others
vary significantly.
Optimal values are also different for each RANSAC variant.
\edit{We use the ratio test with the threshold recommended by the authors of each feature, or a reasonable value if no recommendation exists, and the \edit{``both''} matching strategy -- this cuts down on the number of outliers.}

\subsection{Ratio test: One feature at a time}
\label{imc:matching}

\begin{figure}
\centering
\renewcommand{\hspacer}{0.3mm}
\renewcommand{\vspacer}{0.1mm}
\renewcommand{\imscl}{0.01}
\includegraphics[scale=\imscl,width=\linewidth]{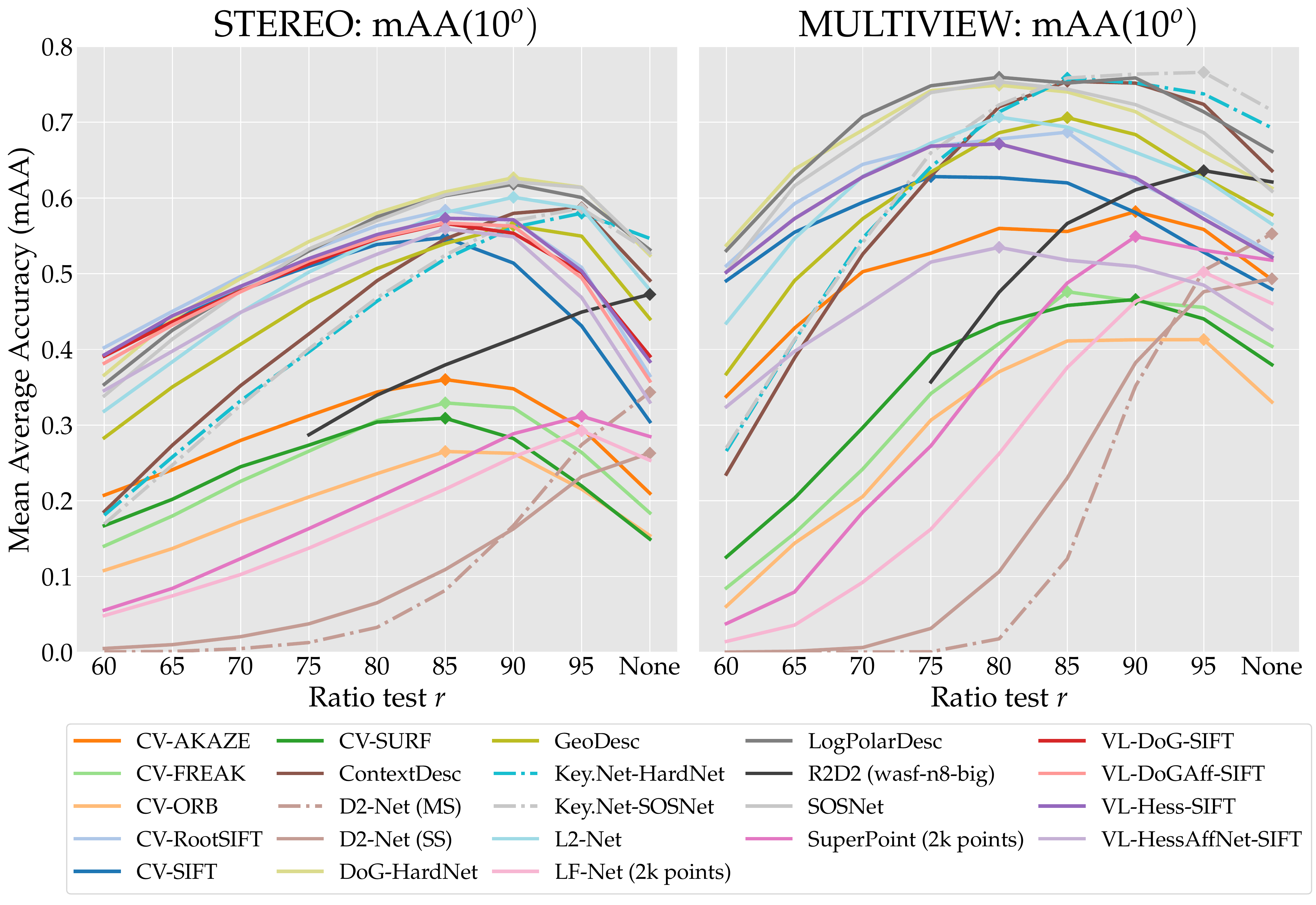} \\
\caption{
\edit{{\bf Validation -- Optimal ratio test $r$ for matching with ``both''.}
We evaluate bidirectional matching with the ``both'' strategy (the best
one), and different ratio test thresholds $r$, for each feature type.
We use 8k features (2k for SuperPoint and LF-Net).
For stereo, we use PyRANSAC.}
}
\label{fig:results_both_vs_any_and_ratio_test}
\vspace{0mm}
\end{figure}

\begin{figure}
\centering
\renewcommand{\hspacer}{0.3mm}
\renewcommand{\vspacer}{0.1mm}
\renewcommand{\imscl}{0.01}
\includegraphics[scale=\imscl,width=\linewidth]{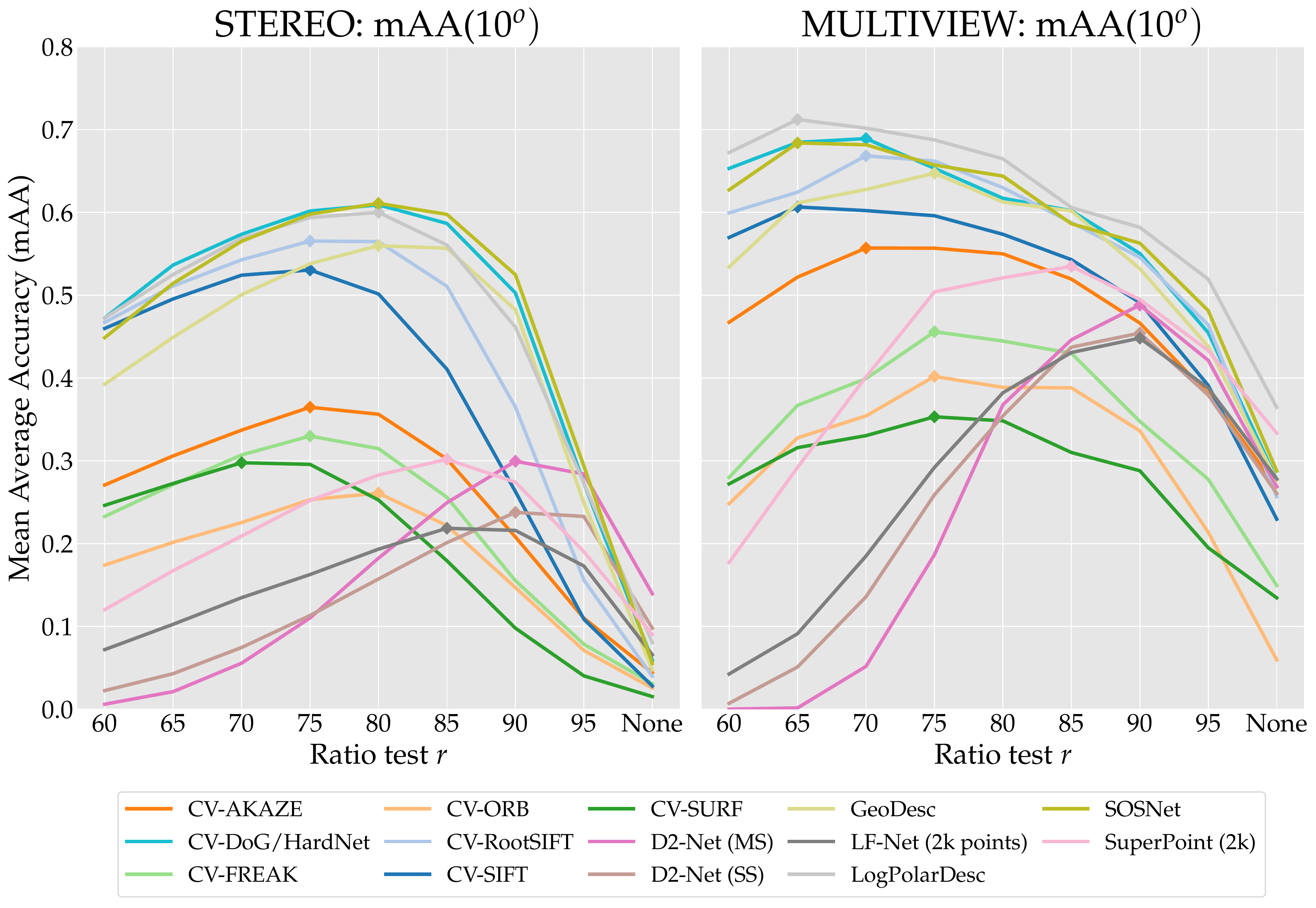} \\
\caption{
{\bf Validation -- Optimal ratio test $r$ for matching with ``either''.}
Equivalent to \fig{results_both_vs_any_and_ratio_test} but with the ``either'' matching strategy.
\edit{This strategy requires 
aggressive filtering and does not reach the performance of ``both'',
we thus explore only a subset of the methods.}
}
\label{fig:results_both_vs_any_and_ratio_test_extra}
\end{figure}

\edit{Having 
``frozen''
RANSAC, we turn to the feature matcher -- \edit{note that it comes {\em before} RANSAC, but it cannot be evaluated in isolation}.
\edit{We select PyRANSAC as a ``baseline'' RANSAC and evaluate different ratio test thresholds, separately for the stereo and multiview tasks.}
\edit{For this experiment, we use 8k features with all methods, except for those which cannot work on this regime -- SuperPoint and LF-Net. This choice will be substantiated in \secref{number-of-features}.}
We report the results for bidirectional matching with the ``both'' strategy in \fig{results_both_vs_any_and_ratio_test}, and with the ``either'' strategy in \fig{results_both_vs_any_and_ratio_test_extra}.
We find that ``both'' -- the method we have used so far -- performs best overall.
Bidirectional matching with the ``either'' strategy produces many (false) matches, increasing the computational cost in the estimator, and requires very small ratio test thresholds -- as low as $r{=}0.65$.
Our experiments with unidirectional matching indicate that is slightly worse, and it depends on the order of the images, so we did not explore it further.}

As expected, each feature requires different settings, as the distribution of their descriptors is different.
We also observe that optimal values vary significantly between stereo and multiview, even though one might expect that bundle adjustment should be able to better deal with outliers. 
We suspect \revision{that this indicates that there is room for improvement in COLMAP's implementation of RANSAC.}

Note how the ratio test is critical for performance, and one could arbitrarily select a threshold that favours one method over another, which shows the importance of proper benchmarking.
Interestingly, D2-Net is the {\em only} method that \edit{clearly} performs best without the ratio test. \edit{It also performs poorly overall in our evaluation, despite reporting state-of-the-art results in other benchmarks~\cite{MikoDescEval2003,hpatches2017,ActiveSearch2012,taira2018inloc} 
-- without the ratio test, the number of tentative matches might be too high for RANSAC or COLMAP to perform well.}

Additionally, we implement the first-geometric-inconsistent ratio threshold, or FGINN \cite{MODS2015}. 
We find that although it improves over unidirectional matching, its gains \edit{mostly} disappear against matching with ``both''. 
We report these results in \secref{fginn}.

\begin{figure}
\centering
\renewcommand{\hspacer}{0.3mm}
\renewcommand{\vspacer}{0.1mm}
\renewcommand{\imscl}{0.01}
\includegraphics[scale=\imscl,width=\linewidth]{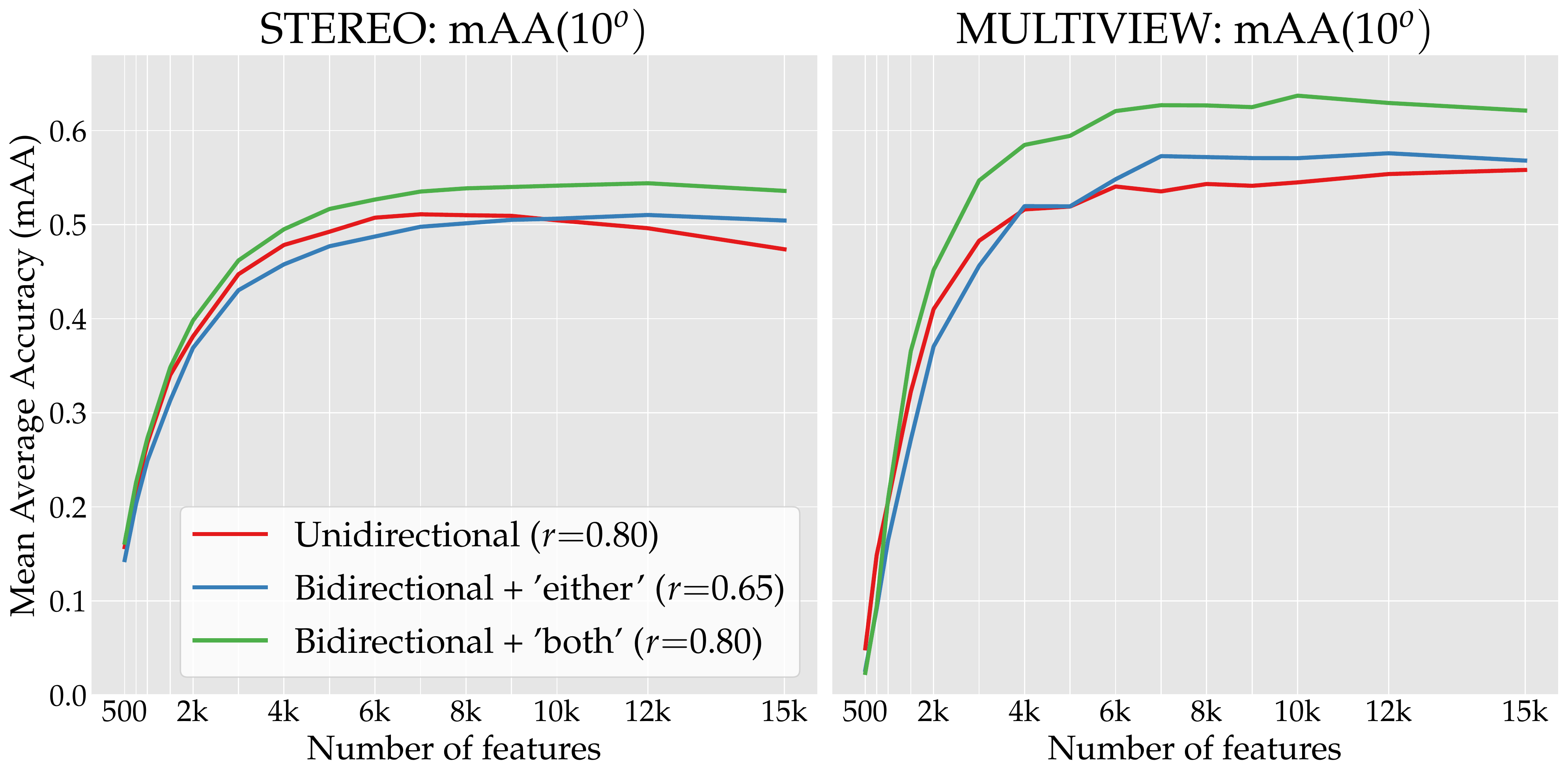} \\
\caption{
{\bf Validation -- Number of features.}
Performance on the stereo and multi-view tasks while varying the number of SIFT features, with three matching strategies, and \edit{reasonable defaults} for the ratio test $r$.
}
\label{fig:results_num_features_val}
\end{figure}

\subsection{Choosing the number of features}
\label{imc:number-of-features}

The ablation tests in this section use (up to) $K{=}8000$ feature (2k for SuperPoint and LF-Net, as they are trained to extract fewer keypoints). This number is commensurate with that used by SfM frameworks \cite{Wu13,COLMAP2016}.
\edit{We report performance for different values of $K$ in \fig{results_num_features_val}.}
We use PyRANSAC with \edit{reasonable defaults} \edit{for all three matching strategies, with SIFT features.}

As expected, performance is strongly correlated with the number of features.
We find 8k to be a good compromise between performance and cost, and also consider 2k (actually 2048) as a `cheaper' alternative -- this also provides a fair comparison with some learned methods which only operate on that regime.
\edit{We choose these two values as valid categories for the open challenge\footnoteref{website} linked to the benchmark, and do the same here for consistency.}

\subsection{Additional experiments}
\label{imc:additional-validation-experiments}

\edit{Some methods require additional considerations before evaluating them on the test set.
We briefly discuss them in this section.
Further experiments are available in \secref{appendix}.
}

\begin{figure}
\centering
\renewcommand{\imsize}{4.4cm}
\renewcommand{\hspacer}{0.1mm}
\renewcommand{\vspacer}{0.1mm}
\renewcommand{\imscl}{0.01}
\includegraphics[scale=\imscl,height=\imsize]{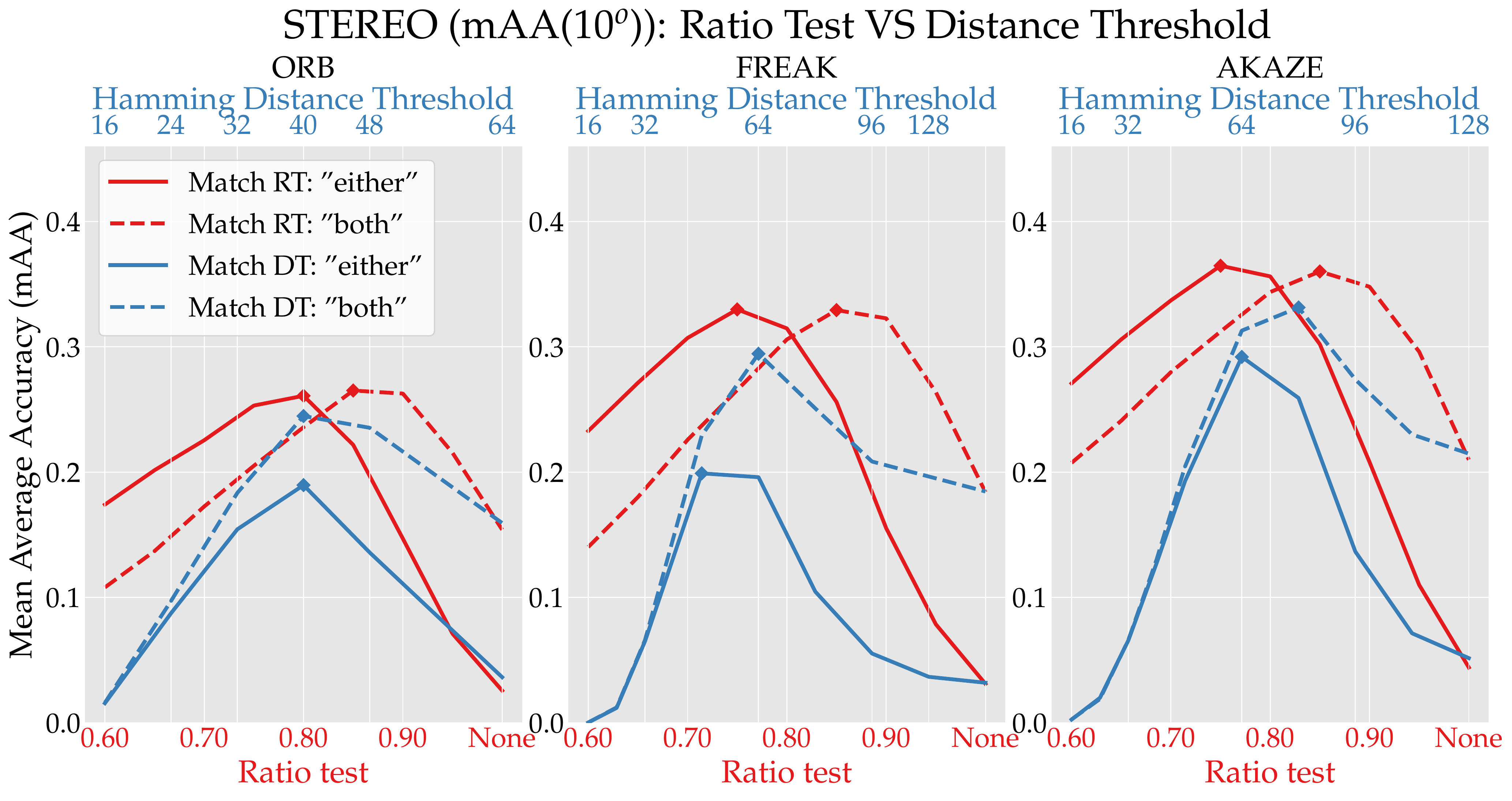}\hspace{\hspacer}%
\caption{
{\bf Validation -- Matching binary descriptors.}
We filter out non-discriminative matches with the ratio test or a distance threshold. The latter (the standard) performs worse in our experiments.%
}
\label{fig:matching_binary}
\end{figure}

\paragraph{Binary features (\fig{matching_binary}).}
\edit{We consider three binary descriptors: ORB \cite{ORB2011}, AKAZE \cite{AKAZE2013}, and FREAK \cite{FREAK2012}
}
Binary descriptor papers \edit{historically} favour a distance threshold in place of the ratio test to reject non-discriminative matches \cite{ORB2011}, although some papers have used the ratio test for ORB descriptors~\cite{saddle2019}.
We evaluate both in~\fig{matching_binary} -- \edit{as before, we use up to 8k features and matching with the ``both'' strategy}.
\edit{The ratio test works better for all three methods -- we use it instead of a distance threshold for all experiments in the Chapter, including those in the previous sections.}

\begin{figure}
\centering
\renewcommand{\imsize}{5.2cm}
\renewcommand{\hspacer}{0.3mm}
\renewcommand{\vspacer}{0.1mm}
\renewcommand{\imscl}{0.01}
\includegraphics[scale=\imscl,width=0.49\linewidth]{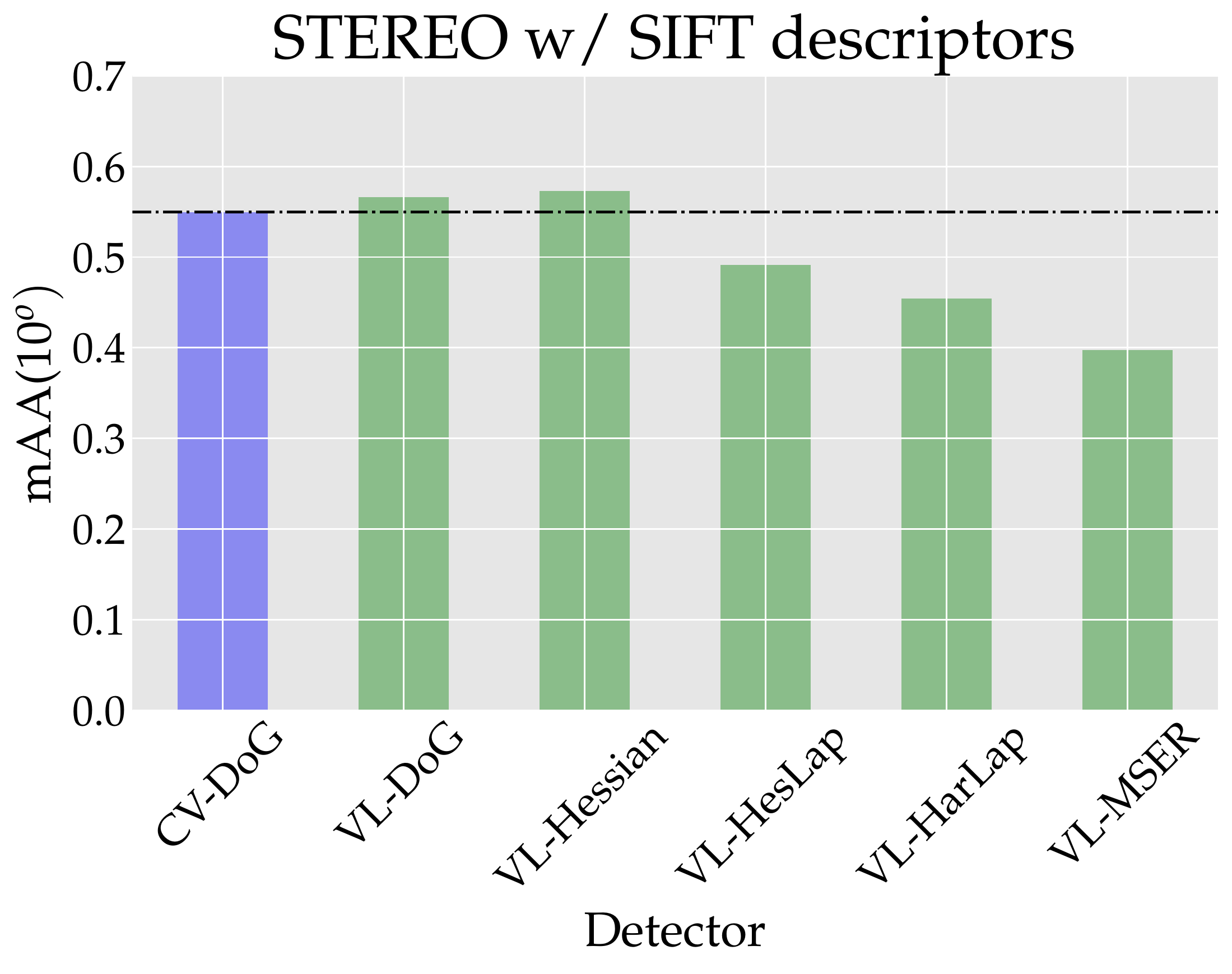}%
\hspace{\hspacer}%
\includegraphics[scale=\imscl,width=0.49\linewidth]{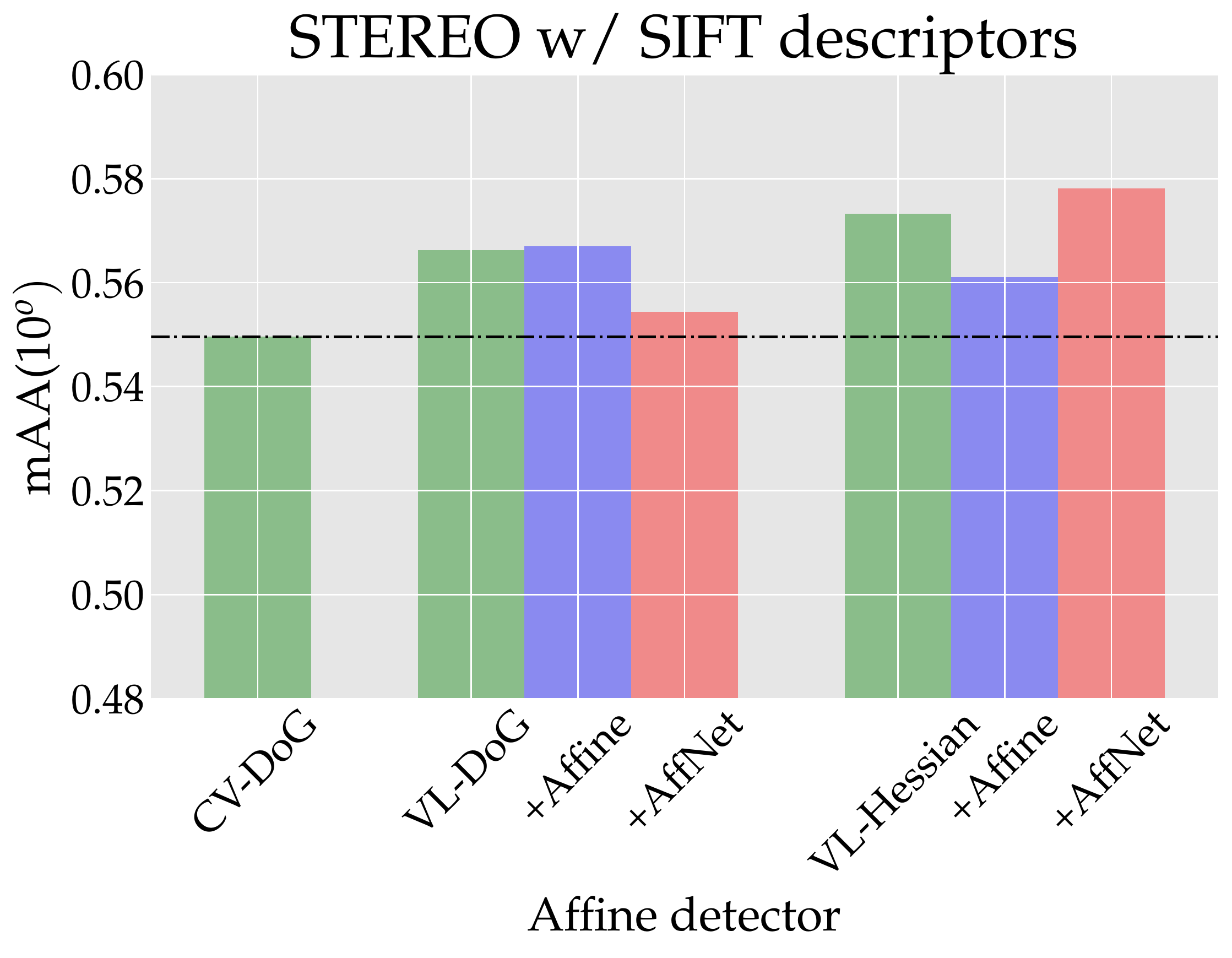}%
\caption{
{\bf Validation -- Benchmarking detectors.}
We evaluate the performance on the stereo task while pairing different detectors with SIFT descriptors.
The dashed, black line indicates OpenCV SIFT -- the baseline.
{\bf Left:} OpenCV DoG vs. VLFeat implementations of blob detectors (DoG, Hessian, HesLap) and corner detectors (Harris, HarLap), and MSER.
{\bf Right:} Affine shape estimation for DoG and Hessian keypoints, against the plain version. We consider a classical approach, Baumberg (Affine)~\cite{Baumberg2000}, and the recent, learned AffNet~\cite{AffNet2018}
-- they provide a small but inconsistent boost.
}
\label{fig:results_vlfeat}
\end{figure}
\paragraph{On the influence of the detector (\fig{results_vlfeat}).}
We %
embed several popular blob and corner detectors into our pipeline, with OpenCV's DoG \cite{SIFT2004} as a baseline.
We combine multiple methods, taking advantage of the VLFeat library:
\edit{Difference of Gaussians} (DoG), Hessian \cite{Hessian78}, HessianLaplace \cite{HesHarAff2004}, HarrisLaplace \cite{HesHarAff2004}, MSER \cite{MSER2002}, DoGAffine, Hessian-Affine \cite{HesHarAff2004,Baumberg2000}, DoG-AffNet \cite{AffNet2018}, and Hessian-AffNet \cite{AffNet2018}.
We pair them with SIFT descriptors, also computed with VLFeat, as OpenCV cannot process affine keypoints, and report the results in \fig{results_vlfeat}.
VLFeat's DoG performs marginally better than OpenCV's.
Its affine version gives a small boost.
Given the small gain and the infrastructure burden of interacting with a Matlab/C library, we use OpenCV's DoG implementation for most of this Chapter.

\begin{figure}
\centering
\renewcommand{\imsize}{4.3cm}
\renewcommand{\hspacer}{0.3mm}
\renewcommand{\vspacer}{0.1mm}
\renewcommand{\imscl}{0.01}
\includegraphics[scale=\imscl,width=0.8\linewidth]{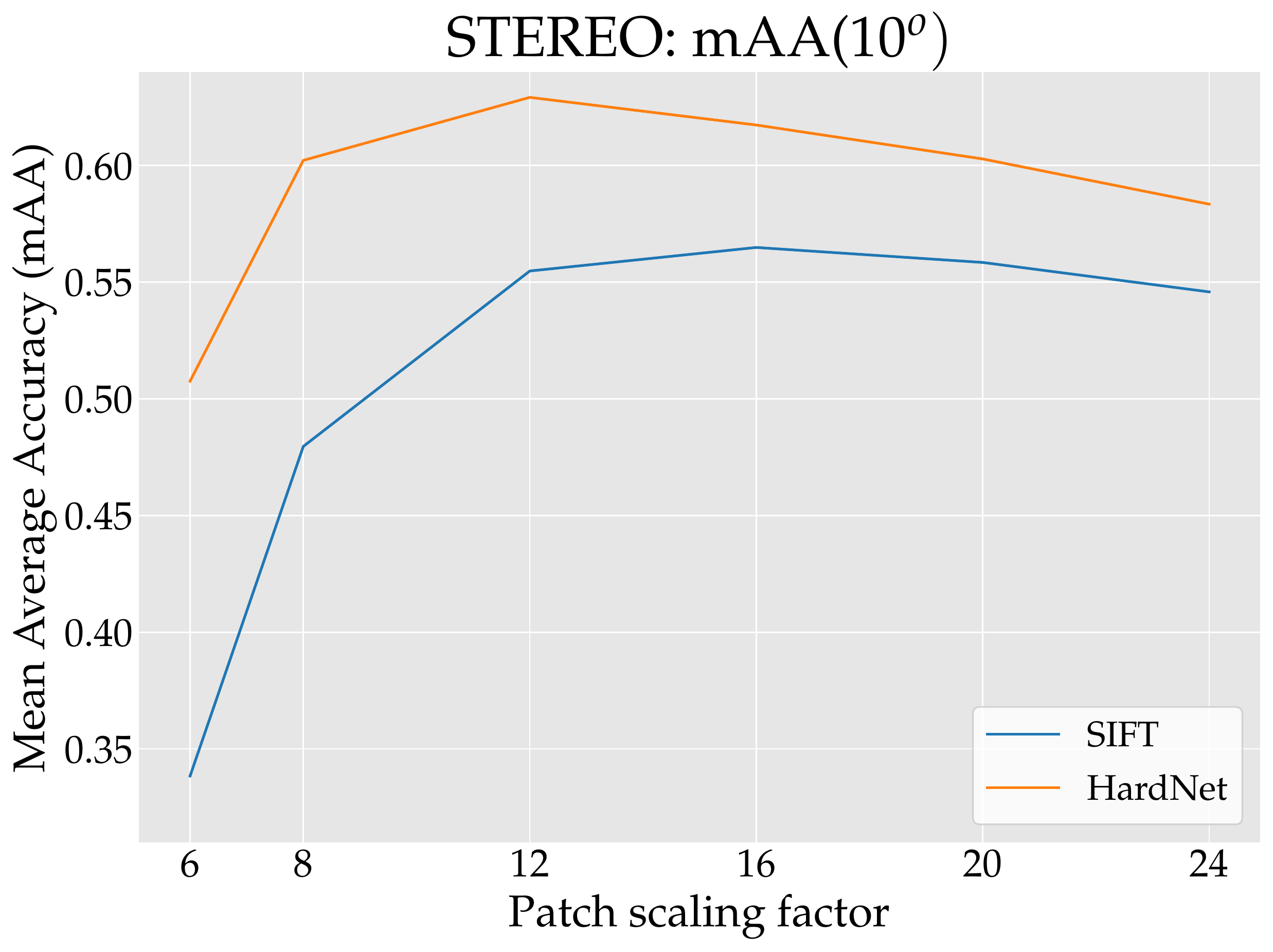}
\caption{
{\bf Validation -- Scaling the descriptor support region.}
Performance with SIFT and HardNet descriptors while applying a scaling factor $\lambda$ to the keypoint scale (note that OpenCV's default value is $\lambda{=}12$). 
\edit{We consider SIFT and HardNet. 
Default values are optimal or near-optimal.}%
}
\label{fig:results_mrsize}
\end{figure}
\paragraph{On increasing the support region (\fig{results_mrsize}).}
The size (``scale'') of the support region used to compute a descriptor can significantly affect its performance \cite{DSPSIFT2015,DeepComp2015,Aanaes02}.
We experiment with different scaling factors, using DoG with SIFT and HardNet \cite{HardNet2017}, and find that 12$\times$ the OpenCV scale \ky{(the default value)} is already nearly optimal, confirming the findings \edit{reported} in \cite{LogPolar2019}.
We show these results in \fig{results_mrsize}.
Interestingly, SIFT descriptors do benefit from increasing the scaling factor from 12 to 16, \edit{but the difference is very small -- we thus use the recommended value of 12 for the rest of the experiments}.
This, however, suggests that deep descriptors such as HardNet might be able to increase performance slightly by training on larger patches.

\section{Establishing the State of the Art}
\label{imc:results}

With the findings \edit{and the optimal parameters} found in \secref{ablation}, we move on to the test set, evaluating many methods with their optimal settings.
All experiments in this section use bidirectional matching with the `'both'' strategy.
We consider a large feature budget (up to 8k features) and a smaller one (up to 2k), and evaluate many detector/descriptor combinations.

We make three changes with respect to the validation experiments of the previous section.
(1) We double the RANSAC budget from 0.5 seconds (used for validation) to 1 second per image pair, and adjust the maximum number of iterations $\ransacMaxIters$ accordingly -- we made this decision to encourage participants to the challenge based on this benchmark to use built-in methods rather than run RANSAC themselves to squeeze out a little extra performance, and use the same values for consistency.
(2) We run each stereo and multiview evaluation three times and average the results, in order to decrease the potential randomness in the results -- \edit{in general, we found the variations within these three runs to be negligible}.
(3) We use brute-force to match descriptors instead of FLANN, as we observed a drop in performance. \edit{For more details, see Section~\ref{imc:flann} and Table~\ref{tbl:test_flann}}.

For stereo, we consider DEGENSAC and MAGSAC, which perform the best in the validation set, and PyRANSAC as a `baseline' RANSAC.
We report the results with 
\edit{both 8k features and 2k features in the following subsections.}
All observations are in terms of mAA, our primary metric, unless stated otherwise.

\definecolor{light-gray}{gray}{0.9}
\renewcommand{\arraystretch}{0.95}
\renewcommand{\tabcolsep}{0.8mm}
\begin{table}
\begin{center}
\begin{adjustbox}{width=\linewidth,center}
\footnotesize
\begin{tabular}{@{}lcccccccc@{}}
\toprule
\multicolumn{2}{c}{} & \multicolumn{2}{c}{PyRANSAC} & \multicolumn{2}{c}{DEGENSAC} & \multicolumn{2}{c}{MAGSAC} & \\
Method & NF & NI$^\uparrow$ & mAA(10$^o$)$^\uparrow$ & NI$^\uparrow$ & mAA(10$^o$)$^\uparrow$ & NI$^\uparrow$ & mAA(10$^o$)$^\uparrow$ & Rank \\
\midrule
CV-SIFT & 7861.1 & 167.6 & .3996 & 243.6 & .4584 & 297.4 & .4583 & 14 \\
VL-SIFT & 7880.6 & 179.7 & .3999 & 261.6 & .4655 & 326.2 & .4633 & 13 \\
VL-Hessian-SIFT & 8000.0 & 204.4 & .3695 & 290.2 & .4450 & 348.9 & .4335 & 15 \\
VL-DoGAff-SIFT & 7892.1 & 171.6 & .3984 & 250.1 & .4680 & 317.1 & .4666 & 11 \\
VL-HesAffNet-SIFT & 8000.0 & 209.3 & .3933 & 299.0 & .4679 & 350.0 & .4626 & 12 \\
\midrule
CV-$\sqrt{\text{SIFT}}$ & 7860.8 & 192.3 & .4228 & 281.7 & .4930 & 347.5 & .4941 & 10 \\
CV-SURF & 7730.0 & 107.9 & .2280 & 113.6 & .2593 & 145.3 & .2552 & 19 \\
CV-AKAZE & 7857.1 & 131.4 & .2570 & 246.8 & .3074 & 301.8 & .3036 & 17 \\
CV-ORB & 7150.2 & 123.7 & .1220 & 150.0 & .1674 & 178.9 & .1570 & 22 \\
CV-FREAK & 8000.0 & 123.3 & .2273 & 131.0 & .2711 & 196.7 & .2656 & 18 \\
\midrule
L2-Net & 7861.1 & 213.8 & .4621 & 366.0 & .5295 & 481.0 & .5252 & 5 \\
DoG-HardNet & 7861.1 & 286.5 & .4801 & 432.3 & {\bf \textcolor{Green}{.5543}} & 575.1 & .5502 & 2 \\
DoG-HardNetAmos+ & 7861.0 & 265.7 & .4607 & 398.6 & .5385 & 528.7 & .5329 & 3 \\
Key.Net-HardNet & 7997.6 & 448.1 & .3997 & 598.3 & .4986 & 815.4 & .4739 & 9 \\
Key.Net-SOSNet & 7997.6 & 275.5 & .4236 & 587.4 & .5019 & 766.4 & .4780 & 8 \\
GeoDesc & 7861.1 & 205.4 & .4328 & 348.5 & .5111 & 453.4 & .5056 & 7 \\
ContextDesc & 7859.0 & 278.2 & .4684 & 493.6 & .5098 & 544.1 & .5143 & 6 \\
DoG-SOSNet & 7861.1 & 281.6 & .4784 & 424.6 & {\bf \textcolor{red}{.5587}} & 563.3 & {\bf \textcolor{NavyBlue}{.5517}} & 1 \\
LogPolarDesc & 7861.1 & 254.4 & .4574 & 441.8 & .5340 & 591.2 & .5238 & 4 \\
\midrule
D2-Net (SS) & 5665.3 & 280.8 & .1933 & 482.3 & .2228 & 781.3 & .2032 & 21 \\
D2-Net (MS) & 6924.1 & 278.2 & .2160 & 470.6 & .2506 & 741.2 & .2321 & 20 \\
R2D2 (wasf-n8-big) & 7940.5 & 457.6 & .3683 & 842.2 & .4437 & 998.9 & .4236 & 16 \\
\midrule
DoG-AffNet-HardNet & 7834.0 & 267.9 & .4505 & 403.4 & .5447 & 516.8 & .5412 & $3^{*}$ \\
DoG-MKD-Concat & 7860.8 & 208.0 & .4061 & 305.8 & .4846 & 381.4 & .4810 & $11^{*}$  \\
DoG-TFeat & 7860.8 & 160.8 & .4008 & 234.8 & .4649 & 292.6 & .4668 & $13^{*}$ \\
\bottomrule
\end{tabular}
\end{adjustbox}
\end{center}
\caption{
{\bf Test -- Stereo results with 8k features.}
We report: {\bf (NF)} Number of Features; {\bf (NI)} Number of Inliers produced by RANSAC; and {\bf mAA(10$^o$)}.
Top three methods by mAA marked in
\textcolor{red}{red}, \textcolor{Green}{green} and \textcolor{NavyBlue}{blue}.
 $^{*}$ {\bf The last group of results is obtained after the paper publication and not described in the text of the paper. Their rank does not influence other entries ranks.}
}
\label{tbl:test_stereo}
\end{table}

\renewcommand{\arraystretch}{0.6}
\renewcommand{\tabcolsep}{1.2mm}
\begin{table}
\begin{center}
\begin{adjustbox}{width=\linewidth,center}
\footnotesize
\begin{tabular}{@{}lcccccccc@{}}
\toprule
Method & NL$^\uparrow$ & SR$^\uparrow$ & RC$^\uparrow$ & TL$^\uparrow$ & mAA(5$^o$)$^\uparrow$ & mAA(10$^o$)$^\uparrow$ & ATE$^\downarrow$ & Rank \\
\midrule
CV-SIFT & 2577.6 & 96.7 & 94.1 & 3.95 & .5309 & .6261 & .4721 & 14 \\
VL-SIFT & 3030.7 & 97.9 & 95.4 & 4.17 & .5273 & .6283 & .4669 & 13 \\
VL-Hessian-SIFT & 3209.1 & 97.4 & 94.1 & 4.13 & .4857 & .5866 & .5175 & 16 \\
VL-DoGAff-SIFT & 3061.5 & 98.0 & 96.2 & 4.11 & .5263 & .6296 & .4751 & 12 \\
VL-HesAffNet-SIFT & 3327.7 & 97.7 & 95.2 & 4.08 & .5049 & .6069 & .4897 & 15 \\
\midrule
CV-$\sqrt{\text{SIFT}}$ & 3312.1 & 98.5 & 96.6 & 4.13 & .5778 & .6765 & .4485 & 9 \\
CV-SURF & 2766.2 & 94.8 & 92.6 & 3.47 & .3897 & .4846 & .6251 & 18 \\
CV-AKAZE & 4475.9 & 99.0 & 95.4 & 3.88 & .4516 & .5553 & .5715 & 17 \\
CV-ORB & 3260.3 & 97.2 & 91.1 & 3.45 & .2697 & .3509 & .7377 & 22 \\
CV-FREAK & 2859.1 & 92.9 & 91.7 & 3.53 & .3735 & .4653 & .6229 & 20 \\
\midrule
L2-Net & 3424.9 & 98.6 & 96.2 & 4.21 & .5661 & .6644 & .4482 & 10 \\
DoG-HardNet & 4001.4 & 99.5 & {\bf \textcolor{NavyBlue}{97.7}} & {\bf \textcolor{NavyBlue}{4.34}} & {\bf \textcolor{red}{.6090}} & {\bf \textcolor{red}{.7096}} & {\bf \textcolor{red}{.4187}} & 1 \\
DoG-HardNetAmos+ & 3550.6 & 98.8 & 96.9 & 4.28 & .5879 & .6888 & .4428 & 6 \\
Key.Net-HardNet & 3366.0 & 98.9 & 96.7 & 4.32 & .5391 & .6483 & .4622 & 11 \\
Key.Net-SOSNet & {\bf \textcolor{NavyBlue}{5505.5}} & 100.0 & {\bf \textcolor{red}{98.7}} & {\bf \textcolor{Green}{4.46}} & .5989 & {\bf \textcolor{Green}{.7038}} & .4286 & 2 \\
GeoDesc & 3839.0 & 99.1 & 97.2 & 4.26 & .5782 & .6803 & .4445 & 8 \\
ContextDesc & 3732.5 & 99.3 & 97.6 & 4.22 & {\bf \textcolor{Green}{.6036}} & {\bf \textcolor{NavyBlue}{.7035}} & {\bf \textcolor{NavyBlue}{.4228}} & 3 \\
DoG-SOSNet & 3796.0 & 99.3 & 97.4 & 4.32 & {\bf \textcolor{NavyBlue}{.6032}} & .7021 & {\bf \textcolor{Green}{.4226}} & 4 \\
LogPolarDesc & 4054.6 & 99.0 & 96.4 & 4.32 & .5928 & .6928 & .4340 & 5 \\
\midrule
D2-Net (SS) & {\bf \textcolor{Green}{5893.8}} & {\bf \textcolor{Green}{99.8}} & 97.5 & 3.62 & .3435 & .4598 & .6361 & 21 \\
D2-Net (MS) & {\bf \textcolor{red}{6759.3}} & {\bf \textcolor{NavyBlue}{99.7}} & {\bf \textcolor{Green}{98.2}} & 3.39 & .3524 & .4751 & .6283 & 19 \\
R2D2 (wasf-n8-big) & 4432.9 & {\bf \textcolor{NavyBlue}{99.7}} & 97.2 & {\bf \textcolor{red}{4.59}} & .5775 & .6832 & .4333 & 7 \\
\midrule
DoG-AffNet-HardNet & 4671.3 & {\bf \textcolor{Green}{99.9}} & {\bf \textcolor{NavyBlue}{98.1}} & {\bf \textcolor{Green}{4.56}} & {\bf \textcolor{red}{.6296}} & {\bf \textcolor{red}{.7267}} & {\bf \textcolor{red}{.4021}} & $1^{*}$ \\
DoG-MKD-Concat & 3507.4 & 98.5 & 96.1 & 4.17 & .5461 & .6476 & .4668 & $11^{*}$ \\
DoG-TFeat & 2905.3 & 97.1 & 94.8 & 4.04 & .5270 & .6261 & .4873 & $14^{*}$ \\
\bottomrule
\end{tabular}
\end{adjustbox}
\end{center}
\caption{
{\bf Test -- Multiview results with 8k features.}
We report:
{\bf (NL)} Number of 3D Landmarks;
{\bf (SR)} Success Rate (\%) in the 3D reconstruction across ``bags'';
{\bf (RC)} Ratio of Cameras (\%) registered in a ``bag'';
{\bf (TL)} Track Length or number of observations per landmark;
{\bf mAA} at 5 and 10$^o$;
and {\bf (ATE)} Absolute Trajectory Error.
All metrics are averaged across different ``bag'' sizes, as explained in \secref{benchmark}.
We rank them by mAA at 10$^o$ and color-code them as in \tbl{test_stereo}.
 $^{*}$ {\bf The last group of results is obtained after the paper publication and not described in the text of the paper. Their rank does not influence other entries ranks.}
}
\label{tbl:test_multiview}
\end{table}

\subsection[Results with 8k features]{Results with 8k features \edit{--- Tables~\ref{tbl:test_stereo}~and~\ref{tbl:test_multiview}}}

On the stereo task, deep descriptors extracted on DoG keypoints are at the top in terms of mAA, with SOSNet being \#1, closely followed by HardNet.
Interestingly, `HardNetAmos+'~\cite{HardNetAMOS2019}, a version trained on more datasets -- Brown~\cite{Brown10}, HPatches~\cite{hpatches2017}, and AMOS-patches~\cite{HardNetAMOS2019} -- performs worse than the original models, trained only on the ``Liberty'' scene from Brown's dataset.
On the multiview task, HardNet edges out ContextDesc, SOSNet and LogpolarDesc by a small margin. Affine features bring benefits in the multivuiew scenario, where AffNet improves DoG-HardNet results in terms of both mAA and success rate. However, the plain DoG detector is better for the stereo task.
We also pair HardNet \revision{and SOSNet} with Key.Net, a learned detector, which performs worse than with DoG when extracting a large number of features, \revision{with the exception of Key.Net + SOSNet on the multiview task.}

R2D2, the best performing end-to-end method, does well on multiview (\revision{\#7}), but performs worse than SIFT on stereo -- it produces a much larger number of ``inliers'' (which may be correct or incorrect) than most other methods. This suggests that, like D2-Net, its lack of compatibility with the ratio test may be a problem when paired with sample-based robust estimators, due to a lower inlier ratio.
Note that D2-net performs poorly on our benchmark, despite state-of-the-art results on others.
On the multiview task it creates many more \ky{3D} landmarks than any other method.
\edit{Both issues may be related to its poor localization (pixel) accuracy, due to operating on downsampled feature maps.}

Out of the handcrafted methods, SIFT -- RootSIFT specifically -- \edit{remains competitive}, \edit{being}
\revision{\#10} on stereo and \revision{\#9} on multiview, within 13.1\% and 4.9\% relative of the top performing method, respectively,
\edit{while previous benchmarks report differences in performance of {\em orders of magnitude.}}
Other ``classical'' features do not fare so well. 
\edit{One interesting observation is that among these, their ranking on validation and test set is not consistent -- Hessian is better on validation than DoG, but significantly worse on the test set, especially in the multiview setup.}
\revision{While this is a special case, this nonetheless demonstrates that a small-scale benchmark can be misleading, and a method needs to be tested on a variety of scenes, which is what our test set aims to provide.}

Regarding the robust estimators, DEGENSAC and MAGSAC both perform very well, with the former edging out the latter for most local feature methods.
This may be due to the nature of the scenes, which \edit{often} contain dominant planes.

\subsection[Results with 2k features]{Results with 2k features \edit{--- Tables~\ref{tbl:test_stereo_2k} and~\ref{tbl:test_multiview_2k}}}

Results change slightly on the low-budget regime, \revision{where the top two spots on both tasks are occupied by Key.Net+SOSNet and DoG-AffNet-HardNet}. They are closely followed by LogPolarDesc (\revision{\#3 on stereo and \#4 on multiview}), \edit{a method} trained on DoG keypoints -- but using a much larger support region, resampled into log-polar patches. 
R2D2 performs very well on the multiview task (\revision{\#3}), while once again falling a bit short on the stereo task \edit{(\revision{\#8, and 14.5\%} relative below the \#1 method)}, for which it \edit{retrieves} a number of inliers significantly larger than its competitors.
The rest of the end-to-end methods do not perform so well, other than SuperPoint, which obtains competitive results on the multiview task.

The difference between classical and learned methods is more pronounced than with 8k points, with RootSIFT once again at the top, but now within 31.4\% relative of the \#1 method on stereo, and \revision{26.9\%} on multiview.
\edit{This is somewhat to be expected, given that with fewer keypoints, the quality of each individual point matters more.}

\begin{figure}
\centering
\renewcommand{\hspacer}{0.3mm}
\renewcommand{\vspacer}{0.1mm}
\renewcommand{\imscl}{0.01}
\includegraphics[scale=\imscl,width=\linewidth]{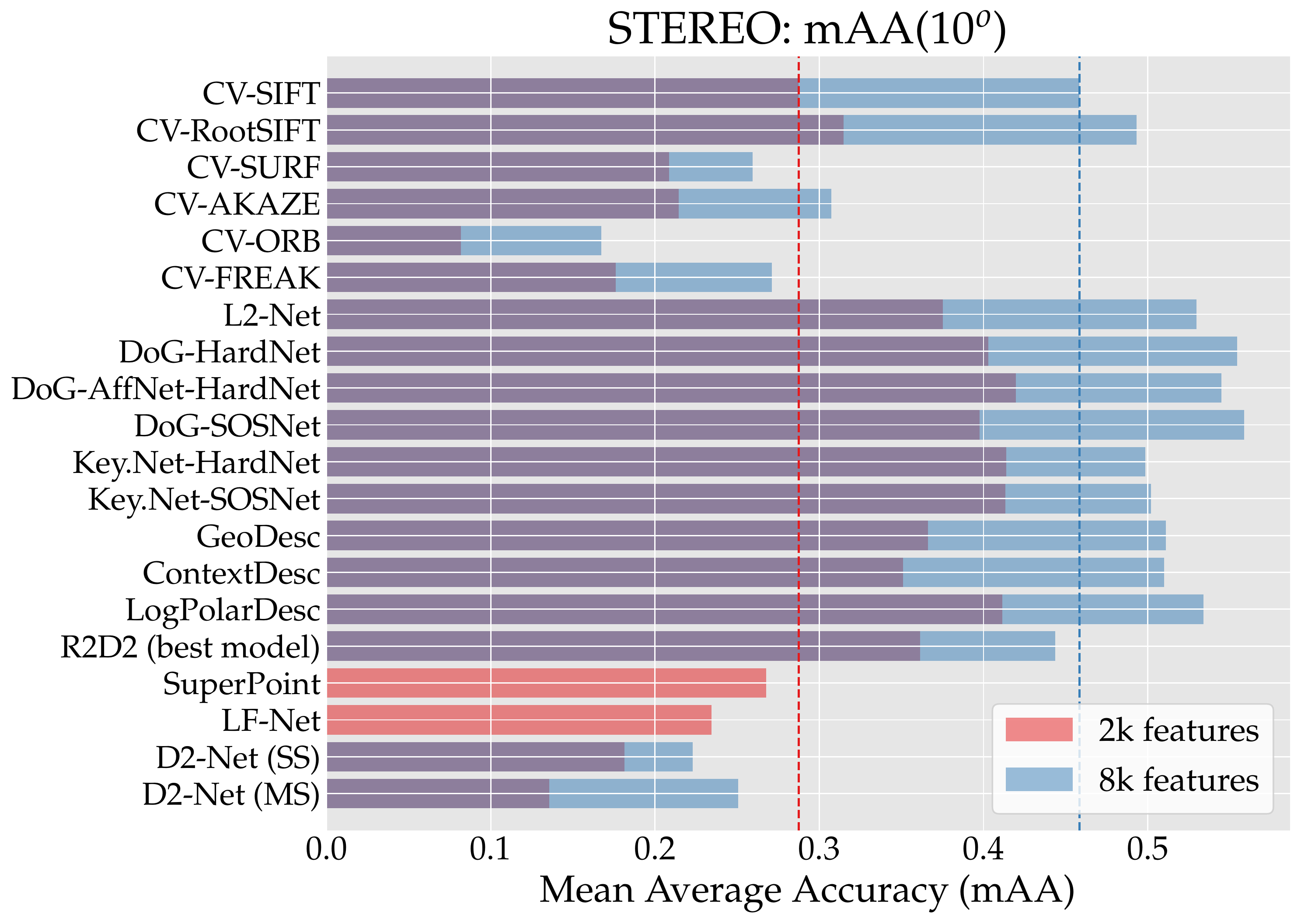}
\caption{
{\bf Test -- Stereo performance: 2k vs 8k features.}
\edit{We compare the results obtained with different methods using either 2k or 8k features -- we use DEGENSAC, which performs better than other RANSAC variants under most circumstances. 
Dashed lines indicate SIFT's performance.
For LF-Net and SuperPoint we do not include results with 8k features, as we failed to obtain meaningful results.
For R2D2, we use the best model for each setting.}
}
\label{fig:rest_2k_vs_8k}
\vspace{-3mm}
\end{figure}

\begin{figure}
\centering
\renewcommand{\hspacer}{0.3mm}
\renewcommand{\vspacer}{0.1mm}
\renewcommand{\imscl}{0.01}
\includegraphics[scale=\imscl,width=\linewidth]{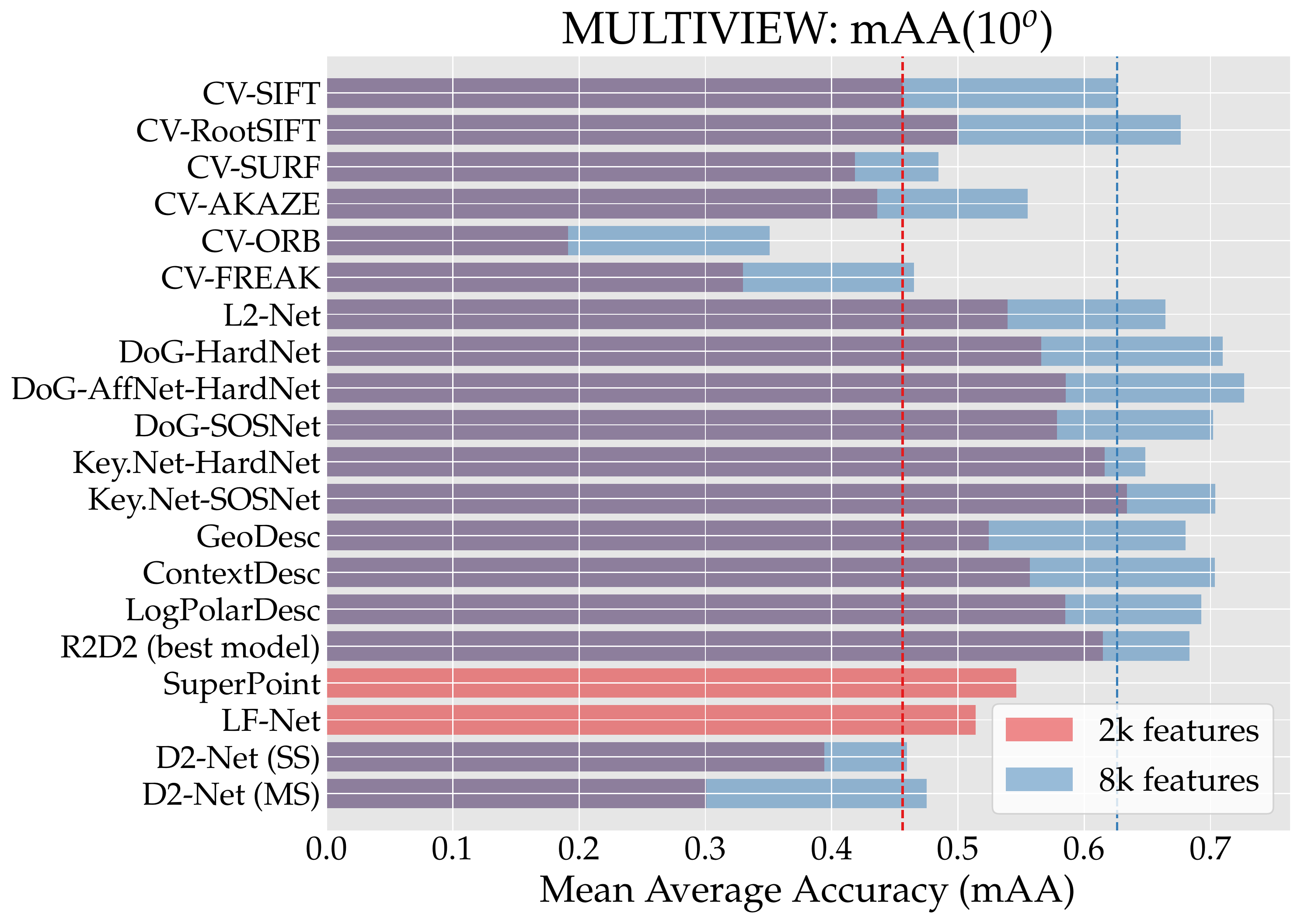}
\caption{
{\bf Test -- Multiview performance: 2k vs 8k features.}
\edit{Same as \fig{rest_2k_vs_8k}, for multiview.
}
}
\label{fig:rest_2k_vs_8k_multiview}
\vspace{-3mm}
\end{figure}

\subsection[2k features vs 8k features]{2k features vs 8k features \edit{--- Figs.~\ref{fig:rest_2k_vs_8k}~and~\ref{fig:rest_2k_vs_8k_multiview}}}

We compare the results between the low- and high-budget regimes in \fig{rest_2k_vs_8k}, for stereo (with DEGENSAC), and \fig{rest_2k_vs_8k_multiview}, for multiview.
Note how methods can behave quite differently. 
\edit{Those} based on DoG \edit{significantly benefit from an increased feature budget}, whereas those learned end-to-end \edit{may} require re-training -- this is exemplified by the difference in performance between 2k and 8k for Key.Net+Hardnet, specially on multiview, which is very narrow despite quadrupling the budget.
Overall, learned detectors -- KeyNet, SuperPoint, R2D2, LF-Net -- show relatively better results on multiview setup than on stereo. Our hypothesis is that they have good robustness, but low localization precision \revision{which is later corrected during bundle adjustment}.
\definecolor{light-gray}{gray}{0.9}
\renewcommand{\arraystretch}{0.95}
\renewcommand{\tabcolsep}{0.8mm}
\begin{table}
\begin{center}
\begin{adjustbox}{width=\linewidth,center}
\footnotesize
\begin{tabular}{@{}lcccccccc@{}}
\toprule
\multicolumn{2}{c}{} & \multicolumn{2}{c}{PyRANSAC} & \multicolumn{2}{c}{DEGENSAC} & \multicolumn{2}{c}{MAGSAC} & \\
Method & NF & NI$^\uparrow$ & mAA(10$^o$)$^\uparrow$ & NI$^\uparrow$ & mAA(10$^o$)$^\uparrow$ & NI$^\uparrow$ & mAA(10$^o$)$^\uparrow$ & Rank \\
\midrule
CV-SIFT & 2048.0 & 84.9 & .2489 & 79.0 & .2875 & 99.2 & .2805 & 12 \\
\midrule
CV-$\sqrt{\text{SIFT}}$ & 2048.0 & 84.2 & .2724 & 88.3 & .3149 & 106.8 & .3125 & 10 \\
CV-SURF & 2048.0 & 37.9 & .1725 & 72.7 & .2086 & 87.0 & .2081 & 15 \\
CV-AKAZE & 2048.0 & 96.1 & .1780 & 91.0 & .2144 & 115.5 & .2127 & 14 \\
CV-ORB & 2031.8 & 56.3 & .0610 & 63.5 & .0819 & 71.5 & .0765 & 19 \\
CV-FREAK & 2048.0 & 62.5 & .1461 & 65.6 & .1761 & 78.4 & .1698 & 17 \\
\midrule
L2-Net & 1936.3 & 66.1 & .3131 & 92.4 & .3752 & 114.7 & .3691 & 6 \\
DoG-HardNet & 1936.3 & 111.9 & .3508 & 117.7 & .4029 & 150.5 & .4033 & 5 \\
Key.Net-HardNet & 2048.0 & 134.4 & .3272 & 174.8 & {\bf \textcolor{red}{.4139}} & 228.4 & .3897 & 1 \\
Key.Net-SOSNet & 2048.0 & 88.0 & .3279 & 171.9 & {\bf \textcolor{Green}{.4132}} & 212.9 & .3928 & 2 \\
GeoDesc & 1936.3 & 98.9 & .3127 & 103.9 & .3662 & 129.7 & .3640 & 7 \\
ContextDesc & 2048.0 & 118.8 & .2965 & 124.1 & .3510 & 146.4 & .3485 & 9 \\
DoG-SOSNet & 1936.3 & 111.1 & .3536 & 132.1 & .3976 & 149.6 & .4092 & 4 \\
LogPolarDesc & 1936.3 & 118.8 & .3569 & 124.9 & {\bf \textcolor{NavyBlue}{.4115}} & 161.0 & .4064 & 3 \\
\midrule
D2-Net (SS) & 2045.6 & 107.6 & .1157 & 134.8 & .1355 & 259.3 & .1317 & 18 \\
D2-Net (MS) & 2038.2 & 149.3 & .1524 & 188.4 & .1813 & 302.9 & .1703 & 16 \\
LF-Net & 2020.3 & 100.2 & .1927 & 106.5 & .2344 & 141.0 & .2226 & 13 \\
SuperPoint & 2048.0 & 120.1 & .2577 & 126.8 & .2964 & 127.3 & .2676 & 11 \\
R2D2 (wasf-n16) & 2048.0 & 191.0 & .2829 & 215.6 & .3614 & 215.6 & .3614 & 8 \\
\midrule
DoG-AffNet-HardNet & 2047.8 & 105.6 & .3589 & 152.1 & {\bf \textcolor{red}{.4197}} & 195.2 & {\bf \textcolor{Green}{.4175}} & $1^{*}$ \\
\bottomrule
\end{tabular}
\end{adjustbox}
\end{center}
\caption{
{\bf Test -- Stereo results with 2k features.}
Same as \tbl{test_stereo}.
 $^{*}$ {\bf The last group of results is obtained after the paper publication and not described in the text of the paper. Their rank does not influence other entries ranks.}
}
\label{tbl:test_stereo_2k}
\end{table}

\renewcommand{\arraystretch}{0.6}
\renewcommand{\tabcolsep}{1.2mm}
\begin{table}
\begin{center}
\begin{adjustbox}{width=\linewidth,center}
\footnotesize
\begin{tabular}{@{}lcccccccc@{}}
\toprule
Method & NL$^\uparrow$ & SR$^\uparrow$ & RC$^\uparrow$ & TL$^\uparrow$ & mAA(5$^o$)$^\uparrow$ & mAA(10$^o$)$^\uparrow$ & ATE$^\downarrow$ & Rank \\
\midrule
CV-SIFT & 1081.2 & 87.6 & 87.4 & 3.70 & .3718 & .4562 & .6136 & 13 \\
\midrule
CV-$\sqrt{\text{SIFT}}$ & 1174.7 & 90.3 & 89.4 & 3.82 & .4074 & .4995 & .5589 & 12 \\
CV-SURF & 1186.6 & 90.2 & 88.6 & 3.55 & .3335 & .4184 & .6701 & 15 \\
CV-AKAZE & 1383.9 & 94.7 & 90.9 & 3.74 & .3393 & .4361 & .6422 & 14 \\
CV-ORB & 683.3 & 74.9 & 73.0 & 3.21 & .1422 & .1914 & .8153 & 19 \\
CV-FREAK & 1075.2 & 87.2 & 86.3 & 3.52 & .2578 & .3297 & .7169 & 17 \\
\midrule
L2-Net & 1253.3 & 94.7 & 92.6 & 3.96 & .4369 & .5392 & .5419 & 9 \\
DoG-HardNet & 1338.2 & 96.3 & 93.7 & 4.03 & .4624 & .5661 & .5093 & 6 \\
Key.Net-HardNet & 1276.3 & 97.8 & {\bf \textcolor{NavyBlue}{95.7}} & {\bf \textcolor{red}{4.49}} & {\bf \textcolor{Green}{.5050}} & {\bf \textcolor{Green}{.6161}} & {\bf \textcolor{Green}{.4902}} & 2 \\
Key.Net-SOSNet & 1475.5 & {\bf \textcolor{Green}{99.3}} & {\bf \textcolor{red}{96.5}} & {\bf \textcolor{Green}{4.42}} & {\bf \textcolor{red}{.5229}} & {\bf \textcolor{red}{.6340}} & {\bf \textcolor{red}{.4853}} & 1 \\
GeoDesc & 1133.6 & 93.6 & 91.3 & 4.02 & .4246 & .5244 & .5455 & 10 \\
ContextDesc & 1504.9 & 95.6 & 93.3 & 3.92 & .4529 & .5568 & .5327 & 7 \\
DoG-SOSNet & 1317.4 & 96.0 & 93.8 & 4.05 & .4739 & .5784 & .5194 & 5 \\
LogPolarDesc & 1410.2 & 96.0 & 93.8 & 4.05 & .4794 & .5849 & .5090 & 4 \\
\midrule
D2-Net (SS) & {\bf \textcolor{red}{2357.9}} & {\bf \textcolor{NavyBlue}{98.9}} & 94.7 & 3.39 & .2875 & .3943 & .7010 & 16 \\
D2-Net (MS) & {\bf \textcolor{Green}{2177.3}} & 98.2 & 93.4 & 3.01 & .1921 & .3007 & .7861 & 20 \\
LF-Net & 1385.0 & 95.6 & 90.4 & 4.14 & .4156 & .5141 & .5738 & 11 \\
SuperPoint & 1184.3 & 95.6 & 92.4 & {\bf \textcolor{NavyBlue}{4.34}} & .4423 & .5464 & .5457 & 8 \\
R2D2 (wasf-n16) & 1228.4 & {\bf \textcolor{red}{99.4}} & {\bf \textcolor{Green}{96.2}} & 4.29 & {\bf \textcolor{NavyBlue}{.5045}} & {\bf \textcolor{NavyBlue}{.6149}} & {\bf \textcolor{NavyBlue}{.4956}} & 3 \\
\midrule
DoG-AffNet-HardNet & {\bf \textcolor{NavyBlue}{1788.7}} & 98.7 & {\bf \textcolor{NavyBlue}{95.7}} & 4.19 & .4771 & .5854 & .5114 & $4^{*}$\\ 
\bottomrule
\end{tabular}
\end{adjustbox}
\end{center}
\caption{
{\bf Test -- Multiview results with 2k features.}
Same as \tbl{test_multiview}.
 $^{*}$ {\bf The last group of results is obtained after the paper publication and not described in the text of the paper. Their rank does not influence other entries ranks.}
}
\label{tbl:test_multiview_2k}
\end{table}

\subsection[Outlier pre-filtering with deep networks]{Outlier pre-filtering with deep networks \edit{--- \tbl{test_cne}}}

Next, we study the performance of CNe~\cite{CNe2018} for outlier rejection, paired with PyRANSAC, DEGENSAC, and MAGSAC.
Its training data does not use the ratio test, so we omit it here too -- note that because of this, it expects a relatively large number of input matches. 
We thus evaluate it only for the 8k feature setting, \edit{while using the ``both'' matching strategy}.

\definecolor{light-gray}{gray}{0.9}
\renewcommand{\arraystretch}{0.9}
\renewcommand{\tabcolsep}{0.3mm}
\begin{table}
\begin{center}
\begin{adjustbox}{width=\linewidth,center}
\footnotesize
\begin{tabular}{@{}lcccccccc@{}}
\toprule
& \multicolumn{6}{c}{Stereo Task} & \multicolumn{2}{c}{\multirow{2}{*}{Multi-view Task}} \\
& \multicolumn{2}{c}{PyRANSAC} & \multicolumn{2}{c}{DEGENSAC} & \multicolumn{2}{c}{MAGSAC} & \\
Method & mAA(10$^o$)$^\uparrow$ & $\Delta$(\%)$^\uparrow$ & mAA(10$^o$)$^\uparrow$ & $\Delta$(\%)$^\uparrow$ & mAA(10$^o$)$^\uparrow$ & $\Delta$(\%)$^\uparrow$ & mAA(10$^o$)$^\uparrow$ & $\Delta$(\%)$^\uparrow$ \\
\midrule
CV-SIFT & .4086 & +2.24 & .4751 & +3.65 & .4694 & +2.42 & .6815 & +8.85 \\
\midrule
CV-$\sqrt{\text{SIFT}}$ & .4205 & -0.53 & .4927 & -0.06 & .4848 & -1.87 & {\bf \textcolor{NavyBlue}{.6978}} & +3.16 \\
CV-SURF & .2490 & +9.18 & .3071 & +18.42 & .2954 & +15.75 & .5750 & +18.67 \\
CV-AKAZE & .2857 & +11.18 & .3417 & +11.18 & .3316 & +9.23 & .6026 & +8.51 \\
CV-ORB & .1323 & +8.49 & .1856 & +10.87 & .1748 & +11.34 & .4171 & +18.88 \\
CV-FREAK & .2532 & +11.36 & .3204 & +18.18 & .3053 & +14.93 & .5574 & +19.79 \\
\midrule
L2-Net & .4377 & -5.27 & .5012 & -5.35 & .4937 & -5.99 & .6951 & +4.62 \\
DoG-HardNet & .4427 & -7.80 & {\bf \textcolor{Green}{.5156}} & -6.98 & .5056 & -8.11 & {\bf \textcolor{Green}{.7061}} & -.50 \\
Key.Net-HardNet & .3081 & -22.92 & .4226 & -15.23 & .4012 & -15.36 & .6620 & +2.11 \\
GeoDesc & .4239 & -2.05 & .4924 & -3.67 & .4807 & -4.93 & .6956 & +2.25 \\
ContextDesc & .3976 & -15.11 & .4482 & -12.09 & .4535 & -11.83 & .6900 & -1.91 \\
DoG-SOSNet & .4439 & -7.21 & {\bf \textcolor{red}{.5187}} & -7.15 & {\bf \textcolor{NavyBlue}{.5073}} & -8.04 & {\bf \textcolor{red}{.7103}} & +1.18 \\
LogPolarDesc & .4259 & -6.89 & .4898 & -8.27 & .4808 & -8.22 & .6871 & -.82 \\
\midrule
D2-Net (SS) & .1231 & -36.32 & .1717 & -22.95 & .1608 & -20.86 & .4639 & +0.89 \\
D2-Net (MS) & .0998 & -53.78 & .1370 & -45.33 & .1316 & -43.29 & .4132 & -13.02 \\
R2D2 (wasf-n8-big) & .2218 & -39.78 & .3141 & -29.21 & .3032 & -28.43 & .6229 & -8.83 \\
\bottomrule
\end{tabular}
\end{adjustbox}
\end{center}
\caption{
{\bf Test -- Outlier pre-filtering with CNe (8k features).}
We report mAP at 10$^o$ with CNe, on stereo and multi-view, and its increase in performance w.r.t. \tbl{test_multiview} -- positive $\Delta$ meaning CNe helps.
When using CNe, we disable the ratio test.
}
\label{tbl:test_cne}
\end{table}

Our experiments with SIFT, \edit{the local feature used to train CNe}, are encouraging: CNe aggressively filters out about 80\% of the matches in a single forward pass, boosting mAA at 10$^\circ$ by \edit{2-4\%} relative for stereo task and 8\% for multiview task.
In fact, it is surprising that nearly all classical methods benefit from it, with gains of up to \et{20\%} relative.
By contrast, it damages performance with most learned descriptors, even those operating on DoG keypoints, and particularly for methods learned end-to-end, such as D2-Net and R2D2.
We hypothesize this might be because the models performed better on the ``classical'' keypoints it was trained with
-- \cite{Sun19} reports that re-training them for a specific feature helps.

\renewcommand{\arraystretch}{1}
\renewcommand{\tabcolsep}{0.9mm}
\begin{table}
\begin{center}
\tiny
\begin{tabular}{@{}lcccccccc@{}}
\toprule
& \multicolumn{2}{c}{CV-$\sqrt{\text{SIFT}}$} & \multicolumn{2}{c}{HardNet} & \multicolumn{2}{c}{SOSNet} & \multicolumn{2}{c}{LogPolarDesc} \\
& NI$^\uparrow$ & mAA(10$^o$)$^\uparrow$ & NI$^\uparrow$ & mAA(10$^o$)$^\uparrow$ & NI$^\uparrow$ & mAA(10$^o$)$^\uparrow$ & NI$^\uparrow$ & mAA(10$^o$)$^\uparrow$ \\
\midrule
Standard & 281.7 & 0.4930 & 432.3 & 0.5543 & 424.6 & 0.5587 & 441.8 & 0.5340 \\
\midrule
Upright & 270.0 & 0.4878 & 449.2 & 0.5542 & 432.9 & 0.5554 & 461.8 & 0.5409 \\
$\Delta$ (\%) & -4.15 & -1.05 & +3.91 & -0.02 & +1.95 & -0.59 & +4.53 & +1.29 \\
\midrule
Upright++ & 358.9 & 0.5075 & 527.6 & 0.5728 & 508.4 & 0.5738 & 543.2 & 0.5510 \\
$\Delta$ (\%) & +27.41 & +2.94 & +22.04 & +3.34 & +19.74 & +2.70 & +22.95 & +3.18 \\
\bottomrule
\end{tabular}
\end{center}
\caption{
{\bf Test -- Stereo performance with upright descriptors (8k features).}
\edit{We report {\bf (NI)} the number of inliers and {\bf mAA} at 10$^o$ for the stereo task, using DEGENSAC. As DoG may return multiple orientations for the same point~\cite{SIFT2004} (up to 30\%), we report: {\bf (top)} with orientation estimation; {\bf (middle)} setting the orientation to zero while removing duplicates; and {\bf (bottom)} adding new points until hitting the 8k-feature budget.}
}
\label{tbl:test_upright}
\end{table}
\renewcommand{\arraystretch}{1}
\renewcommand{\tabcolsep}{0.9mm}
\begin{table}
\begin{center}
\tiny
\begin{tabular}{@{}lcccccccc@{}}
\toprule
& \multicolumn{2}{c}{CV-$\sqrt{\text{SIFT}}$} & \multicolumn{2}{c}{HardNet} & \multicolumn{2}{c}{SOSNet} & \multicolumn{2}{c}{LogPolarDesc} \\
& NL$^\uparrow$ & mAA(10$^o$)$^\uparrow$ & NL$^\uparrow$ & mAA(10$^o$)$^\uparrow$ & NL$^\uparrow$ & mAA(10$^o$)$^\uparrow$ & NL$^\uparrow$ & mAA(10$^o$)$^\uparrow$ \\
\midrule
Standard & 3312.1 & 0.6765 & 4001.4 & 0.7096 & 3796.0 & 0.7021 & 4054.6 & 0.6928 \\
\midrule
Upright & 3485.1 & 0.6572 & 3594.6 & 0.6962 & 4025.1 & 0.7054 & 3737.4 & 0.6934 \\
$\Delta$ (\%) & +5.22 & -2.85 & -10.17 & -1.89 & +6.04 & +0.47 & -7.82 & +0.09 \\
\midrule
Upright++ & 4404.6 & 0.6792 & 4250.4 & 0.7231 & 3988.6 & 0.7129 & 4414.1 & 0.7109 \\
$\Delta$ (\%) & +32.99 & +0.40 & +6.22 & +1.90 & +5.07 & +1.54 & +8.87 & +2.61 \\
\bottomrule
\end{tabular}
\end{center}
\caption{
{\bf Test -- Multiview performance with upright descriptors (8k features).}
Analogous to \tbl{test_upright}, on the multiview task. We report the number of landmarks (NL) instead of the number of inliers (NI).
}
\label{tbl:test_upright_multiview}
\end{table}
\renewcommand{\arraystretch}{1}
\renewcommand{\tabcolsep}{0.9mm}
\begin{table}
\begin{center}
\tiny
\begin{tabular}{@{}lcccccccc@{}}
\toprule
& \multicolumn{2}{c}{CV-$\sqrt{\text{SIFT}}$} & \multicolumn{2}{c}{HardNet} & \multicolumn{2}{c}{SOSNet} & \multicolumn{2}{c}{D2Net} \\
& NI$^\uparrow$ & mAA(10$^o$)$^\uparrow$ & NI$^\uparrow$ & mAA(10$^o$)$^\uparrow$ & NI$^\uparrow$ & mAA(10$^o$)$^\uparrow$ & NI$^\uparrow$ & mAA(10$^o$)$^\uparrow$ \\
\midrule
Exact & 281.7 & 0.4930 & 432.0 & 0.5532 & 424.3 & 0.5575 & 470.6 & 0.2506 \\
\midrule
FLANN & 274.6 & 0.4879 & 363.3 & 0.5222 & 339.8 & 0.5179 & 338.9 & 0.2046 \\
$\Delta$ (\%) & -2.52 & -1.03 & -15.90 & -5.60 & -19.92 & -7.10 & -27.99 & -18.36 \\
\bottomrule
\end{tabular}
\end{center}
\caption{
\edit{{\bf Test -- Stereo performance with OpenCV FLANN, using kd-tree approximate nearest neighbors (8k features).}
kd-tree parameters are: 4 trees, 128 checks. We report {\bf (NI)} the number of inliers and {\bf mAA} at 10$^o$ for the stereo task, using DEGENSAC. }}
\label{tbl:test_flann}
\end{table}
\subsection[On the effect of 
local feature 
orientation]{On the effect of 
local feature 
orientation \edit{estimation --- Tables~\ref{tbl:test_upright}~and~\ref{tbl:test_upright_multiview}}}

In contrast with classical methods, which estimate the orientation of each keypoint, modern, end-to-end pipelines \cite{SuperPoint2017,D2Net2019,R2D22019} often skip this step, assuming that the images are roughly aligned (upright), with the descriptor shouldering the increased invariance requirements.
As our images meet this condition, we experiment with setting the orientation of keypoints to a fixed value (zero).
DoG often returns multiple orientations for the same keypoint, so we consider two variants: one where we simply remove keypoints which become duplicates after setting the orientation to a constant value \edit{(Upright)}, and a second one where we fill out the budget with new keypoints \edit{(Upright++)}.
We list the results in \tbl{test_upright}, for stereo, and \tbl{test_upright_multiview}, for multiview.
Performance increases across the board \edit{with Upright++} -- albeit by a small margin.

\subsection[On the effect of approximate nearest neighbor matching]{On the effect of approximate nearest neighbor matching \edit{--- \tbl{test_flann}}}
\label{imc:flann}

While it is known that approximate nearest neighbor search algorithms have non-perfect recall~\cite{FLANN2009}, it is not clear how their usage influences the downstream performance.
We thus compare the exact (brute-force) nearest neighbor search with a popular choice for the approximate nearest neighbor search, FLANN~\cite{FLANN2009}, as implemented in OpenCV.
We experimented with different parameters and found that 4 trees and 128 checks provide a reasonable trade-off between precision and runtime.

We report the results with and without FLANN in \tbl{test_flann}.

The performance drop varies for different methods: from moderate 1$\%$ for RootSIFT, to 5-7$\%$ for HardNet and SOSNet, and 18\% for D2-Net.

\begin{figure}
\centering
\renewcommand{\imsize}{5.3cm}
\renewcommand{\hspacer}{0.3mm}
\renewcommand{\vspacer}{0.1mm}
\renewcommand{\imscl}{0.01}
\includegraphics[scale=\imscl,width=\linewidth]{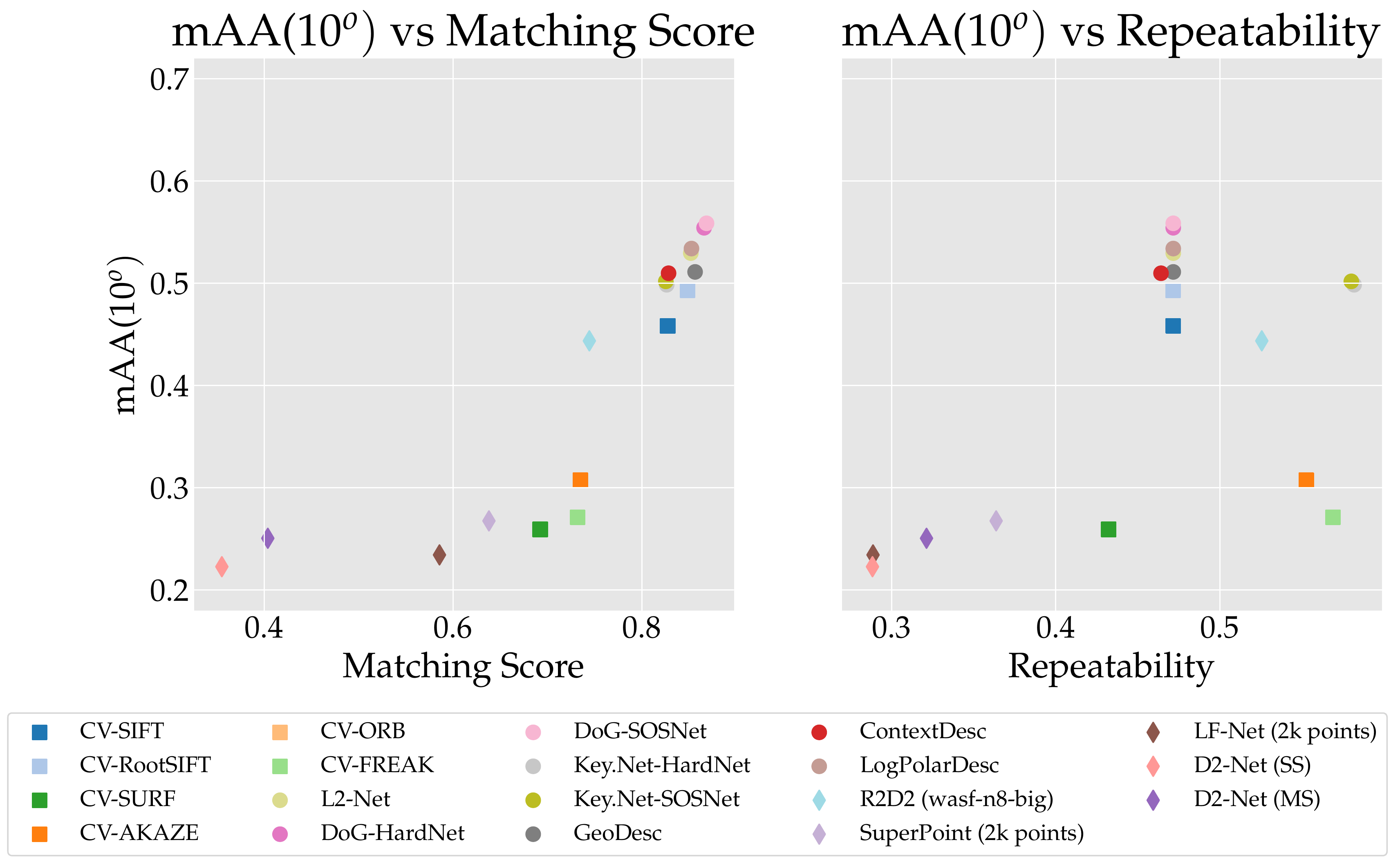}
\caption{{\bf Test -- Downstream vs. traditional metrics \revision{(8k features)}.}
We cross-reference stereo mAA at 10$^\circ$ with repeatability and matching score, with a 3-pixel threshold.
}
\label{fig:results_correlation}
\end{figure}

\subsection[Pose mAA vs. traditional metrics]{Pose mAA vs. traditional metrics \edit{--- \fig{results_correlation}}}
\label{imc:metrics}

\begin{figure}
\centering
\includegraphics[width=\linewidth]{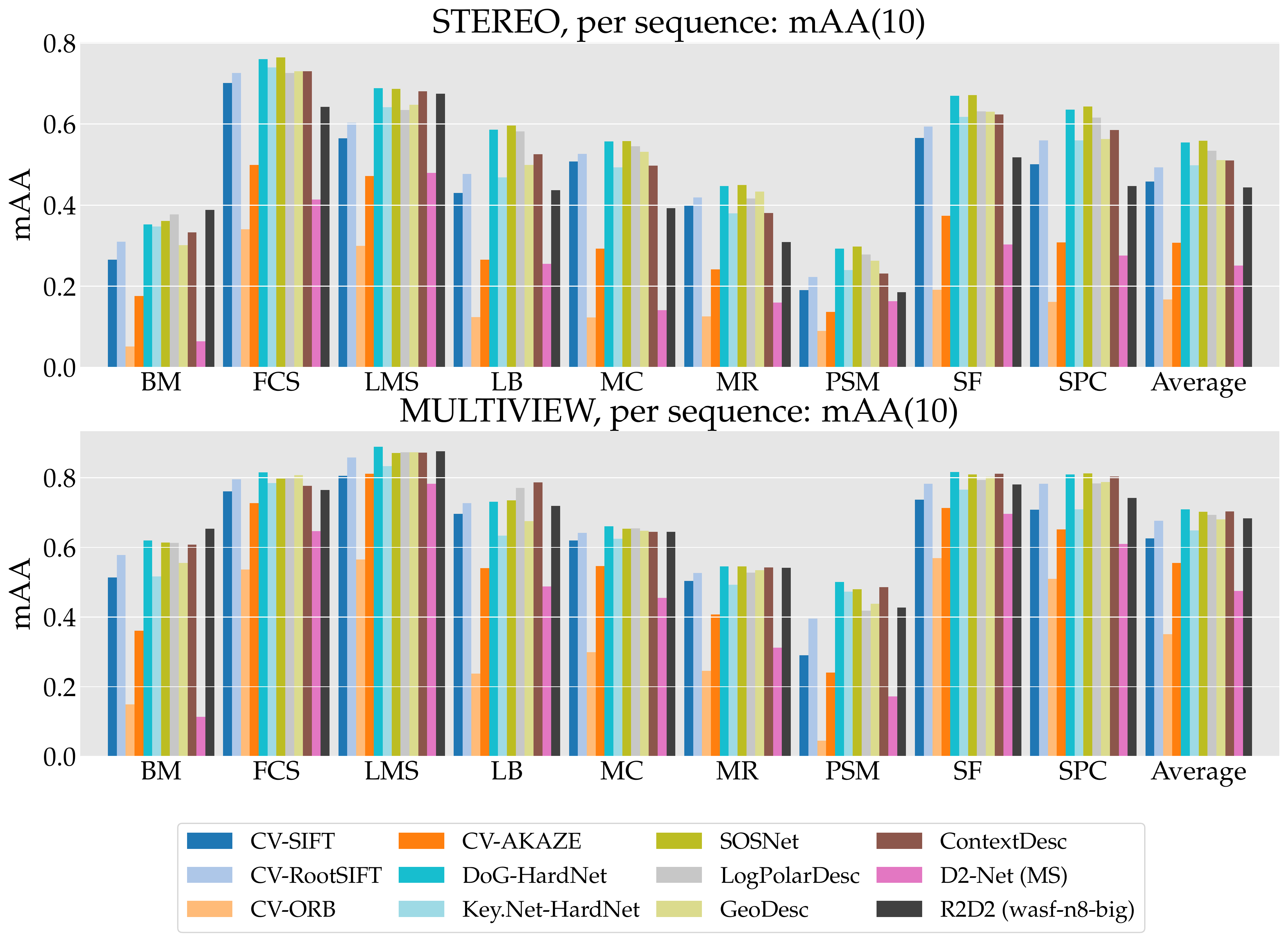}
\caption{
{\bf Test -- Breakdown by scene \revision{(8k features)}.}
\edit{For the stereo task, we use DEGENSAC.} Note how performance can vary drastically between scenes, and the relative rank of a given local feature fluctuates as well.
Please refer to \tbl{sequences_trainval} for a legend. 
}
\label{fig:sequences}
\vspace{-3mm}
\end{figure}

To examine the relationship between our pose metric and traditional metrics, we compare mAA against repeatability and matching score on the stereo task, with DEGENSAC.
While the matching score seems to correlate with mAA, repeatability is harder to interpret.
However, note that even for the matching score, which shows correlation, higher value does not guarantee high mAA -- see \eg RootSIFT vs ContextDesc.
We remind the reader that, as explained in \secref{benchmark}, our implementation differs from the classical formulation, as many methods do not have a strict notion of a support region.
We compute these metrics at a 3-pixel threshold, and provide more granular results in \secref{appendix}.

As shown, all methods based on DoG are clustered, as they operate on the same keypoints. 
Key.Net obtains the best repeatability, but performs worse than DoG in terms of mAA with the same descriptors (HardNet). 
AKAZE and FREAK perform surprisingly well in terms of repeatability -- \#2 and \#3, respectively -- but obtain a low mAA, which may be related to their descriptors, which are binary.
R2D2 shows good repeatability but a poor matching score and is outperformed by DoG-based features.

\subsection[Breakdown by scene]{Breakdown by scene \edit{--- \fig{sequences}}}
Results may vary drastically from scene to scene, as shown in \fig{sequences}.
A given method may also perform better on some than others -- for instance, D2-Net nears the state of the art on ``Lincoln Memorial Statue'', but is 5x times worse on ``British Museum''.
AKAZE and ORB show similar behaviour.
This can provide valuable insights on limitations and failure cases.

\begin{figure}
\centering
\renewcommand{\imsize}{5.2cm}
\renewcommand{\hspacer}{0.3mm}
\renewcommand{\vspacer}{0.1mm}
\renewcommand{\imscl}{0.01}
\includegraphics[width=\linewidth]{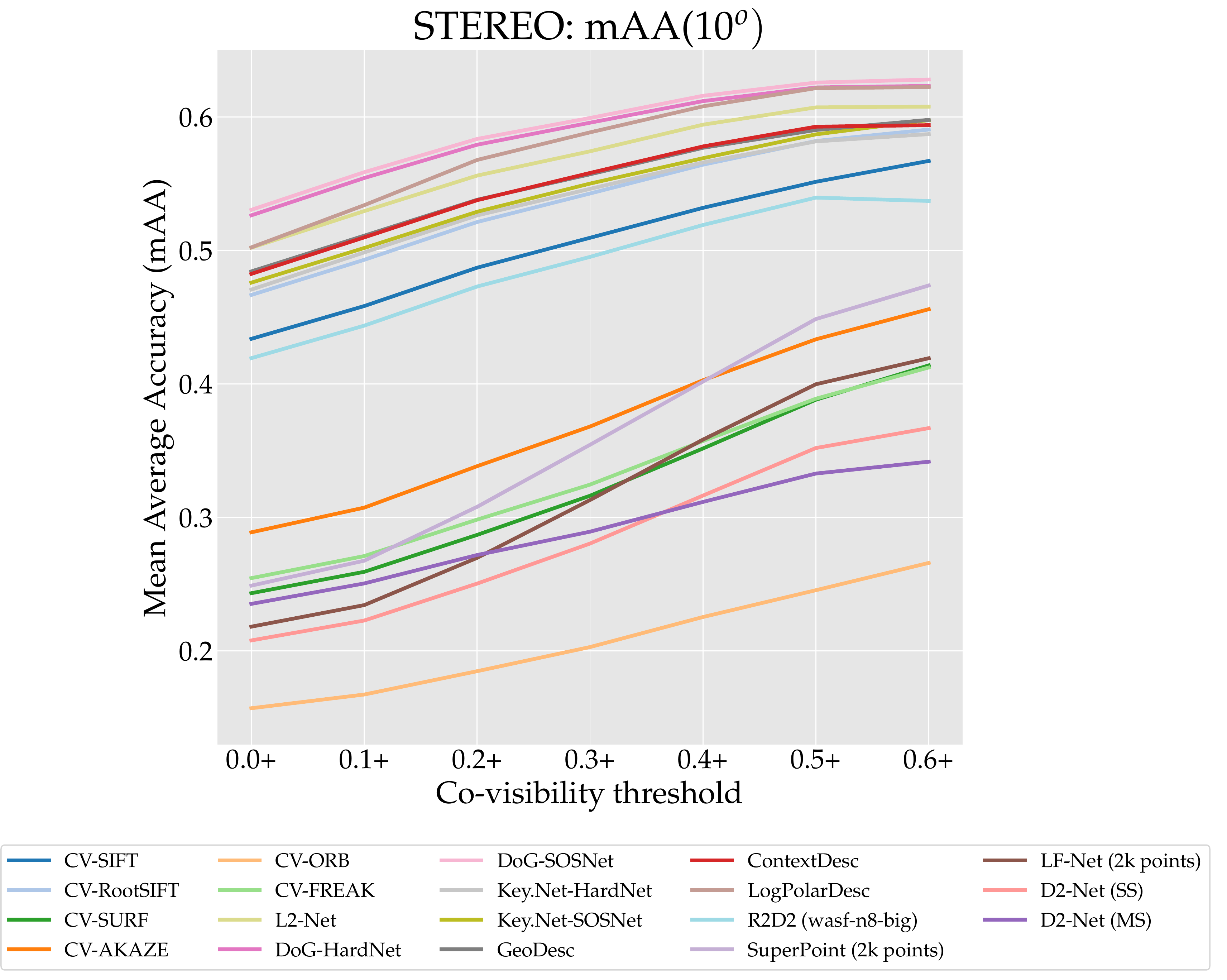}
\caption{{\bf Test -- Local features vs. co-visibility \revision{(8k features)}.}
\edit{We plot mAA at 10$^\circ$ on the stereo task -- using DEGENSAC -- at different co-visibility thresholds for different local feature types. 
Bin ``0+'' consists of all possible image pairs, including potentially unmatchable ones.
Bin ``0.1+'' includes pairs with a minimum co-visibility value of 0.1 -- this is the default value we use for all other experiments in this paper -- and so forth.
Results are mostly consistent, with end-to-end methods performing better at higher than lower co-visibility.}
}
\label{fig:results_visibility}
\end{figure}
\begin{figure}
\centering
\renewcommand{\imsize}{5.2cm}
\renewcommand{\hspacer}{0.3mm}
\renewcommand{\vspacer}{0.1mm}
\renewcommand{\imscl}{0.01}
\includegraphics[width=\linewidth]{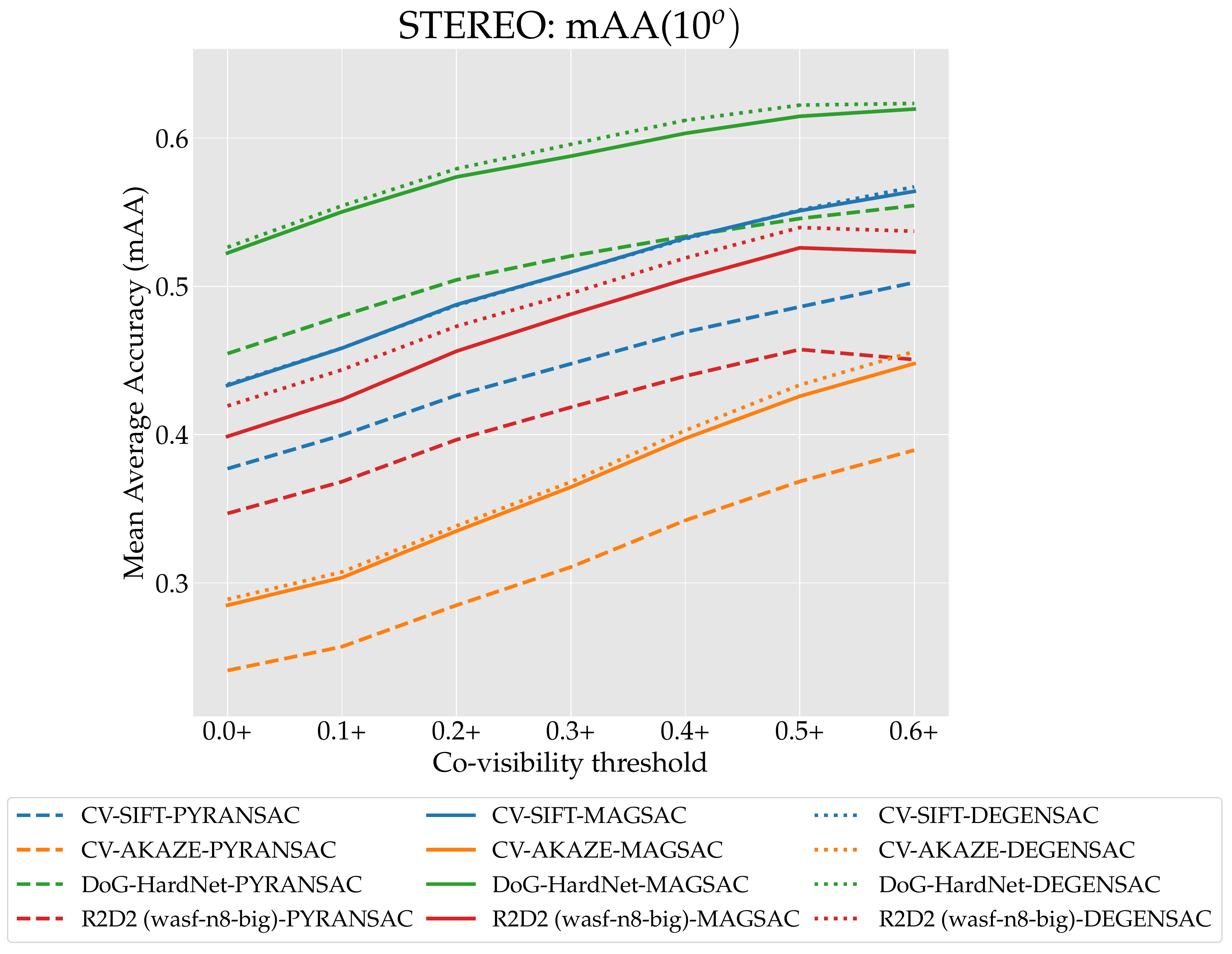}
\caption{{\bf Test -- RANSAC vs. co-visibility \revision{(8k features)}.} \edit{We plot mAA at 10$^\circ$ on the stereo task at different co-visibility thresholds for different RANSAC variants, binning the results as in \fig{results_visibility}. 
We pair them with different local features methods.
The difference in performance between RANSAC variants seems consistent across pairs, regardless of their difficulty.}
}
\label{fig:results_visibility_ransac}
\end{figure}

\subsection[Breakdown by co-visibility ]{Breakdown by co-visibility 
\edit{--- Figs.~\ref{fig:results_visibility}~and~\ref{fig:results_visibility_ransac}}}

\edit{\edit{Next, in \fig{results_visibility} we evaluate stereo performance at different co-visibility thresholds, for several local feature methods, using DEGENSAC.}
Bins are encoded as $v$+, with $v$ the co-visibility threshold, and include all image pairs with a co-visibility value larger or equal than $v$. 
This means that the first bin may contain unmatchable images -- we use 0.1+ for all other experiments in the Chapter. 
We do not report values above 0.6+ as 
\edit{there are only a handful of them and thus the results are very noisy.}

Performance for all local features and RANSAC variants increases with the co-visibility threshold, as expected.
Results are consistent, with end-to-end methods performing better at higher than \edit{co-visibility} than lower, and single-scale D2-Net outperforming its multi-scale counterpart at 0.4+ and above, where the images are more likely aligned in terms of scale.

We also break down different RANSAC methods in \fig{results_visibility_ransac}, along with different local fetures, including AKAZE (binary), SIFT (also handcrafted), HardNet (learned descriptor on DoG points) and R2D2 (learned end-to-end). 
We do not observe significant variations \edit{in the trend of each curve as we \edit{swap} RANSAC and local feature methods}.
DEGENSAC and MAGSAC show very similar performance.}

\begin{figure}
\centering
\renewcommand{\hspacer}{0.3mm}
\renewcommand{\vspacer}{0.1mm}
\renewcommand{\imscl}{0.01}
\includegraphics[scale=\imscl,width=\linewidth]{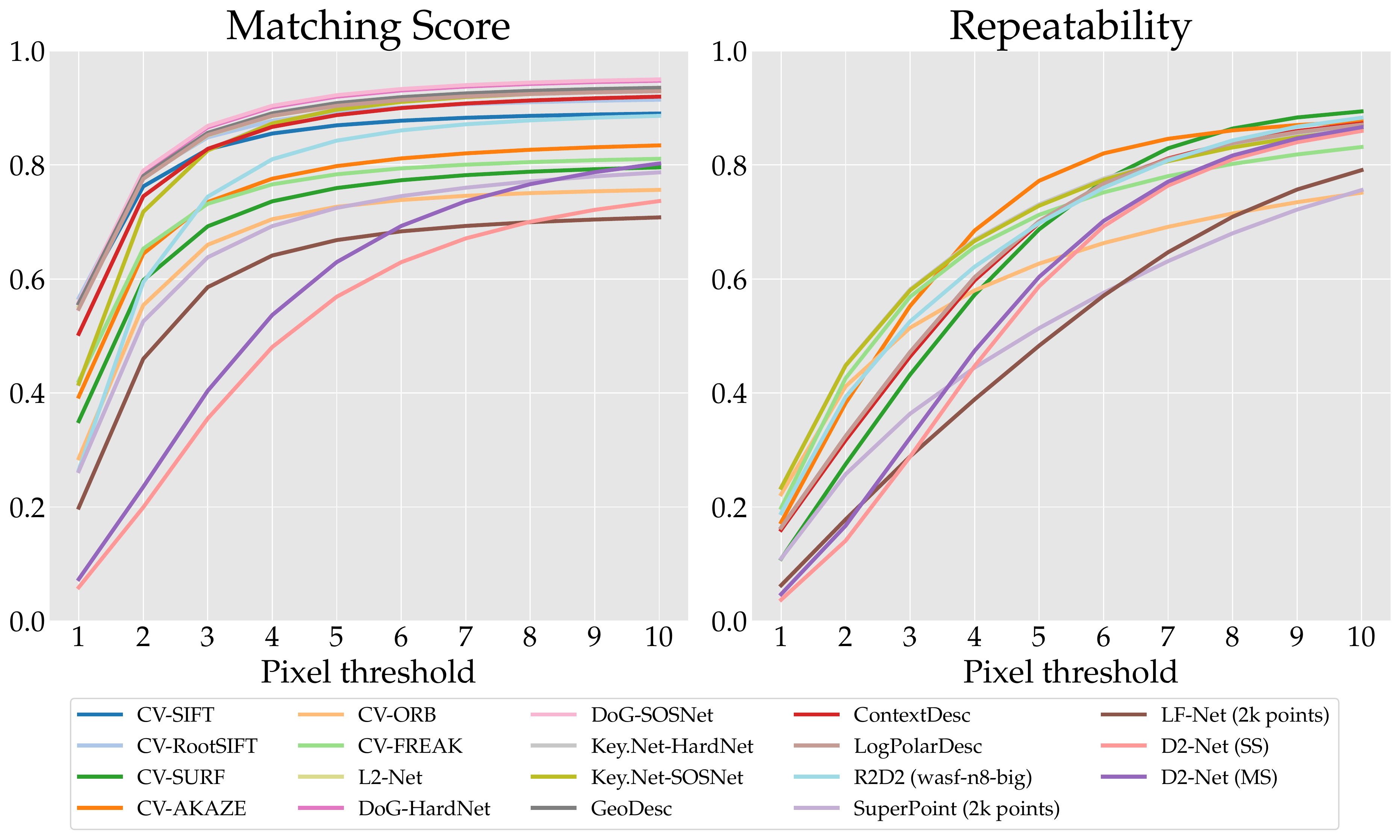}\\
\caption{{\bf Test -- Classical metrics, by pixel threshold \revision{(8k features)}.} Repeatability and matching score computed at different pixel thresholds are shown.}
\label{fig:results_repeatability}
\end{figure}
\subsection[Classical metrics vs. pixel threshold ]{Classical metrics vs. pixel threshold --- \fig{results_repeatability}}

\edit{In \fig{results_correlation} we plot repeatability and matching score against mAA at a fixed error threshold of 3 pixels. 
In \fig{results_repeatability} we show them at different pixel thresholds. 
End-to-end methods tend to perform better at higher pixel thresholds, which is expected -- D2-Net in particular extracts keypoints from downsampled feature maps. 
These results are computed from the depth maps estimated by COLMAP, which are not pixel-perfect, so the results for very low thresholds are not completely trustworthy.

Note that repeatability is typically lower than matching score, which might be counter-intuitive as the latter is more strict -- it requires two features to be nearest-neighbors in descriptor space in addition to being physically close (after reprojection). 
We compute repeatability with the raw set of keypoints, whereas the matching score is computed with optimal matching settings -- bidirectional matching with the ``both'' strategy and the ratio test. This results in a much smaller pool -- 8k features are typically narrowed down to \edit{200--400} matches (see \tbl{results_supp_inliers}).
This better isolates the performance of the detector and the descriptor where it matters.}

\begin{figure}
\centering
\includegraphics[width=\linewidth]{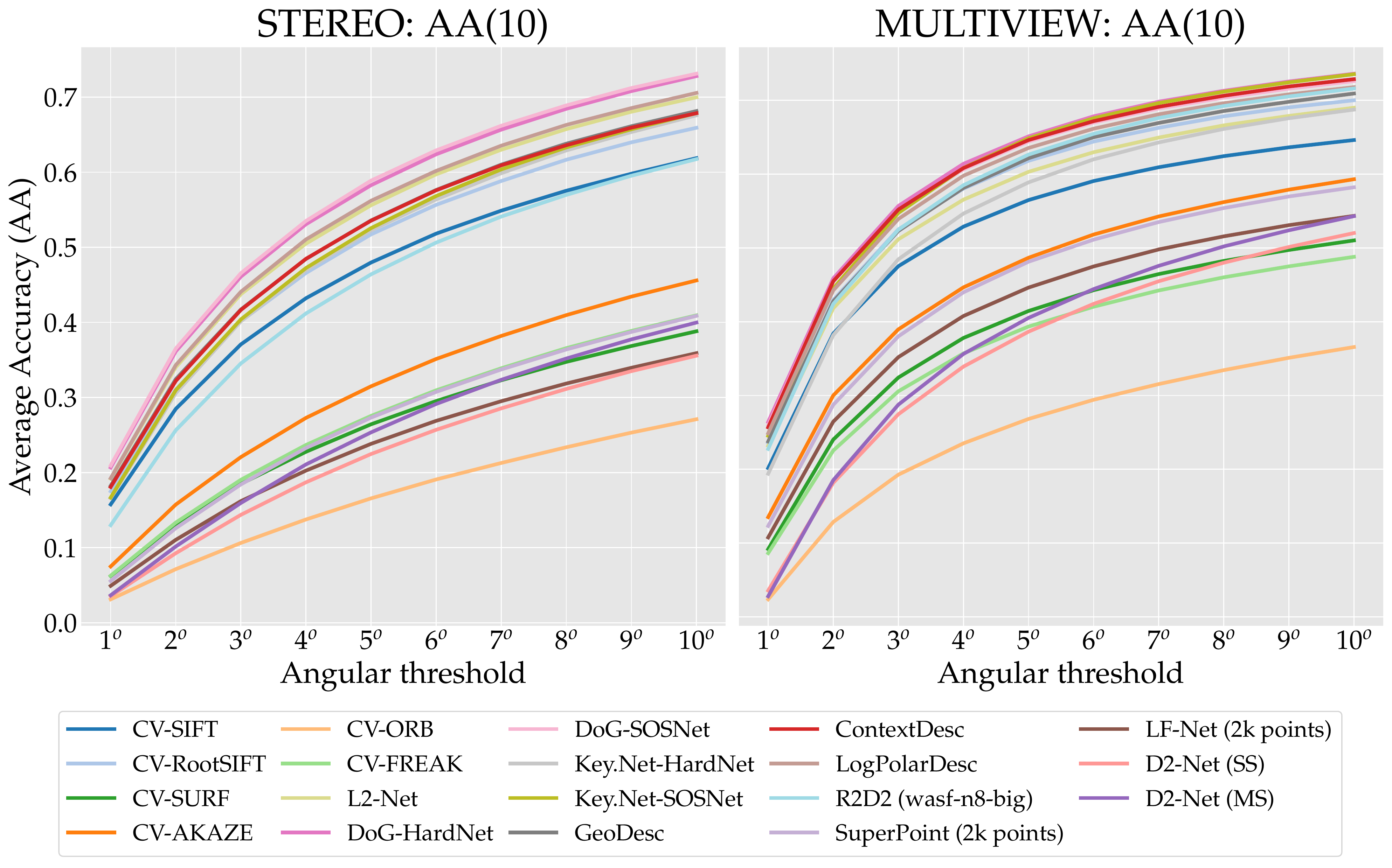}
\caption{
\textbf{Test -- Breakdown by angular threshold, for local features \revision{(8k features)}.}
\edit{Accuracy in pose estimation is evaluated at every error threshold. Mean Average Accuracy used in the rest of paper is the area under this curve. Ranks are consistent across thresholds.}
}
\label{fig:map_granular_features}
\end{figure}
\begin{figure}
\centering
\includegraphics[width=\linewidth]{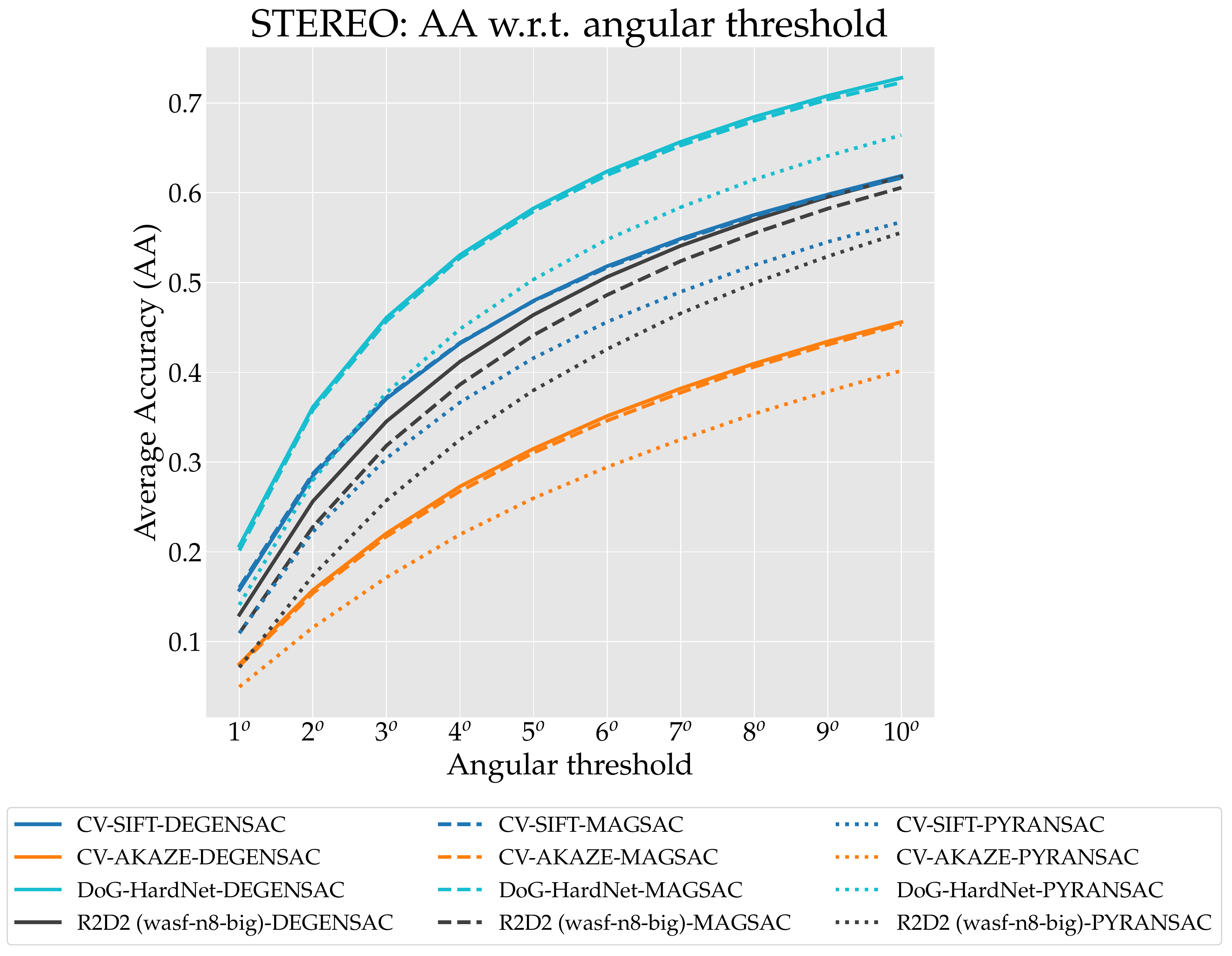}
\caption{
{\bf Test -- Breakdown by angular threshold, for RANSAC \revision{(8k features)}.}
\edit{We plot performance on the stereo task at different angular error thresholds, for \edit{four} different local features and three RANSAC variants: MAGSAC (solid line), DEGENSAC (dashed line), and PyRANSAC (dotted line).
We observe a similar behaviour across all thresholds.}
}
\label{fig:map_granular_geom}
\end{figure}
\subsection[Breakdown by 
angular threshold]{Breakdown by 
angular threshold --- Figs.~\ref{fig:map_granular_features}~and~\ref{fig:map_granular_geom}}

We summarize pose accuracy by mAA at 10$^\circ$ in order to have a single, 
easy-to-interpret number.
In this section we show how performance varies across different error thresholds -- we look at the {\em average accuracy} at every angular error threshold, rather than the mean Average Accuracy. \fig{map_granular_features} plots performance on stereo and multiview for different local feature types, showing that the ranks remain consistent across thresholds.
\fig{map_granular_geom} shows how it affects different RANSAC variants, with four different local features.
Again, ranks do not change. DEGENSAC and MAGSAC perfom nearly identically for all features, except for R2D2.
The consistency in the ranks demonstrate that summarizing the results with a single number, mAA at 10$^\circ$, is reasonable.
\subsection[Qualitative results]{Qualitative results --- Figs.~\ref{fig:supp_stereo},~\ref{fig:supp_stereo2},~\ref{fig:supp_multiview},~and~\ref{fig:supp_multiview2}}
Figs.~\ref{fig:supp_stereo}~and~\ref{fig:supp_stereo2} show qualitative results for the stereo task. 
We draw the inliers produced by DEGENSAC and color-code them using the ground truth depth maps, from green (0) to yellow (5 pixels off) if they are correct, and in red if they are incorrect (more than 5 pixels off). 
Matches including keypoints which fall on occluded pixels are drawn in blue. 
Note that while the depth maps are somewhat noisy and not pixel-accurate, they are sufficient for this purpose.
\edit{Notice how the best performing methods have more correct matches that are well spread across the overlapping region.}

\fig{supp_multiview} shows qualitative results for the multiview task, for handcrafted detectors. 
We illustrate it by drawing keypoints used in the SfM reconstruction in blue and the rest in red. 
It showcases the importance of the detector, specially on unmatchable regions such as the sky. 
ORB and Hessian keypoints are too concentrated in the high-contrast regions, failing to provide evenly-distributed features.
In contrast, SURF fails to filter-out background and sky points. 
\edit{\fig{supp_multiview2} shows results for learned methods: DoG+HardNet, Key.Net+HardNet, SuperPoint, R2D2, and D2-Net (multiscale). 
They have different detection patterns: Key.Net resembles a ``cleaned'' version of DoG, while R2D2 seems to be evenly distributed. D2Net looks rather noisy, while SuperPoint fires precisely on corner points on structured parts, and sometimes form a regular grid on sky-like homogeneous regions, which might be due to \ky{the method} running out of locations to place points at -- note however that the best results were obtained with larger NMS (non-maxima suppression) thresholds}.

\section{Further results and considerations}
\label{imc:appendix}

In this section we provide additional results on the validation set.
\edit{These include}
a study of the typical outlier ratios under optimal settings in \secref{appendix-inliers}, matching with FGINN in \secref{fginn}, image-preprocessing techniques for feature extraction in \secref{appendix-clahe}, and a breakdown of the optimal settings in \secref{optimal-settings}, provided \edit{to serve as a} reference.

\renewcommand{\arraystretch}{1}
\renewcommand{\tabcolsep}{1.2mm}
\begin{table}
\small
\begin{center}
\begin{tabular}{@{}lcccc@{}}
\toprule
Method & \# matches & \# inliers & Ratio (\%) & mAA(10$^\circ$) \\
\midrule
CV-SIFT & 328.3 & 113.0 & 34.4 & 0.548 \\
\midrule
CV-$\sqrt{\text{SIFT}}$& 331.6& 131.4& 39.6& 0.584 \\
CV-SURF& 221.5& 77.4& 35.0& 0.309 \\
CV-AKAZE& 369.1& 143.5& 38.9& 0.360 \\
CV-ORB& 193.4& 74.7& 38.6& 0.265 \\
CV-FREAK & 216.6 & 75.5 & 34.8 & 0.329 \\
\midrule
DoG-HardNet& 433.1& 173.1& 40.0& 0.627 \\
Key.Net-HardNet & 644.1 & 166.6 & 25.9 & 0.580 \\
L2Net& 368.0& 144.0& 39.1& 0.601 \\
GeoDesc& 322.6& 133.4& 41.3& 0.564 \\
ContextDesc & 560.8 & 165.9 & 29.6 & 0.587 \\
SOSNet& 391.7& 161.8& 41.3& 0.621 \\
LogPolarDesc& 522.8& 175.2& 33.5& 0.618 \\
\midrule
SuperPoint (2k points)& 202.0& 62.6& 31.0& 0.312 \\
LF-Net (2k points)& 165.9& 73.2& 44.1& 0.293 \\
D2-Net (SS)& 657.3& 148.9& 22.7& 0.263 \\
D2-Net (MS)& 601.7& 161.7& 26.9& 0.343 \\
R2D2 (wasf-n8-big) & 901.5 & 477.3 & 52.9 & 0.473 \\
\bottomrule
\end{tabular}
\end{center}
\caption{
\edit{{\bf Validation -- Number of inliers with optimal settings.} We use 8k features with optimal parameters, and  PyRANSAC as a robust estimator. The number of inliers varies significantly between methods, despite tuning the matcher and the ratio test, and lower inlier ratios tend to correlate with low performance.}
}
\label{tbl:results_supp_inliers}
\end{table}
\subsection[Number of inliers per step]{Number of inliers per step --- \tbl{results_supp_inliers}}
\label{imc:appendix-inliers}

\edit{We list the number of input matches and their \emph{resulting} inliers for the stereo task, in \tbl{results_supp_inliers}.
\edit{As before, we remind the reader that these inliers are what each method reports, \ie, the \edit{matches} that are actually used to estimate the poses.}
We list the number of input matches, the number of inliers produced by each method (which may still contain outliers), their ratio, and the mAA at 10$^\circ$.
We use PyRANSAC with optimal settings for each method, the ratio test, and bidirectional matching with the ``both'' strategy.

We see that inlier-to-outlier ratios hover around 35--40\% for all features relying on classical detectors. 
Key.Net with HardNet descriptors sees a significant drop in inlier ratio and mAA, when compared to its DoG counterpart.
D2-Net similarly has inlier ratios around 25\%. 
R2D2 has the largest inlier ratio by far at 53\%, but is outperformed by many other methods in terms of mAA, suggesting that many of these are not actual inliers.
In general, we observe that the methods which produce a large number of matches, such as Key.Net (600+), D2-Net (600+) or R2D2 (900+) are less accurate in terms of pose estimation.}

\begin{figure}
\centering
\renewcommand{\imsize}{4.3cm}
\renewcommand{\hspacer}{0.3mm}
\renewcommand{\vspacer}{0.1mm}
\renewcommand{\imscl}{0.01}
\includegraphics[scale=\imscl,width=0.99\linewidth]{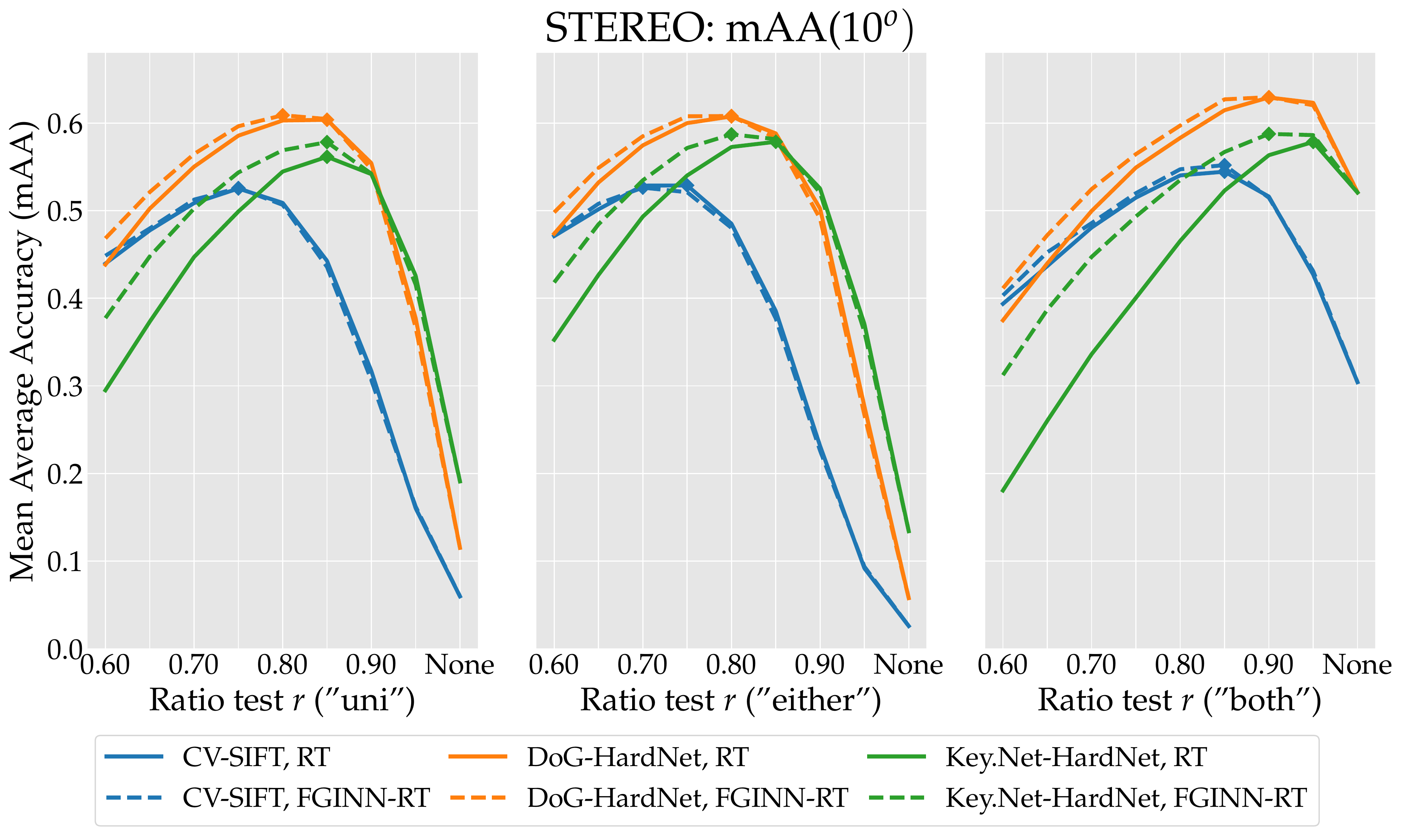}
\caption{{\bf Validation -- FGINN vs. ratio test  \revision{(8k features)}.} We evaluate the ratio test with FGINN~\cite{MODS2015} (dashed line), and the standard ratio test (solid line). With FGINN, the valid range for $r$ -- the ratio test threshold -- is significantly wider, but the best performance with the ``both'' matching strategy is not significantly better than for the standard ratio test.
}
\label{fig:results_fginn}
\end{figure}

\subsection[Feature matching with FGINN]{Feature matching with an advanced ratio test --- \fig{results_fginn}}
\label{imc:fginn}

We also compare the benefits of applying first-geometrically-inconsistent-neighbor-ratio~(FGINN) \cite{MODS2015} to DoG/SIFT, DoG/HardNet and Key.Net/HardNet, against Lowe's standard ratio test~\cite{SIFT2004}.
FGINN performs the ratio-test with second-nearest neighbors that are ``far enough'' from the tentative match (10 pixels in \cite{MODS2015}).
In other words, it loosens the test to allow for nearby-thus-similar points. 
We test it for 3 matching strategies: unidirectional (``uni''), ``both'' and ``either''.
We report the results in \fig{results_fginn}.
As shown, FGINN provides minor improvements over the standard ratio test in case of unidirectional matching, and not as much when ``both'' is used. 
It also behaves differently compared to the standard strategy, in that performance at stricter thresholds degrades less.

\begin{figure}
\centering
\renewcommand{\imsize}{5.2cm}
\renewcommand{\hspacer}{0.3mm}
\renewcommand{\vspacer}{0.1mm}
\renewcommand{\imscl}{0.01}
\includegraphics[scale=\imscl,width=0.49\linewidth]{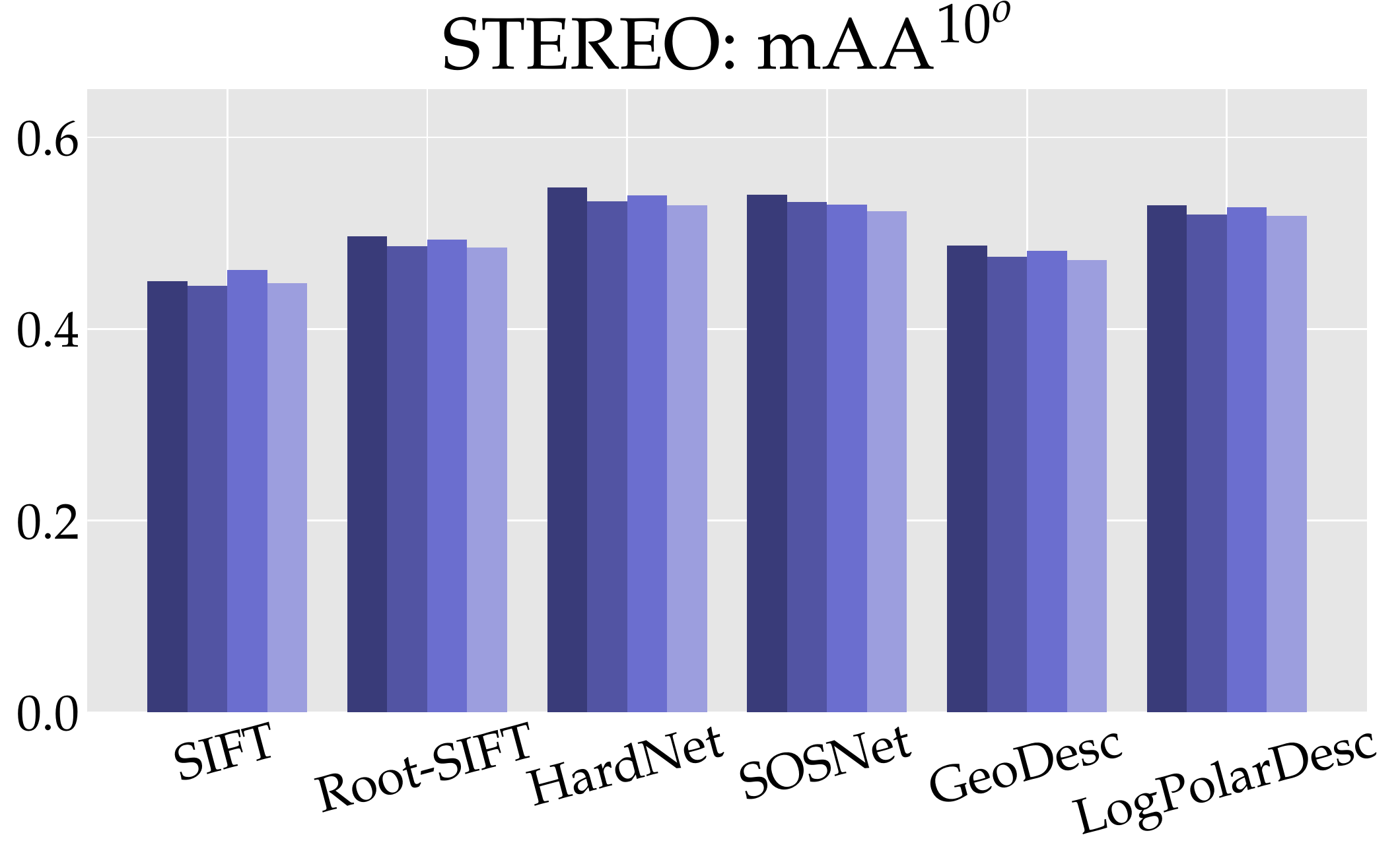}
\includegraphics[scale=\imscl,width=0.49\linewidth]{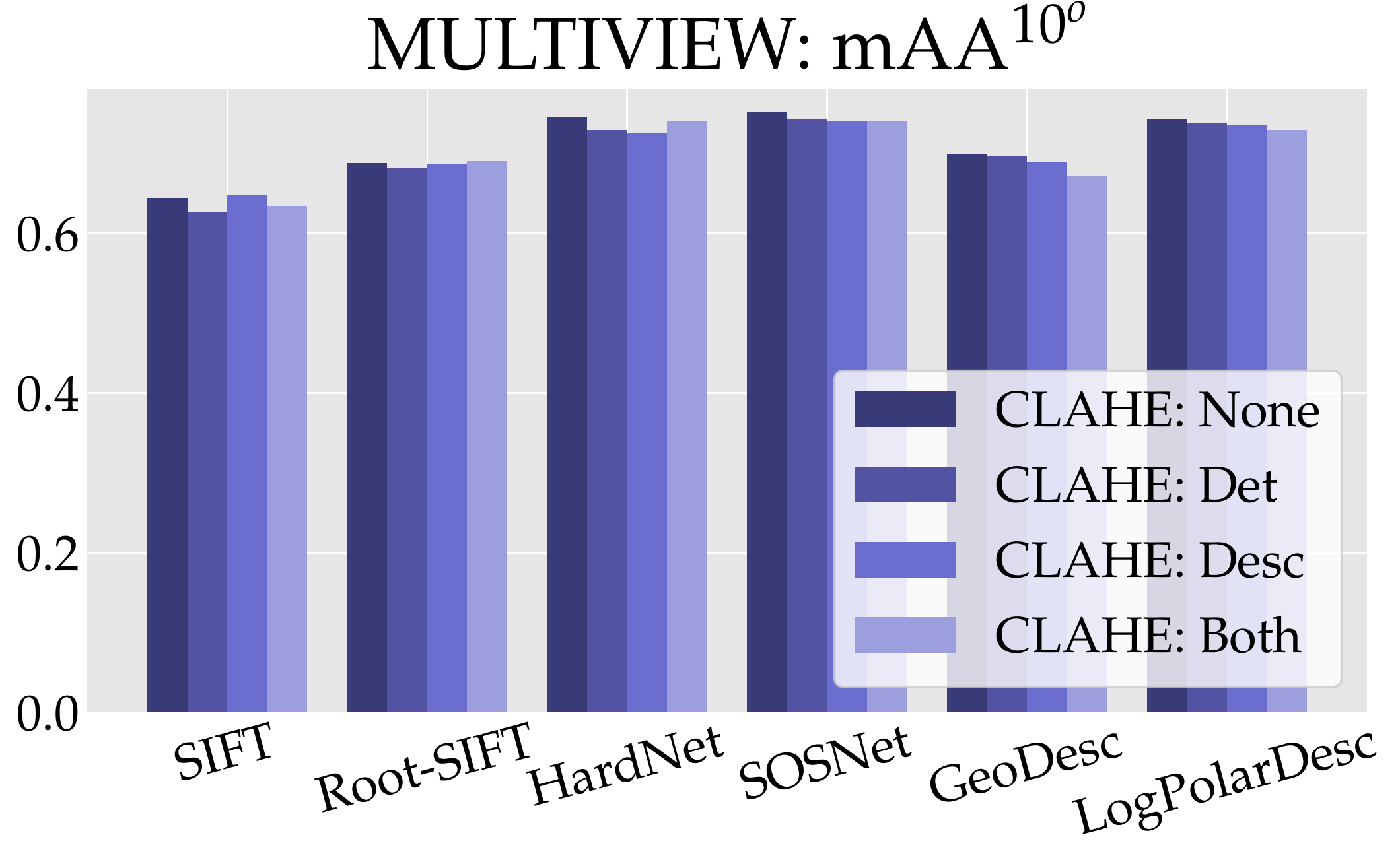}%
\caption{
\edit{{\bf Validation -- Image pre-processing with CLAHE \revision{(8k features)}.} 
We experiment with contrast normalization for keypoint and descriptor extraction.
Results are very similar with or without it.}}
\label{fig:results_clahe}
\end{figure}
\subsection[Image pre-processing]{Image pre-processing --- \fig{results_clahe}}
\label{imc:appendix-clahe}

Contrast normalization is key to invariance against illumination changes -- local feature methods typically apply some normalization strategy over small patches~\cite{SIFT2004,HardNet2017}.
Therefore, we experiment with contrast-limited adaptive histogram equalization (CLAHE) \cite{CLAHE1994}, as implemented in OpenCV.
We apply it prior to feature detection and/or description with SIFT and several learned descriptors, and display the results in \fig{results_clahe}.
Performance decreases for all learned methods, presumably because they are not trained for it. 
Contrary to our initial expectations, SIFT does not benefit much from it either: the only increase in performance comes from applying it for descriptor extraction, at 2.5\% relative for stereo task and 0.56\% relative for multi-view. 
This might be due to the small number of night-time images in our data. It also falls in line with the observations in~\cite{Dong_2015_CVPR}, which show that SIFT descriptors are actually optimal under certain assumptions.
\definecolor{light-gray}{gray}{0.9}
\renewcommand{\arraystretch}{0.95}
\renewcommand{\tabcolsep}{1.5mm}
\begin{table}
\begin{center}
\scriptsize
\begin{tabular}{@{}lcccccc@{}}
\toprule
& $\ransacReprojTh_{\text{PyR}}$ & $\ransacReprojTh_{\text{DEGEN}}$ & $\ransacReprojTh_{\text{GCR}}$ & $\ransacReprojTh_{\text{MAG}}$ & $r_{\text{stereo}}$ & $r_{\text{multiview}}$ \\
\midrule
CV-SIFT & 0.25 & 0.5 & 0.5 & 1.25 & 0.85 & 0.75 \\
\midrule
CV-$\sqrt{\text{SIFT}}$ & 0.25 & 0.5 & 0.5 & 1.25 & 0.85 & 0.85 \\
CV-SURF & 0.75 & 0.75 & 0.75 & 2 & 0.85 & 0.90 \\
CV-AKAZE & 0.25 & 0.75 & 0.75 & 1.5 & 0.85 & 0.90 \\
CV-ORB & 0.75 & 1 & 1.25 & 2 & 0.85 & 0.95 \\
CV-FREAK & 0.5 & 0.5 & 0.75 & 2 & 0.85 & 0.85 \\
\midrule
VL-DoG-SIFT & 0.25 & 0.5 & 0.5 & 1.5 & 0.85 & 0.80 \\
VL-DoGAff-SIFT & 0.25 & 0.5 & 0.5 & 1.5 & 0.85 & 0.80 \\
VL-Hess-SIFT & 0.2 & 0.5 & 0.5 & 1.5 & 0.85 & 0.80 \\
VL-HessAffNet-SIFT & 0.25 & 0.5 & 0.5 & 1& 0.85 & 0.80 \\
\midrule
CV-DoG/HardNet & 0.25 & 0.5 & 0.5 & 1.5 & 0.90 & 0.80 \\
KeyNet/Hardnet & 0.5 & 0.75 & 0.75 & 2 & 0.95 & 0.85 \\
KeyNet/SOSNet & 0.25 & 0.75 & 0.75 & 1.5 & 0.95 & 0.95  \\
CV-DoG/L2Net & 0.2 & 0.5 & 0.5 & 1.5 & 0.90 & 0.80 \\
CV-DoG/GeoDesc & 0.2 & 0.5 & 0.75 & 1.5 & 0.90 & 0.85 \\
ContextDesc & 0.25 & 0.75 & 0.5 & 1 & 0.95 & 0.85 \\
CV-DoG/SOSNet & 0.25 & 0.5 & 0.75 & 1.5 & 0.90 & 0.80 \\
CV-DoG/LogPolarDesc & 0.2 & 0.5 & 0.5 & 1.5 & 0.90 & 0.80 \\
\midrule
D2-Net (SS) & 1 & 2 & 2 & 7.5 & --- & --- \\
D2-Net (MS) & 1 & 2 & 2 & 5 & --- & --- \\
R2D2 (wasf-n8-big) & 0.75 & 1.25 & 1.25 & 2 & --- & 0.95 \\
\midrule
CV-DoG/TFeat & 0.25 & 0.5 & -- & 1.25 & 0.85 & 0.80 \\
CV-DoG/MKD-Concat & 0.25 & 0.5 & -- & 1.25 & 0.85 & 0.80 \\
CV-DoGAffNet/HardNet & 0.25 & 0.5 & -- & 1.25 & 0.85 & 0.85 \\
\bottomrule
\end{tabular}
\end{center}
\caption{
{\bf Optimal hyper-parameters with 8k features.}
\edit{We summarize the optimal hyperparameters -- the maximum number of RANSAC iterations $\ransacReprojTh$ and the ratio test threshold $r$ -- for each combination of methods.
The number of RANSAC iterations $\ransacMaxIters$ is set to 250k for PyRANSAC, 50k for DEGENSAC, and 10k for both GC-RANSAC and MAGSAC (for the experiments on the validation set).
We use bidirectional matching with the ``both'' strategy.}
}
\label{tbl:optimal_parameters_8k}
\end{table}

\definecolor{light-gray}{gray}{0.9}
\renewcommand{\arraystretch}{0.95}
\renewcommand{\tabcolsep}{1.5mm}
\begin{table}
\begin{center}
\footnotesize
\begin{tabular}{@{}lccccc@{}}
\toprule
& $\ransacReprojTh_{\text{PyR}}$ & $\ransacReprojTh_{\text{DEGEN}}$ & $\ransacReprojTh_{\text{MAG}}$ & $r_{\text{stereo}}$ & $r_{\text{multiview}}$ \\
\midrule
CV-SIFT & 0.75 & 0.5 & 2 & 0.90 & 0.90 \\
\midrule
CV-$\sqrt{\text{SIFT}}$ & 0.5 & 0.5 & 1.25 & 0.90 & 0.90 \\
CV-SURF & 0.25 & 1 & 3 & 0.90 & 0.95 \\
CV-AKAZE & 1 & 0.75 & 2 & 0.90 & 0.95 \\
CV-ORB & 1 & 1.25 & 3 & 0.90 & 0.90 \\
CV-FREAK & 0.75 & 0.75 & 2 & 0.9 & 0.95 \\
\midrule
CV-DoG/HardNet & 0.5 & 0.5 & 1.5 & 0.95 & 0.90 \\
KeyNet/HardNet & 0.5 & 0.75 & 2.5 & 0.95 & 0.90 \\
KeyNet/SOSNet & 0.5 & 0.75 & 1.5 & 0.95 & 0.95 \\
CV-DoG/L2Net & 0.25 & 0.5 & 1.5 & 0.90 & 0.90 \\
CV-DoG/GeoDesc & 0.5 & 0.5 & 1.5 & 0.95 & 0.90 \\
ContextDesc& 0.75 & 0.75 & 2 & 0.95 & 0.95  \\
CV-DoG/SOSNet & 0.5 & 0.75 & 1.5 & 0.95 & 0.90 \\
CV-DoG/LogPolarDesc & 0.5 & 0.5 & 1.5 & 0.95 & 0.90 \\
\midrule
D2-Net (SS) & 1.5 & 2 & 7.5 & --- & --- \\
D2-Net (MS) & 1.5 & 2 & 10 & --- & --- \\
SuperPoint & 0.75 & 1 & 3 & 0.95 & 0.90 \\
LF-Net & 1 & 1 & 4 & 0.95 & 0.95 \\
R2D2 (wasf-n16) & 1.5 & 1.25 & 2 & --- & --- \\
\midrule
CV-DoGAffNet/HardNet & 0.25 & 0.5 &  1.25 & 0.95 & 0.95 \\
\bottomrule
\end{tabular}
\end{center}
\caption{
{\bf Optimal hyper-parameters with 2k features.}
\edit{Equivalent to \tbl{optimal_parameters_8k}. We do not evaluate GC-RANSAC, as it is always outperformed by DEGENSAC and MAGSAC, but keep PyRANSAC as a baseline RANSAC.}
}
\label{tbl:optimal_parameters_2k}
\end{table}

\subsection[Optimal settings breakdown]{Optimal settings breakdown --- Tables~\ref{tbl:optimal_parameters_8k}~and~\ref{tbl:optimal_parameters_2k}}
\label{imc:optimal-settings}
For the sake of clarity, we summarize the optimal hyperparameter combinations from Figs. \ref{fig:plots-ransac-cost-vs-map}, \ref{fig:results_ransac_nort} and \ref{fig:results_both_vs_any_and_ratio_test} in \tbl{optimal_parameters_8k} (for 8k features) and \tbl{optimal_parameters_2k} (for 2k features). 
We set the confidence value to $\ransacConf{=}0.999999$ for all RANSAC variants. 
\edit{Notice how it is better to have more features and a stricter ratio test threshold to filter them out, than having fewer features from the beginning.}
\begin{figure*}
\centering
\renewcommand{\arraystretch}{0.5}
\renewcommand{\tabcolsep}{0.2mm}
\renewcommand{\imsize}{3.0cm}
\renewcommand{\hspacer}{0.3mm}
\renewcommand{\vspacer}{0.1mm}
\renewcommand{\imscl}{0.01}
\small
\begin{center}
\begin{tabular}{@{}ccccc@{}}
\includegraphics[scale=\imscl,width=\imsize]{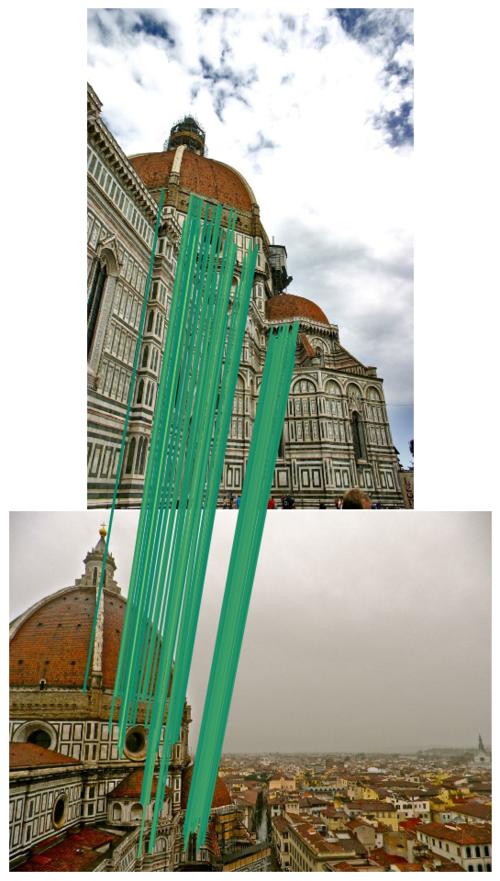} &
\includegraphics[scale=\imscl,width=\imsize]{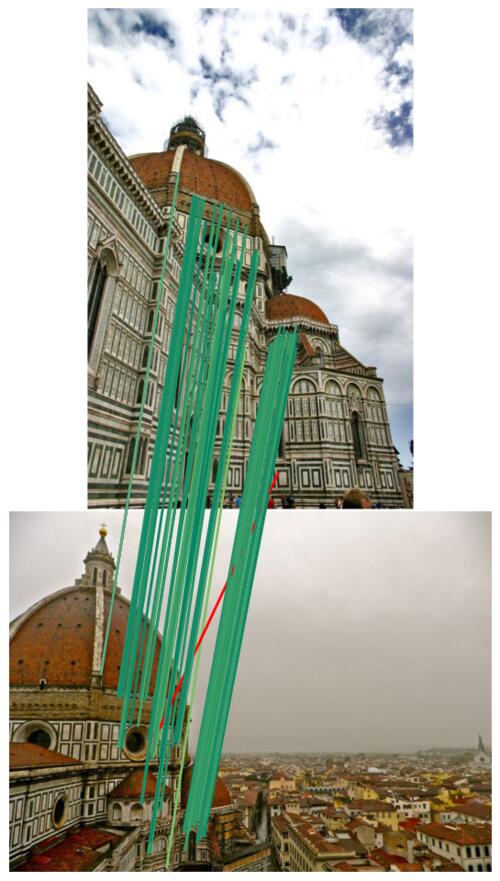} &
\includegraphics[scale=\imscl,width=\imsize]{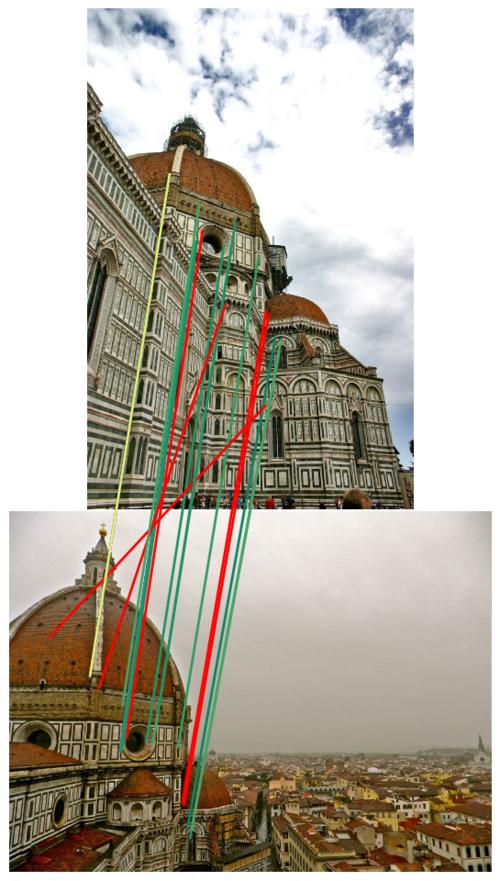} &
\includegraphics[scale=\imscl,width=\imsize]{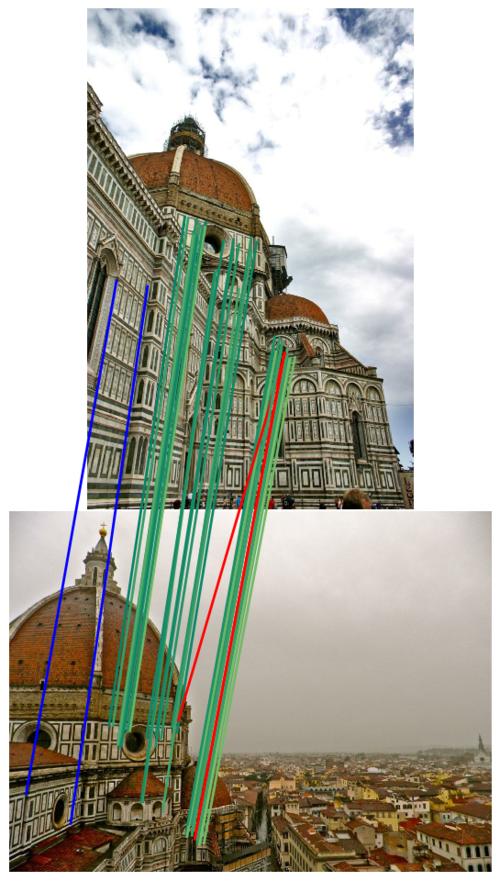} &
\includegraphics[scale=\imscl,width=\imsize]{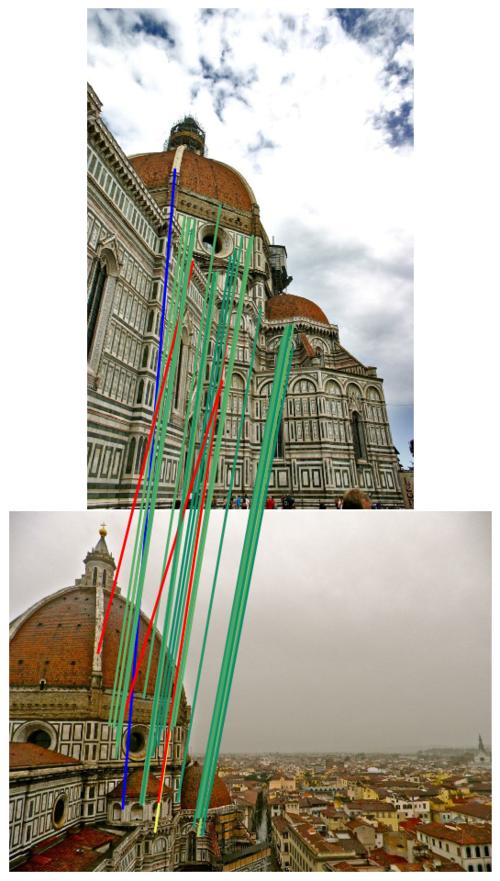}\\

\includegraphics[scale=\imscl,width=\imsize]{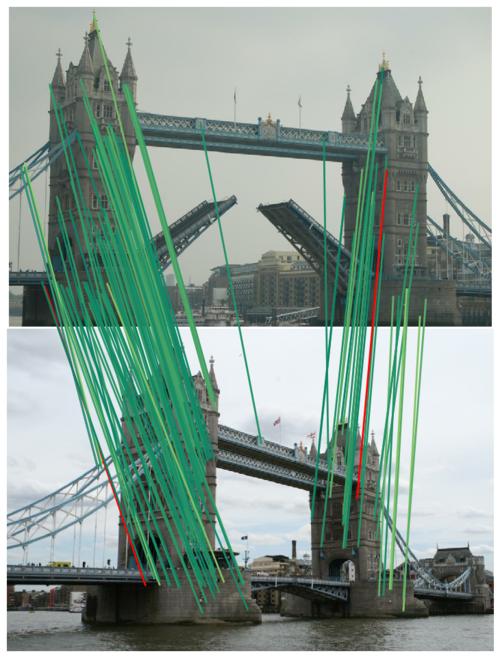} &
\includegraphics[scale=\imscl,width=\imsize]{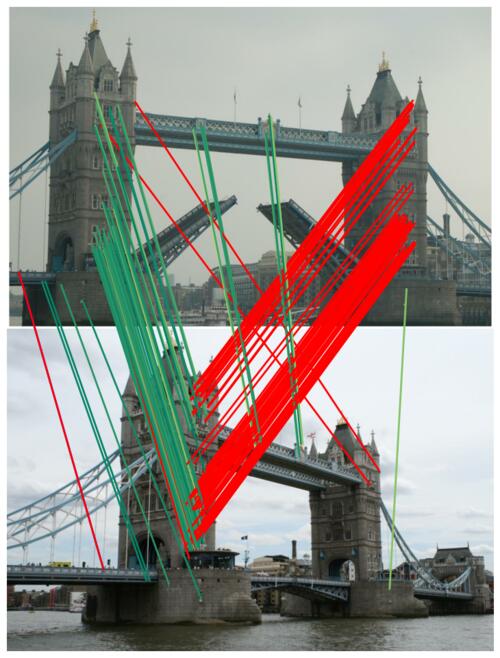} &
\includegraphics[scale=\imscl,width=\imsize]{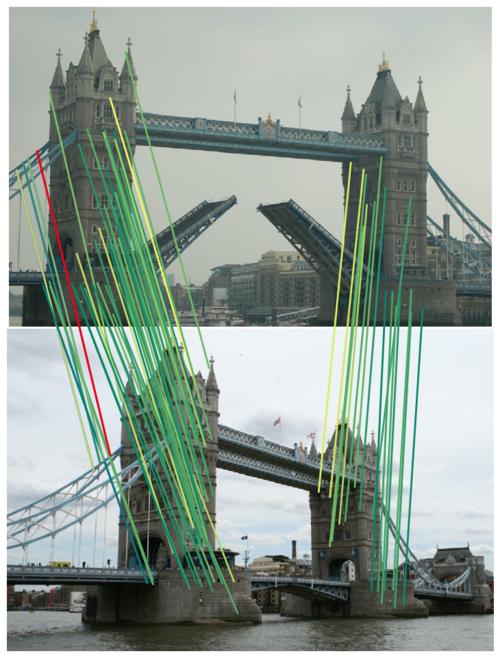} &
\includegraphics[scale=\imscl,width=\imsize]{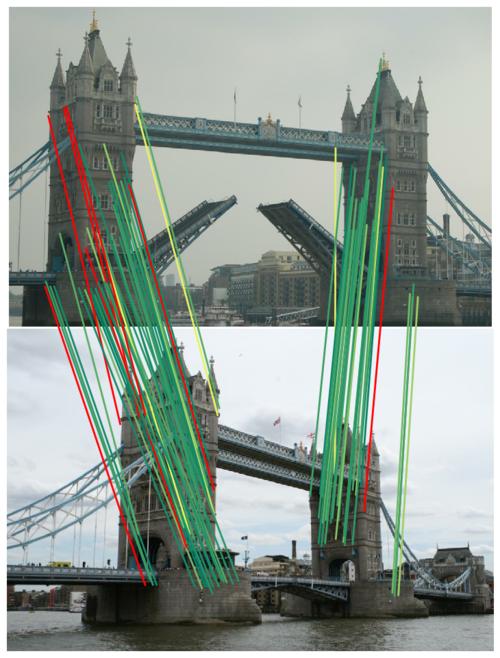} &
\includegraphics[scale=\imscl,width=\imsize]{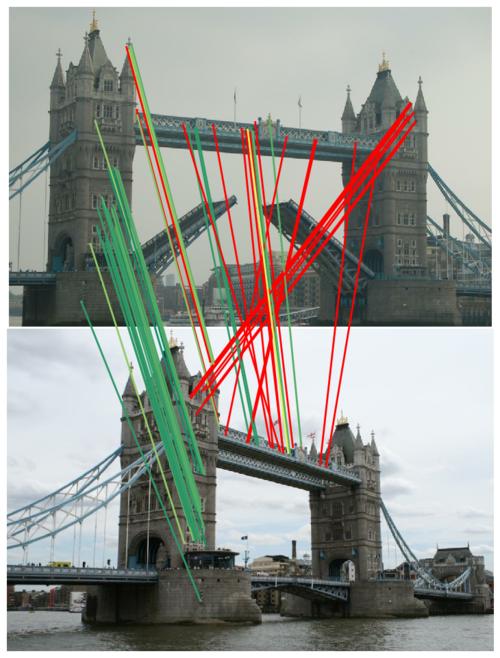}\\

\includegraphics[scale=\imscl,width=\imsize]{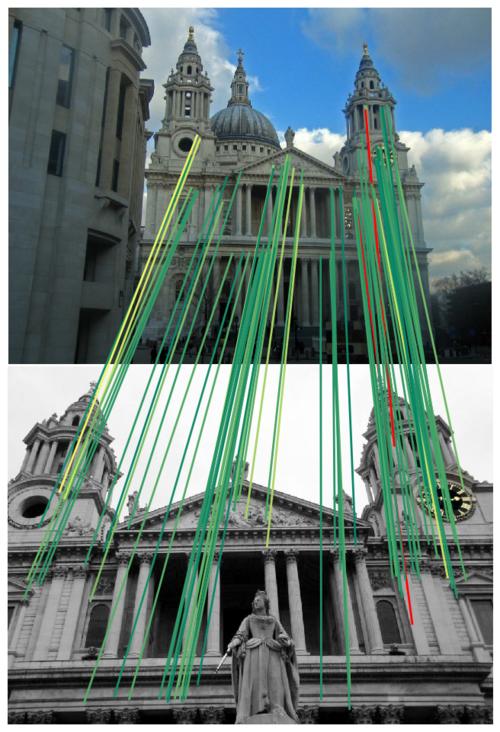} &
\includegraphics[scale=\imscl,width=\imsize]{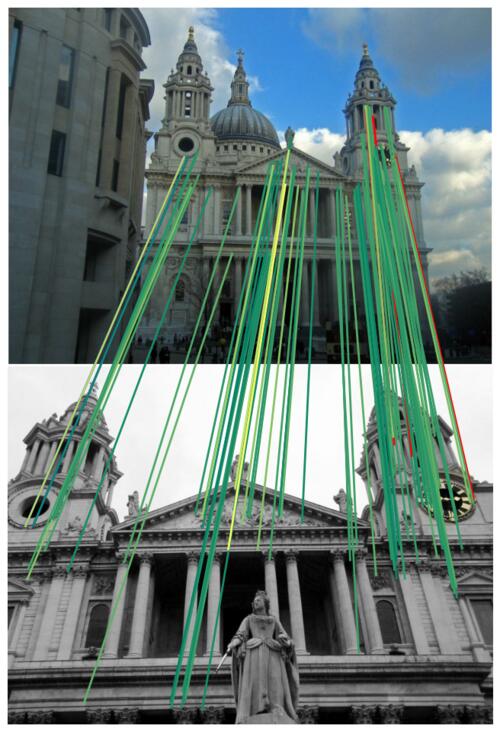} &
\includegraphics[scale=\imscl,width=\imsize]{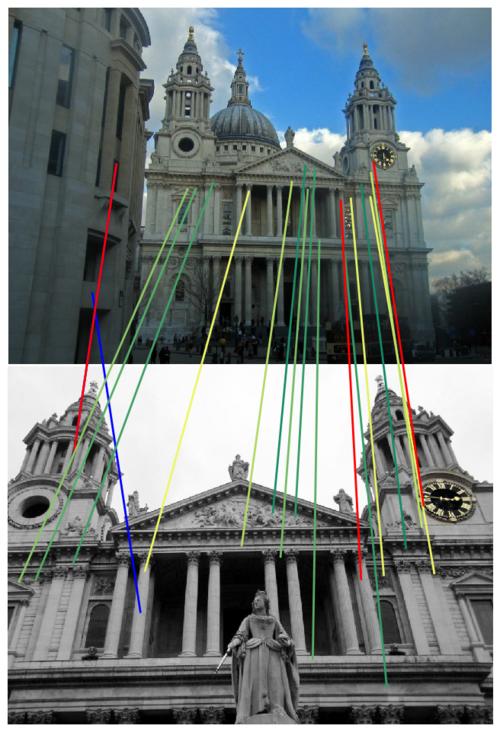} &
\includegraphics[scale=\imscl,width=\imsize]{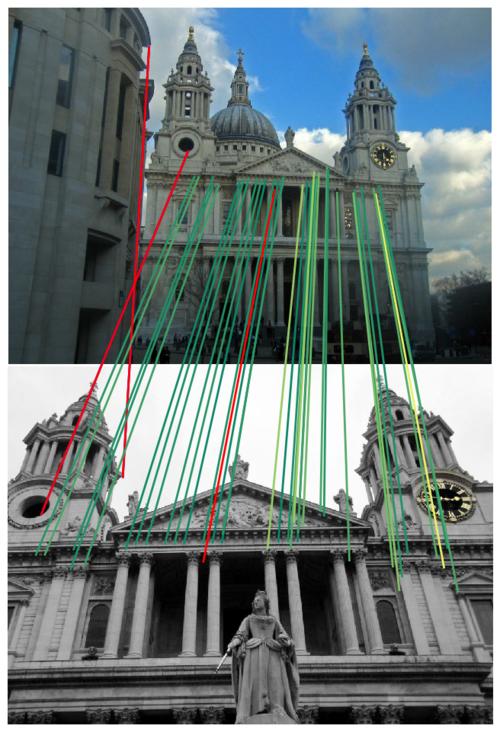} &
\includegraphics[scale=\imscl,width=\imsize]{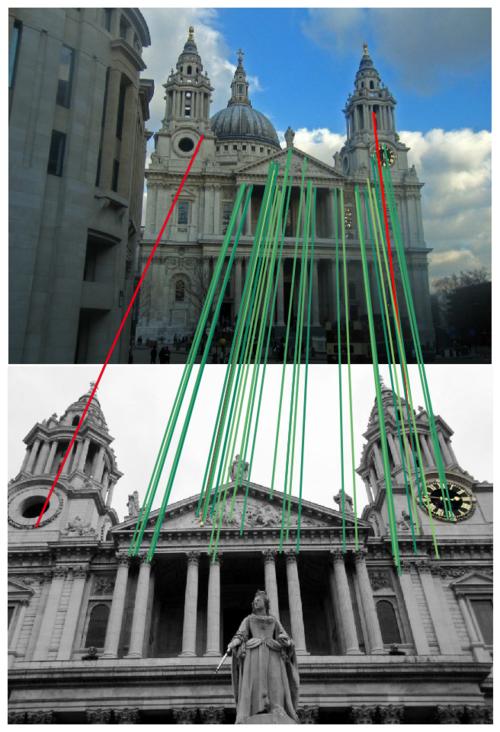}\\

\includegraphics[scale=\imscl,width=\imsize]{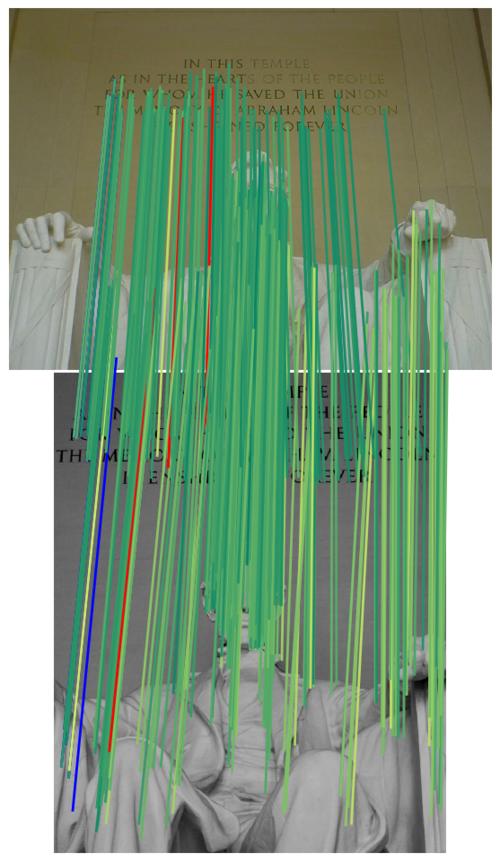} &
\includegraphics[scale=\imscl,width=\imsize]{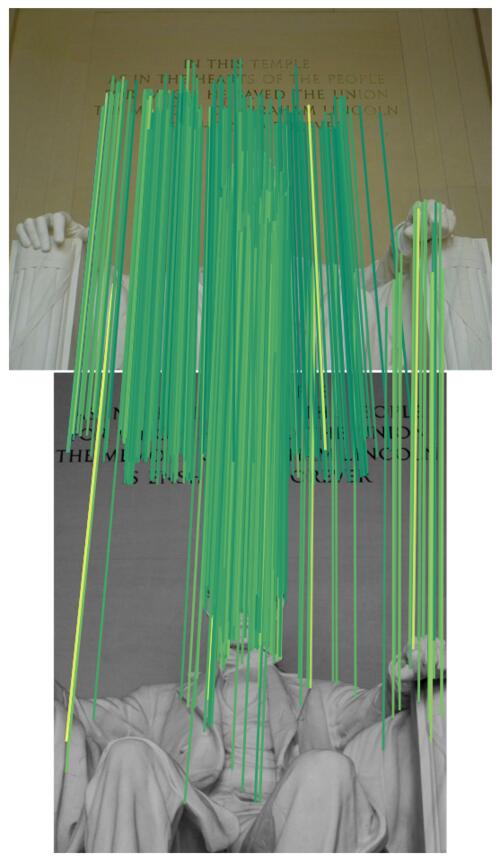} &
\includegraphics[scale=\imscl,width=\imsize]{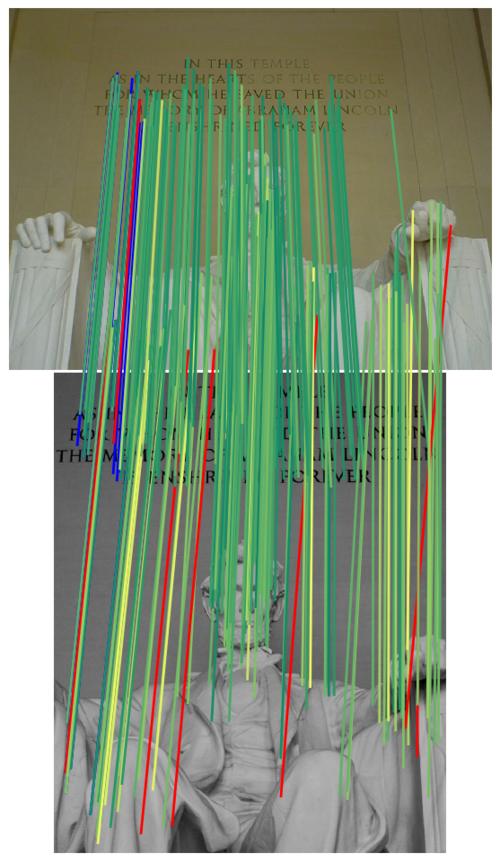} &
\includegraphics[scale=\imscl,width=\imsize]{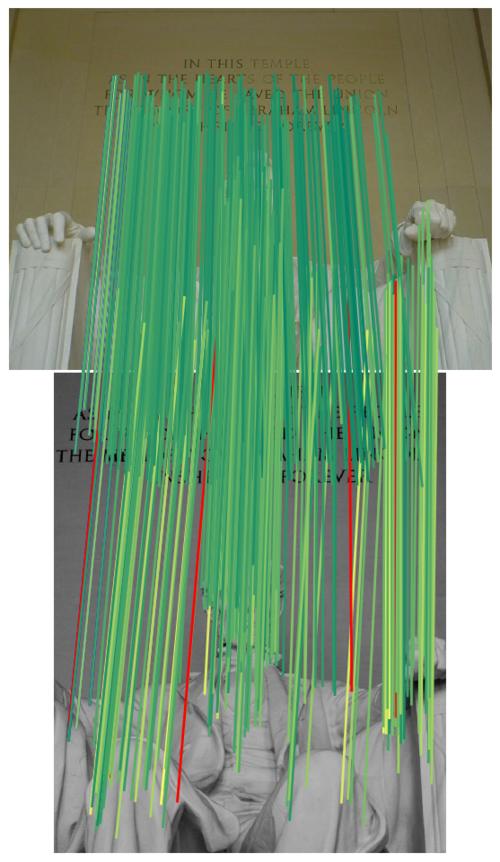} &
\includegraphics[scale=\imscl,width=\imsize]{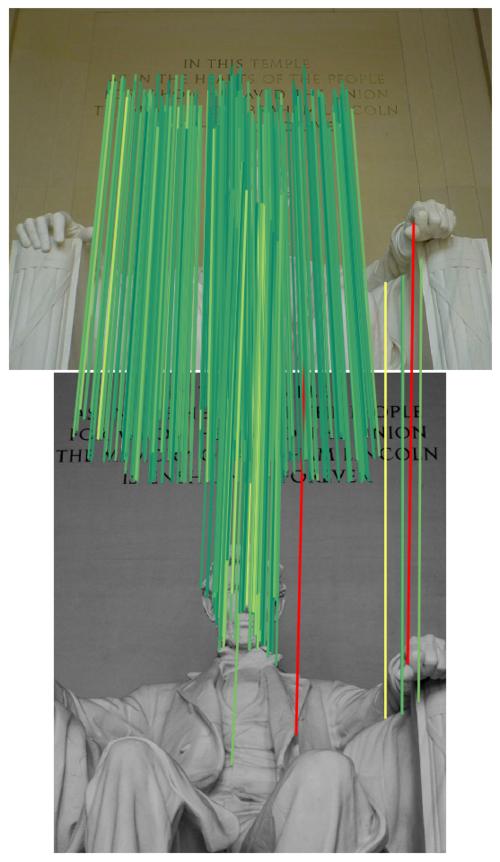}\\

(a) RootSIFT & (b) Hessian-SIFT & (c) SURF & (d) AKAZE & (e) ORB \\

\end{tabular}
\end{center}
\caption{{\bf Qualitative results for the stereo task -- \edit{``Classical''} features.} We plot the matches predicted by each \edit{local feature, with DEGENSAC}. \edit{Matches above a 5-pixel error threshold are drawn in \textcolor{red}{{\bf red}}, and those below are color-coded by their error, from 0 (\textcolor{ForestGreen}{{\bf green}}) to 5 pixels (\textcolor{Goldenrod}{{\bf yellow}}). Matches for which we do not have depth estimates are drawn in \textcolor{blue}{{\bf blue}}.}}
\label{fig:supp_stereo}
\end{figure*}

\begin{figure*}
\centering
\renewcommand{\arraystretch}{0.5}
\renewcommand{\tabcolsep}{0.2mm}
\renewcommand{\imsize}{3.0cm}
\renewcommand{\hspacer}{0.3mm}
\renewcommand{\vspacer}{0.1mm}
\renewcommand{\imscl}{0.01}
\small
\begin{center}
\begin{tabular}{@{}ccccc@{}}
\includegraphics[scale=\imscl,width=\imsize]{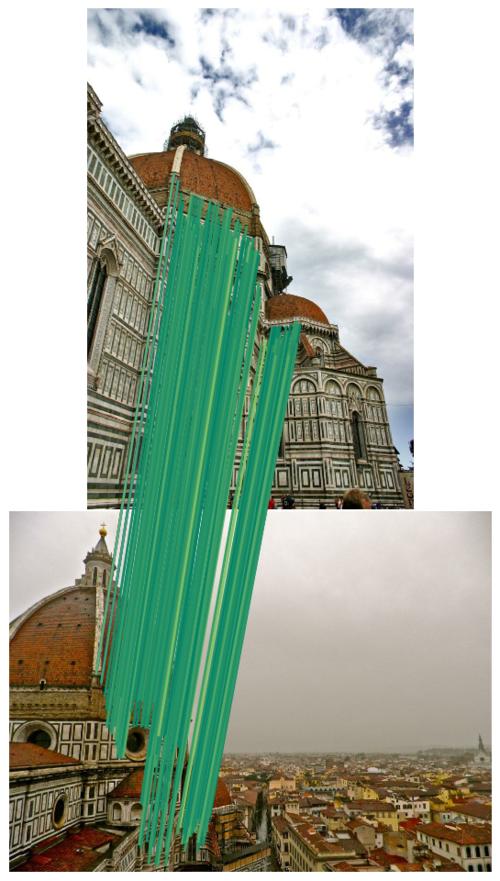} &
\includegraphics[scale=\imscl,width=\imsize]{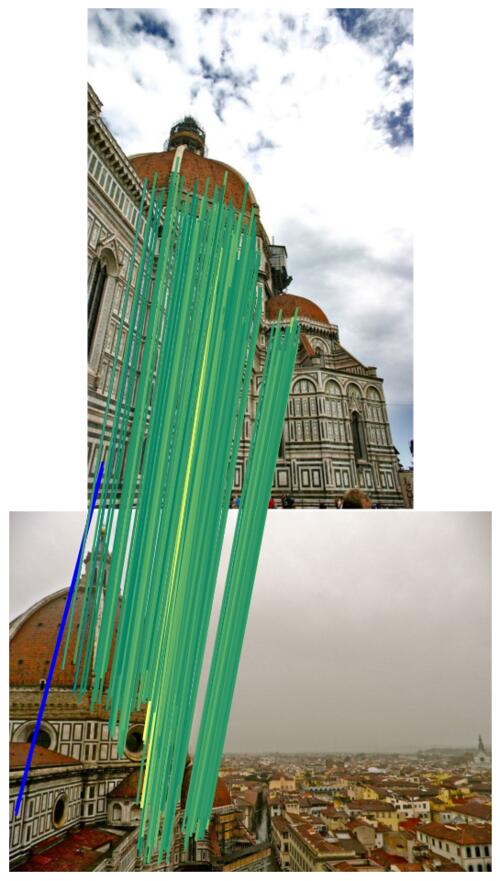} &
\includegraphics[scale=\imscl,width=\imsize]{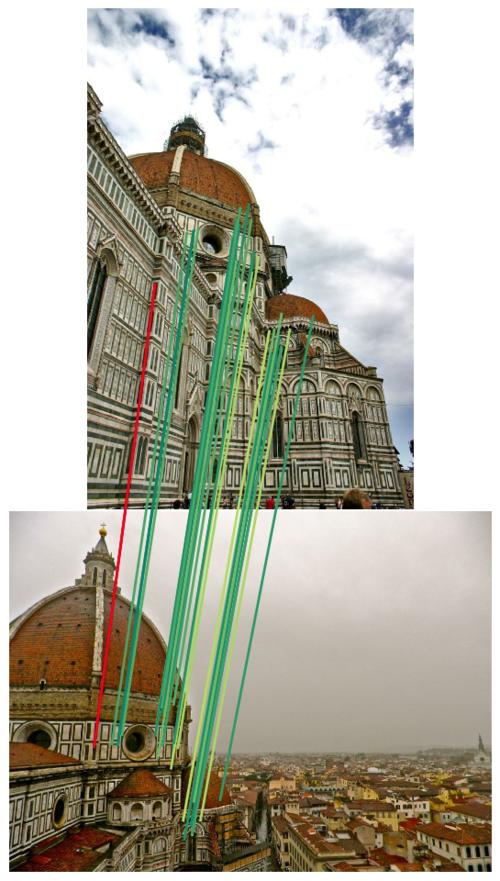} &
\includegraphics[scale=\imscl,width=\imsize]{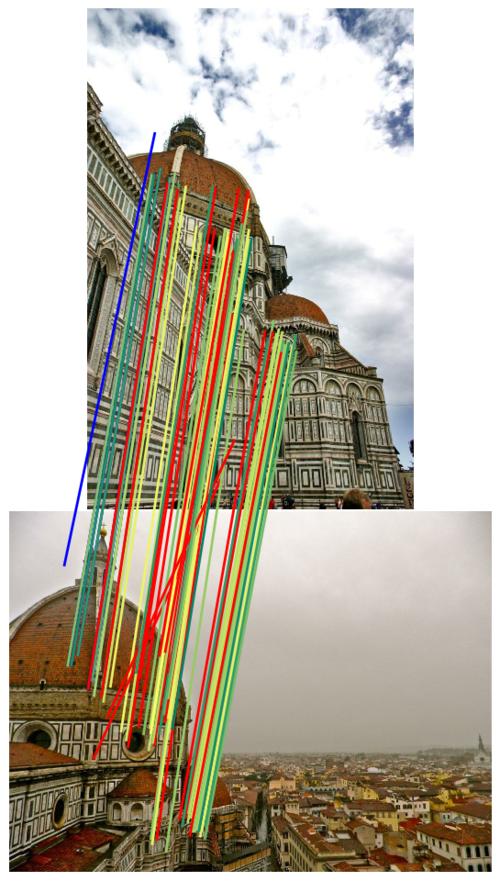} &
\includegraphics[scale=\imscl,width=\imsize]{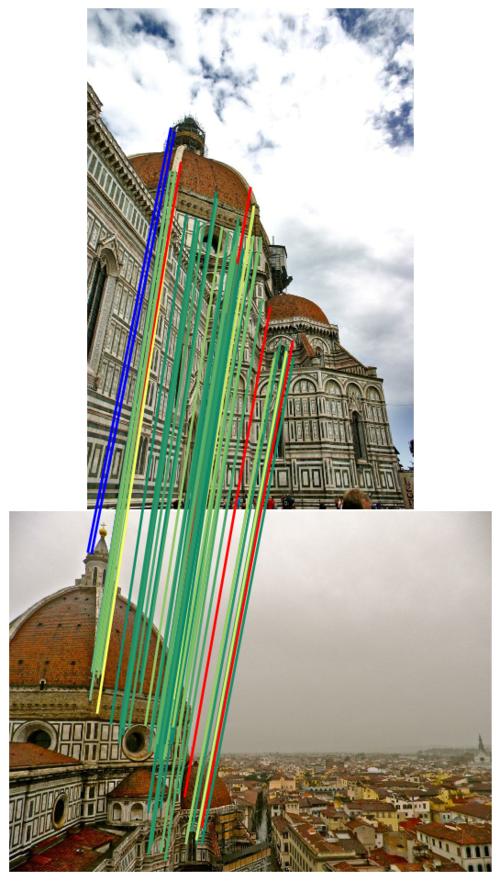}\\

\includegraphics[scale=\imscl,width=\imsize]{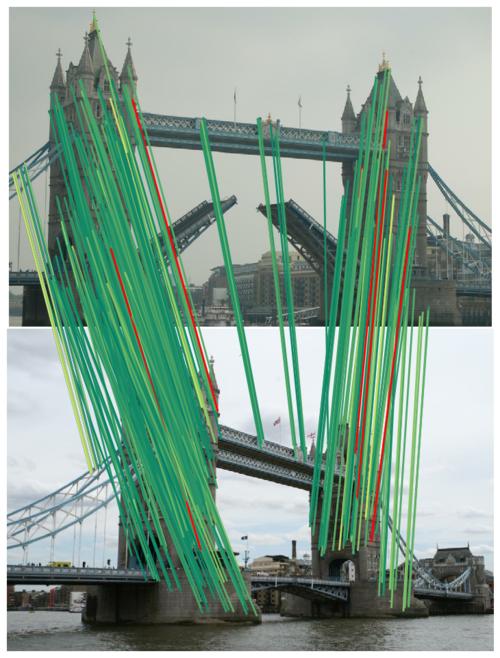} &
\includegraphics[scale=\imscl,width=\imsize]{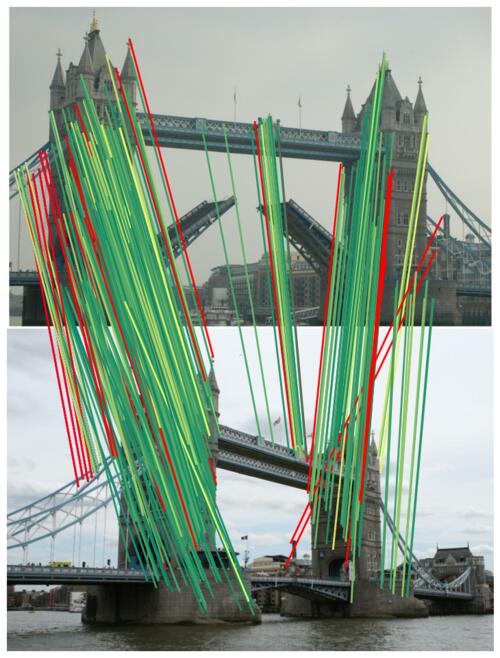} &
\includegraphics[scale=\imscl,width=\imsize]{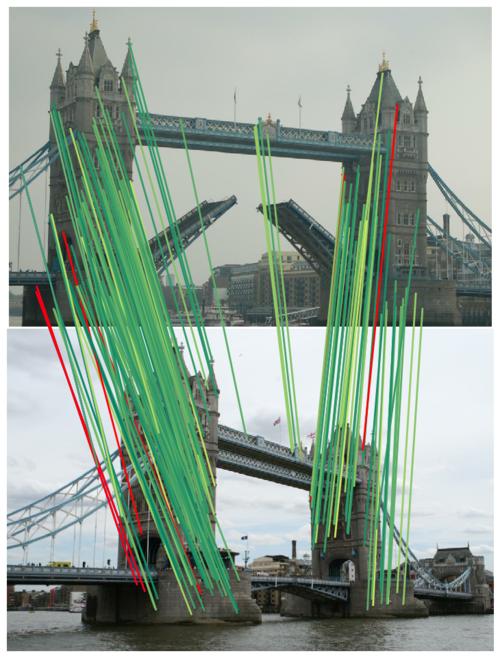} &
\includegraphics[scale=\imscl,width=\imsize]{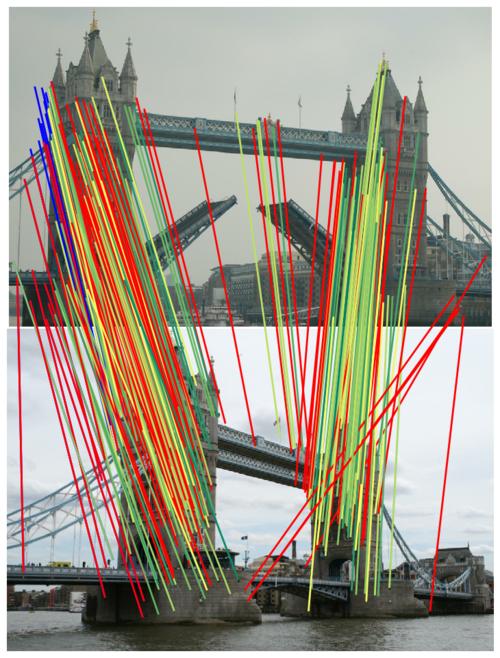} &
\includegraphics[scale=\imscl,width=\imsize]{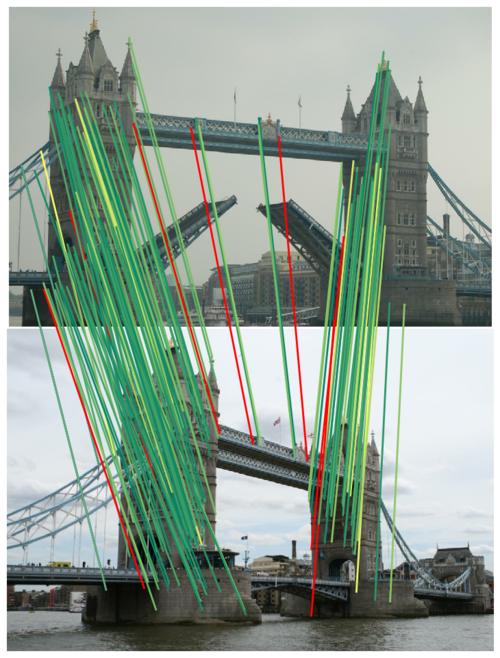}\\

\includegraphics[scale=\imscl,width=\imsize]{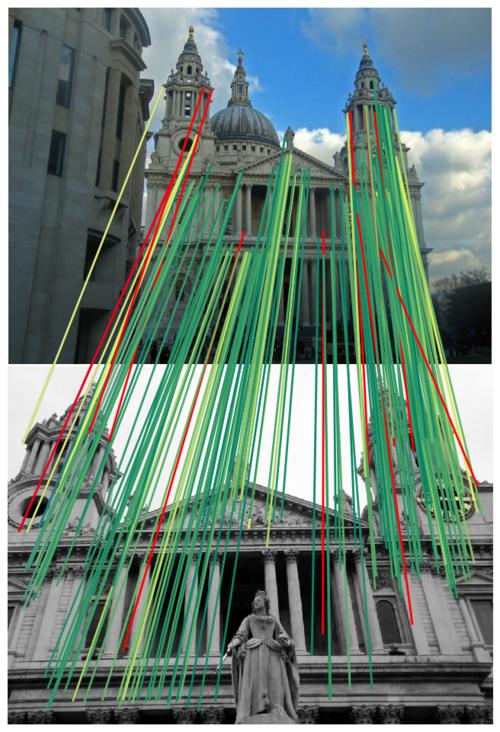} &
\includegraphics[scale=\imscl,width=\imsize]{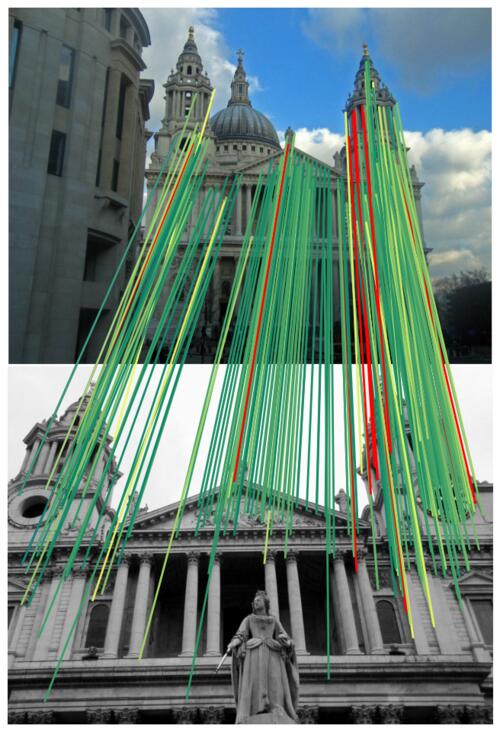} &
\includegraphics[scale=\imscl,width=\imsize]{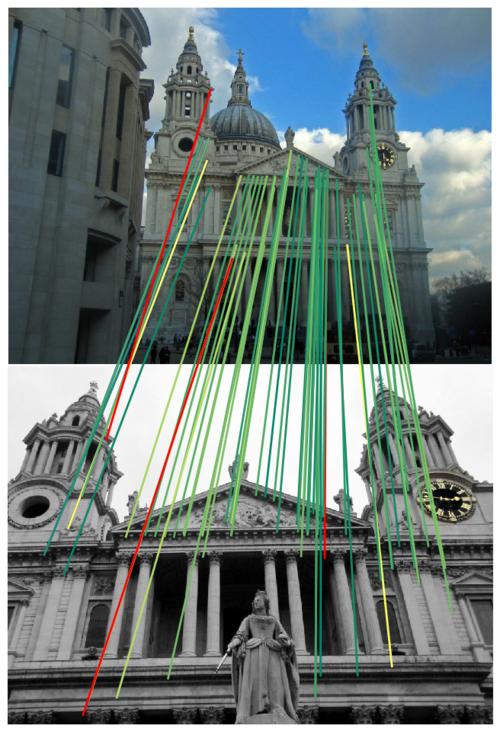} &
\includegraphics[scale=\imscl,width=\imsize]{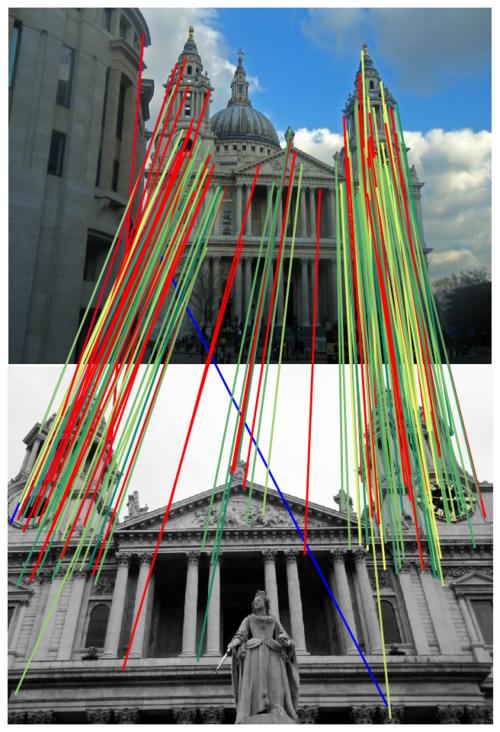} &
\includegraphics[scale=\imscl,width=\imsize]{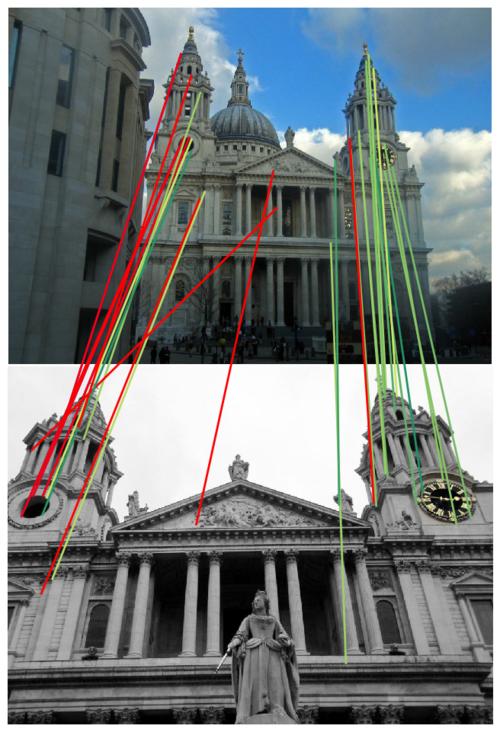}\\

\includegraphics[scale=\imscl,width=\imsize]{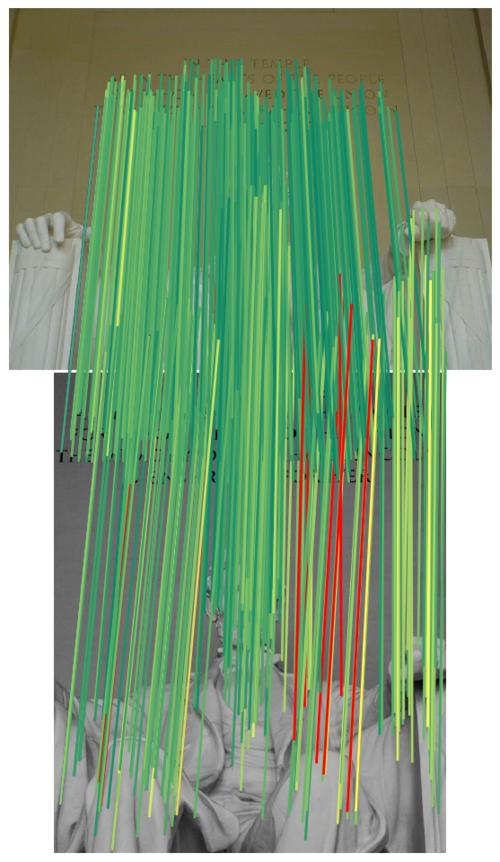} &
\includegraphics[scale=\imscl,width=\imsize]{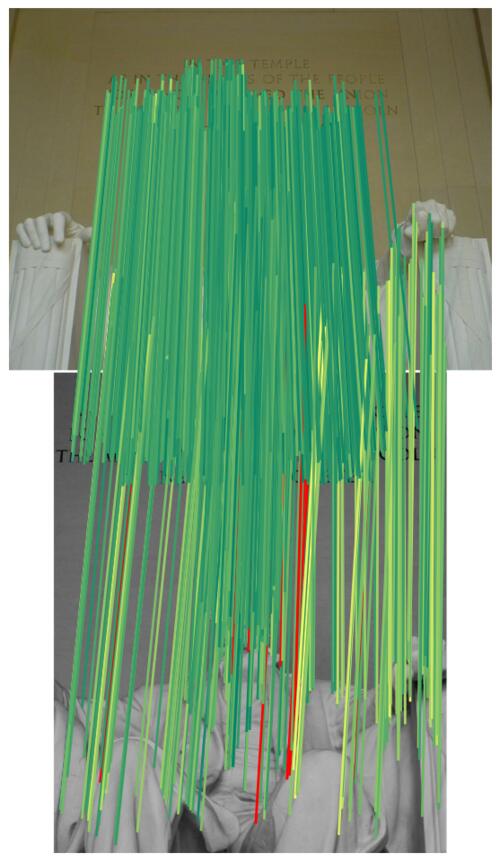} &
\includegraphics[scale=\imscl,width=\imsize]{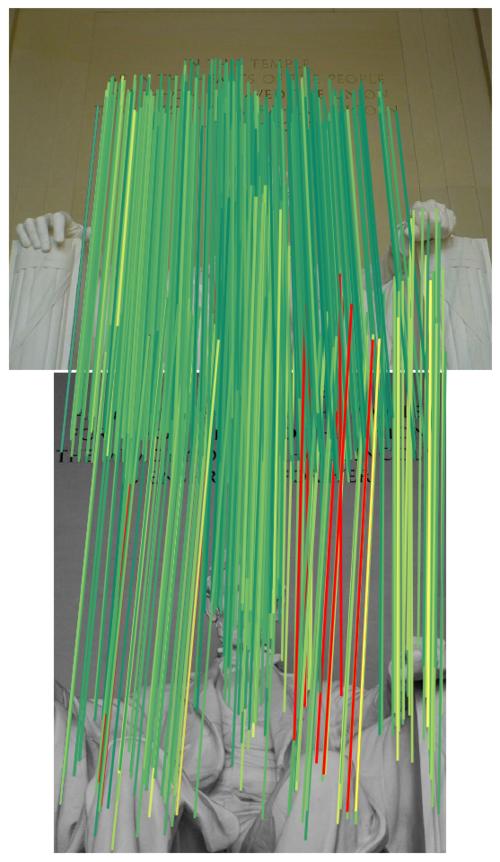} &
\includegraphics[scale=\imscl,width=\imsize]{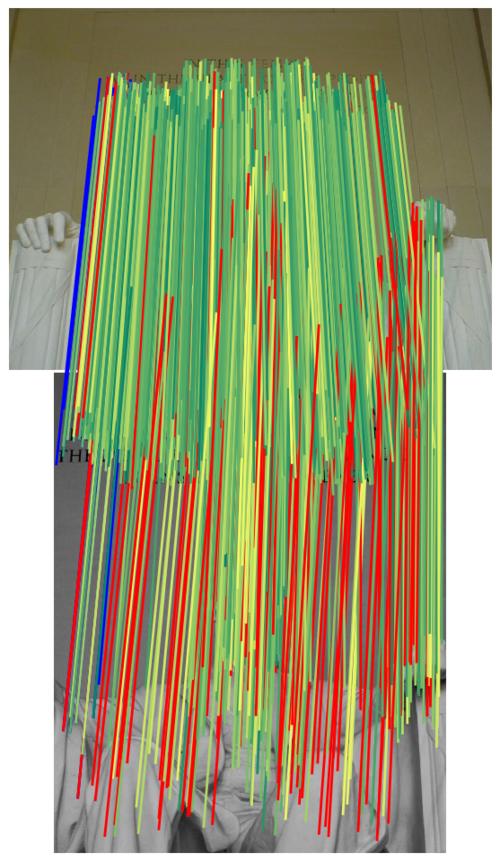} &
\includegraphics[scale=\imscl,width=\imsize]{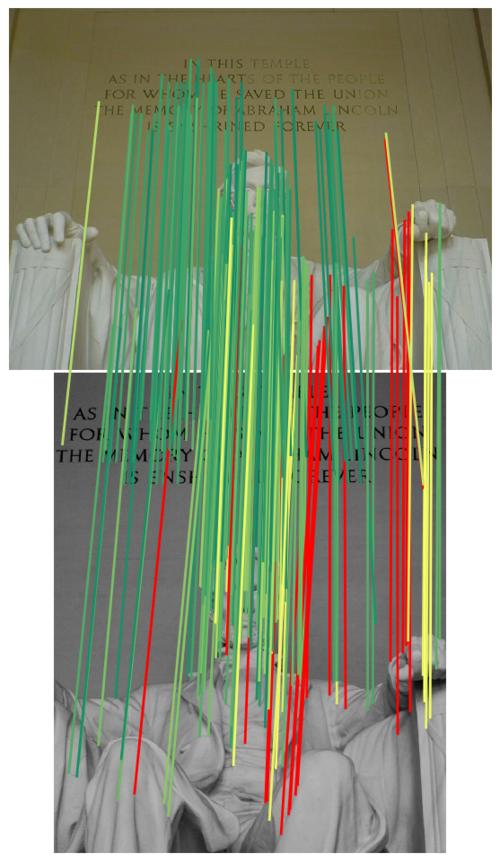}\\
(a) DoG+HardNet & (b) KeyNet+HardNet & (c) R2D2 & (d) D2-Net (MS) & (e) SuperPoint (2k) \\

\end{tabular}
\end{center}
\caption{{\bf Qualitative results for the stereo task -- Learned features.}
Color-coded as in \fig{supp_stereo}.}
\label{fig:supp_stereo2}
\end{figure*}

\begin{figure*}
\centering
\renewcommand{\arraystretch}{0.5}
\renewcommand{\tabcolsep}{0.2mm}
\renewcommand{\imsize}{3.0cm}
\renewcommand{\hspacer}{0.3mm}
\renewcommand{\vspacer}{0.1mm}
\renewcommand{\imscl}{0.01}
\small
\begin{center}
\begin{tabular}{@{}ccccc@{}}

\includegraphics[scale=\imscl,width=\imsize]{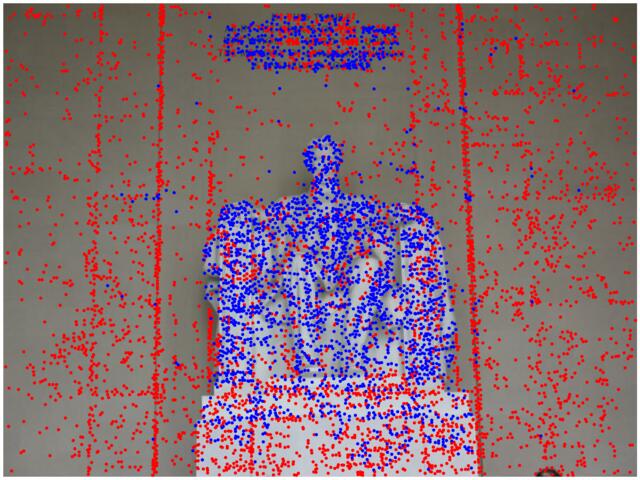} &
\includegraphics[scale=\imscl,width=\imsize]{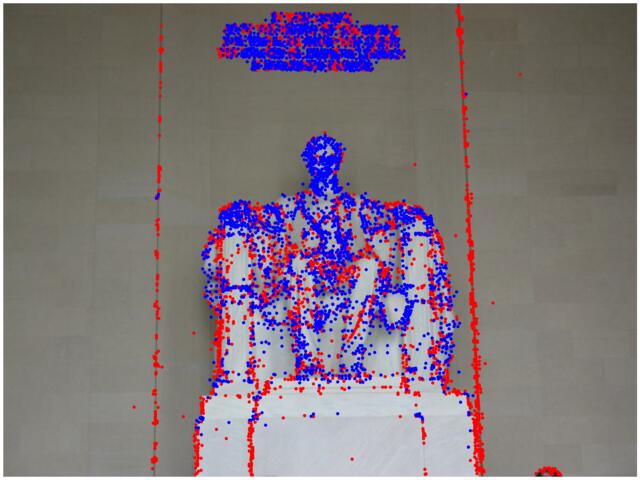} &
\includegraphics[scale=\imscl,width=\imsize]{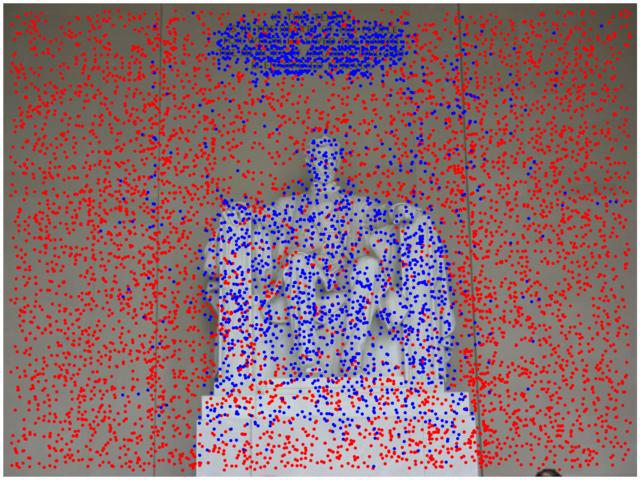} &
\includegraphics[scale=\imscl,width=\imsize]{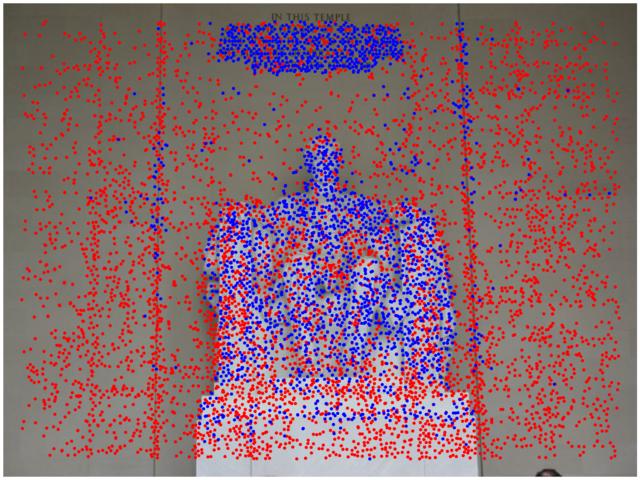} &
\includegraphics[scale=\imscl,width=\imsize]{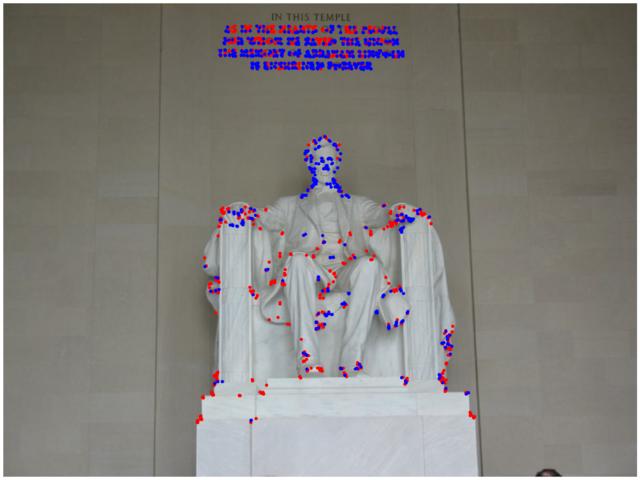} \\

\includegraphics[scale=\imscl,width=\imsize]{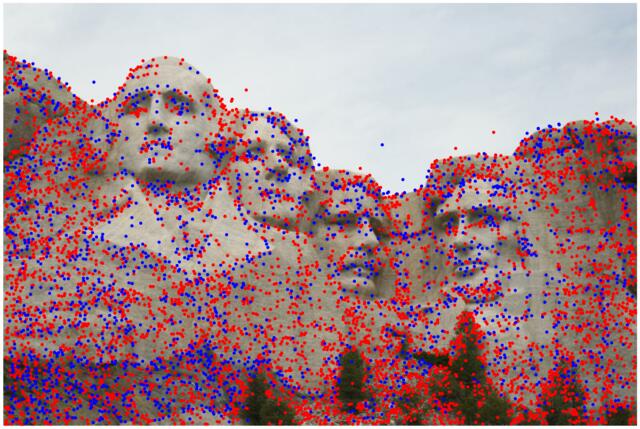} &
\includegraphics[scale=\imscl,width=\imsize]{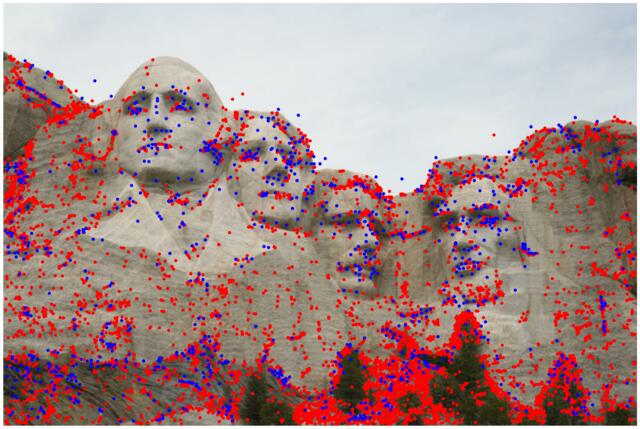} &
\includegraphics[scale=\imscl,width=\imsize]{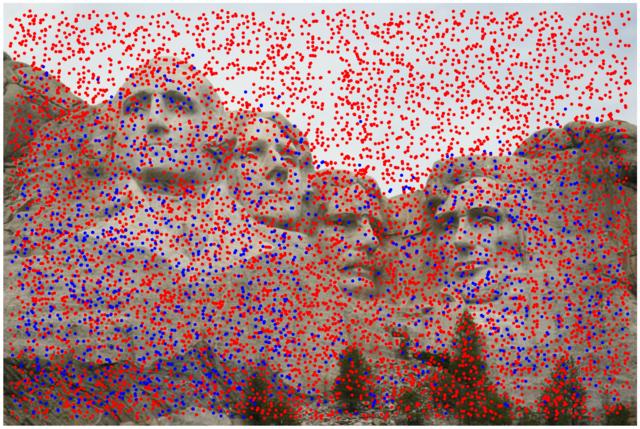} &
\includegraphics[scale=\imscl,width=\imsize]{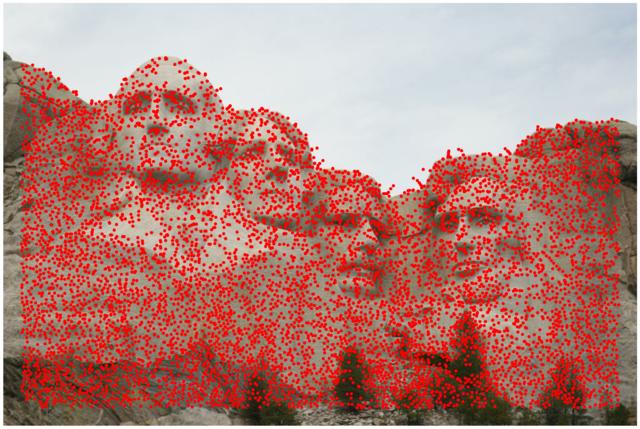} &
\includegraphics[scale=\imscl,width=\imsize]{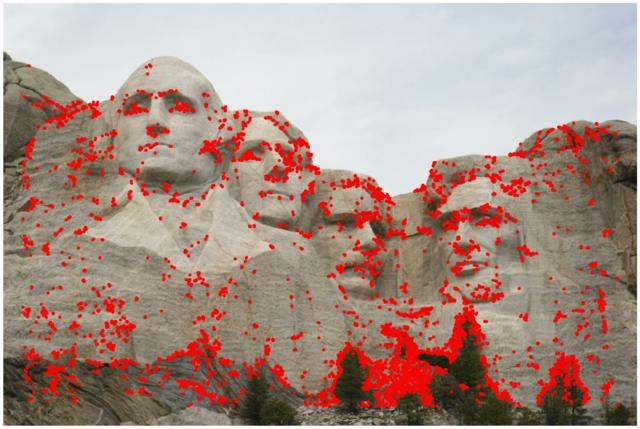} \\

\includegraphics[scale=\imscl,width=\imsize]{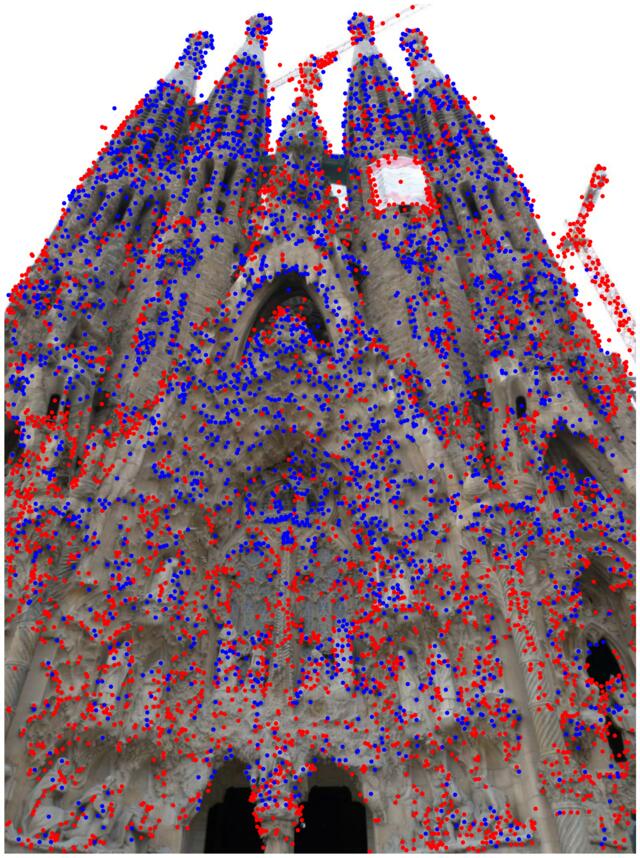} &
\includegraphics[scale=\imscl,width=\imsize]{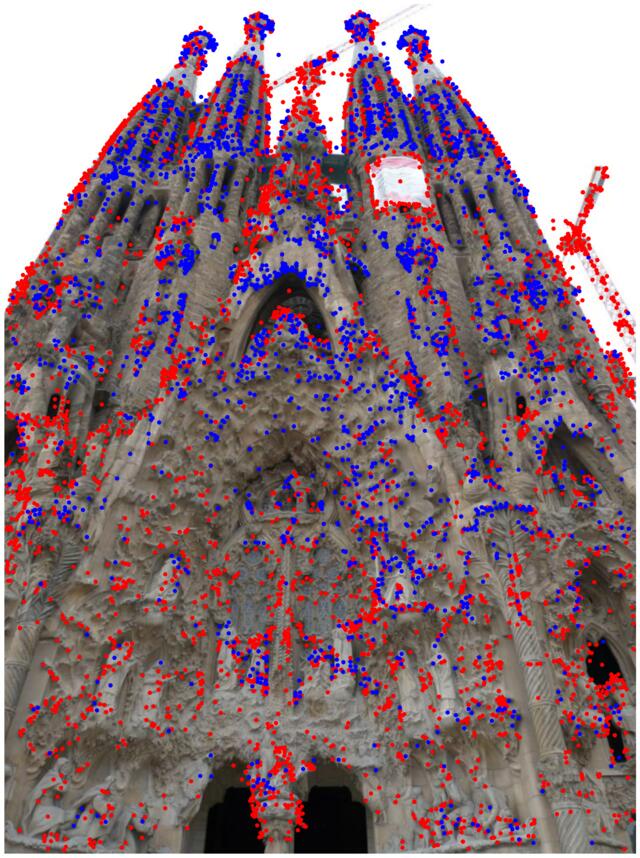} &
\includegraphics[scale=\imscl,width=\imsize]{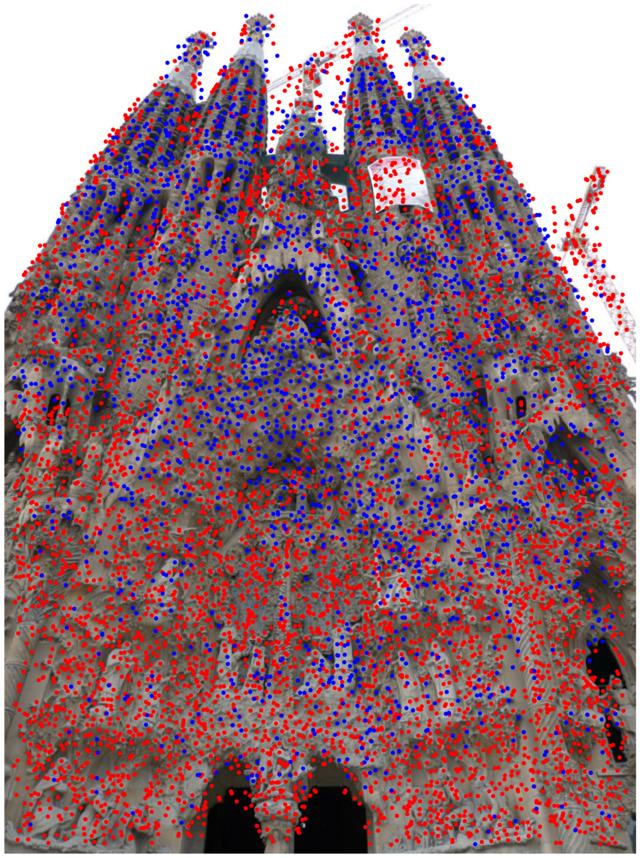} &
\includegraphics[scale=\imscl,width=\imsize]{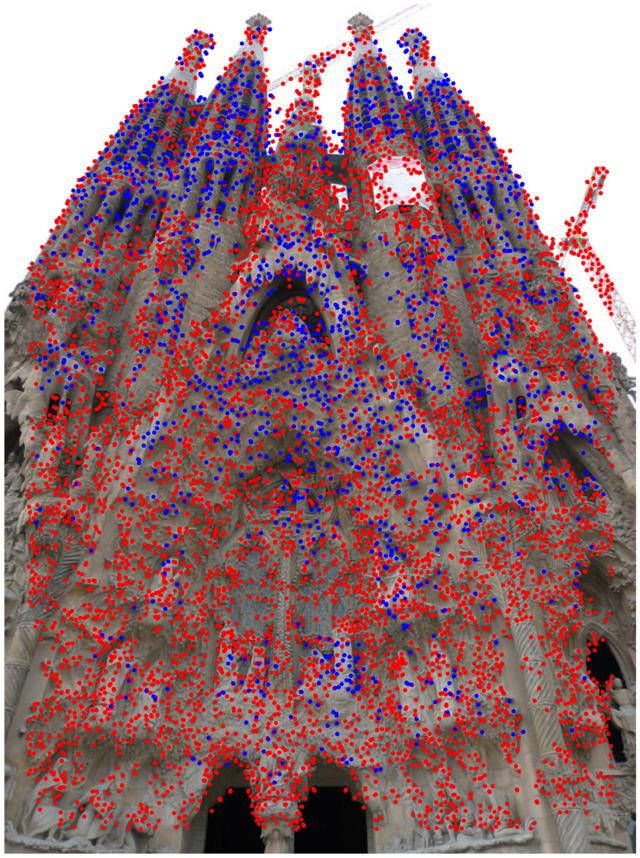} &
\includegraphics[scale=\imscl,width=\imsize]{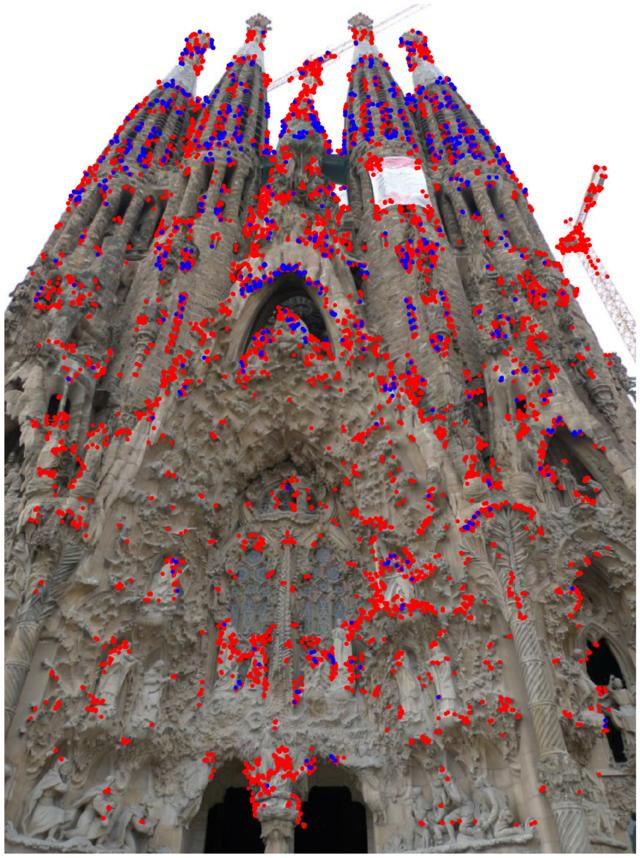} \\

\includegraphics[scale=\imscl,width=\imsize]{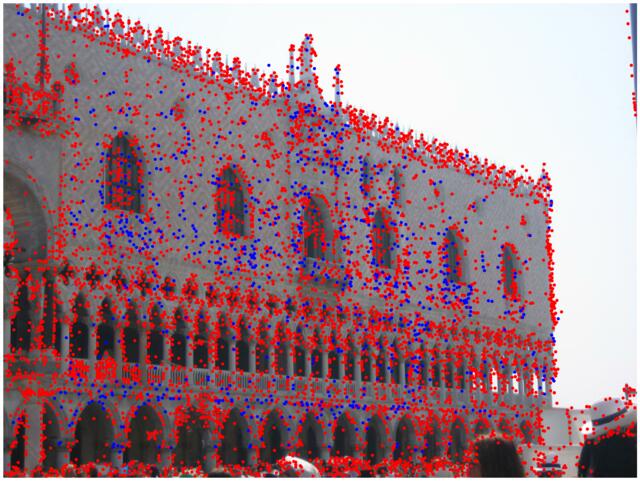} &
\includegraphics[scale=\imscl,width=\imsize]{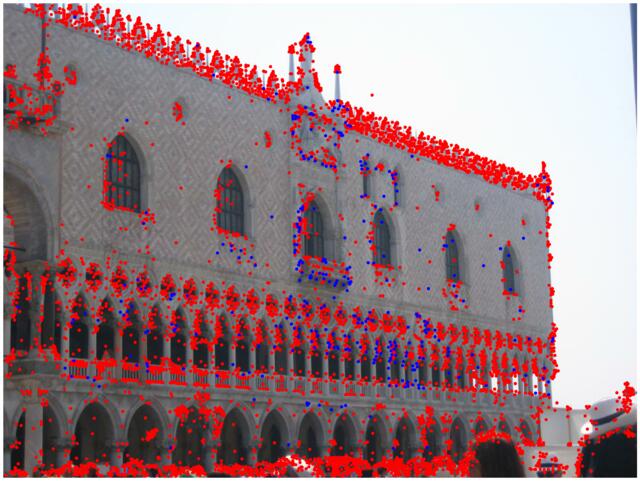} &
\includegraphics[scale=\imscl,width=\imsize]{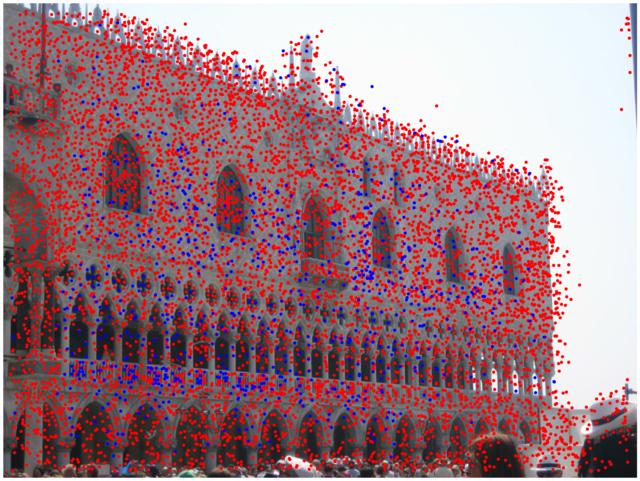} &
\includegraphics[scale=\imscl,width=\imsize]{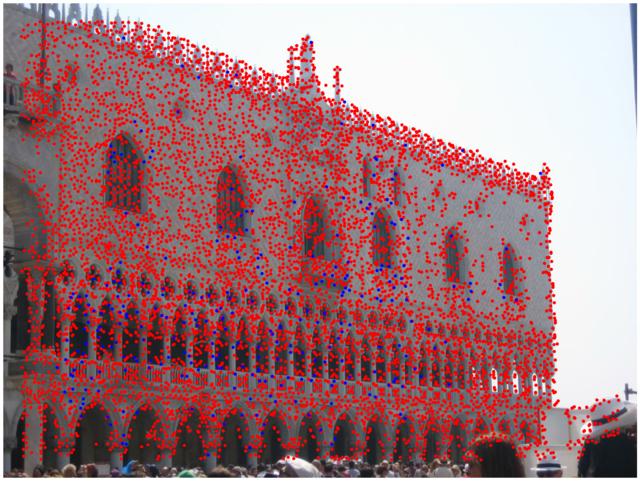} &
\includegraphics[scale=\imscl,width=\imsize]{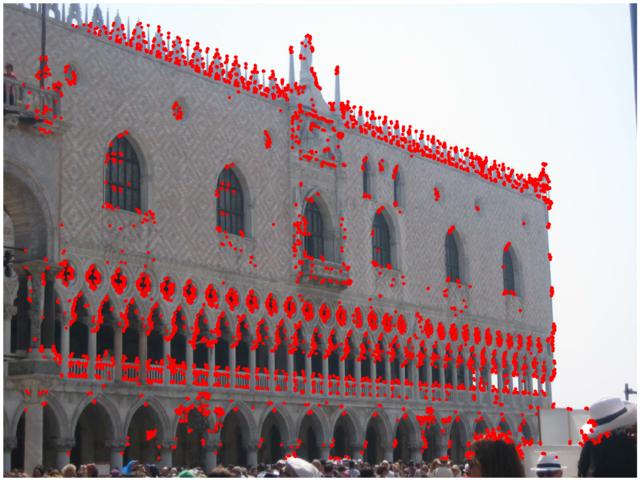} \\

\includegraphics[scale=\imscl,width=\imsize]{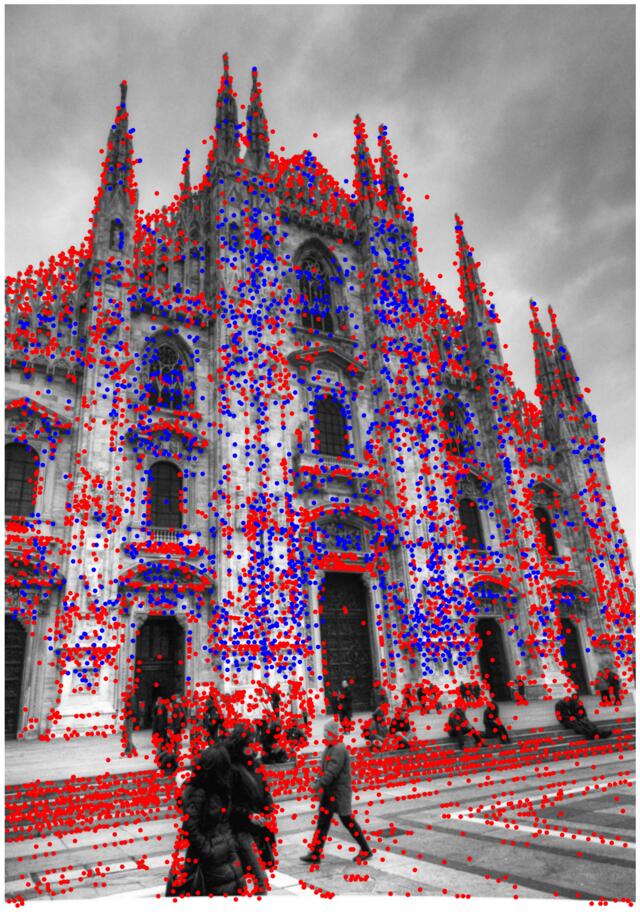} &
\includegraphics[scale=\imscl,width=\imsize]{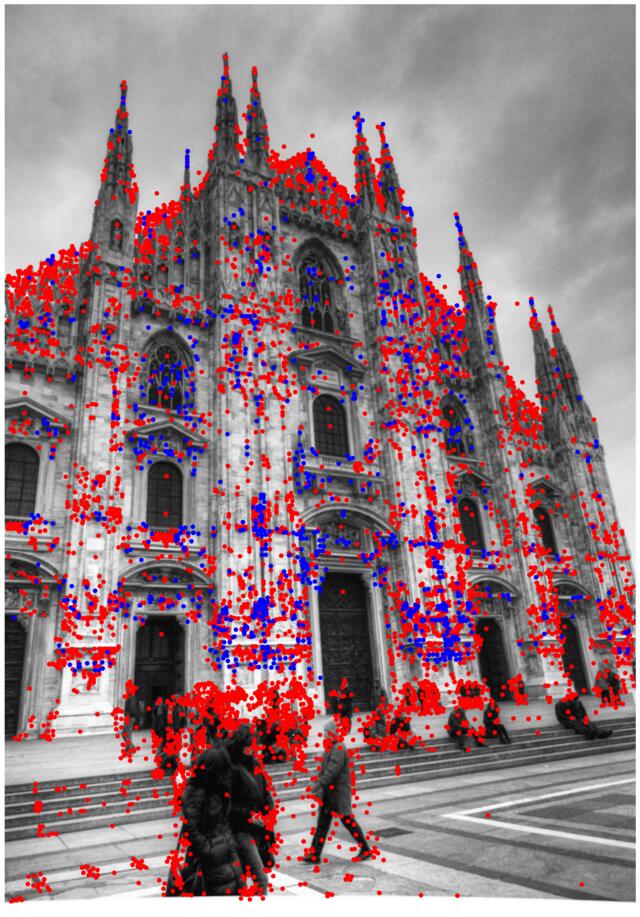} &
\includegraphics[scale=\imscl,width=\imsize]{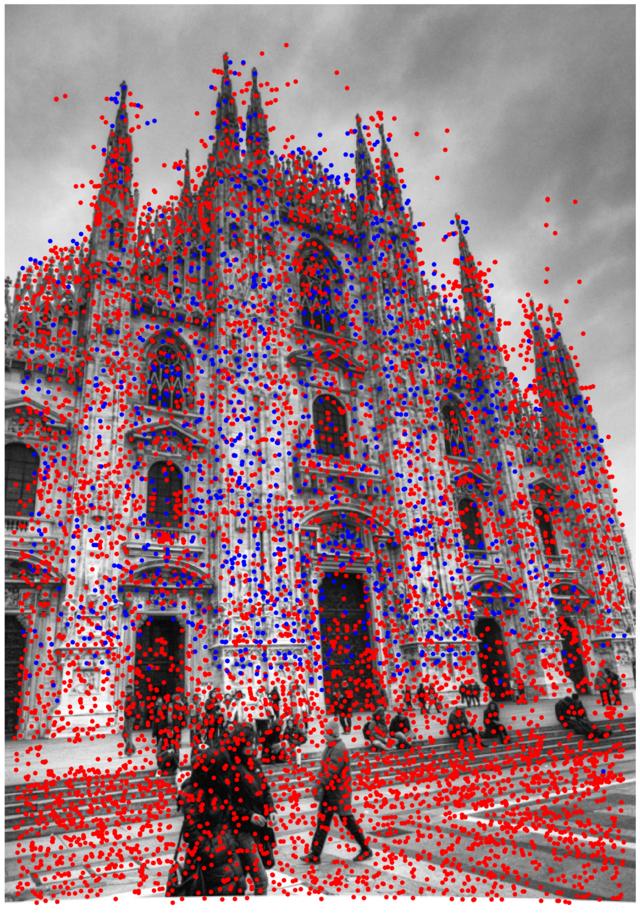} &
\includegraphics[scale=\imscl,width=\imsize]{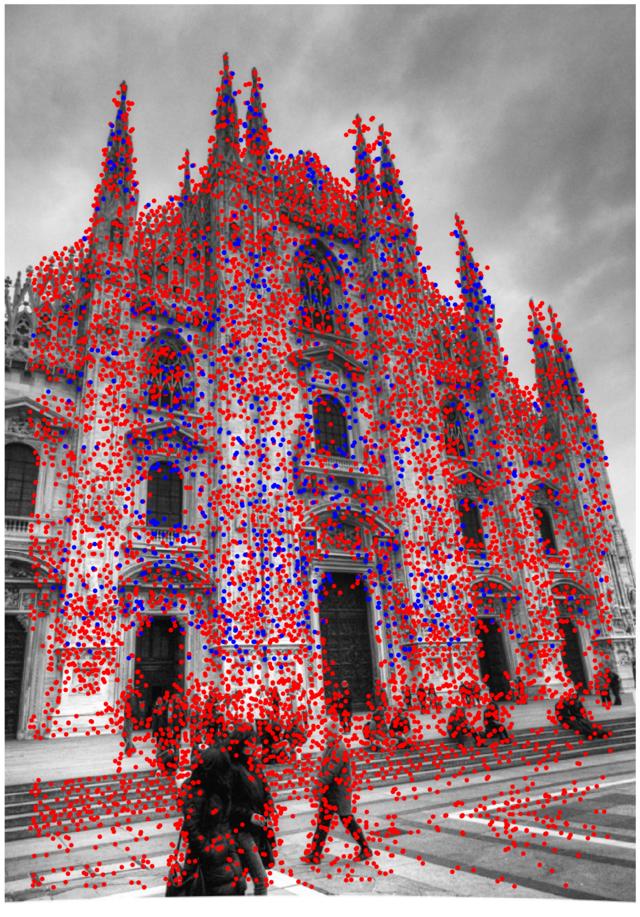} &
\includegraphics[scale=\imscl,width=\imsize]{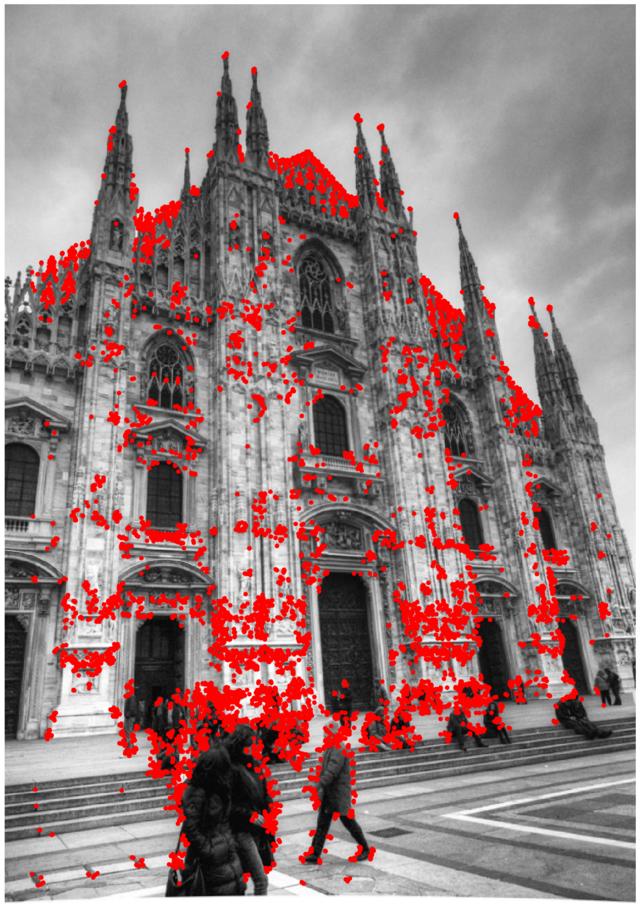} \\
(a) RootSIFT & (b) Hessian-SIFT & (c) SURF & (d) AKAZE & (e) ORB \\
\end{tabular}
\end{center}
\caption{{\bf Qualitative results for the multi-view task -- \edit{``Classical''} features.}
We show images and the keypoints detected on them for different methods, with the points reconstructed by COLMAP in \textcolor{blue}{{\bf blue}}, and the ones that are not used in the 3D model in \textcolor{red}{{\bf red}} -- more blue points indicate denser 3D models. These results correspond the multiview-task for a 25-image subset.
}
\label{fig:supp_multiview}
\end{figure*}

\begin{figure*}
\centering
\renewcommand{\arraystretch}{0.5}
\renewcommand{\tabcolsep}{0.2mm}
\renewcommand{\imsize}{3.0cm}
\renewcommand{\hspacer}{0.3mm}
\renewcommand{\vspacer}{0.1mm}
\renewcommand{\imscl}{0.01}
\small
\begin{center}
\begin{tabular}{@{}ccccc@{}}

\includegraphics[scale=\imscl,width=\imsize]{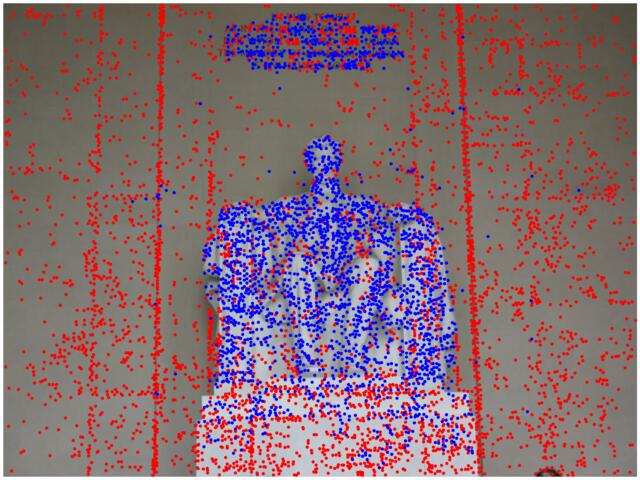} &
\includegraphics[scale=\imscl,width=\imsize]{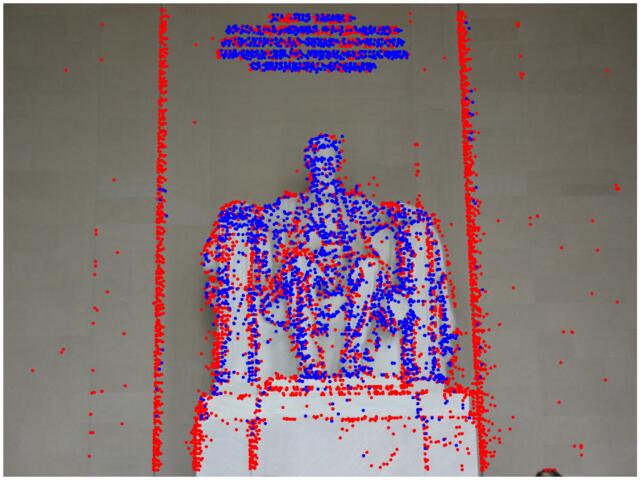} &
\includegraphics[scale=\imscl,width=\imsize]{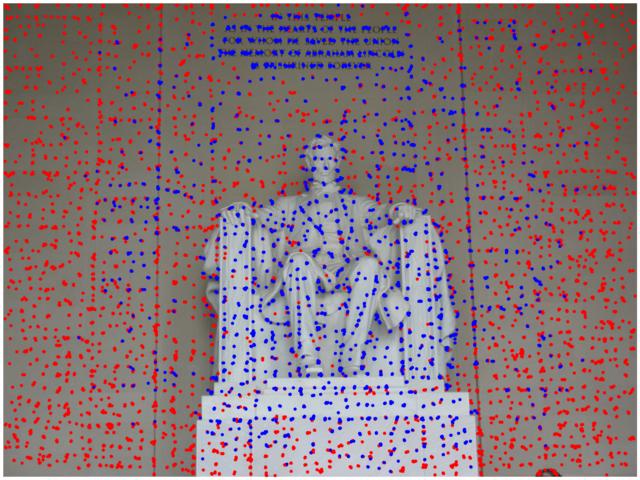} &
\includegraphics[scale=\imscl,width=\imsize]{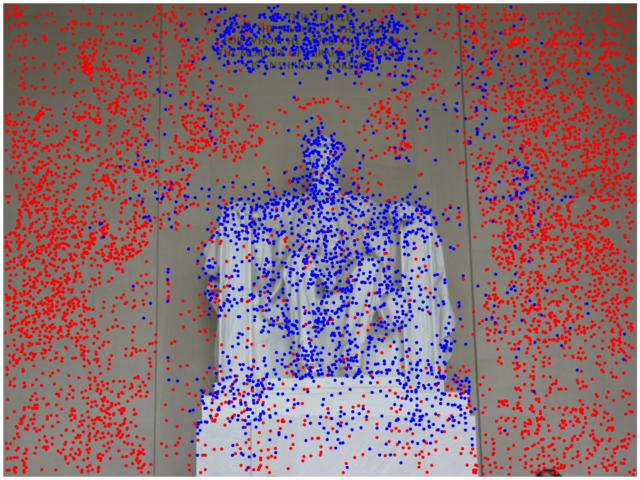} &
\includegraphics[scale=\imscl,width=\imsize]{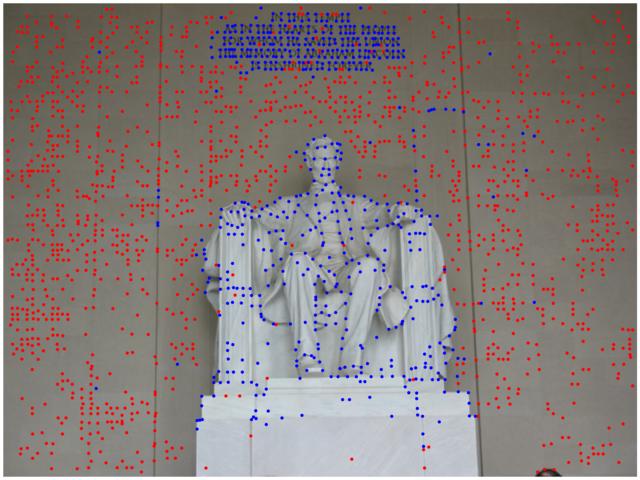} \\

\includegraphics[scale=\imscl,width=\imsize]{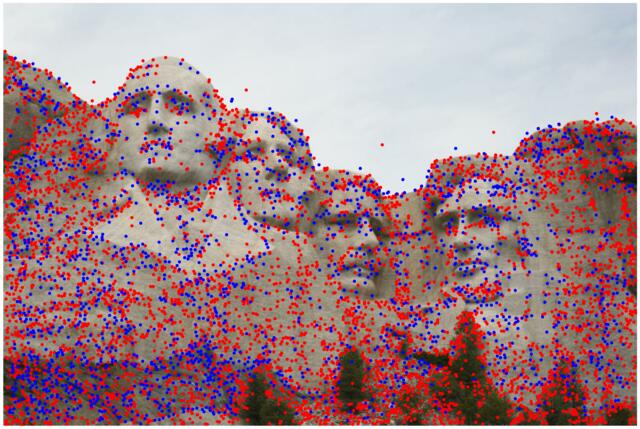} &
\includegraphics[scale=\imscl,width=\imsize]{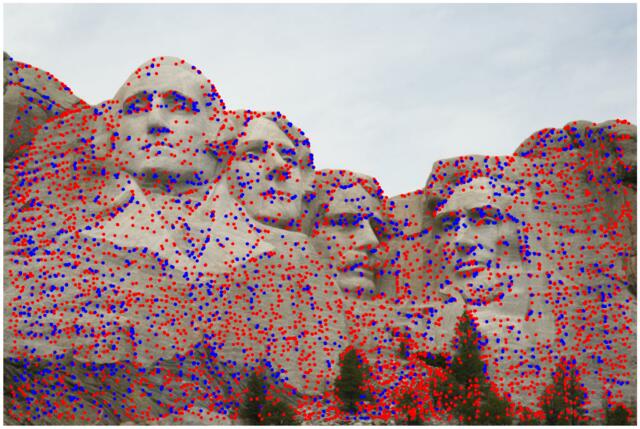} &
\includegraphics[scale=\imscl,width=\imsize]{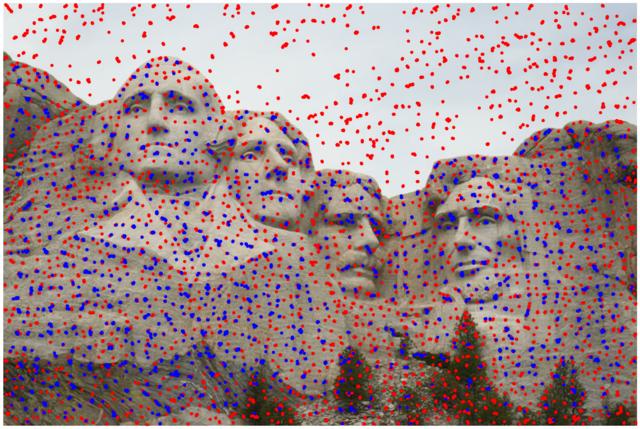} &
\includegraphics[scale=\imscl,width=\imsize]{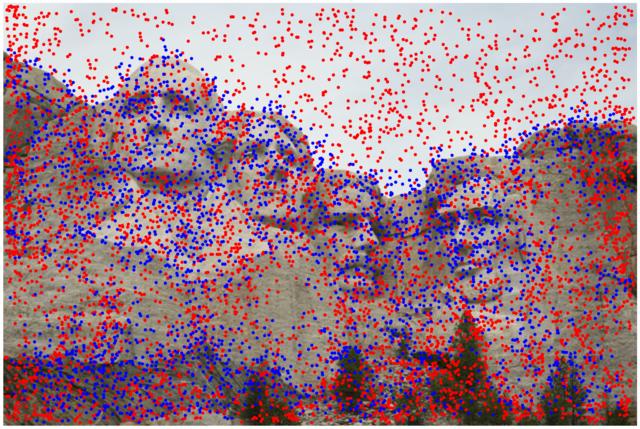} &
\includegraphics[scale=\imscl,width=\imsize]{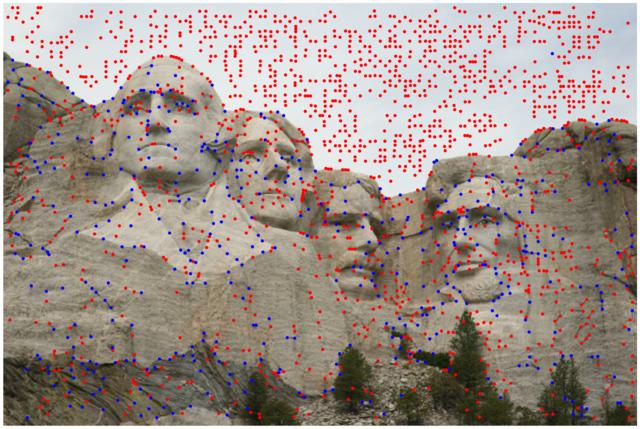} \\

\includegraphics[scale=\imscl,width=\imsize]{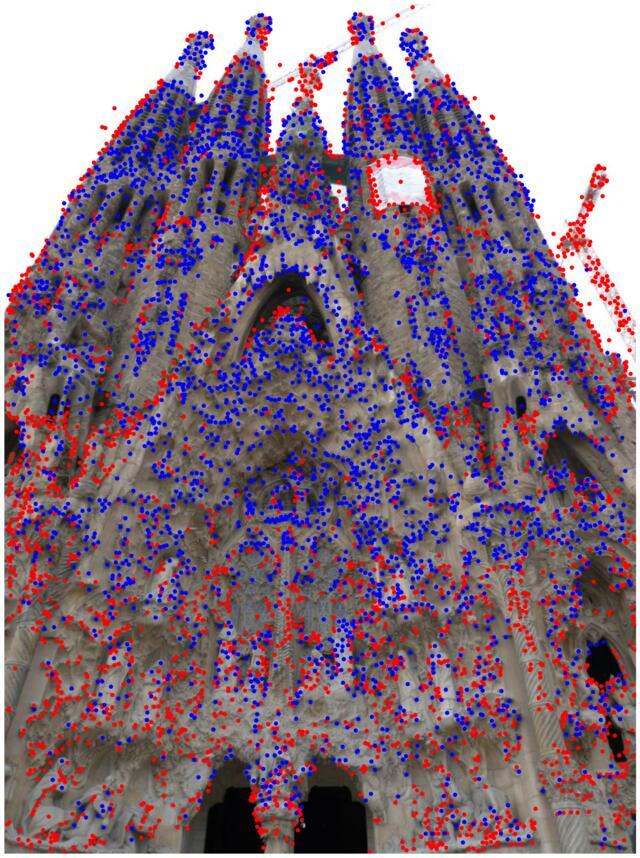} &
\includegraphics[scale=\imscl,width=\imsize]{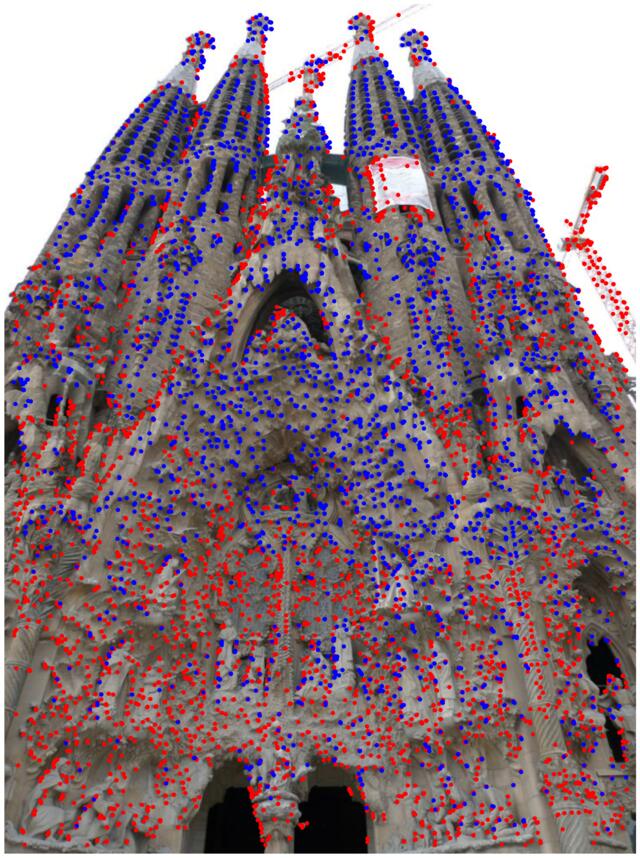} &
\includegraphics[scale=\imscl,width=\imsize]{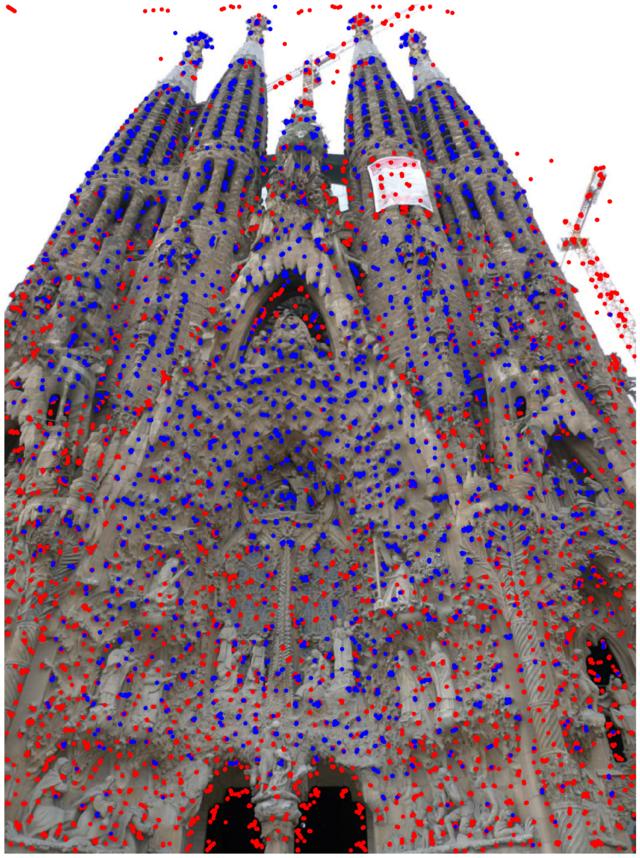} &
\includegraphics[scale=\imscl,width=\imsize]{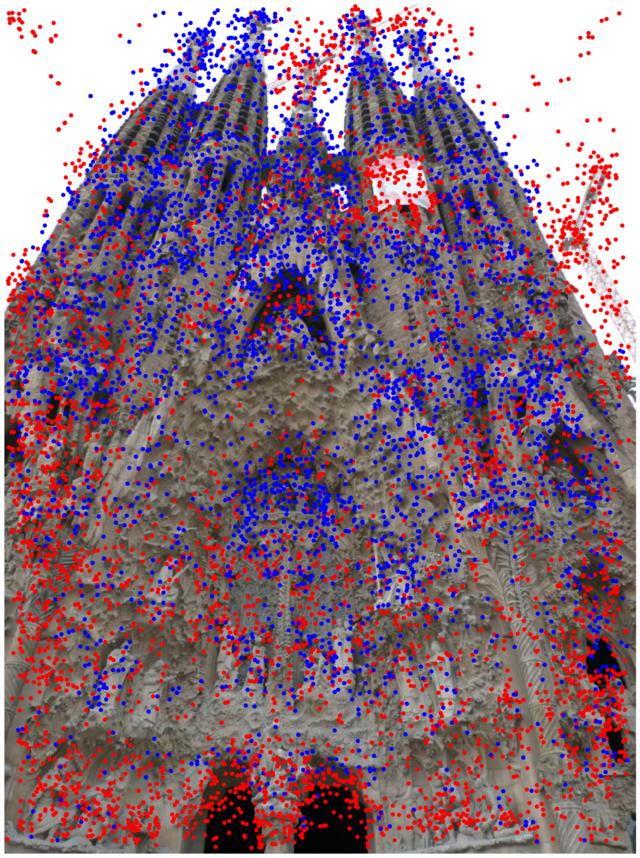} &
\includegraphics[scale=\imscl,width=\imsize]{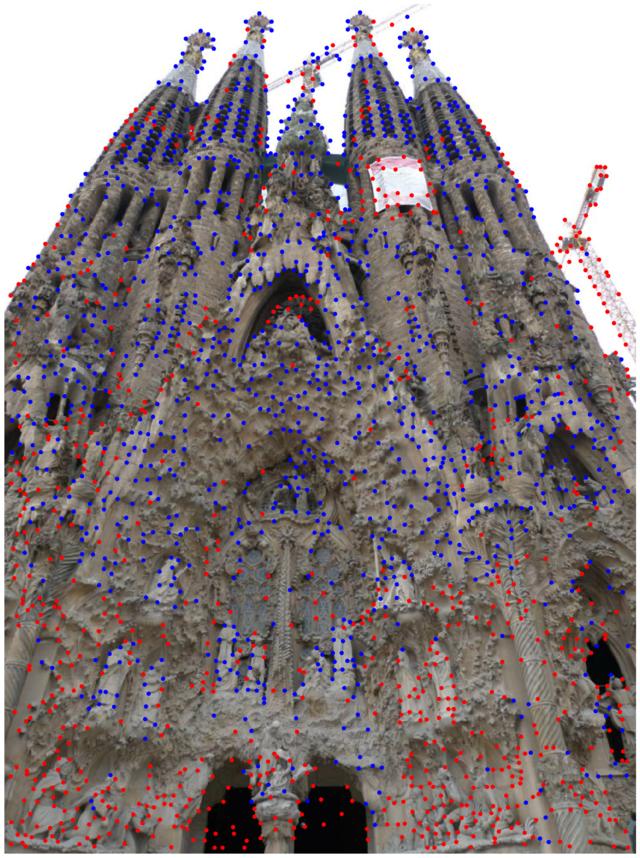} \\

\includegraphics[scale=\imscl,width=\imsize]{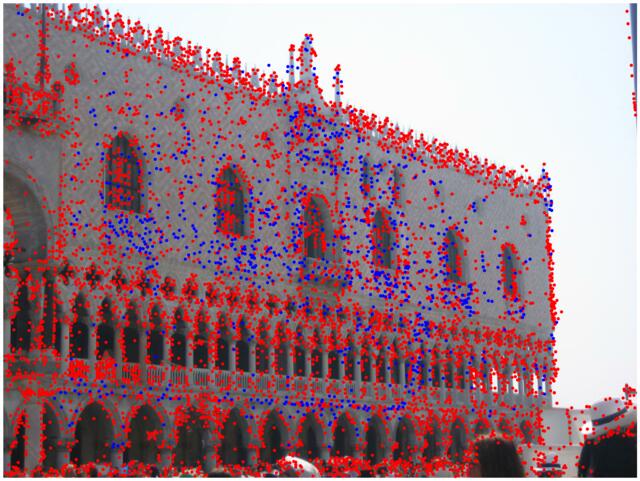} &
\includegraphics[scale=\imscl,width=\imsize]{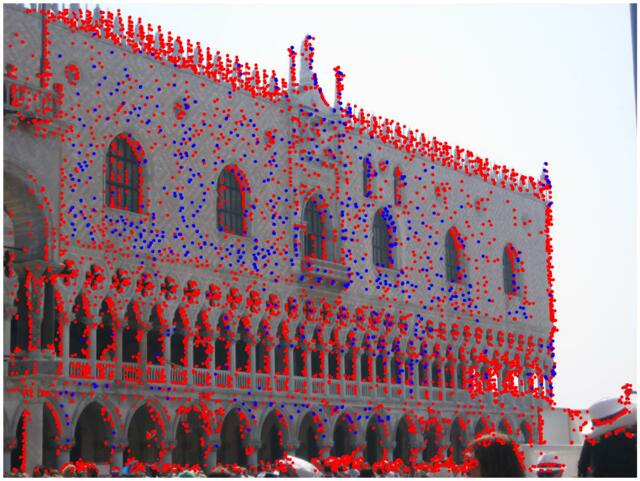} &
\includegraphics[scale=\imscl,width=\imsize]{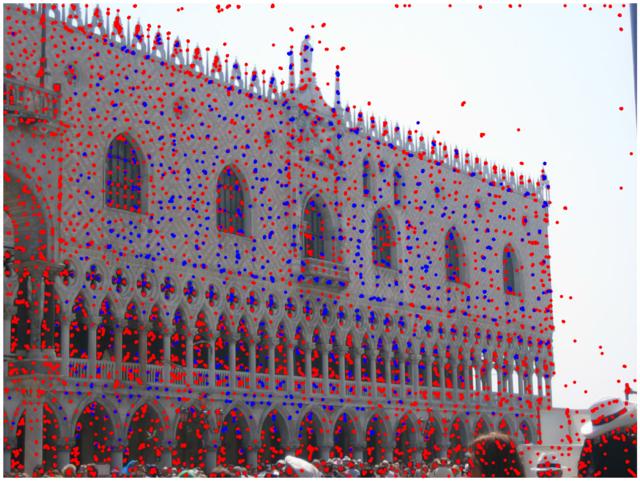} &
\includegraphics[scale=\imscl,width=\imsize]{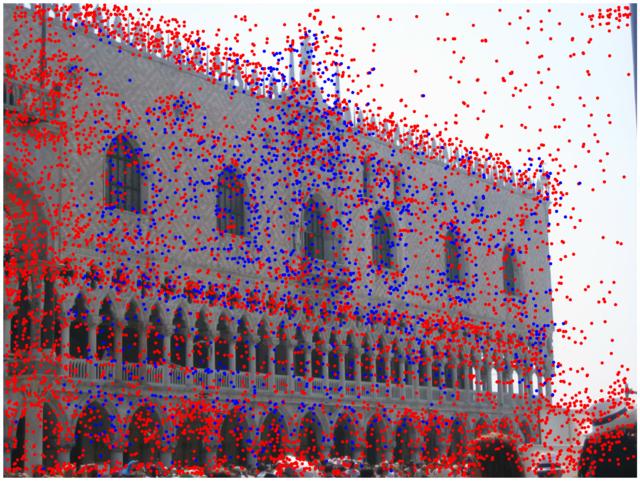} &
\includegraphics[scale=\imscl,width=\imsize]{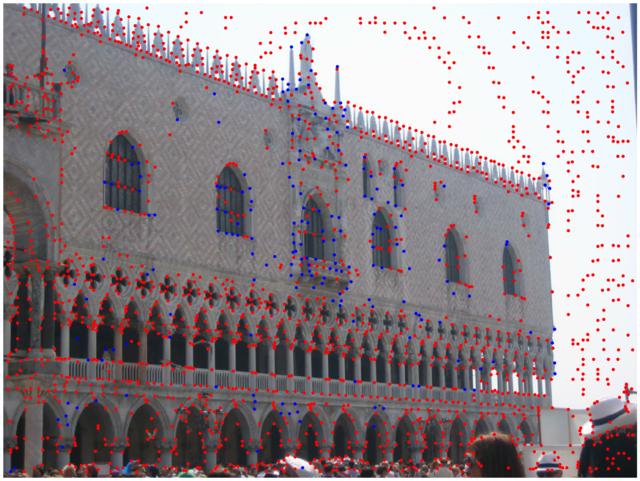} \\

\includegraphics[scale=\imscl,width=\imsize]{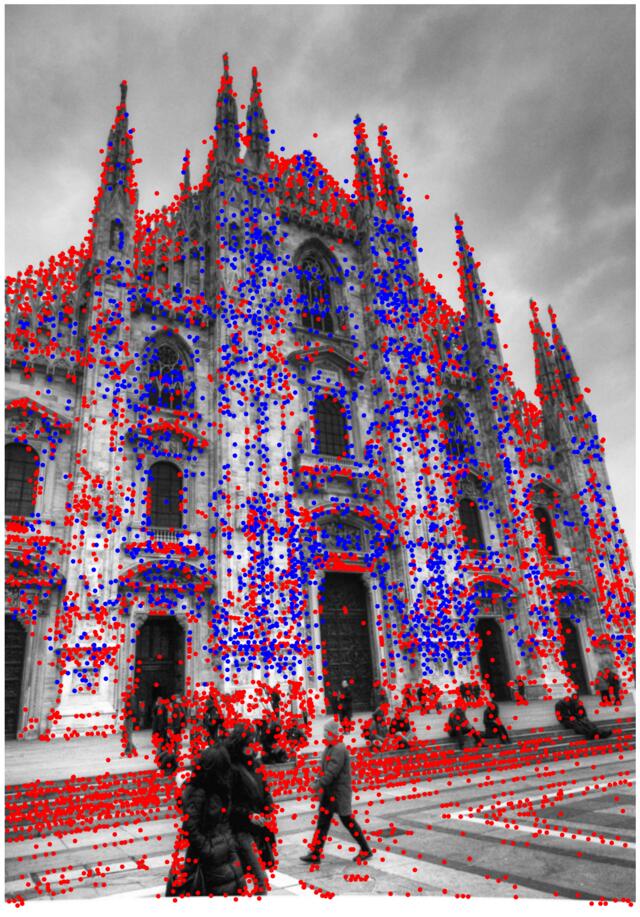} &
\includegraphics[scale=\imscl,width=\imsize]{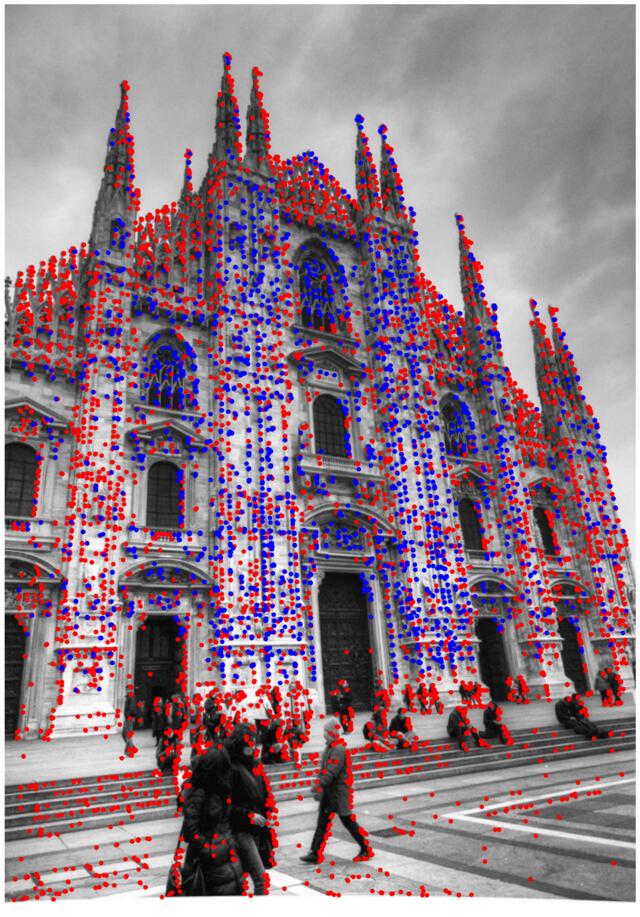} &
\includegraphics[scale=\imscl,width=\imsize]{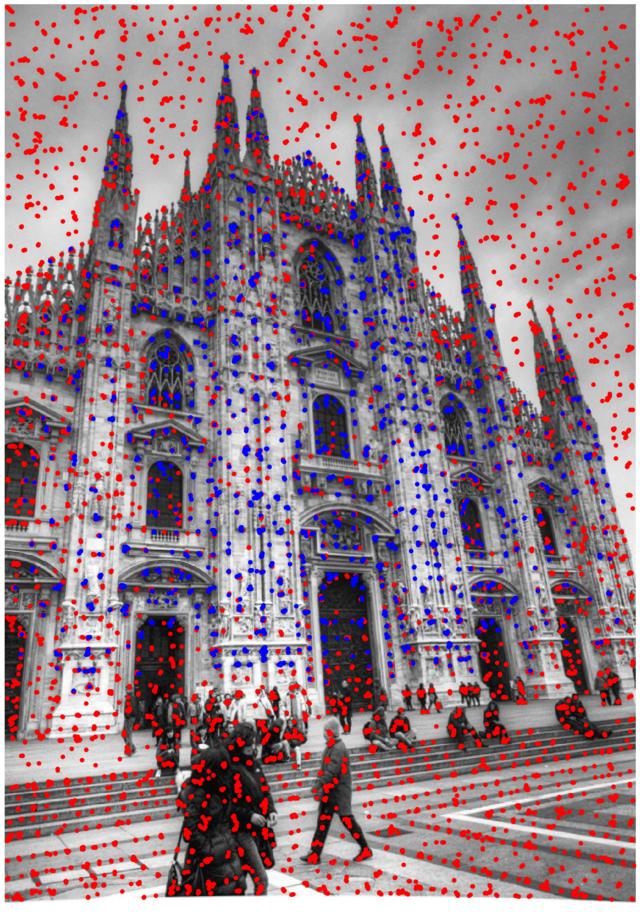} &
\includegraphics[scale=\imscl,width=\imsize]{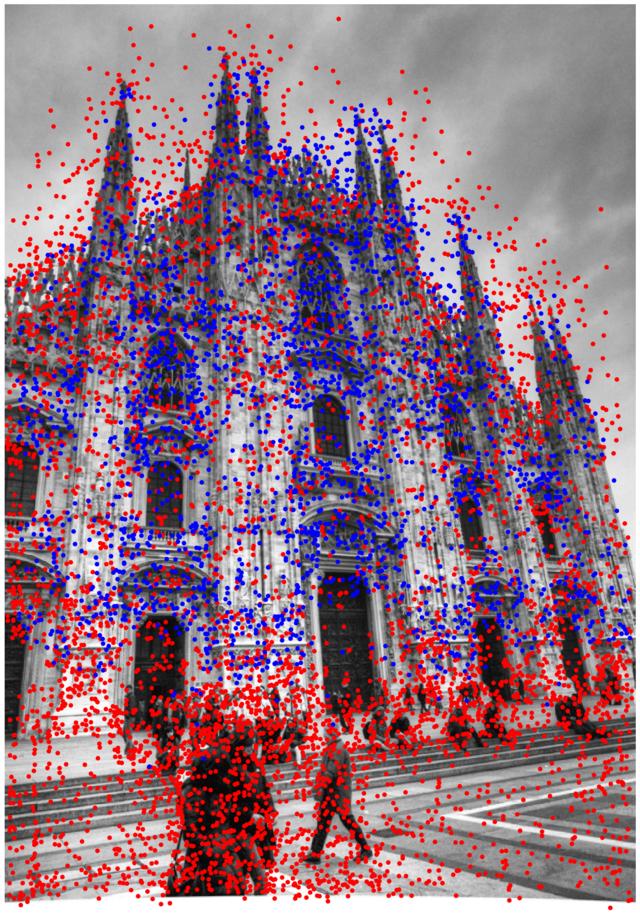} &
\includegraphics[scale=\imscl,width=\imsize]{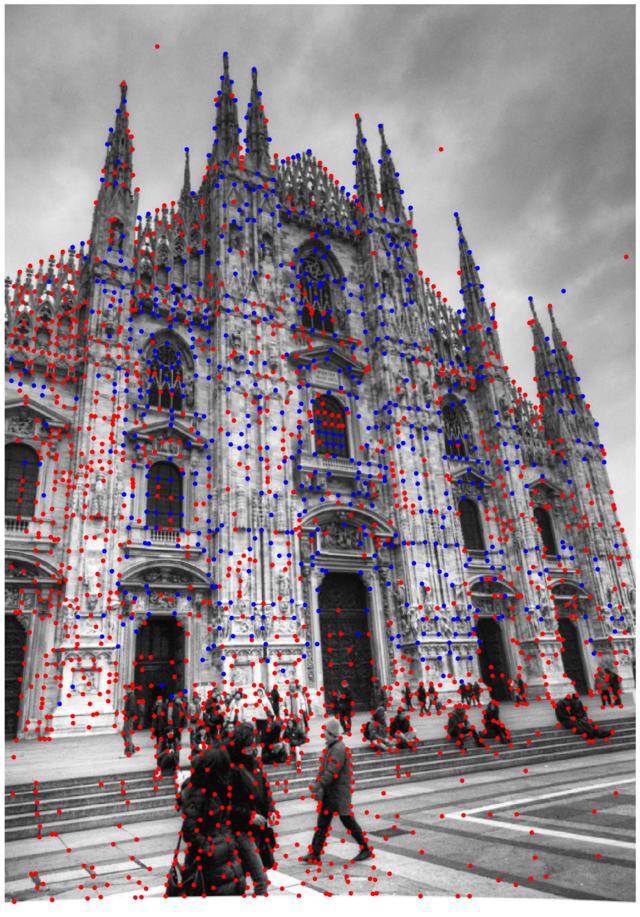} \\

(a) DoG+HardNet & (b) KeyNet+HardNet & (c) R2D2 & (d) D2-Net (MS) & (e) SuperPoint (2k) \\
\end{tabular}
\end{center}
\caption{{\bf Qualitative results for the multi-view task -- Learned features.}
\edit{Color-coded as in \fig{supp_multiview}.}
}
\label{fig:supp_multiview2}
\end{figure*}

\section{Summary}
\label{imc:conclusions}
We introduce a comprehensive benchmark for local features and robust estimation algorithms.
The modular structure of its pipeline allows to easily integrate, configure, and combine methods and heuristics.
We demonstrate this by evaluating dozens of popular algorithms, from seminal works to the cutting edge of machine learning research, and show that classical solutions may still outperform the perceived state of the art with proper settings.

The experiments carried out through the benchmark and reported in this paper have already revealed unexpected, non-intuitive properties of various components of the SfM pipeline, which will benefit SfM development, \eg, the need to tune RANSAC to the particular feature detector \emph{and} descriptor and to select specific settings for a particular RANSAC variant.
Other 
\edit{interesting}
facts have been uncovered by our tests, such as that the optimal set-ups across different tasks (stereo and multiview) may differ, or \edit{that methods that perform better on proxy tasks, like patch \edit{retrieval} or repeatability, may be inferior on the downstream task}.
Our work is open-sourced and makes the basis of an open challenge for image matching with sparse methods.

\chapter{Questions from Past and Ideas for Future Research}
\label{ch:conclusions}

Each Chapter ends with its conclusion. To conclude the thesis as a whole, I would like to present some open questions that I find to be puzzling and important as well as research directions, which I find to be worth exploring in the short and long term. These questions and ideas arise while I was working on the thesis and are directly related to it. 
As you may noticed, I have switched ``we'' to ``I'' to underline the subjectiveness of the open directions I am going to present.

\section{Questions}
\label{concl:lessons-learned}

\subsection{Should one make the training-time setup as close as possible to the test time setup?}
\label{concl:test-vs-training}
While it is believed to be a good idea to remove the gap between training and test conditions with an evidence for doing so~\cite{testtime20219}, it has not always led to the best results. For example, HardNet~\cite{HardNet2017} borrows the idea of using the second nearest neighbor (SNN) descriptor ratio~\cite{SIFT2004} from the SIFT paper. However, we have discovered that using SNN ratio for descriptor learning leads to inferior results, confirming the results first reported in~\cite{TFeat2016}. We had to modify the loss function to become the triplet margin loss, which optimizes the distance difference instead of the ratio. Another example is the patch sampling procedure for HardNetAmos training~\cite{Pultar2020improving} -- there is little to no difference between random points as ``keypoints'' and actual detections of the Hessian detector when used for local descriptor training. Yet another example is the non-maxima supression method applied to the DISK~\cite{tyszkiewicz2020disk} local features. The paper reports that it is beneficial to replace train-time grid-based approach with a more classical window non-maxima supression. 
The question is -- could we tell when to reduce train-test gap and when not to, and to what extent? 

\subsection{End-to-end or stop gradients?}
\label{concl:test-vs-training}
Some papers -- LIFT~\cite{LFNet2018}, SuperPoint~\cite{SuperPoint2017}, SuperGlue~\cite{sarlin2019superglue} -- report the benefits of end-to-end gradient propagation for model performance. 
Others -- AffNet~\cite{AffNet2018}, DELF~\cite{DELF2017}, DELG~\cite{DELG2020} use the stop-gradient operation and report problems otherwise. DELF and DELG stop the gradients of the descriptor-related loss function and not propagate them to the detector part. Otherwise, the overall representation quality is worse and overfitting is observed. AffNet does not propagate the gradient to the negative examples in its modified triplet margin loss.  Otherwise, AffNet training does not converge when trained with HardNet, see Chapter~\ref{ch:local-affine-features} for more details.

On the other hand, SuperPoint~\cite{SuperPoint2017} and R2D2~\cite{R2D22019} optimize both detector and descriptor losses jointly without stopping the gradients.  Why do some methods benefit from the joint optimization and other methods benefit from the stop gradient? How does stop-gradient operation help and why? A recent paper by Tian~\etal~\cite{tian2021understanding} starts exploring the influence of the stop-gradient operation in the context of a specific case of contrastive learning. However, there are more questions than answers in that direction so far. 

\subsection{Evaluation: typical use-case sampling or focus on corner cases?}
\label{concl:test-vs-training}
 In the past, there was a tendency to create small datasets, which were explicitly curated w.r.t. of the nuisance factors they contain. Oxford-Affine dataset~\cite{MikoDescEval2003, Mikolajczyk05} is a good example of it -- the problems were split into ``categories'' like ``blur'', ``rotation'', ``viewpoint change''. %
 If the model or algorithm can deal with the hardest cases, one could assume that it will work in the less challenging conditions as well.  This might not be the case, because the combination of mild illumination and viewpoint change could be harder to handle than the case, which contains extreme changes, but only in a single nuisanse factor. In other words, one might be able to match a day versus night photo taken from the static camera, but not able to match day versus evening, when the camera moved slightly. 
 
 Now the tendency is to gather as much data as possible from the ``real world'' and hope that it would cover the nuisance factors, which are common in the real world~\cite{DELF2017, IMW2020}. Sometimes it is true. For example, early research in image matching tends to design methods to be rotation invariant~\cite{SIFT2004, ORB2011}. Later, the introduction of the larger scale datasets, scraped from the internet~\cite{Philbin07, DELF2017, IMW2020} led to the the spread of ``up-is-up'' assumption~\cite{Perdoch-CVPR2009efficient}, baked into the local features pipeline~\cite{SuperPoint2017, D2Net2019}. Rotational invariance in such scenarios hurts more than helps -- the most general method would always work worse than those, which are tailored to the specific situation. 
 
 The question is -- how best to combine both approaches -- curated datasets, which contain corner cases and the ``typical use-case'' datasets?

\section{Future research directions}
\label{concl:future-research-directions}

\subsection{Application-specific local
features}
\label{concl:application-specific-local-features}

Current local features are designed to be general. SIFT works reasonably well for various environments, as well as applications: 3D reconstruction, SLAM, image retrieval, etc. Learned features, like SuperPoint or R2D2 are biased towards the data they are trained on, but still have nothing domain specific in their design.

Despite the lack of domain-specific design, various local features have different properties: they are more or less robust to nuisance factors like illumination and camera position. Some of them, e.g., DELF, are poorly localized, while others -- SIFT, SuperPoint -- are localized well. They can be more or less dense and/or evenly distributed over the image. For example, in image retrieval, one does not really care about precise localization, the robustness is much more important. In 3D reconstruction, it is beneficial to have a lot of features for better reconstruction. On the other hand, for the SLAM/relocalization application, sparse features would be more advantageous because of the smaller memory footprint and computational cost. There are actually some steps in that direction. Let me name a few.

\begin{enumerate}
\item
  HOW local features~\cite{HOW2020} are designed for image retrieval. Localization precision is not considered at all and repeated patterns, unlike for WxBS are encouraged, not discouraged, because they increase robustness of the represenation. For the image retrieval it is not a problem, if one mismatch one window for almost identical neighboring window, as long as it helps finding the image.
\item SIPs~\cite{SIPS2019} -- as sparse as possible local features for the SLAM.

\item Reinforced Feature Points: Optimizing Feature Detection and Description for a High-Level Task~\cite{ReinforcedPoints2020} -- finetuning the exisiting local feature with reinforcement learning for the downstream task. 
\end{enumerate}

I believe that it is only the beginning and we are yet to experience AlphaZero moment for the local features.

\subsection{Rethinking the overall wide baseline stereo pipeline,
optimized for specific applications}
\label{concl:rethinking-pipeline}

It is often perceived that image matching of an unordered collection of images is a task with quadratic complexity w.r.t. number of images. Some operations can be done separately, e.g.~feature detection, but others, like feature matching and RANSAC cannot. However, it turns out that the quadratic part can be done on the level of nearest neighbor search among global descriptors alone. This part scales very well~\cite{FAISS2017}. The rest of the steps -- feature matching and RANSAC can be avoided for more than 90\% of image pairs with clever preprocessing, ordering, and re-using results from the previous matching. Moreover, to do the whole task faster (matching image collection), one may need to introduce additional steps, which are slowing things down for the two images case, e.g., calculating a global descriptor of the image. That is what we have done for the initial pose estimation for the global SfM~\cite{barath2020efficient}, which reduced the matching runtime on 1dSfM dataset~\cite{wilson_eccv2014_1dsfm} from 200 hours to 29. Another example would be SuperGlue~\cite{sarlin2019superglue}, where the authors abandoned traditional descriptor matching and instead leveraged all information about keypoints and descriptors from both images altogether processed by the graph neural network.

\subsection{Rethinking and improving the process of training data generation for WxBS}
\label{concl:rethinking-training}

So far, all local feature papers I have read rely on one of these sources of supervision:

\begin{enumerate}
\item
  SfM data, obtained using COLMAP with, possibly, a cleaned depth information.
\item
  Affine and color augmentation.
\item
  Synthetic images (e.g.~corners for SuperPoint).
\end{enumerate}

There are several problems with these sources of supervision.

\textbf{SfM data} assume that the data is matchable by existing
methods, at least to an extent. That might not always be true for
cross-seasonal, medical and other kinds of data. It is also not
applicable for historical photography and other types of data.
Moreover, SfM data takes a significant time to compute and the significant storage space. %

\textbf{Affine and color augmentation} can take us only this far -- we
actually want our detectors and descriptors to be robust to the changes,
which we do not know how to simulate/augment.

\textbf{Synthetic images} as in, for example, the CARLA~\cite{CARLA2017} simulator, lack fine details and photorealism. However, I am optimistic about using neural renderers and learned wide multiple baseline stereo in a GAN-like self-improving loop, where WxBS is used for improving the rendering part, which is, in its turn, provides the WxBS with new, more challenging examples to train on.

What do I propose instead? The interesting directions to explore are the following. First -- photorealistic renderings of the scenes, like the ones being used for object detection~\cite{PhotorealisticHodan2019} and semantic segmentation~\cite{Hypersim2020}. Currently, they are not tuned to provide the information required for the wide baseline stereo. Second -- probably more expensive -- approach would be to use laser scans registered together with RGB images to get the perfect 3D reconstruction and use it for local features training. One of such examples is the Tanks and Temples dataset~\cite{TanksAndTemples2017} currently mostly used for benchmarking and training geometry estimation models.

\subsection{Matching with On-Demand View Synthesis revisited}
\label{concl:mods-again}

I like the ``on-demand'' principle a lot and I think we can explore it much more that we are now. So far, we have either affine view synthesis (ASIFT~\cite{ASIFT2009}, MODS~\cite{MODS2015}), or GAN-based stylizations~\cite{anoosheh2019nighttoday} for the day-night matching. That is why I am glad to see papers like ``Single-Image Depth Prediction Makes Feature Matching Easier''~\cite{toft-2020-rectified-features}, which generate normalized views based on depth to help the matching. Why not go further? Combine viewpoint, illumination, season, and sensor synthesis? 

\subsection{More inputs!}
\label{concl:moar-inputs}

I have mentioned above that monocular depth may help feature matching or camera pose estimation. However, why stop here? 
Let us use other networks as well, especially given that we will need them on the robot or vehicle anyway. Semantic segmentation? Yes, please. Surface normals? Why not? Intrinsic images? Let it be!

\bibliographystyle{plain}
\bibliography{phd_thesis_bibliography}

\end{document}